\documentclass[a4paper]{scrartcl}
\usepackage[utf8]{inputenc}
\usepackage[T1]{fontenc}
\usepackage[english]{babel} 
\usepackage{lmodern}
\usepackage[]{algorithm2e}
\usepackage[onehalfspacing]{setspace}
\usepackage{amsmath,amssymb,amsthm,amsfonts,amsbsy,latexsym}
\usepackage{graphicx}
\usepackage{geometry}
\usepackage{natbib}
\usepackage{multirow}
\usepackage{float}
\usepackage{fancyhdr}
\usepackage{makecell}
\usepackage{pgf,tikz}
\usepackage{makeidx}
\usetikzlibrary{shapes,snakes}
\usetikzlibrary{fadings}
\usepackage[pdfpagelabels=true]{hyperref}

\newcommand{\R}{\mathbb{R}}

\newcommand{\D}{\mathcal{D}}

\DeclareMathOperator*{\sign}{sign}
\DeclareMathOperator*{\argmin}{argmin}

\DeclareMathOperator*{\SNR}{SNR}

\numberwithin{equation}{section}
\IfFileExists{upquote.sty}{\usepackage{upquote}}{}
\begin{document}

\newtheorem{thm}{Theorem}[section]
\newtheorem{Def}{Definition}[section]
\newtheorem{lem}{Lemma}[section]
\newtheorem{rem}{Remark}[section]
\newtheorem{cor}{Corollary}[section]
\newtheorem{prop}{Proposition}[section]
\newtheorem{ex}{Example}[section]
\newtheorem{ass}{Assumption}[section]
\newtheorem*{bew}{Proof}
\title{Global quantitative robustness of regression feed-forward neural networks}
\author{Tino Werner\footnote{Institute for Mathematics, Carl von Ossietzky University Oldenburg, P/O Box 5634, 26046 Oldenburg (Oldb), Germany, \texttt{tino.werner1@uni-oldenburg.de}}}
\maketitle

\begin{footnotesize} 

\begin{abstract}

Neural networks are an indispensable model class for many complex learning tasks. Despite the popularity and importance of neural networks and many different established techniques from literature for stabilization and robustification of the training, the classical concepts from robust statistics have rarely been considered so far in the context of neural networks. Therefore, we adapt the notion of the regression breakdown point to regression neural networks and compute the breakdown point for different feed-forward network configurations and contamination settings. In an extensive simulation study, we compare the performance, measured by the out-of-sample loss, by a proxy of the breakdown rate and by the training steps, of non-robust and robust regression feed-forward neural networks in a plethora of different configurations. The results indeed motivate to use robust loss functions for neural network training.

\end{abstract}


\end{footnotesize}

\tableofcontents

\newpage

\section{Introduction}

Regarding the ambitious goal of achieving trustworthy AI (e.g., \cite{wing}), one crucial component is the robustness of the AI model itself, sometimes referred to under the notion of security of AI. Robustness is often understood in the sense of adversarial robustness, i.e., robustness of the AI against adversarial attacks (e.g., \cite{goodfellow}). Adversarial attacks are perturbations of the data points which are as small as possible, quantified usually in some vector norm, so that the output of the AI model significantly changes, e.g., in the standard classification setting for adversarial attacks, the output class switches from the original class to another class or even to a target class pre-scribed by the attacker (e.g., \cite{carlini}). Adversarial attacks are crafted after the AI model is trained, therefore, they do not have any effect on AI training unless being used for adversarial training. 

The classical robust statistics (e.g., \cite{huber}, \cite{hampel}, \cite{maronna}, \cite{rieder}) consider the robustness of an estimator against contamination in the data set, so the contamination appears before training, often by natural reasons like model misspecification. The robustness of an estimator is quantified either in terms of local robustness which amounts to the influence function (\cite{hampel74}, \cite{huber83}), measuring the infinitesimal influence of one single data point onto the estimator, or in terms of global robustness which corresponds to the breakdown point (BDP). As for the BDP, there exists a functional version (\cite{hampel71}), but when working with data, the empirical version, also called finite-sample BDP (\cite{huber83}), is often more appropriate, which quantifies the number of instances in the data set that have to be modified arbitrarily in order to let the estimator break down, i.e., take unreasonable values. 

Robust statistics is related to the concept of poisoning attacks (\cite{biggio12}), which is a counterpart of adversarial attacks, where also perturbations of data points are computed in order to distort a model, but in contrast to adversarial attacks, the perturbation takes place before AI training, hence poisoning attacks indeed have an impact on the trained model. Note that one can either consider poisoning attacks as additional samples (\cite{biggio12}) or as perturbations of existing samples that replace these original samples (\cite{lai20f}). In contrast to the perturbations usually considered in robust statistics which either affect a subset of the instances or instance-specific subsets of the cells (\cite{alqallaf}), the perturbations for poisoning attacks are computed onto the whole training data subject to an overall bound, often quantified in some matrix norm. Therefore, the perturbations considered in robust statistics can be interpreted as poisoning attack perturbations that are bounded by the $l_0$-norm, either w.r.t. the row indices or w.r.t. the cell indices. 

In order to safeguard against perturbations, robust statistics considers means to bound the influence of data points, for example, by downweighting or even trimming outlying instances or cells. It is important to note that contamination in the response vector is fundamentally different from contamination in the predictor matrix due to the additional leverage effect. In fact, while the first situation may be handled using loss functions with bounded gradient, contamination in the predictor matrix amounts to using loss functions whose gradient even redescends to zero for input values with large norm, bounding the loss function itself (e.g., \cite{hampel}). Usually, due to low efficiency, the resulting M-estimator often enters MM-estimation as starting estimator (e.g., \cite{maronna}). Another opportunity is to robustify the aggregation procedure of the losses in the sense that a trimmed mean or the median of the losses is minimized, which, for the squared loss function, leads to the least trimmed squares (LTS) or the least median of squares (LMS) estimator, see \cite{rous84}. Techniques based on robust loss functions or aggregations have indeed also been proposed for deep learning, e.g., \cite{rusiecki}, \cite{kalina19b} or \cite{melegy}. 

In this paper, we aim at systematically investigating the effects of different types of contamination and to which extent a suitably robust loss function can safeguard against them. Moreover, we provide a more formal description of the meaning of robustness in the setting of deep learning and propose an adapted version of the regression BDP in order to quantify the global robustness of feed-forward neural networks. 

The paper is organized as follows. Sec. \ref{trimnn:prelim} compiles the basic components of feed-forward neural networks, robust statistics and robustifications of neural networks by the standard concepts of robust statistics from literature. In Sec. \ref{trimnn:bdpsec}, we propose an adapted breakdown point notion tailored to neural networks and analyze it for different contamination configurations. Sec. \ref{trimnn:simsec} provides the details of our simulation study. The summarized results are provided in Sec. \ref{trimnn:ressec} while the graphics that depict the results are moved to the Appendix due to their sheer number.

\section{Preliminaries} \label{trimnn:prelim}

\subsection{Feed-forward neural networks}

Neural networks are complex structures that do not directly map an input vector onto an output like a standard linear model but that include numerous potentially non-linear intermediate transformations. We refer to \cite{goodfellow16} for an excellent overview. 

In this work, we restrict ourselves to the analysis of feed-forward neural networks (FFNNs) which can be considered to be the most simple ones. Let a data matrix $\D=(X,Y) \in \bigotimes_{i=1}^n \mathcal{X} \times \mathcal{Y} \subset \R^{n \times (p+1)}$ be given for predictors $X_i \in \mathcal{X} \subset \R^p$ and responses $Y_i \in \mathcal{Y} \subset \R$. An FFNN consists of $(H+1)$ layers of which $H$ ones are interpreted as hidden layers since they contain the intermediate transformations while the $(H+1)$-th one is the output layer that maps the final transformation onto the output space. Each of the layers consists of so-called hidden nodes which are $L_h$ ones for layer $h$, $h=1,...,H+1$. In the hidden nodes, the actual transformations are computed. 

Starting from an input vector $x \in \mathcal{X}$ which can be interpreted to define the input layer (layer 0), a fully-connected FFNN computes \begin{equation} \label{trimnn:trafo} z_l^{(1)}=\sigma^{(1)}(a_l^{(1)}):=\sigma^{(1)}\left(\sum_{j=1}^p w_{lj}^{(1)}x_j+b_l^{(1)}\right), \ \ \ l=1,...,L_1, \end{equation} in the first hidden layer, for a so-called activation function $\sigma^{(1)}: \R \rightarrow \R$ and weights $b_l^{(1)}$, $w_{lj}^{(1)} \in \R$, $j=1,...,p$. The $z_l^{(1)}$, $l=1,...,L_1$, serve as the input for the second hidden layer and so forth. In order to streamline the notation, we write $z_j^{(0)}:=x_j$, so that for layer $h=1,...,H+1$, the computation in the $l$-th hidden node, $l=1,...,L_h$, can be written as \begin{center} $ \displaystyle z_l^{(h)}=\sigma^{(h)}(a_l^{(h)}):=\sigma^{(h)}\left(\sum_{j=1}^{L_{h-1}} w_{lj}^{(h)}z_j^{(h-1)}+b_l^{(h)}\right). $ \end{center} Classical intermediate activation functions include sigmoid ones like $tanh$ or the piece-wise linear ReLU activation $z \mapsto \max(z,0)$, although many variants already have been proposed (\cite{goodfellow16}). The output activation function $\sigma^{(H+1)}$ is essentially defined by the learning task, for example, a regression NN with real-valued outputs would require no output activation (artificially, one can speak of an ``identity activation''), besides having only one output node ($L_{H+1}=1$), while classification problems usually invoke a mapping of the activations in layer $H$ onto the, say $K$, different classes, leading to $K$ output nodes, which amounts to using a softmax activation, or a sigmoid activation for $K=2$. 

During training, the intercepts $((b_l^{(h)})_{l=1}^{L_h})_{h=1}^{H+1}$ and weights $(((w_{lj}^{(h)})_{j=1}^{L_h-1})_{l=1}^{L_h})_{h=1}^{H+1}$ (we will usually refer to both intercepts and weights just as ``weights'') are updated via gradient steps. Having a loss function $L: \mathcal{Y} \times \mathcal{Y} \rightarrow [0,\infty[$, one updates the weights in the output layer via standard gradient steps. Provided that the activation functions are sufficiently smooth, this gradient step can be used to update the weights of the previous layer in a backward updating scheme that relies on the chain rule, therefore the procedure is called backward propagation (BP). Based on an initialization of all weights, the first training epoch consists of a forward pass where the inputs $X_i$ are fed into the NN, resulting in predictions $\hat Y_i$. Using these $\hat Y_i$, a backward pass based on the losses $L(\hat Y_i,Y_i)$ is performed, updating the weights via BP. Starting with the output layer, denoting \begin{center} $ \displaystyle \partial_{a_l^{(H+1)}}L=\partial_{z_l^{(H)}}L(\sigma^{(H+1)})'(a_l^{(H+1)})=:\delta_l^{(H+1)},$ \end{center} one computes the partial derivaties \begin{center} $ \displaystyle \partial_{b_l^{(H+1)}}L=\partial_{a_l^{(H+1)}}L\partial_{b_l^{(H+1)}}a_l^{(H+1)}=\delta_l^{(H+1)}$, \end{center} \begin{center} $ \displaystyle \partial_{w_{lj}^{(H+1)}}L=\partial_{a_l^{(H+1)}}L\partial_{w_{lj}^{(H+1)}}a_l^{(H+1)}=\delta_l^{(H+1)}z_j^{(H)}, $ \end{center} and using $\partial_{b_l^{(h)}}a_l^{(h)}=1 \ \forall j,l,h$ and $\partial_{w_{lj}^{(h)}}a_l^{(h)}=z_j^{(h-1)} \ \forall j,h$ leads to the recursion \begin{center} $ \displaystyle \delta_l^{(h)}=w_{lj}^{(h)} \delta_l^{(h+1)}(\sigma^{(h)})'(a_l^{(h)}), $ \end{center} \begin{equation} \label{trimnn:bp34} \partial_{b_l^{(h)}}L=\delta_l^{(h)}, \ \ \ \partial_{w_{lj}^{(h)}}L=z_j^{(h-1)}\delta_l^{(h)}. \end{equation} The gradients in Eq. \ref{trimnn:bp34} are then used for a gradient step of the form \begin{center} $ \displaystyle w_{lj}^{(h)}=w_{lj}-\frac{\eta}{m}\sum_{i=1}^m \partial_{w_{lj,i}^{(h)}}L, \ \ \ b_l^{(h)}=b_l^{(h)}-\frac{\eta}{m}\sum_{i=1}^m \partial_{b_{l,i}^{(h)}}L, $ \end{center} where the index $i$ is the respective derivative on instance $i$ from a sample of size $m \le n$ of instances from the training data and where $\eta$ is a learning rate.

This is done until numerical convergence or until a maximum number of epochs is reached.

\subsection{Robust statistics}

The following definition formalizes the understanding of contamination in robust statistics (cf. \cite[Sec. 4.2]{rieder}). 

\begin{Def} \label{trimnn:contball} Let $(\Omega, \mathcal{A})$ be a measurable space and denote $\mathcal{P}:=\{P_{\theta} \ | \ \theta \in \Theta\}$ as a parametric family so that each element $P_{\theta} \in \mathcal{P}_{\theta}$ is a distribution on $(\Omega, \mathcal{A})$. Let $P_{\theta_0}$ denote the ideal distribution. Let $\Theta \subset \R^p$ be a parameter space. A \textbf{contamination model} is a family $\mathcal{U}_*(\theta_0):=\{U_*(\theta_0,r) \ |\ r \in [0,\infty[\}$ of \textbf{contamination balls} $U_*(\theta_0,r)=\{Q \in \mathcal{M}_1(\mathcal{A}) \ | \ d_*(P_{\theta_0},Q) \le r \}$ for the set $\mathcal{M}_1(\mathcal{A})$ of probability distributions on $\mathcal{A}$. The radius $r$ is sometimes referred to as ``contamination radius''. \end{Def}

\begin{ex} \label{trimnn:convcont} The convex contamination model $\mathcal{U}_c(\theta_0)$ is given by the set of convex contamination balls \begin{center} $ \displaystyle U_c(\theta_0,r)=\{(1-r)_+P_{\theta_0}+\min(1,r)Q \ | \ Q \in \mathcal{M}_1(\mathcal{A}) \}$. \end{center}  \end{ex} 

The finite-sample regression BDP has been introduced in \cite{huber83} and considers linear models, i.e., $Y_i=X_i\beta+\epsilon_i$.

\begin{Def} \label{trimnn:bdp} Let $Z_n:=\{(X_1,Y_1),...,(X_n,Y_n)\}$ for instances $(X_i,Y_i) \in \R^{p+1}$ in the given data set. The \textbf{case-wise finite-sample breakdown point} for a regression estimator $\hat \beta \in \R^p$ is defined by \begin{equation} \label{trimnn:fsbdp} \varepsilon^*(\hat \beta,Z_n)=\min\left\{\frac{m}{n} \ \bigg| \ \sup_{Z_n^m}(||\hat \beta(Z_n^m)||)=\infty \right\} \end{equation} where the set $Z_n^m$ denotes any sample with exactly $(n-m)$ instances in common with $Z_n$. The coefficient $\hat \beta(Z_n)$ is the estimated regression coefficient on $Z_n^m$. \end{Def} 

The contamination procedure considered in Def. \ref{trimnn:bdp} is a case-wise or instance-wise contamination as in the convex contamination setting in Ex. \ref{trimnn:convcont} where the distributions $P$ and $Q$ are defined on $\mathcal{X} \times \mathcal{Y}$. One can restrict the contamination to $Y$-contamination, indicating that one is only allowed to contaminate the response vector, so $P$ and $Q$ would only be defined on $\mathcal{Y}$ while the predictor matrix $X$ would stay unharmed. Contamination of the predictor matrix is referred to as $X$-contamination. An even more flexible setting has been proposed in \cite{alqallaf} where one does not necessarily have to contaminate whole instances but where one is allowed to contaminate single cells. This contamination model can be described by the following definition, see also \cite{agos15}: 

\begin{Def} \label{trimnn:cell-wise} Let $P_{\theta_0}$ be a distribution on some measurable space $(\Omega,\mathcal{A})$ and let $X_i \sim P_{\theta_0}$ for $i=1,...,n$. For the $n \times p$-matrix $X$ consisting of the $X_i$ as rows and random variables $U_1,...,U_p \sim Bin(1,r)$ i.i.d., the \textbf{cell-wise convex contamination model} consists of all sets of the form \begin{center} $ \displaystyle U^{cell}(\theta_0,r):=\{Q \ |\ Q=\mathcal{L}(UX+(1-U) \tilde X)\} $ \end{center} for $\tilde X \sim \tilde Q$ where $\tilde Q$ is any distribution on $(\Omega,\mathcal{A})$ and where $U$ is the matrix with diagonal entries $U_i$. \end{Def} 

\begin{rem} \label{trimnn:bdprem} The notion of the BDP can be lifted to the cell-wise contamination setting by just considering the relative frequency of contaminated cells necessary in order to let the estimator break down, as for example proposed in \cite{velasco}. \end{rem}

\subsection{Robust neural networks}

There are tons of literature considering the robustness of NNs. The term ``robustness'' is often understood as robustness against adversarial attacks (\cite{goodfellow}), which aim at finding perturbations of the training points that lead to significantly different outputs (while training the model on the non-perturbed training points), or poisoning or backdoor attacks (\cite{biggio12}), which are injections of malicious inputs into the training data, either by replacing existing ones (e.g., \cite{lai20f}) or by adding new points. While adversarial robustness relates to the smoothness of the NN itself for regression and also considers a proper placement of the decision boundaries for classification, robustness against poisoning attacks is closely related to the classical understanding of robustness in robust statistics. More precisely, poisoning attacks are crafted subject to a $||\cdot||_q$-bound on the perturbation matrix itself, $1 \le q \le \infty$, while the classical robustness considers mostly the sparse variant, i.e., the $||\cdot||_0$-bound, meaning that only a few instances or cells can be perturbed, but with arbitrary perturbation magnitude which is bounded in poisoning attacks. 

The application of robust loss functions or trimming for NNs has been proposed in several works, e.g., \cite{melegy}, \cite{rusiecki}, \cite{chuang} and \cite{kalina19b}. Although these works are accompanied by simulation studies on simulated and real data, to the best of our knowledge, a systematic robustness analysis by distinguishing between $Y$-outliers, case-wise and cell-wise $X$-outliers has not yet been done so far, and $Y$-outliers and $X$-outliers have been handled separately very rarely, e.g., in \cite{kordos}. Moreover, although the notion of the BDP has been used in many works, a concise definition in the context of neural networks has not been given except for the work of \cite{chen21b} which is however tailored to the aggregation procedure in convolutional filters.

Further robustifications have been proposed for the training process itself, for example, by gradient clipping which essentially corresponds to a Huberization of the losses (\cite{menon19}), resilient backpropagation where the weights are not updated according to the gradients themselves but by a fixed amount where the gradient only enters via its sign (\cite{riedmiller}), i.e., \begin{center} $ \displaystyle w_{lj}^{(h)}=w_{lj}-\eta\sign\left(\frac{1}{m}\sum_{i=1}^m \partial_{w_{lj,i}^{(h)}}L\right), \ \ \ b_l^{(h)}=b_l^{(h)}-\eta\sign\left(\frac{1}{m}\sum_{i=1}^m \partial_{b_{l,i}^{(h)}}L\right), $ \end{center} robust aggregation in GNNs (\cite{geisler}) or bounded activation functions, just to name a few. Proper regularization like Dropout (\cite{sriva}) or knowledge constraints as in informed machine learning (e.g., \cite{rueden19b}) can also be interpreted as implicit robustification, although generally not safeguarding against large outliers.

\section{Breakdown of regression feed-forward neural networks} \label{trimnn:bdpsec}

\subsection{Robust loss functions and robust aggregation} 

In order to safeguard, at least partially, against contamination, several ``robust loss functions'' have been proposed (see, e.g., \cite{hampel}). Well-known ones include Huber's loss function, introduced in \cite{huber64}, and Tukey's biweight function, \begin{center} $ \displaystyle L^{Huber}_{\delta}(r)=\begin{cases} r^2/2, \ \ \ |r| \le \delta \\ \delta |r|-\delta^2/2, \ \ \ |r|>\delta \end{cases}, \ \ \ L^{Tukey}_k(r)=\begin{cases} 1-[1-(r/k)^2]^3, \ \ \ |r| \le k \\ 1, \ \ \ |r|>k \end{cases}. $ \end{center} While Huber's loss function only has bounded gradients, Tukey's loss function is a redescender (e.g., \cite{huber}) since the gradients attain zero for large inputs (in absolute value). For regression, $r$ denotes the residuals, i.e., for an instance $(X_i,Y_i)$ of the data set, the residual is given by $r_i=Y_i-\hat Y_i$ for the prediction $\hat Y_i$ of the regression model.

\begin{figure}
\label{trimnn:hubtuk}
\begin{center}
\includegraphics[width=5cm]{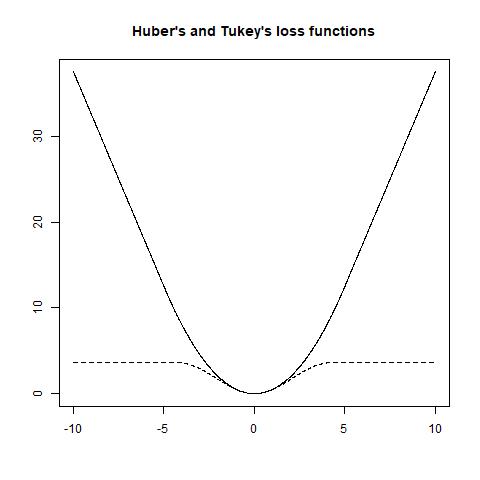} \includegraphics[width=5cm]{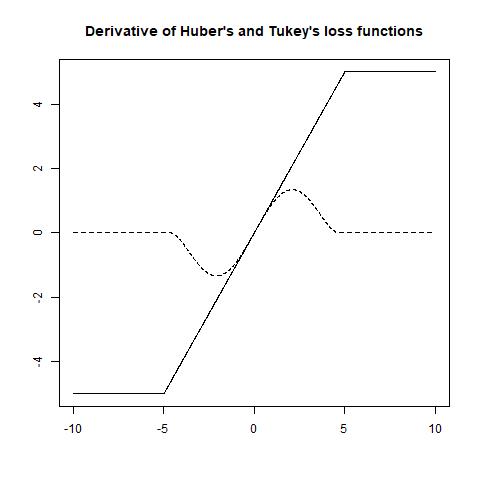}
\end{center}
\caption{Huber's and Tukey's loss functions and derivatives}
\end{figure}

Another common technique to safeguard against contamination is to robustify the aggregation procedure when computing average losses or gradients during training. \cite{rous84} proposed to minimize the median of the squares residuals, leading to the least median of squares (LMS) estimator, or the trimmed mean of the squared residuals, resulting in the least trimmed squares (LTS) estimator. Trimming is understood in the sense that one only considers the $\lceil n/2 \rceil \le h<n$ instances with the smallest losses during training. Although finding the optimal $h$-subset is a combinatorial problem and therefore infeasible, a clever strategy suggested in \cite{rous00}, \cite{rous06} based on so-called concentration steps where one iteratively finds the $h$ instances with the smallest losses w.r.t. the current model fit and updates the model by re-computing it on the current $h$ best instances allows for a fast computation of the LTS. 

It is important to point out that trimming does not correspond to a robust loss function because for each instance, either the standard loss function is used or the loss function $L \equiv 0$ is used, in contrast to a skipped loss function (e.g., \cite[Sec. 2.5, Ex. 1]{hampel}) which, based on the squared loss, would be $L_k(r)=|r|^2I(|r| \le k)$, i.e., this loss function would indeed non-continuously fall back to zero for large absolute values of the input. We highlighted this fact because \cite{kordos} mentioned non-differentiability of certain loss functions, including a trimmed absolute loss. The trimmed absolute loss is clearly non-differentiable due to the absolute loss already being non-differentiable (in zero), but this does not jeopardize BP in NN training as it just corresponds to a zero set. This fact has been proven in \cite{lee20d} who show that automatic differentiation as done in BP is correct even for non-differentiable functions as long as they satisfy certain regularity conditions which is true for piece-wise smooth functions, including Huber's loss, a skipped squared loss or an absolute loss.

\subsection{Breakdown point} 

Although we do not propose a new BDP concept, we intend to clarify the meaning of a regression BDP for neural networks:

\begin{Def} \label{trimnn:bdpnn} Let $Z_n$ and $Z_n^m$ be defined as in Def. \ref{trimnn:bdp} and let $B \in \R^{L_1 \times L_2 \times ... \times L_{H+1}}$ denote the intercept array and let $W \in \R^{(L_1 \times L_0) \times ... \times (L_{H+1} \times L_H)}$ denote the weight array that correspond to the FFNN. Then, the case-wise finite-sample BDP for the FFNN is given by \begin{equation} \label{trimnn:ffnnbdp} \varepsilon^*(\hat B, \hat W,Z_n)=\min\left\{\frac{m}{n} \ \bigg| \ \sup_{Z_n^m}(\sup_{k \in \mathbb{N}}(||(vec(\hat B^{(k)}(Z_n^m)),vec(\hat W^{(k)}(Z_n^m)))||))=\infty \right\} \end{equation} where $vec: \R^{n_1 \times ... \times n_K} \rightarrow \R^{n_1+...+n_K}$ denotes the operator that flattens an array to a vector and where $\hat B^{(k)}(Z_n^m)$, $\hat W^{(k)}(Z_n^m)$ denote the estimated intercepts and weights on $Z_n^m$ in training epoch $k$. \end{Def}

The situation of cell-wise contamination can be captured as described in Rem. \ref{trimnn:bdprem}, i.e. the relative number of contaminated cells is considered.

We have to elaborate why the epoch enters the BDP definition here. BDP computations in machine learning usually consider the underlying empirical risk in order to show the effect of outliers, for example, one can show that the loss of the zero coefficient is smaller than the loss of a broken coefficient in order to prove that no breakdown occurs (e.g., \cite[Thm. 1]{alfons13}, \cite[Thm. 2]{zhao18}). As NN training considers techniques like gradient clipping or resilient backpropagation which bounds gradients even if the underlying loss function in fact leads to unbounded gradients, in combination with early stopping, such NNs would never break down by design, although not converging and being in fact highly non-robust. In our BDP definition \ref{trimnn:bdpnn}, we therefore do not allow for early stopping since it does not seem to be reasonable in this setting because, in consequence, one could stop any iterative algorithm directly after the initialization which would also never lead to a breakdown for proper initializations and artificially create robustness where there is in fact no robustness. 

\begin{rem} We implicitly made the assumption that only the instances, i.e., the predictors $X_i$ or the responses $Y_i$ can be contaminated, so a direct manipulation of the hidden nodes is excluded. Although it may technically be possible to intercept a training procedure by modifying and freezing hidden neurons arbitrarily, it does not seem to be reasonable in this work where we want to examine robustness of FFNNs against training data contamination only. Note that if one had access to the internal training process, one just could modify the activations $z_j^{(H)}$ in the final hidden layer and freeze them which would correspond to the standard linear regression or GLM setting where these final activations act as the usual predictors. \end{rem}

Note that we consistently cover non-fully connected FFNNs where some neuron connections do not exist, which can be described by weights that are frozen at zero.

\subsection{Breakdown point for $Y$-contamination}

\begin{prop} \label{trimnn:bdpycont} A regression FFNN with an unbounded output activation function $\sigma$ with $|\sigma(z)| \rightarrow \infty$ for $|z| \rightarrow \infty$ whose gradient does not vanish for $|z| \rightarrow \infty$ and unbounded loss function $L$ with $L(r) \rightarrow \infty$ for $|r| \rightarrow \infty$ for residuals $r$ and non-redescending gradient which may be regularized with a penalty term of the form $\lambda J(W)$ for $J \ge 0$, $J(W) \rightarrow \infty$ for $||W|| \rightarrow \infty$ and $J(0)=0$, has a FSBDP of $1/n$ w.r.t. $Y$-contamination if non-trimmed average gradients are computed in BP. \end{prop}

\begin{bew} Let wlog. $Y_1$ be contaminated. 

\textbf{i)} In the case of a loss function $L$ with unbounded gradient and non-manipulated gradients, $Y_1$ can be manipulated so that $|L'(u-u')|_{u=Y_1,u'=z^{(H+1)}}| \rightarrow \infty$. Then, the first gradient in the first BP step, i.e., $\nabla_{a^{(H+1)}}L \ \circ \ \sigma'(z^{(H+1)})$ is already unbounded almost-surely, leading to unbounded output weights via the first update. This is also true if a regularization term of the form $\lambda J(W)$ is added to the loss function, since, due to $L$ being unbounded, one can always find $Y_1$ so that the loss term is dominant due to the penalty parameter $\lambda$ being static. Regardless of the remaining backward pass, the following forward pass leads to unbounded output activations by assumption. More precisely, due to non-vanishing gradients of the output activation function and the unbounded gradients of the loss function, the total gradients in the BP pass that update the output weights are unbounded, always ending with unbounded output weights, therefore, to a breakdown. Note that due to Def. \ref{trimnn:bdpnn}, minibatch GD that randomly selects $m<n$ instances whose gradients are aggregated does not safeguard against contamination as the probability that only clean instances are selected converges to zero for a growing number of iterations, so eventually, i.e., for epoch number $k \rightarrow \infty$, such an instance will almost-surely be selected.

\textbf{ii)} The case of clipped gradients, resilient backpropagation (Rprop) and bounded gradients can be handled simultaneously. Contamination of $Y_1$ has the effect that the maximum possible gradient $|L'(u-u')|_{u=Y_1,u'=z^{(H+1)}}|$ in absolute value can be achieved. Therefore, although the gradient is bounded, either by bounded gradients of the loss function, gradient clipping or Rprop which only considers the sign of the gradient, $Y_1$ always induces a non-zero maximum gradient (whose absolute value is $\eta$ in Rprop) due to the assumption that the gradient is non-redescending. As pointed out in the discussion after Def. \ref{trimnn:bdpnn}, robustness solely achieved by early stopping is ignored here, therefore, letting $k \rightarrow \infty$ for the number $k$ of training epochs, at least the output weights eventually diverge as a non-zero constant (positive for $Y_1 \rightarrow \infty$, negative for $Y_1 \rightarrow -\infty$) is added in each iteration to the output weights, hence, a breakdown in the sense of Def. \ref{trimnn:bdpnn} occurs. \vspace{-0.5cm} \begin{flushright} $_\Box $ \end{flushright} \end{bew}

Prop. \ref{trimnn:bdpycont} shows that the robustification of the training procedure, altough undoubtly solving or at least safeguarding against the exploding gradient problem, does not lead to robustness in the classical sense of quantitative robustness in our modified notion in Def. \ref{trimnn:bdpnn} if neither the loss function nor the aggregation of the losses is robust, hence the outliers can persistently pull the weights eventually to unbounded values. 

The tides turn if one robustifies the aggregation procedure of either the loss or the gradients by the classical trimming.

\begin{prop} \label{trimnn:bdpyconttrim} A regression FFNN with an unbounded output activation function $\sigma$ with $|\sigma(z)| \rightarrow \infty$ for $|z| \rightarrow \infty$ whose gradient does not vanish for $|z| \rightarrow \infty$ and unbounded loss function $L$ with $L(r) \rightarrow \infty$ for $|r| \rightarrow \infty$ and non-redescending gradient and where one computes the average of the $h$ gradients with the smallest norm in BP in each training epoch, has a BDP of $(n-h+1)/n$. \end{prop}

\begin{bew} Assume that there are $(n-h)$ large outliers which, as in Prop. \ref{trimnn:bdpycont} produce unbounded losses and gradients. Due to trimming the largest (in absolute value) $(n-h)$ gradients away, these outliers cannot harm the training procedure, therefore, the BDP is at least $(n-h+1)/n$. 

Conversely, assume that the are $(n-h+1)$ large outliers. Even trimming away $(n-h)$ of them results in one remaining large outlier, so a breakdown occurs due to the unboundedness of the gradients, as detailed out in the proof of Prop. \ref{trimnn:bdpycont}. Hence, the BDP is at most $(n-h+1)/n$. \vspace{-0.5cm} \begin{flushright} $_\Box $ \end{flushright} \end{bew}

The difference between trimmed means of gradients and mini-batch GD is that mini-batch GD randomly samples gradients while trimming targetedly removes the largest gradients. Note that one cannot treat the case of a trimmed loss aggregation and a trimmed gradient aggregation simultaneously in Prop. \ref{trimnn:bdpyconttrim} as both procedures are in general not equivalent. For example, if one would use Tukey's biweight loss (which is not covered by Prop. \ref{trimnn:bdpyconttrim} due to the vanishing gradients) and an additional trimming in the aggregation and if there are $(n-h)$ large outliers which fall into the constant area in the biweight loss, one would trim away these instances when trimming the loss aggregation, however, as these large outliers correspond to zero gradients, one would not trim away them during trimmed gradient aggregation (which would not be harmful in the sense of the BDP as these gradients are zero). For classical loss functions like the squared loss that satisfy both $L=L(r)$ with $L(r) \rightarrow \infty$ for $|r| \rightarrow \infty$ as well as $\nabla L=\nabla L(r)$ with $|\nabla L(r)| \rightarrow \infty$ for $|r| \rightarrow \infty$, the equivalence obviously holds, hence upper $\alpha$-trimmed squared losses would indeed induce a BDP of $\alpha$. As consequence, when using loss functions like the Tukey loss, one large outlier, even if not trimmed away during gradient or loss aggregation, may not necessarily be able to already lead to a breakdown.

\begin{rem} We want to point out that we only consider outlier robustness here due to the upper trimming, i.e., trimming away the largest losses or gradients. Inlier robustness would require also a lower trimming. In this work, we consider only the classical \textbf{static contamination} in the sense that the attacker can only manipulate a selected subset of instances before the training starts. One could, due to the iterative training procedure, also consider an attacker who can intercept this process and adapt the contamination (at least for the initially selected instances), leading to \textbf{iterative contamination}. 

In the setting of iterative contamination, trimming away a large fraction of the highest losses can indeed backfire when concerning robustness. For example, consider an upper $\alpha$-trimmed squared loss with $\alpha=0.5$ and assume for simplicity that $n$ is even. Then, provided that the attacker can query the output of the neural network in each training epoch, the attacker can, for the selected half of the instances, manipulate the responses in a way that they are always slightly larger than the currently predicted values but that they correspond to the lowest half of the losses. More precisely, let $L_i^{(k)}=L(Y_i^{(k)},\hat Y_i^{(k)})$ for all $i=1,...,n$ in iteration $k$ and let $L_{(1)}^{(k)} \le ... \le L_{(n)}^{(k)}$. Let $I^{(k)}$ be the index set corresponding to the $n/2$ smallest losses. The iterative replacement outliers $\hat Y^{(k+1)}$ after finishing iteration $k$ would be taken from the set \begin{center} $ \displaystyle \{Y \ | \ L(Y,\hat Y_i^{(k)})<L_{(n/2+1)}^{(k)}, Y>\hat Y_i^{(k)}\} $\end{center} for all $i \in I^{(k)}$. This will result in trimming away all instances $j$ for $j \in \{1,...,n\} \setminus I^{(k)}$ and, by construction, a (potentially small but) positive gradient. After the following forward pass, the attacker adapts the responses again according to this scheme and so forth, making the output weights eventually arbitrarily large. \end{rem}

\subsection{The case of $X$-contamination} 

The effect of $X$-contamination is much more difficult to handle than that of $Y$-contamination due to the strong dependence of the influence of such contamination on the architecture and the weights. We can elaborate it more precisely in the following proposition.

\begin{prop} \label{trimnn:xcontprop} A FFNN that uses standard BP has a case-wise BDP of $1/n$ if the initial weights allow for the error terms $\delta_l^{(1)}$, corresponding to the input layer and considered as a function of an input vector $x$, to decrease with at most a sublinear rate. \end{prop}

\begin{bew} In BP, the gradients in the input layer are given by $\partial_{w_{lj}^{(1)}} L=z_j^{(0)}\delta_l^{(1)}=x_j\delta_l^{(1)}$ (see. Eq. \ref{trimnn:bp34}), so if the assumptions are satisfied, a large $X$-outlier, wlog. $X_1$, can produce unbounded gradients here and, therefore, unbounded weights for the first hidden layer in the first BP step.  \vspace{-0.5cm} \begin{flushright} $_\Box $ \end{flushright} \end{bew}

\begin{cor} In the same setting as in Prop. \ref{trimnn:xcontprop}, cell-wise $X$-contamination would lead to a cell-wise BDP of $\frac{1}{np}$. \end{cor}

This extreme setting considered in the proof above then carries over to the following forward pass where either all predicted responses (not only $\hat Y_1$!) become unbounded, leading to the problems we examined for $Y$-contamination, or where the large weights of the first hidden layer cause intermediate linear terms $a_l^{(h)}$ to be located at near-flat regions of the following intermediate activation function $\sigma^{(h)}$ that, for $X_{1,j} \rightarrow \infty$, in fact prevent from any further weight updates in the subsequent BP steps when using standard BP, while it may still allow for updates using Rprop if the gradients of the activation functions are not exactly zero. Nevertheless, the problem of diverging initial weights in the setting of Prop. \ref{trimnn:xcontprop} cannot be remedied. 

However, it is not evident under which conditions the assumptions hold unless a particular NN with fixed initial weights are given. This is true since even ReLU activation functions have flat regions. For example, a simple NN with one hidden layer and two hidden nodes with ReLU activation is not automatically affected by large $X$-outliers. Let $p=1$ for simplicity. Then, letting $X_1 \rightarrow \infty$, a negative initial weight $w_{11}^{(1)}$ would make the activation $a_{1}^{(1)}$ negative and therefore censored by the ReLU activation, leading to a vanishing gradient w.r.t. instance 1 in the BP, so the aggregated gradient in BP will not be affected by the outlier. However, due to the gradients w.r.t. the other instances, the weight $w_{11}^{(1)}$ may become positive in later iterations. Due to this fact, a rigorous assessment of the effect of $X$-contamination is out of scope for this paper. 

We restrict ourselves to a practical analysis of the effect of $X$-contamination to FFNN training which, in our opinion, has not yet been sufficiently done in literature. Especially the case of cell-wise contamination does not seem to yet have been considered at all for FFNNs. Although the effect of $X$-contamination may be bounded if bounded activation functions are used, it is not evident how a NN behaves in the presence of cell-wise outliers which easily affect all instances while keeping the cell-wise contamination rate rather low, especially for high-dimensional predictors.

\section{Simulation study} \label{trimnn:simsec}

In order to experimentally examine the quantitative robustness of regression FFNNs, we set up an extensive simulation study that shall shed light on the performance, the number of training steps and the breakdown behaviour of robust and non-robust FFNNs for non-contaminated data, $Y$-contaminated data, case-wise $X$-contaminated data and cell-wise $(X,Y)$-contaminated data, all with convex contamination, for different dimensions, underlying regression structures and contamination radii.

\subsection{Data generation}

The data are generated by first drawing $X_i \sim \mathcal{N}_p(\mu 1_p, I_p)$ i.i.d., $i=1,...,n$, for $1_p=(1,...,1) \in \R^p$ and the identity matrix $I_p \in \R^{p \times p}$, and coefficients $\beta_j \sim \mathcal{N}(0,1)$ for $j=1,...,p$ i.i.d.. We consider three different regression structures:

\textbf{1.) Structure ``lin'':} We have a standard linear model of the form $Y_i=X_i\beta+\epsilon_i$.

\textbf{2.) Structure ``poly'':} We first compute $\tilde Y_i=X_i\beta$ and transform the responses according to $Y_i=|\tilde Y_i|^{-2/3}+\epsilon_i$ afterwards. This structure has been proposed in literature, e.g., \cite{rusiecki14}.

\textbf{3.) Structure ``trig'':} We again first compute $\tilde Y_i=X_i\beta$ and transform the responses according to $Y_i=\frac{\sin(|\tilde Y_i|)}{|\tilde Y_i|}+\epsilon_i$, which has also been proposed e.g. in \cite{rusiecki14}.

Evidently, for the linear structure, one would not need neural networks, however, we aim at comparing the performance of the FFNNs also in this setting. The errors $\epsilon_i$ are drawn i.i.d. according to a Gaussian distribution where the variance is adapted according to a pre-scribed signal-to-noise ratio ($\SNR$). The different configurations are collected in Tab. \ref{trimnn:scen}. 

\begin{table}
\begin{center}
\begin{tabular}{|p{2cm}|p{2cm}|p{2cm}|p{2cm}|p{2cm}|} \hline 
$p$ & $n_{train}$ & $n_{test}$ & $\SNR$ & $\mu$ \\ \hline
5 & 150 & 50 & 2  & 0 \\ \hline
20 & 500 & 200 & 2  & 0 \\ \hline
50 & 1000 & 500 & 2  & 0 \\ \hline
\end{tabular}
\end{center}
\caption[Data generation specification]{Data generation specifications} \label{trimnn:scen}
\end{table}

The training sets contain $n_{train}$ instances, the (independent) test sets $n_{test}$ instances. For each configuration, we draw $V=100$ independent data sets.

\subsection{Perturbation generation}

In the non-contaminated setting, the training set is used for training as it is. Note that the test data are never perturbed.

As for $Y$-contamination, we randomly select $\lceil rn_{train} \rceil$ responses for the contamination radius $r$ in the convex contamination model (see Ex. \ref{trimnn:convcont}) and replace them with i.i.d. realizations of a $\mathcal{N}(\mu_{out},1)$-distribution. In the $X$-contamination setting, we randomly select $\lceil rn_{train} \rceil$ instances and replace the $p$-dimensional predictor vectors with i.i.d. realizations from a $\mathcal{N}_p(\mu_{out}1_p,I_p)$-distribution. Finally, in the cell-wise $(X,Y)$-contamination setting, we randomly select $\lceil rn_{train}(p+1) \rceil$ cells and replace them with i.i.d. realizations from a $\mathcal{N}(\mu_{out},1)$-distribution. We set $r \in \{0.1,0.25,0.4\}$ and $\mu_{out} \in \{10,100,1000\}$.

\subsection{Network training}

We use the implementation from the $\mathsf{R}$-package \texttt{neuralnet} (\cite{neuralnet}). In order to allow for robust networks, we added Huber's loss function, Tukey's loss function and the upper trimmed squared loss as well as their respective gradients to the \texttt{convert.error.function} function.

As for network training, we only modify the parameters \texttt{err.fct}, \texttt{act.fct}, \texttt{stepmax} and \texttt{hidden}, for the other parameters, the default configurations are used. In particular, we use the resilient BP algorithm here, since experiments with large outliers lead to exploding gradients when using the standard BP algorithm.

We consider two different activation functions (that are used in each hidden node, note that the output activation is the identity function as we are in a regression setting), namely, the logistic activation function and the softplus activation function $x \mapsto \ln(1+e^x)$ which is a classical smooth approximation of the ReLU activation function (e.g., \cite{goodfellow16}). 

We consider six loss functions/aggregations, i.e., the squared loss, Huber's loss, Tukey's loss and the upper $\alpha$-trimmed squared loss (just trimmed (squared) loss henceforth) with a trimming rate $\alpha \in \{0.1, 0.25, 0.5\}$. As for the depth and size of the FFNNs, we use shallow networks with $H=2$ hidden layers with $L_h=10$ hidden nodes, $h=1,2$, and deep networks with $H=10$ and $L_h=5$ hidden nodes, $h=1,...,10$. The resulting number of weights is listed in Tab. \ref{trimnn:nnweights}.  As for the Tukey loss, we use the value $k=4.685$ from literature, corresponding to 95\% efficiency (e.g., \cite{maronna}). As for the Huber loss, we compute $\delta=med((|Y_i-\hat Y_i|)_{i=1}^n)$ in each iteration.

\begin{table}
\begin{center}
\begin{tabular}{|p{1cm}|p{1cm}|p{3cm}|p{2.5cm}|p{4cm}|} \hline 
$p$ & $L_h$ & \# Intercepts $b_l^{(h)}$ & \# Weights $w_{jl}^{(h)}$ & Total number of weights  \\ \hline
5 & (10,10) & 21 & 160  & 181 \\ 
5 & $5 \cdot 1_{10}$ & 51 & 255 & 306 \\ \hline
20 & (10,10) & 21 & 310  & 331 \\ 
20 & $5 \cdot 1_{10}$ & 51 & 330 & 381 \\ \hline
50 & (10,10) & 21 & 610  & 681 \\ 
50 & $5 \cdot 1_{10}$ & 51 & 480 & 531 \\ \hline
\end{tabular}
\end{center}
\caption{Number of neural network weights} \label{trimnn:nnweights}
\end{table}

For the shallow neural networks, we allow for a maximum epoch number \texttt{stepmax} of 100000, while deep networks are allowed to iterate through 250000 epochs. 

Finally, we consider both standardized and non-standardized responses, i.e., in the case of a standardization, we compute \begin{center} $ \displaystyle Y_i \mapsto \frac{Y_i-\min_j(Y_j)}{\max_j(Y_j)-\min_j(Y_j)} $ \end{center} as a pre-processing step before training which is a common technique to stabilize the training procedure.

Summarizing, we consider a factorial design of the following configurations listed in Tab. \ref{trimnn:config}. 

\begin{table}[H]
\begin{center}
\begin{tabular}{|p{1.25cm}||p{1cm}||p{1cm}||p{0.5cm}||p{0.5cm}||p{1.5cm}||p{1.75cm}||p{2cm}||p{1.1cm}|} \hline 
$(n,p)$ & Struc\-ture & Cont. & $r$ & $\mu_{out}$ & Activation function & Loss function & Standar\-dization & Network  \\ \hline
(150,5) & ``lin'' & None & 0.1 & 10 & logistic & squared & True & Shallow \\ \hline
(500,20) & ``poly'' & $Y$ & 0.25 & 100 & softplus & Huber & False & Deep \\ \hline
(1000,50) & ``trig'' & $X$ & 0.4 & 1000& & Tukey & &  \\ \hline
&& $(X,Y)$ (cell-wise) &&&& Trimmed squared ($\alpha=0.1$) & &  \\ \hline
&&&&&& Trimmed squared ($\alpha=0.25$) &  &  \\ \hline
&&&&&& Trimmed squared ($\alpha=0.5$) & & \\ \hline
\end{tabular}
\end{center}
\caption{Simulation configurations} \label{trimnn:config}
\end{table}

Due to the factorial design, we have 15552 different configurations in total.

\begin{rem}[Trimmed loss] Trimming has been successfully applied to regression tasks in the Least Trimmed Squares (LTS, \cite{rous84}) and Sparse Least Trimmed Squares (SLTS, \cite{alfons13}) algorithm. Both invoke so-called C-steps (\cite{rous00}, \cite{rous06}), i.e., concentration steps, based on the following argument: Having initially trained the regression model on a random $h$-subset of the data set, identify the $h$ instances with the smallest losses (originally, with the smallest absolute residuals, however, for a loss function of the form $L(u,u')=L(|u-u'|)$ which holds for the losses we use in this work, both statements are equivalent) and re-compute the estimator on these clean instances. Denote by $R_H(\hat \theta)$ the average loss on the $h$-subset $H \subset \{1,...,n_{train}\}$ for regression parameter $\hat \theta$. In iteration $k$, for the current parameter $\hat \theta^{(k-1)}$, by definition of the clean $h$-subset $H^{(k)}$, it holds that $R_{H^{(k)}}(\hat \theta^{(k-1)}) \le R_{H^{(k-1)}}(\hat \theta^{(k-1)})$. Then, by computing $\hat \theta^{(k)}=\argmin_{\theta}(R_{H^{(k)}}(\theta))$, it holds that $R_{H^{(k)}}(\hat \theta^{(k)}) \le R_{H^{(k)}}(\hat \theta^{(k-1)})$, so the loss monotonically decreases. 

In the setting of neural networks, due to the iterative and time-consuming training procedure, the pragmatic solution which is also used for the computations in this paper does not enable C-steps. If one would indeed compute $\hat \theta^{(k)}=(\hat B^{(k)}, \hat W^{(k)})$ for each $k$ by training the corresponding neural network until convergence, one would have the same argument as in LTS or SLTS that one at least attains a local minimum of the trimmed objective. The practical problem is however that this procedure would be very time-consuming. The computation of multiple neural networks until convergence is usually avoided, for example, instead of Bagging neural networks, one applies Dropout (\cite{sriva}) where in each iteration, one randomly drops nodes so that one artificially creates multiple networks without having to train multiple networks until convergence.

For the trimmed squared loss however, we effectively only localize the loss and hence the gradient to the clean $h$-subset with $h=\lceil (1-\alpha)n_{train}\rceil$ for one single epoch. One can interpret this as an approximate C-step, however, one single gradient step is not guaranteed to lead to a better solution as, for example, a too large step size in the gradient update may cause the updated parameter to jump over the local optimum in the parameter space and be in fact worse than the previous one. \end{rem}

\subsection{Evaluation}

In our simulations, we always use batches of all 24 configurations concerning the loss function and the contamination for all of the 648 configurations of the other parameters. For each of these 648 configurations, we evaluate the performance of the neural networks trained w.r.t. the six different loss functions in each of the four different contamination scenarios. 

The performance consists of three aspects: The test loss, the number of training epochs and the number of successfully terminated training procedures. A training procedure is successfully terminated if convergence of the partial derivatives up to a given threshold (we use the default of 0.01) happened within the maximum allowed number of training epochs. In case of successful training, we can compute the test loss, for which, since we are in a regression setting, we simply compute the average squared test loss, i.e., \begin{center} $ \displaystyle \frac{1}{n_{test}}\sum_{i=1}^{n_{test}} (\hat Y_i-Y_i)^2. $ \end{center} Moreover, if training was successful, we extract the number of epochs required.

The test losses and training epoch numbers are then averaged over all (out of $V=100$) successful trainings so that for each of the 648 configurations, we can compute the average test loss and average training epochs for all of the 24 (loss function, contamination)-configurations. These averages are depicted graphically, separately for the test losses and training epochs. These graphics also contain the number of successful trainings which are written on top of the individual bars. As a by-product, the relative number of failed trainings can be interpreted as a surrogate of the breakdown rate. In some cases, in particular, for softplus activation, the test loss can be that large that $\mathsf{R}$ reports \texttt{Inf} so that the average test loss would also be $\infty$ which would be misleading. Therefore, we compute the average of the finite test losses and report in the graphics that there was at least one infinite loss by writing ``Inf'' over the respective error bar.

Due to the large number of 1296 graphics, we moved them to the appendix.

\section{Results} \label{trimnn:ressec}

In this section, we summarize the results of our simulation study. Due to the large number of different parameters, we analyze the impact of each parameter separately in an own subsection.

\subsection{Output standardization}

Regarding the test losses, evidently, the mean losses are smaller in the scenarios where an output standardization was used than in scenarios without output standardization. Note that the ranking of the models in terms of their mean test losses in the scenarios with output standardization does not need to coincide in the corresponding scenarios without output standardization since this scaling of the responses does not directly carry over into a scaling of the model as we do not have linear models here. 

Notably, the number of NNs that have converged is considerably higher for the scenarios with output standardization than in the scenarios without output standardization. Due to the standardization, large outliers have the effect that the original responses are crowded near the left side of the interval $[0,1]$ into which they are transformed while the large outliers are crowded near the right side. It is important to emphasize that robustness has to be defined carefully when operating in bounded spaces, see \cite{davies05}, as, in this standardized setting, it would be impossible to achieve a breakdown in the classical sense of an unbounded norm of the coefficients/weights of the model (in our particular setting, if only $Y$-contamination is allowed as the regressors are not standardized). Although a breakdown would also not be achievable in our breakdown point notion in Def. \ref{trimnn:bdpnn}, the results show that, given a maximum number of iterations, the NN algorithm was unable to converge in many cases in settings with a large contamination magnitude and a large contamination radius.

We want to point out that the number of converged networks does not necessarily need to be higher in a setting with standardized responses than in the corresponding setting without output standardization, see Fig. \ref{trimnn:n200p5r25m1polynonrelu} with a $\mu_{out}=10^3$ for one example where, in the case of $Y$-contamination, the number of Tukey-NNs that converged is higher in the non-standardized case (76) than in the standardized case (9). The reason is that the Tukey loss can lead to the vanishing gradient problem which we will refer to in more detail later in Sec. \ref{trimnn:sec:loss}. If the softplus activation function is used, as expected, output standardization cannot deal with large $X$- and $(X,Y)$-contamination, see e.g., Fig. \ref{trimnn:n1000p50r40m1trignonrelu} since we only use the output standardization once as data preprocessing step, so after the first iteration, the predicted responses suffer from the $X$-outliers, preventing the NN from converging within the maximum number of iterations.

The number of iterations is generally lower if output standardization is used and often even dramatically lower, although the mean number of iterations does not necessarily have to be lower in a scenario with output standardization than in the corresponding setting without output standardization, see e.g. Fig. \ref{trimnn:n200p5r10m1trignonreluStep} where, on the non-contaminated data, the average number of iterations for the Tukey NN has increased from less than 30000 to more than 40000 due to the output standardization.

\subsection{Contamination}

Of course, a higher contamination magnitude and a higher contamination radius generally lead to a lower number of converged NNs and a higher test loss. However, occasionally, a higher amount of contamination may even lead to a better test loss, in particular, this can be the case for the Tukey NN, see e.g. Fig. \ref{trimnn:n200p5r25m1trignonlog}, \ref{trimnn:n200p5r40m1trignonlog}, or a higher number of converged NNs, e.g., in Fig. \ref{trimnn:n200p5r10m1linnonrelu}, \ref{trimnn:n200p5r25m1linnonrelu}, \ref{trimnn:n200p5r40m1linnonrelu}, both for the non-standardized case and $Y$-contamination.

It is important to point out that the number of converged NNs does not necessarily have do decrease for less robust NNs with a growing amount of contamination, in particular, the NN based on the squared loss. One can observe that, oddly, the number of converged NNs w.r.t. the squared loss and the $0.1$-trimmed squared loss increases with the contamination magnitude and the contamination radius, at least for the logistic activation function, e.g., Fig. \ref{trimnn:n200p5r10m1linnonlogdeep}, \ref{trimnn:n200p5r25m1linnonlogdeep}, \ref{trimnn:n200p5r40m1linnonlogdeep}. However, note that the test loss increases. One may interpret this observation by a gradient enhancement thanks to the contamination so that oscillating gradients (w.r.t. the subsequent epochs) that may prevent convergence (within the allowed maximum number of epochs) are avoided, nevertheless, one can clearly not speak of a reasonable fit due to the large test losses, strongly indicating that the NNs were highly distorted by the outliers. For other regression structures, due to the low test losses of the more robust models, the bad generalization performance of the distorted fit of the NNs w.r.t. the squared loss can be dramatically high, for example, up to more than 6 orders of magnitude higher than for the robust models in Fig. \ref{trimnn:n1000p50r10m1trignonlog}, \ref{trimnn:n1000p50r25m1trignonlog}, \ref{trimnn:n1000p50r40m1trignonlog}. This effect of an increased number of converged NNs w.r.t. non-robust losses for higher contamination radii and magnitudes cannot be observed for the softplus loss function, not even if output standardization is used, most likely due to exploding gradients against which the Rprop method does not safeguard in the case that the raw gradient whose sign is computed is already infinite (or at least no longer representable by the software).

As for the number of iterations, there is no clear effect of contamination or the different types of contamination.

\subsection{Activation function}

Evidently, the logistic activation function bounds the activations of the respective hidden nodes, but clearly does not prevent the NN from making large-valued predictions due to the weights in the output layer and the identity output activation for regression, neither does it safeguard against $Y$-contamination.

Most interestingly, $X$-contamination generally only leads to a slight increase of the test loss for small data for logistic activation, and can in some cases even lead to a better out-of-sample performance than in the corresponding non-contaminated situation, see Fig. \ref{trimnn:n1000p50r10m1polynonlog}, \ref{trimnn:n1000p50r25m1polynonlog}, \ref{trimnn:n1000p50r40m1polynonlog}, in the case of output standardization. However, $X$-contamination has a strong effect for softplus activation, leading to a large test loss and a decreasing number of converged NNs. Due to the bounded logistic activation and the redescending gradient of this function, the predicted responses for instances containing $X$-outliers cannot become infinite and hence do not distort the training process much, in contrast to the unbounded softplus activation.

In this sense, one can conclude that a bounded activation function in general can provide a robustification against $X$-contamination but not against $Y$-contamination.

\subsection{Network structure} \label{sec:struc}

As for the network structure, there is one striking unexpected behaviour: Although the squared loss function induces a highly non-robust network, the number of converged NNs is generally much larger for deep networks than for shallow networks and, oddly, may be even higher on contaminated data than on non-contaminated data. This behaviour can sometimes also be observed for shallow networks (e.g., Fig. \ref{trimnn:n500p20r10m1polynonlog} for $\mu_{out}=10^3$), but it is characteristic for deep networks for every regression structure, but only without output standardization and for the logistic activation function. We interpret this behaviour as a clear \textbf{overfitting} problem of deep networks. Although current research indicates that there is a so-called ``interpolating regime'' (e.g., \cite{fissler22}) or ``modern regime'' (e.g., \cite{berner}) where the statistical risk corresponding to a model, after increasing with increasing model complexity due to overfitting, may decrease again if the model complexity becomes very large, leading to a double descent risk curve (\cite{berner}). However, although this behaviour has been proven for data from an ideal distribution with certain assumptions on the tail behaviour (e.g., \cite{bartlett20}, \cite{belkin}, \cite{tsigler}) for linear models and, in particular, in \cite{frei} for non-linear models, a tight approximation of the training data may certainly have to be understood as overfitting if the training data are vastly contaminated as near-interpolating such data cannot lead to a well-generalizing model, which is confirmed by the large test losses. Note that the work of \cite{frei} indeed shows that for neural networks trained according to the logistic loss, even adversarial label flipping is allowed while still achieving a double descent risk curve, so a certain amount of contamination could still be coped with, although our setting is a regression setting and hence structurally different.

In addition, it seems that in combination with the softplus activation function, deep neural networks are much more brittle than shallow networks, even on non-contaminated data (see Sec. \ref{trimnn:secloss1505}, \ref{trimnn:secloss50020} and \ref{trimnn:secloss100050} in comparison with Sec. \ref{trimnn:secloss1505deep}, \ref{trimnn:secloss50020deep} and \ref{trimnn:secloss100050deep}), in particular, the cases with output standardization where the test loss is generally higher for the deeper networks. The difference is particularly striking for the Huber NNs. This could be explained by the fact that a deep network contains even more summations, and due to the unboundedness of the softplus activation, the activations grow in each hidden layer and if they do not become negative so that they are clipped, they de-stabilize the training. Another reason could be that the bound of 250000 epochs we fixed may not be sufficient for convergence in these settings.

Concerning the training steps, one can clearly observe that in the cases where the deep network nearly interpolates the contaminated training data for non-robust networks, the number of training epochs is rather small, indicating that, due to the large number of network parameters, the network can quickly adapt to the data due to the squared or $0.1$-trimmed squared losses not or rarely trimming gradients away in contrast to the other loss functions.

\subsection{Dimensionality and regression structure}

Contamination shows similar effects concerning the number of converged NNs, the required training epochs and the test losses, disregarding the dimensionality of the data. One has to be careful when comparing the results among the respective sections as for a growing dimensionality of the data, the NNs tend to even not converge within our fixed maximum number of training epochs on non-contaminated data. Nevertheless, the fact that, in average, the number of converged NNs decreases with the amount of contamination holds for all dimensionalities.

The regression structure leads to a different loss scale which is smaller for the trigonometric structure than for the linear and the polynomial structure. The effect of contamination however is similar among all regression structures.

\subsection{Loss function} \label{trimnn:sec:loss}

This is the main part of the analysis of the results.

The most non-robust loss functions, i.e., the non-trimmed and the $0.1$-trimmed squared loss, suffer most from contamination which can be clearly seen in the simulation results. Even for logistic activation, the number of converged NNs quickly decreases with the contamination magnitude and radius. As already highlighted in Sec. \ref{sec:struc}, sometimes the number of converged NNs even increases with the amount of contamination, e.g., Fig. \ref{trimnn:n200p5r40m1linnonlog} for $\mu_{out}=10^3$ and without output standardization, but for the price of an immense test loss due to overfitting. Concerning the test loss, for the non-robust NNs, they are prevented from becoming extremely large in our simulations when using output standardization, nevertheless, the effect of contamination is clearly visible. A very important additional result is that the NN based on the non-trimmed squared loss does not necessarily correspond to the smallest test losses when trained on non-contaminated data. There are configurations for which the NNs based on the non-trimmed squared loss even lead to the largest test losses if trained on non-contaminated data, see Fig. \ref{trimnn:n500p20r10m1trignonlog} for the case without response standardization. On top of that, there are many configurations in which this NN was not able to converge within the maximum number of training epochs, even on non-contaminated data.

As for the number of training epochs, one can observe that the NNs based on the non-trimmed squared loss and the $0.1$-trimmed squared loss are often among the networks with the highest required number of training epochs (together with the Tukey NNs), on both non-contaminated and contaminated data, in particular, when using output standardization. Nevertheless, there are also situations where the number of iterations is smaller on contaminated data than on non-contaminated data and lower than for NNs that are trained w.r.t. other loss functions, see, e.g., Fig. \ref{trimnn:n200p5r40m1trignonlogStep} for $\mu_{out}=10^3$ and without output standardization. Unexpectedly, on the already mentioned situations where the NN overfits to contaminated data, e.g., Fig. \ref{trimnn:n500p20r10m1polynonlogdeep}, \ref{trimnn:n500p20r25m1polynonlogdeep}, \ref{trimnn:n500p20r40m1polynonlogdeep}, \ref{trimnn:n500p20r10m1polynonlogdeepStep}, \ref{trimnn:n500p20r25m1polynonlogdeepStep}, \ref{trimnn:n500p20r40m1polynonlogdeepStep}, for $\mu_{out}=10^3$, the number of training epochs is rather low. An explanation could be that, although the Rprop algorithm is used that only considers the sign of the gradients, due to the large number of network weights and the fact that no gradients for any training instance are clipped for the non-trimmed squared loss, the NN is able to quickly adapt to the contaminated data so that the training loss decreases quickly, leading to small residuals and hence small gradients and hence numerical convergence, in contrast to NNs trained w.r.t. robust loss functions where the gradients for large outliers are trimmed or zero during a training epoch but where the set of trimmed instances may change from training epoch to training epoch, delaying convergence. Note that this near-interpolating behaviour of the NN trained w.r.t. the non-trimmed squared loss does not necessarily have to imply a low number of training epochs, see, e.g., Fig. \ref{trimnn:n500p20r25m1trignonlogStep} for $\mu_{out}=10^3$ without output standardization where the average number of training epochs for the case of $Y$-contamination is higher than for any other loss. \ \\

The $0.25$- and $0.5$-trimmed squared loss lead to models which show a quite well performance throughout our simulations and which can handle large contamination radii. If the contamination radius is 0.4, exceeding the trimming rate for the $0.25$-trimmed squared loss, one can occasionally indeed see that the performance is much worse than that of the NN trained according to the $0.5$-trimmed squared loss, e.g., Fig. \ref{trimnn:n200p5r40m1polynonlog} for the cases without output standardization for $\mu_{out}=10^2$ and $\mu_{out}=10^3$, but, although $25\%$ of the instances are trimmed away for the $0.25$-trimmed squared loss, it does not necessarily safeguard against settings with a contamination radius of $0.25$, see, e.g., Fig. \ref{trimnn:n200p5r25m1trignonlog} for $(X,Y)$-contamination and without output standardization where the NN trained according to the $0.5$-trimmed squared loss performs still remarkably well in contrast to the NN trained according to the $0.25$-trimmed squared loss. In the settings where the softplus activation function is used, the $0.5$-trimmed loss usually leads to models that have converged, if any model has converged at all, often accompanied with Tukey NNs. Notably, the $0.5$-trimmed squared loss shows sometimes outstanding behaviour in comparison with the other losses, see, e.g., Fig. \ref{trimnn:n500p20r10m1trignonlog}, \ref{trimnn:n500p20r25m1trignonlog}, \ref{trimnn:n500p20r40m1trignonlog}, where it leads to models which are clearly the best ones in any setting without output contamination while being still competitive in settings with output standardization. As the contamination radius increases, the corresponding models are among the best models. Of course, trimming too much information away can also result in undesired behaviour, see e.g. Fig. \ref{trimnn:n500p20r10m1polynonlogdeep}, \ref{trimnn:n500p20r25m1polynonlogdeep}, \ref{trimnn:n500p20r40m1polynonlogdeep}, where the trimmed losses with a high trimming rate induce weakly performing models on the non-contaminated data. 

As for the number of training epochs, intuitively, NNs trained according to a trimmed loss with a high trimming rate should require less iterations on contaminated data in order to converge due to trimming away the gradients of outliers. Of course, there are counterexamples, e.g., Fig. \ref{trimnn:n200p5r40m1linnonlogdeepStep} where these NNs require more epochs than most of the other NNs for the cases without output standardization and without contamination resp. with $Y$-contamination, but generally, the number of training epochs is lower than for most of the other NNs, often considerably lower, see again Fig. \ref{trimnn:n200p5r40m1linnonlogdeepStep} for the cases with output standardization for the $0.5$-trimmed loss or Fig. \ref{trimnn:n500p20r10m1trignonlog}, \ref{trimnn:n500p20r25m1trignonlog}, \ref{trimnn:n500p20r40m1trignonlog}, \ref{trimnn:n500p20r10m1trignonlogStep}, \ref{trimnn:n500p20r25m1trignonlogStep}, \ref{trimnn:n500p20r40m1trignonlogStep} mentioned above where the corresponding NN performed very well concerning the test losses in the cases without output standardization. However, too quick convergence due to excessive trimming can also be misleading, see, e.g., Fig. \ref{trimnn:n1000p50r40m1trignonlogdeep} and the corresponding Fig. \ref{trimnn:n1000p50r40m1trignonlogdeepStep} for the cases with output standardization where the NN trained according to the $0.5$-trimmed squared loss corresponds to high test losses in comparison with other models (except for the Huber NN) and where the number of training epochs is very low (as for the Huber NN) which may have prevented these models from approximating the (mostly clean part of the data) well.  \ \\

The Huber loss function does not lead to zero gradients, in contrast to the Tukey loss function where large residuals correspond to zero gradients or trimmed loss functions where losses (and hence gradients) are trimmed away. The robustification effect of the Huber loss function originates from the constant gradients for large residuals while the magnitude of the residuals still contributes to the gradients. However, as we use the Rprop algorithm, one can ask whether the robustification effect is still valid as only the sign of the averaged gradients is considered. Indeed, due to bounding the gradients in the Huber loss function, a single outlier cannot contribute much to this average gradient, i.e., the robustification effect here means that single outliers cannot have full control over the average gradient and hence of its sign. Therefore, the Huber loss also has a robustification effect in our simulations, but weaker than that of loss functions like the Tukey loss or the $0.5$-trimmed loss because in situations with high contamination radii, the outliers can still pull the average gradients and hence control their sign. 

Overall, the Huber NNs clearly suffer from contamination in the settings without output standardization, but perform very well in settings with output standardization and logistic activation and, for shallow networks, also with softplus contamination (see Sec. \ref{sec:struc} where we pointed out that for softplus activation, shallow networks tend to perform better than deep ones), often even leading to a better performance than the NNs trained according to the non-trimmed squared loss on non-contaminated data and being competitive on contaminated data. Note again that the effect of deep networks is particularly striking for the Huber NNs which perform generally very well for softplus activation and output standardization for shallow networks, but not for deep networks. However, in some settings, they do not perform as desired even for logistic activation, see, e.g., Fig. \ref{trimnn:n200p5r40m1trignonlogdeep}. 

As for the number of training epochs, in the cases without output standardization, their number is still lower than for the squared loss and often comparable or even better than for the Tukey loss or a $0.1$-trimmed squared loss, but for standardized output, one can clearly observe than the Huber NNs generally require much less training epochs than the NNs trained according to the squared loss while requiring generally more epochs than the NNs trained according to the $0.5$-trimmed squared loss. Interestingly, this performance aspect also degrades for deep networks where the number of training epochs is relatively high in cases without output standardization but while being in general still very low when using output standardization. \ \\

The Tukey NNs seem anomalous in many of our figures. In Sec. \ref{trimnn:secloss1505}, one can observe that without output standardization, the performance of the Tukey NNs is competitive for $Y$-contamination but the number of converged Tukey NNs quickly decreases with growing contamination magnitudes and radii if $X$-contamination is included. For data with output standardization, the test loss corresponding to the Tukey NNs is visibly higher than for most of the other NN, in particular, if $X$-contamination is involved. For deep networks in Sec. \ref{trimnn:secloss1505deep}, the behaviour is similar, although for low contamination radii, the performance of the Tukey NNs on data with output standardization and $X$- and $(X,Y)$-contamination is comparable to that of the other NNs and decreases significantly on data with $(X,Y)$-contamination when the contamination radius increases, see, e.g., Fig. \ref{trimnn:n200p5r10m1linnonlogdeep}, \ref{trimnn:n200p5r25m1linnonlogdeep}, \ref{trimnn:n200p5r40m1linnonlogdeep}. These characteristics carry over to Sec. \ref{trimnn:secloss50020} and \ref{trimnn:secloss100050} where the performance is even worse than on the smaller data, both in terms of test losses and converged NNs. In Sec. \ref{trimnn:secloss50020deep} however, at least the performance of Tukey NNs with logistic activation on data with $(X,Y)$-contamination and standardized responses is competitive and sometimes even best among the six different loss functions. The performance however generally decreases again in Sec. \ref{trimnn:secloss100050deep}.

As for the number of training epochs, the Tukey NNs converge rather quickly in the settings where Tukey NNs perform well, i.e., on data with $Y$-contamination and standardized responses and where the NNs use logistic activation functions. If $X$-contamination is included, the number of training epochs is in general very high, which may be a consequence of trimming many instances implicitly away due to zero gradients (as the threshold of $k=4.685$ is fixed during training), slowing down the training progress and, on top of that, misleading the NNs since the test losses are high in these cases. In particular in situations with softplus activation, one can also observe contrary behaviour, e.g., Fig. \ref{trimnn:n200p5r25m1linnonreludeep}, \ref{trimnn:n200p5r40m1linnonreludeep} and \ref{trimnn:n200p5r25m1linnonreludeepStep}, \ref{trimnn:n200p5r40m1linnonreludeepStep} for the case with output standardization and in particular with $\mu_{out}=10$. The test loss is large, but the number of training epochs is extremely low, evidently due to the large residuals resulting from the unbounded activation function so that a high number of instances receives a zero gradient, stopping training far too early. This issue obviously has occurred in very challenging situations like in Fig. \ref{trimnn:n500p20r10m1linnonreludeepStep} and \ref{trimnn:n500p20r25m1linnonreludeepStep} or Fig. \ref{trimnn:n1000p50r10m1linnonreludeepStep}, \ref{trimnn:n1000p50r25m1linnonreludeepStep}, \ref{trimnn:n1000p50r40m1linnonreludeepStep} for non-standardized responses where only Tukey NNs have converged, but with an awkwardly low average number of training epochs of around 1 or 2. In other words, while any other NN did not converge here, evidently due to exploding gradients, some Tukey NNs were able to perform a very few training epochs until any residual was so large that each instance has been trimmed, leading to zero gradients and hence numerical convergence.

\subsection{Discussion}

The goal of this work was to investigate the effect of different types of contamination on feed-forward regression networks with different network configurations, in particular, different loss functions. We are aware of the fact that in real data analysis, extreme outliers like the ones we generated in the settings with at least $\mu_{out}=10^3$ would very likely be detected when applying an outlier detection algorithm first. Nevertheless, contamination can have a masking effect (\cite{hampel}) in the sense that contaminated data points let other contaminated data points appear to be non-contaminated, making the detection of all outliers in truth very complicated. 

The results of our simulations indicate, as expected, that contamination decreases the performance of the trained NNs and prevents some NNs even from converging within a given number of training epochs. This observation holds throughout the whole simulation study, but on has to point out that the effect of $X$-contamination depends on the activation function since logistic activation decreased th effect of $X$-outliers due to the boundedness of the activation function so that a growing magnitude of contamination essentially has no further effect. Our results are consistent with those from literature, in particular, in \cite{rusiecki}, it also has been observed that the effect of $X$-contamination when using a sigmoid activation function is limited on the loss scale, however, NNs trained according to robust approaches (least median and least trimmed squares) still have shown a better performance in these settings.

Output standardization is essentially always done as it speeds up the training procedure. In this study, we nevertheless also wanted to examine cases without output standardization. As expected, the effect of contamination leads to a drastically decrease in performance, in particular, sometimes nearly no NN converges. As for output standardization, one could ask whether contamination has an effect as the responses are standardized to $[0,1]$. Nevertheless, contamination indeed significantly decreases the performance of the models, motivating to use robust NNs.

As for the robust NNs, we can conclude that the Tukey loss is not suitable for network training due to the vanishing gradients, preventing the NN from learning. This issue either results in a way too fast convergence or in a very high number of training epochs required for convergence. In any case, the performance of the Tukey NNs was not convincing throughout the simulation study. One could try to empirically adapt the parameter $k$ of the Tukey loss, but it does not seem necessary to focus on the Tukey loss here because, in contrast, the Huber NNs generally perform very well in the settings with output standardization and logistic activation, where the parameter $\delta$ has been adaptively chosen in each epoch. Moreover, they require generally less training epochs than the NNs trained according to the non-trimmed squared loss. As elaborated in the previous subsection, although the gradients of the Huber loss function are bounded and although we use the Rprop algorithm, the Huber NNs are not robust against large contamination amounts since a large group of outliers can have full control over the average gradients and hence the sign. Trimmed losses with a high trimming rate turned out to be highly competitive throughout the whole simulation study, both in terms of test losses, the number of converged NNs and the required training epochs. However, excessive trimming can also mislead the training process, apart from the fact that the trimming rate is another hyperparameter whose optimal value cannot easily be selected.

Summarizing, we recommend to replace the squared loss function in regression applications of feed-forward NNs by either the Huber loss function or a trimmed squared loss function.

\section{Conclusion}

In this paper, we formalized and investigated the global quantitative robustness of feed-forward regression neural networks.

First, we argued why the classical regression BDP from robust statistics is not suitable for measuring the quantitative robustness of a regression NN due to the iterative training procedure in combination with techniques like gradient clipping or resilient backpropagation that erroneously could let highly non-robust NNs appear to be robust. After having proposed an adapted version of the regression BDP for neural networks, we formally proved that a feed-forward NN with an unbounded output activation and an unbounded loss function has a BDP of zero while a trimmed loss aggregation indeed induces NNs with a positive BDP.

We conducted an extensive simulation study where we compared the performance of NNs trained according to the squared loss, the Huber loss, the Tukey loss and several trimmed squared losses on a plethora of different scenarios which differ by the amount of contamination, the size of the data, the activation functions, the standardization of the responses and the underlying true regression structure. We computed the average squared loss on independent and non-contaminated test data, the average number of training epochs and the number of converged NNs. It turned out that, as expected, the performance of the NNs trained according to the non-trimmed squared loss suffers from contamination, in particular, if $Y$-contamination is included, while the effect of $X$-contaminaton is bounded if logistic activation is used. One can observe that the overall behaviour of the Huber NNs and the NNs trained according to a trimmed squared loss with a high trimming rate ($0.25$, $0.5$) is best in all three aspects. 

We conclude that one should consider the usage of the Huber loss function or a trimmed squared loss for regression FFNNs.

Robustness of graphical convolutional networks has been considered in \cite{chen21b} who propose a median or trimmed mean aggregation of the information of the neighbors in order to robustify the networks. Future work should consider about a formal notion of robustness and the robustification of other types of neural networks.

\renewcommand\ref{References}
\bibliography{Biblio}
\bibliographystyle{abbrvnat}
\setcitestyle{authoryear,open={((},close={))}}

\newpage

\appendix
\counterwithin{figure}{section}

\section{Simulation results}
\sectionmark{Simulation results}

In the following, we compare the performance of the neural networks trained according to the six different loss functions on the $V=100$ clean, $Y$-, $X$- and $(X,Y)$-contaminated data sets. 

Sec. B-G illustrate the out of sample performance. The height of the bars in the bar plots represents the average out-of-sample loss where the average is taken over all data sets (out of $V=100$) where the algorithm has converged and where the loss is finite due to the fact that, in particular for softplus networks, the loss may become that large that $\mathsf{R}$ does no longer report a finite number. 

If there is a least one data set on which the network has produced an infinite loss, we highlight this fact with an ``Inf'' over the corresponding error bar. The number of converged networks is reported on top of the corresponding error bar. In cases where none of the $V=100$ networks has converged, the corresponding bar does not exist. 

Note that the $y$-axis is scaled logarithmically. The abbreviations ``Trim10'', ``Trim25'' and ``Trim50'' on the $x$-axis indicate that the $\alpha$-trimmed squared loss for $\alpha=0.1$, $\alpha=0.25$, $\alpha=0.5$, respectively, has been used for training. The test loss is always the non-trimmed squared loss. 

In Sec. H-M, we report the number of training epochs, averaged over all out of $V=100$ networks where the algorithm has converged. Again, the number of converged networks is reported on top of the corresponding bar. 

It is important to note that since we substract the number of infinite losses from the number of successful trainings, there can be an epoch number bar if there is no corresponding loss bar in the case that at least one network has converged but that the corresponding test loss is always infinite.

\section{Simulation results for $n=150$ and $p=5$: Test loss} \label{trimnn:secloss1505}

\subsection{Logistic activation function}

\subsubsection{Linear function}

\begin{figure}[H]
\label{trimnn:n200p5r10m1linnonlog}
\begin{center}
\includegraphics[width=6.75cm,height=6.25cm]{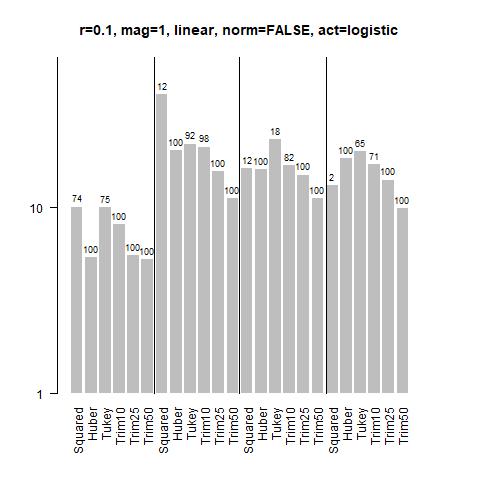}
\includegraphics[width=6.75cm,height=6.25cm]{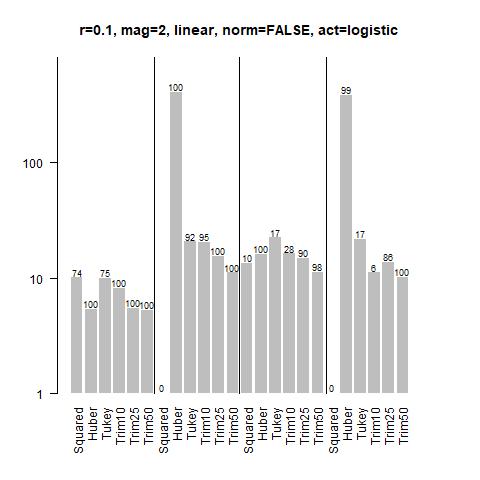} \\
\includegraphics[width=6.75cm,height=6.25cm]{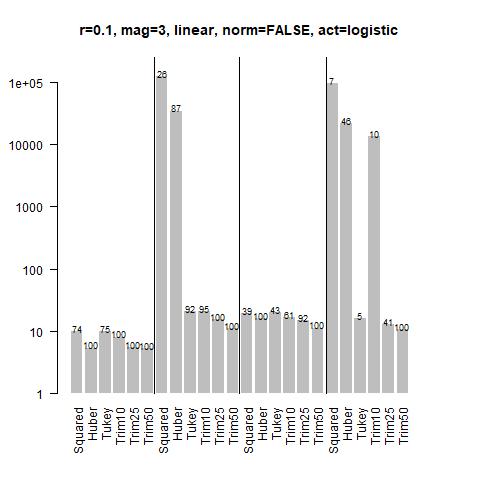} 
\includegraphics[width=6.75cm,height=6.25cm]{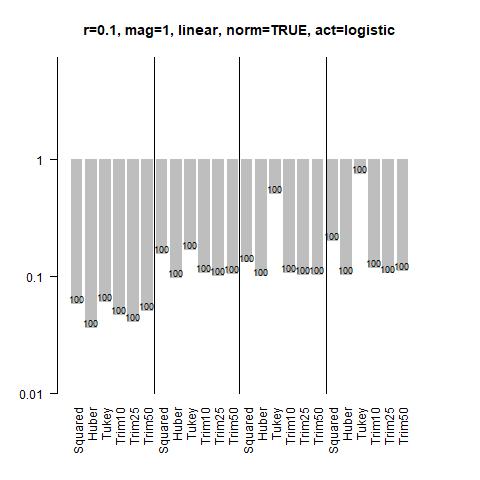}\\
\includegraphics[width=6.75cm,height=6.25cm]{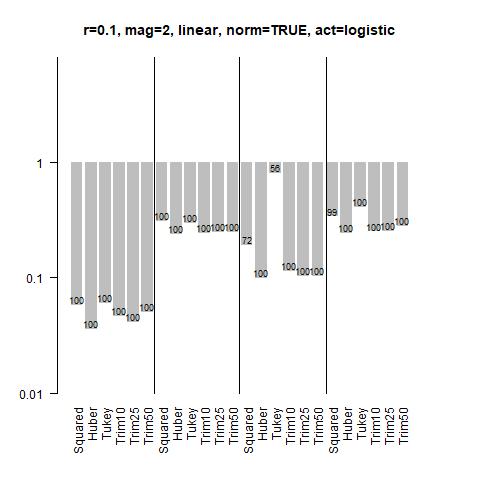} 
\includegraphics[width=6.75cm,height=6.25cm]{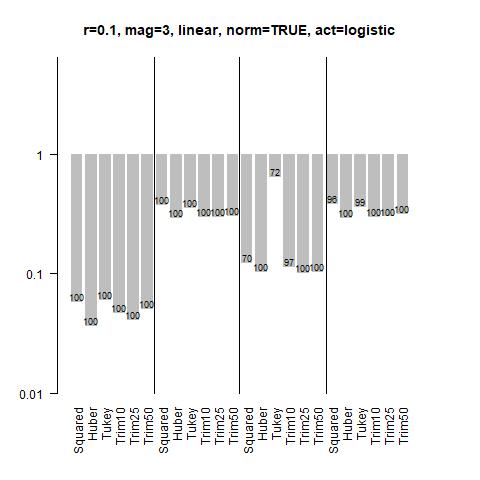} 
\end{center}
\caption{Results for $r=0.1$}
\end{figure}

\begin{figure}[H]
\label{trimnn:n200p5r25m1linnonlog}
\begin{center}
\includegraphics[width=6.75cm,height=6.25cm]{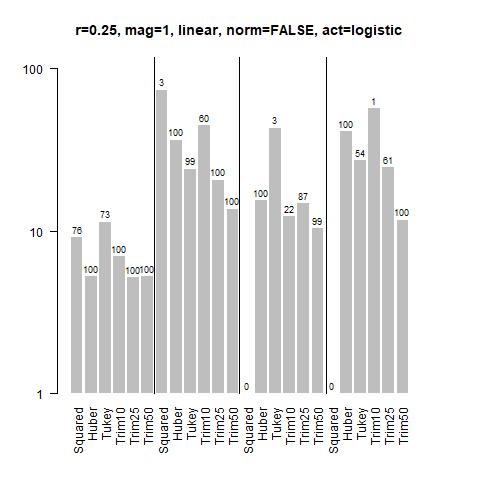}
\includegraphics[width=6.75cm,height=6.25cm]{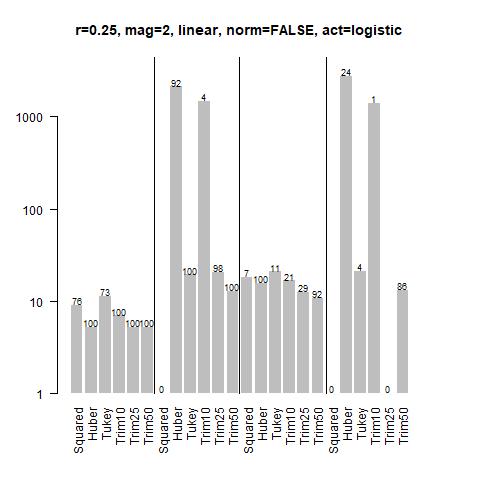} \\
\includegraphics[width=6.75cm,height=6.25cm]{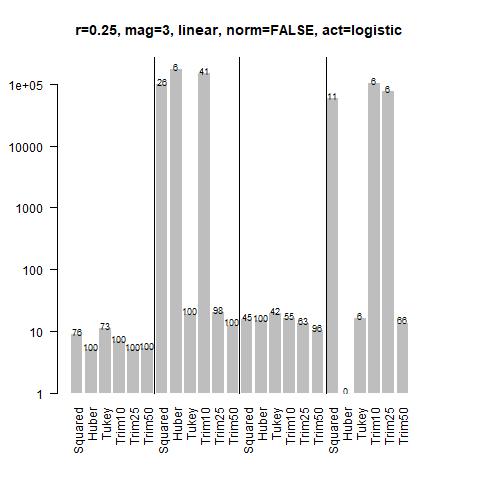} 
\includegraphics[width=6.75cm,height=6.25cm]{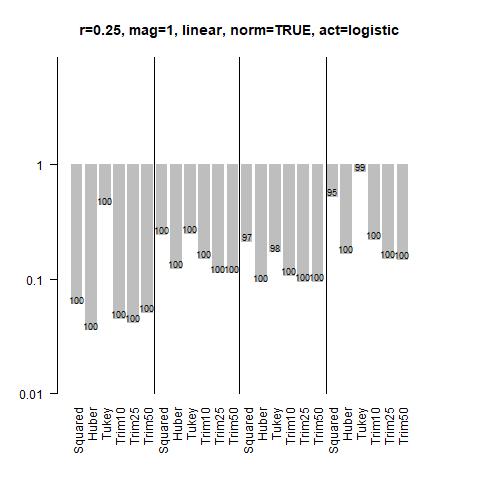}\\
\includegraphics[width=6.75cm,height=6.25cm]{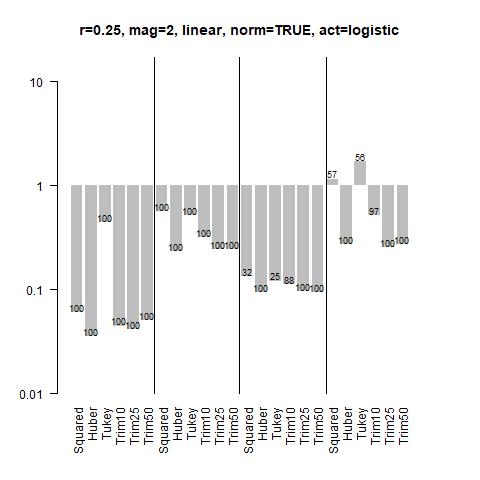} 
\includegraphics[width=6.75cm,height=6.25cm]{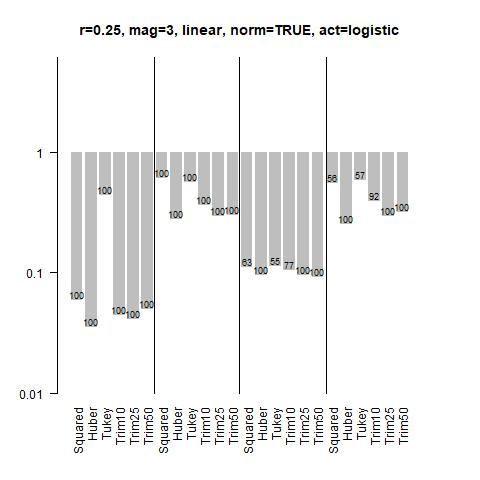} 
\end{center}
\caption{Results for $r=0.25$}
\end{figure}

\begin{figure}[H]
\begin{center}
\includegraphics[width=6.75cm,height=6.25cm]{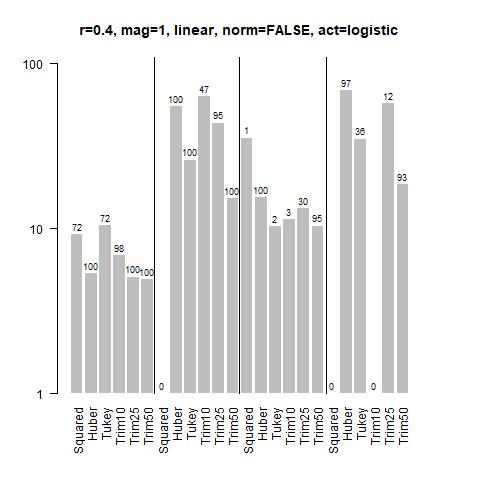}
\includegraphics[width=6.75cm,height=6.25cm]{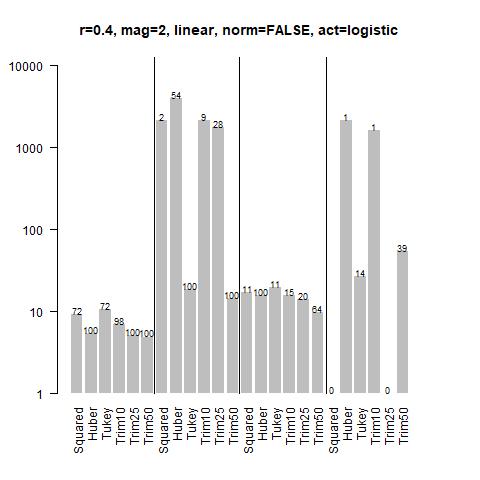} \\
\includegraphics[width=6.75cm,height=6.25cm]{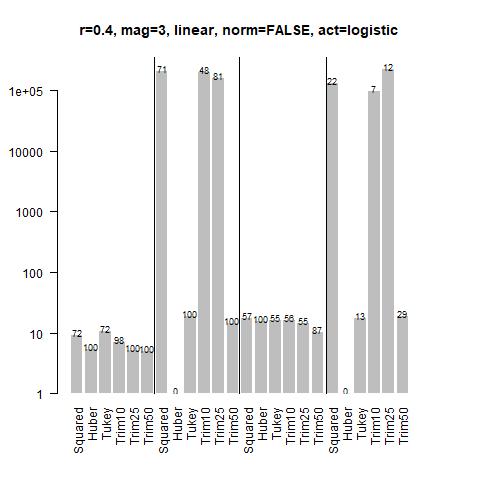} 
\includegraphics[width=6.75cm,height=6.25cm]{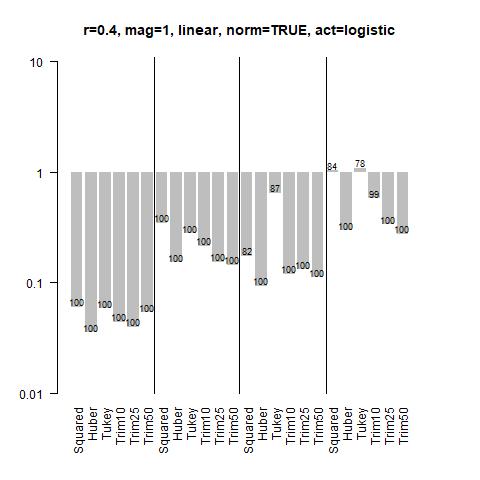}\\
\includegraphics[width=6.75cm,height=6.25cm]{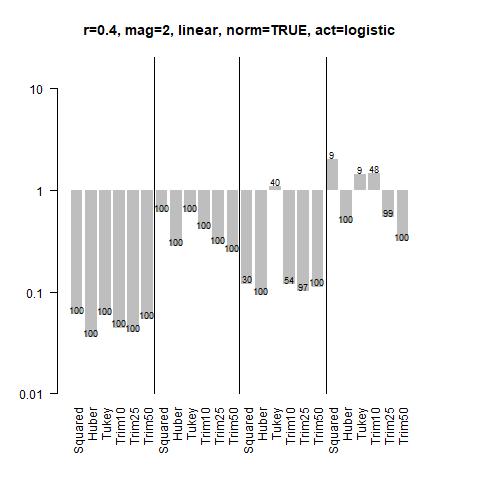} 
\includegraphics[width=6.75cm,height=6.25cm]{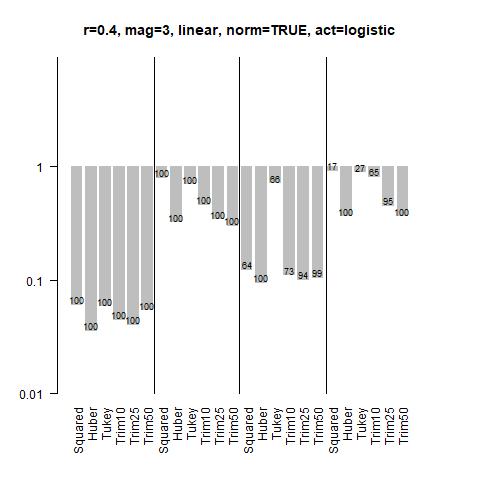} 
\end{center}
\caption{Results for $r=0.4$}\label{trimnn:n200p5r40m1linnonlog}
\end{figure}

\subsubsection{Polynomial function}

\begin{figure}[H]
\label{trimnn:n200p5r10m1polynonlog}
\begin{center}
\includegraphics[width=6.75cm,height=6.25cm]{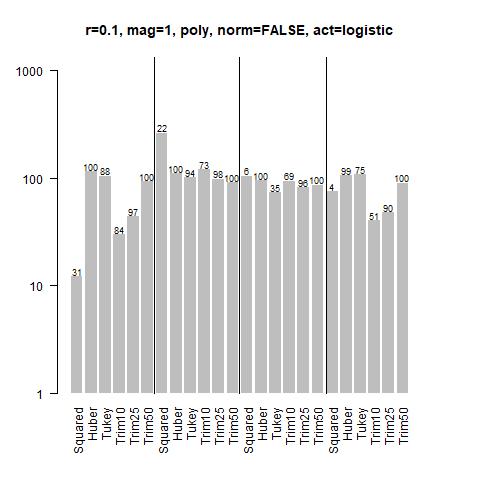}
\includegraphics[width=6.75cm,height=6.25cm]{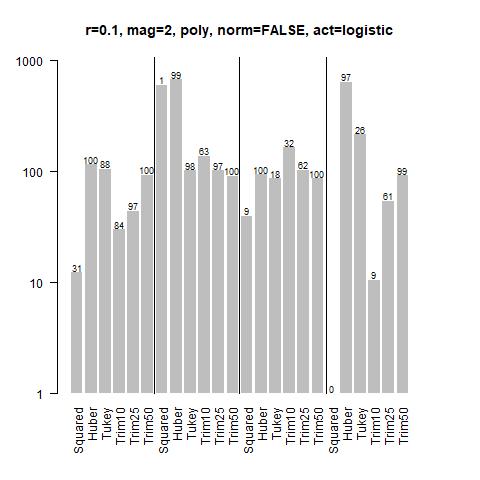} \\
\includegraphics[width=6.75cm,height=6.25cm]{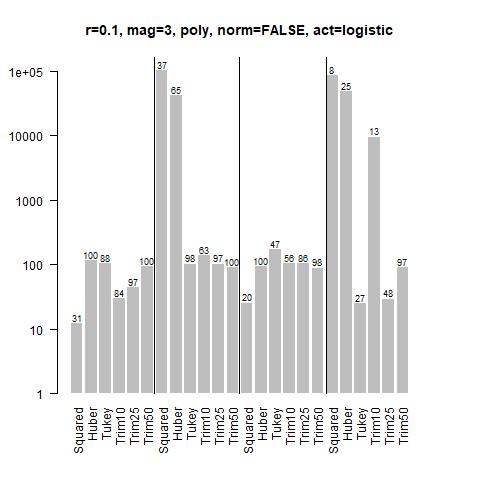} 
\includegraphics[width=6.75cm,height=6.25cm]{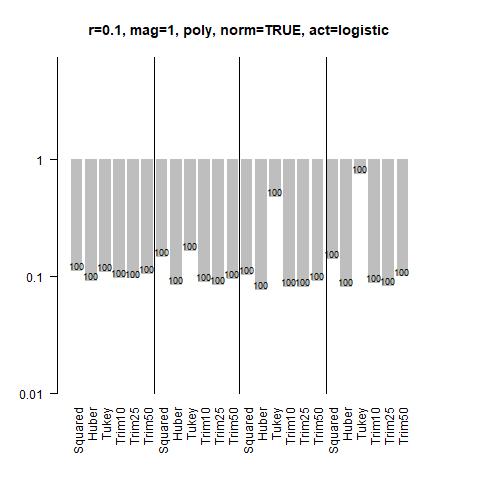}\\
\includegraphics[width=6.75cm,height=6.25cm]{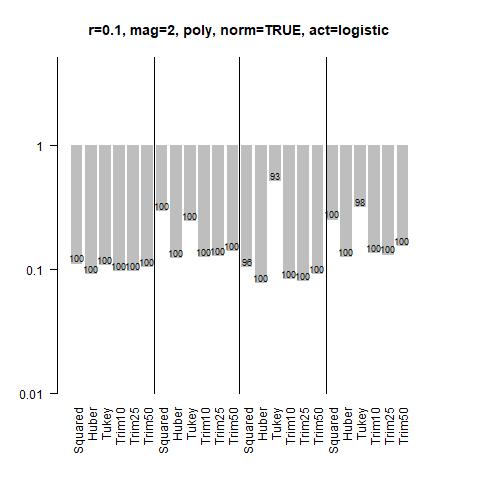} 
\includegraphics[width=6.75cm,height=6.25cm]{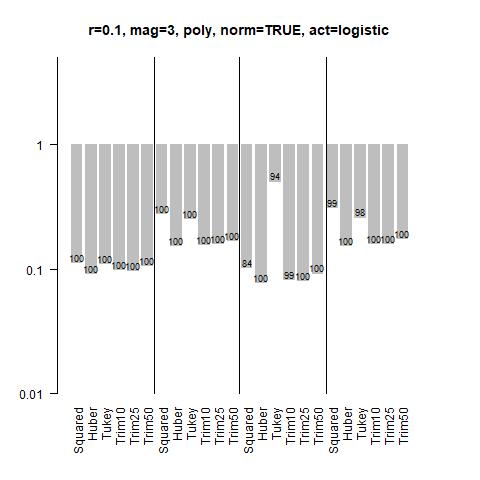} 
\end{center}
\caption{Results for $r=0.1$}
\end{figure}

\begin{figure}[H]
\label{trimnn:n200p5r25m1polynonlog}
\begin{center}
\includegraphics[width=6.75cm,height=6.25cm]{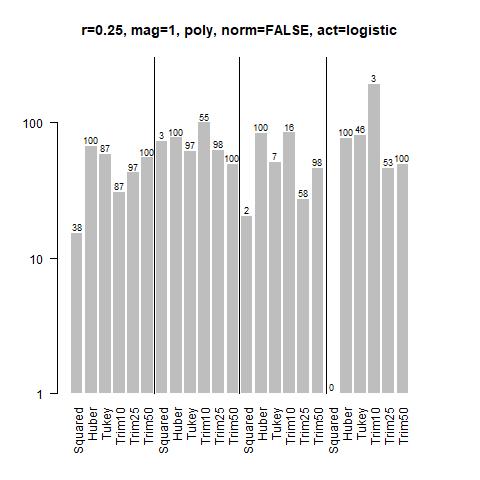}
\includegraphics[width=6.75cm,height=6.25cm]{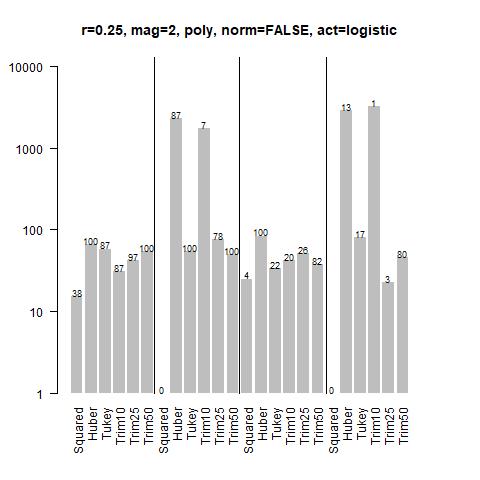} \\
\includegraphics[width=6.75cm,height=6.25cm]{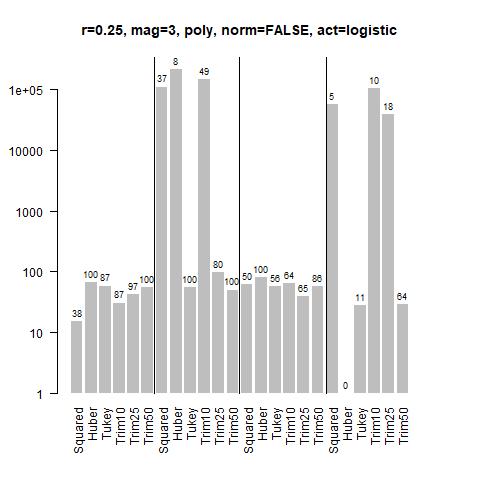} 
\includegraphics[width=6.75cm,height=6.25cm]{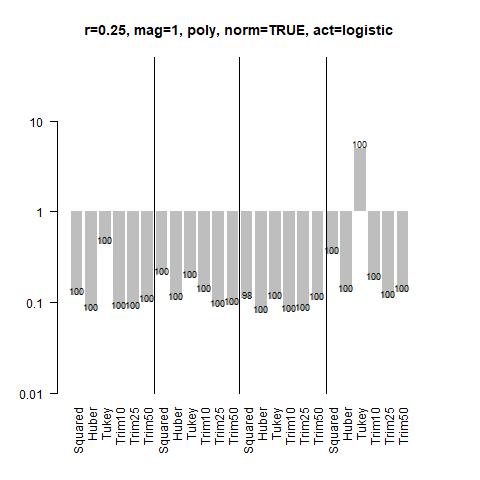}\\
\includegraphics[width=6.75cm,height=6.25cm]{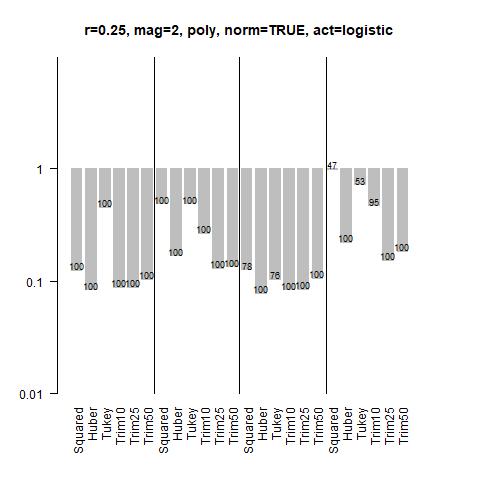} 
\includegraphics[width=6.75cm,height=6.25cm]{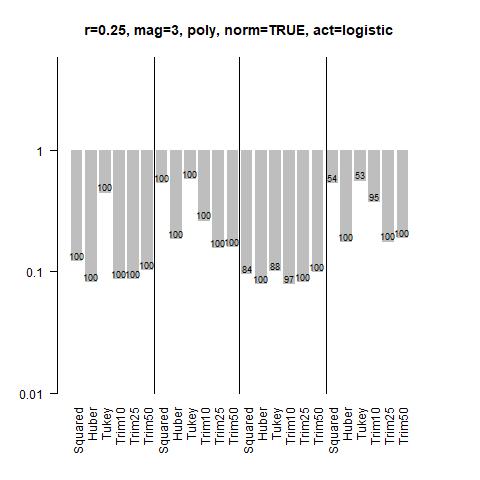} 
\end{center}
\caption{Results for $r=0.25$}
\end{figure}

\begin{figure}[H]
\begin{center}
\includegraphics[width=6.75cm,height=6.25cm]{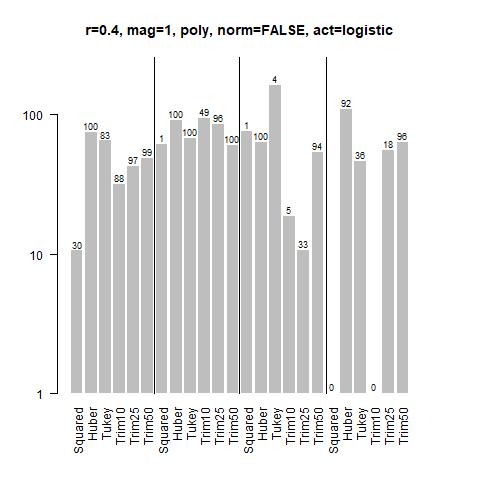}
\includegraphics[width=6.75cm,height=6.25cm]{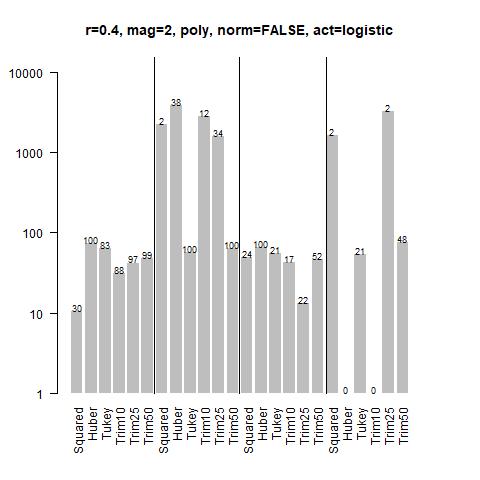} \\
\includegraphics[width=6.75cm,height=6.25cm]{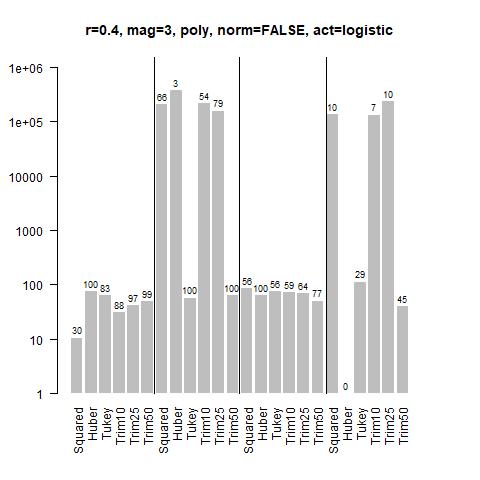} 
\includegraphics[width=6.75cm,height=6.25cm]{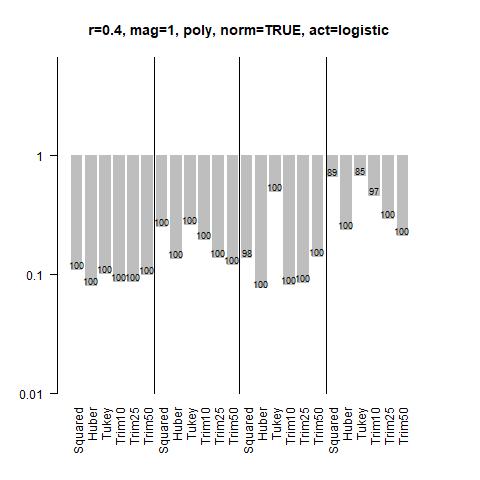}\\
\includegraphics[width=6.75cm,height=6.25cm]{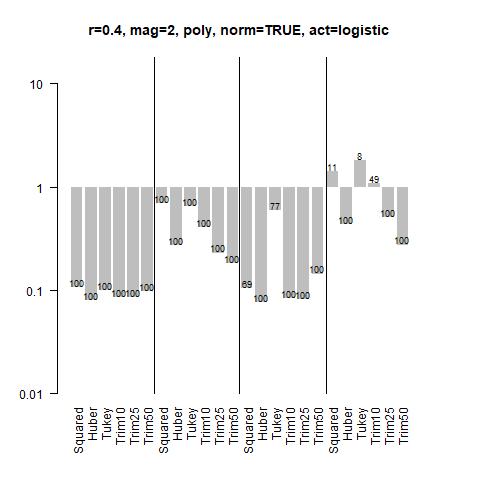} 
\includegraphics[width=6.75cm,height=6.25cm]{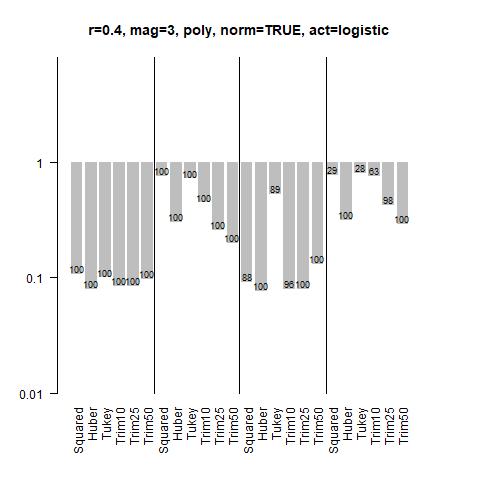} 
\end{center}
\caption{Results for $r=0.4$}\label{trimnn:n200p5r40m1polynonlog}
\end{figure}

\subsubsection{Trigonometric function}

\begin{figure}[H]
\begin{center}
\includegraphics[width=6.75cm,height=6.25cm]{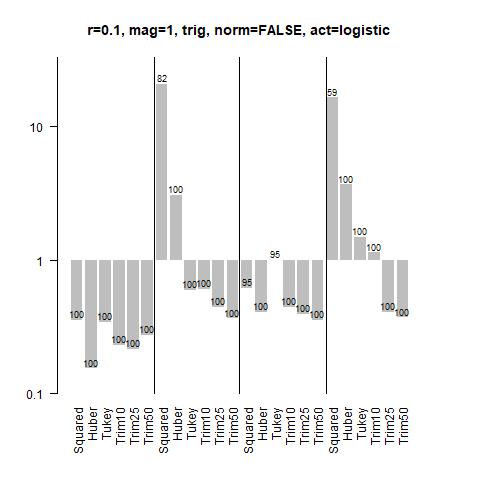}
\includegraphics[width=6.75cm,height=6.25cm]{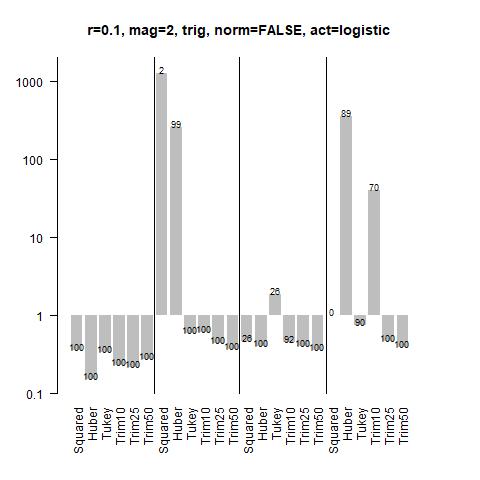} \\
\includegraphics[width=6.75cm,height=6.25cm]{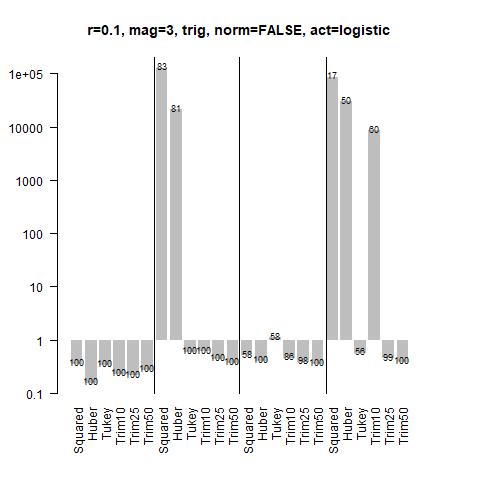} 
\includegraphics[width=6.75cm,height=6.25cm]{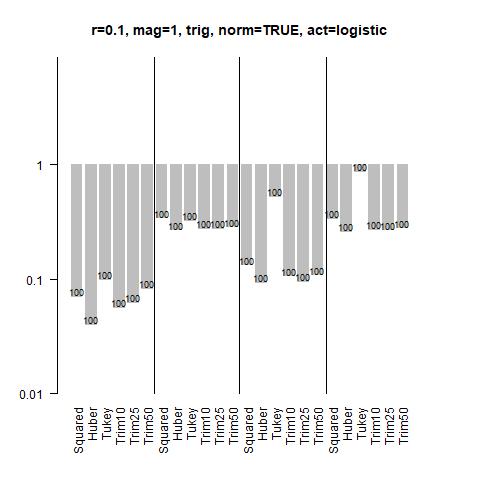}\\
\includegraphics[width=6.75cm,height=6.25cm]{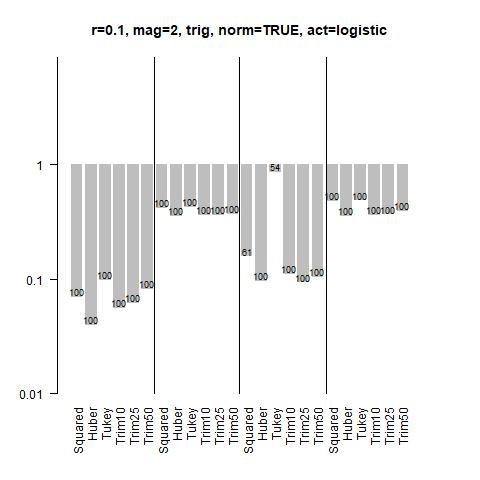} 
\includegraphics[width=6.75cm,height=6.25cm]{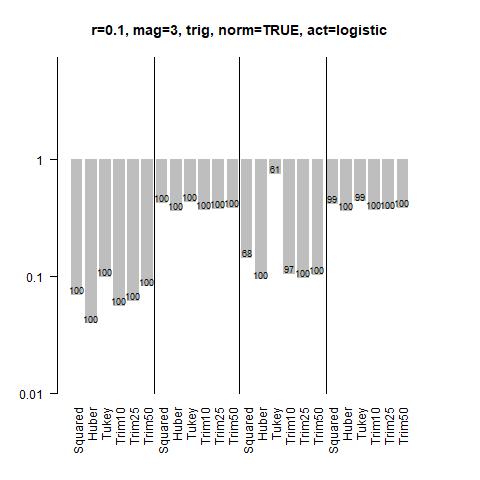} 
\end{center}
\caption{Results for $r=0.1$}\label{trimnn:n200p5r10m1trignonlog}
\end{figure}

\begin{figure}[H]
\begin{center}
\includegraphics[width=6.75cm,height=6.25cm]{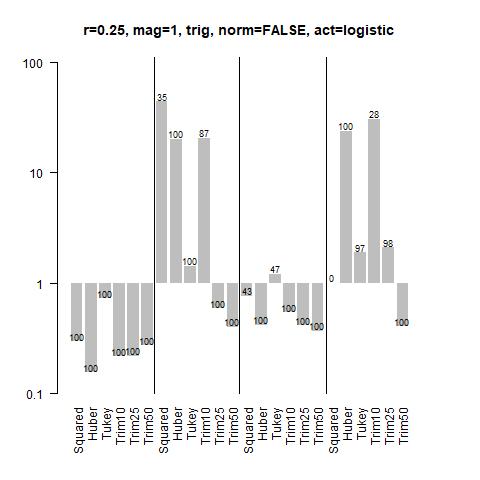}
\includegraphics[width=6.75cm,height=6.25cm]{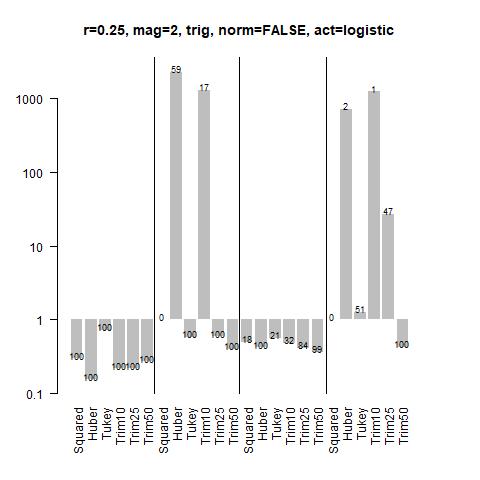} \\
\includegraphics[width=6.75cm,height=6.25cm]{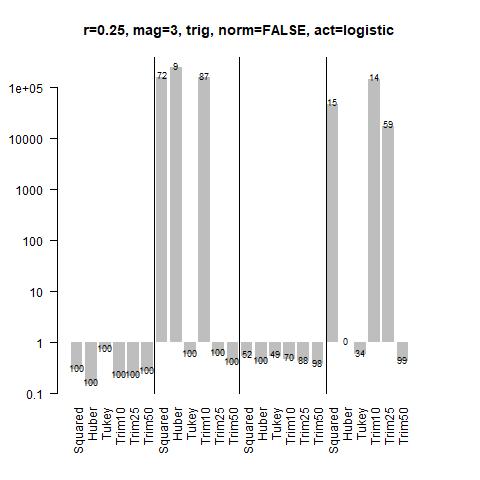} 
\includegraphics[width=6.75cm,height=6.25cm]{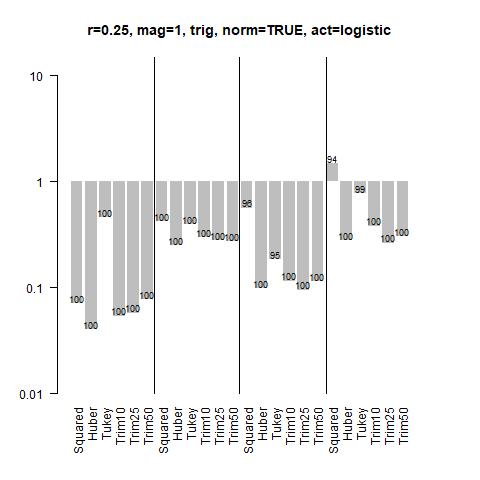}\\
\includegraphics[width=6.75cm,height=6.25cm]{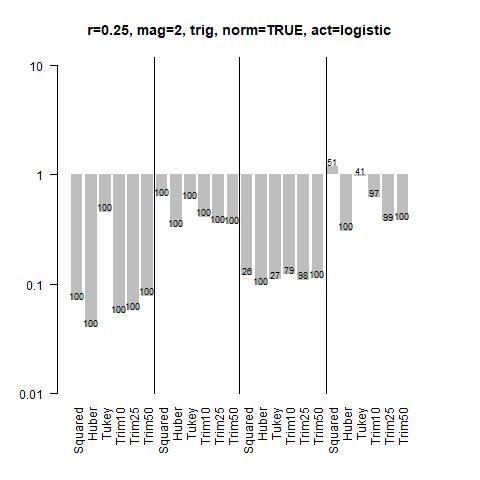} 
\includegraphics[width=6.75cm,height=6.25cm]{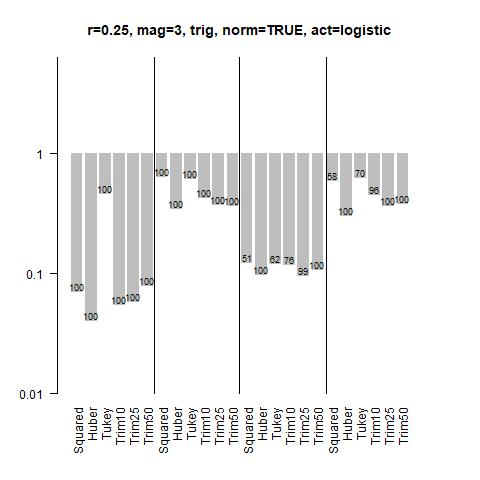} 
\end{center}
\caption{Results for $r=0.25$}\label{trimnn:n200p5r25m1trignonlog}
\end{figure}

\begin{figure}[H]
\begin{center}
\includegraphics[width=6.75cm,height=6.25cm]{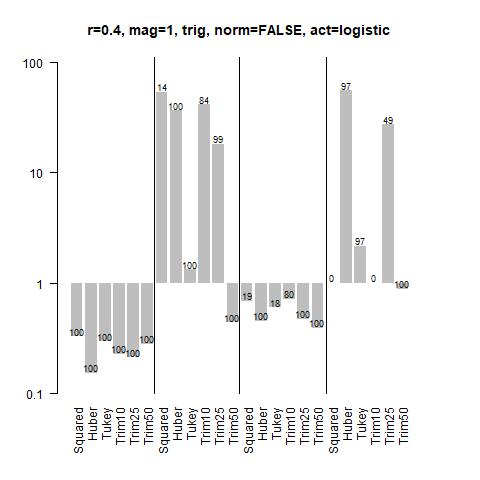}
\includegraphics[width=6.75cm,height=6.25cm]{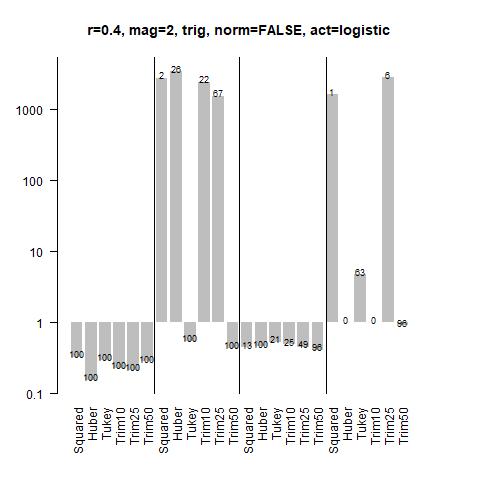} \\
\includegraphics[width=6.75cm,height=6.25cm]{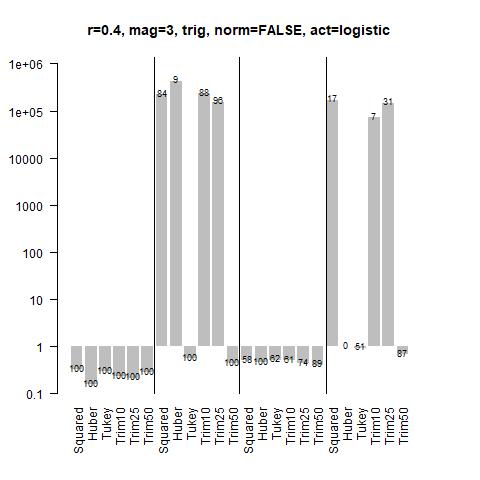} 
\includegraphics[width=6.75cm,height=6.25cm]{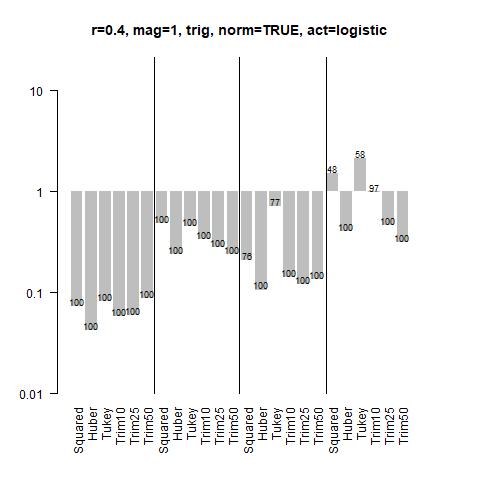}\\
\includegraphics[width=6.75cm,height=6.25cm]{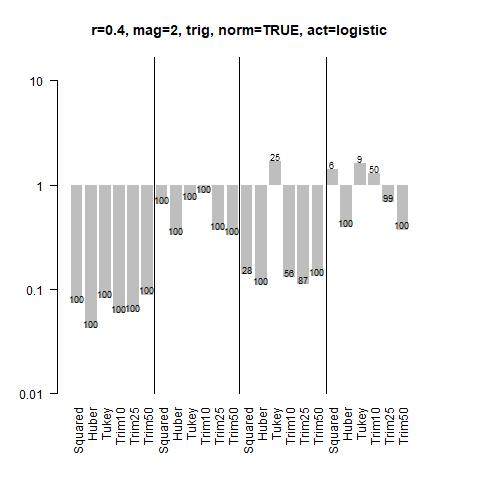} 
\includegraphics[width=6.75cm,height=6.25cm]{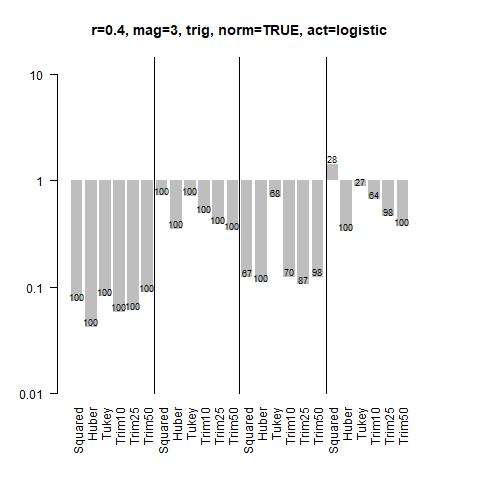} 
\end{center}
\caption{Results for $r=0.4$}\label{trimnn:n200p5r40m1trignonlog}
\end{figure}

\subsection{Softplus activation function}  

\subsubsection{Linear function}

\begin{figure}[H]
\begin{center}
\includegraphics[width=6.75cm,height=6.25cm]{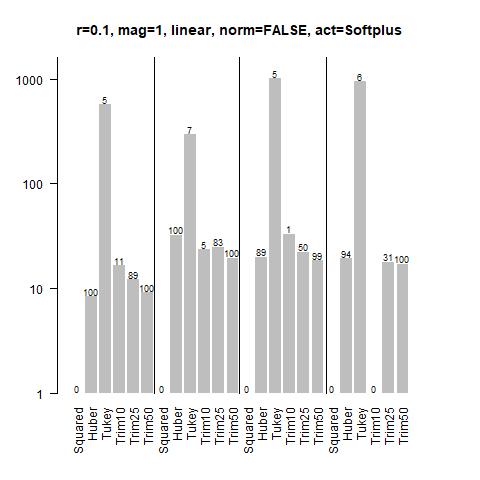}
\includegraphics[width=6.75cm,height=6.25cm]{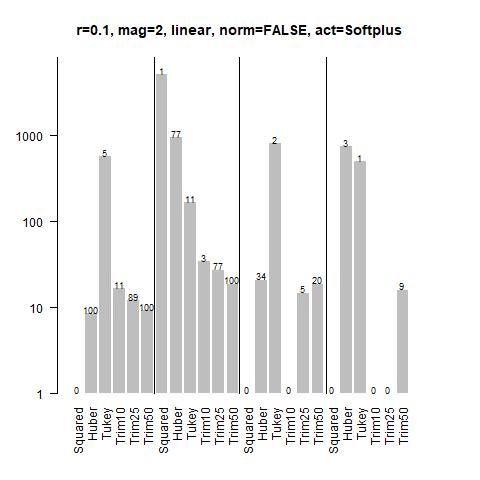} \\
\includegraphics[width=6.75cm,height=6.25cm]{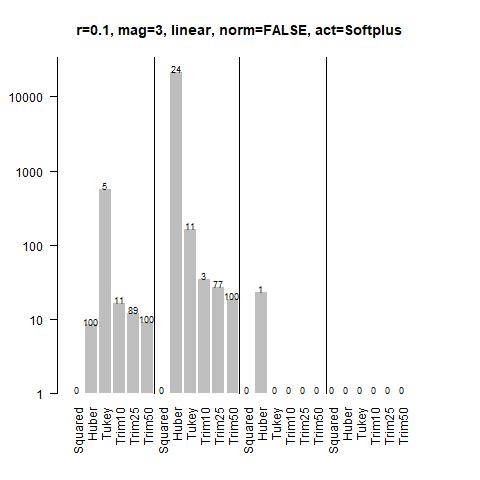} 
\includegraphics[width=6.75cm,height=6.25cm]{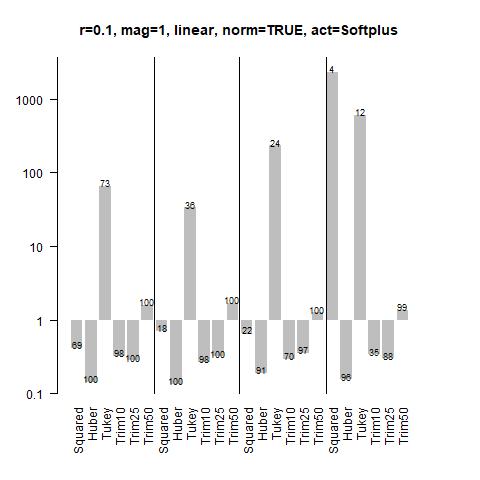}\\
\includegraphics[width=6.75cm,height=6.25cm]{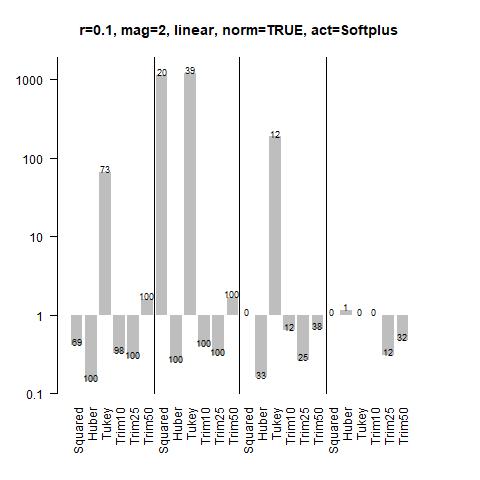} 
\includegraphics[width=6.75cm,height=6.25cm]{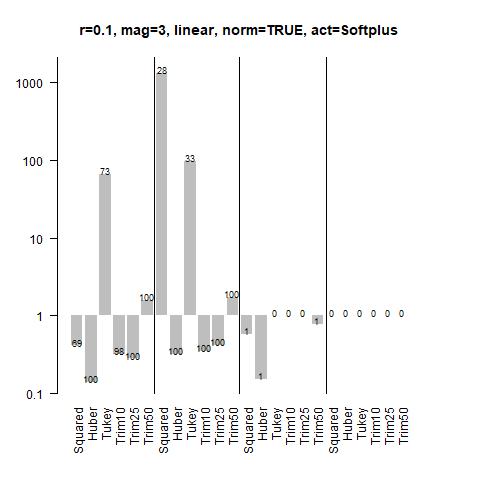} 
\end{center}
\caption{Results for $r=0.1$}\label{trimnn:n200p5r10m1linnonrelu}
\end{figure}

\begin{figure}[H]
\begin{center}
\includegraphics[width=6.75cm,height=6.25cm]{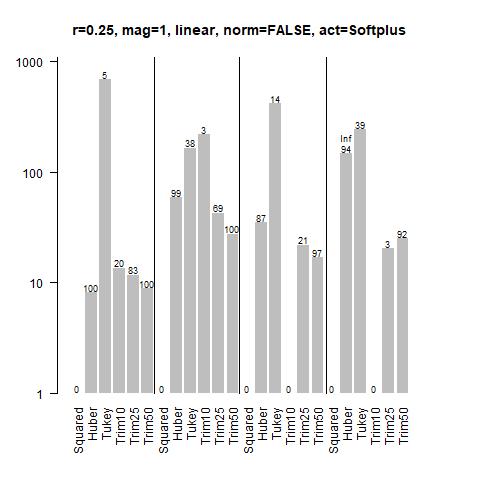}
\includegraphics[width=6.75cm,height=6.25cm]{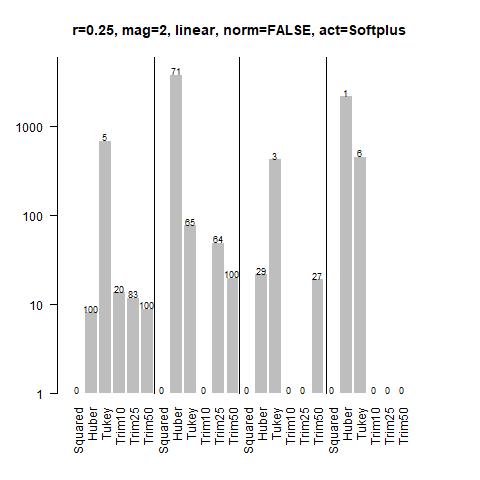} \\
\includegraphics[width=6.75cm,height=6.25cm]{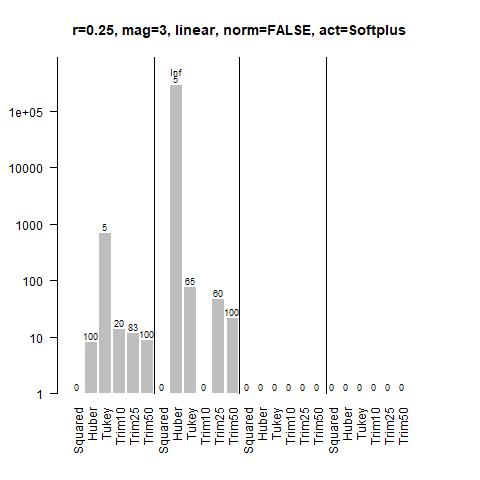} 
\includegraphics[width=6.75cm,height=6.25cm]{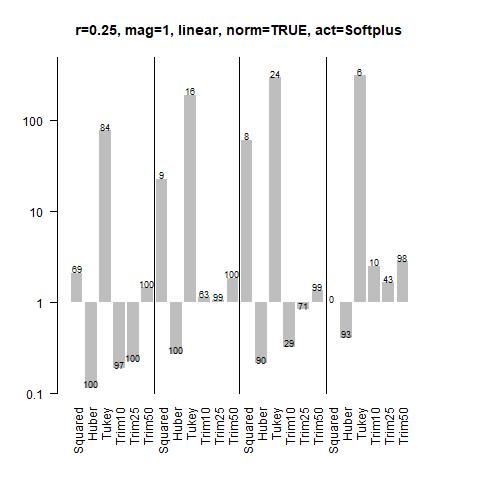}\\
\includegraphics[width=6.75cm,height=6.25cm]{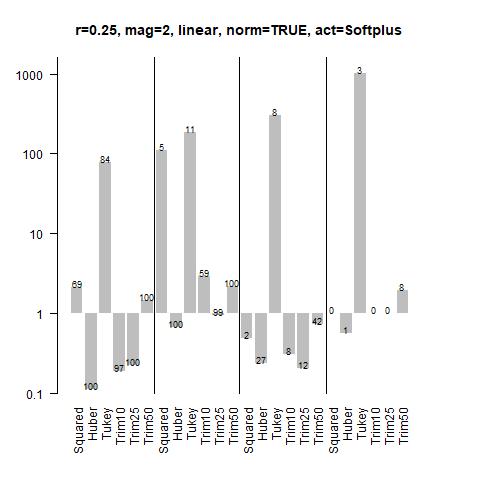} 
\includegraphics[width=6.75cm,height=6.25cm]{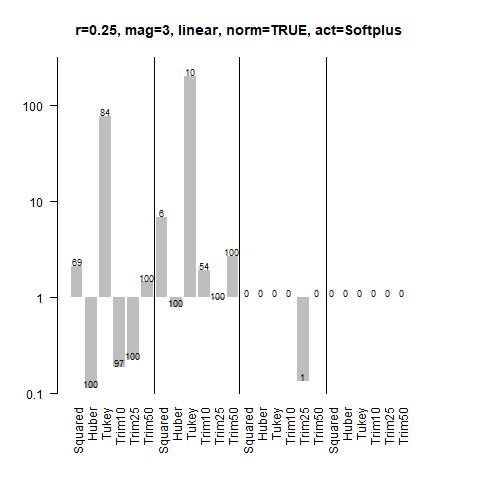} 
\end{center}
\caption{Results for $r=0.25$}\label{trimnn:n200p5r25m1linnonrelu}
\end{figure}

\begin{figure}[H]
\begin{center}
\includegraphics[width=6.75cm,height=6.25cm]{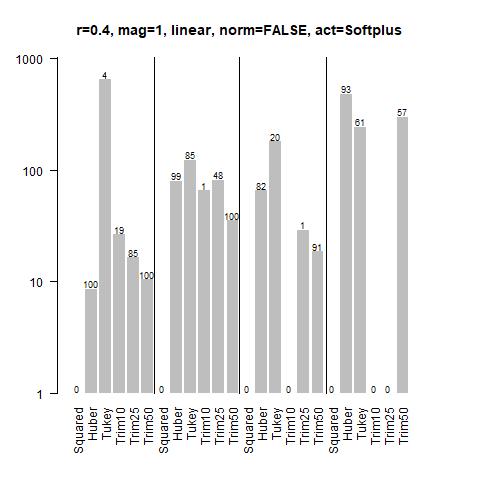}
\includegraphics[width=6.75cm,height=6.25cm]{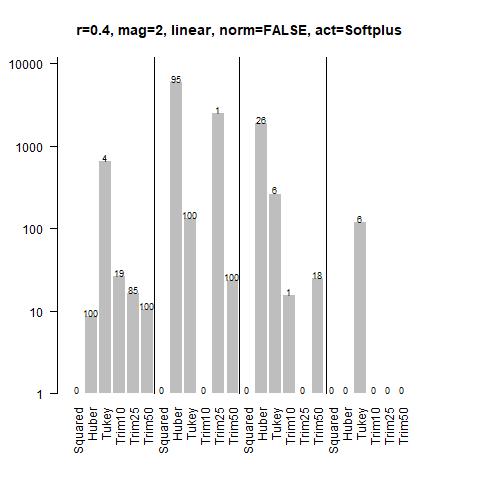} \\
\includegraphics[width=6.75cm,height=6.25cm]{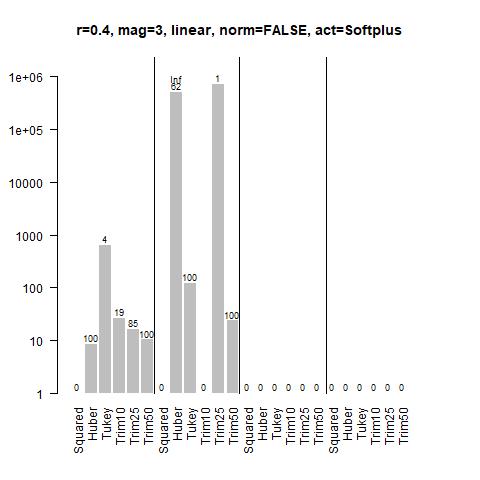} 
\includegraphics[width=6.75cm,height=6.25cm]{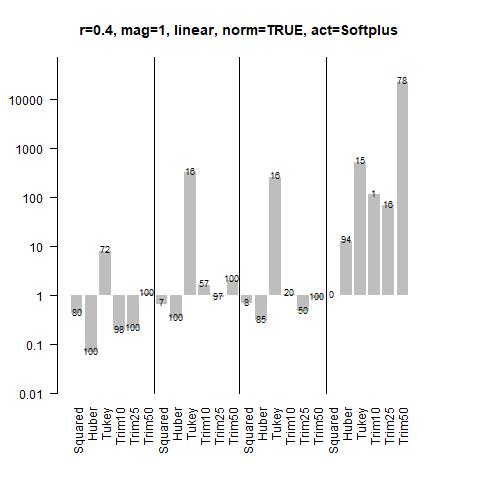}\\
\includegraphics[width=6.75cm,height=6.25cm]{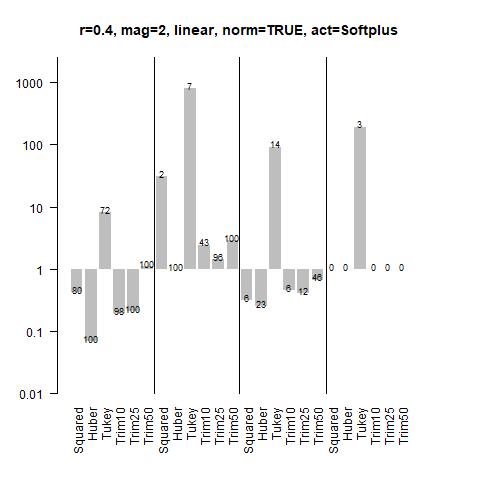} 
\includegraphics[width=6.75cm,height=6.25cm]{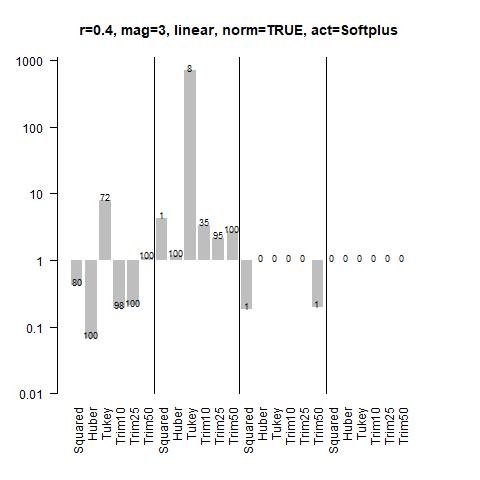} 
\end{center}
\caption{Results for $r=0.4$}\label{trimnn:n200p5r40m1linnonrelu}
\end{figure}

\subsubsection{Polynomial function}

\begin{figure}[H]
\label{trimnn:n200p5r10m1polynonrelu}
\begin{center}
\includegraphics[width=6.75cm,height=6.25cm]{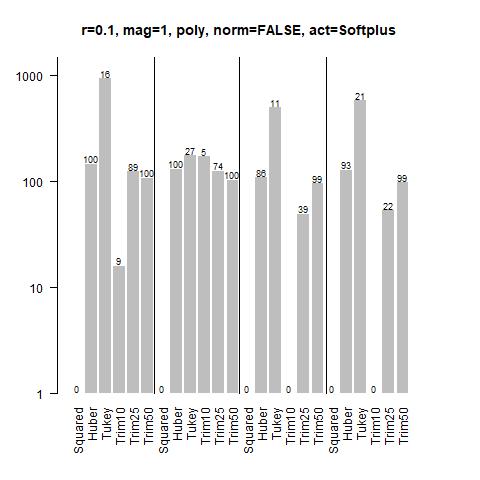}
\includegraphics[width=6.75cm,height=6.25cm]{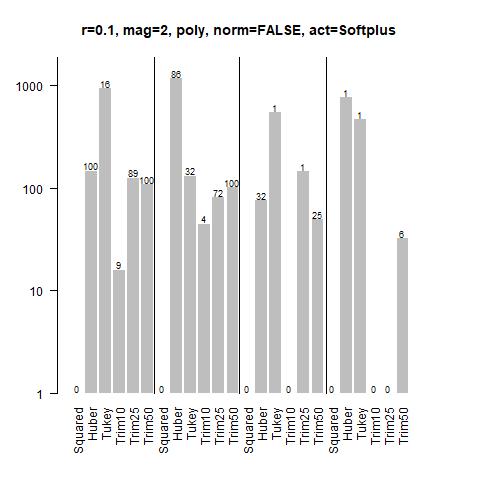} \\
\includegraphics[width=6.75cm,height=6.25cm]{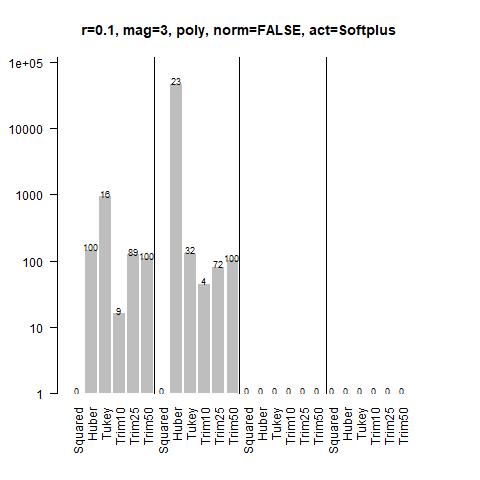} 
\includegraphics[width=6.75cm,height=6.25cm]{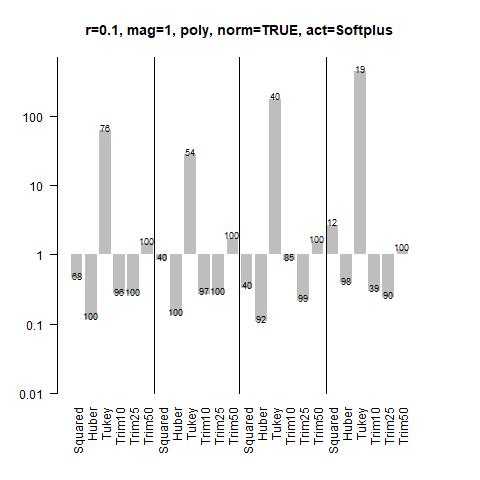}\\
\includegraphics[width=6.75cm,height=6.25cm]{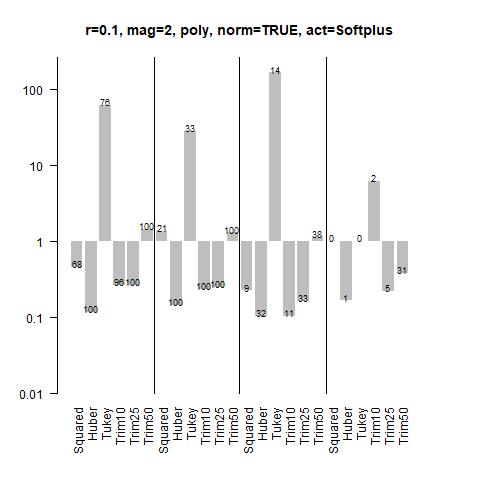} 
\includegraphics[width=6.75cm,height=6.25cm]{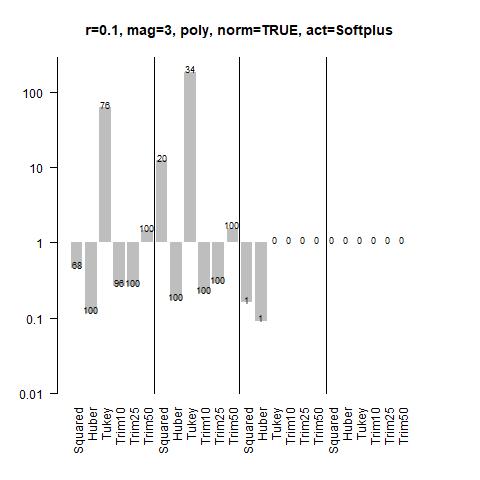} 
\end{center}
\caption{Results for $r=0.1$}
\end{figure}

\begin{figure}[H]
\begin{center}
\includegraphics[width=6.75cm,height=6.25cm]{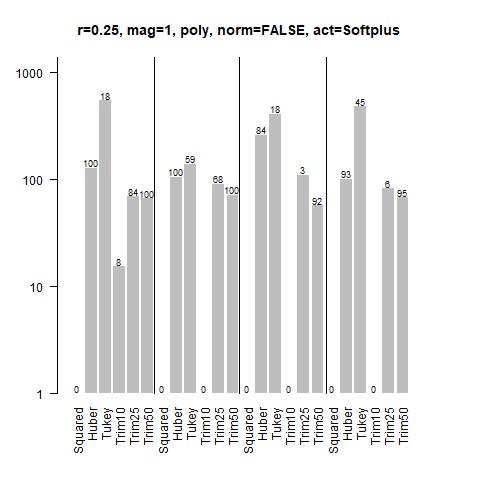}
\includegraphics[width=6.75cm,height=6.25cm]{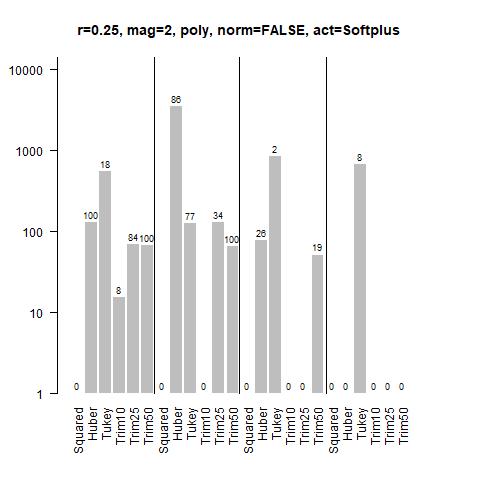} \\
\includegraphics[width=6.75cm,height=6.25cm]{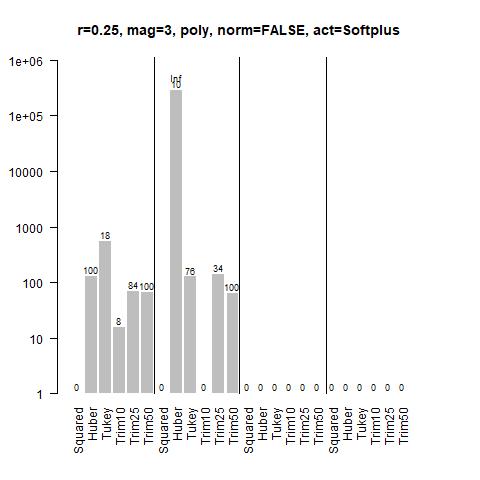} 
\includegraphics[width=6.75cm,height=6.25cm]{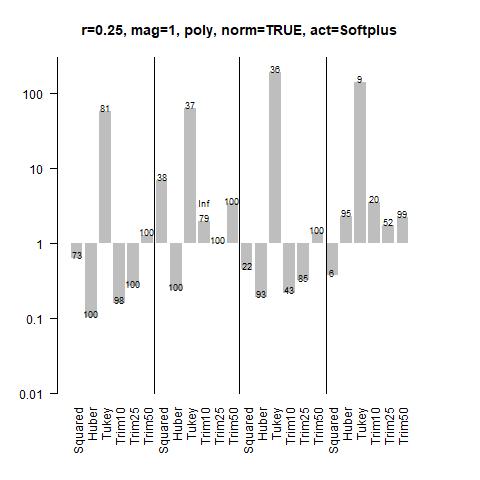}\\
\includegraphics[width=6.75cm,height=6.25cm]{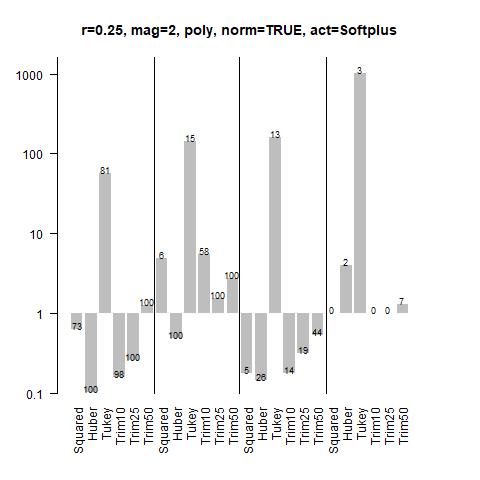} 
\includegraphics[width=6.75cm,height=6.25cm]{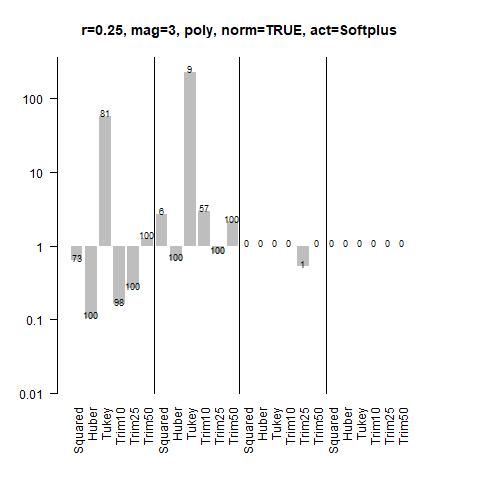} 
\end{center}
\caption{Results for $r=0.25$}\label{trimnn:n200p5r25m1polynonrelu}
\end{figure}

\begin{figure}[H]
\label{trimnn:n200p5r40m1polynonrelu}
\begin{center}
\includegraphics[width=6.75cm,height=6.25cm]{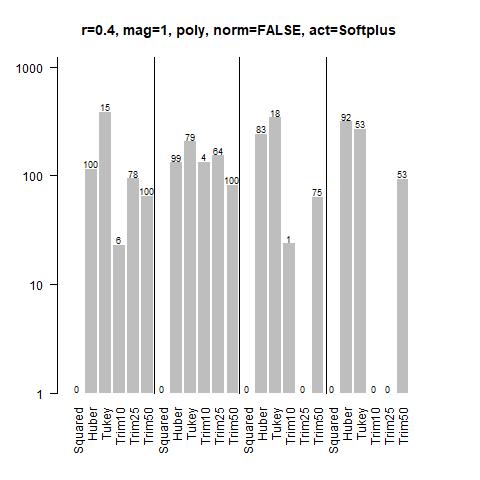}
\includegraphics[width=6.75cm,height=6.25cm]{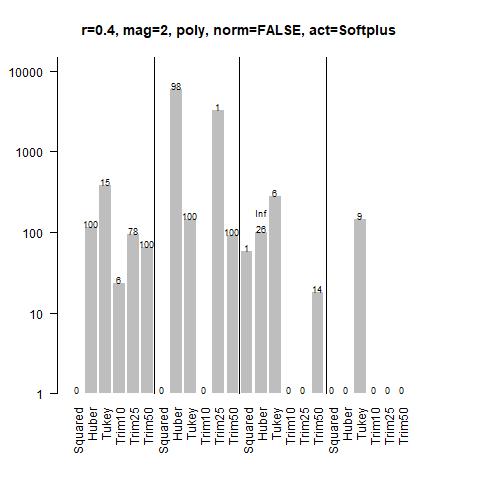} \\
\includegraphics[width=6.75cm,height=6.25cm]{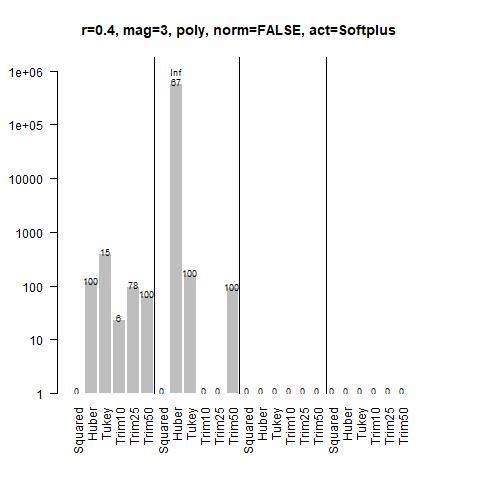} 
\includegraphics[width=6.75cm,height=6.25cm]{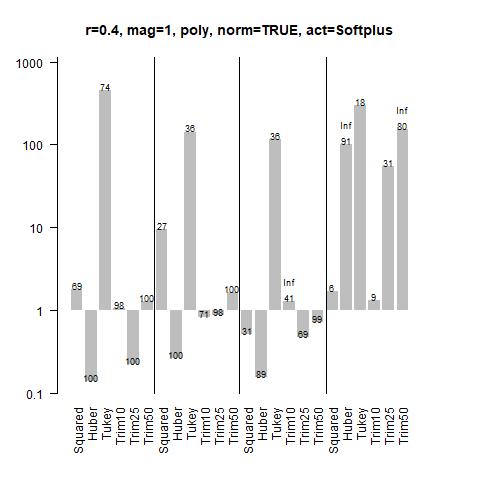}\\
\includegraphics[width=6.75cm,height=6.25cm]{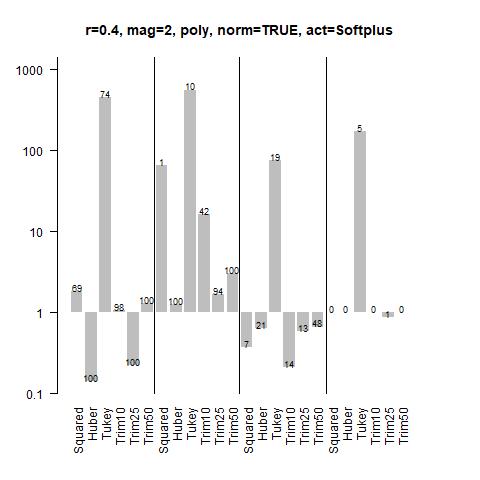} 
\includegraphics[width=6.75cm,height=6.25cm]{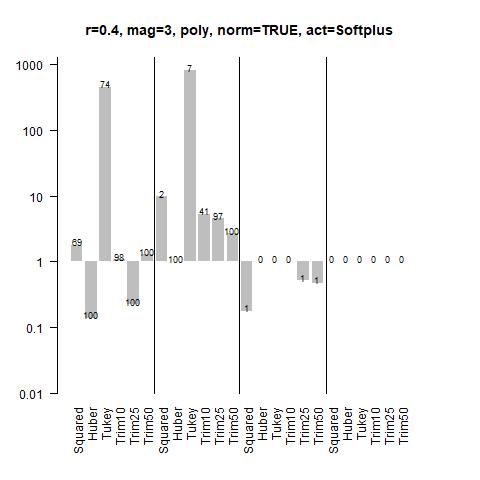} 
\end{center}
\caption{Results for $r=0.4$}
\end{figure}

\subsubsection{Trigonometric function}

\begin{figure}[H]
\label{trimnn:n200p5r10m1trignonrelu}
\begin{center}
\includegraphics[width=6.75cm,height=6.25cm]{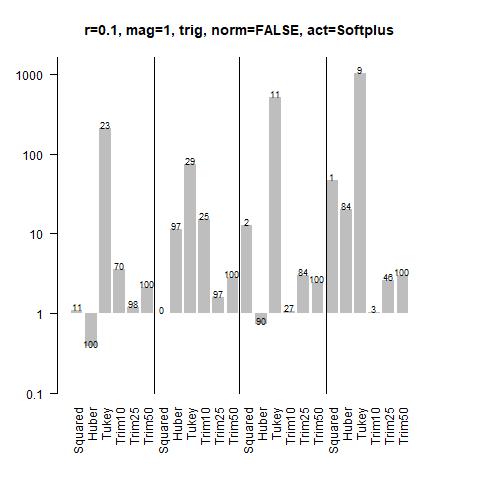}
\includegraphics[width=6.75cm,height=6.25cm]{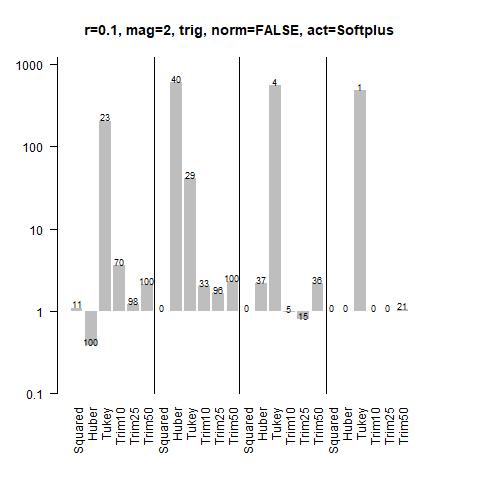} \\
\includegraphics[width=6.75cm,height=6.25cm]{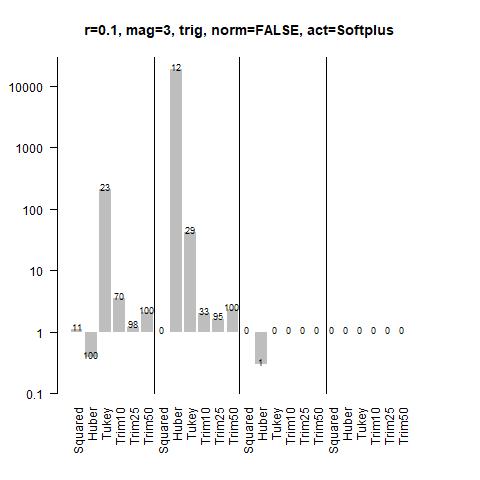} 
\includegraphics[width=6.75cm,height=6.25cm]{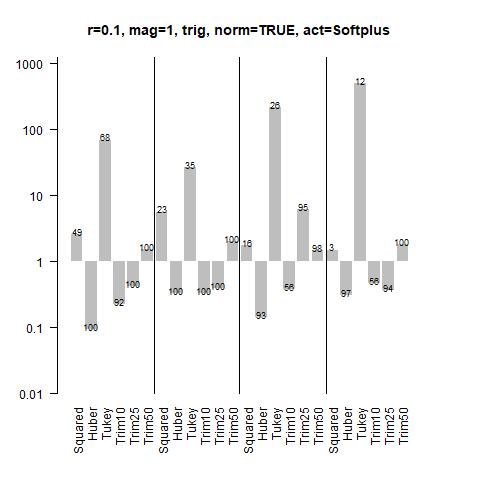}\\
\includegraphics[width=6.75cm,height=6.25cm]{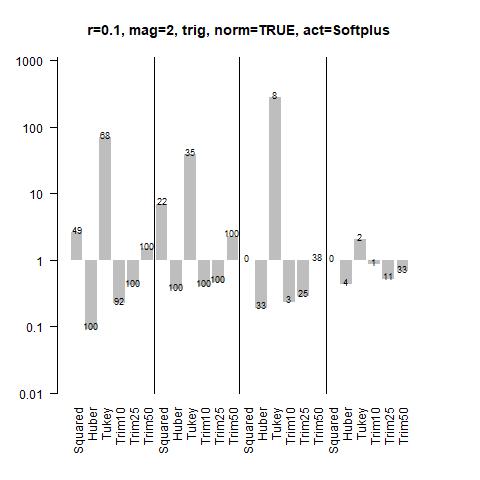} 
\includegraphics[width=6.75cm,height=6.25cm]{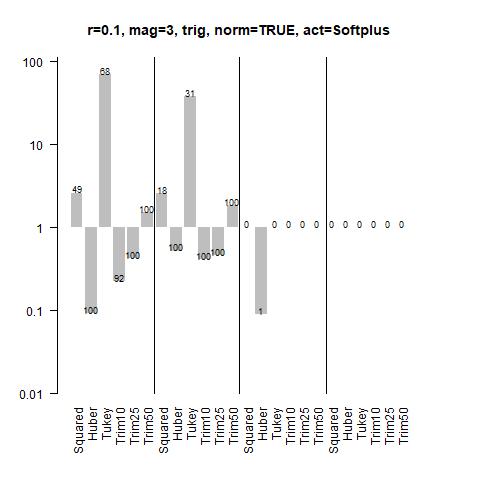} 
\end{center}
\caption{Results for $r=0.1$}
\end{figure}

\begin{figure}[H]
\label{trimnn:n200p5r25m1trignonrelu}
\begin{center}
\includegraphics[width=6.75cm,height=6.25cm]{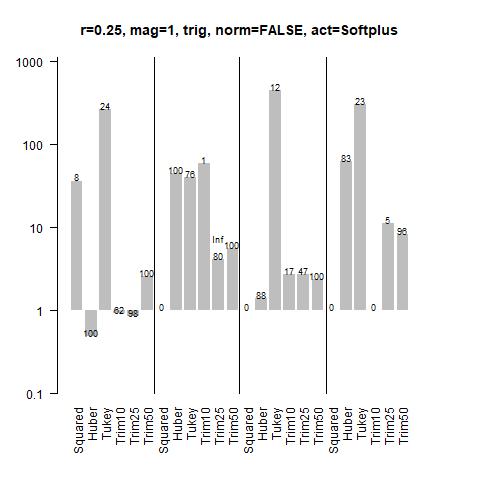}
\includegraphics[width=6.75cm,height=6.25cm]{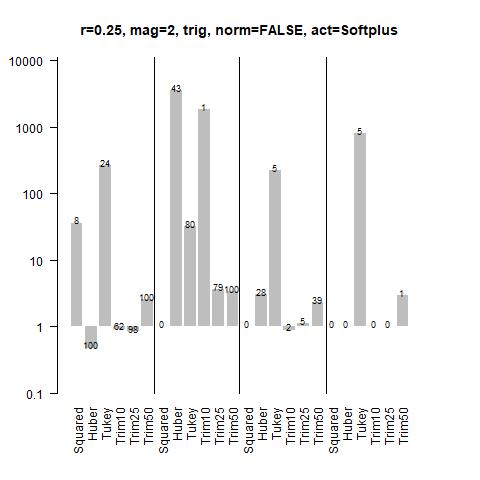} \\
\includegraphics[width=6.75cm,height=6.25cm]{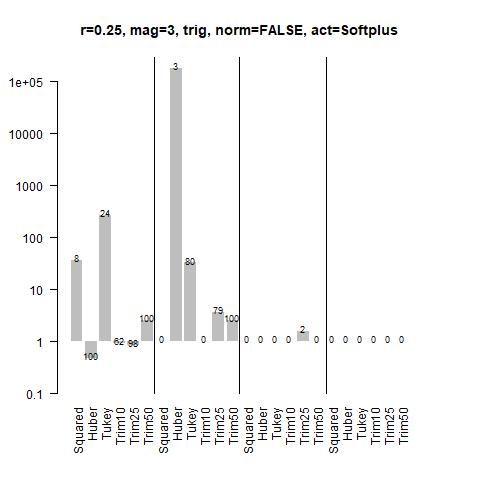} 
\includegraphics[width=6.75cm,height=6.25cm]{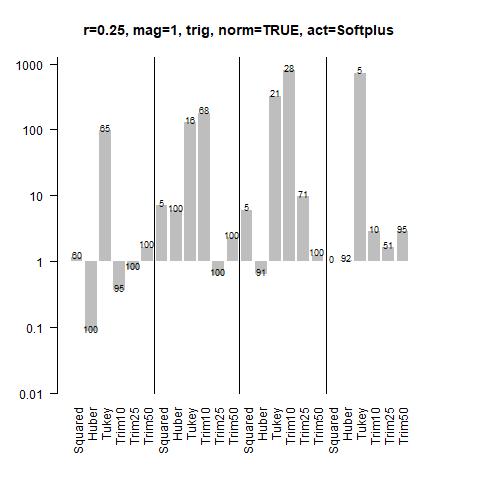}\\
\includegraphics[width=6.75cm,height=6.25cm]{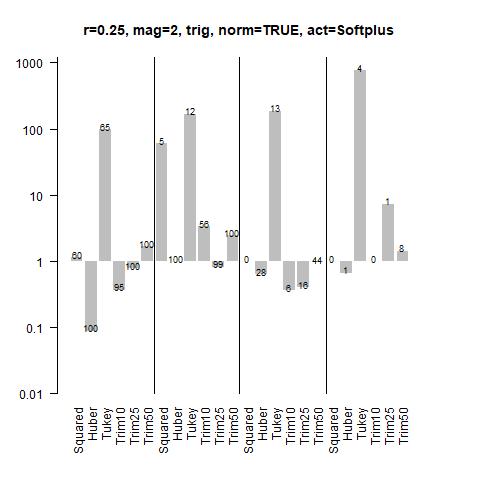} 
\includegraphics[width=6.75cm,height=6.25cm]{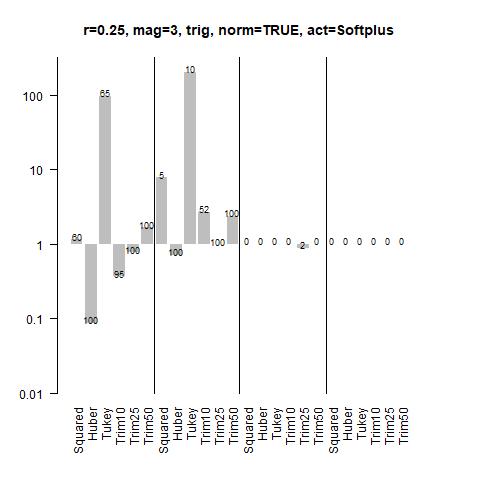} 
\end{center}
\caption{Results for $r=0.25$}
\end{figure}

\begin{figure}[H]
\label{trimnn:n200p5r40m1trignonrelu}
\begin{center}
\includegraphics[width=6.75cm,height=6.25cm]{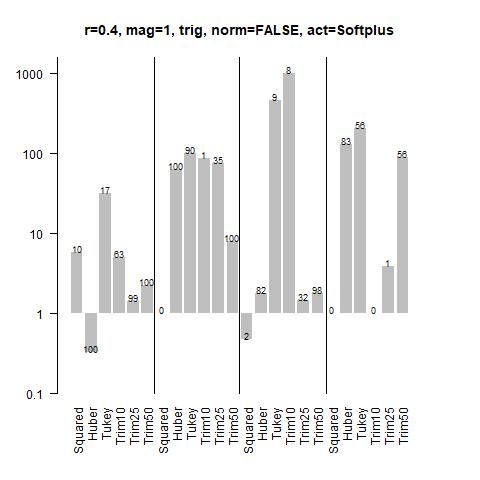}
\includegraphics[width=6.75cm,height=6.25cm]{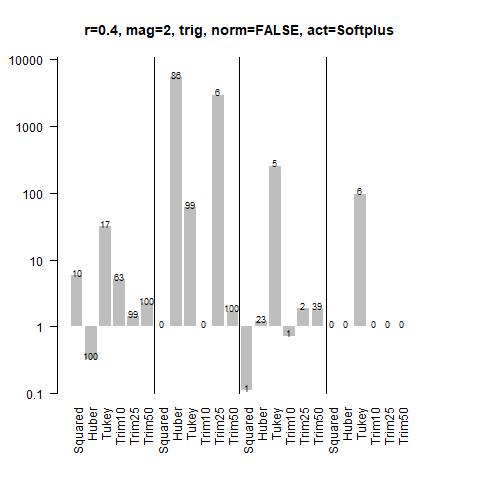} \\
\includegraphics[width=6.75cm,height=6.25cm]{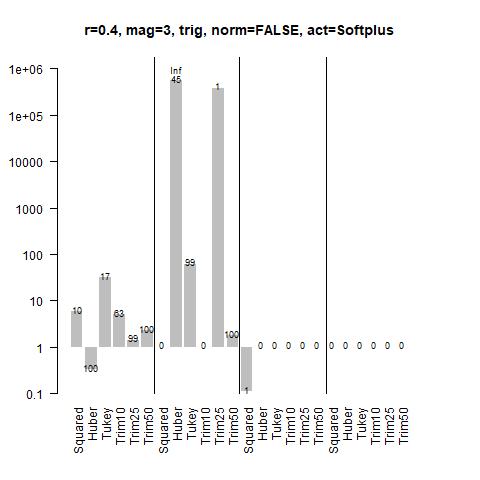} 
\includegraphics[width=6.75cm,height=6.25cm]{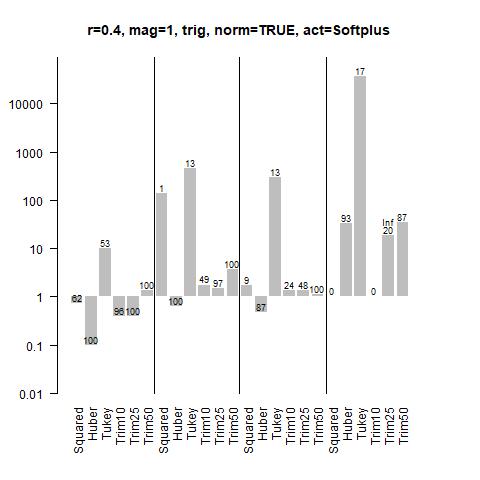}\\
\includegraphics[width=6.75cm,height=6.25cm]{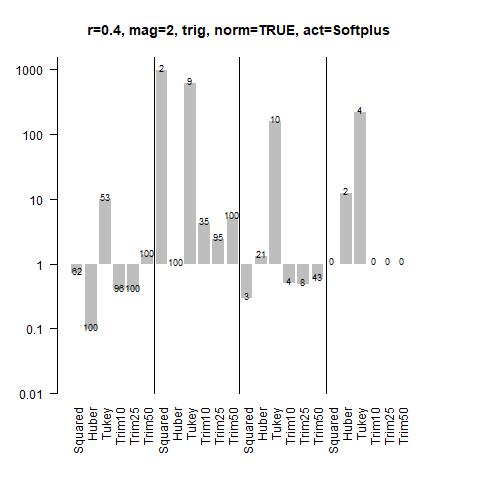} 
\includegraphics[width=6.75cm,height=6.25cm]{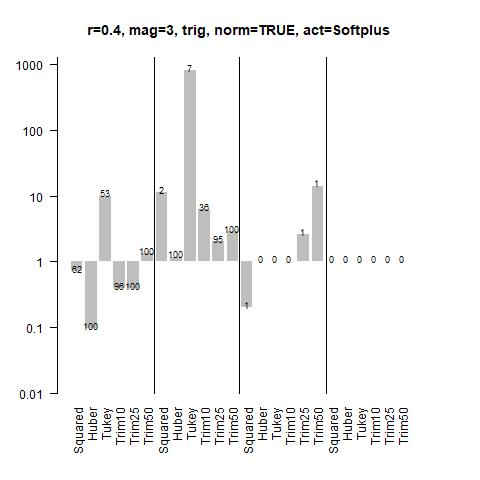} 
\end{center}
\caption{Results for $r=0.4$}
\end{figure}

\section{Simulation results for $n=150$ and $p=5$, deep network: Test loss} \label{trimnn:secloss1505deep}

\subsection{Logistic activation function}

\subsubsection{Linear function}

\begin{figure}[H]
\begin{center}
\includegraphics[width=6.75cm,height=6.25cm]{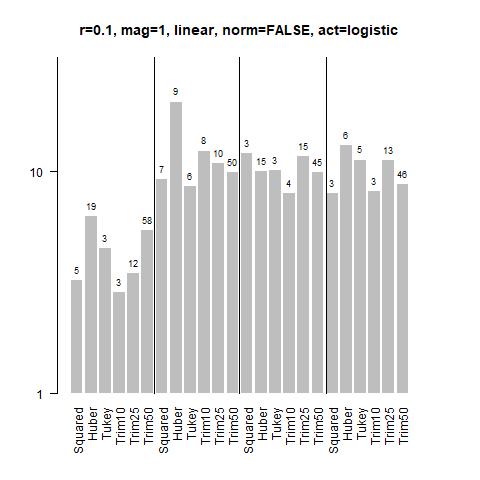}
\includegraphics[width=6.75cm,height=6.25cm]{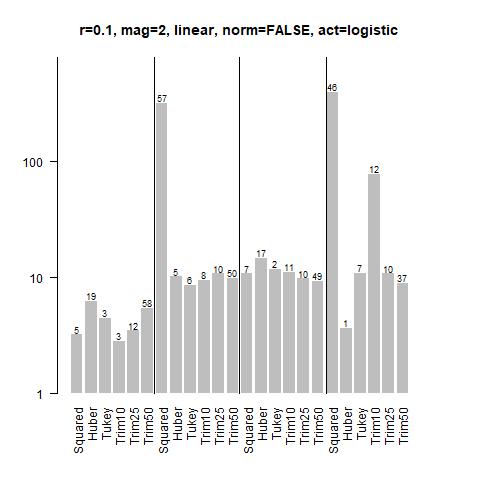} \\
\includegraphics[width=6.75cm,height=6.25cm]{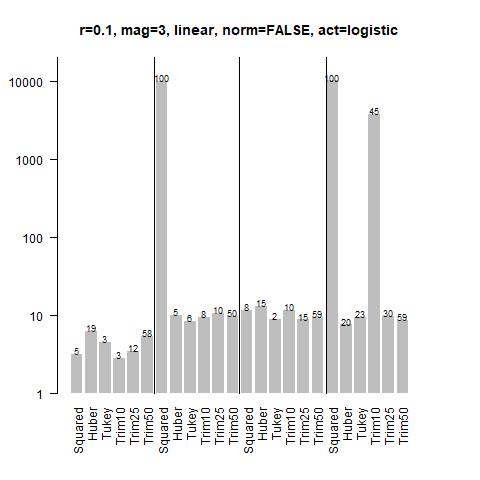} 
\includegraphics[width=6.75cm,height=6.25cm]{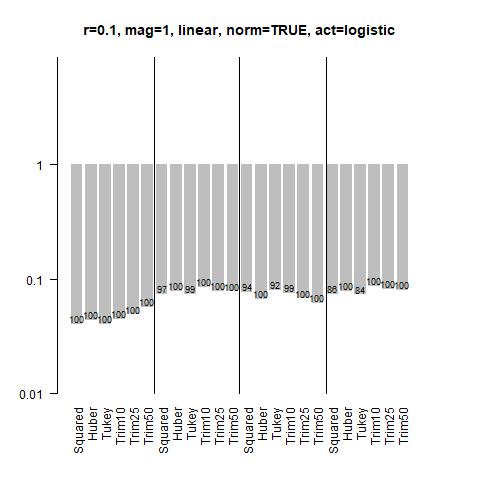}\\
\includegraphics[width=6.75cm,height=6.25cm]{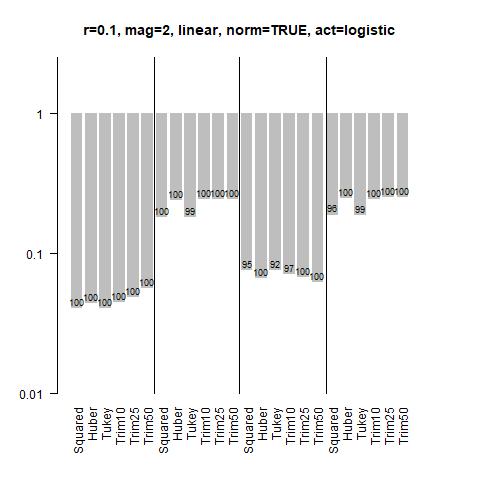} 
\includegraphics[width=6.75cm,height=6.25cm]{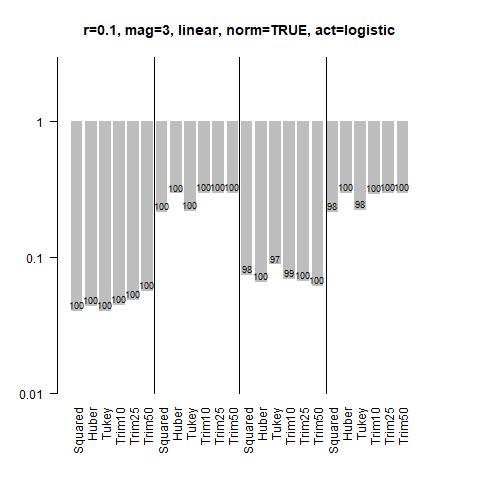} 
\end{center}
\caption{Results for $r=0.1$}\label{trimnn:n200p5r10m1linnonlogdeep}
\end{figure}

\begin{figure}[H]
\begin{center}
\includegraphics[width=6.75cm,height=6.25cm]{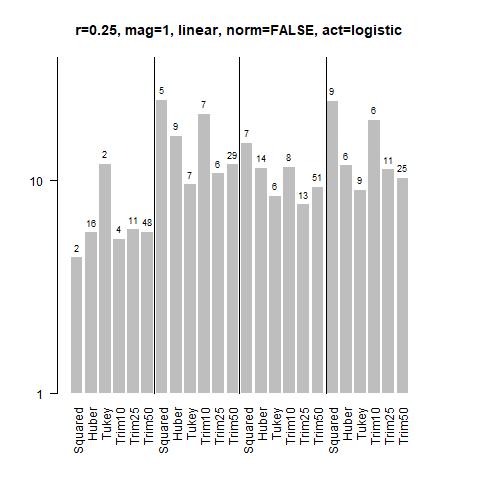}
\includegraphics[width=6.75cm,height=6.25cm]{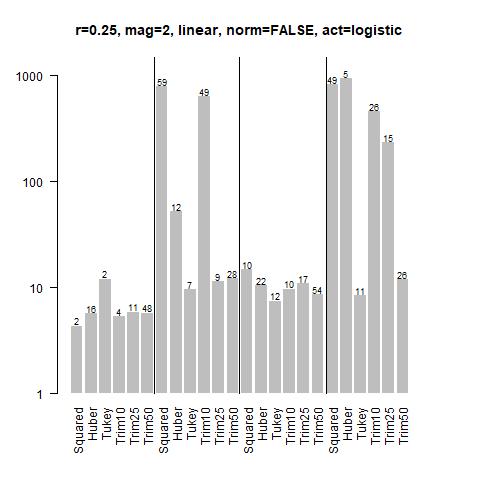} \\
\includegraphics[width=6.75cm,height=6.25cm]{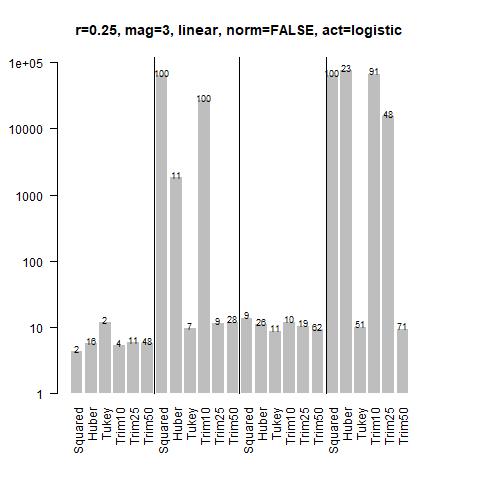} 
\includegraphics[width=6.75cm,height=6.25cm]{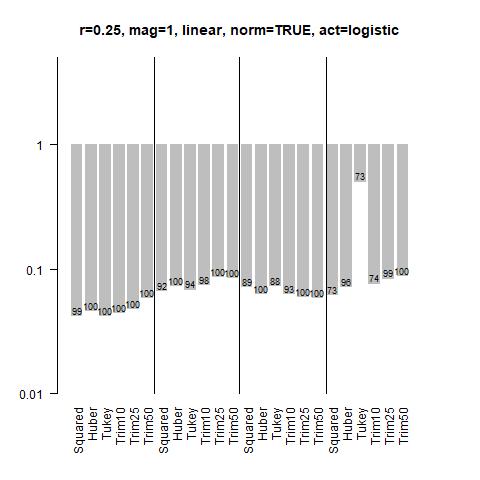}\\
\includegraphics[width=6.75cm,height=6.25cm]{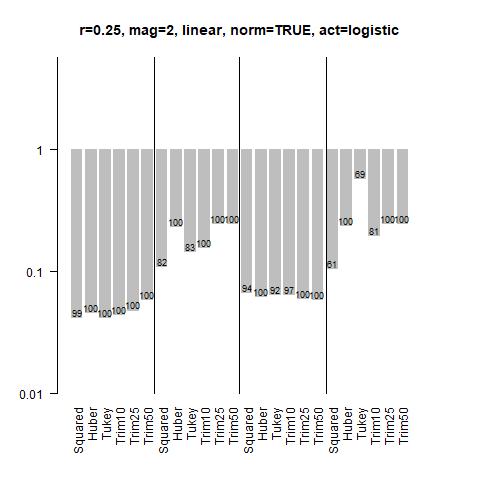} 
\includegraphics[width=6.75cm,height=6.25cm]{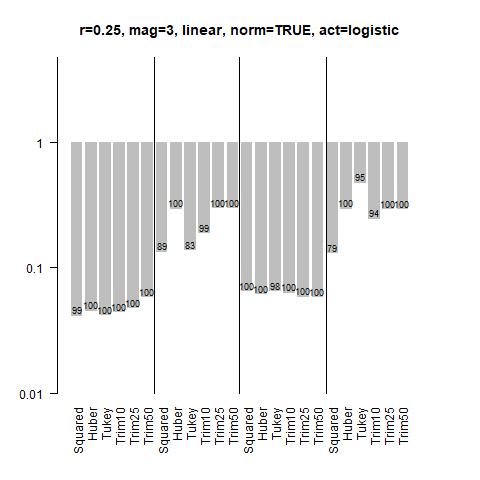} 
\end{center}
\caption{Results for $r=0.25$}\label{trimnn:n200p5r25m1linnonlogdeep}
\end{figure}

\begin{figure}[H]
\begin{center}
\includegraphics[width=6.75cm,height=6.25cm]{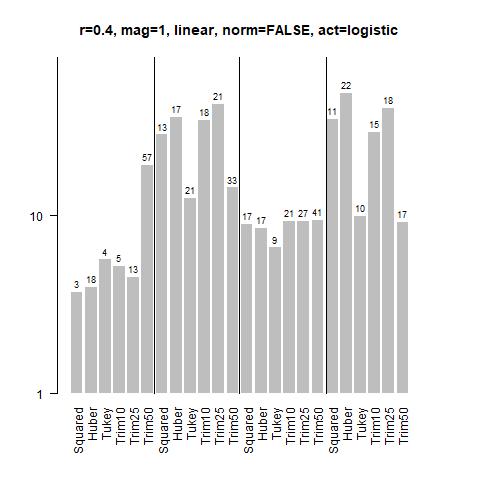}
\includegraphics[width=6.75cm,height=6.25cm]{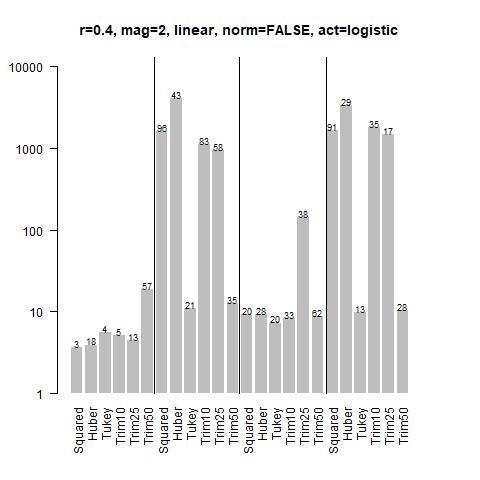} \\
\includegraphics[width=6.75cm,height=6.25cm]{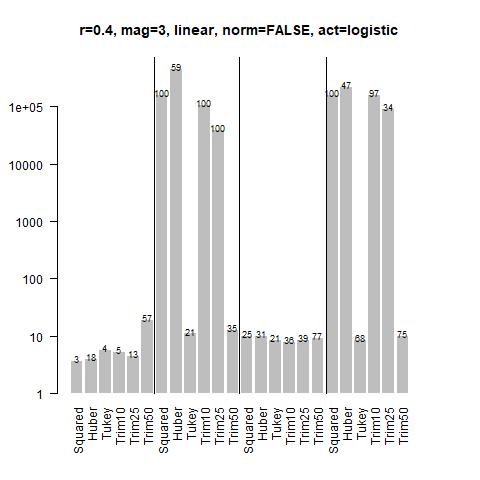} 
\includegraphics[width=6.75cm,height=6.25cm]{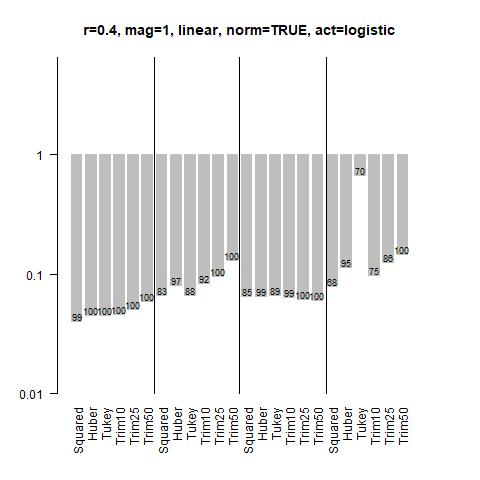}\\
\includegraphics[width=6.75cm,height=6.25cm]{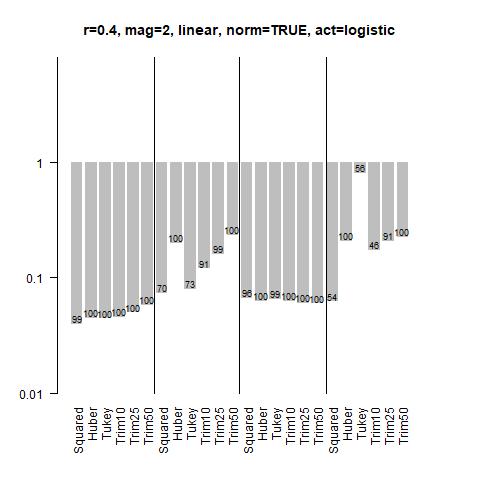} 
\includegraphics[width=6.75cm,height=6.25cm]{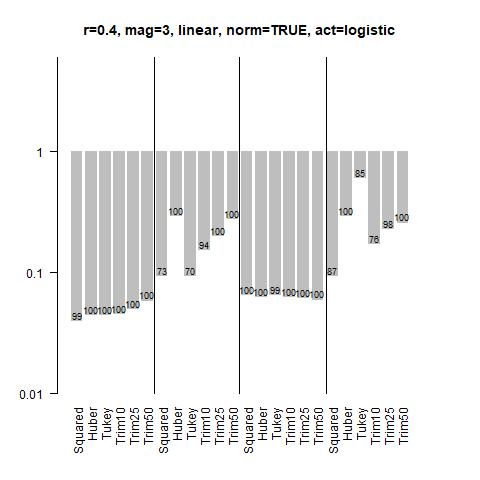} 
\end{center}
\caption{Results for $r=0.4$}\label{trimnn:n200p5r40m1linnonlogdeep}
\end{figure}

\subsubsection{Polynomial function}

\begin{figure}[H]
\label{trimnn:n200p5r10m1polynonlogdeep}
\begin{center}
\includegraphics[width=6.75cm,height=6.25cm]{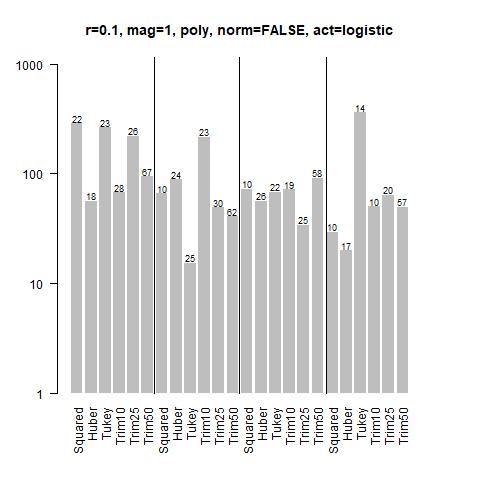}
\includegraphics[width=6.75cm,height=6.25cm]{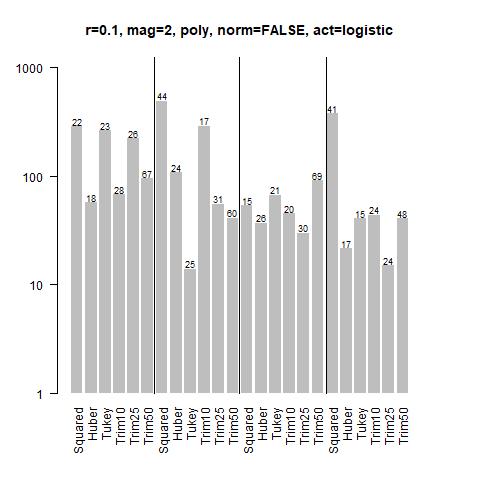} \\
\includegraphics[width=6.75cm,height=6.25cm]{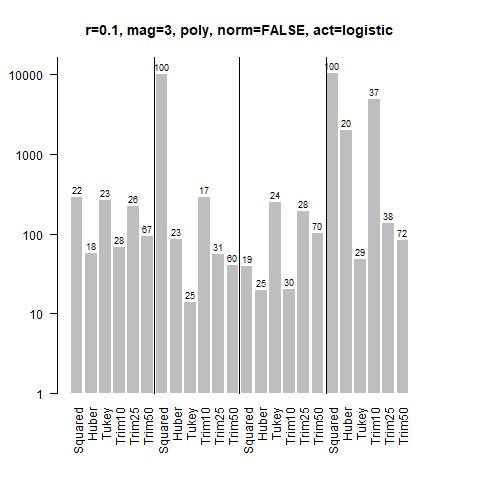} 
\includegraphics[width=6.75cm,height=6.25cm]{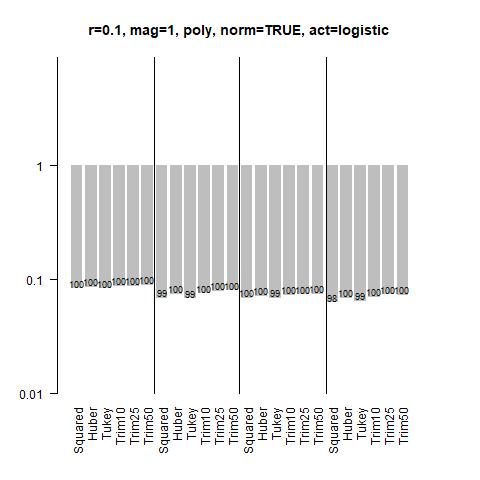}\\
\includegraphics[width=6.75cm,height=6.25cm]{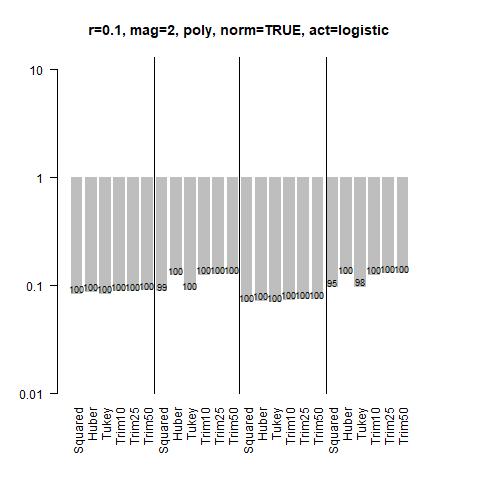} 
\includegraphics[width=6.75cm,height=6.25cm]{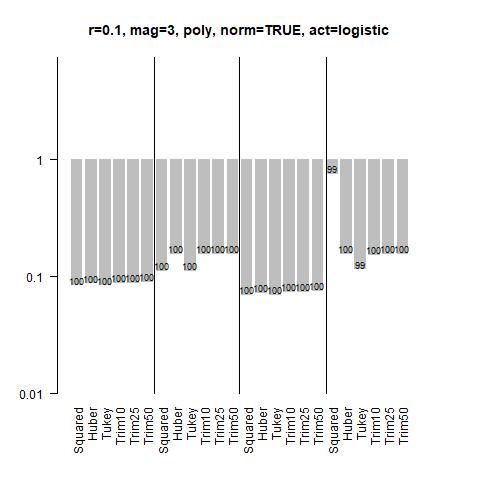} 
\end{center}
\caption{Results for $r=0.1$}
\end{figure}

\begin{figure}[H]
\label{trimnn:n200p5r25m1polynonlogdeep}
\begin{center}
\includegraphics[width=6.75cm,height=6.25cm]{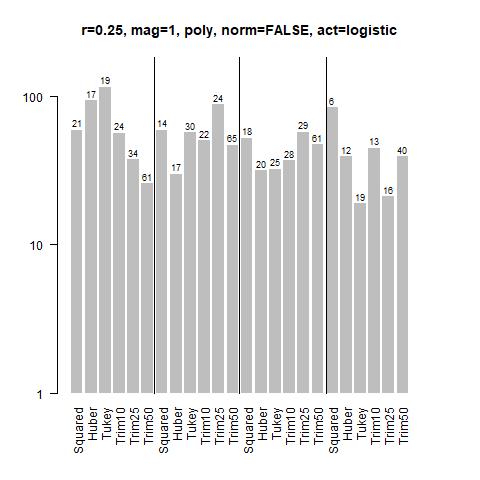}
\includegraphics[width=6.75cm,height=6.25cm]{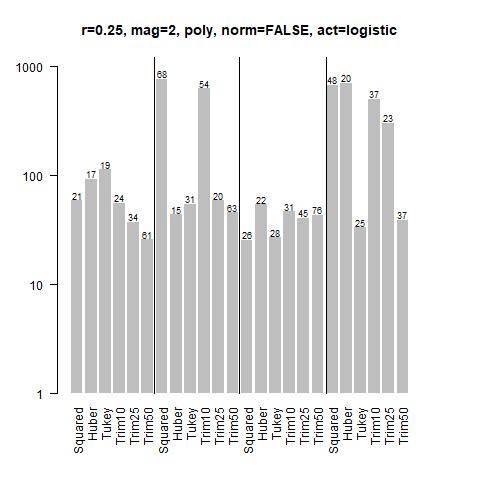} \\
\includegraphics[width=6.75cm,height=6.25cm]{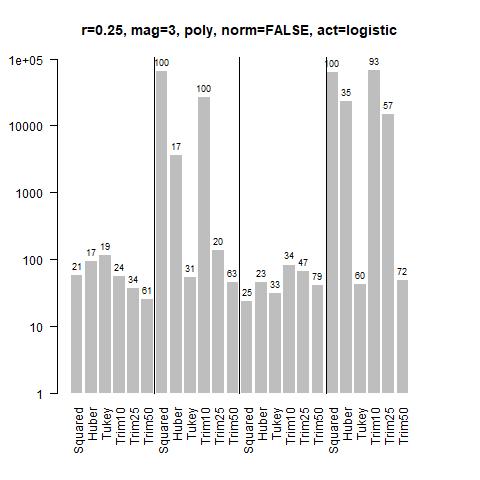} 
\includegraphics[width=6.75cm,height=6.25cm]{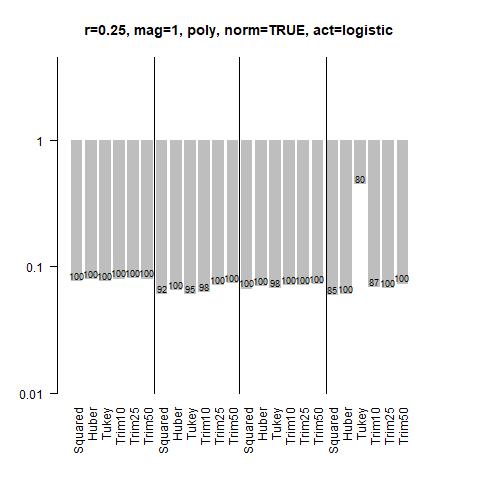}\\
\includegraphics[width=6.75cm,height=6.25cm]{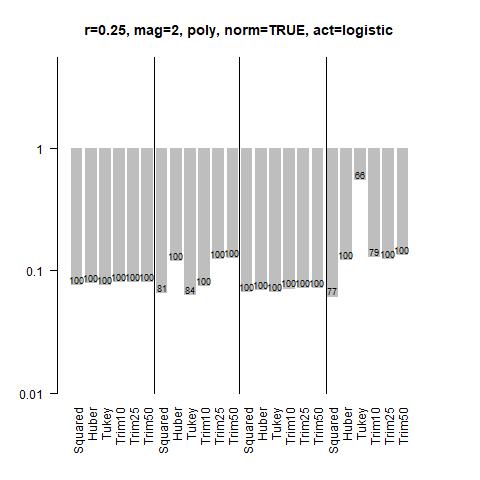} 
\includegraphics[width=6.75cm,height=6.25cm]{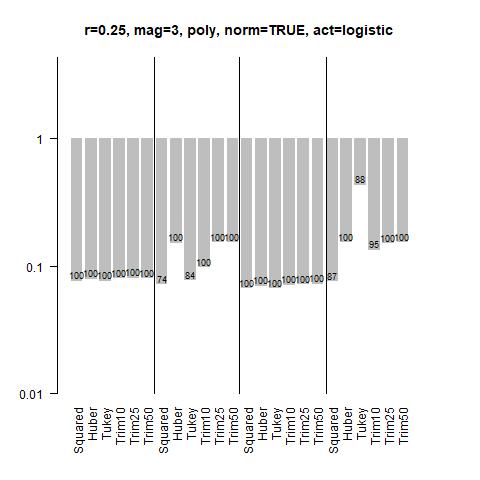} 
\end{center}
\caption{Results for $r=0.25$}
\end{figure}

\begin{figure}[H]
\label{trimnn:n200p5r40m1polynonlogdeep}
\begin{center}
\includegraphics[width=6.75cm,height=6.25cm]{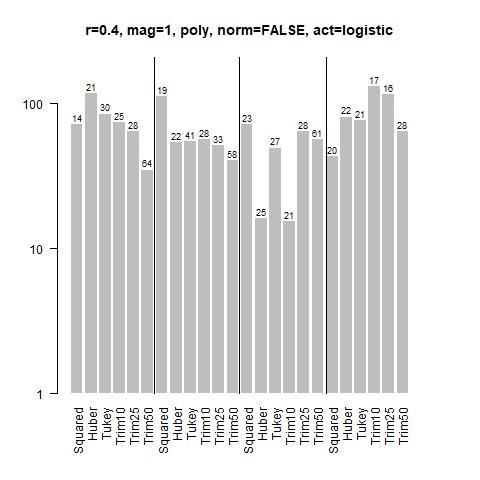}
\includegraphics[width=6.75cm,height=6.25cm]{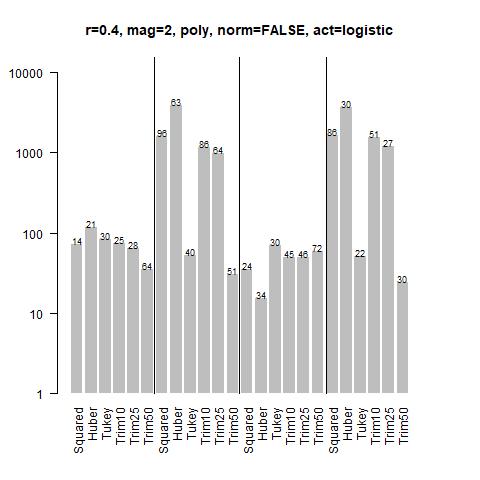} \\
\includegraphics[width=6.75cm,height=6.25cm]{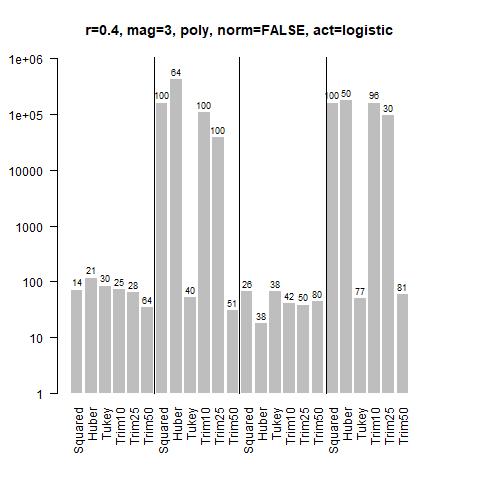} 
\includegraphics[width=6.75cm,height=6.25cm]{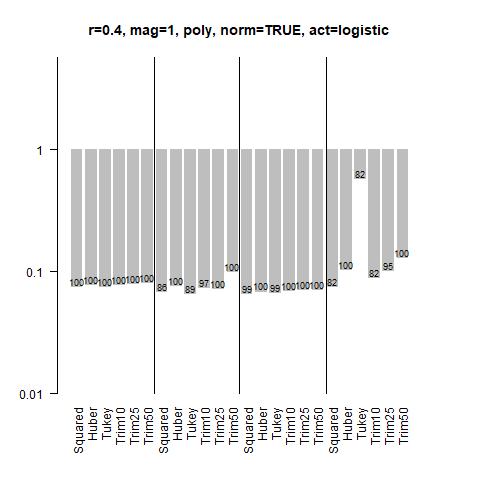}\\
\includegraphics[width=6.75cm,height=6.25cm]{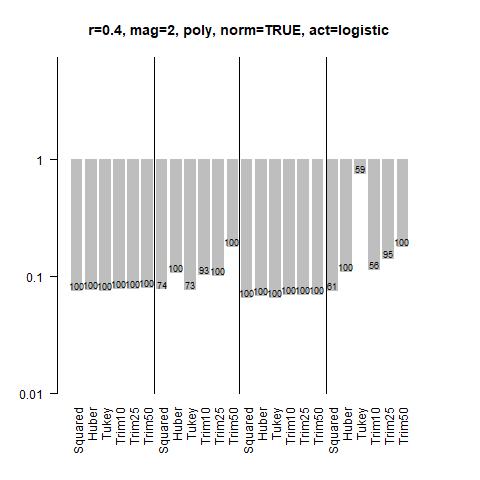} 
\includegraphics[width=6.75cm,height=6.25cm]{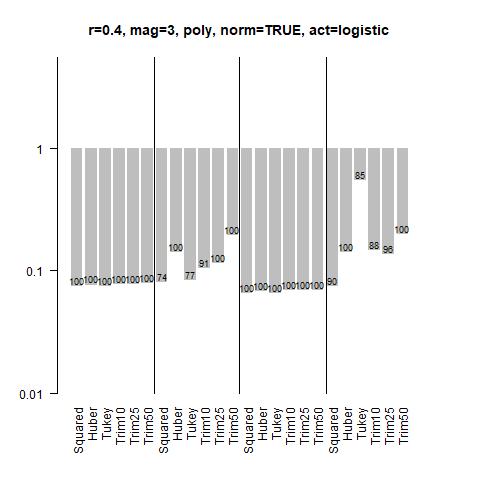} 
\end{center}
\caption{Results for $r=0.4$}
\end{figure}

\subsubsection{Trigonometric function}

\begin{figure}[H]
\label{trimnn:n200p5r10m1trignonlogdeep}
\begin{center}
\includegraphics[width=6.75cm,height=6.25cm]{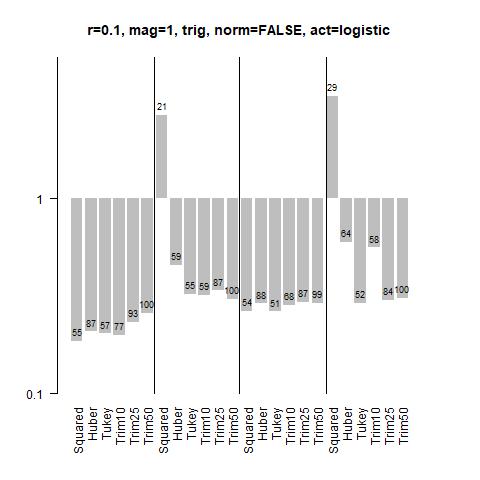}
\includegraphics[width=6.75cm,height=6.25cm]{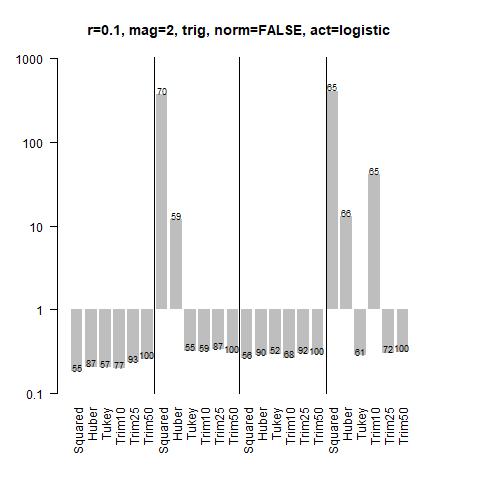} \\
\includegraphics[width=6.75cm,height=6.25cm]{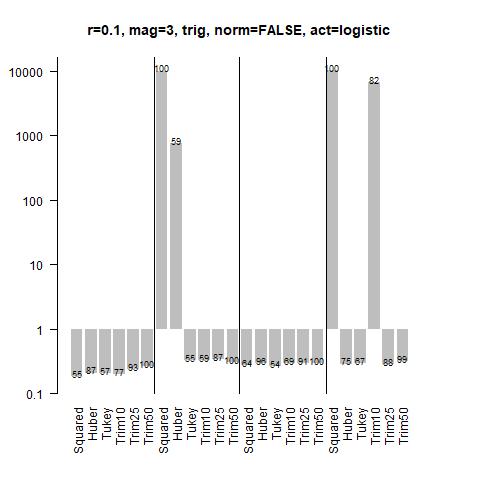} 
\includegraphics[width=6.75cm,height=6.25cm]{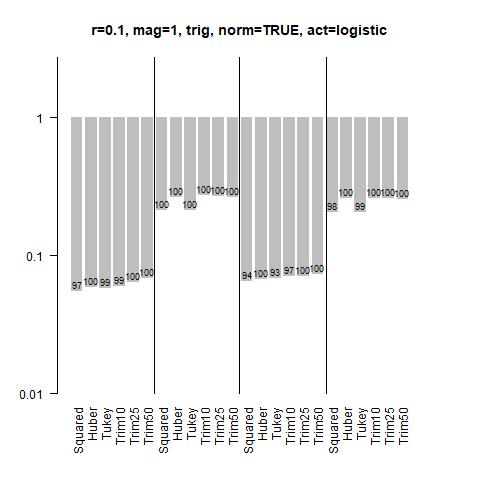}\\
\includegraphics[width=6.75cm,height=6.25cm]{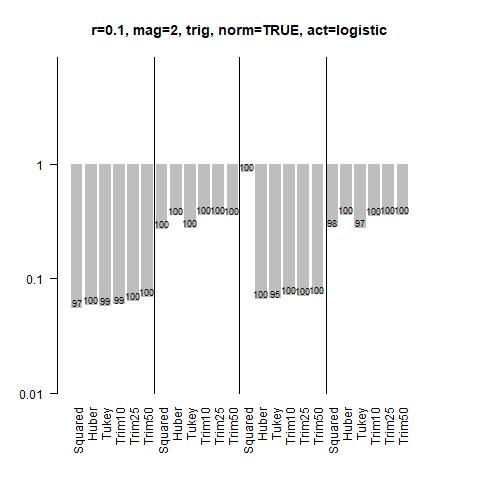} 
\includegraphics[width=6.75cm,height=6.25cm]{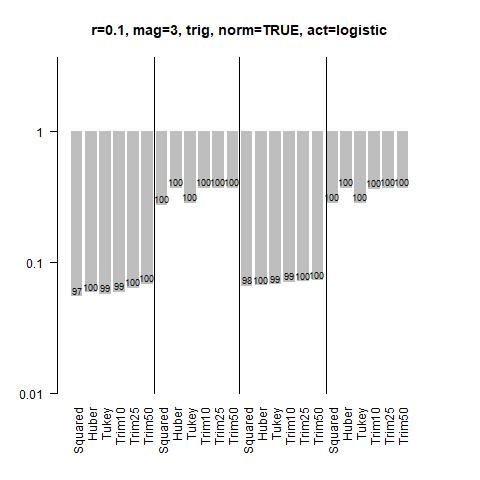} 
\end{center}
\caption{Results for $r=0.1$}
\end{figure}

\begin{figure}[H]
\label{trimnn:n200p5r25m1trignonlogdeep}
\begin{center}
\includegraphics[width=6.75cm,height=6.25cm]{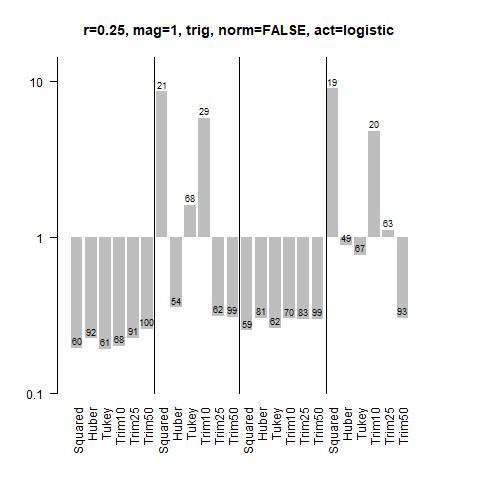}
\includegraphics[width=6.75cm,height=6.25cm]{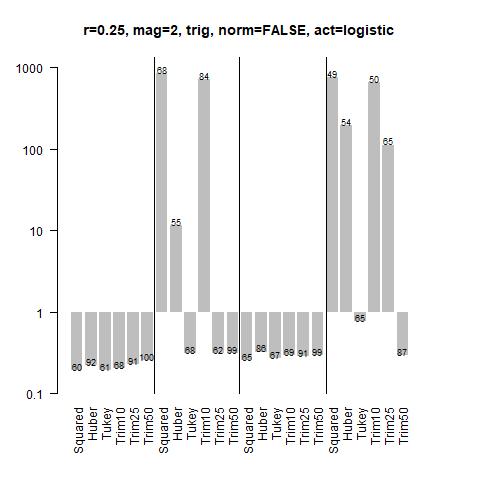} \\
\includegraphics[width=6.75cm,height=6.25cm]{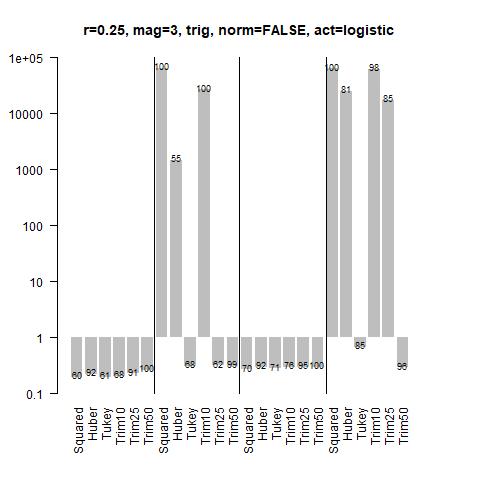} 
\includegraphics[width=6.75cm,height=6.25cm]{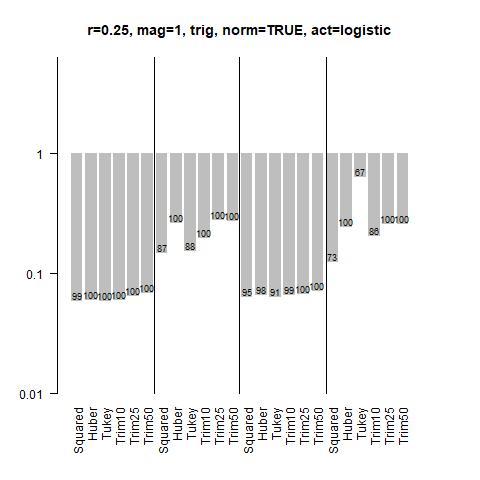}\\
\includegraphics[width=6.75cm,height=6.25cm]{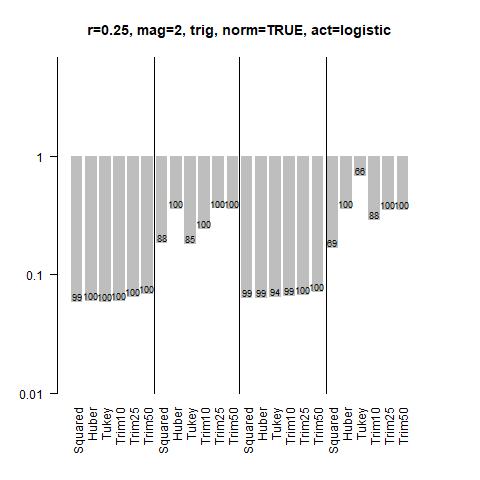} 
\includegraphics[width=6.75cm,height=6.25cm]{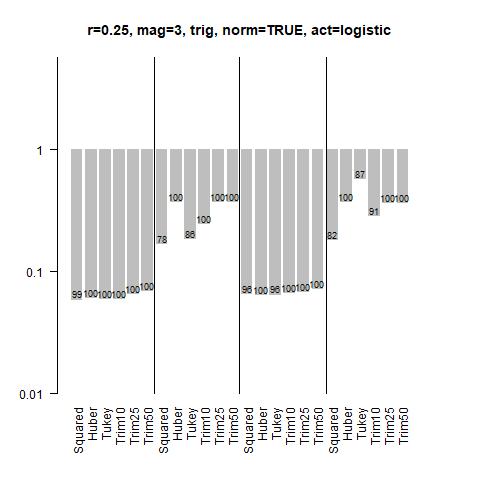} 
\end{center}
\caption{Results for $r=0.25$}
\end{figure}

\begin{figure}[H]
\begin{center}
\includegraphics[width=6.75cm,height=6.25cm]{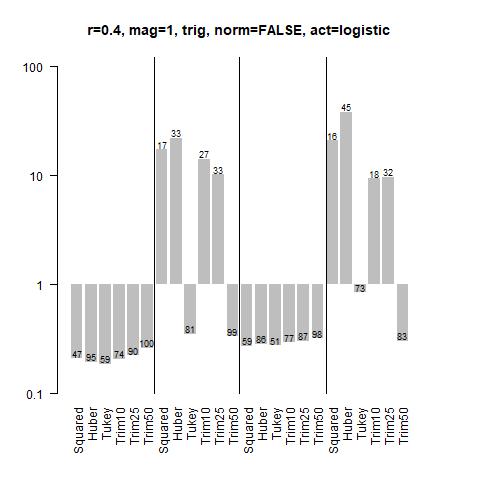}
\includegraphics[width=6.75cm,height=6.25cm]{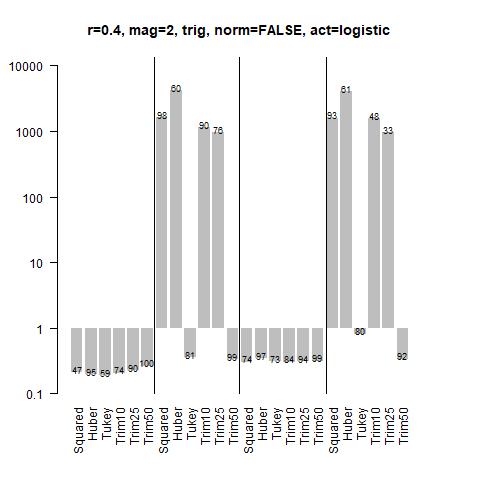} \\
\includegraphics[width=6.75cm,height=6.25cm]{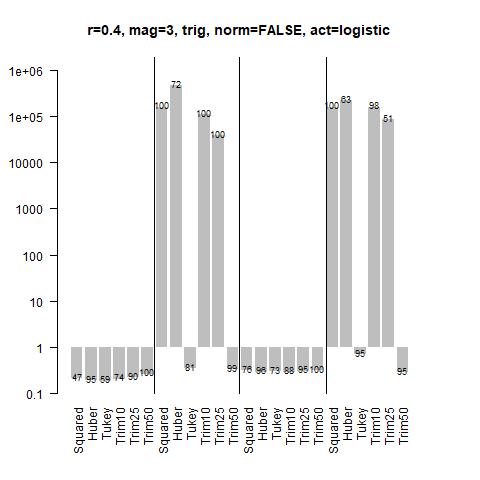} 
\includegraphics[width=6.75cm,height=6.25cm]{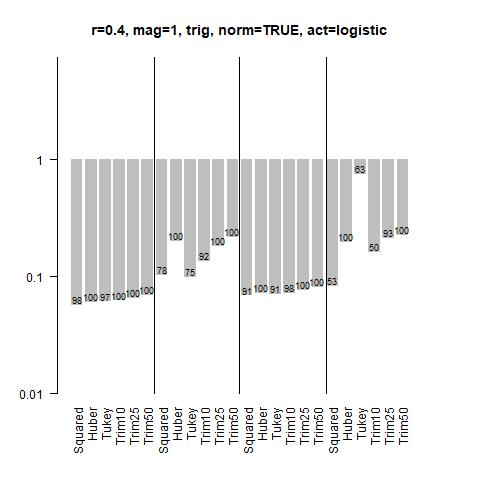}\\
\includegraphics[width=6.75cm,height=6.25cm]{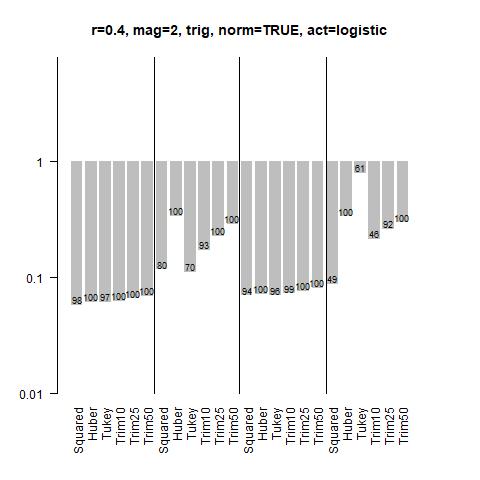} 
\includegraphics[width=6.75cm,height=6.25cm]{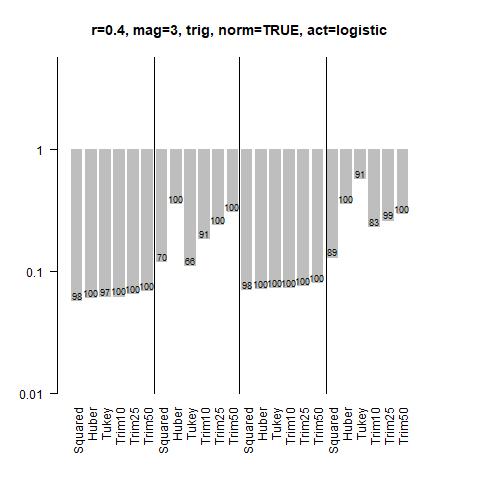} 
\end{center}
\caption{Results for $r=0.4$}\label{trimnn:n200p5r40m1trignonlogdeep}
\end{figure}

\subsection{Softplus activation function}

\subsubsection{Linear function}

\begin{figure}[H]
\label{trimnn:n200p5r10m1linnonreludeep}
\begin{center}
\includegraphics[width=6.75cm,height=6.25cm]{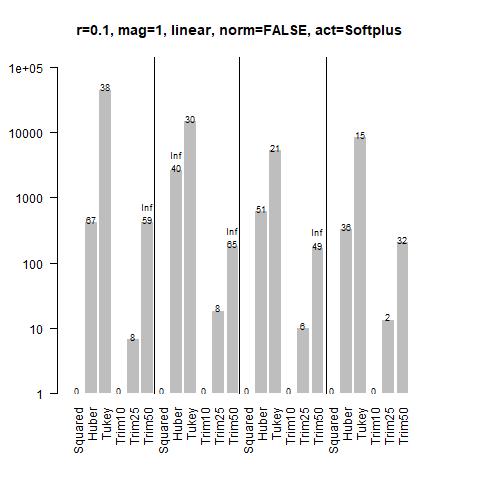}
\includegraphics[width=6.75cm,height=6.25cm]{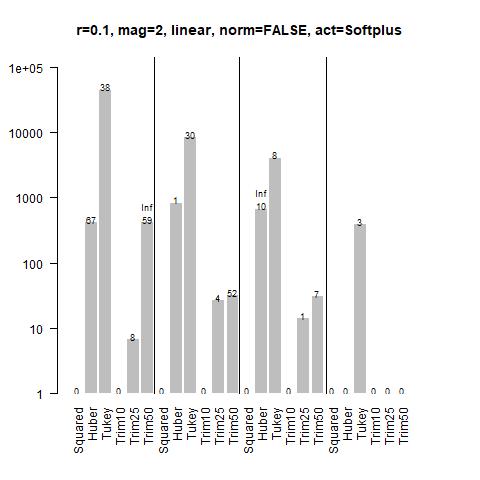} \\
\includegraphics[width=6.75cm,height=6.25cm]{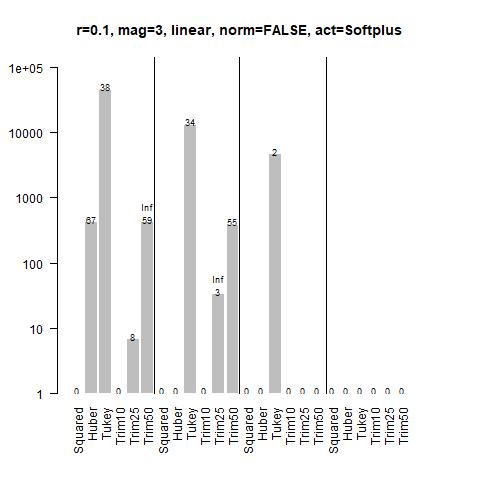} 
\includegraphics[width=6.75cm,height=6.25cm]{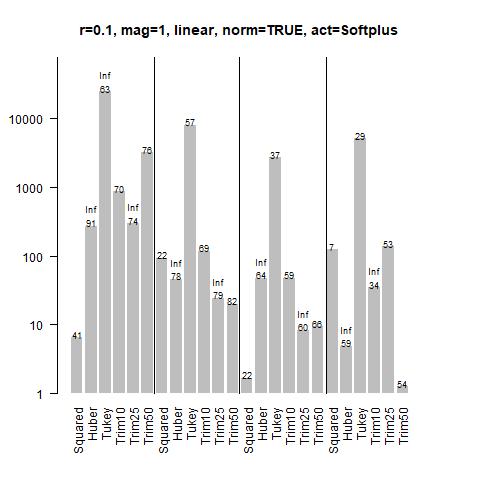}\\
\includegraphics[width=6.75cm,height=6.25cm]{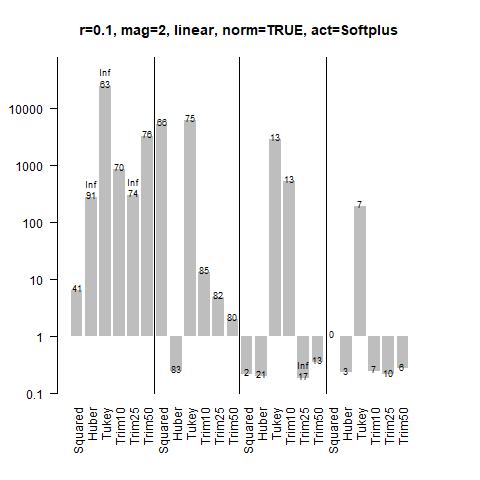} 
\includegraphics[width=6.75cm,height=6.25cm]{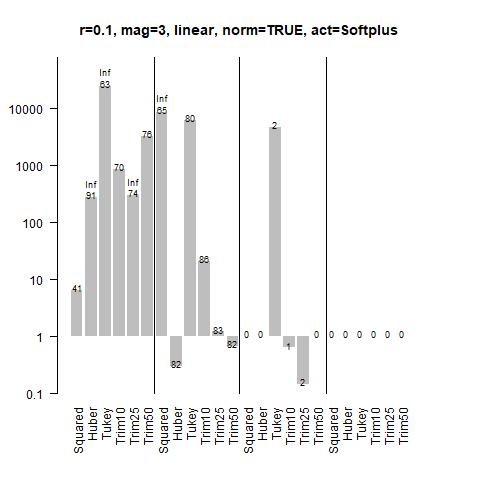} 
\end{center}
\caption{Results for $r=0.1$}
\end{figure}

\begin{figure}[H]
\begin{center}
\includegraphics[width=6.75cm,height=6.25cm]{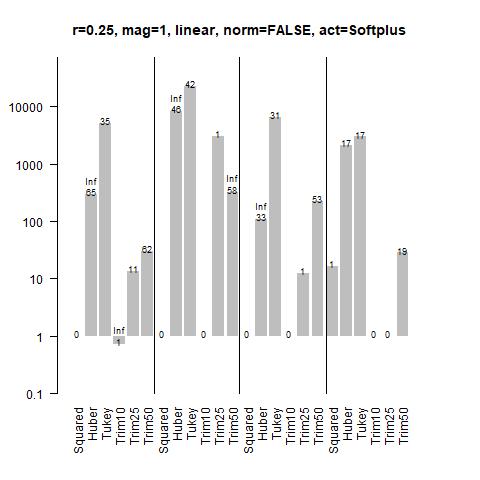}
\includegraphics[width=6.75cm,height=6.25cm]{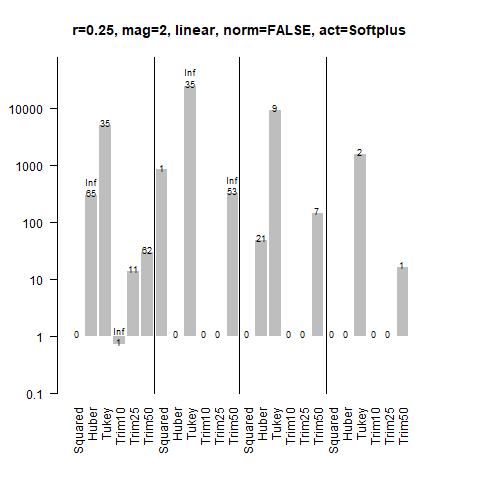} \\
\includegraphics[width=6.75cm,height=6.25cm]{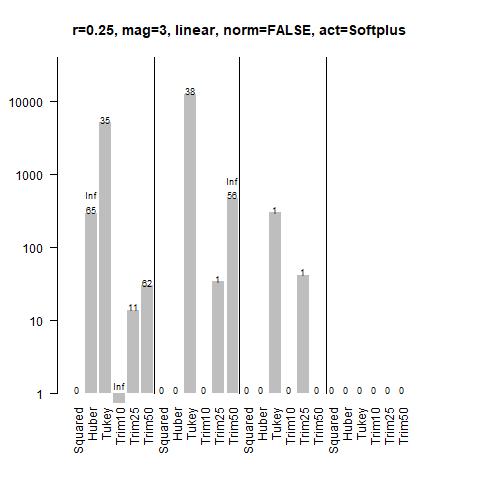} 
\includegraphics[width=6.75cm,height=6.25cm]{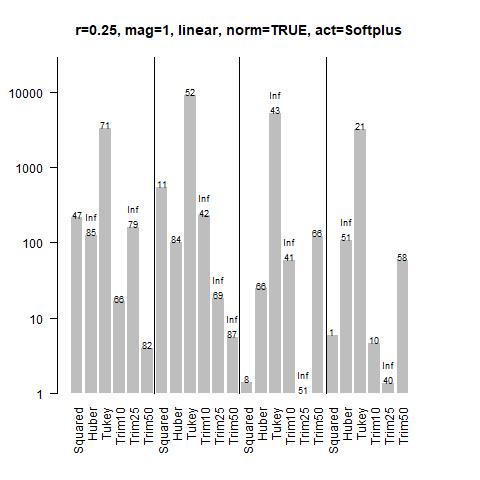}\\
\includegraphics[width=6.75cm,height=6.25cm]{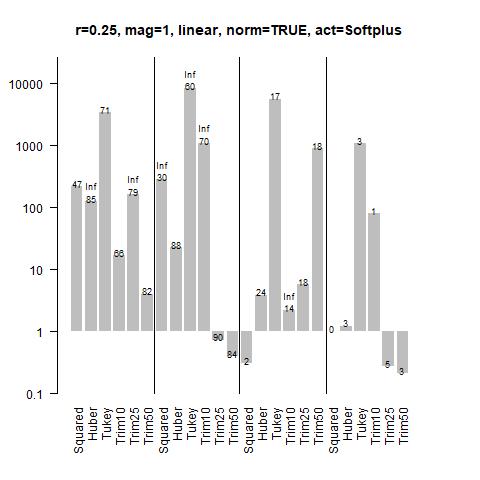} 
\includegraphics[width=6.75cm,height=6.25cm]{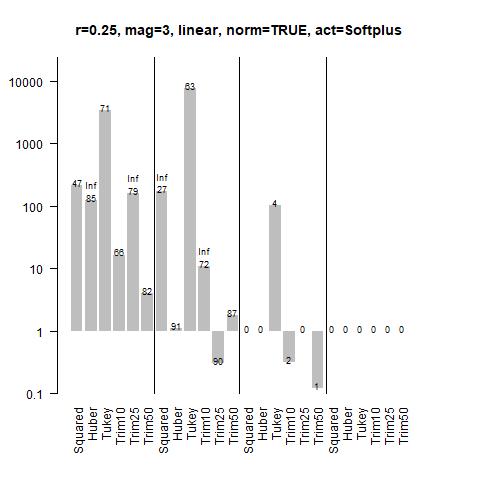} 
\end{center}
\caption{Results for $r=0.25$}\label{trimnn:n200p5r25m1linnonreludeep}
\end{figure}

\begin{figure}[H]
\begin{center}
\includegraphics[width=6.75cm,height=6.25cm]{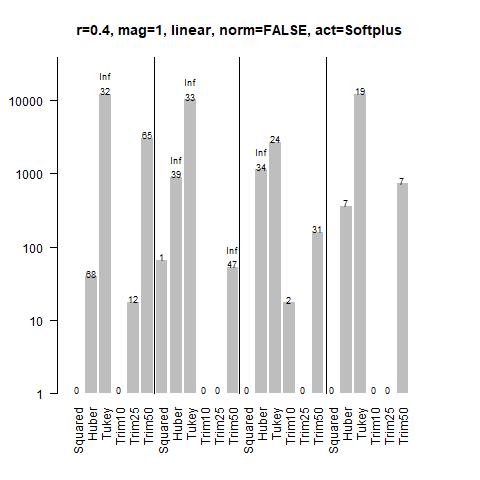}
\includegraphics[width=6.75cm,height=6.25cm]{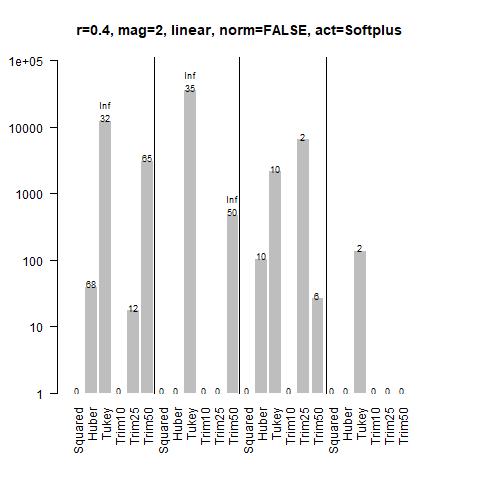} \\
\includegraphics[width=6.75cm,height=6.25cm]{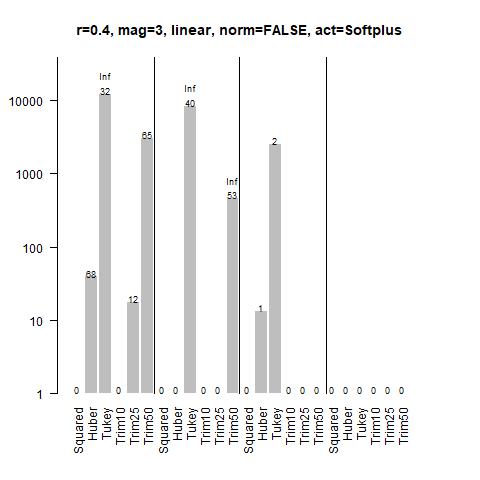} 
\includegraphics[width=6.75cm,height=6.25cm]{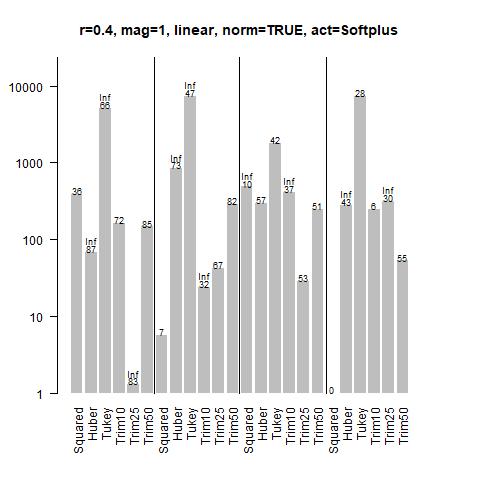}\\
\includegraphics[width=6.75cm,height=6.25cm]{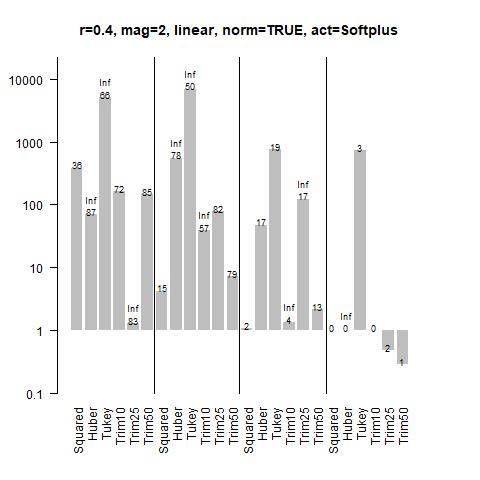} 
\includegraphics[width=6.75cm,height=6.25cm]{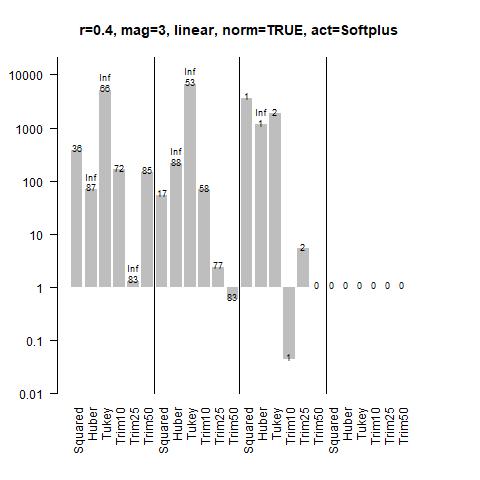} 
\end{center}
\caption{Results for $r=0.4$}\label{trimnn:n200p5r40m1linnonreludeep}
\end{figure}

\subsubsection{Polynomial function}

\begin{figure}[H]
\label{trimnn:n200p5r10m1polynonreludeep}
\begin{center}
\includegraphics[width=6.75cm,height=6.25cm]{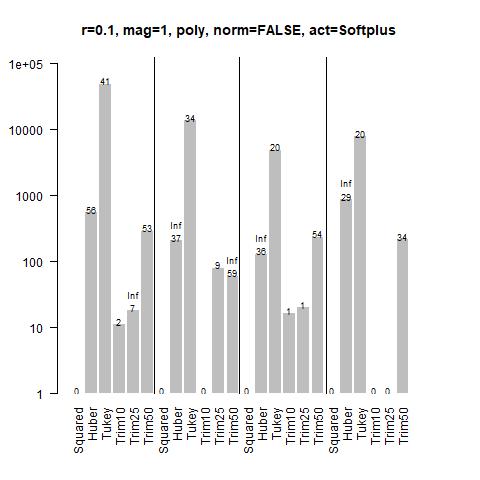}
\includegraphics[width=6.75cm,height=6.25cm]{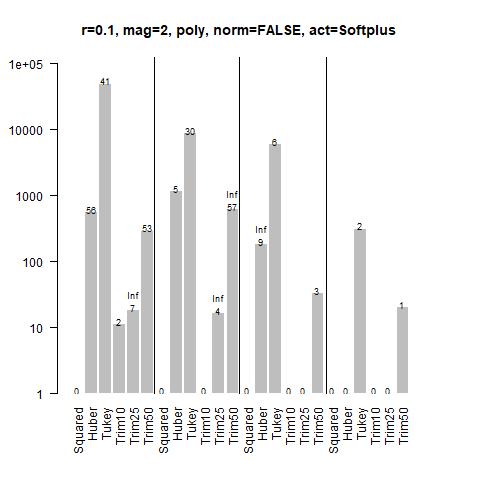} \\
\includegraphics[width=6.75cm,height=6.25cm]{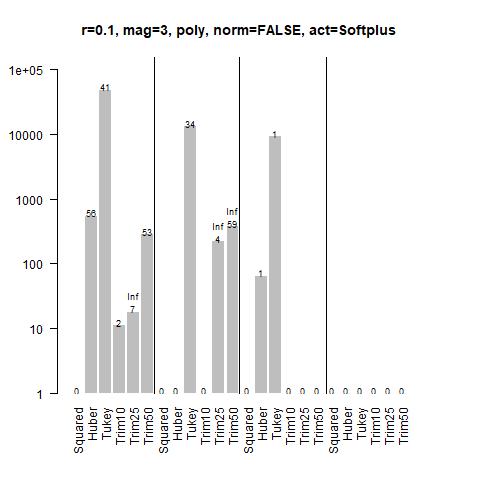} 
\includegraphics[width=6.75cm,height=6.25cm]{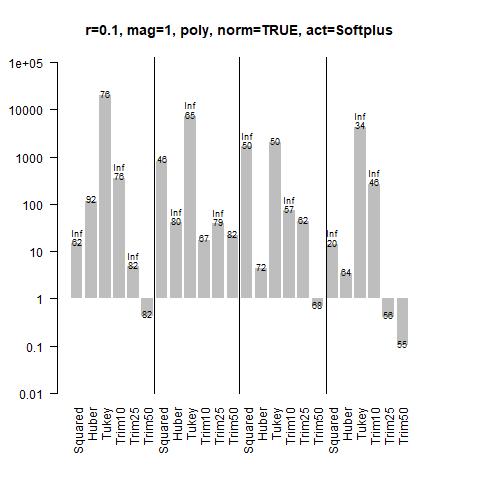}\\
\includegraphics[width=6.75cm,height=6.25cm]{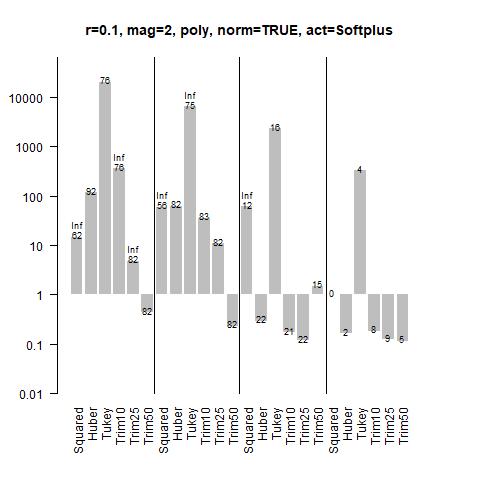} 
\includegraphics[width=6.75cm,height=6.25cm]{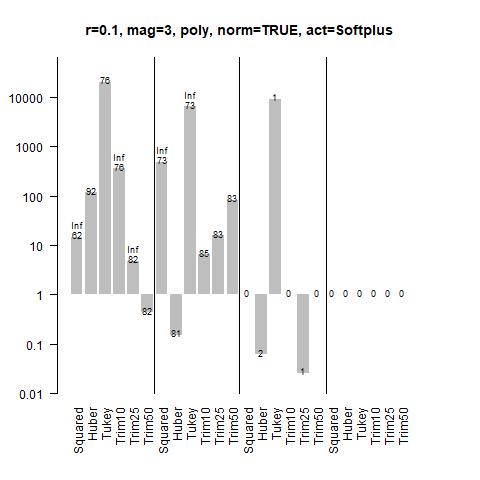} 
\end{center}
\caption{Results for $r=0.1$}
\end{figure}

\begin{figure}[H]
\label{trimnn:n200p5r25m1polynonreludeep}
\begin{center}
\includegraphics[width=6.75cm,height=6.25cm]{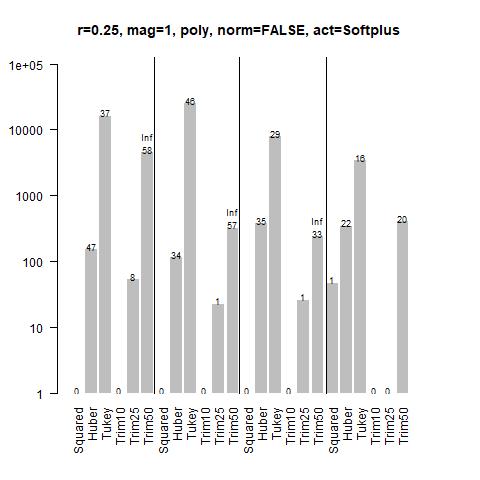}
\includegraphics[width=6.75cm,height=6.25cm]{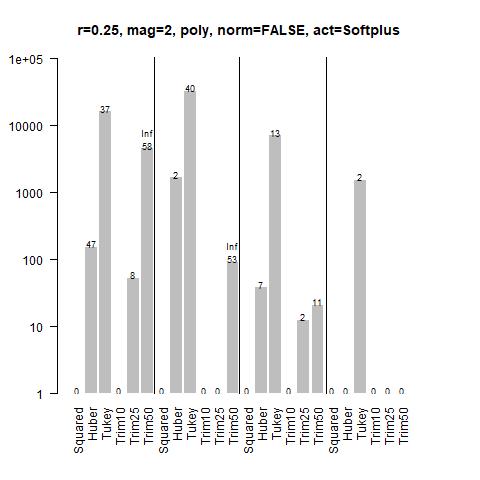} \\
\includegraphics[width=6.75cm,height=6.25cm]{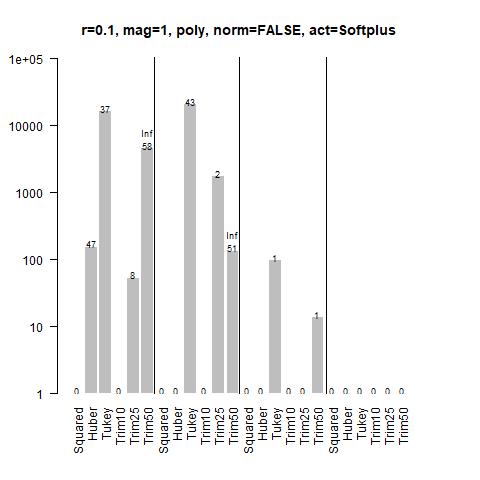} 
\includegraphics[width=6.75cm,height=6.25cm]{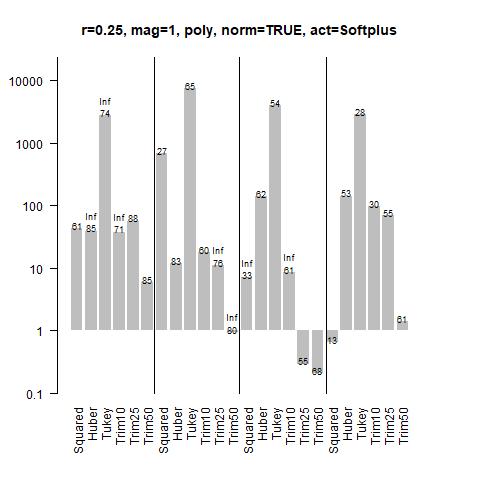}\\
\includegraphics[width=6.75cm,height=6.25cm]{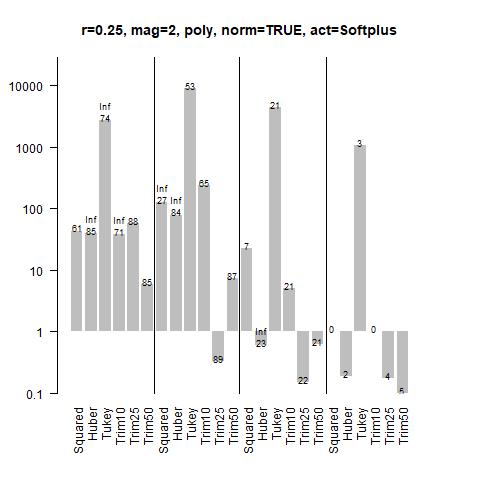} 
\includegraphics[width=6.75cm,height=6.25cm]{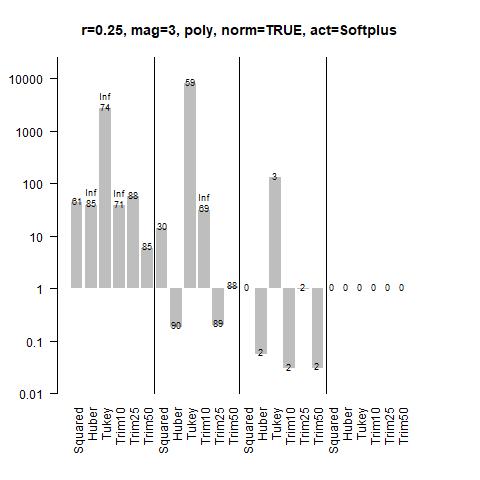} 
\end{center}
\caption{Results for $r=0.25$}
\end{figure}

\begin{figure}[H]
\label{trimnn:n200p5r40m1polynonreludeep}
\begin{center}
\includegraphics[width=6.75cm,height=6.25cm]{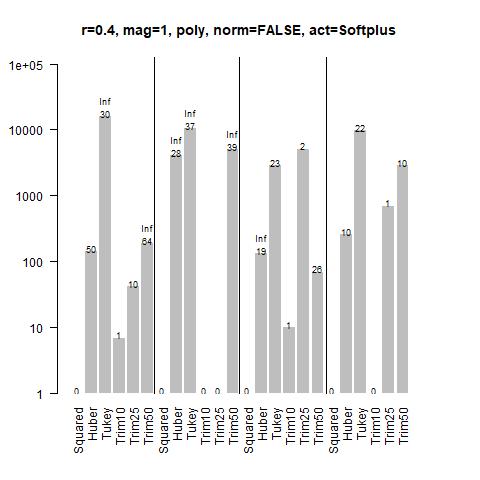}
\includegraphics[width=6.75cm,height=6.25cm]{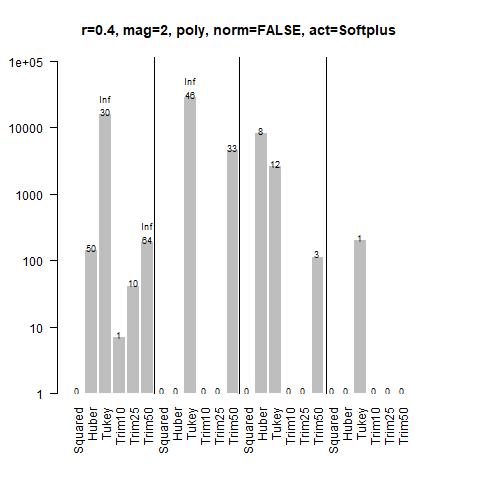} \\
\includegraphics[width=6.75cm,height=6.25cm]{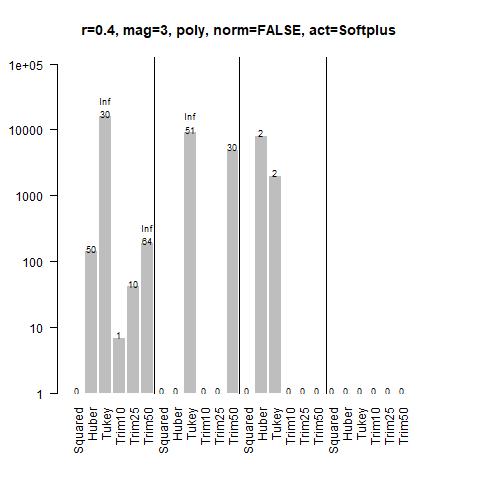} 
\includegraphics[width=6.75cm,height=6.25cm]{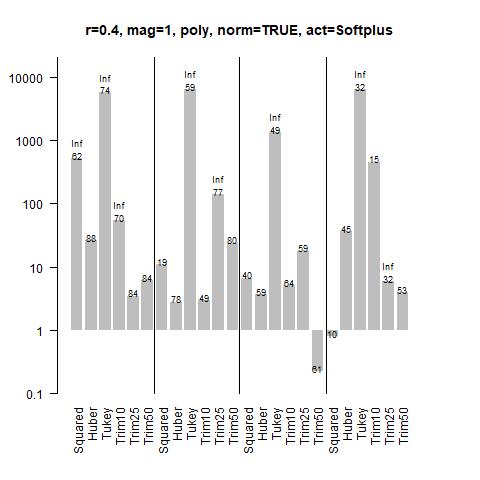}\\
\includegraphics[width=6.75cm,height=6.25cm]{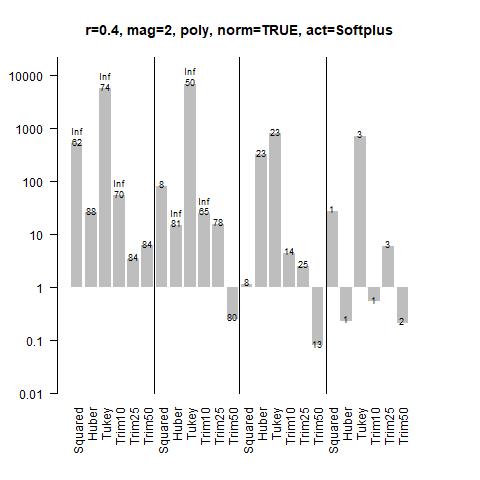} 
\includegraphics[width=6.75cm,height=6.25cm]{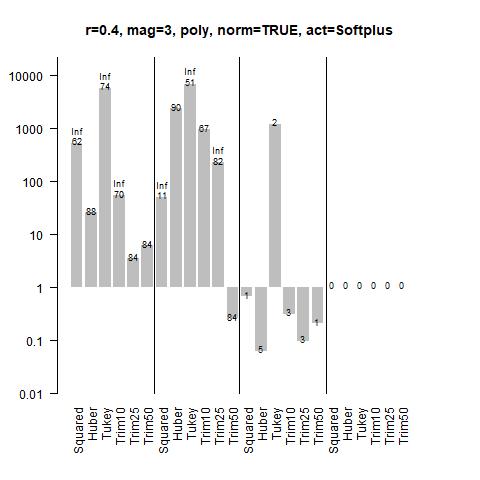} 
\end{center}
\caption{Results for $r=0.4$}
\end{figure}

\subsubsection{Trigonometric function}

\begin{figure}[H]
\label{trimnn:n200p5r10m1trignonreludeep}
\begin{center}
\includegraphics[width=6.75cm,height=6.25cm]{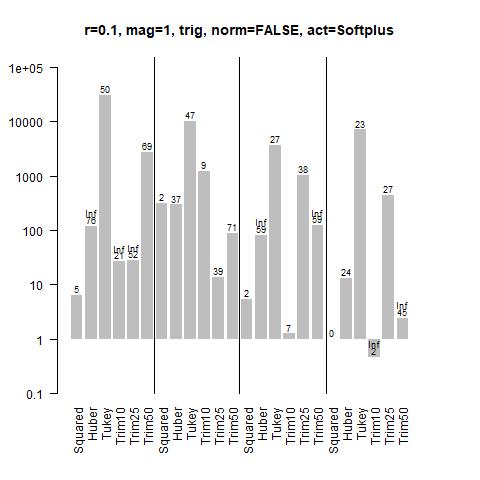}
\includegraphics[width=6.75cm,height=6.25cm]{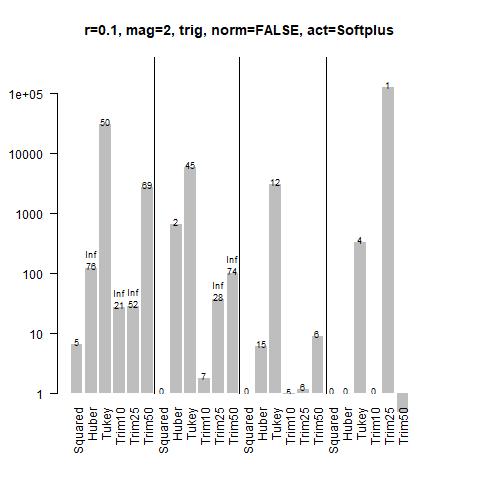} \\
\includegraphics[width=6.75cm,height=6.25cm]{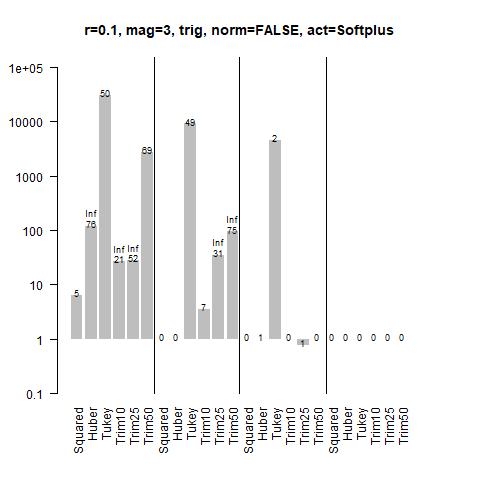} 
\includegraphics[width=6.75cm,height=6.25cm]{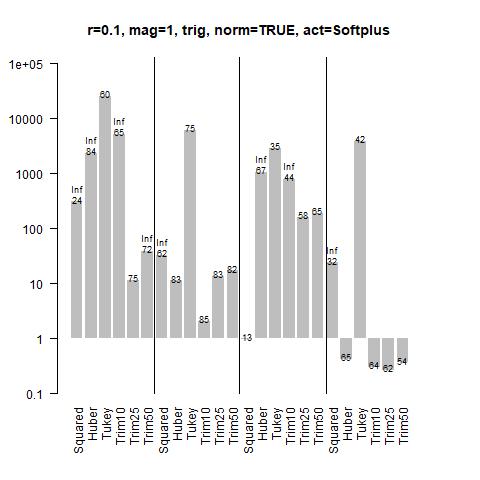}\\
\includegraphics[width=6.75cm,height=6.25cm]{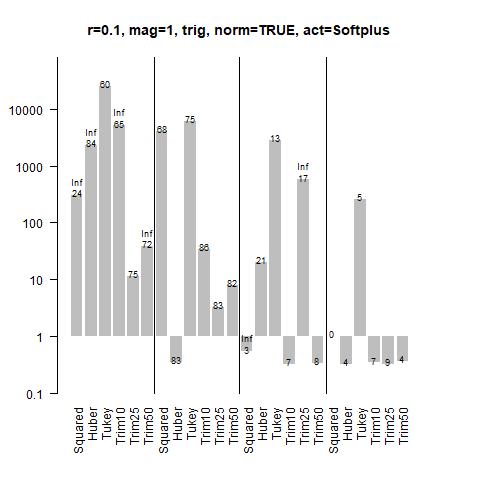} 
\includegraphics[width=6.75cm,height=6.25cm]{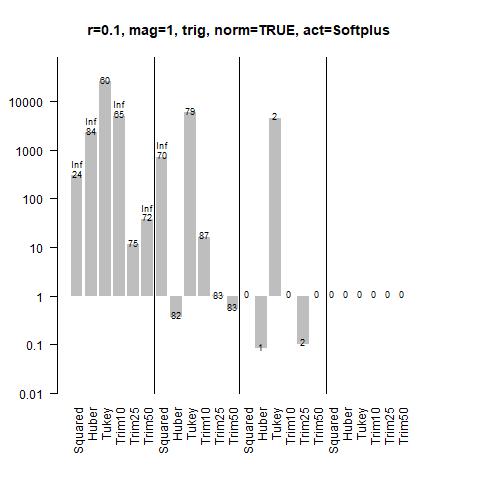} 
\end{center}
\caption{Results for $r=0.1$}
\end{figure}

\begin{figure}[H]
\label{trimnn:n200p5r25m1trignonreludeep}
\begin{center}
\includegraphics[width=6.75cm,height=6.25cm]{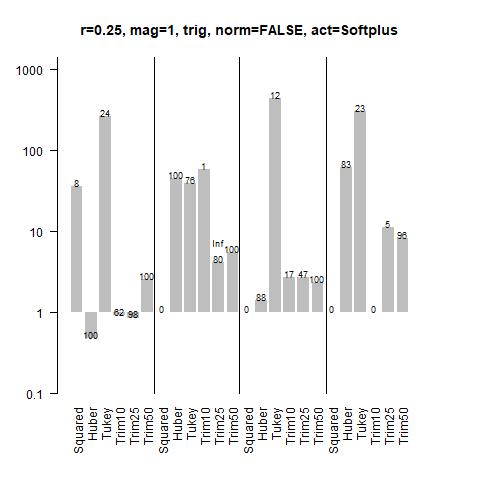}
\includegraphics[width=6.75cm,height=6.25cm]{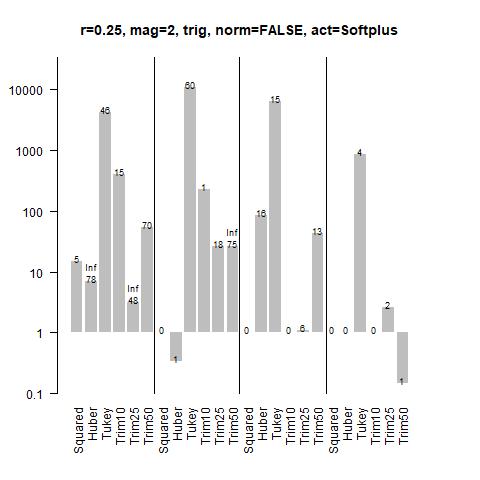} \\
\includegraphics[width=6.75cm,height=6.25cm]{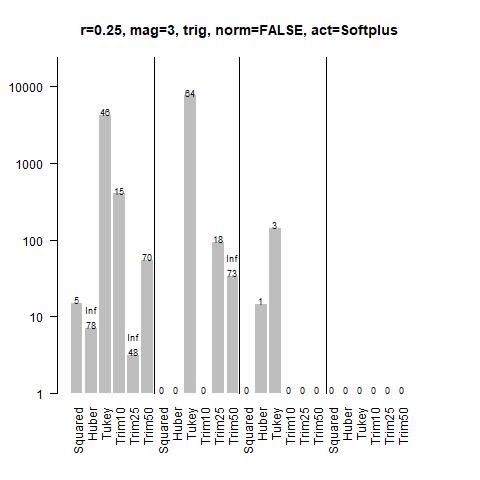} 
\includegraphics[width=6.75cm,height=6.25cm]{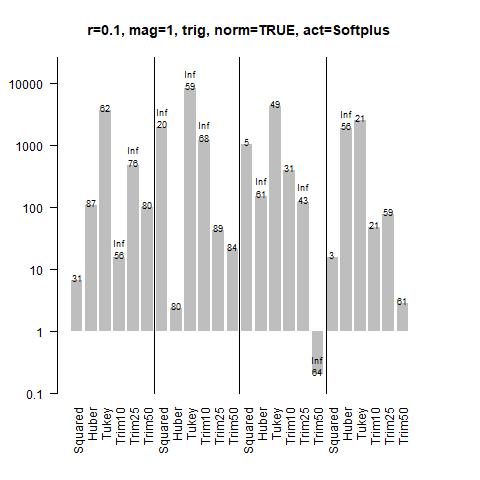}\\
\includegraphics[width=6.75cm,height=6.25cm]{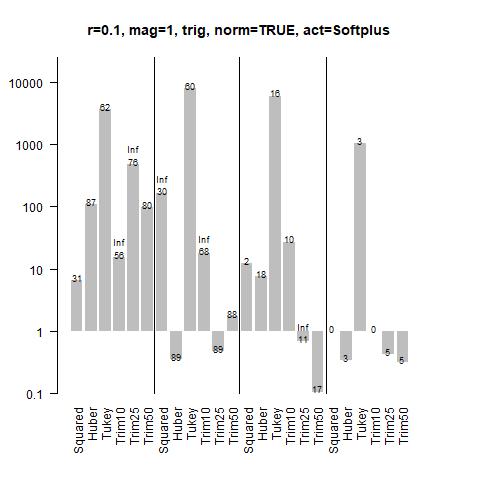} 
\includegraphics[width=6.75cm,height=6.25cm]{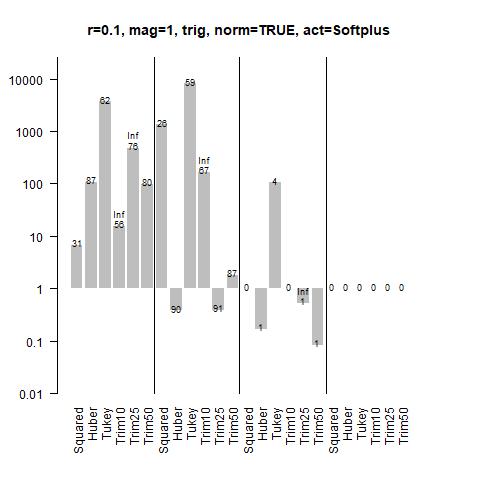} 
\end{center}
\caption{Results for $r=0.25$}
\end{figure}

\begin{figure}[H]
\label{trimnn:n200p5r40m1trignonreludeep}
\begin{center}
\includegraphics[width=6.75cm,height=6.25cm]{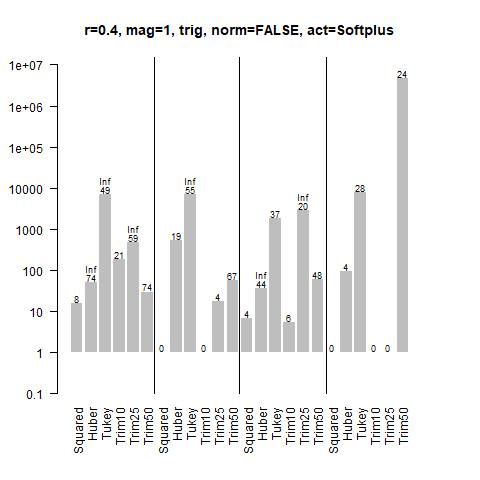}
\includegraphics[width=6.75cm,height=6.25cm]{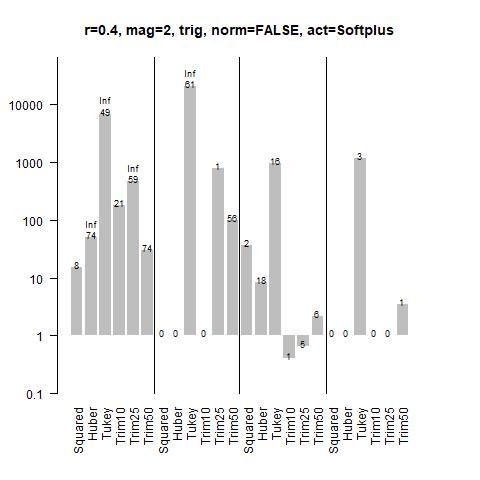} \\
\includegraphics[width=6.75cm,height=6.25cm]{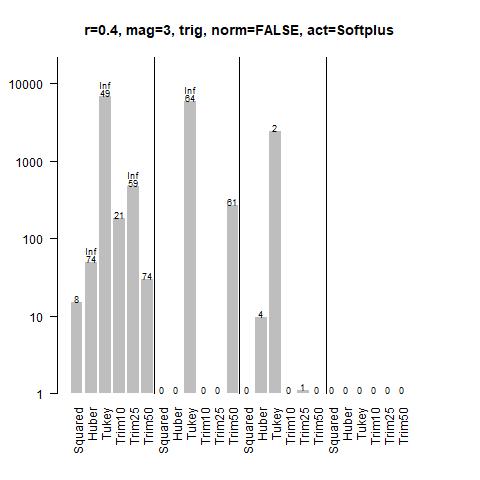} 
\includegraphics[width=6.75cm,height=6.25cm]{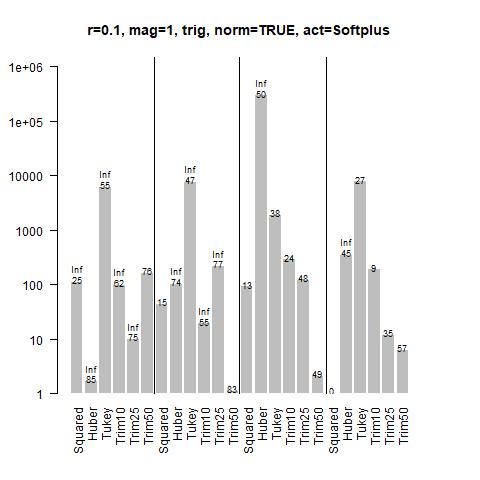}\\
\includegraphics[width=6.75cm,height=6.25cm]{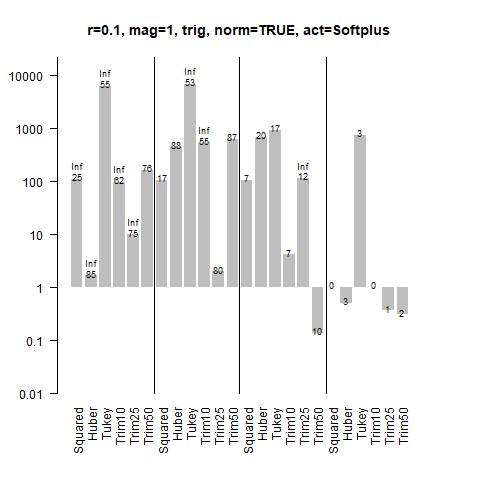} 
\includegraphics[width=6.75cm,height=6.25cm]{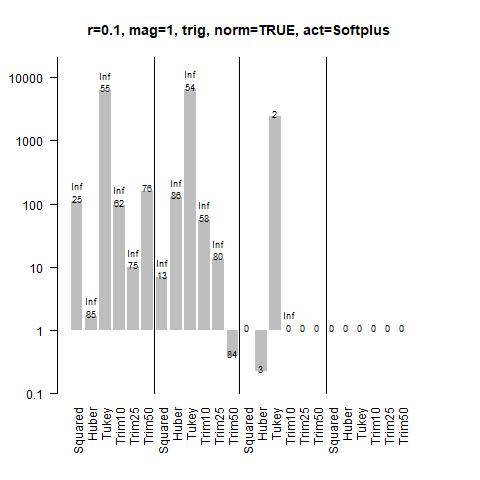} 
\end{center}
\caption{Results for $r=0.4$}
\end{figure}

\section{Simulation results for $n=500$ and $p=20$: Test loss} \label{trimnn:secloss50020}

\subsection{Logistic activation function}

\subsubsection{Linear function}

\begin{figure}[H]
\label{trimnn:n500p20r10m1linnonlog}
\begin{center}
\includegraphics[width=6.75cm,height=6.25cm]{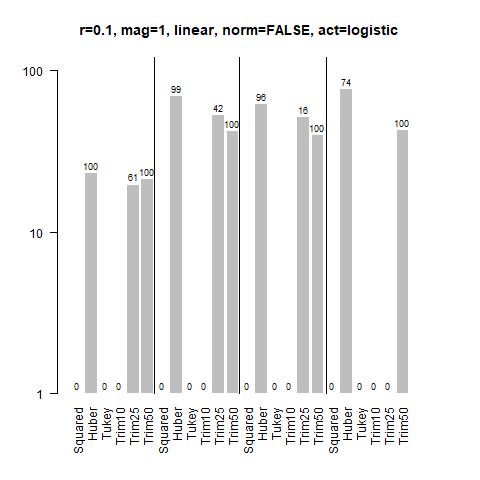}
\includegraphics[width=6.75cm,height=6.25cm]{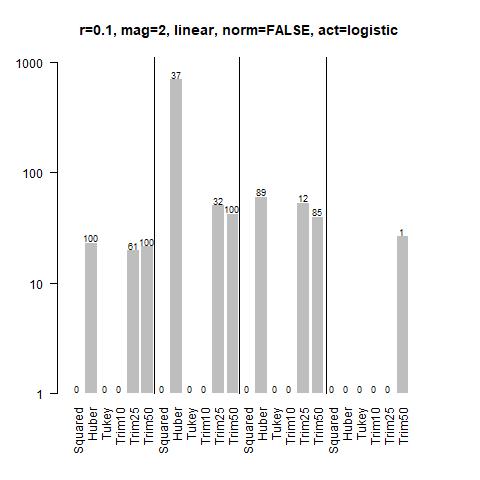} \\
\includegraphics[width=6.75cm,height=6.25cm]{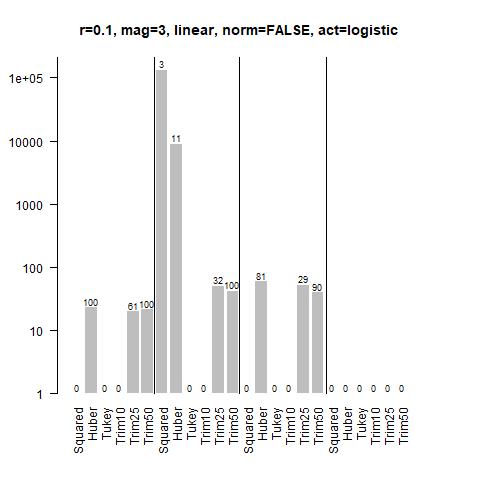} 
\includegraphics[width=6.75cm,height=6.25cm]{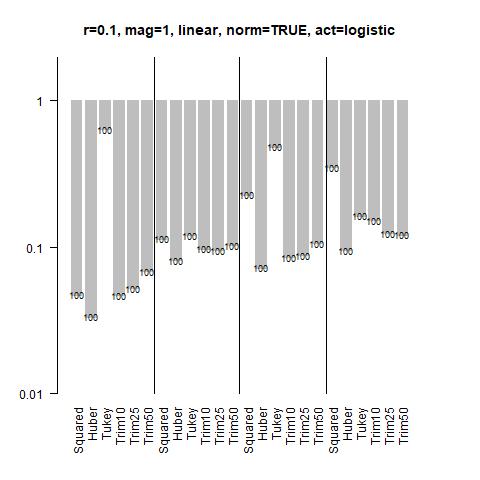}\\
\includegraphics[width=6.75cm,height=6.25cm]{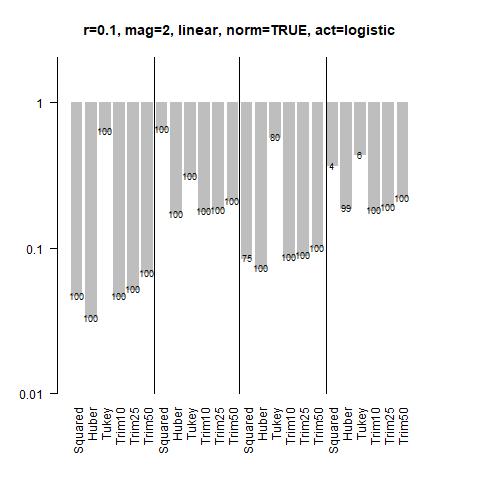} 
\includegraphics[width=6.75cm,height=6.25cm]{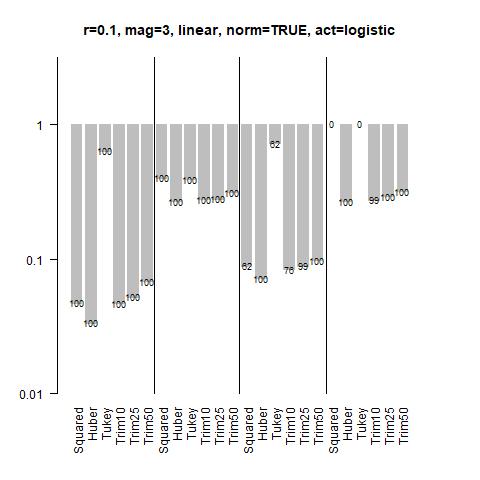} 
\end{center}
\caption{Results for $r=0.1$}
\end{figure}

\begin{figure}[H]
\label{trimnn:n500p20r25m1linnonlog}
\begin{center}
\includegraphics[width=6.75cm,height=6.25cm]{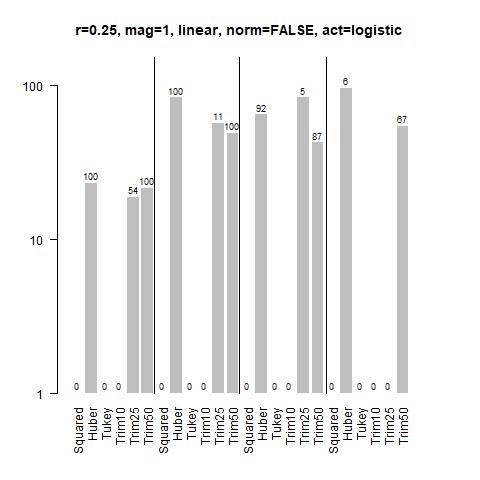}
\includegraphics[width=6.75cm,height=6.25cm]{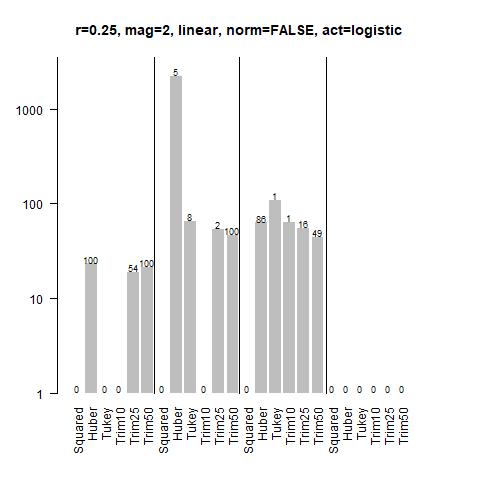} \\
\includegraphics[width=6.75cm,height=6.25cm]{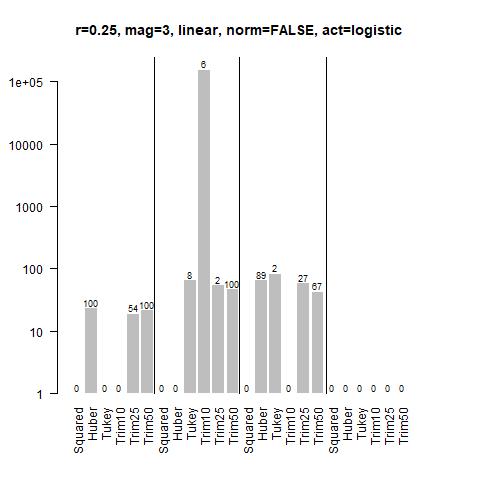} 
\includegraphics[width=6.75cm,height=6.25cm]{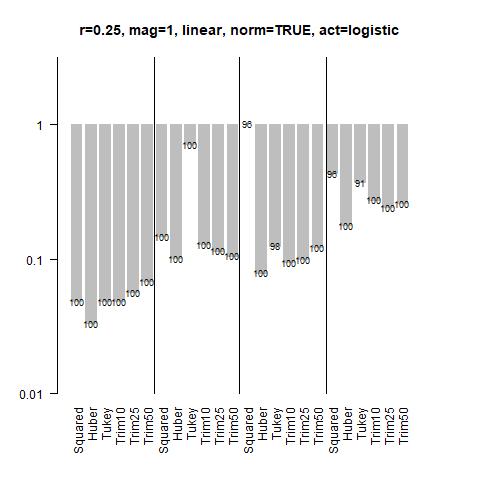}\\
\includegraphics[width=6.75cm,height=6.25cm]{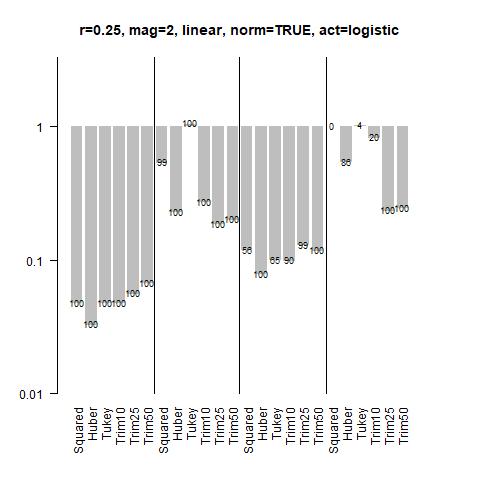} 
\includegraphics[width=6.75cm,height=6.25cm]{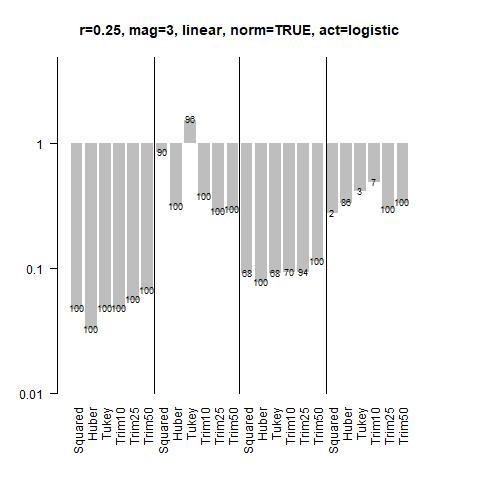} 
\end{center}
\caption{Results for $r=0.25$}
\end{figure}

\begin{figure}[H]
\label{trimnn:n500p20r40m1linnonlog}
\begin{center}
\includegraphics[width=6.75cm,height=6.25cm]{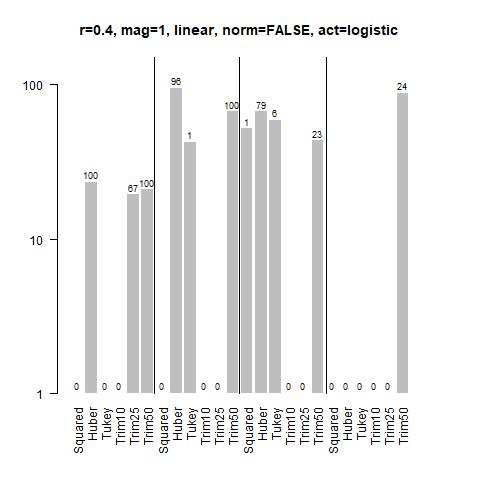}
\includegraphics[width=6.75cm,height=6.25cm]{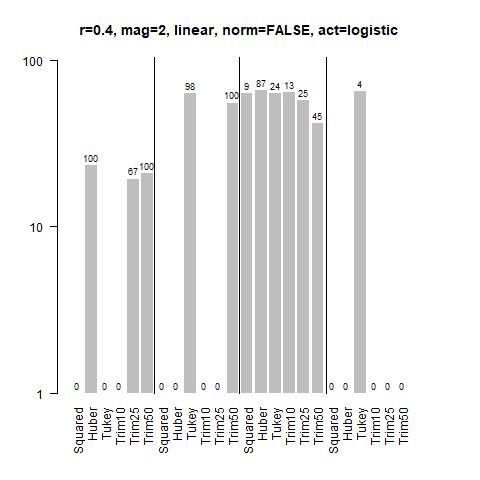} \\
\includegraphics[width=6.75cm,height=6.25cm]{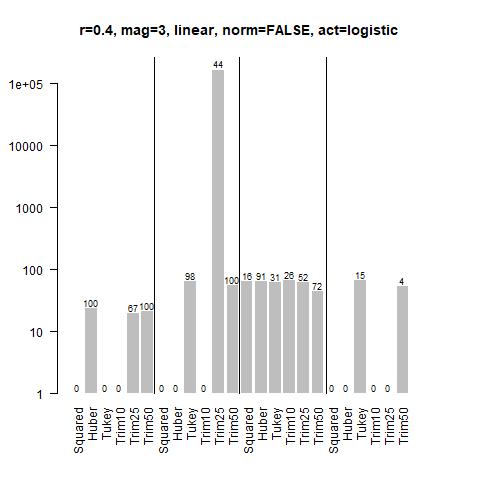} 
\includegraphics[width=6.75cm,height=6.25cm]{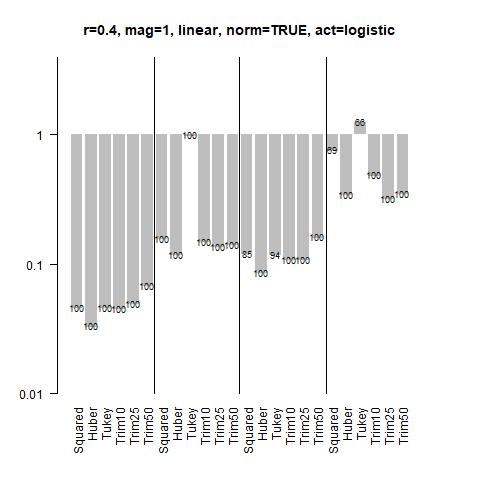}\\
\includegraphics[width=6.75cm,height=6.25cm]{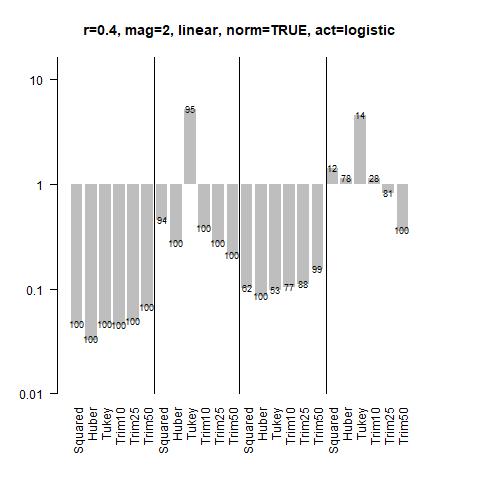} 
\includegraphics[width=6.75cm,height=6.25cm]{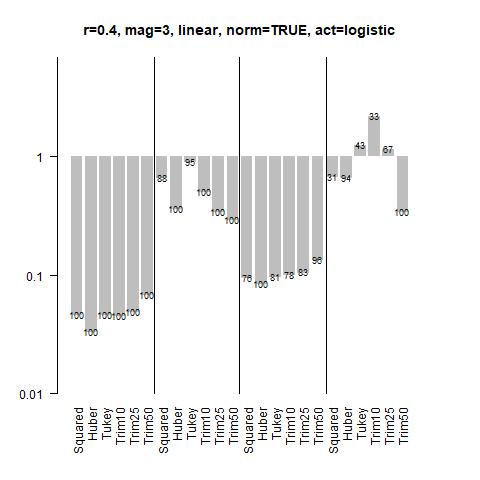} 
\end{center}
\caption{Results for $r=0.4$}
\end{figure}

\subsubsection{Polynomial function}

\begin{figure}[H]
\begin{center}
\includegraphics[width=6.75cm,height=6.25cm]{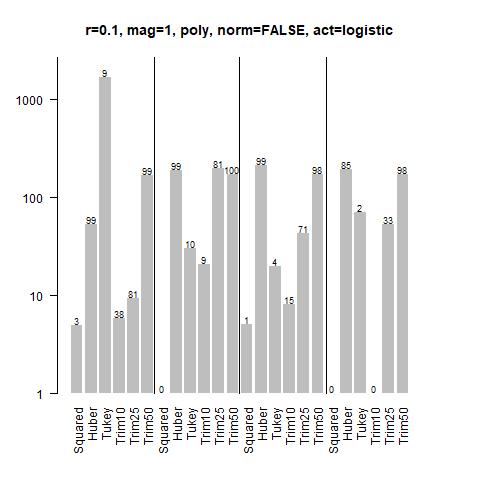}
\includegraphics[width=6.75cm,height=6.25cm]{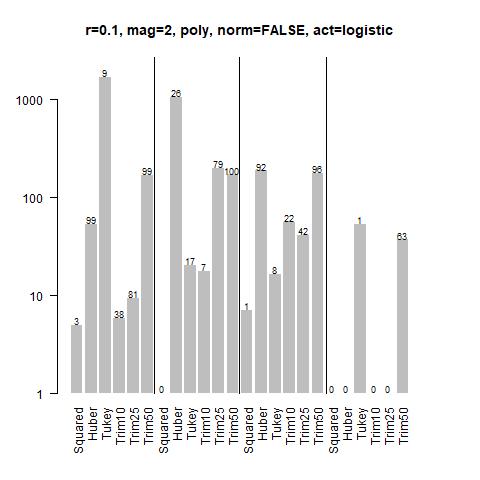} \\
\includegraphics[width=6.75cm,height=6.25cm]{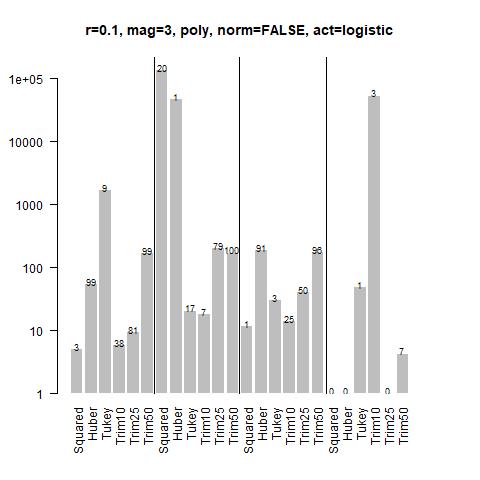} 
\includegraphics[width=6.75cm,height=6.25cm]{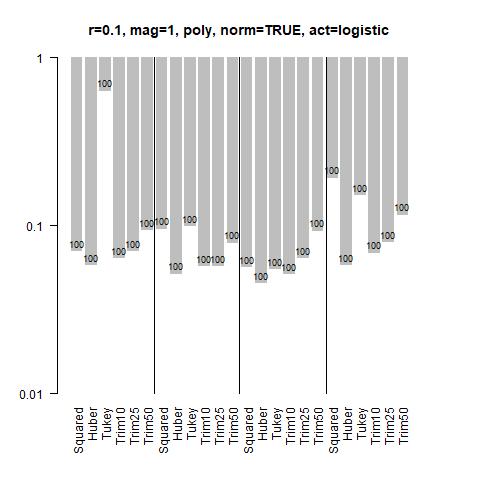}\\
\includegraphics[width=6.75cm,height=6.25cm]{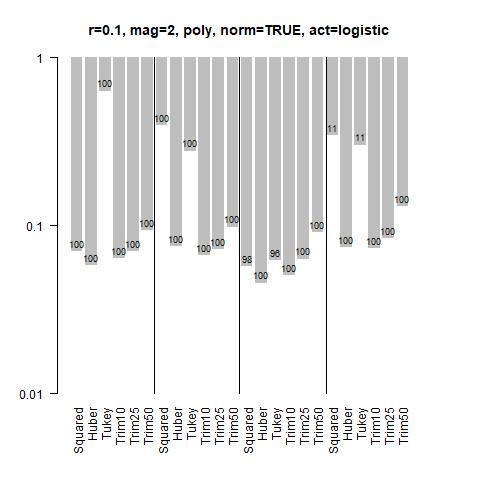} 
\includegraphics[width=6.75cm,height=6.25cm]{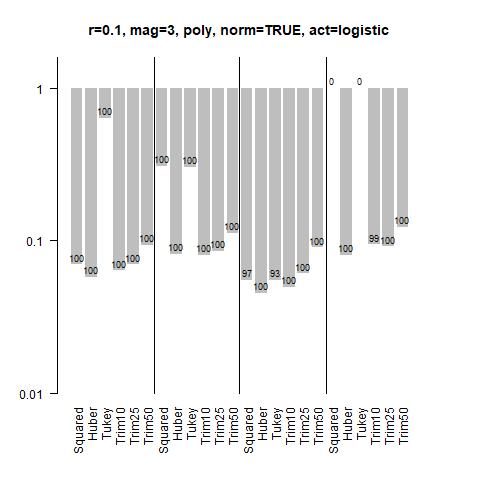} 
\end{center}
\caption{Results for $r=0.1$}\label{trimnn:n500p20r10m1polynonlog}
\end{figure}

\begin{figure}[H]
\begin{center}
\includegraphics[width=6.75cm,height=6.25cm]{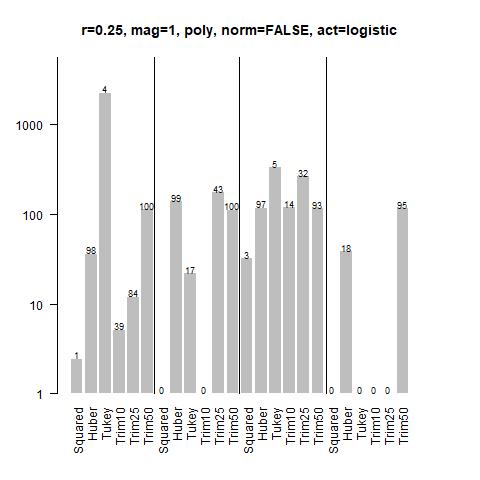}
\includegraphics[width=6.75cm,height=6.25cm]{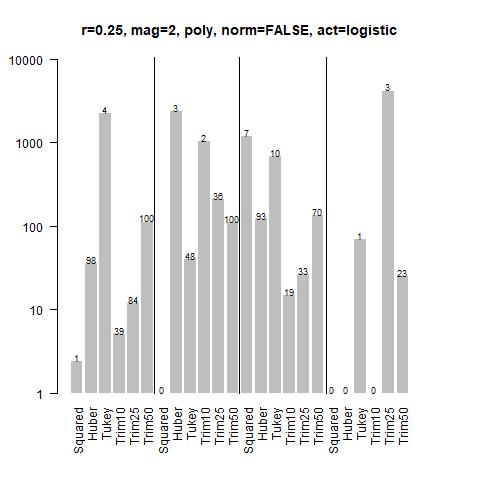} \\
\includegraphics[width=6.75cm,height=6.25cm]{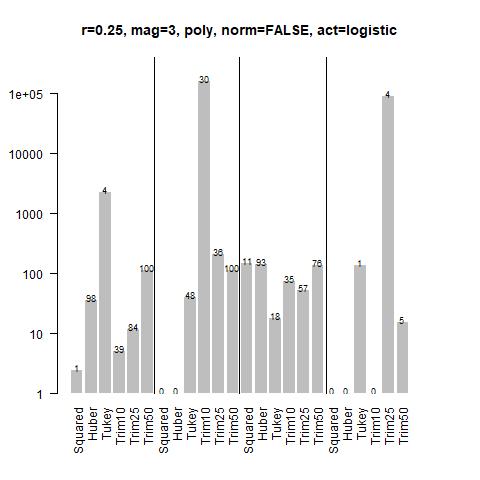} 
\includegraphics[width=6.75cm,height=6.25cm]{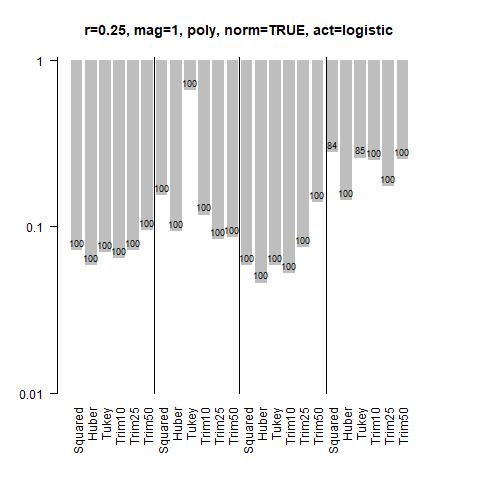}\\
\includegraphics[width=6.75cm,height=6.25cm]{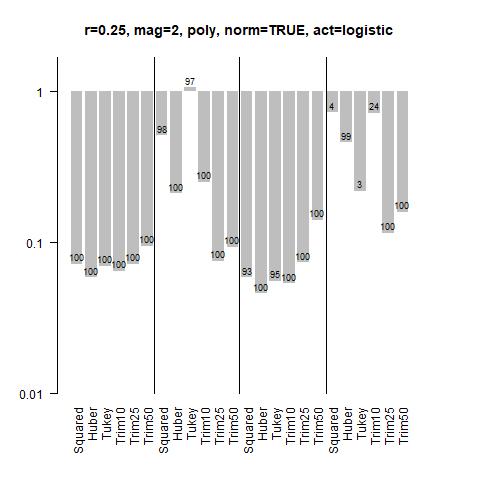} 
\includegraphics[width=6.75cm,height=6.25cm]{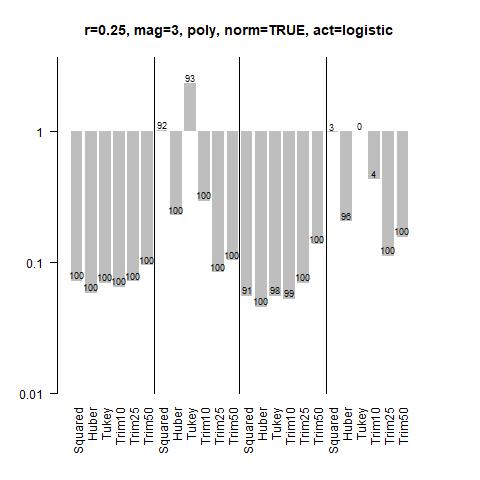} 
\end{center}
\caption{Results for $r=0.25$}\label{trimnn:n500p20r25m1polynonlog}
\end{figure}

\begin{figure}[H]
\begin{center}
\includegraphics[width=6.75cm,height=6.25cm]{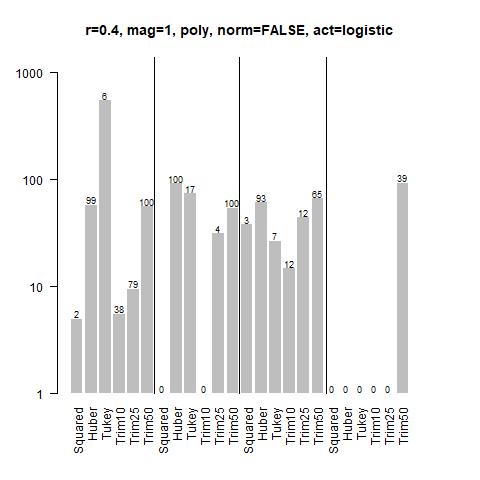}
\includegraphics[width=6.75cm,height=6.25cm]{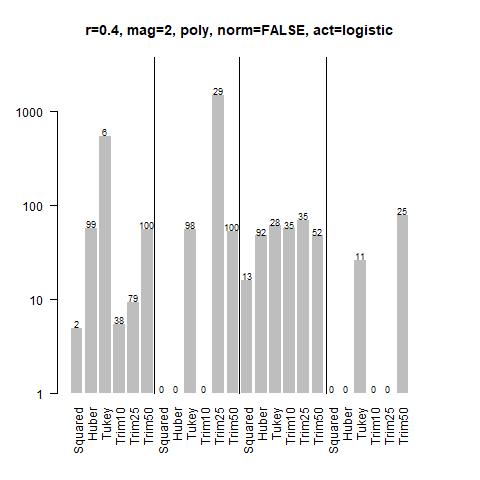} \\
\includegraphics[width=6.75cm,height=6.25cm]{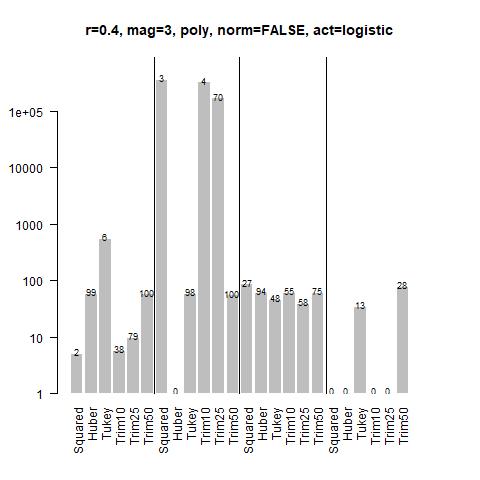} 
\includegraphics[width=6.75cm,height=6.25cm]{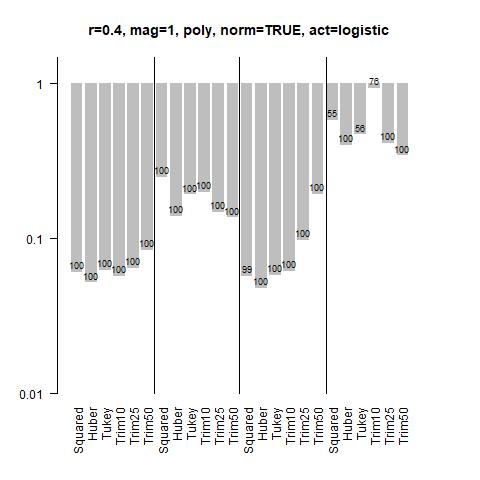}\\
\includegraphics[width=6.75cm,height=6.25cm]{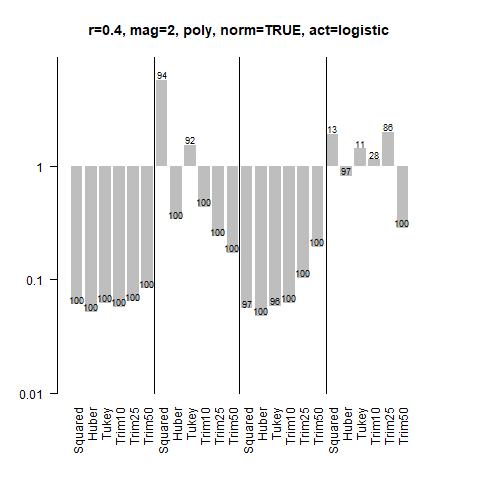} 
\includegraphics[width=6.75cm,height=6.25cm]{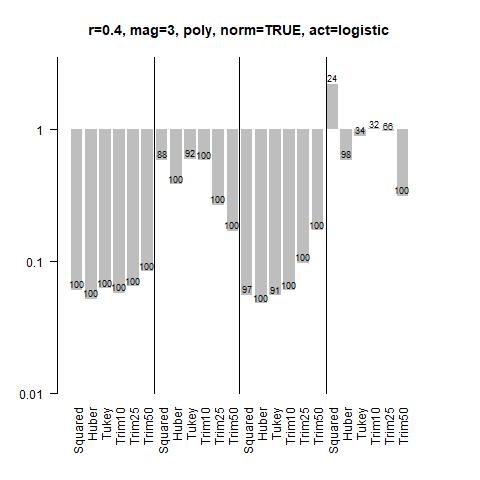} 
\end{center}
\caption{Results for $r=0.4$}\label{trimnn:n500p20r40m1polynonlog}
\end{figure}

\subsubsection{Trigonometric function}

\begin{figure}[H]
\begin{center}
\includegraphics[width=6.75cm,height=6.25cm]{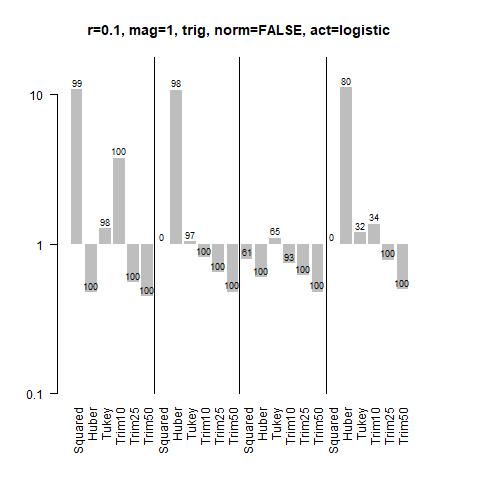}
\includegraphics[width=6.75cm,height=6.25cm]{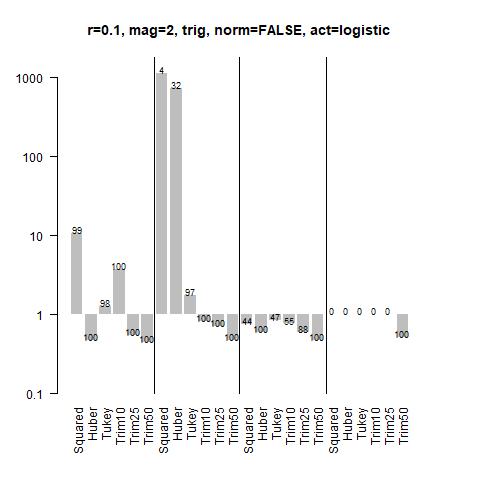} \\
\includegraphics[width=6.75cm,height=6.25cm]{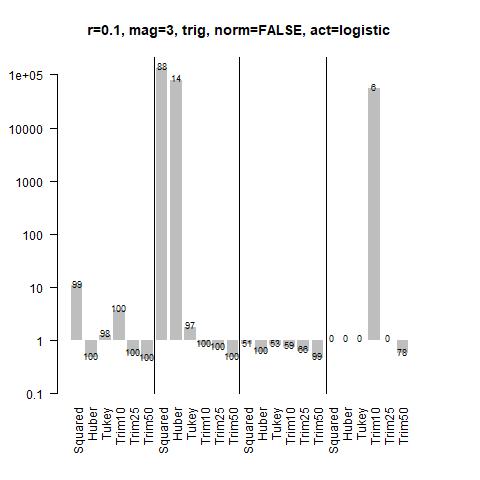} 
\includegraphics[width=6.75cm,height=6.25cm]{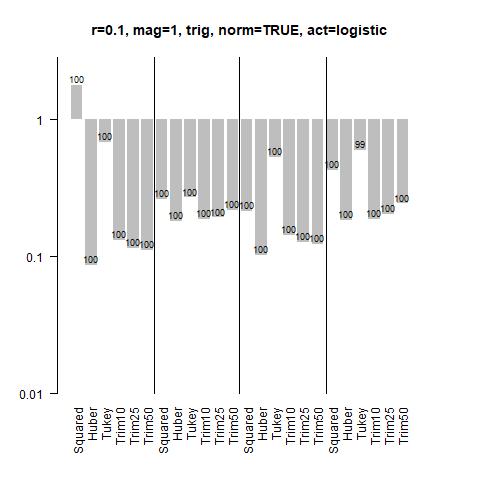}\\
\includegraphics[width=6.75cm,height=6.25cm]{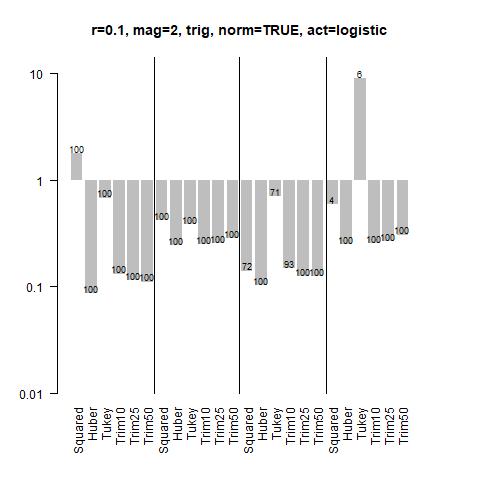} 
\includegraphics[width=6.75cm,height=6.25cm]{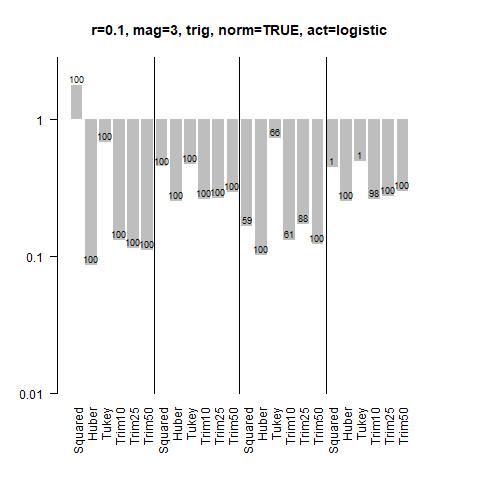} 
\end{center}
\caption{Results for $r=0.1$}\label{trimnn:n500p20r10m1trignonlog}
\end{figure}

\begin{figure}[H]
\begin{center}
\includegraphics[width=6.75cm,height=6.25cm]{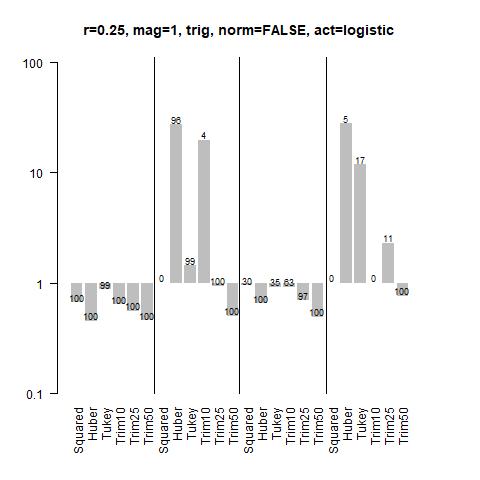}
\includegraphics[width=6.75cm,height=6.25cm]{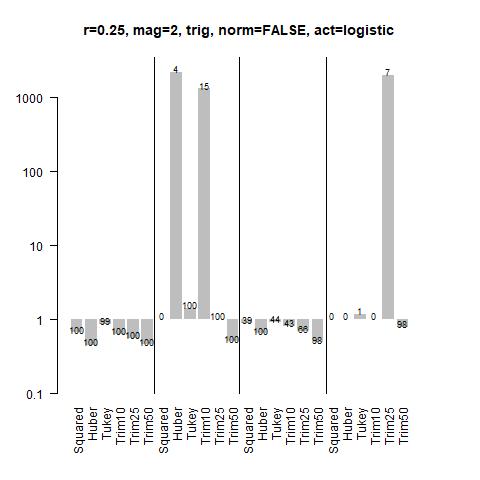} \\
\includegraphics[width=6.75cm,height=6.25cm]{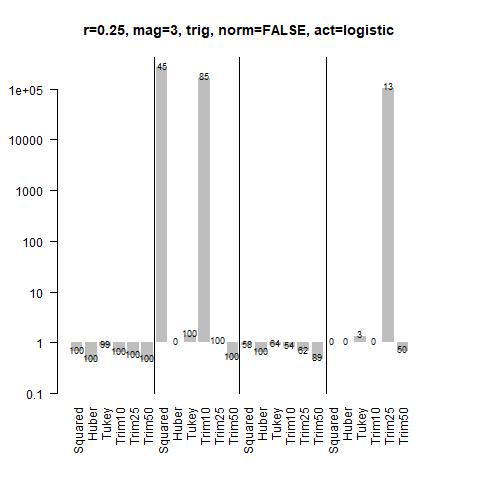} 
\includegraphics[width=6.75cm,height=6.25cm]{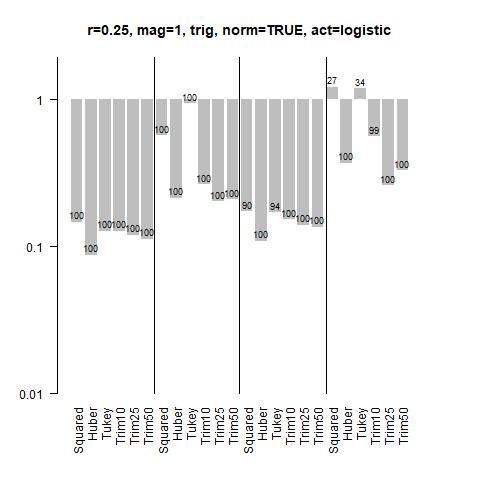}\\
\includegraphics[width=6.75cm,height=6.25cm]{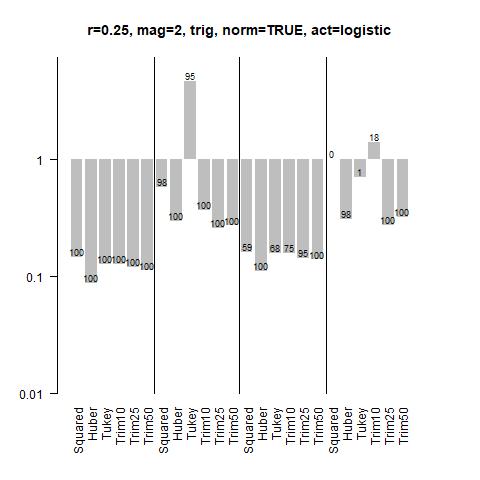} 
\includegraphics[width=6.75cm,height=6.25cm]{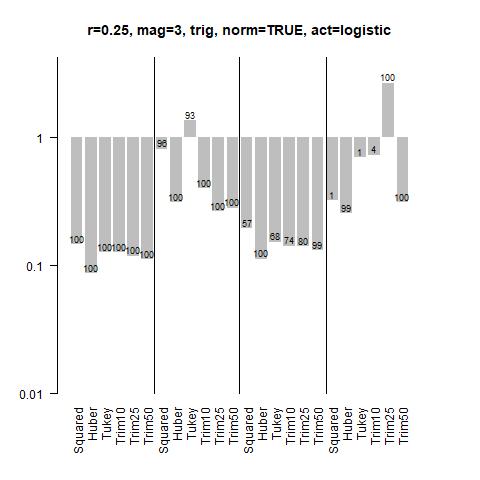} 
\end{center}
\caption{Results for $r=0.25$}\label{trimnn:n500p20r25m1trignonlog}
\end{figure}

\begin{figure}[H]
\begin{center}
\includegraphics[width=6.75cm,height=6.25cm]{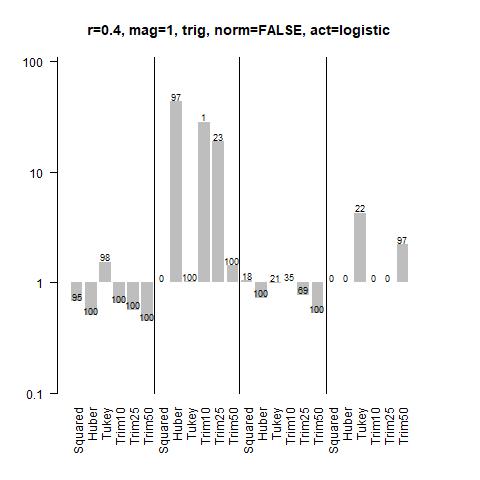}
\includegraphics[width=6.75cm,height=6.25cm]{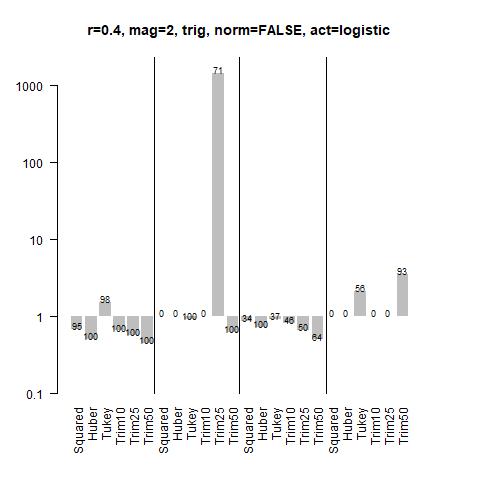} \\
\includegraphics[width=6.75cm,height=6.25cm]{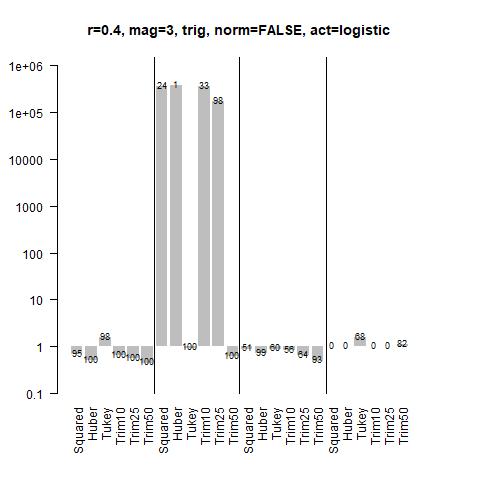} 
\includegraphics[width=6.75cm,height=6.25cm]{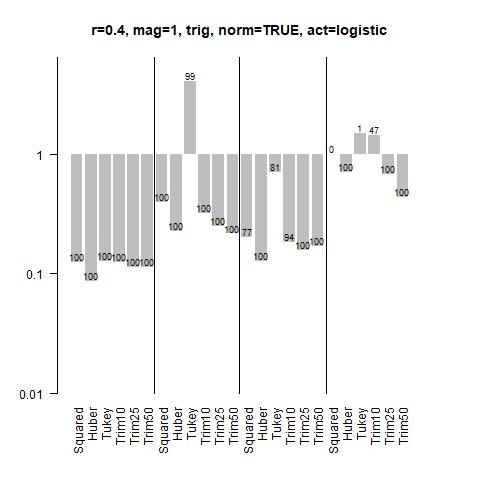}\\
\includegraphics[width=6.75cm,height=6.25cm]{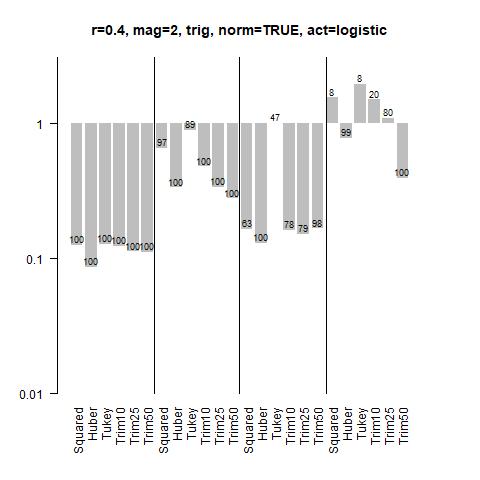} 
\includegraphics[width=6.75cm,height=6.25cm]{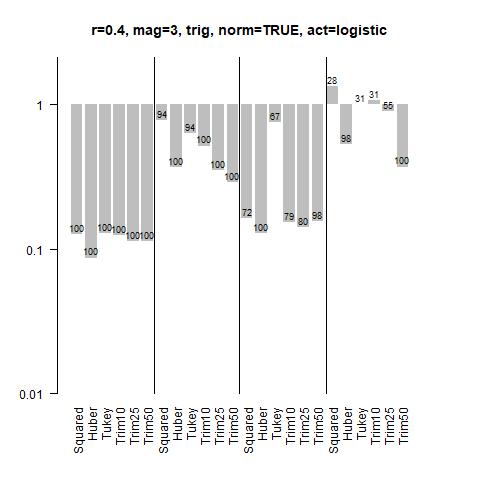} 
\end{center}
\caption{Results for $r=0.4$}\label{trimnn:n500p20r40m1trignonlog}
\end{figure}

\subsection{Softplus activation function}

\subsubsection{Linear function}

\begin{figure}[H]
\label{trimnn:n500p20r10m1linnonrelu}
\begin{center}
\includegraphics[width=6.75cm,height=6.25cm]{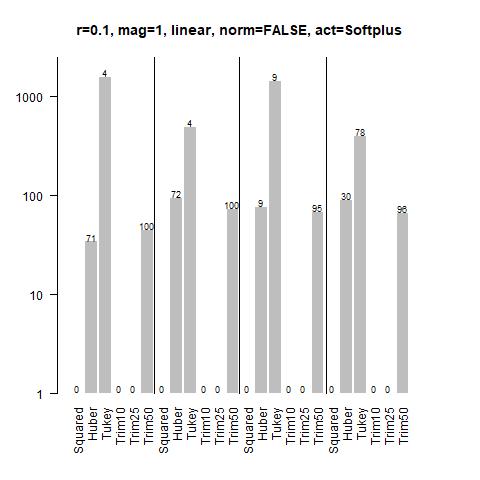}
\includegraphics[width=6.75cm,height=6.25cm]{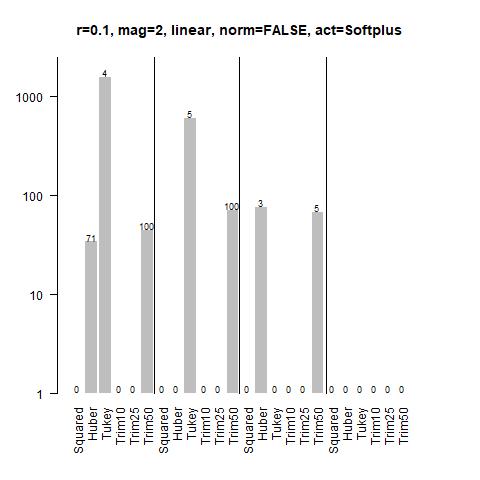} \\
\includegraphics[width=6.75cm,height=6.25cm]{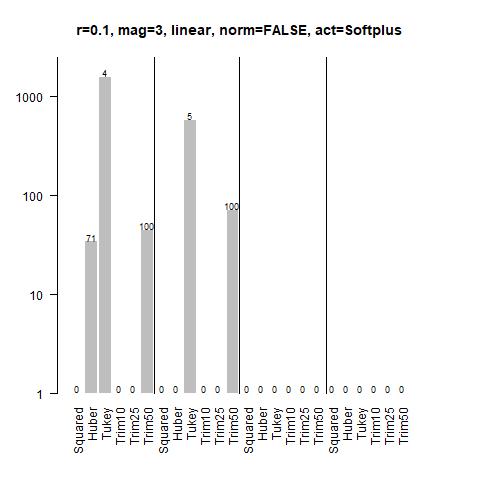} 
\includegraphics[width=6.75cm,height=6.25cm]{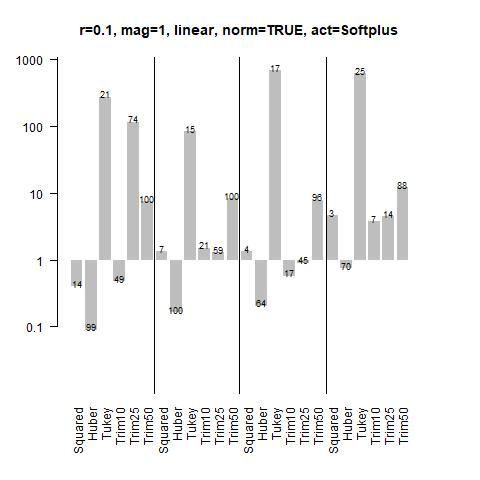}\\
\includegraphics[width=6.75cm,height=6.25cm]{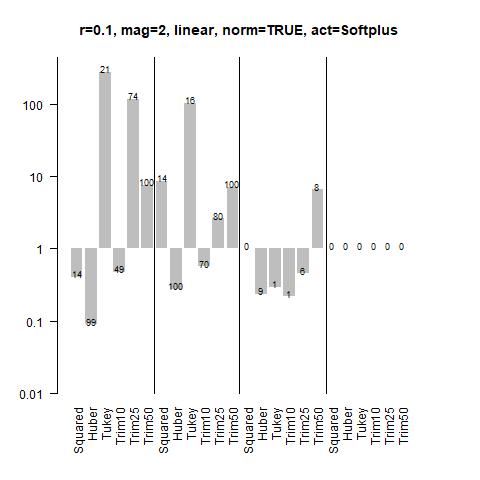} 
\includegraphics[width=6.75cm,height=6.25cm]{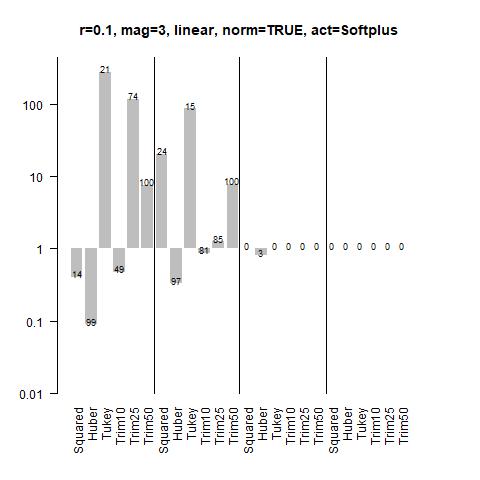} 
\end{center}
\caption{Results for $r=0.1$}
\end{figure}

\begin{figure}[H]
\label{trimnn:n500p20r25m1linnonrelu}
\begin{center}
\includegraphics[width=6.75cm,height=6.25cm]{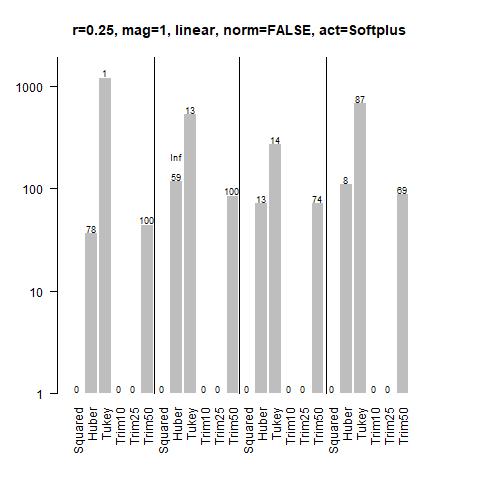}
\includegraphics[width=6.75cm,height=6.25cm]{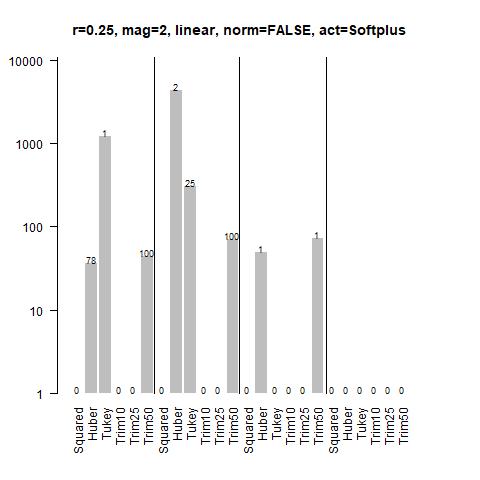} \\
\includegraphics[width=6.75cm,height=6.25cm]{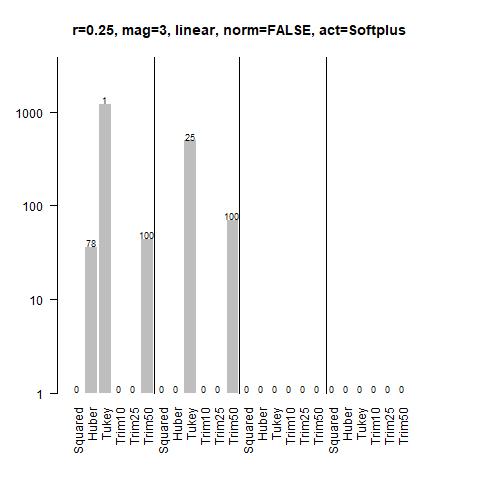} 
\includegraphics[width=6.75cm,height=6.25cm]{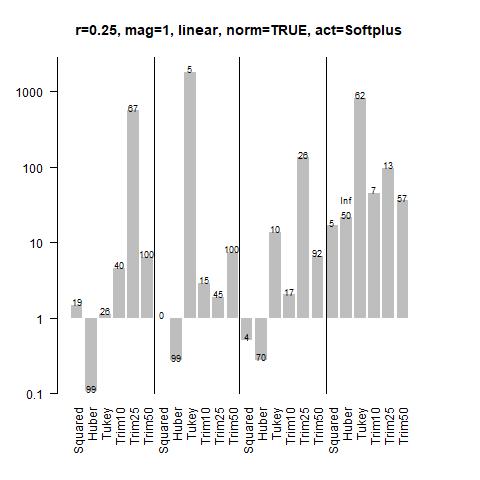}\\
\includegraphics[width=6.75cm,height=6.25cm]{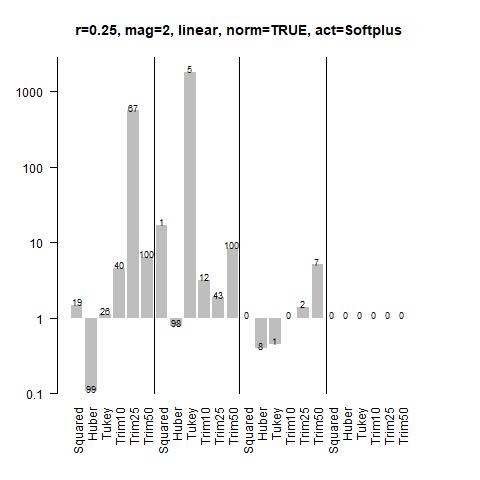} 
\includegraphics[width=6.75cm,height=6.25cm]{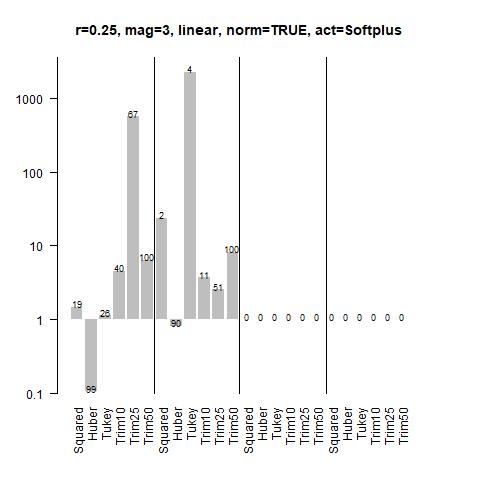} 
\end{center}
\caption{Results for $r=0.25$}
\end{figure}

\begin{figure}[H]
\label{trimnn:n500p20r40m1linnonrelu}
\begin{center}
\includegraphics[width=6.75cm,height=6.25cm]{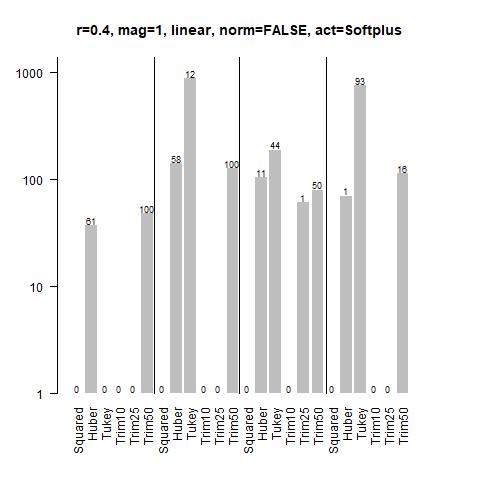}
\includegraphics[width=6.75cm,height=6.25cm]{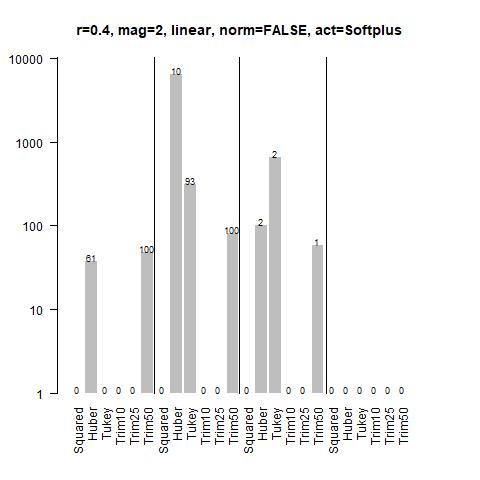} \\
\includegraphics[width=6.75cm,height=6.25cm]{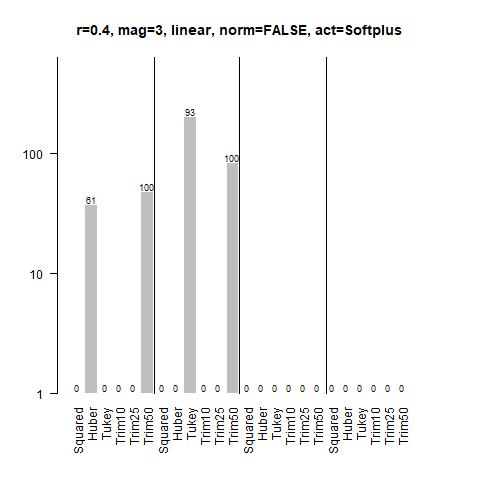} 
\includegraphics[width=6.75cm,height=6.25cm]{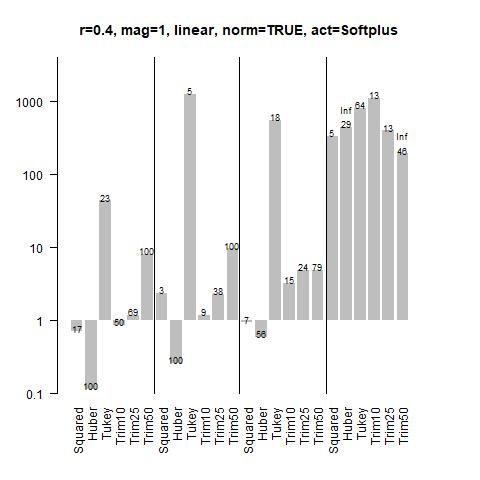}\\
\includegraphics[width=6.75cm,height=6.25cm]{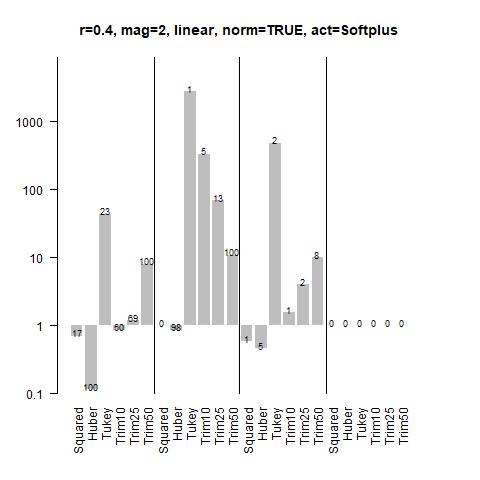} 
\includegraphics[width=6.75cm,height=6.25cm]{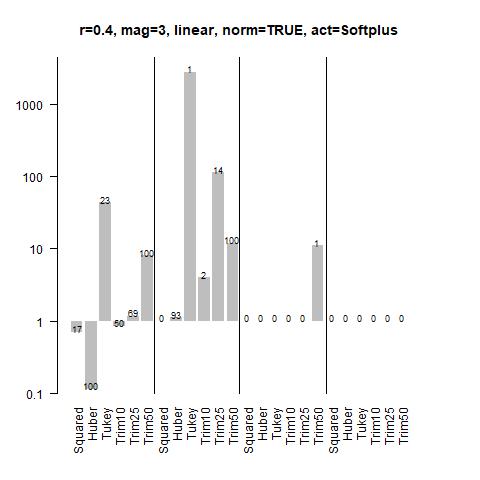} 
\end{center}
\caption{Results for $r=0.4$}
\end{figure}

\subsubsection{Polynomial function}

\begin{figure}[H]
\label{trimnn:n500p20r10m1polynonrelu}
\begin{center}
\includegraphics[width=6.75cm,height=6.25cm]{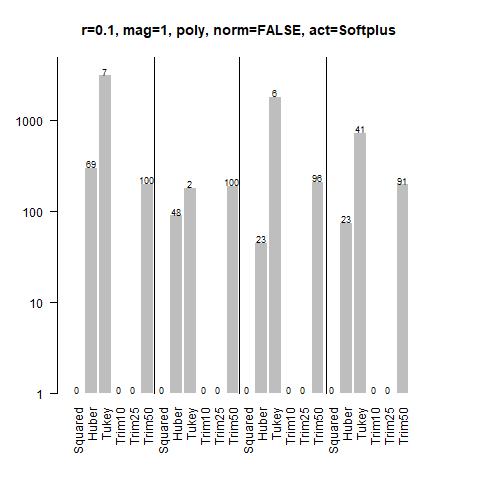}
\includegraphics[width=6.75cm,height=6.25cm]{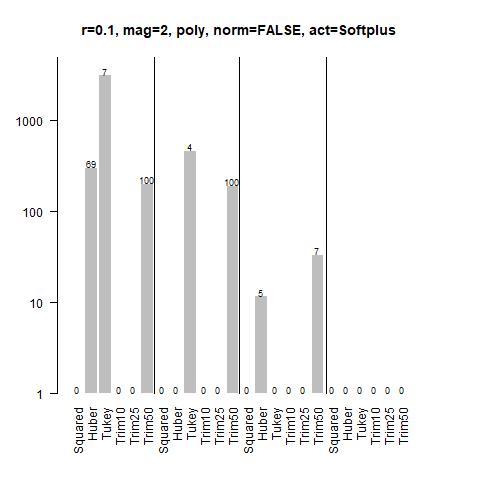} \\
\includegraphics[width=6.75cm,height=6.25cm]{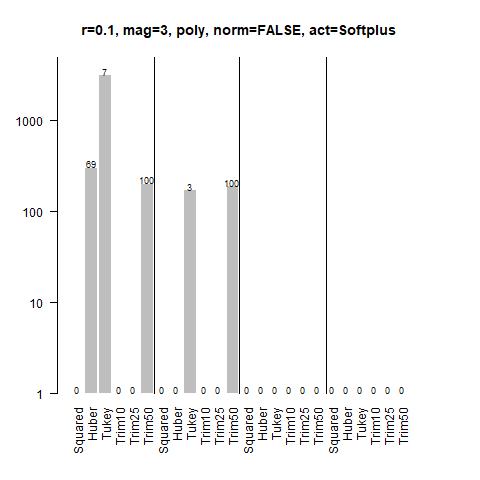} 
\includegraphics[width=6.75cm,height=6.25cm]{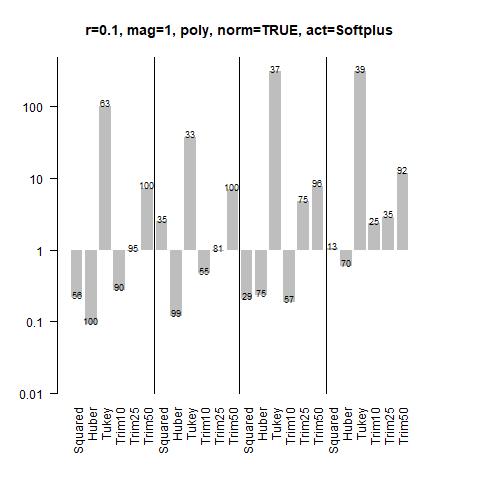}\\
\includegraphics[width=6.75cm,height=6.25cm]{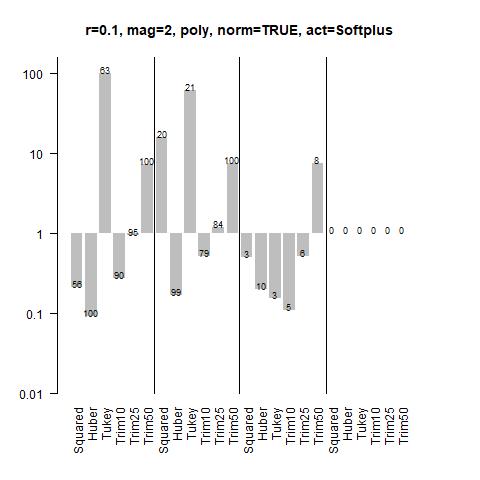} 
\includegraphics[width=6.75cm,height=6.25cm]{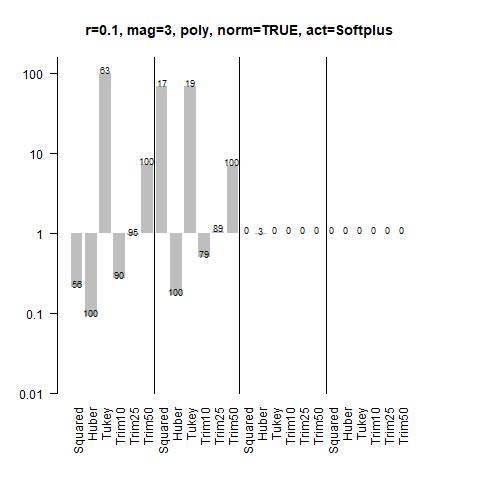} 
\end{center}
\caption{Results for $r=0.1$}
\end{figure}

\begin{figure}[H]
\label{trimnn:n500p20r25m1polynonrelu}
\begin{center}
\includegraphics[width=6.75cm,height=6.25cm]{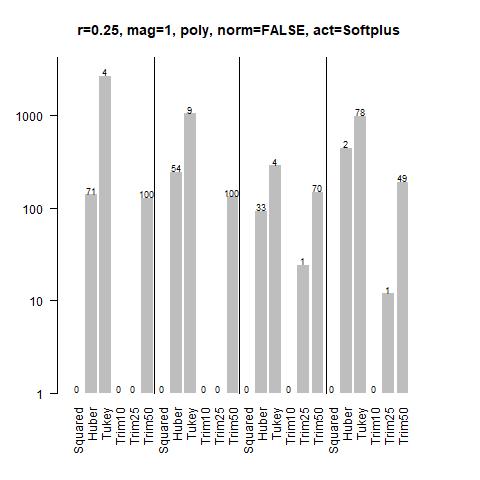}
\includegraphics[width=6.75cm,height=6.25cm]{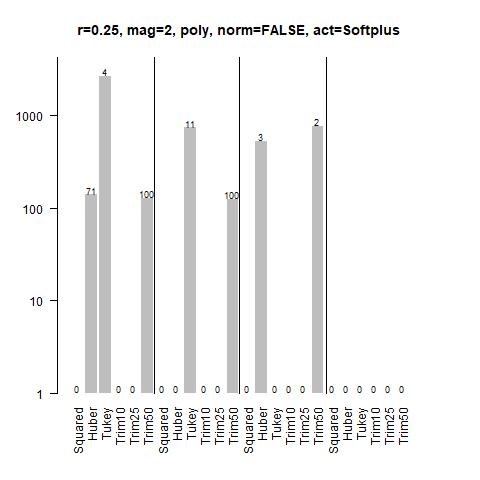} \\
\includegraphics[width=6.75cm,height=6.25cm]{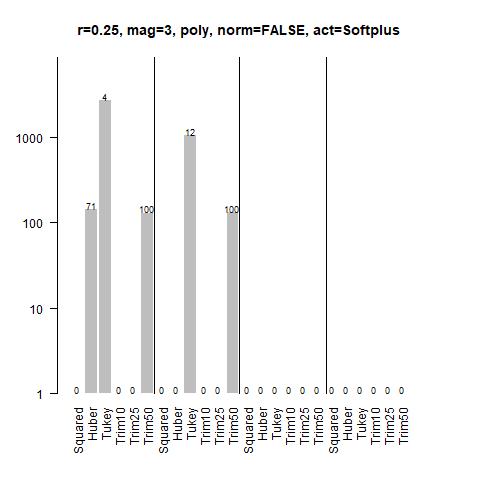} 
\includegraphics[width=6.75cm,height=6.25cm]{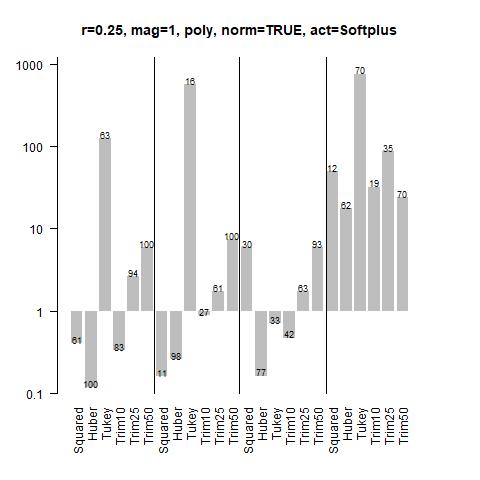}\\
\includegraphics[width=6.75cm,height=6.25cm]{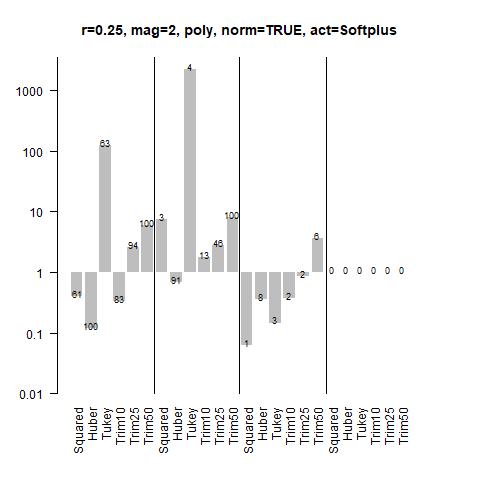} 
\includegraphics[width=6.75cm,height=6.25cm]{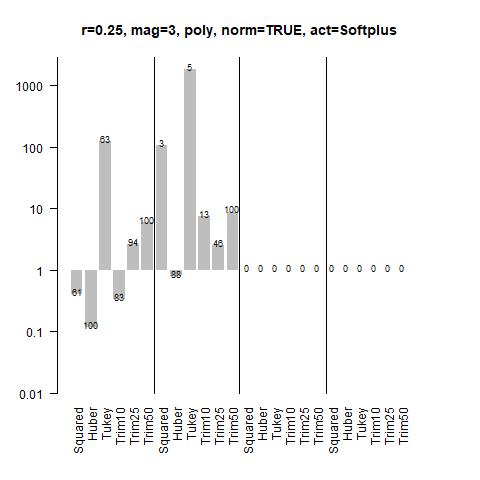} 
\end{center}
\caption{Results for $r=0.25$}
\end{figure}

\begin{figure}[H]
\label{trimnn:n500p20r40m1polynonrelu}
\begin{center}
\includegraphics[width=6.75cm,height=6.25cm]{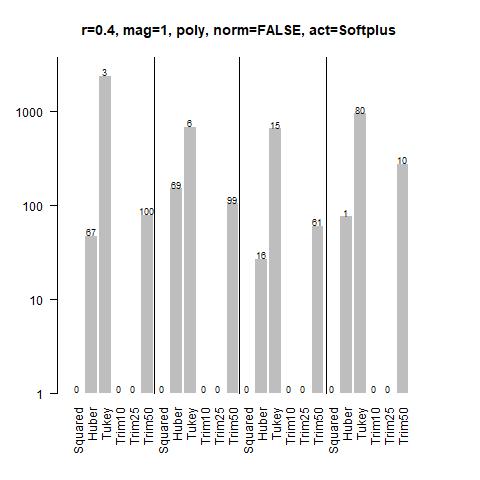}
\includegraphics[width=6.75cm,height=6.25cm]{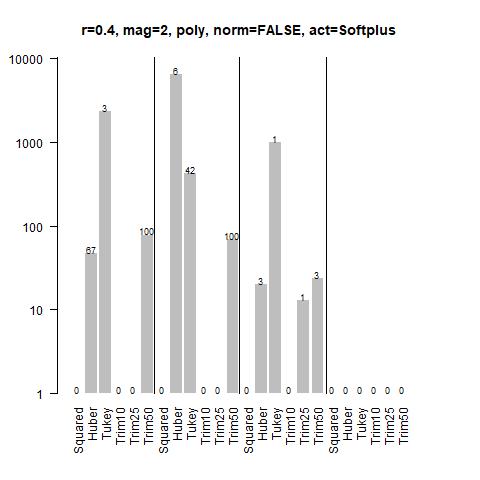} \\
\includegraphics[width=6.75cm,height=6.25cm]{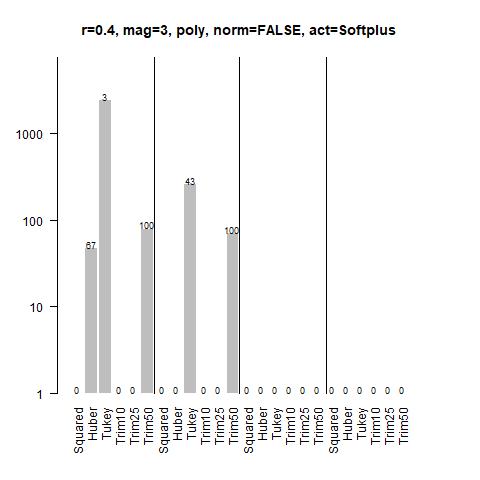} 
\includegraphics[width=6.75cm,height=6.25cm]{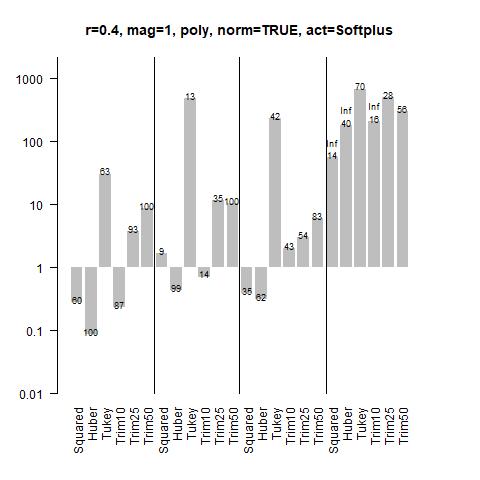}\\
\includegraphics[width=6.75cm,height=6.25cm]{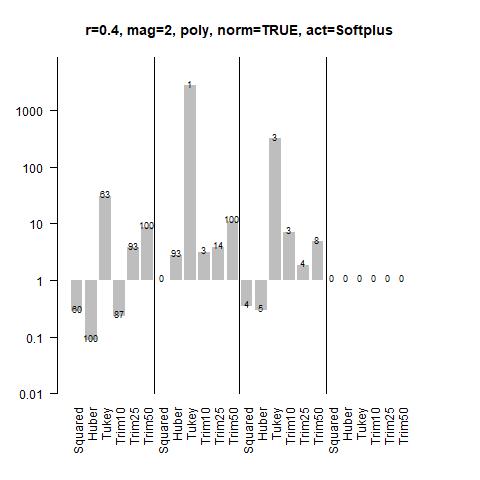} 
\includegraphics[width=6.75cm,height=6.25cm]{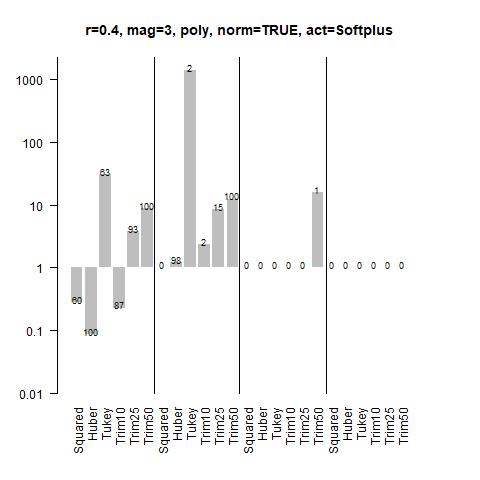} 
\end{center}
\caption{Results for $r=0.4$}
\end{figure}

\subsubsection{Trigonometric function}

\begin{figure}[H]
\label{trimnn:n500p20r10m1trignonrelu}
\begin{center}
\includegraphics[width=6.75cm,height=6.25cm]{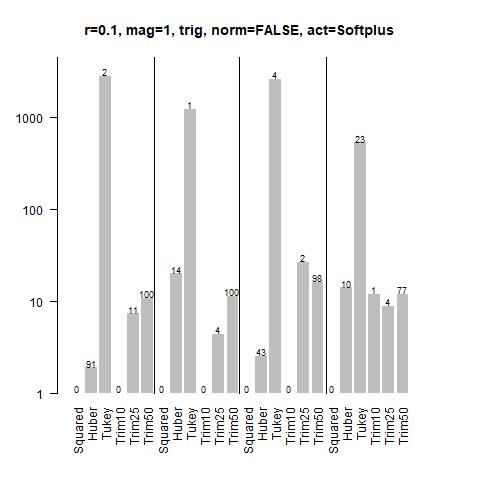}
\includegraphics[width=6.75cm,height=6.25cm]{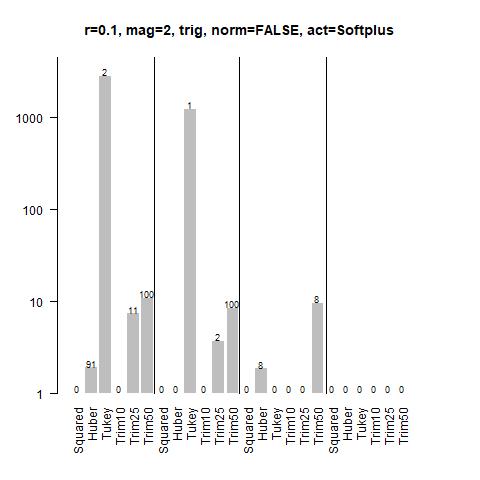} \\
\includegraphics[width=6.75cm,height=6.25cm]{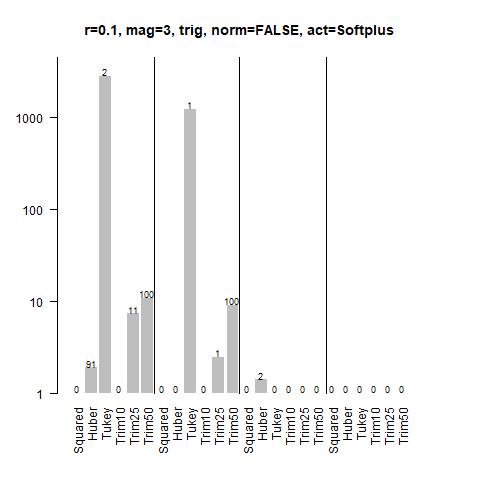} 
\includegraphics[width=6.75cm,height=6.25cm]{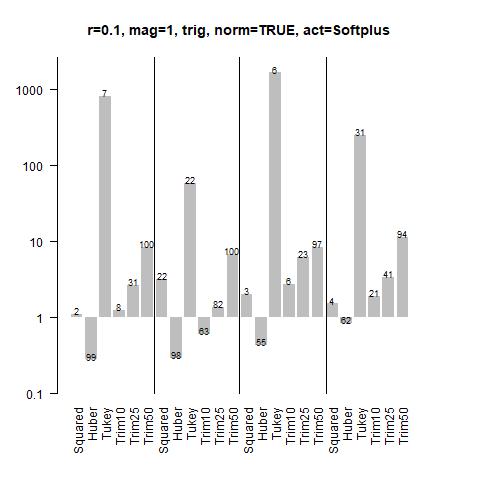}\\
\includegraphics[width=6.75cm,height=6.25cm]{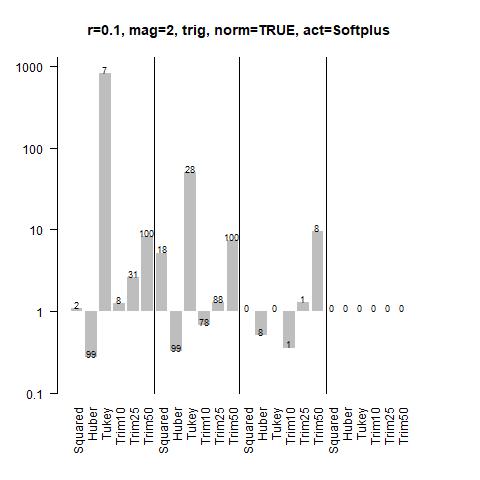} 
\includegraphics[width=6.75cm,height=6.25cm]{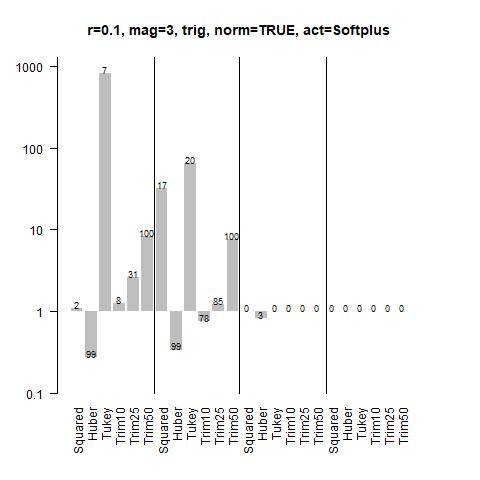} 
\end{center}
\caption{Results for $r=0.1$}
\end{figure}

\begin{figure}[H]
\label{trimnn:n500p20r25m1trignonrelu}
\begin{center}
\includegraphics[width=6.75cm,height=6.25cm]{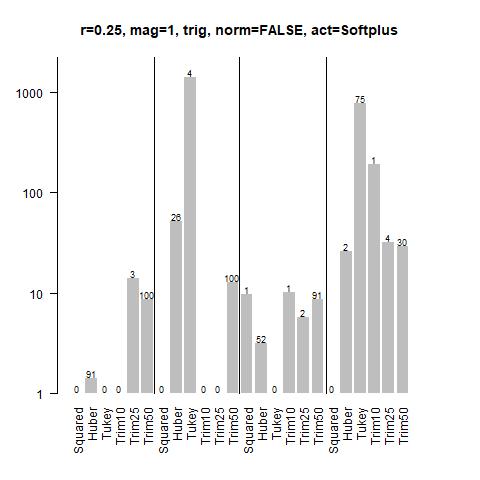}
\includegraphics[width=6.75cm,height=6.25cm]{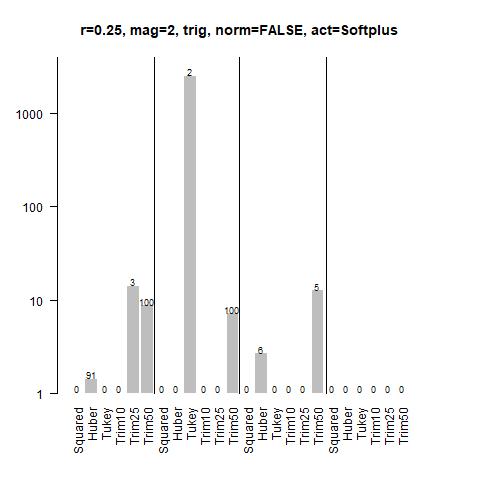} \\
\includegraphics[width=6.75cm,height=6.25cm]{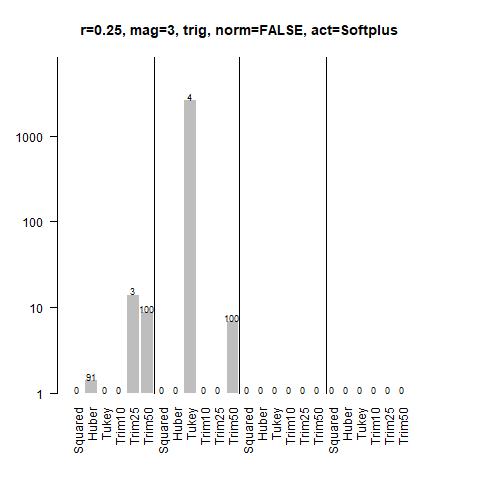} 
\includegraphics[width=6.75cm,height=6.25cm]{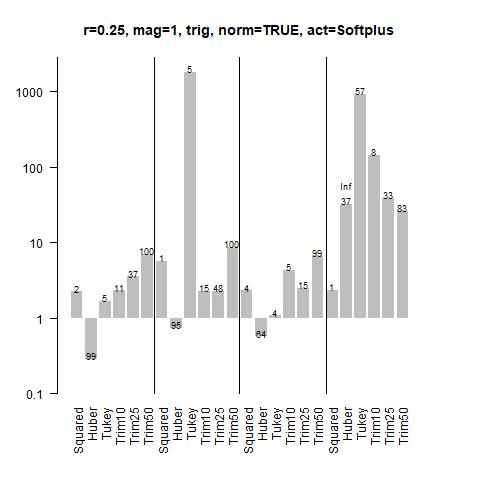}\\
\includegraphics[width=6.75cm,height=6.25cm]{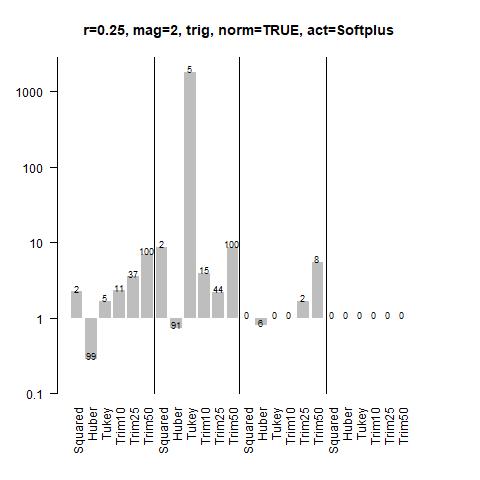} 
\includegraphics[width=6.75cm,height=6.25cm]{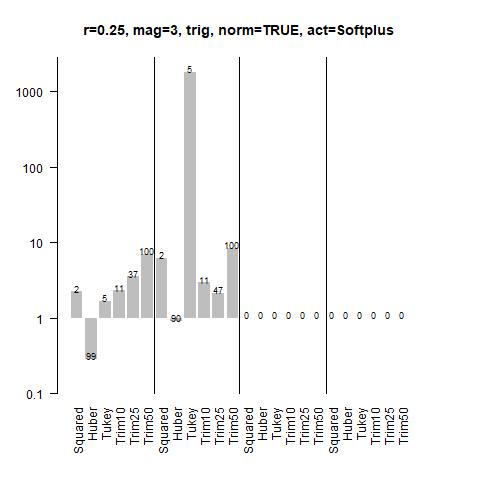} 
\end{center}
\caption{Results for $r=0.25$}
\end{figure}

\begin{figure}[H]
\label{trimnn:n500p20r40m1trignonrelu}
\begin{center}
\includegraphics[width=6.75cm,height=6.25cm]{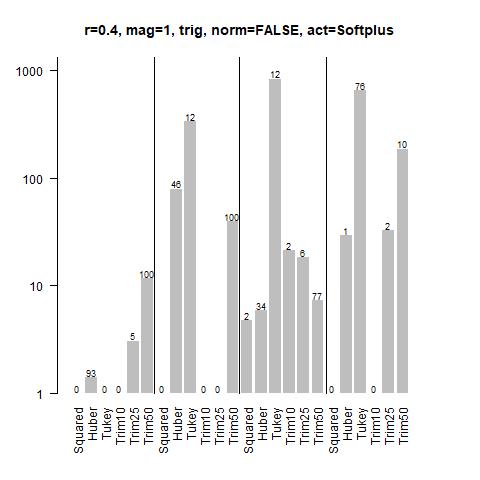}
\includegraphics[width=6.75cm,height=6.25cm]{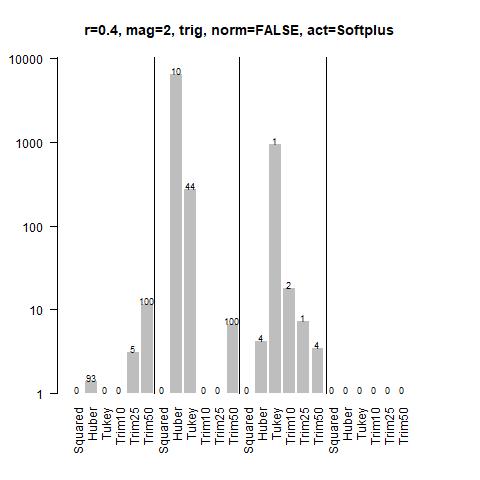} \\
\includegraphics[width=6.75cm,height=6.25cm]{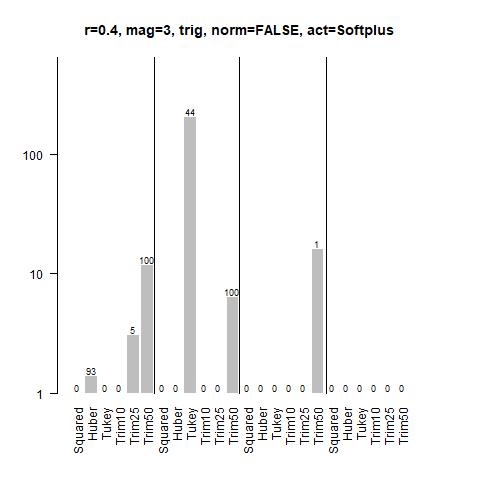} 
\includegraphics[width=6.75cm,height=6.25cm]{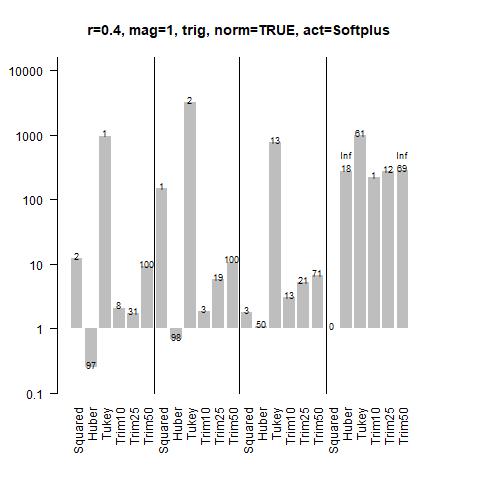}\\
\includegraphics[width=6.75cm,height=6.25cm]{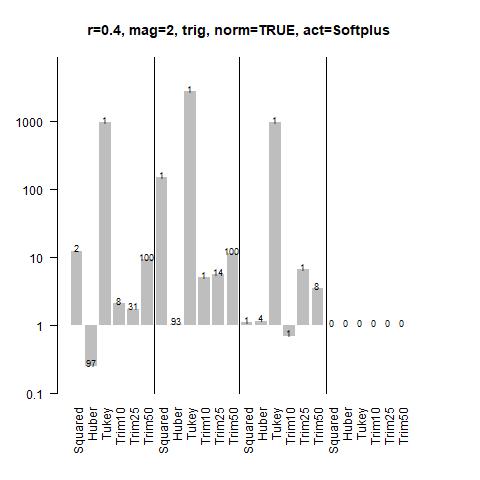} 
\includegraphics[width=6.75cm,height=6.25cm]{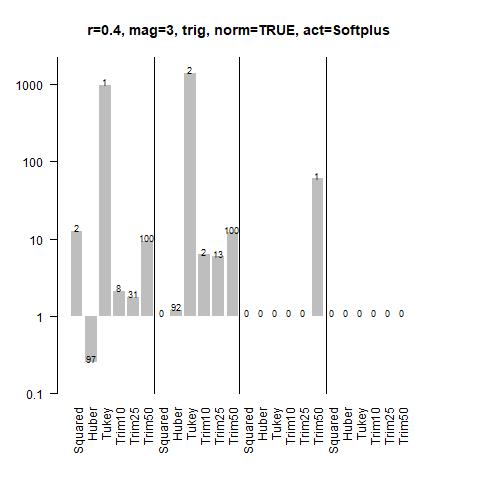} 
\end{center}
\caption{Results for $r=0.4$}
\end{figure}

\section{Simulation results for $n=500$ and $p=20$, deep network: Test loss} \label{trimnn:secloss50020deep}

\subsection{Logistic activation function}

\subsubsection{Linear function}

\begin{figure}[H]
\label{trimnn:n500p20r10m1linnonlogdeep}
\begin{center}
\includegraphics[width=6.75cm,height=6.25cm]{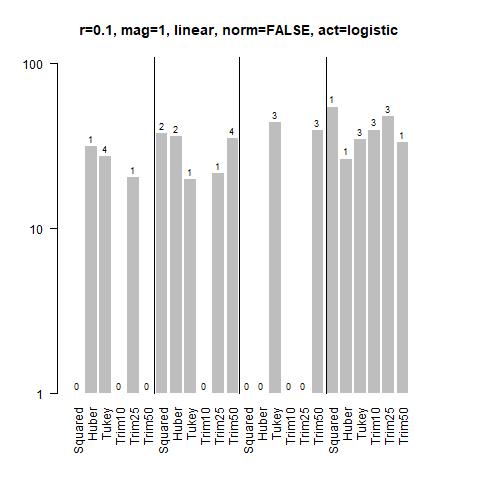}
\includegraphics[width=6.75cm,height=6.25cm]{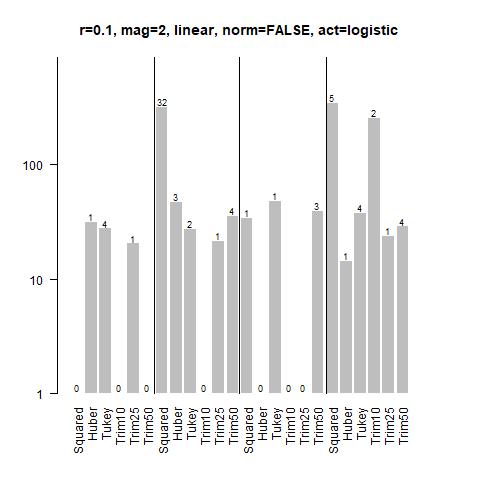} \\
\includegraphics[width=6.75cm,height=6.25cm]{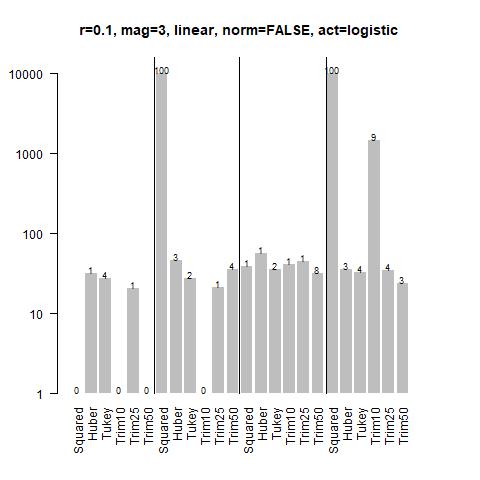} 
\includegraphics[width=6.75cm,height=6.25cm]{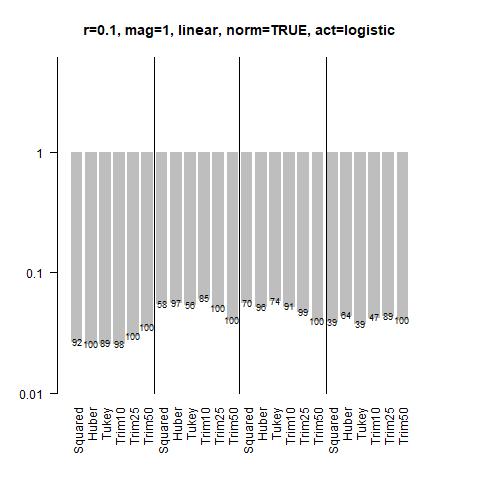}\\
\includegraphics[width=6.75cm,height=6.25cm]{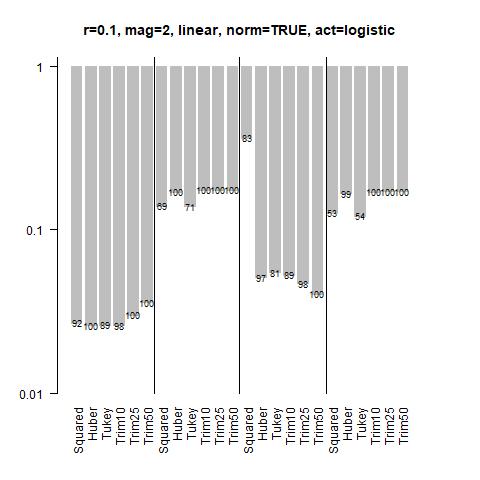} 
\includegraphics[width=6.75cm,height=6.25cm]{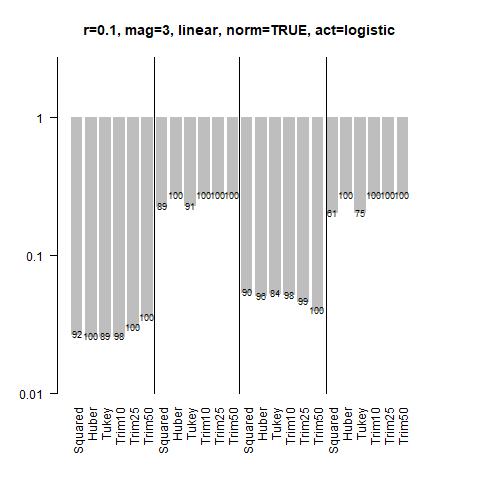} 
\end{center}
\caption{Results for $r=0.1$}
\end{figure}

\begin{figure}[H]
\label{trimnn:n500p20r25m1linnonlogdeep}
\begin{center}
\includegraphics[width=6.75cm,height=6.25cm]{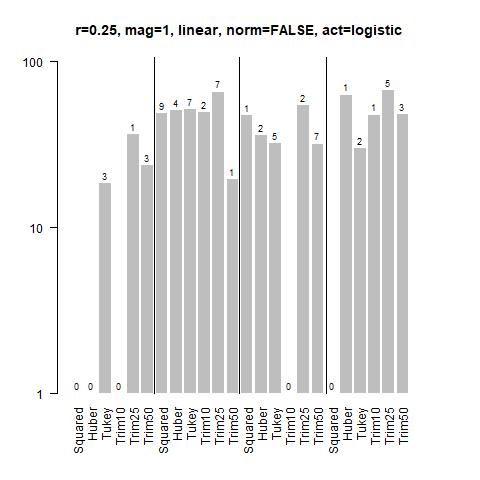}
\includegraphics[width=6.75cm,height=6.25cm]{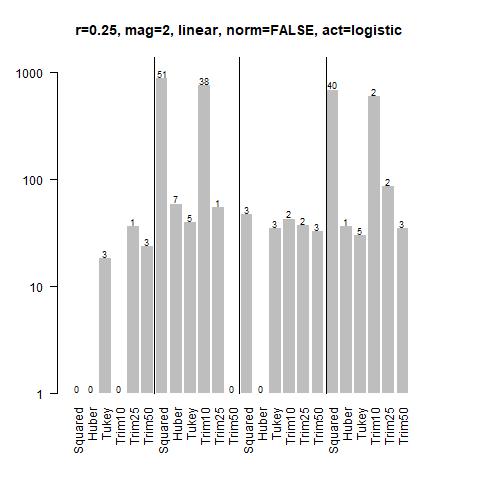} \\
\includegraphics[width=6.75cm,height=6.25cm]{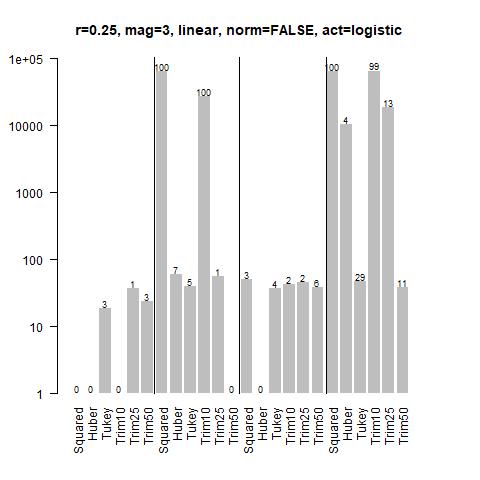} 
\includegraphics[width=6.75cm,height=6.25cm]{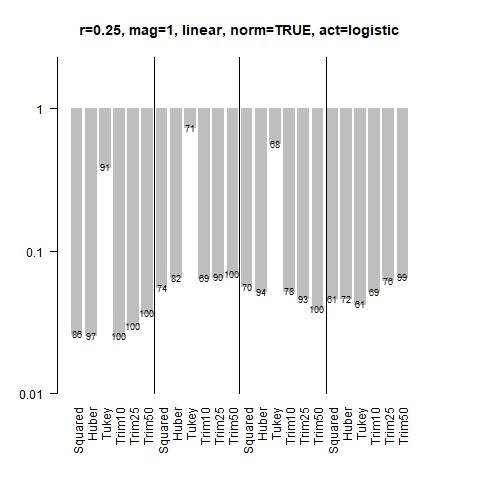}\\
\includegraphics[width=6.75cm,height=6.25cm]{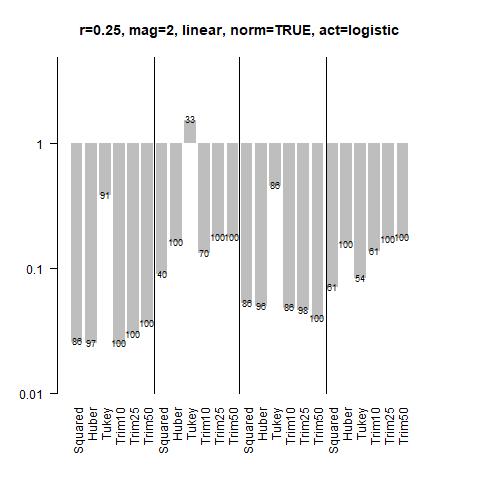} 
\includegraphics[width=6.75cm,height=6.25cm]{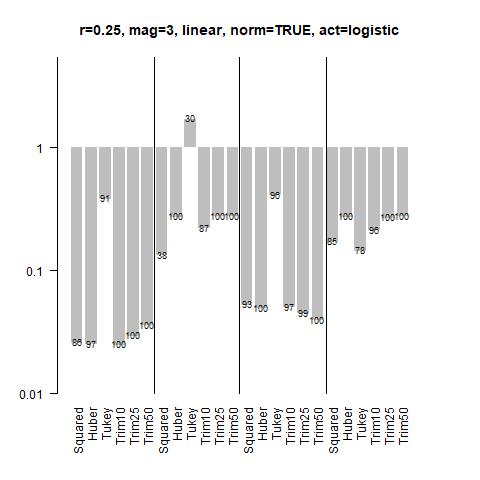} 
\end{center}
\caption{Results for $r=0.25$}
\end{figure}

\begin{figure}[H]
\label{trimnn:n500p20r40m1linnonlogdeep}
\begin{center}
\includegraphics[width=6.75cm,height=6.25cm]{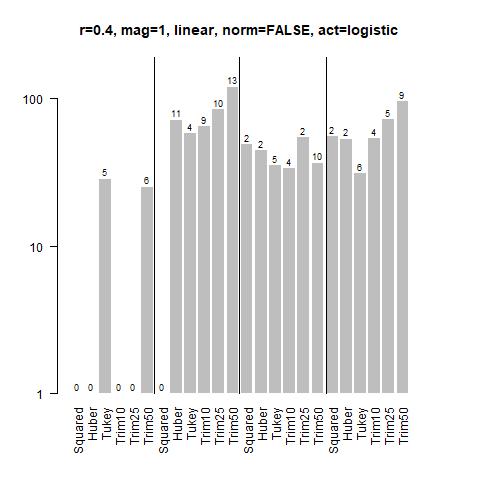}
\includegraphics[width=6.75cm,height=6.25cm]{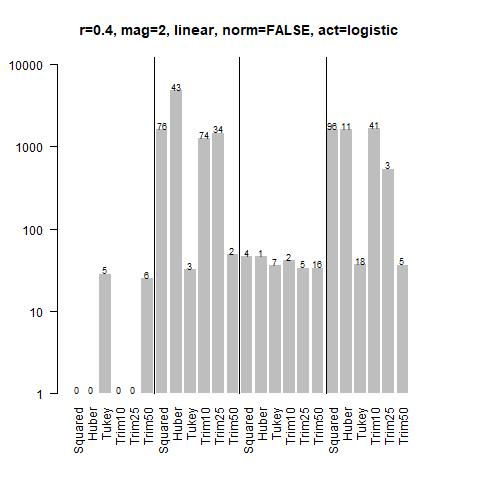} \\
\includegraphics[width=6.75cm,height=6.25cm]{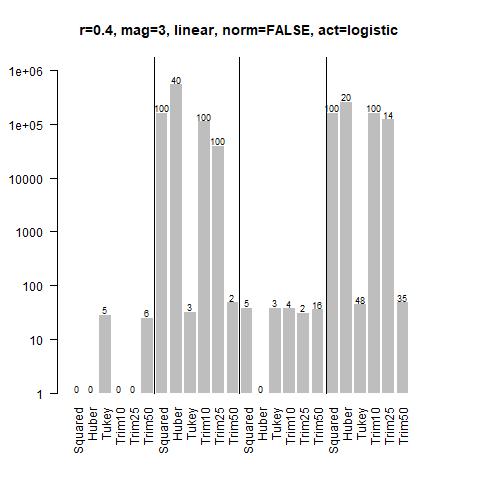} 
\includegraphics[width=6.75cm,height=6.25cm]{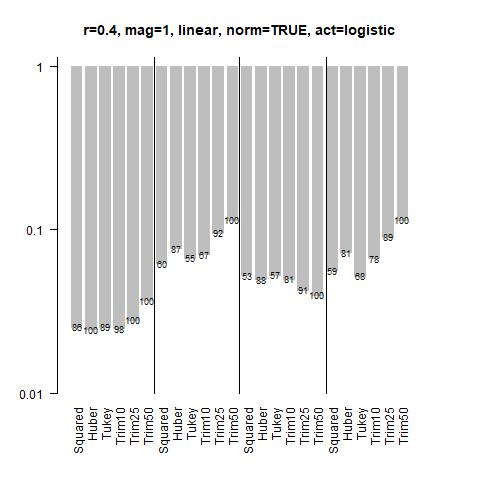}\\
\includegraphics[width=6.75cm,height=6.25cm]{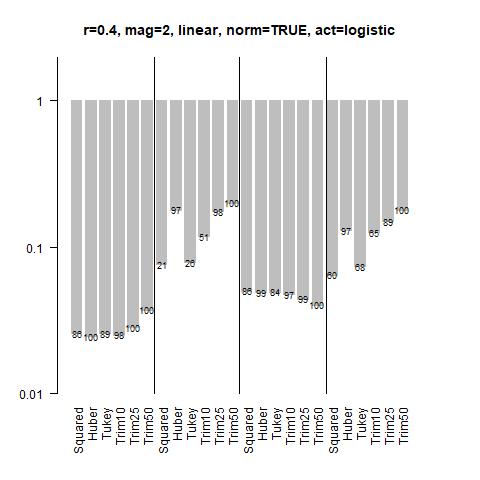} 
\includegraphics[width=6.75cm,height=6.25cm]{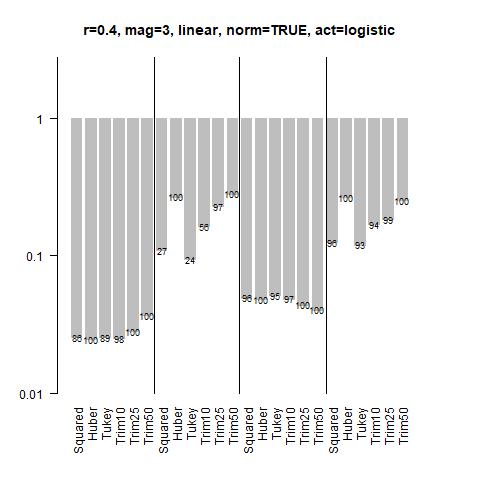} 
\end{center}
\caption{Results for $r=0.4$}
\end{figure}

\subsubsection{Polynomial function}

\begin{figure}[H]
\begin{center}
\includegraphics[width=6.75cm,height=6.25cm]{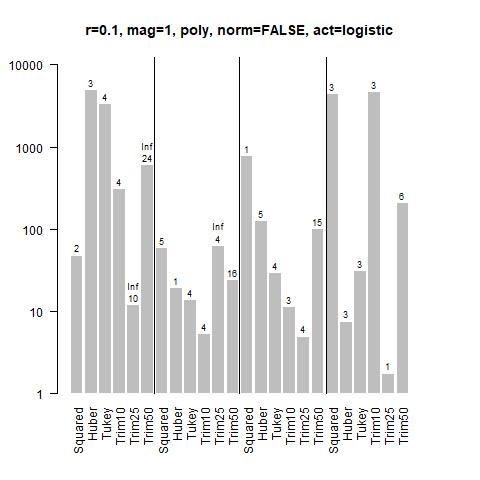}
\includegraphics[width=6.75cm,height=6.25cm]{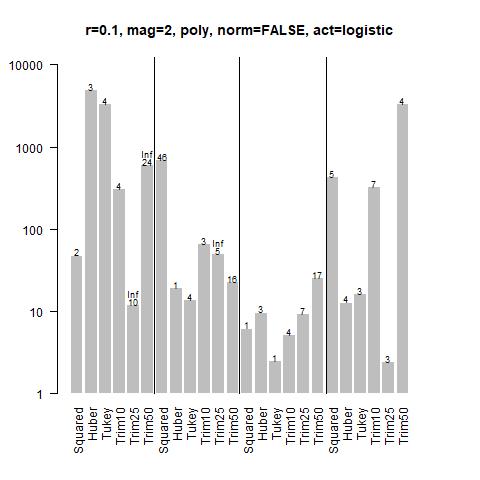} \\
\includegraphics[width=6.75cm,height=6.25cm]{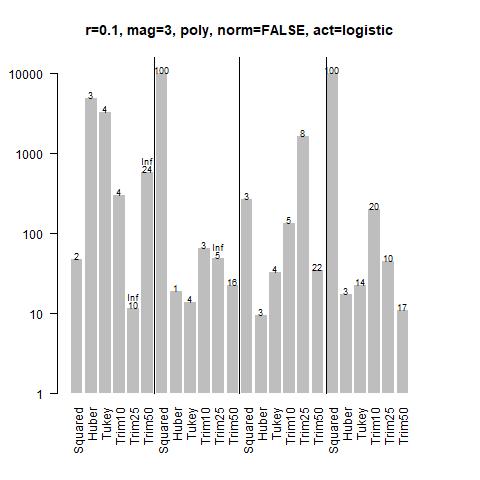} 
\includegraphics[width=6.75cm,height=6.25cm]{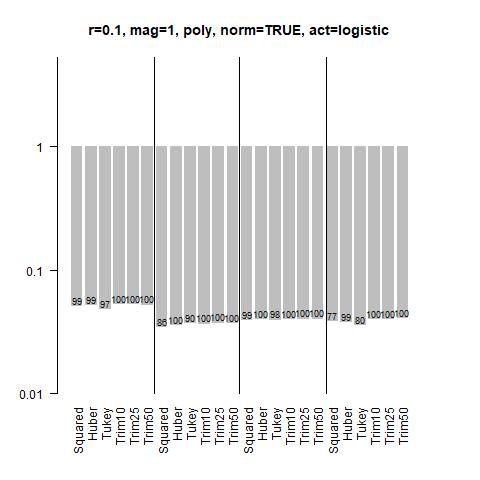}\\
\includegraphics[width=6.75cm,height=6.25cm]{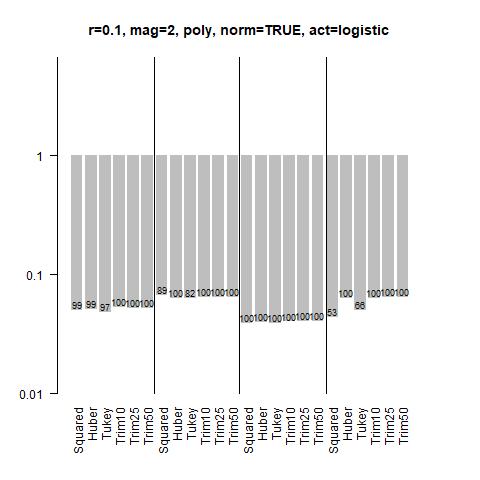} 
\includegraphics[width=6.75cm,height=6.25cm]{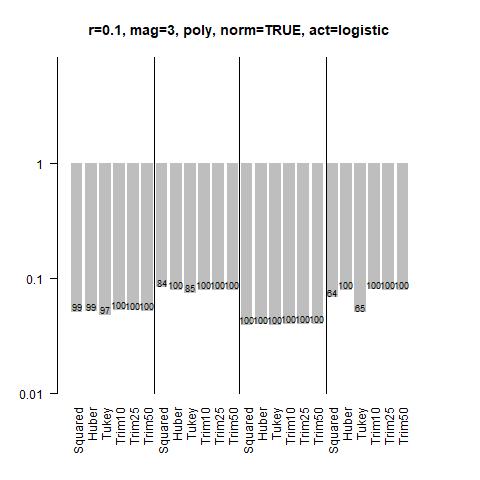} 
\end{center}
\caption{Results for $r=0.1$}\label{trimnn:n500p20r10m1polynonlogdeep}
\end{figure}

\begin{figure}[H]
\begin{center}
\includegraphics[width=6.75cm,height=6.25cm]{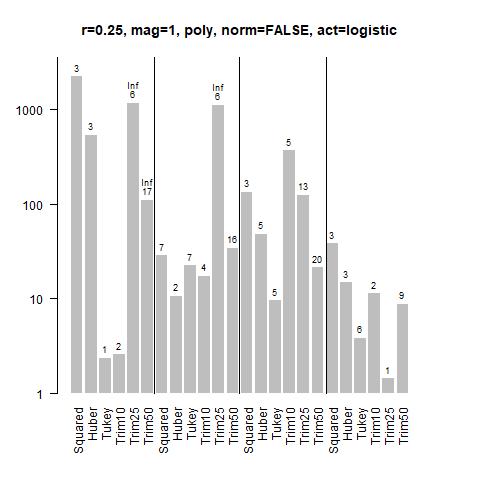}
\includegraphics[width=6.75cm,height=6.25cm]{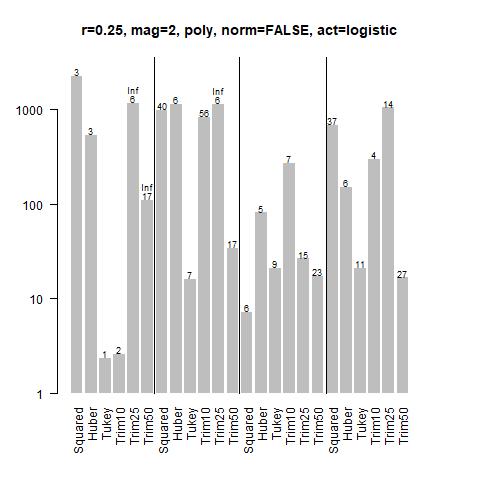} \\
\includegraphics[width=6.75cm,height=6.25cm]{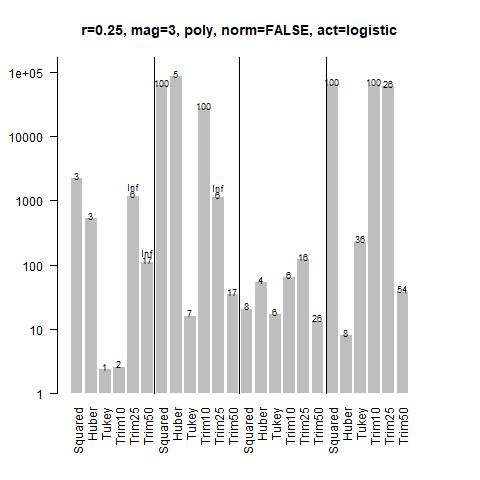} 
\includegraphics[width=6.75cm,height=6.25cm]{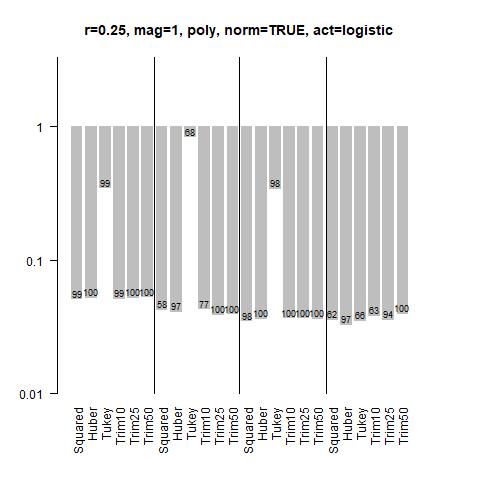}\\
\includegraphics[width=6.75cm,height=6.25cm]{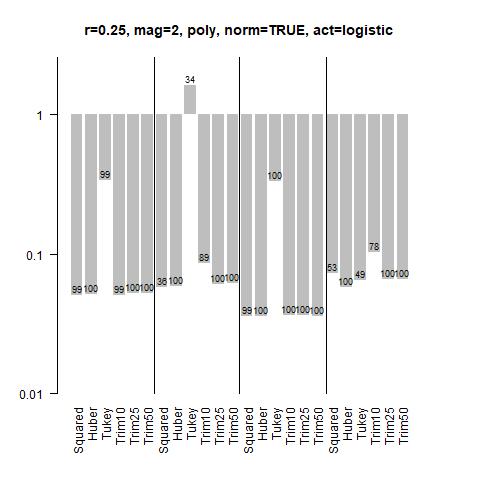} 
\includegraphics[width=6.75cm,height=6.25cm]{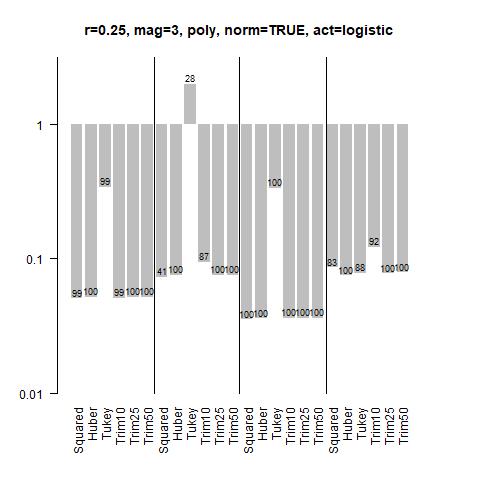} 
\end{center}
\caption{Results for $r=0.25$}\label{trimnn:n500p20r25m1polynonlogdeep}
\end{figure}

\begin{figure}[H]
\begin{center}
\includegraphics[width=6.75cm,height=6.25cm]{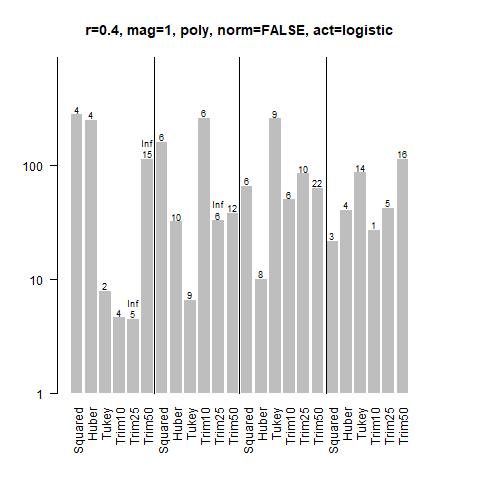}
\includegraphics[width=6.75cm,height=6.25cm]{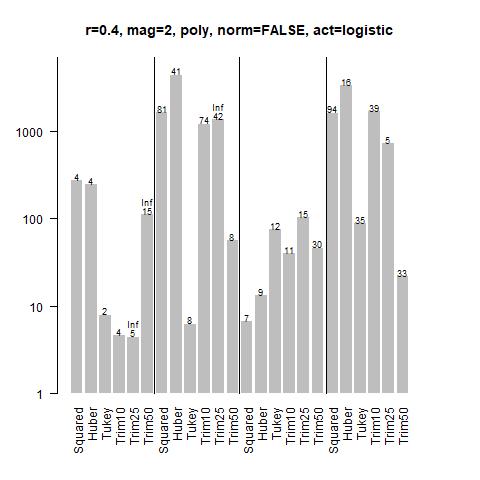} \\
\includegraphics[width=6.75cm,height=6.25cm]{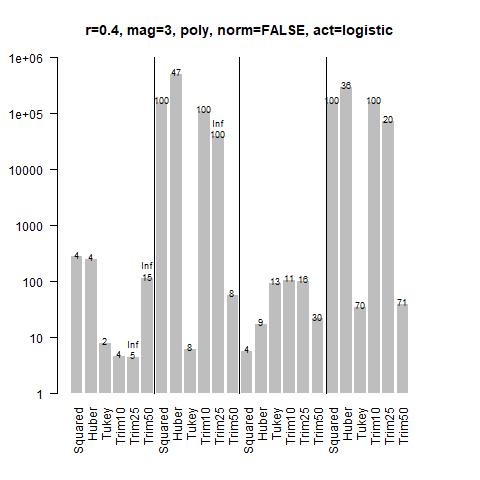} 
\includegraphics[width=6.75cm,height=6.25cm]{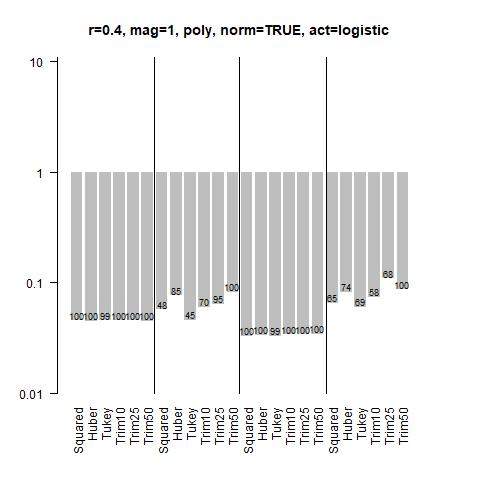}\\
\includegraphics[width=6.75cm,height=6.25cm]{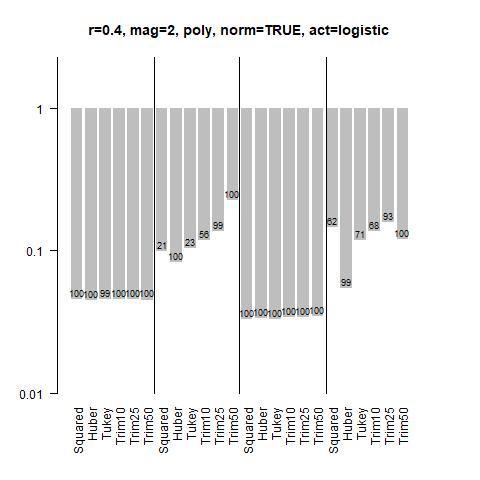} 
\includegraphics[width=6.75cm,height=6.25cm]{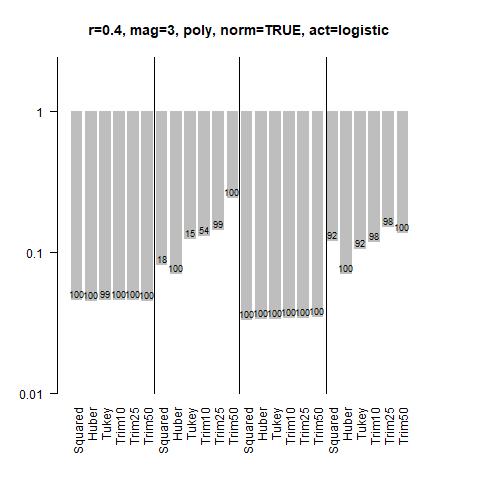} 
\end{center}
\caption{Results for $r=0.4$}\label{trimnn:n500p20r40m1polynonlogdeep}
\end{figure}

\subsubsection{Trigonometric function}

\begin{figure}[H]
\label{trimnn:n500p20r10m1trignonlogdeep}
\begin{center}
\includegraphics[width=6.75cm,height=6.25cm]{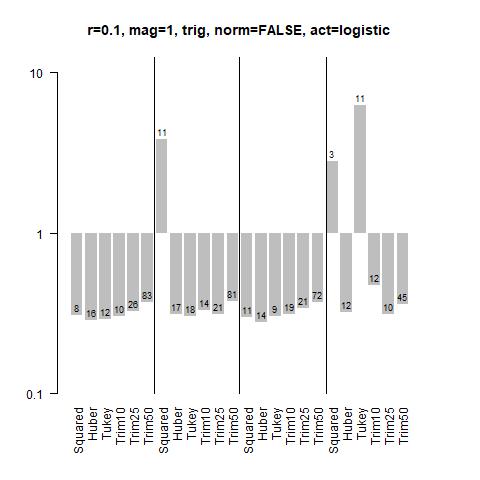}
\includegraphics[width=6.75cm,height=6.25cm]{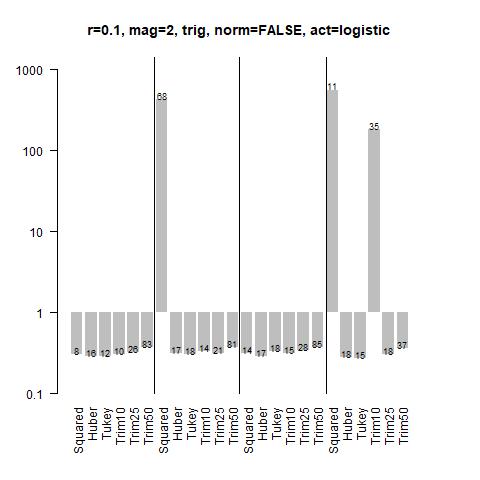} \\
\includegraphics[width=6.75cm,height=6.25cm]{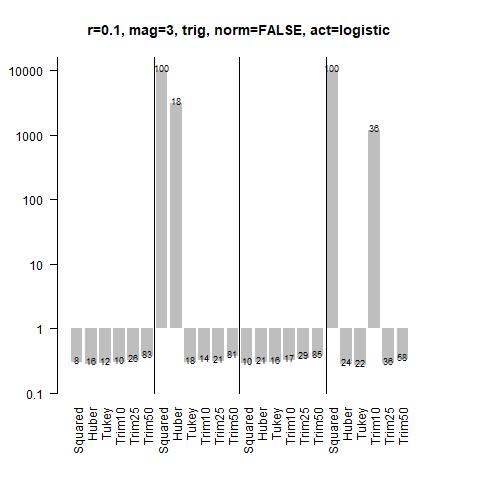} 
\includegraphics[width=6.75cm,height=6.25cm]{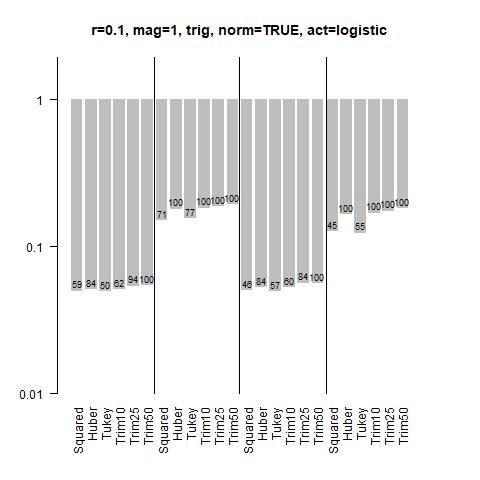}\\
\includegraphics[width=6.75cm,height=6.25cm]{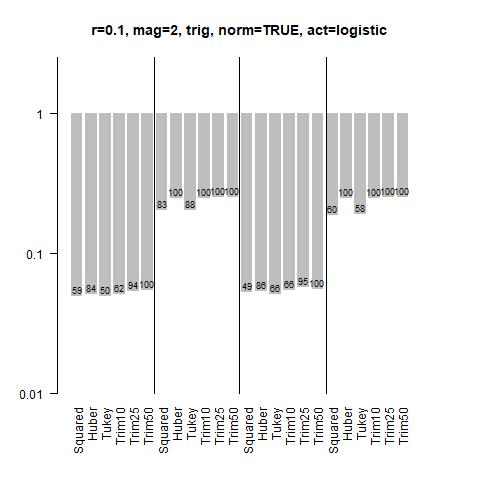} 
\includegraphics[width=6.75cm,height=6.25cm]{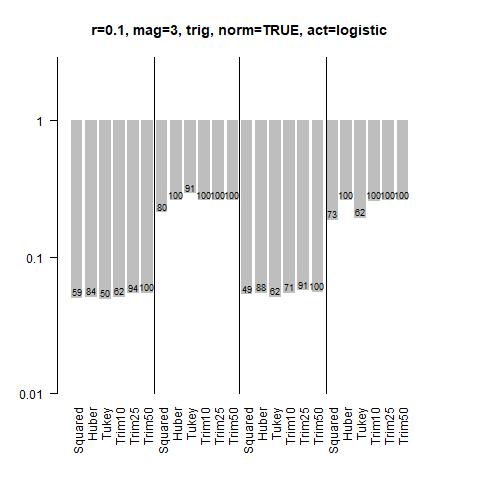} 
\end{center}
\caption{Results for $r=0.1$}
\end{figure}

\begin{figure}[H]
\label{trimnn:n500p20r25m1trignonlogdeep}
\begin{center}
\includegraphics[width=6.75cm,height=6.25cm]{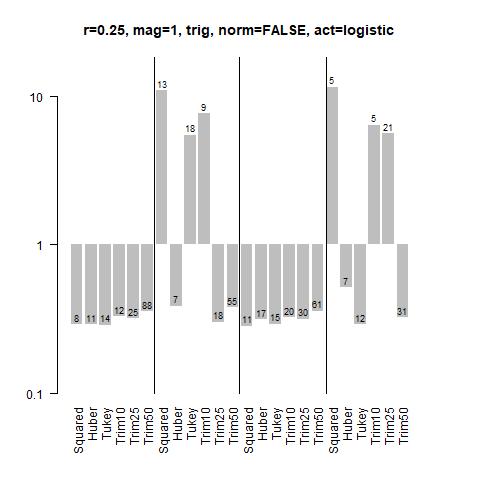}
\includegraphics[width=6.75cm,height=6.25cm]{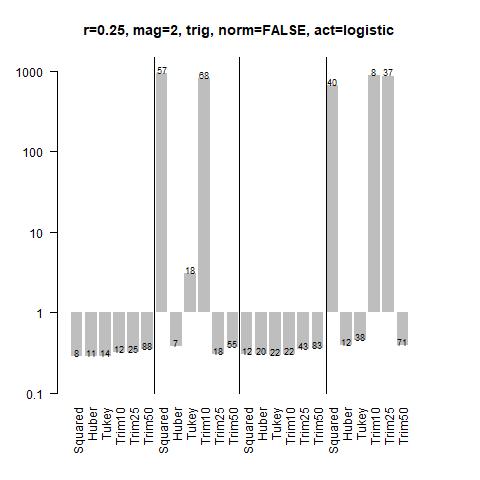} \\
\includegraphics[width=6.75cm,height=6.25cm]{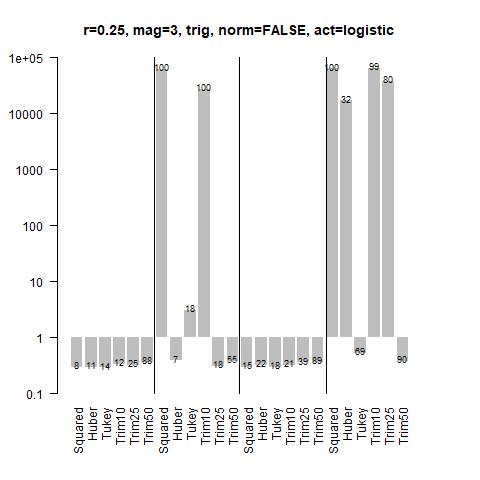} 
\includegraphics[width=6.75cm,height=6.25cm]{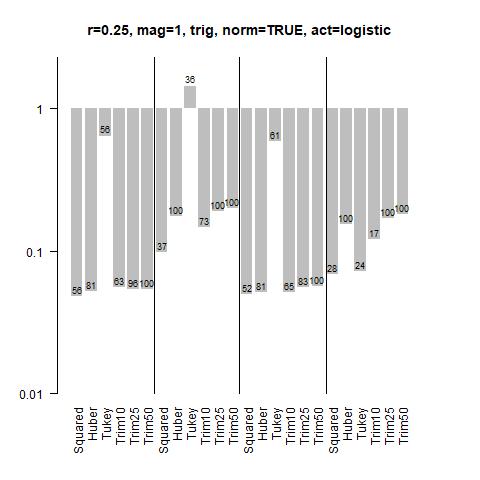}\\
\includegraphics[width=6.75cm,height=6.25cm]{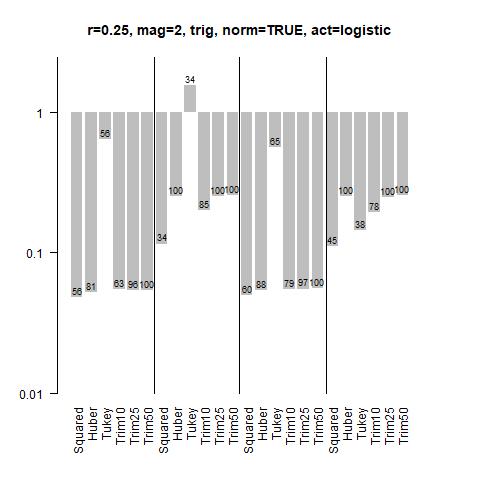} 
\includegraphics[width=6.75cm,height=6.25cm]{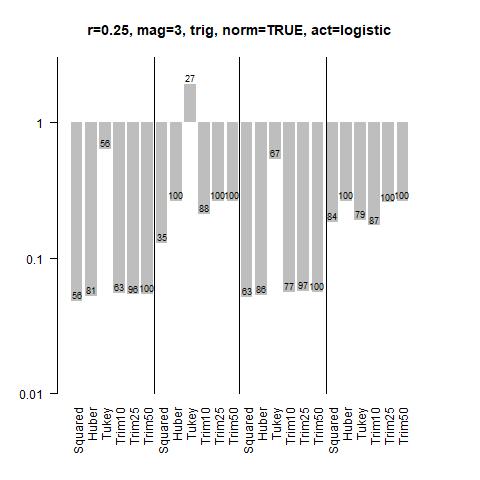} 
\end{center}
\caption{Results for $r=0.25$}
\end{figure}

\begin{figure}[H]
\label{trimnn:n500p20r40m1trignonlogdeep}
\begin{center}
\includegraphics[width=6.75cm,height=6.25cm]{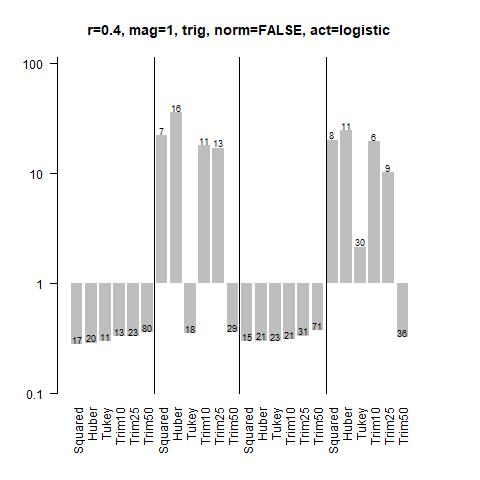}
\includegraphics[width=6.75cm,height=6.25cm]{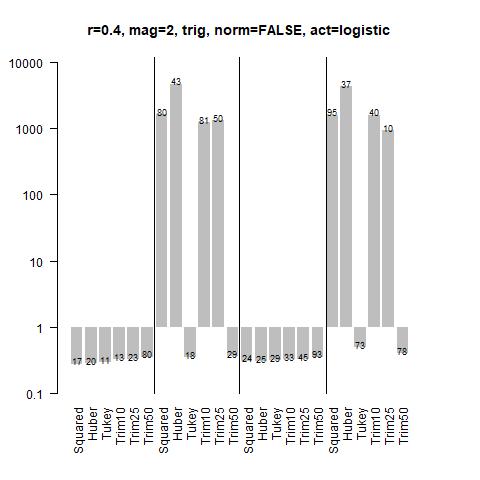} \\
\includegraphics[width=6.75cm,height=6.25cm]{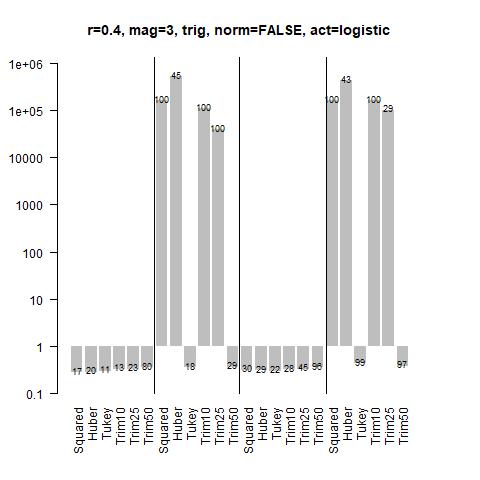} 
\includegraphics[width=6.75cm,height=6.25cm]{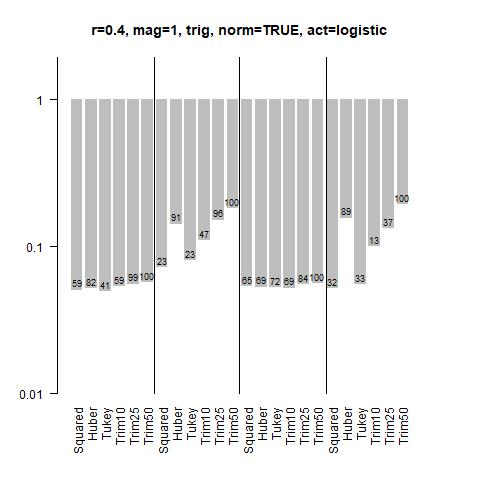}\\
\includegraphics[width=6.75cm,height=6.25cm]{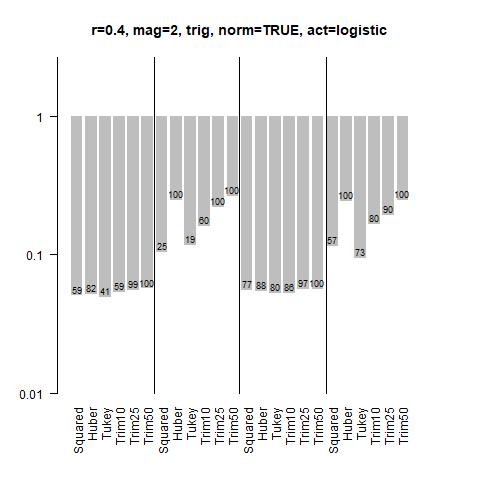} 
\includegraphics[width=6.75cm,height=6.25cm]{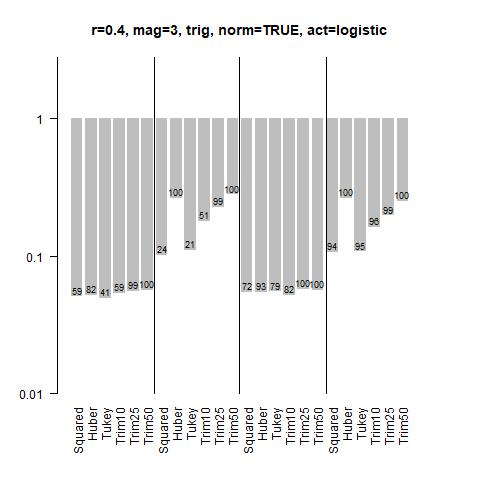} 
\end{center}
\caption{Results for $r=0.4$}
\end{figure}

\subsection{Softplus activation function}

\subsubsection{Linear function}

\begin{figure}[H]
\label{trimnn:n500p20r10m1linnonreludeep}
\begin{center}
\includegraphics[width=6.75cm,height=6.25cm]{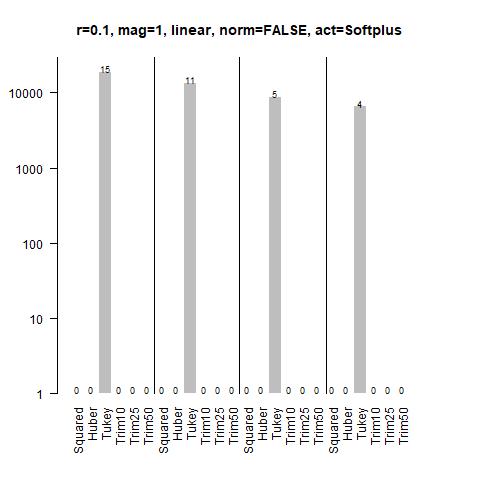}
\includegraphics[width=6.75cm,height=6.25cm]{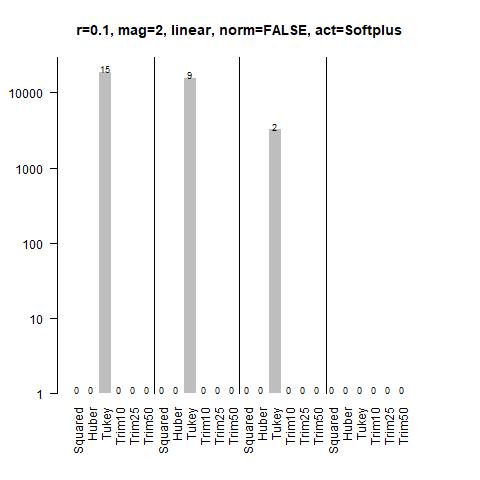} \\
\includegraphics[width=6.75cm,height=6.25cm]{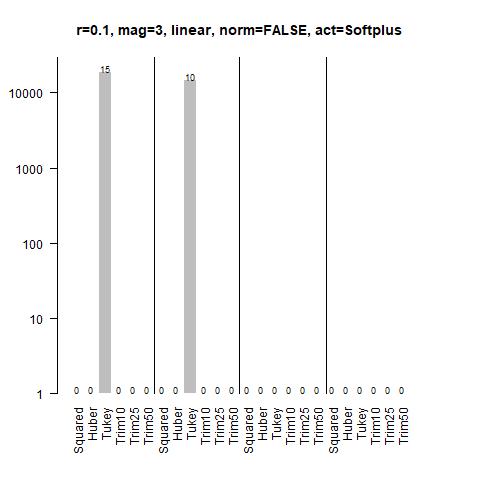} 
\includegraphics[width=6.75cm,height=6.25cm]{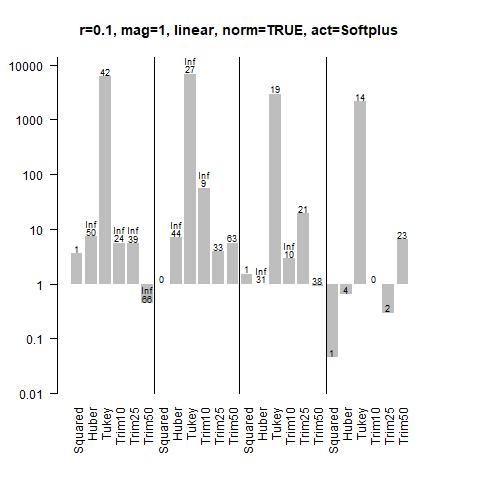}\\
\includegraphics[width=6.75cm,height=6.25cm]{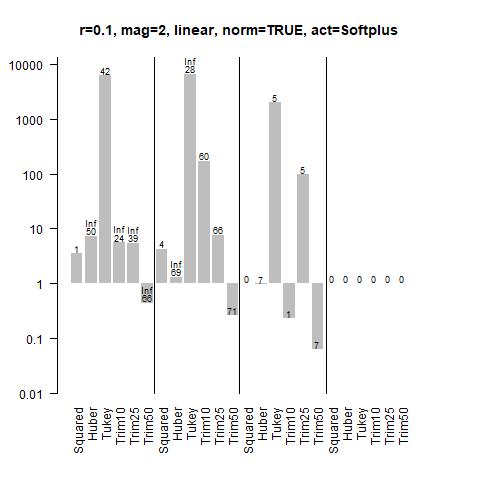} 
\includegraphics[width=6.75cm,height=6.25cm]{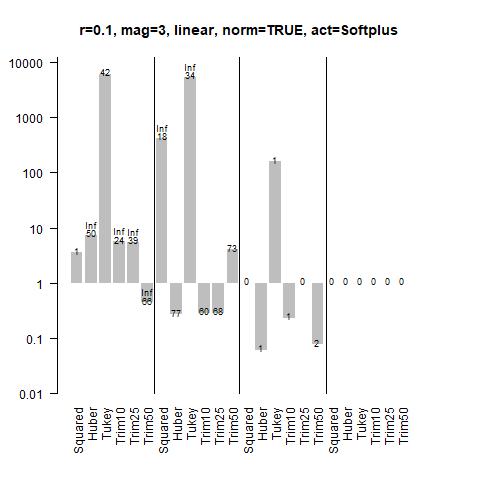} 
\end{center}
\caption{Results for $r=0.1$}
\end{figure}

\begin{figure}[H]
\label{trimnn:n500p20r25m1linnonreludeep}
\begin{center}
\includegraphics[width=6.75cm,height=6.25cm]{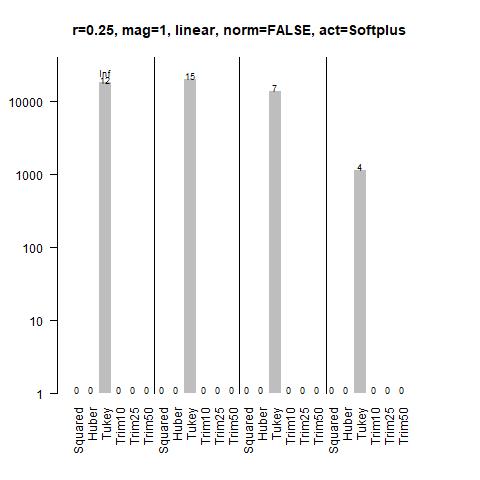}
\includegraphics[width=6.75cm,height=6.25cm]{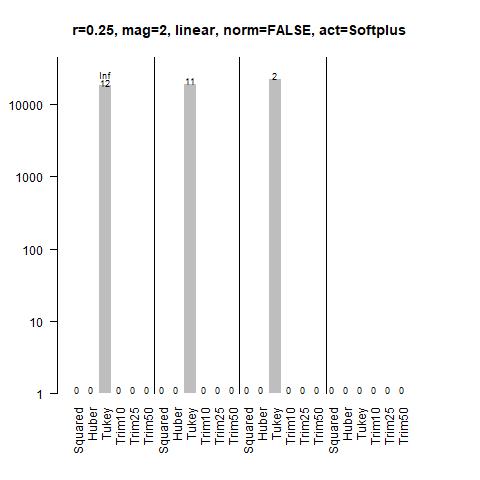} \\
\includegraphics[width=6.75cm,height=6.25cm]{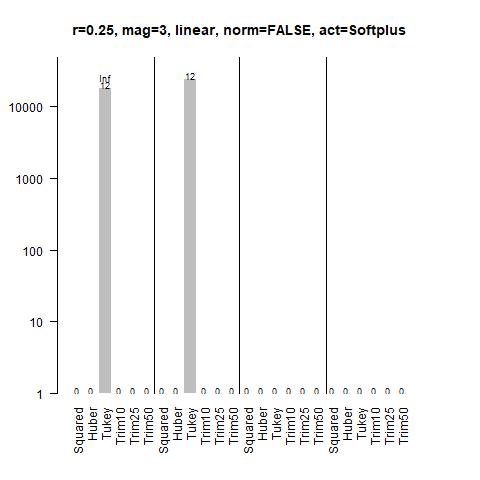} 
\includegraphics[width=6.75cm,height=6.25cm]{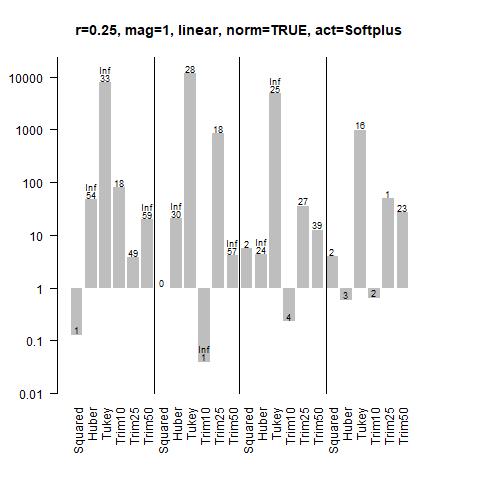}\\
\includegraphics[width=6.75cm,height=6.25cm]{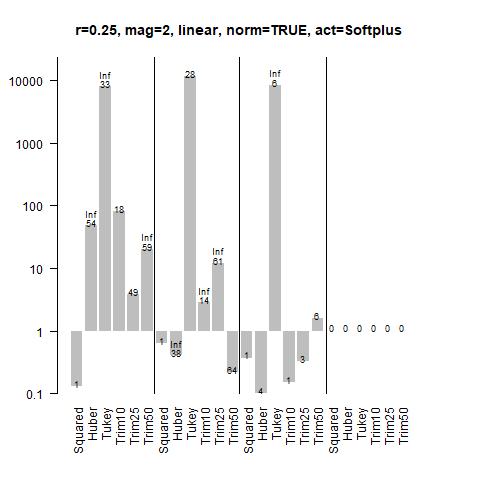} 
\includegraphics[width=6.75cm,height=6.25cm]{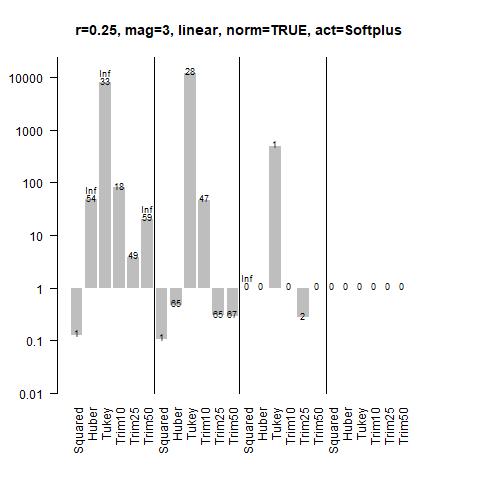} 
\end{center}
\caption{Results for $r=0.25$}
\end{figure}

\begin{figure}[H]
\label{trimnn:n500p20r40m1linnonreludeep}
\begin{center}
\includegraphics[width=6.75cm,height=6.25cm]{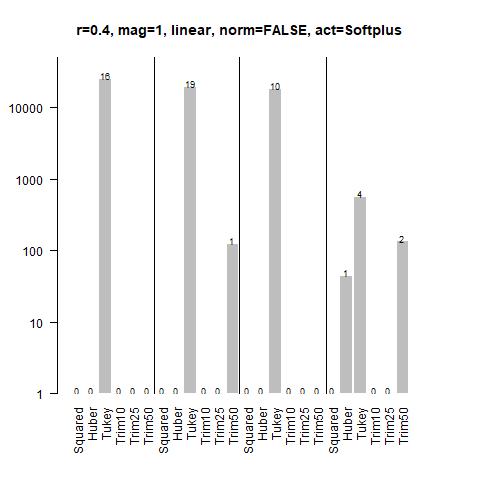}
\includegraphics[width=6.75cm,height=6.25cm]{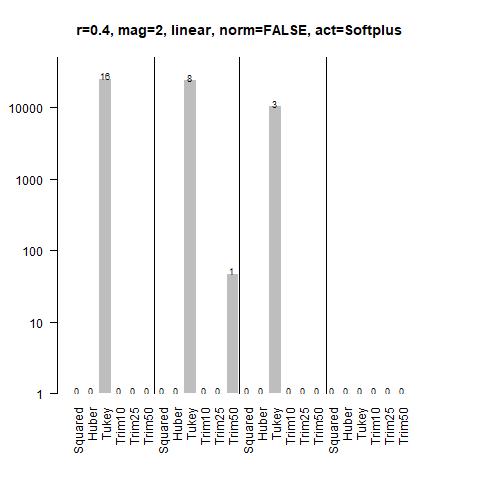} \\
\includegraphics[width=6.75cm,height=6.25cm]{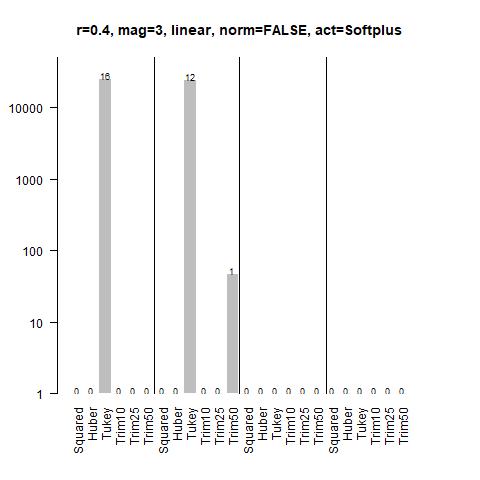} 
\includegraphics[width=6.75cm,height=6.25cm]{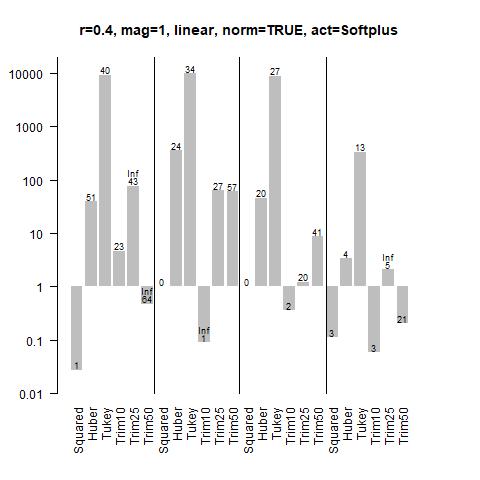}\\
\includegraphics[width=6.75cm,height=6.25cm]{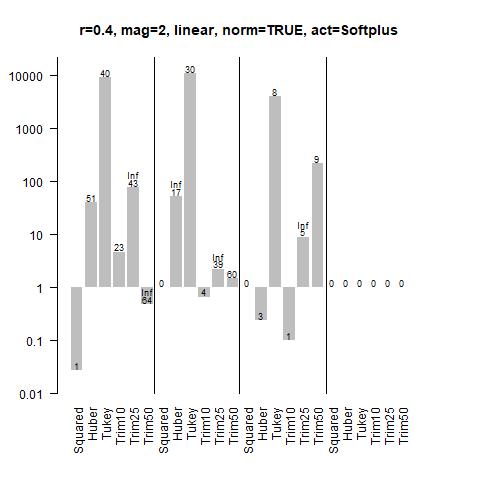} 
\includegraphics[width=6.75cm,height=6.25cm]{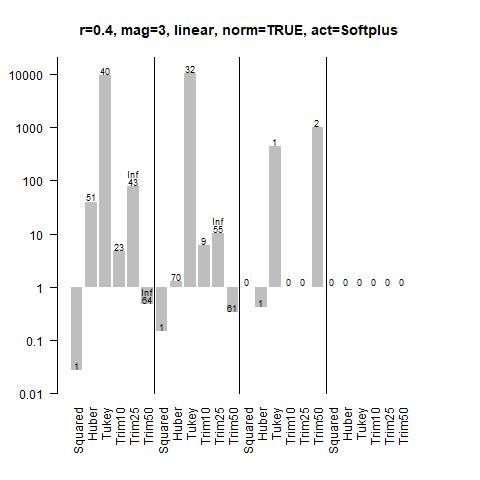} 
\end{center}
\caption{Results for $r=0.4$}
\end{figure}

\subsubsection{Polynomial function}

\begin{figure}[H]
\label{trimnn:n500p20r10m1polynonreludeep}
\begin{center}
\includegraphics[width=6.75cm,height=6.25cm]{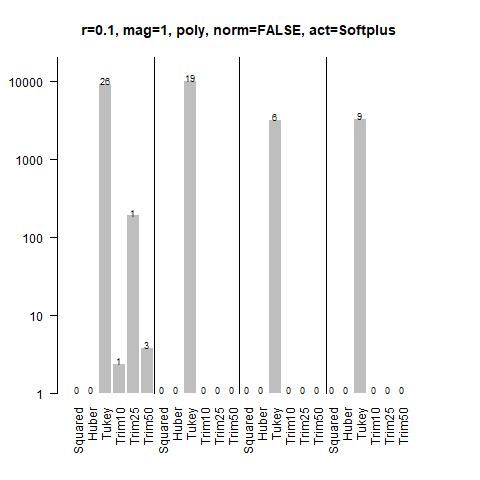}
\includegraphics[width=6.75cm,height=6.25cm]{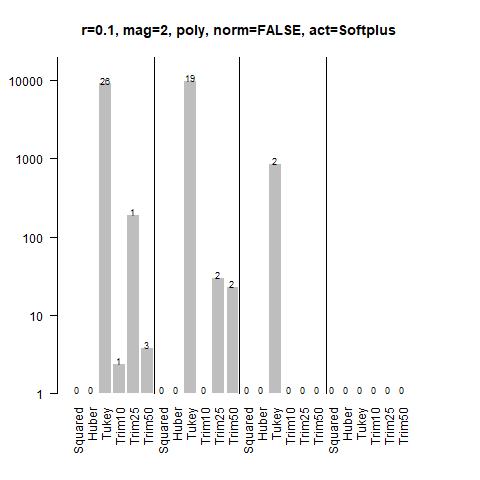} \\
\includegraphics[width=6.75cm,height=6.25cm]{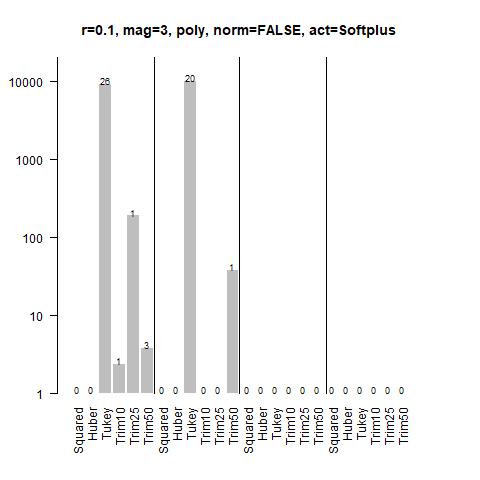} 
\includegraphics[width=6.75cm,height=6.25cm]{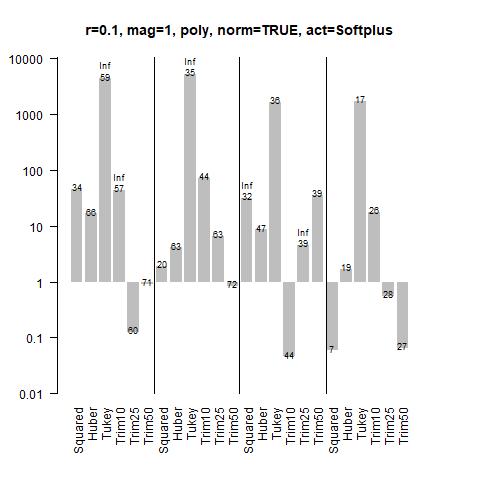}\\
\includegraphics[width=6.75cm,height=6.25cm]{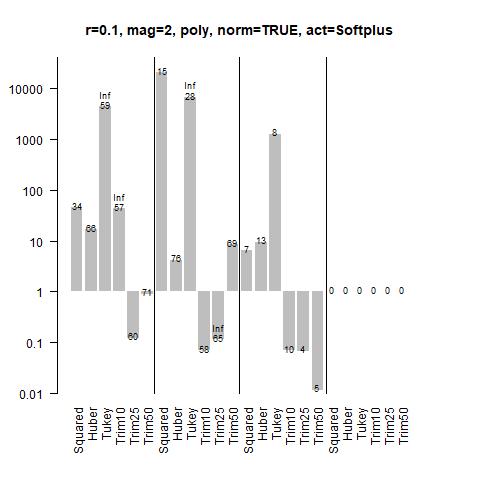} 
\includegraphics[width=6.75cm,height=6.25cm]{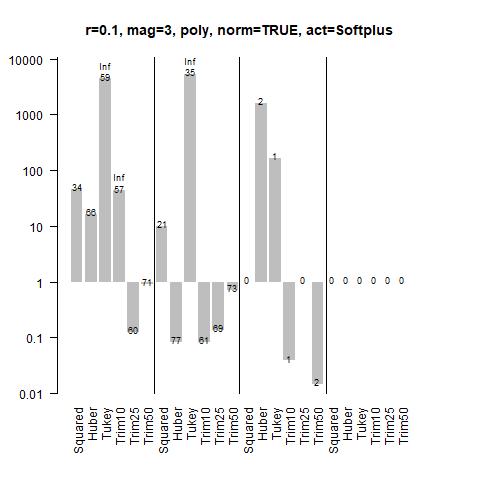} 
\end{center}
\caption{Results for $r=0.1$}
\end{figure}

\begin{figure}[H]
\label{trimnn:n500p20r25m1polynonreludeep}
\begin{center}
\includegraphics[width=6.75cm,height=6.25cm]{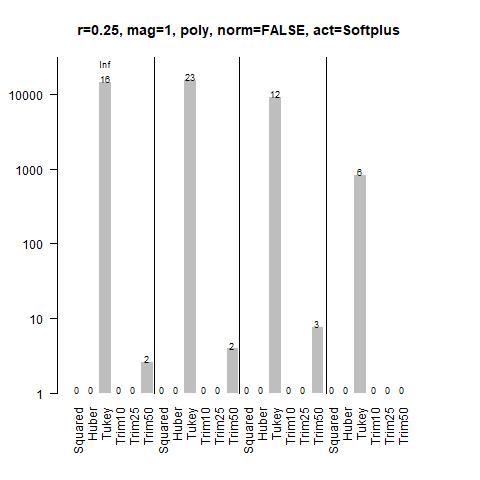}
\includegraphics[width=6.75cm,height=6.25cm]{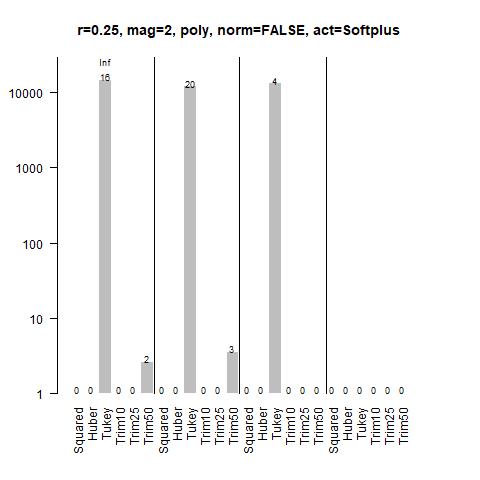} \\
\includegraphics[width=6.75cm,height=6.25cm]{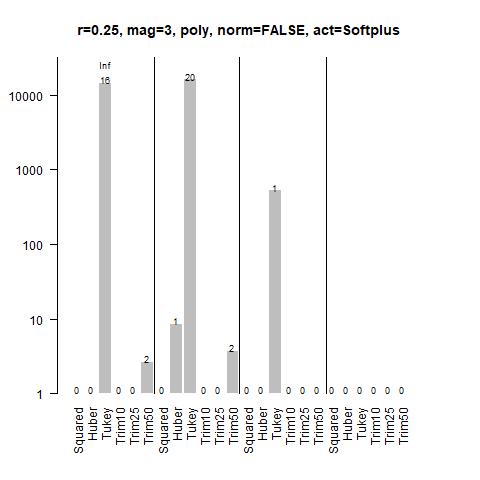} 
\includegraphics[width=6.75cm,height=6.25cm]{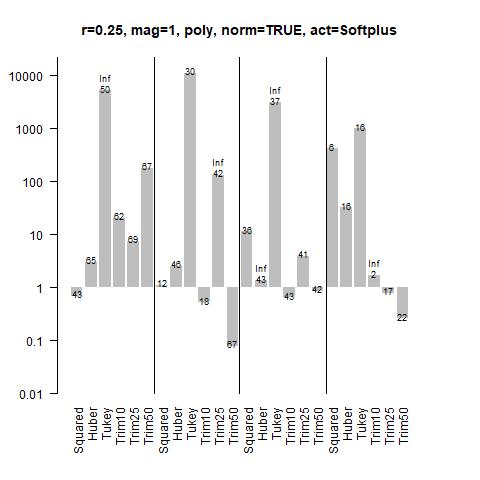}\\
\includegraphics[width=6.75cm,height=6.25cm]{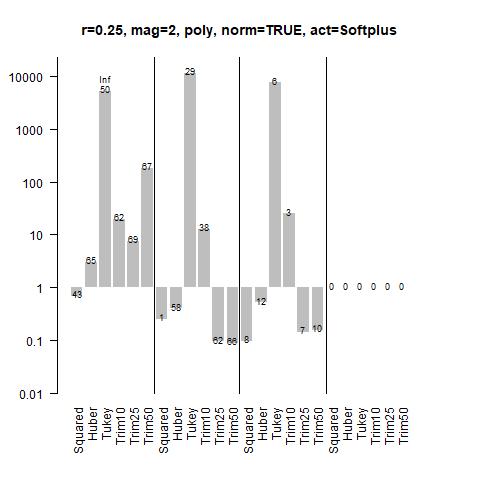} 
\includegraphics[width=6.75cm,height=6.25cm]{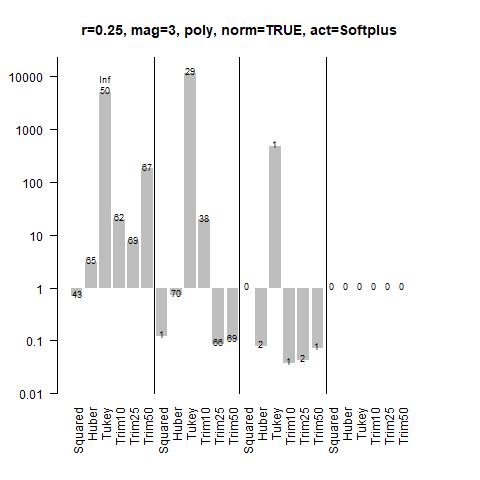} 
\end{center}
\caption{Results for $r=0.25$}
\end{figure}

\begin{figure}[H]
\label{trimnn:n500p20r40m1polynonreludeep}
\begin{center}
\includegraphics[width=6.75cm,height=6.25cm]{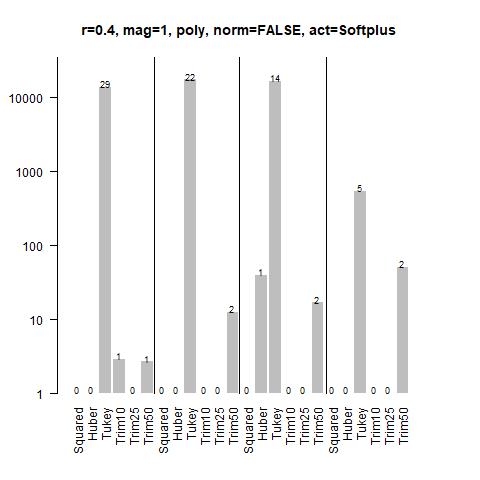}
\includegraphics[width=6.75cm,height=6.25cm]{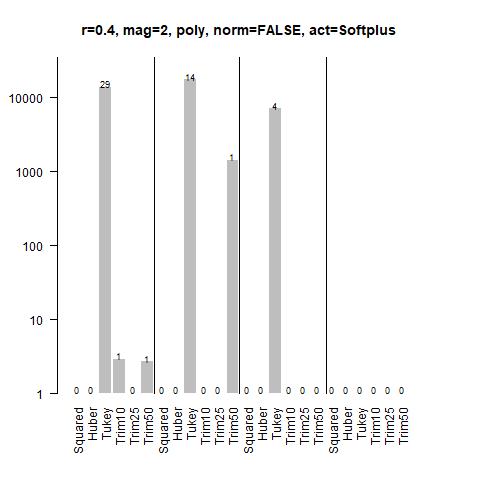} \\
\includegraphics[width=6.75cm,height=6.25cm]{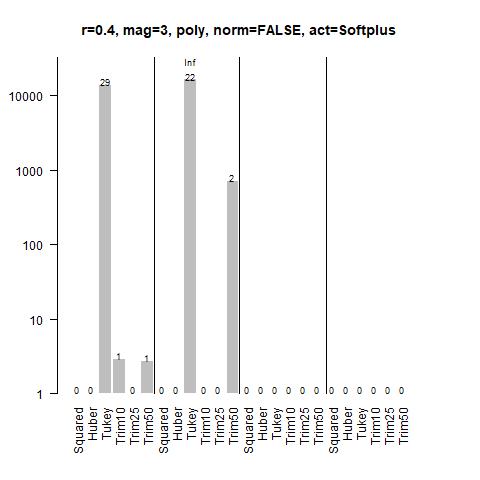} 
\includegraphics[width=6.75cm,height=6.25cm]{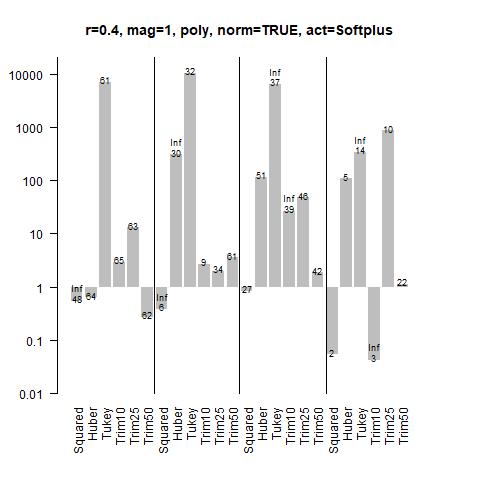}\\
\includegraphics[width=6.75cm,height=6.25cm]{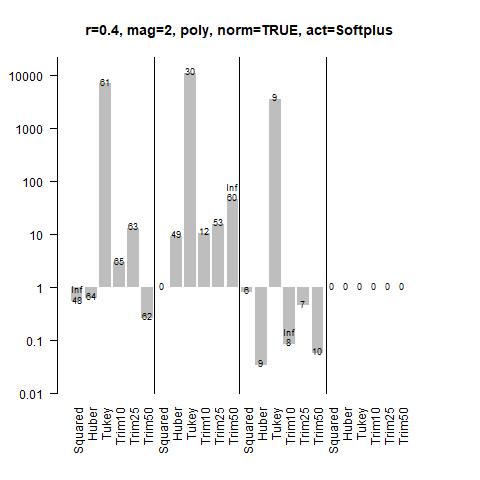} 
\includegraphics[width=6.75cm,height=6.25cm]{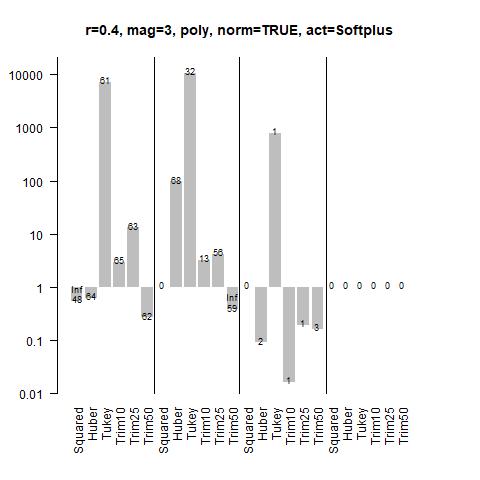} 
\end{center}
\caption{Results for $r=0.4$}
\end{figure}

\subsubsection{Trigonometric function}

\begin{figure}[H]
\label{trimnn:n500p20r10m1trignonreludeep}
\begin{center}
\includegraphics[width=6.75cm,height=6.25cm]{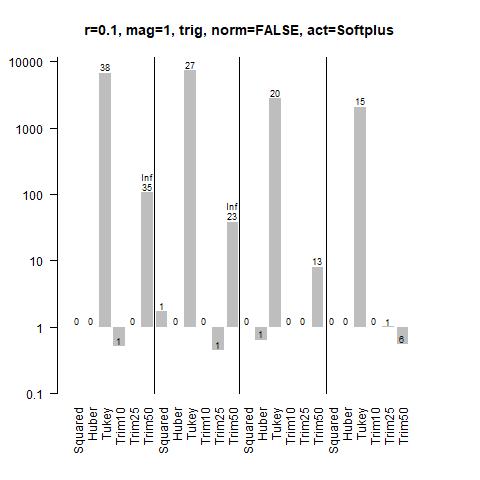}
\includegraphics[width=6.75cm,height=6.25cm]{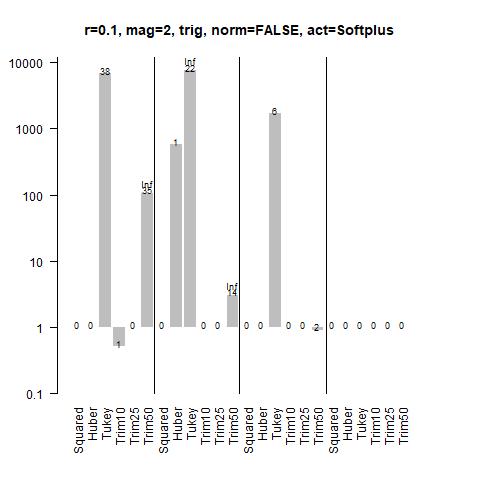} \\
\includegraphics[width=6.75cm,height=6.25cm]{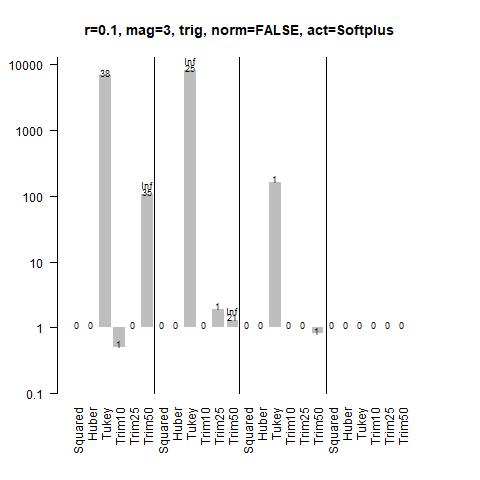} 
\includegraphics[width=6.75cm,height=6.25cm]{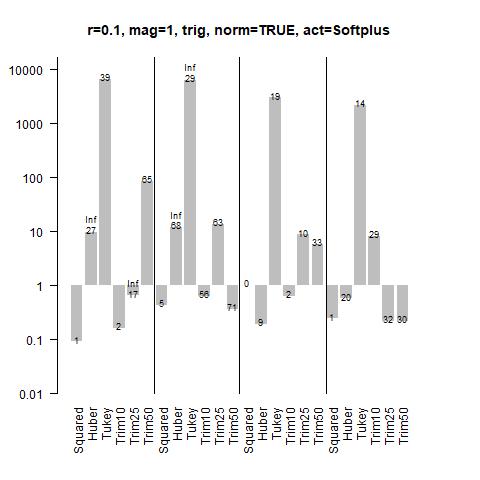}\\
\includegraphics[width=6.75cm,height=6.25cm]{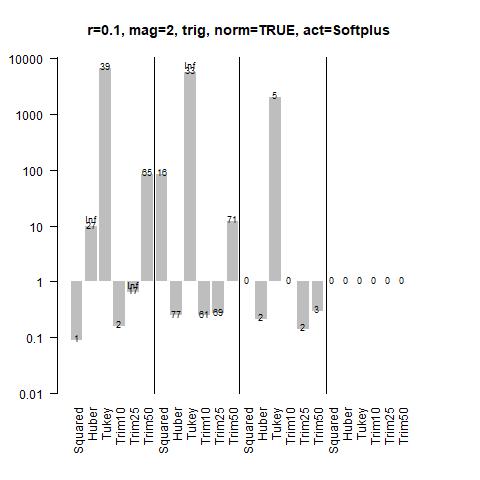} 
\includegraphics[width=6.75cm,height=6.25cm]{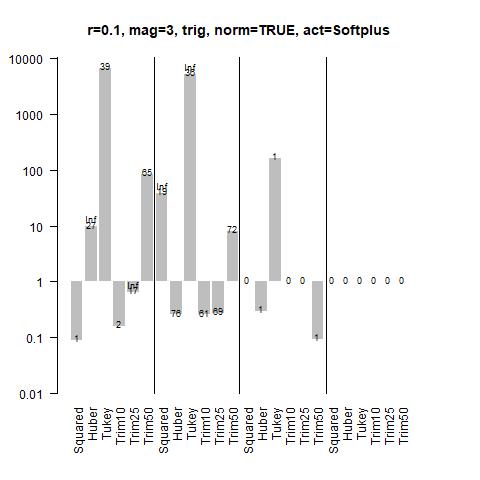} 
\end{center}
\caption{Results for $r=0.1$}
\end{figure}

\begin{figure}[H]
\label{trimnn:n500p20r25m1trignonreludeep}
\begin{center}
\includegraphics[width=6.75cm,height=6.25cm]{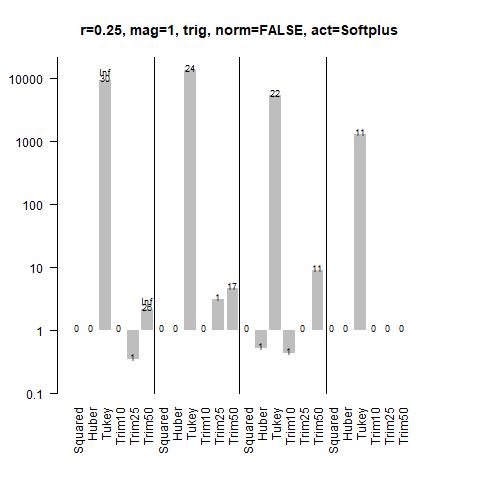}
\includegraphics[width=6.75cm,height=6.25cm]{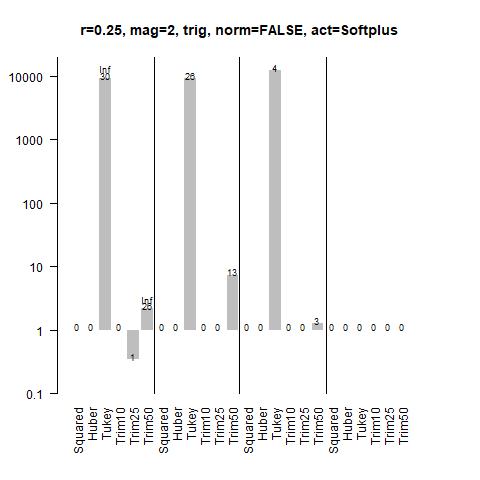} \\
\includegraphics[width=6.75cm,height=6.25cm]{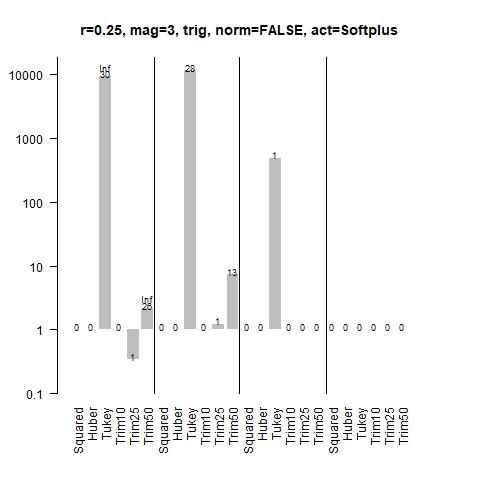} 
\includegraphics[width=6.75cm,height=6.25cm]{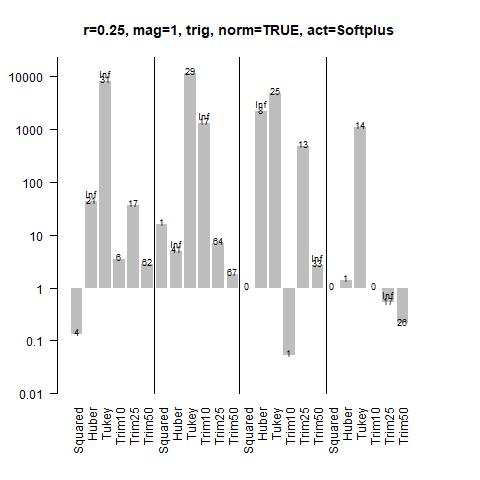}\\
\includegraphics[width=6.75cm,height=6.25cm]{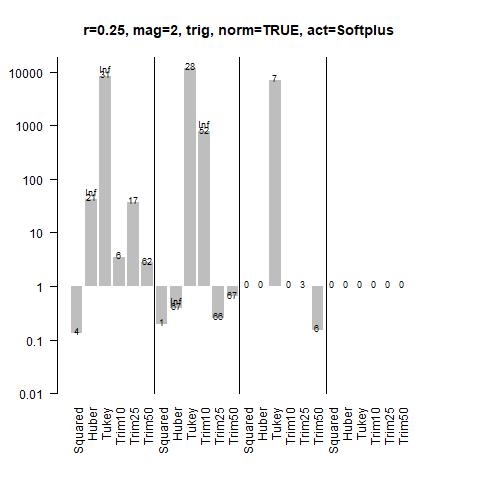} 
\includegraphics[width=6.75cm,height=6.25cm]{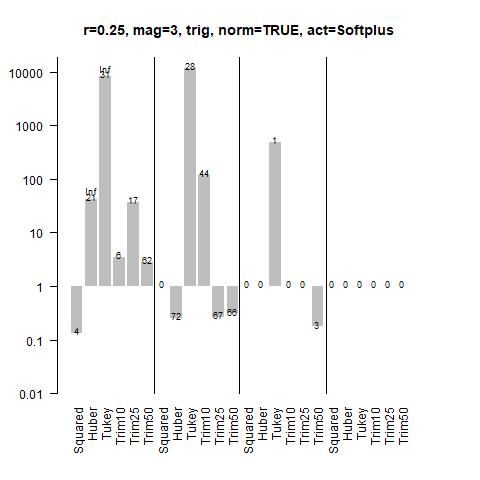} 
\end{center}
\caption{Results for $r=0.25$}
\end{figure}

\begin{figure}[H]
\label{trimnn:n500p20r40m1trignonreludeep}
\begin{center}
\includegraphics[width=6.75cm,height=6.25cm]{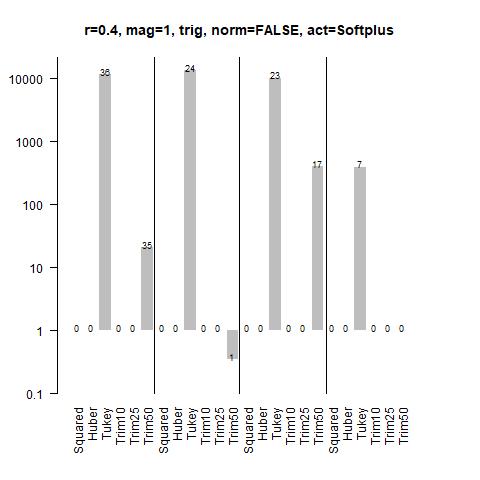}
\includegraphics[width=6.75cm,height=6.25cm]{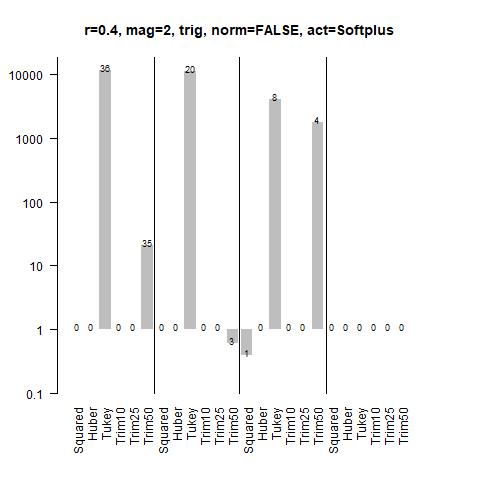} \\
\includegraphics[width=6.75cm,height=6.25cm]{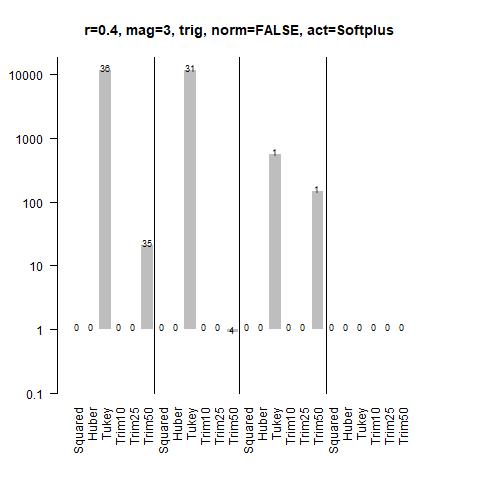} 
\includegraphics[width=6.75cm,height=6.25cm]{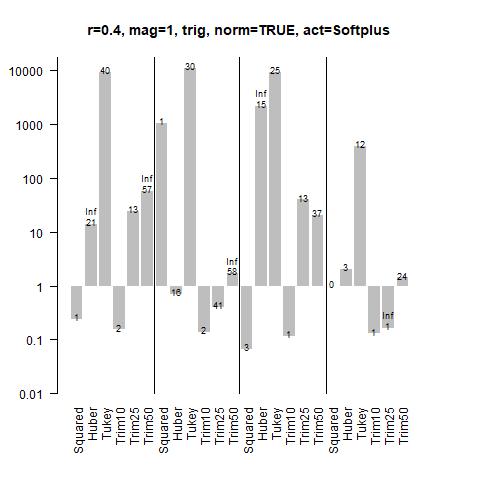}\\
\includegraphics[width=6.75cm,height=6.25cm]{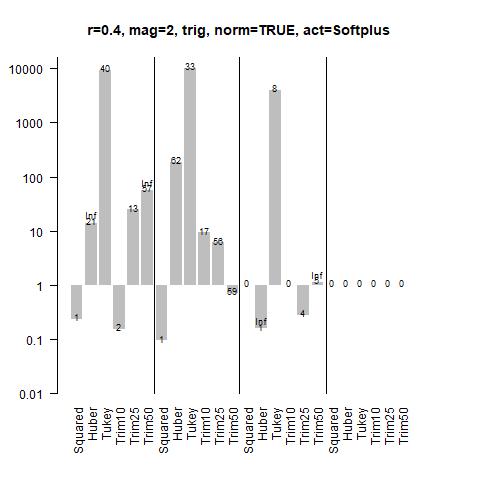} 
\includegraphics[width=6.75cm,height=6.25cm]{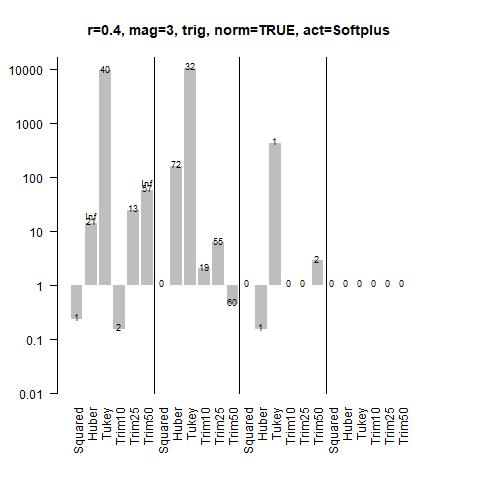} 
\end{center}
\caption{Results for $r=0.4$}
\end{figure}

\section{Simulation results for $n=1000$ and $p=50$: Test loss} \label{trimnn:secloss100050}

\subsection{Logistic activation function}

\subsubsection{Linear function}

\begin{figure}[H]
\label{trimnn:n1000p50r10m1linnonlog}
\begin{center}
\includegraphics[width=6.75cm,height=6.25cm]{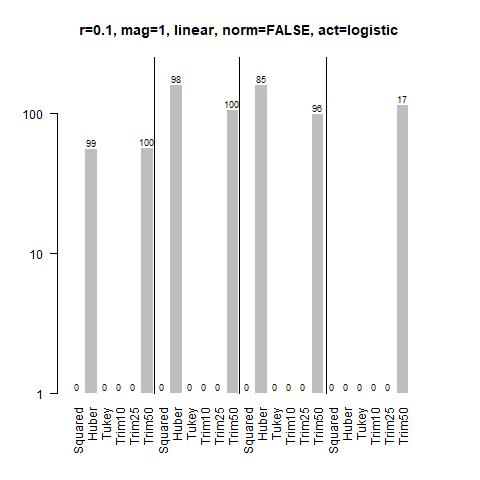}
\includegraphics[width=6.75cm,height=6.25cm]{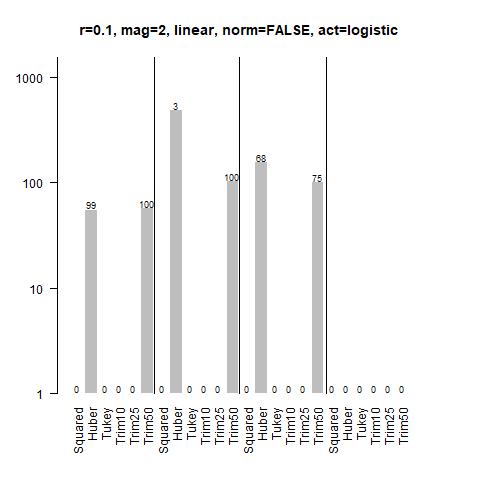} \\
\includegraphics[width=6.75cm,height=6.25cm]{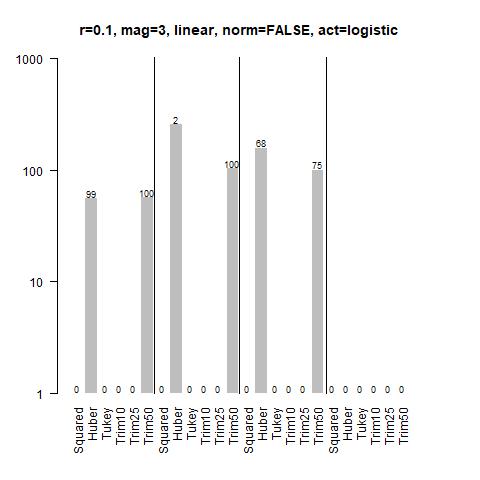} 
\includegraphics[width=6.75cm,height=6.25cm]{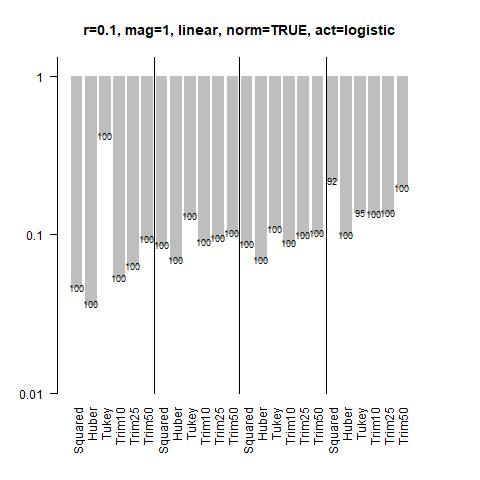}\\
\includegraphics[width=6.75cm,height=6.25cm]{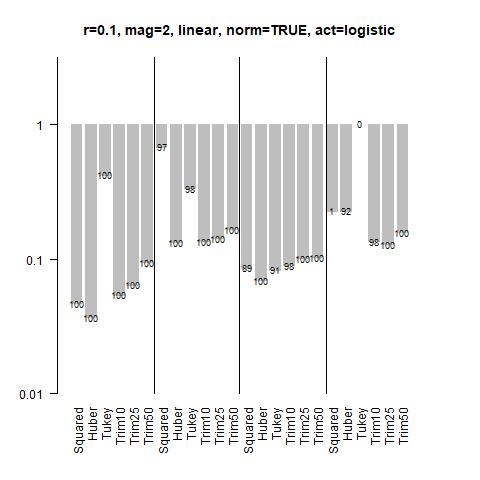} 
\includegraphics[width=6.75cm,height=6.25cm]{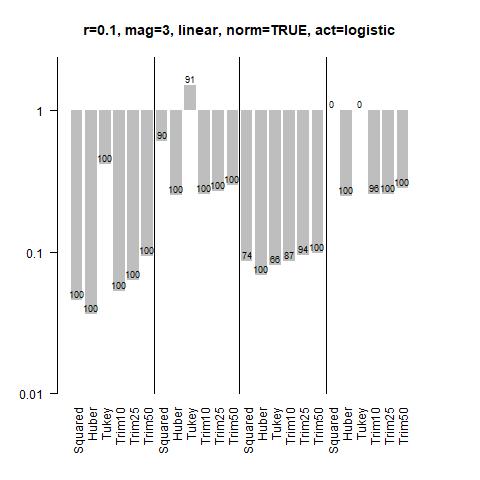} 
\end{center}
\caption{Results for $r=0.1$}
\end{figure}

\begin{figure}[H]
\label{trimnn:n1000p50r25m1linnonlog}
\begin{center}
\includegraphics[width=6.75cm,height=6.25cm]{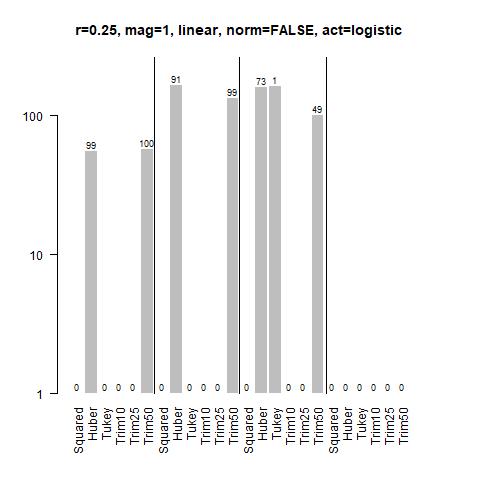}
\includegraphics[width=6.75cm,height=6.25cm]{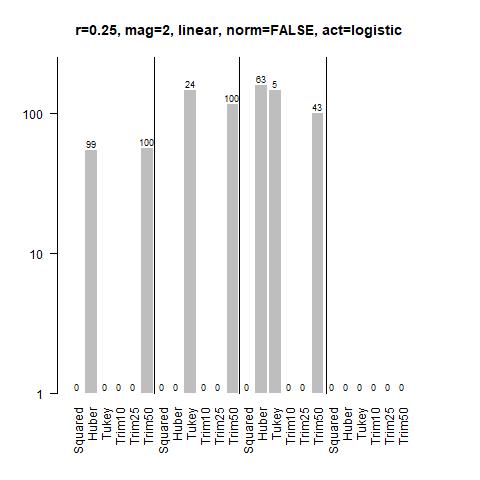} \\
\includegraphics[width=6.75cm,height=6.25cm]{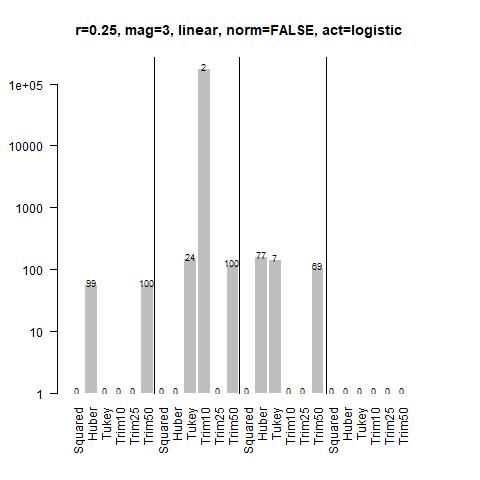} 
\includegraphics[width=6.75cm,height=6.25cm]{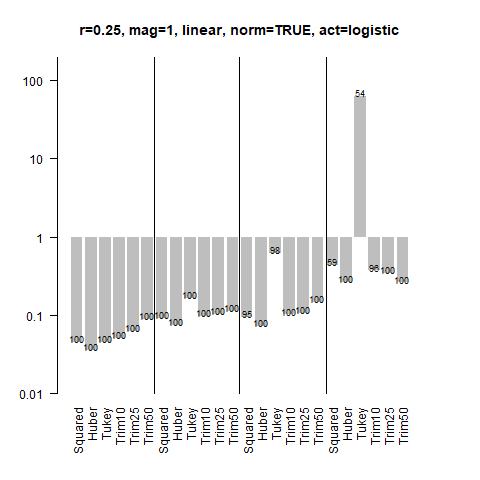}\\
\includegraphics[width=6.75cm,height=6.25cm]{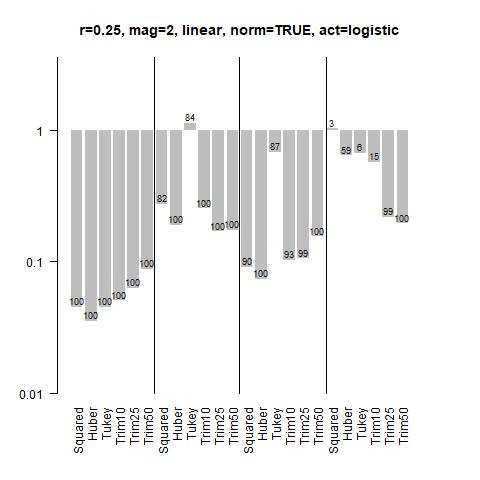} 
\includegraphics[width=6.75cm,height=6.25cm]{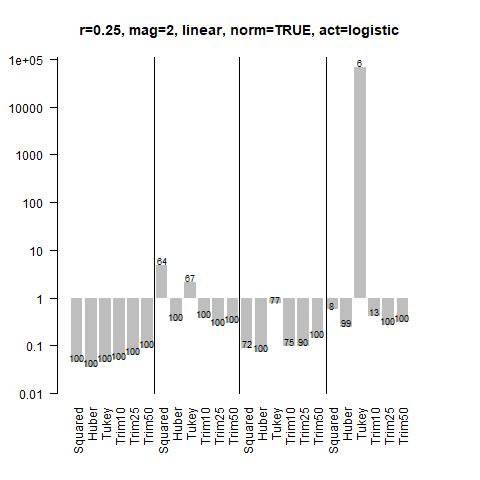} 
\end{center}
\caption{Results for $r=0.25$}
\end{figure}

\begin{figure}[H]
\label{trimnn:n1000p50r40m1linnonlog}
\begin{center}
\includegraphics[width=6.75cm,height=6.25cm]{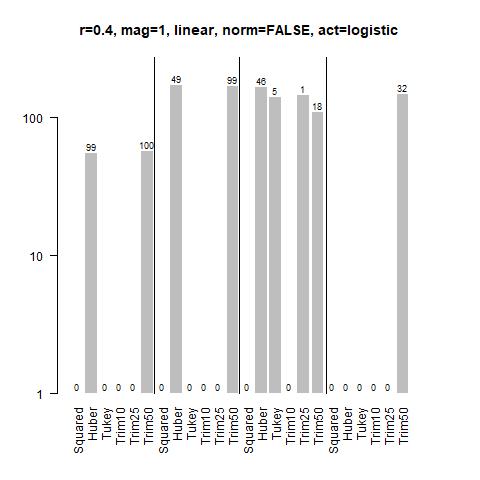}
\includegraphics[width=6.75cm,height=6.25cm]{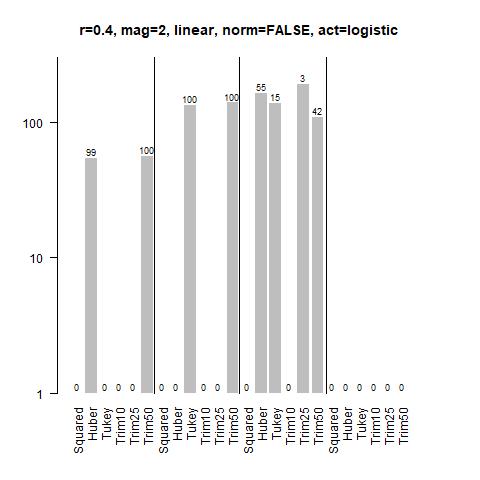} \\
\includegraphics[width=6.75cm,height=6.25cm]{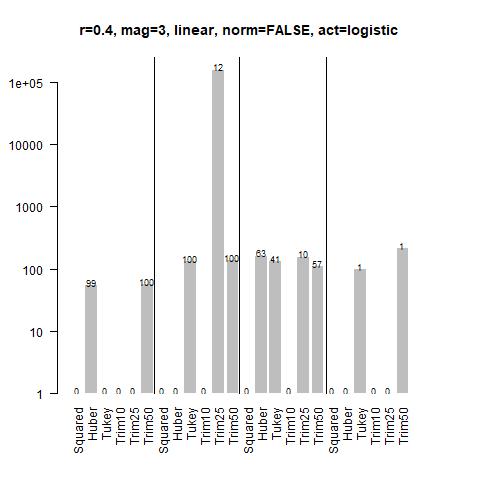} 
\includegraphics[width=6.75cm,height=6.25cm]{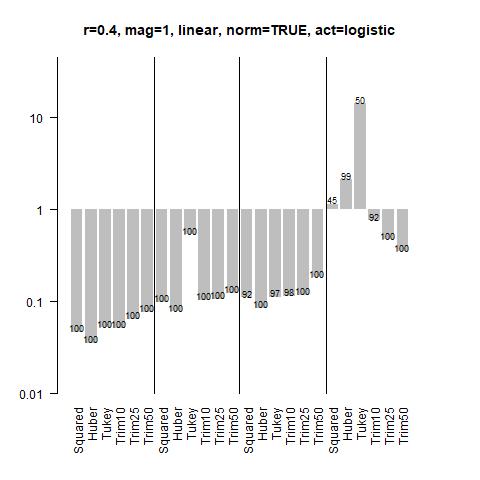}\\
\includegraphics[width=6.75cm,height=6.25cm]{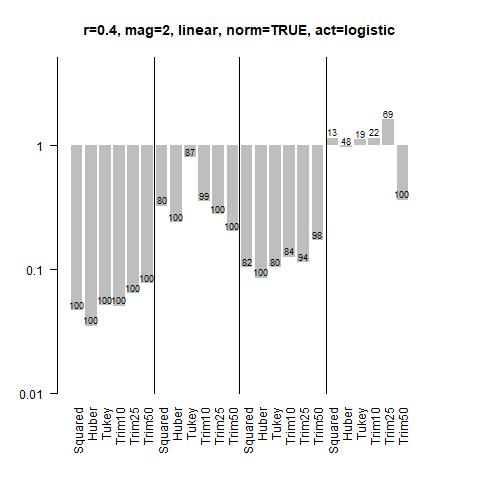} 
\includegraphics[width=6.75cm,height=6.25cm]{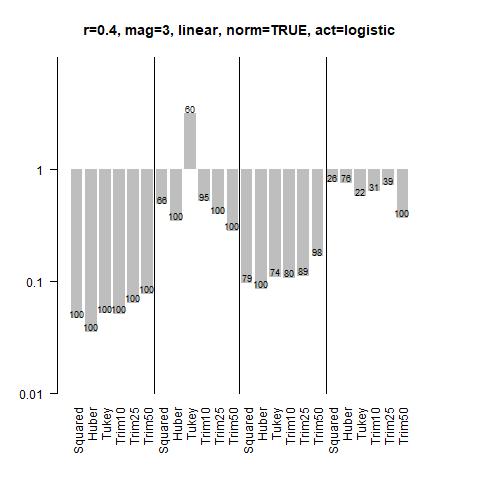} 
\end{center}
\caption{Results for $r=0.4$}
\end{figure}

\subsubsection{Polynomial function}

\begin{figure}[H]
\begin{center}
\includegraphics[width=6.75cm,height=6.25cm]{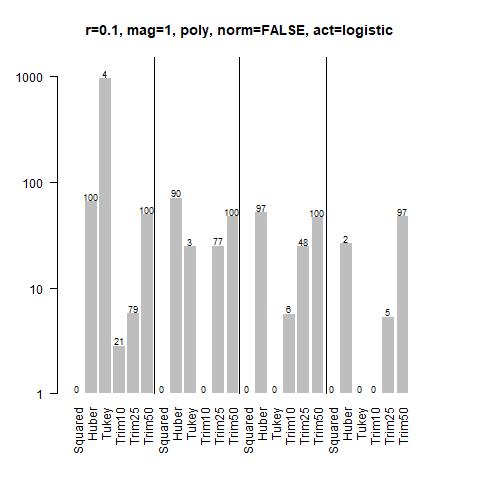}
\includegraphics[width=6.75cm,height=6.25cm]{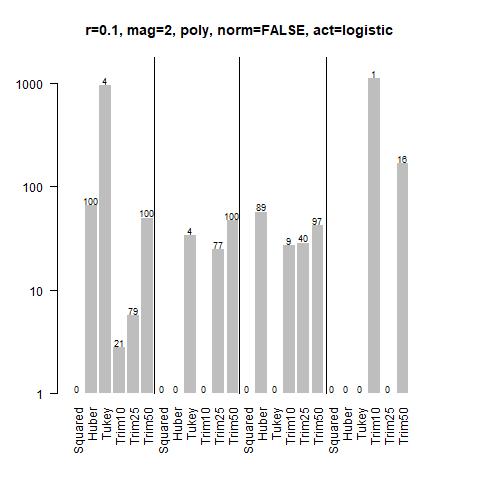} \\
\includegraphics[width=6.75cm,height=6.25cm]{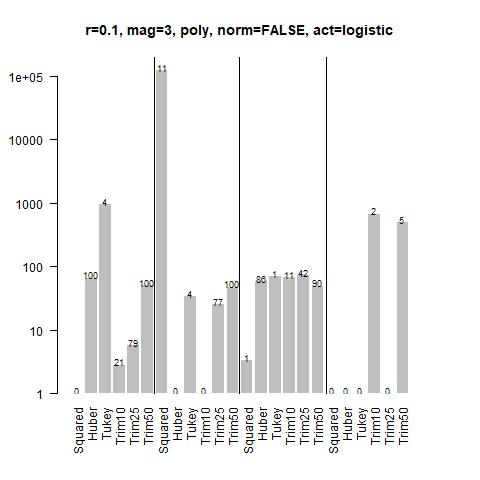} 
\includegraphics[width=6.75cm,height=6.25cm]{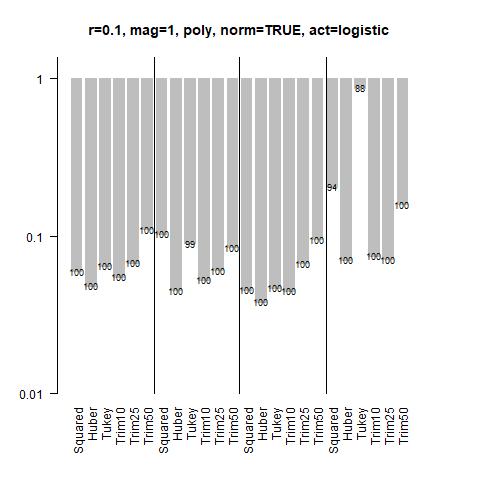}\\
\includegraphics[width=6.75cm,height=6.25cm]{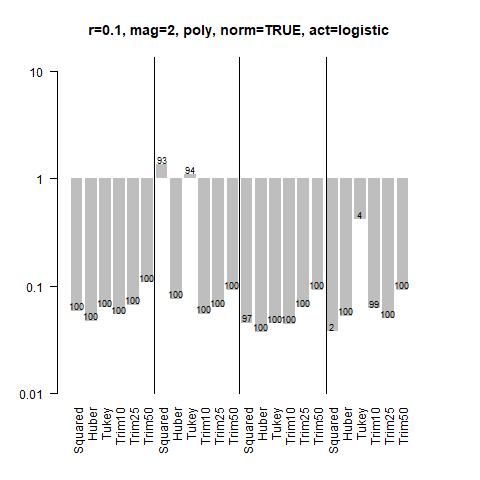} 
\includegraphics[width=6.75cm,height=6.25cm]{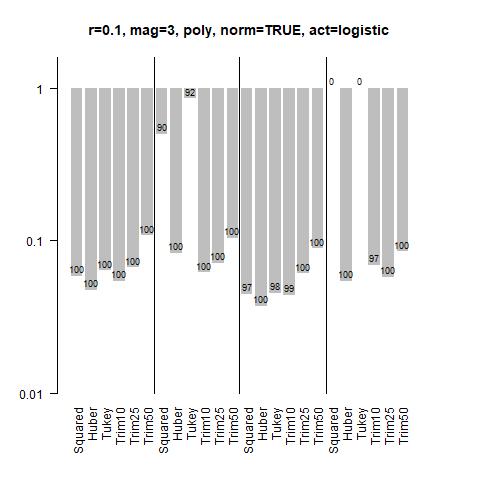} 
\end{center}
\caption{Results for $r=0.1$}\label{trimnn:n1000p50r10m1polynonlog}
\end{figure}

\begin{figure}[H]
\begin{center}
\includegraphics[width=6.75cm,height=6.25cm]{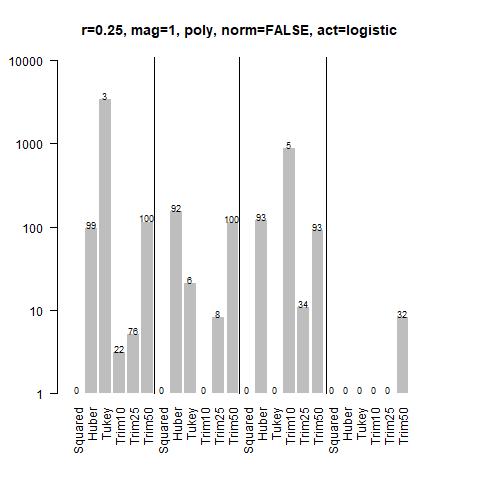}
\includegraphics[width=6.75cm,height=6.25cm]{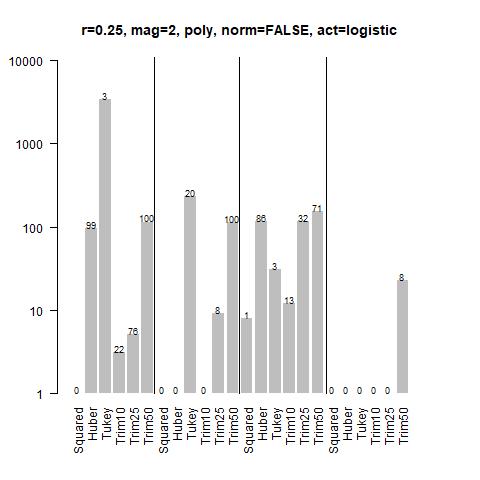} \\
\includegraphics[width=6.75cm,height=6.25cm]{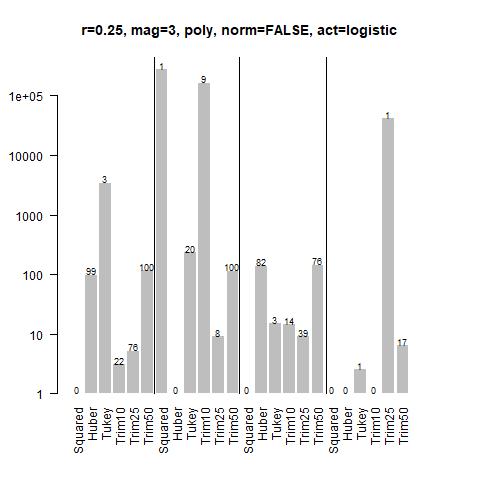} 
\includegraphics[width=6.75cm,height=6.25cm]{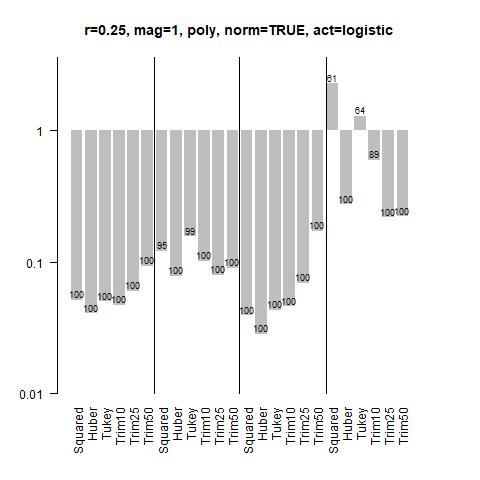}\\
\includegraphics[width=6.75cm,height=6.25cm]{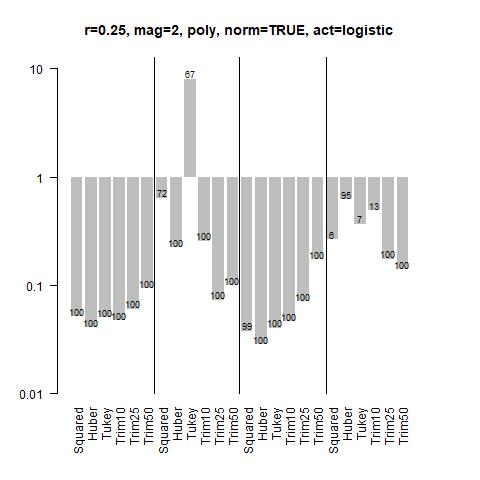} 
\includegraphics[width=6.75cm,height=6.25cm]{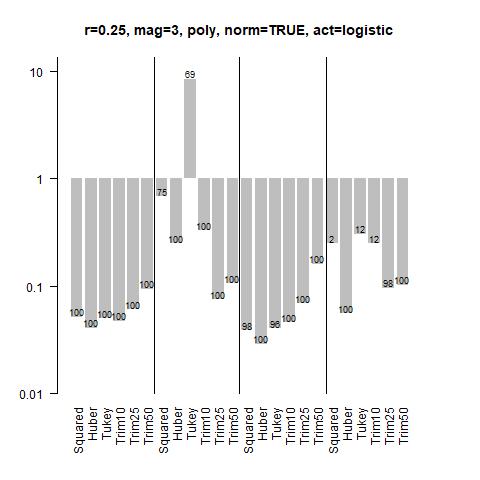} 
\end{center}
\caption{Results for $r=0.25$}\label{trimnn:n1000p50r25m1polynonlog}
\end{figure}

\begin{figure}[H]
\begin{center}
\includegraphics[width=6.75cm,height=6.25cm]{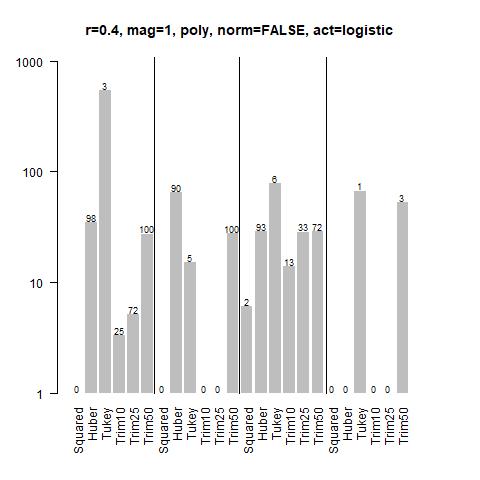}
\includegraphics[width=6.75cm,height=6.25cm]{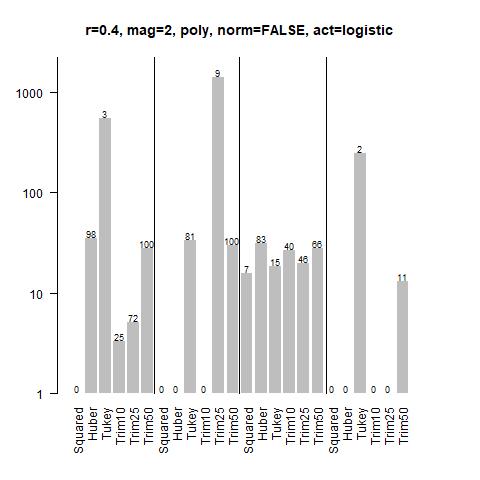} \\
\includegraphics[width=6.75cm,height=6.25cm]{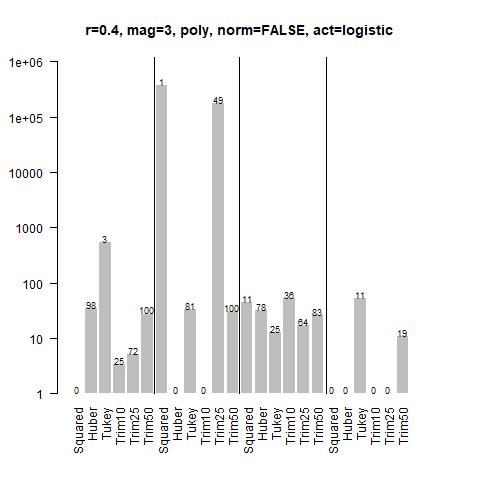} 
\includegraphics[width=6.75cm,height=6.25cm]{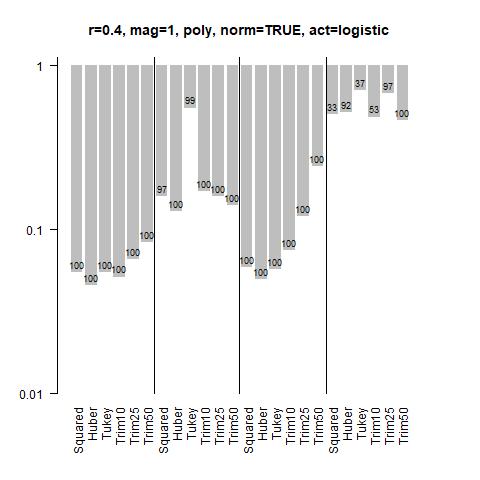}\\
\includegraphics[width=6.75cm,height=6.25cm]{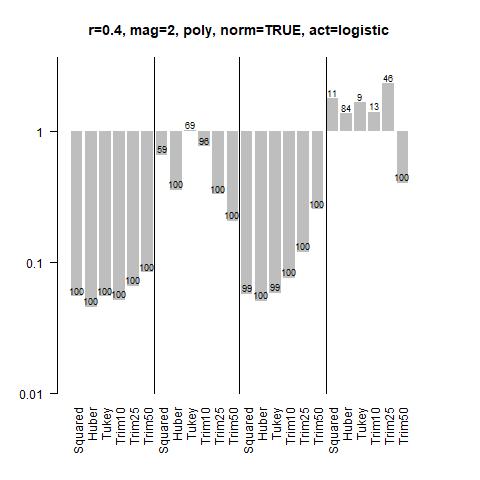} 
\includegraphics[width=6.75cm,height=6.25cm]{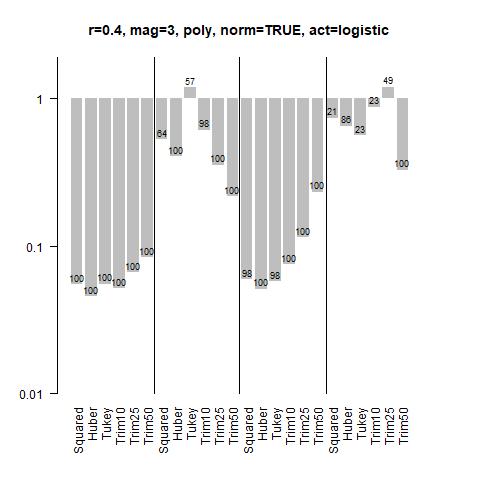} 
\end{center}
\caption{Results for $r=0.4$}\label{trimnn:n1000p50r40m1polynonlog}
\end{figure}

\subsubsection{Trigonometric function}

\begin{figure}[H]
\begin{center}
\includegraphics[width=6.75cm,height=6.25cm]{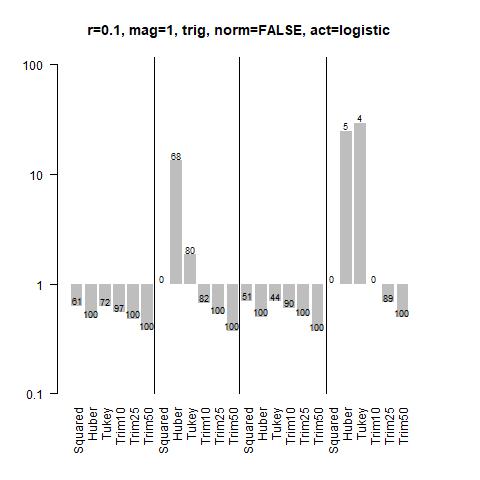}
\includegraphics[width=6.75cm,height=6.25cm]{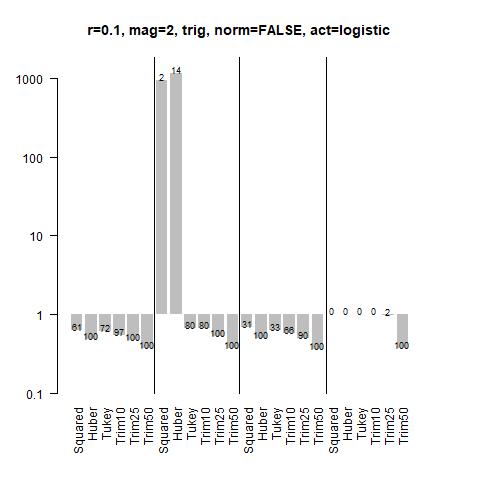} \\
\includegraphics[width=6.75cm,height=6.25cm]{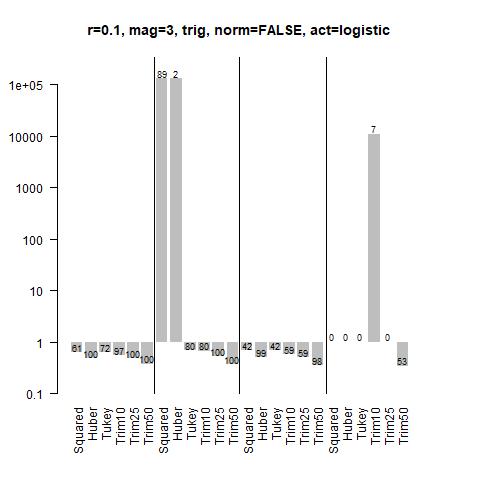} 
\includegraphics[width=6.75cm,height=6.25cm]{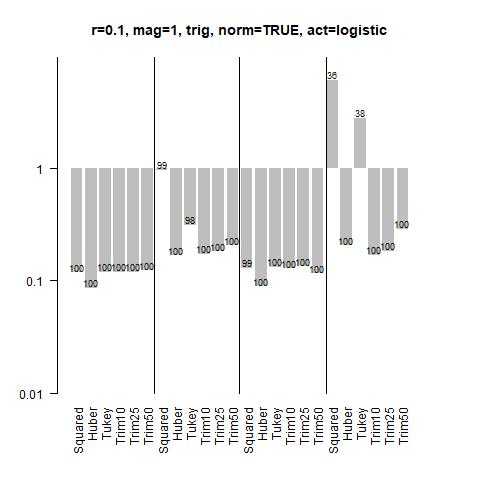}\\
\includegraphics[width=6.75cm,height=6.25cm]{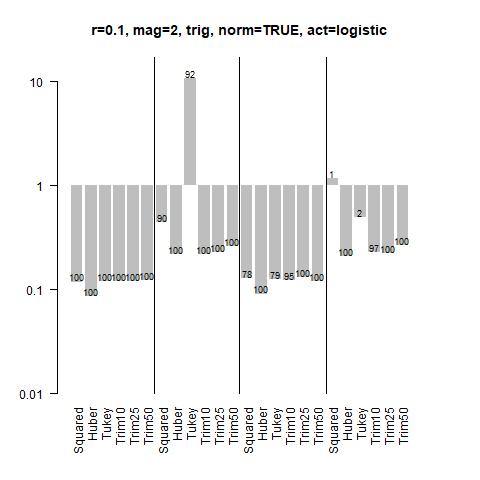} 
\includegraphics[width=6.75cm,height=6.25cm]{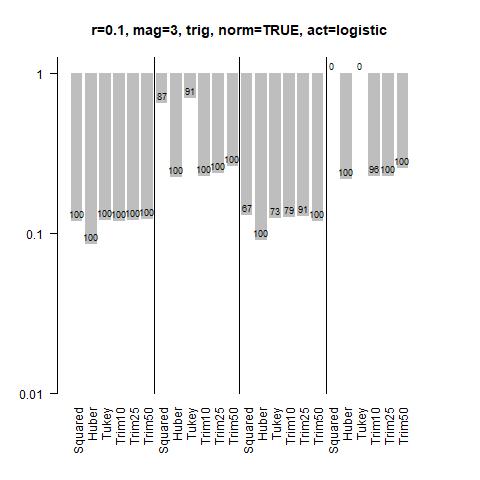} 
\end{center}
\caption{Results for $r=0.1$}\label{trimnn:n1000p50r10m1trignonlog}
\end{figure}

\begin{figure}[H]
\begin{center}
\includegraphics[width=6.75cm,height=6.25cm]{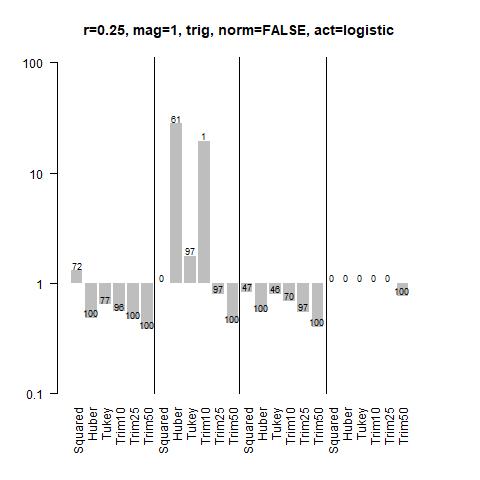}
\includegraphics[width=6.75cm,height=6.25cm]{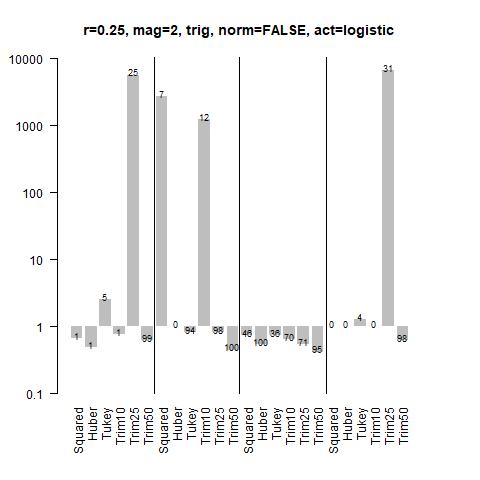} \\
\includegraphics[width=6.75cm,height=6.25cm]{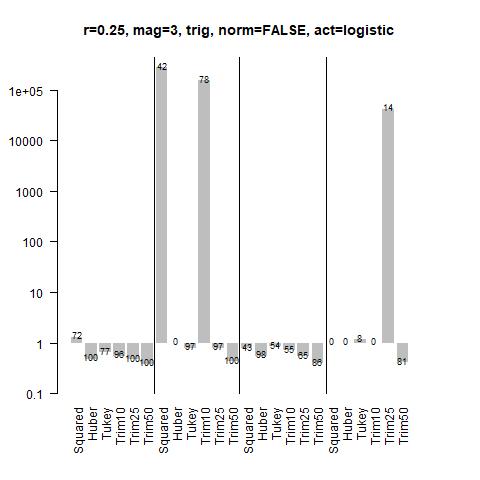} 
\includegraphics[width=6.75cm,height=6.25cm]{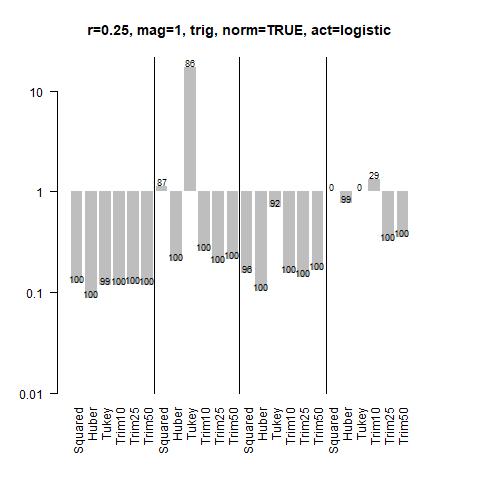}\\
\includegraphics[width=6.75cm,height=6.25cm]{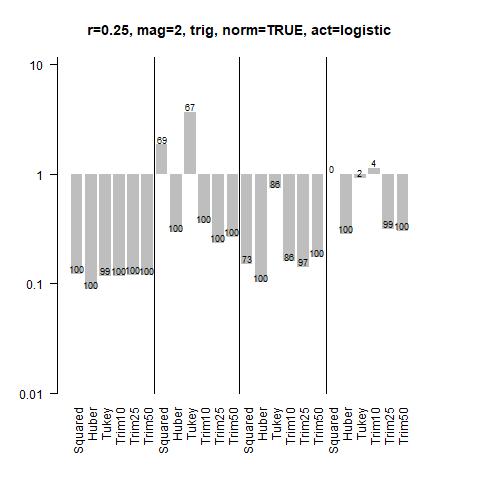} 
\includegraphics[width=6.75cm,height=6.25cm]{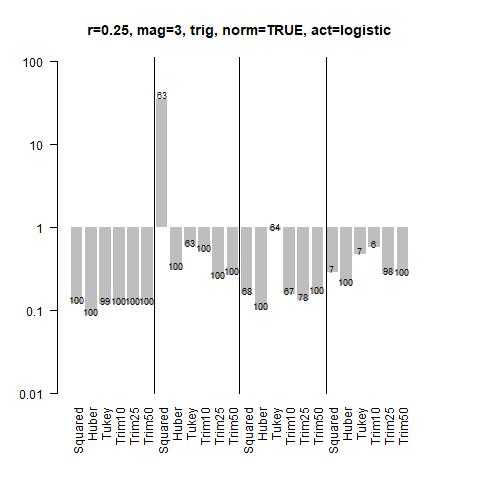} 
\end{center}
\caption{Results for $r=0.25$}\label{trimnn:n1000p50r25m1trignonlog}
\end{figure}

\begin{figure}[H]
\begin{center}
\includegraphics[width=6.75cm,height=6.25cm]{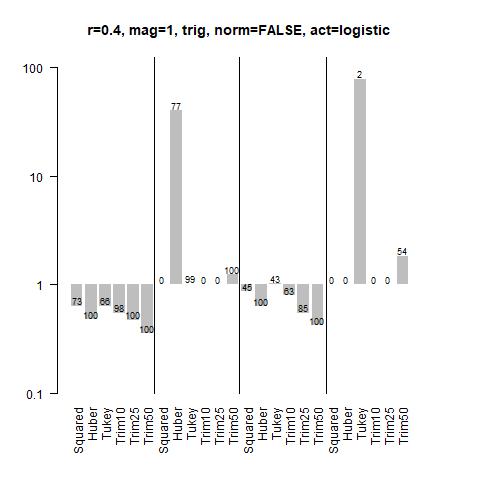}
\includegraphics[width=6.75cm,height=6.25cm]{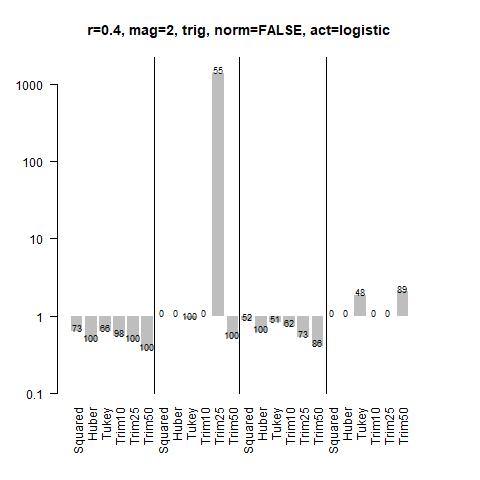} \\
\includegraphics[width=6.75cm,height=6.25cm]{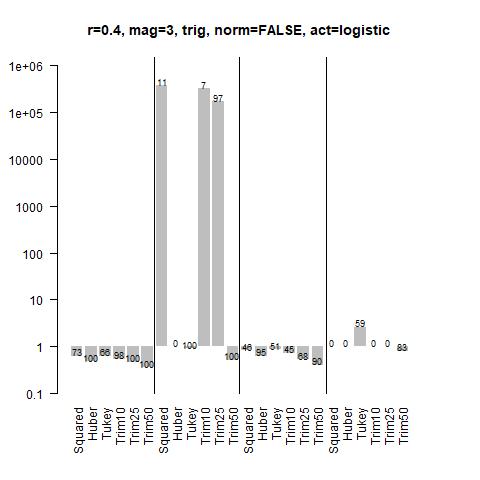} 
\includegraphics[width=6.75cm,height=6.25cm]{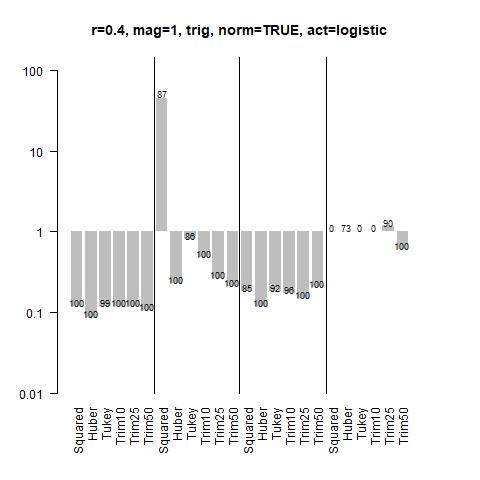}\\
\includegraphics[width=6.75cm,height=6.25cm]{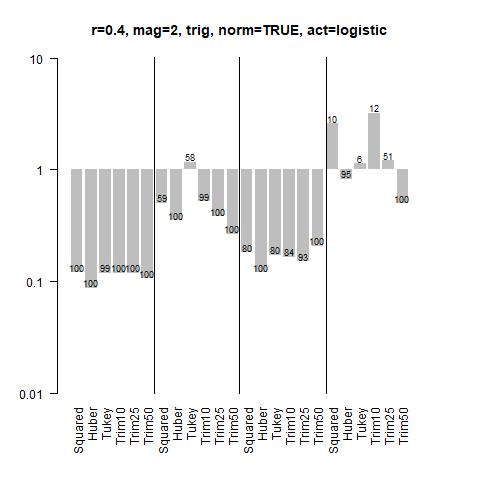} 
\includegraphics[width=6.75cm,height=6.25cm]{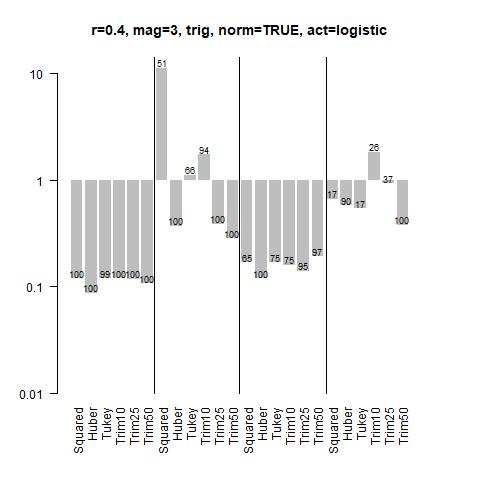} 
\end{center}
\caption{Results for $r=0.4$}\label{trimnn:n1000p50r40m1trignonlog}
\end{figure}

\subsection{Softplus activation function}

\subsubsection{Linear function}

\begin{figure}[H]
\label{trimnn:n1000p50r10m1linnonrelu}
\begin{center}
\includegraphics[width=6.75cm,height=6.25cm]{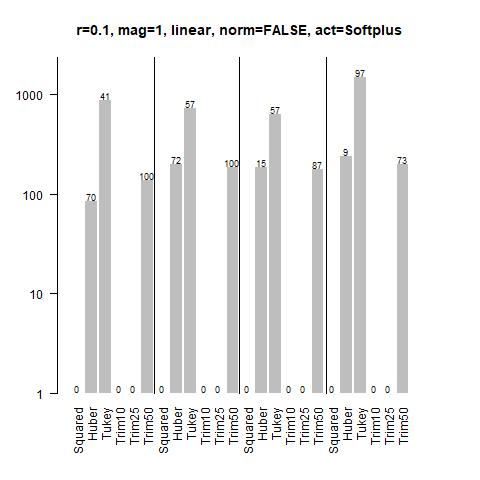}
\includegraphics[width=6.75cm,height=6.25cm]{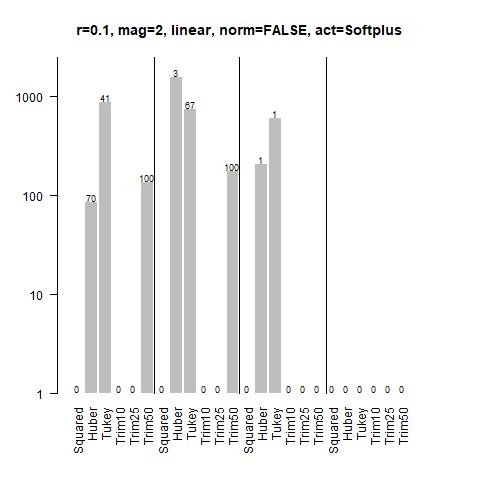} \\
\includegraphics[width=6.75cm,height=6.25cm]{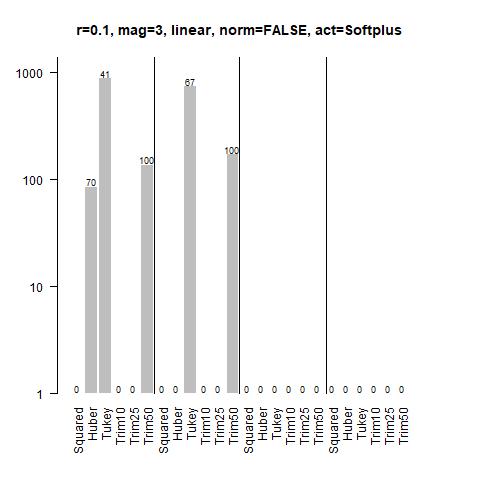} 
\includegraphics[width=6.75cm,height=6.25cm]{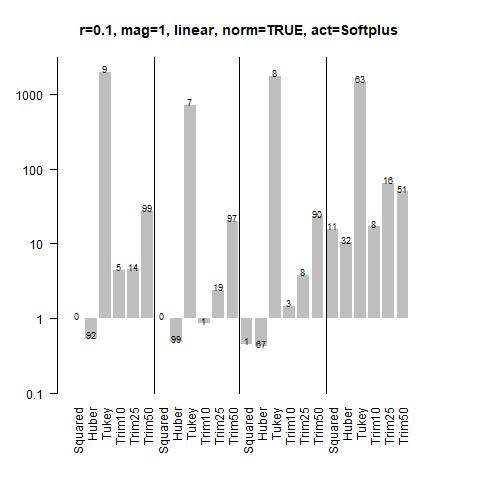}\\
\includegraphics[width=6.75cm,height=6.25cm]{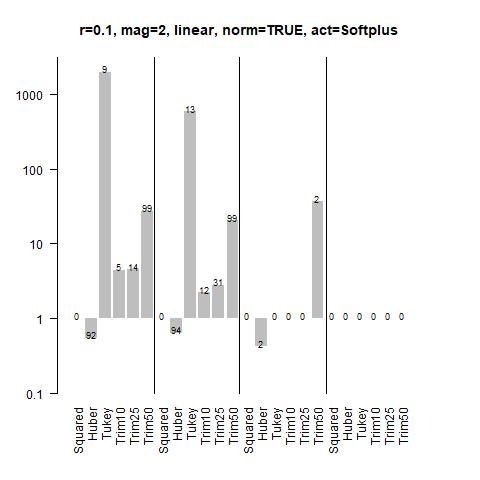} 
\includegraphics[width=6.75cm,height=6.25cm]{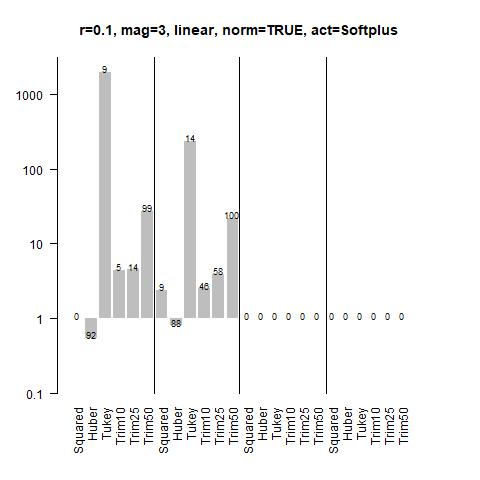} 
\end{center}
\caption{Results for $r=0.1$}
\end{figure}

\begin{figure}[H]
\label{trimnn:n1000p50r25m1linnonrelu}
\begin{center}
\includegraphics[width=6.75cm,height=6.25cm]{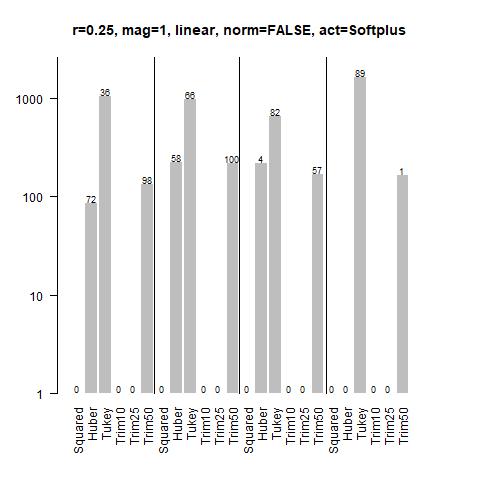}
\includegraphics[width=6.75cm,height=6.25cm]{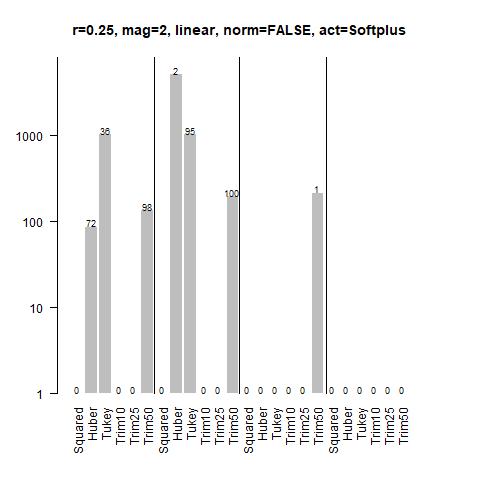} \\
\includegraphics[width=6.75cm,height=6.25cm]{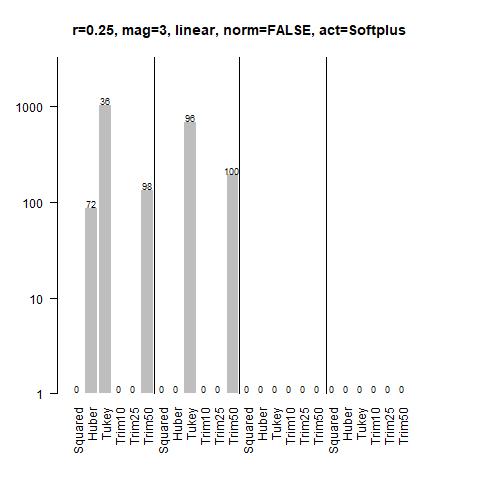} 
\includegraphics[width=6.75cm,height=6.25cm]{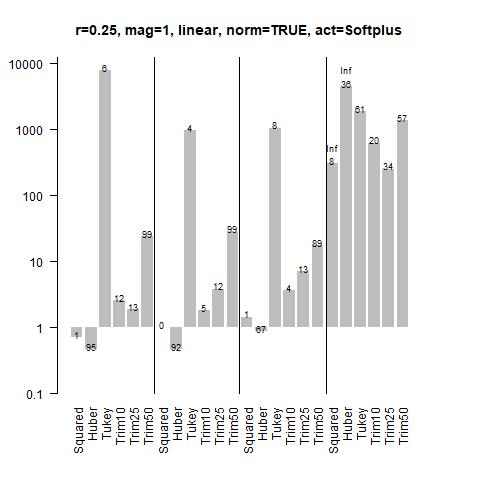}\\
\includegraphics[width=6.75cm,height=6.25cm]{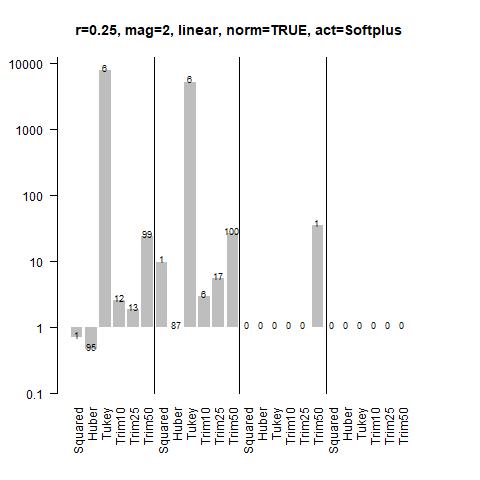} 
\includegraphics[width=6.75cm,height=6.25cm]{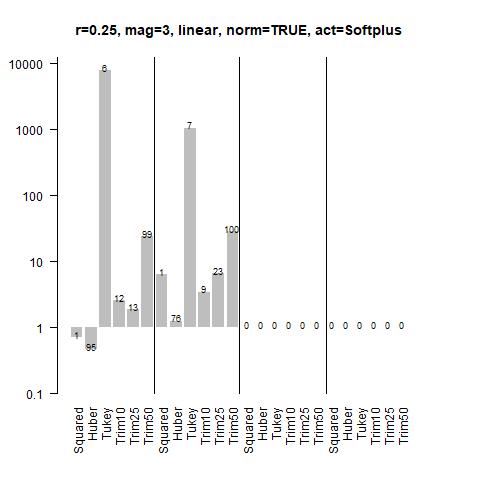} 
\end{center}
\caption{Results for $r=0.25$}
\end{figure}

\begin{figure}[H]
\label{trimnn:n1000p50r40m1linnonrelu}
\begin{center}
\includegraphics[width=6.75cm,height=6.25cm]{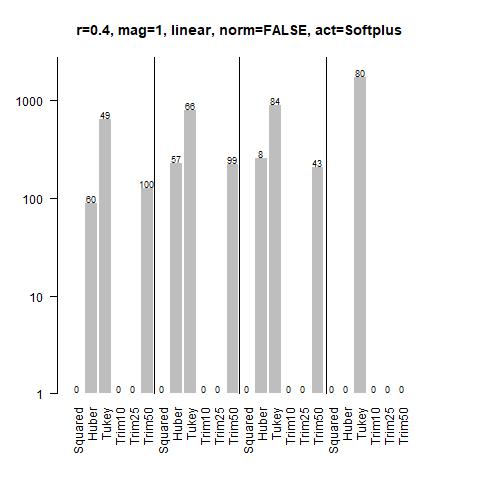}
\includegraphics[width=6.75cm,height=6.25cm]{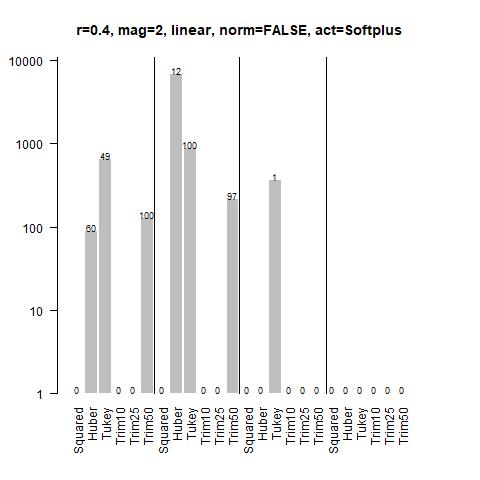} \\
\includegraphics[width=6.75cm,height=6.25cm]{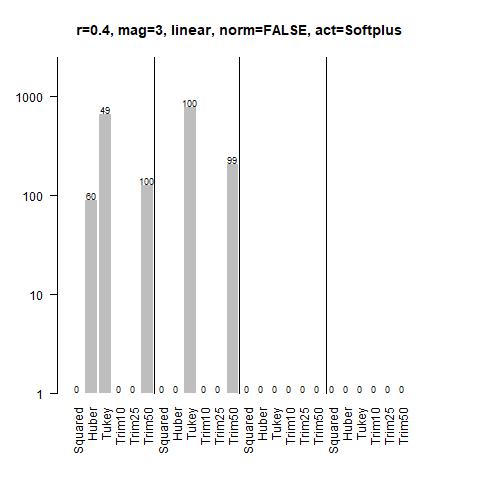} 
\includegraphics[width=6.75cm,height=6.25cm]{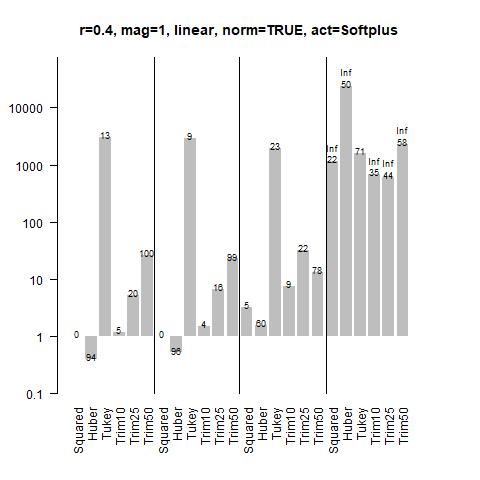}\\
\includegraphics[width=6.75cm,height=6.25cm]{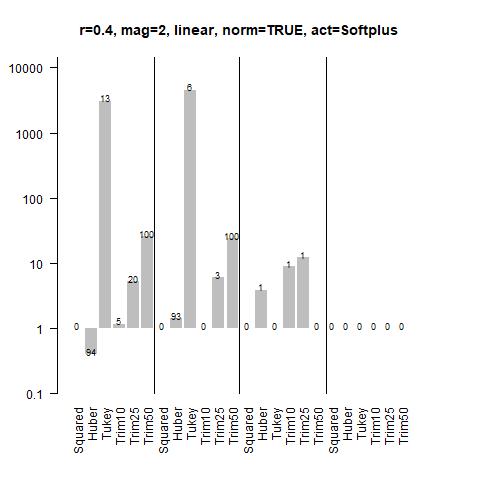} 
\includegraphics[width=6.75cm,height=6.25cm]{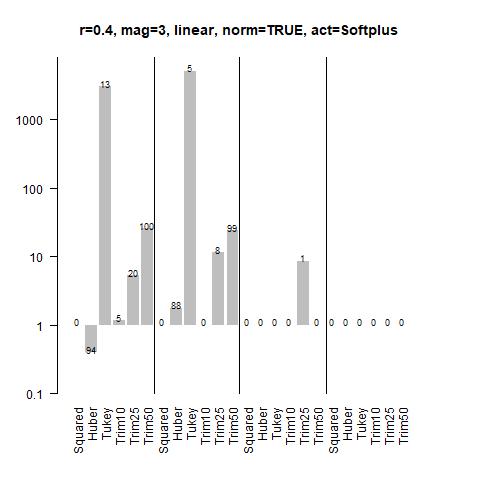} 
\end{center}
\caption{Results for $r=0.4$}
\end{figure}

\subsubsection{Polynomial function}

\begin{figure}[H]
\label{trimnn:n1000p50r10m1polynonrelu}
\begin{center}
\includegraphics[width=6.75cm,height=6.25cm]{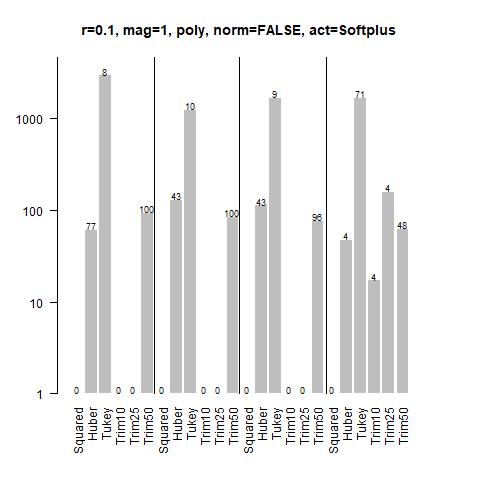}
\includegraphics[width=6.75cm,height=6.25cm]{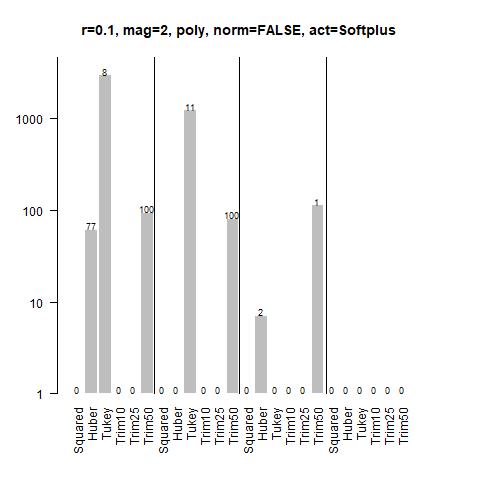} \\
\includegraphics[width=6.75cm,height=6.25cm]{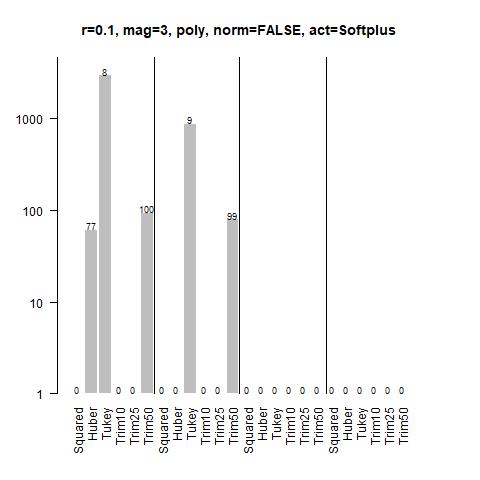} 
\includegraphics[width=6.75cm,height=6.25cm]{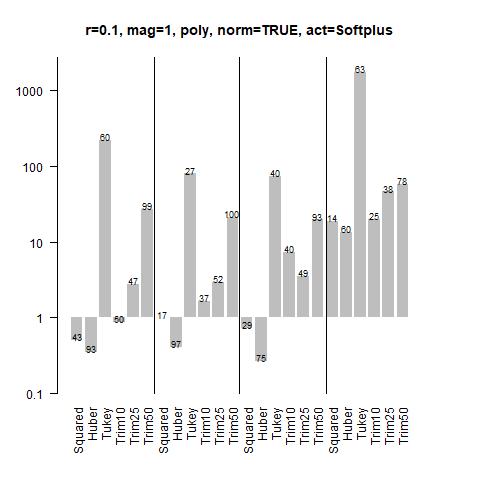}\\
\includegraphics[width=6.75cm,height=6.25cm]{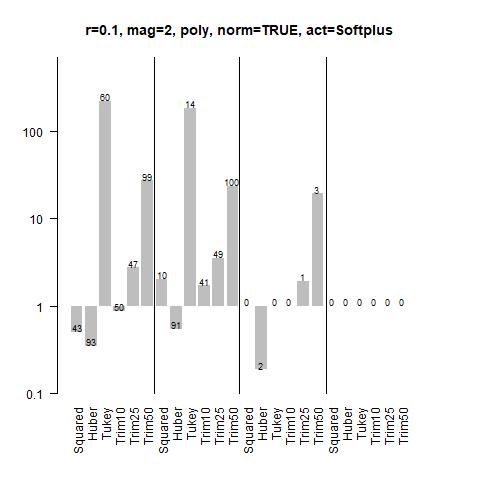} 
\includegraphics[width=6.75cm,height=6.25cm]{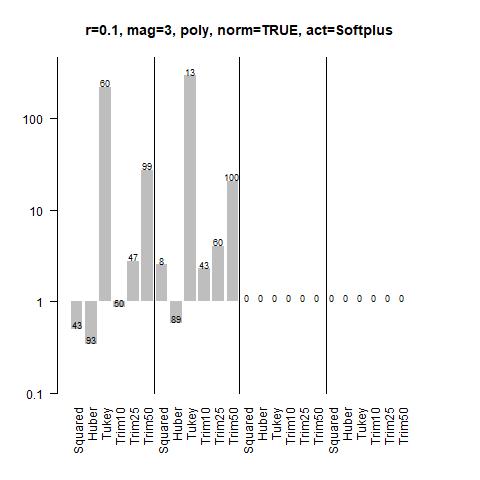} 
\end{center}
\caption{Results for $r=0.1$}
\end{figure}

\begin{figure}[H]
\label{trimnn:n1000p50r25m1polynonrelu}
\begin{center}
\includegraphics[width=6.75cm,height=6.25cm]{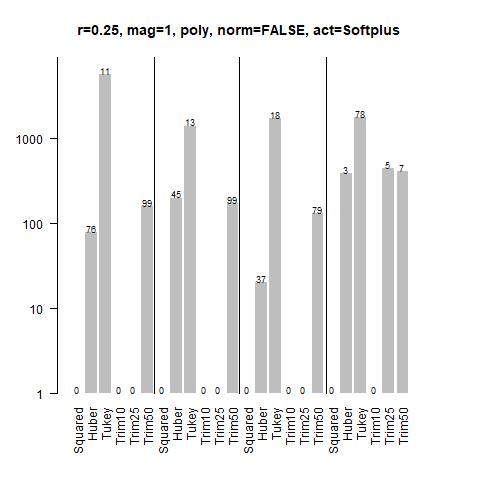}
\includegraphics[width=6.75cm,height=6.25cm]{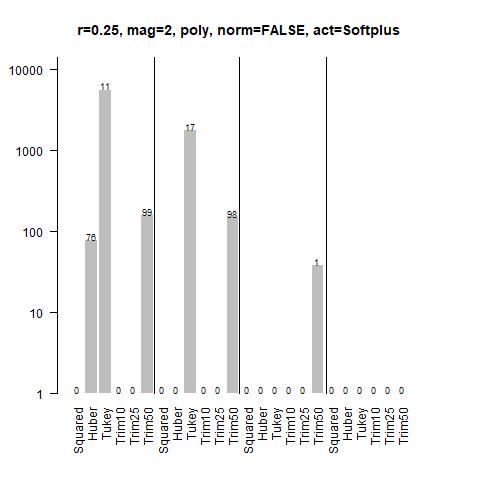} \\
\includegraphics[width=6.75cm,height=6.25cm]{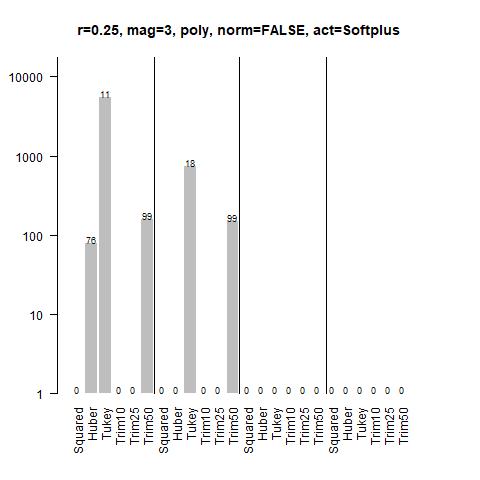} 
\includegraphics[width=6.75cm,height=6.25cm]{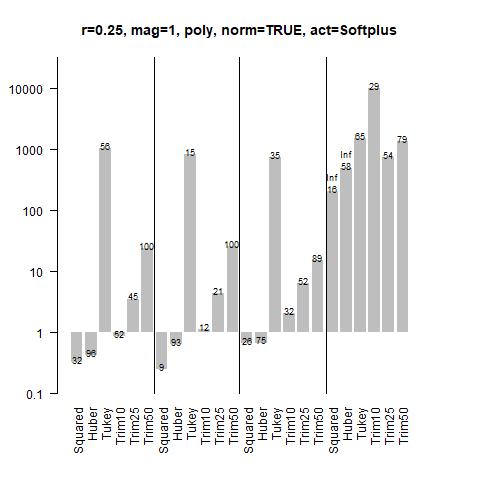}\\
\includegraphics[width=6.75cm,height=6.25cm]{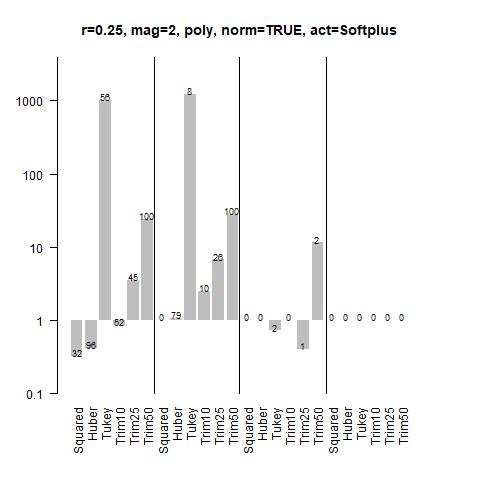} 
\includegraphics[width=6.75cm,height=6.25cm]{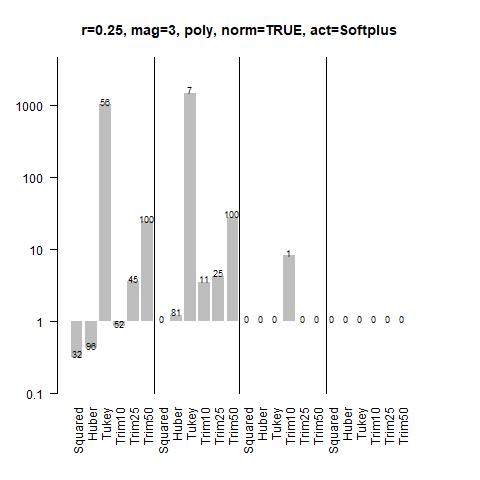} 
\end{center}
\caption{Results for $r=0.25$}
\end{figure}

\begin{figure}[H]
\label{trimnn:n1000p50r40m1polynonrelu}
\begin{center}
\includegraphics[width=6.75cm,height=6.25cm]{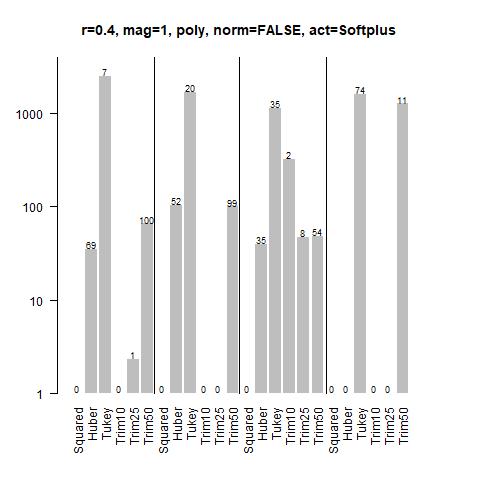}
\includegraphics[width=6.75cm,height=6.25cm]{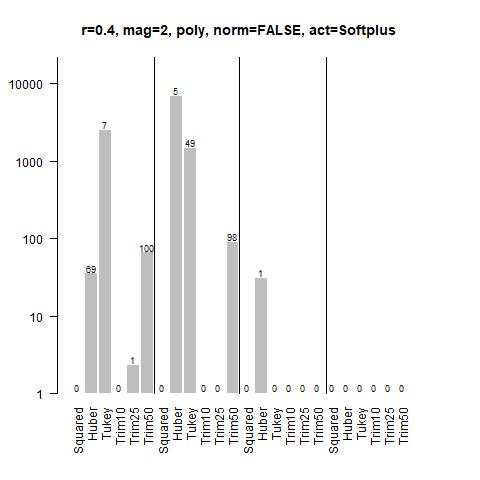} \\
\includegraphics[width=6.75cm,height=6.25cm]{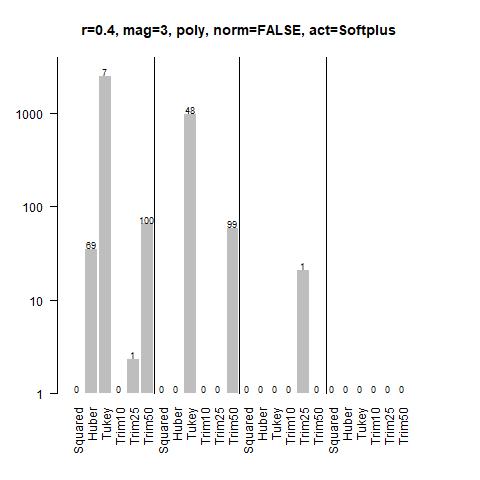} 
\includegraphics[width=6.75cm,height=6.25cm]{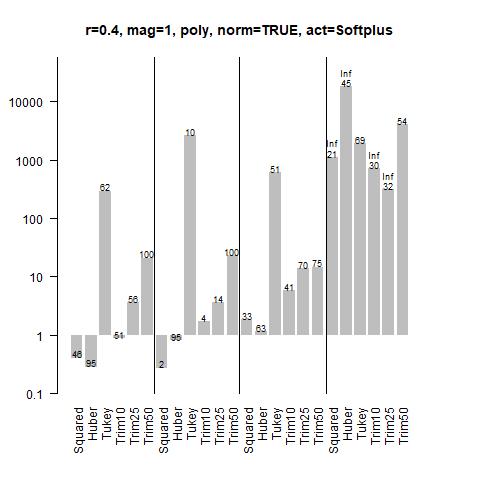}\\
\includegraphics[width=6.75cm,height=6.25cm]{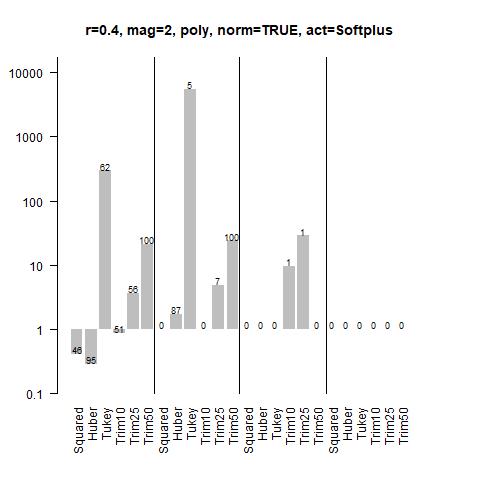} 
\includegraphics[width=6.75cm,height=6.25cm]{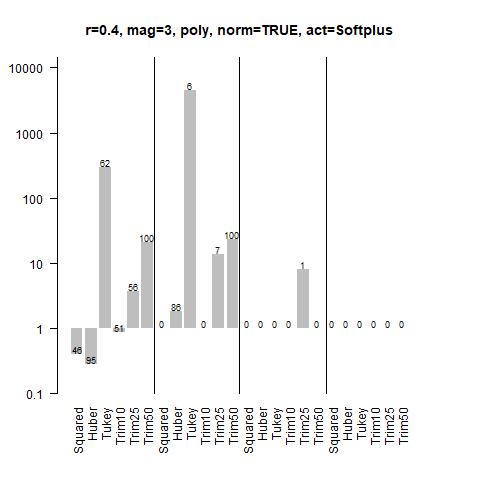} 
\end{center}
\caption{Results for $r=0.4$}
\end{figure}

\subsubsection{Trigonometric function}

\begin{figure}[H]
\label{trimnn:n1000p50r10m1trignonrelu}
\begin{center}
\includegraphics[width=6.75cm,height=6.25cm]{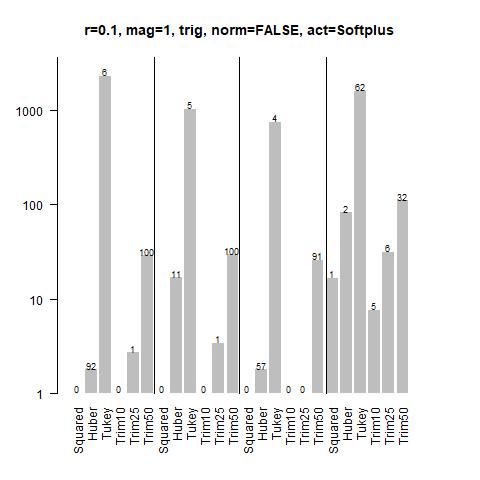}
\includegraphics[width=6.75cm,height=6.25cm]{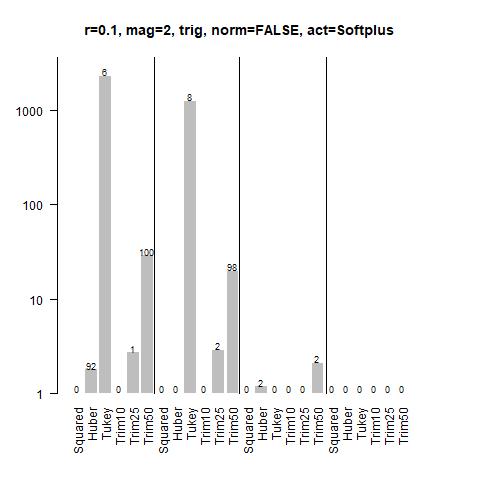} \\
\includegraphics[width=6.75cm,height=6.25cm]{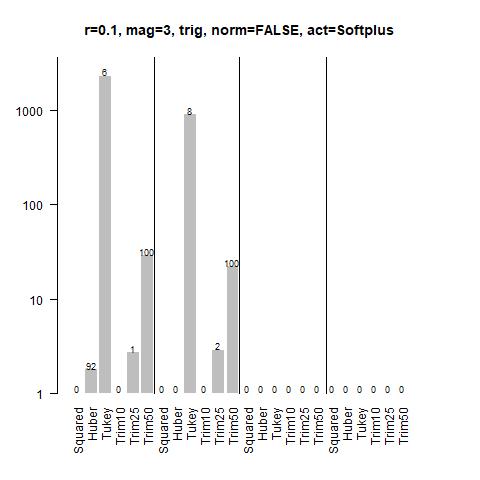} 
\includegraphics[width=6.75cm,height=6.25cm]{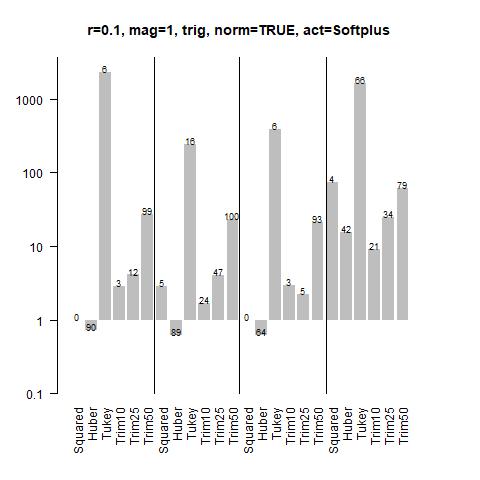}\\
\includegraphics[width=6.75cm,height=6.25cm]{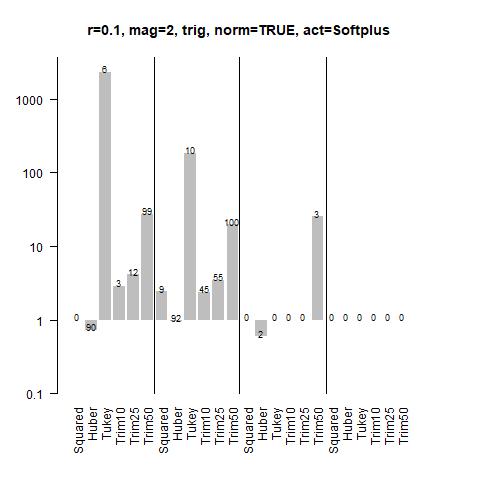} 
\includegraphics[width=6.75cm,height=6.25cm]{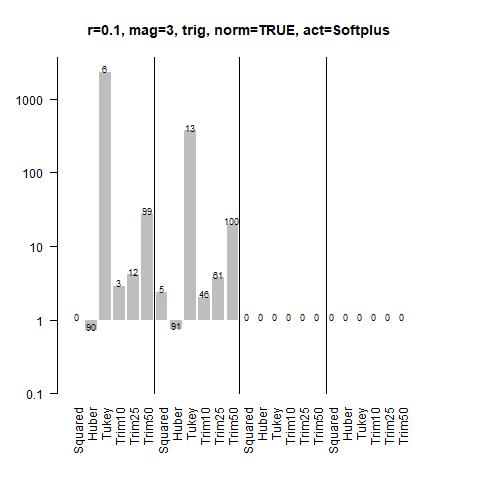} 
\end{center}
\caption{Results for $r=0.1$}
\end{figure}

\begin{figure}[H]
\label{trimnn:n1000p50r25m1trignonrelu}
\begin{center}
\includegraphics[width=6.75cm,height=6.25cm]{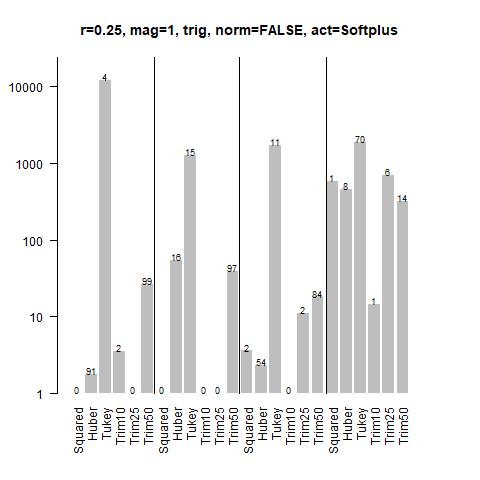}
\includegraphics[width=6.75cm,height=6.25cm]{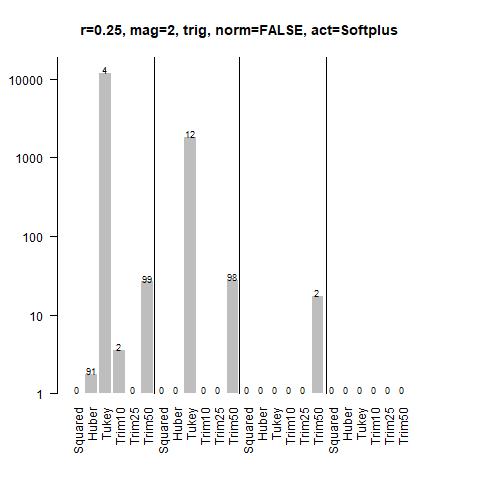} \\
\includegraphics[width=6.75cm,height=6.25cm]{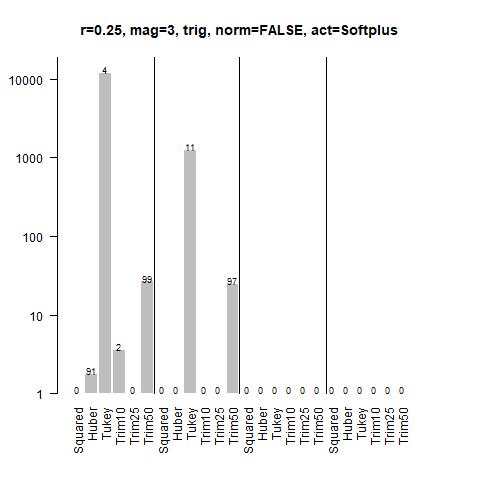} 
\includegraphics[width=6.75cm,height=6.25cm]{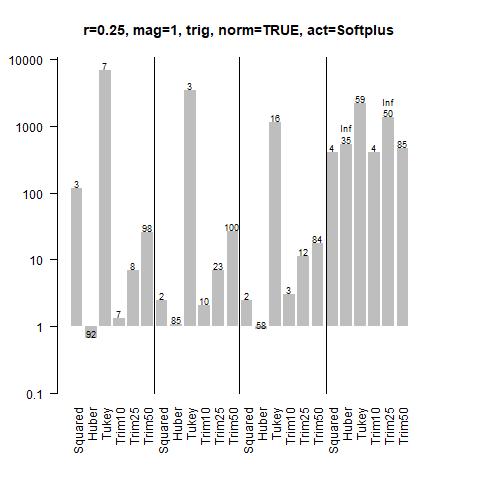}\\
\includegraphics[width=6.75cm,height=6.25cm]{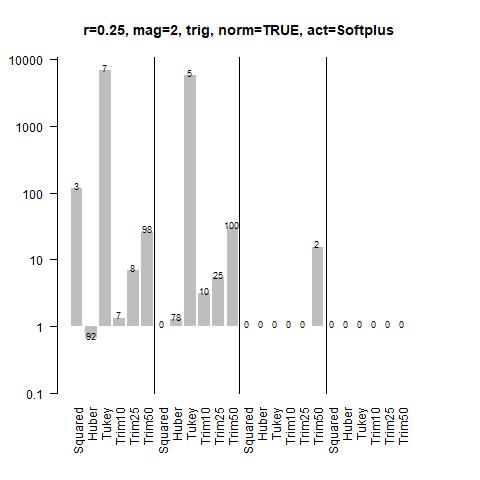} 
\includegraphics[width=6.75cm,height=6.25cm]{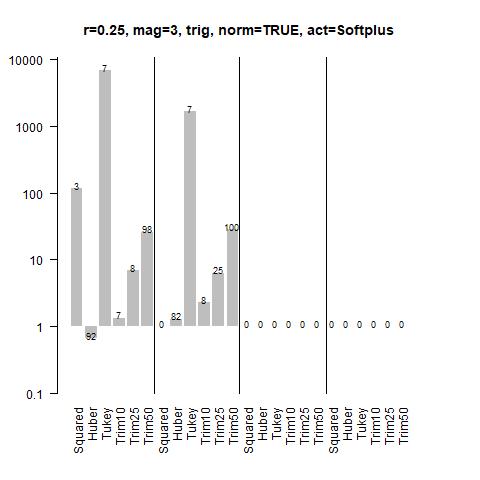} 
\end{center}
\caption{Results for $r=0.25$}
\end{figure}

\begin{figure}[H]
\begin{center}
\includegraphics[width=6.75cm,height=6.25cm]{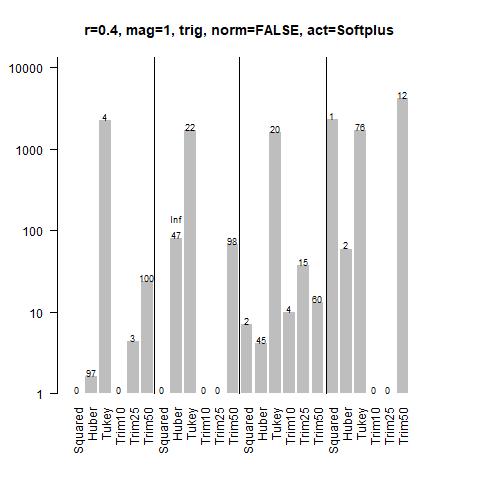}
\includegraphics[width=6.75cm,height=6.25cm]{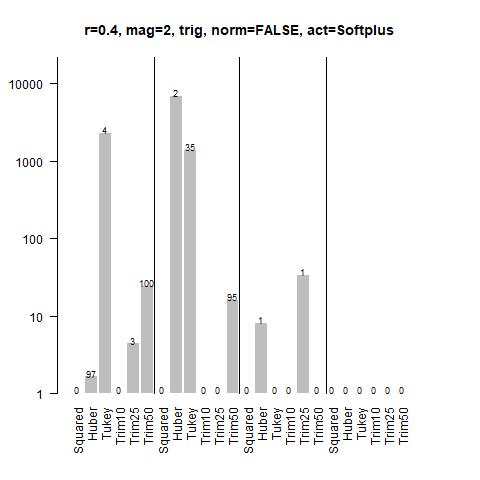} \\
\includegraphics[width=6.75cm,height=6.25cm]{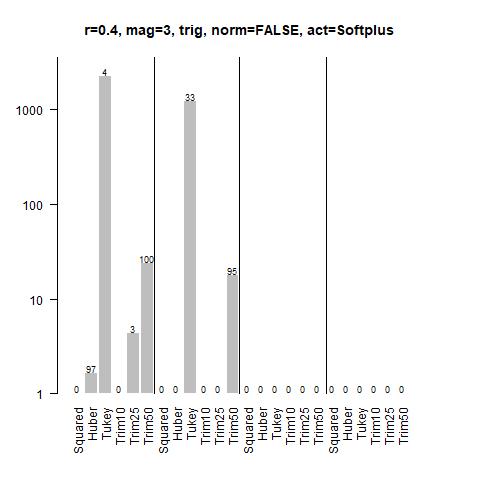} 
\includegraphics[width=6.75cm,height=6.25cm]{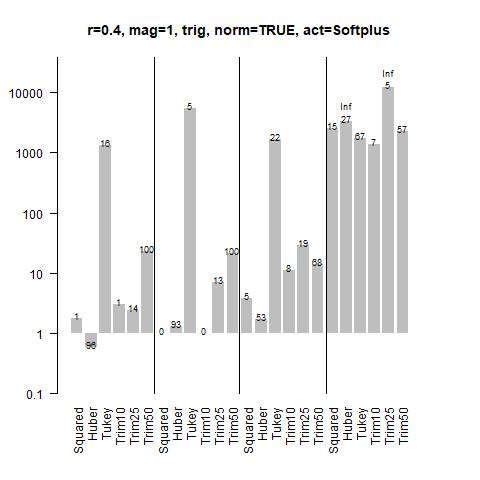}\\
\includegraphics[width=6.75cm,height=6.25cm]{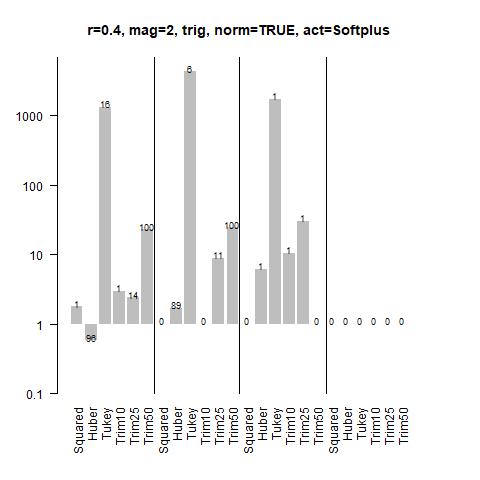} 
\includegraphics[width=6.75cm,height=6.25cm]{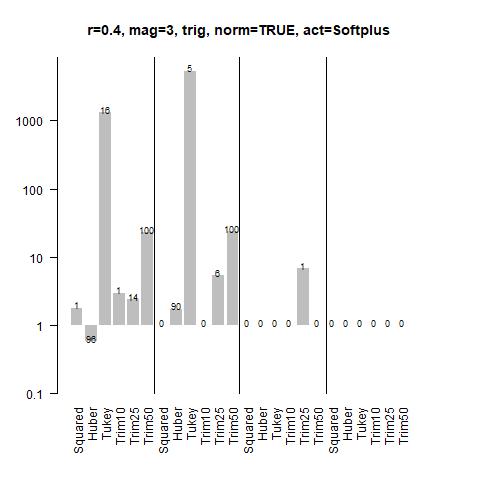} 
\end{center}
\caption{Results for $r=0.4$}\label{trimnn:n1000p50r40m1trignonrelu}
\end{figure}

\section{Simulation results for $n=1000$ and $p=50$, deep network: Test loss} \label{trimnn:secloss100050deep}

\subsection{Logistic activation function}

\subsubsection{Linear function}

\begin{figure}[H]
\label{trimnn:n1000p50r10m1linnonlogdeep}
\begin{center}
\includegraphics[width=6.75cm,height=6.25cm]{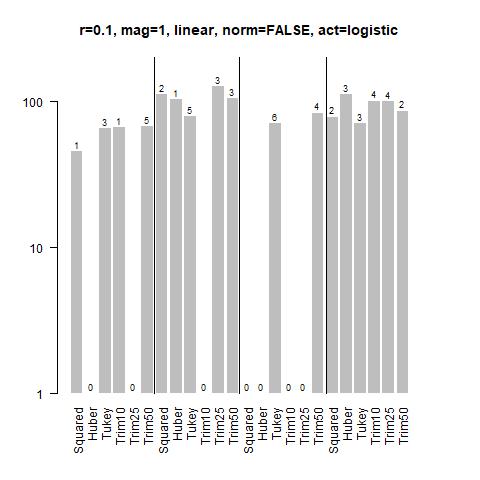}
\includegraphics[width=6.75cm,height=6.25cm]{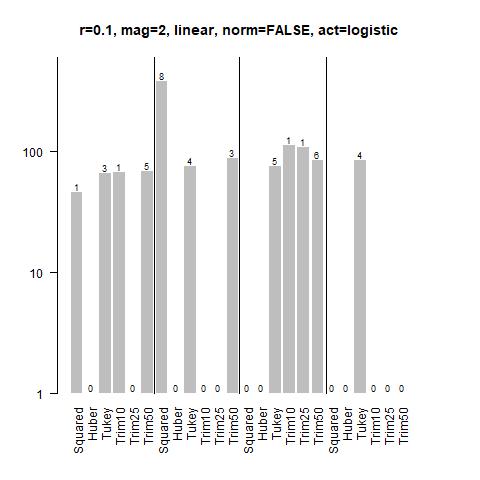} \\
\includegraphics[width=6.75cm,height=6.25cm]{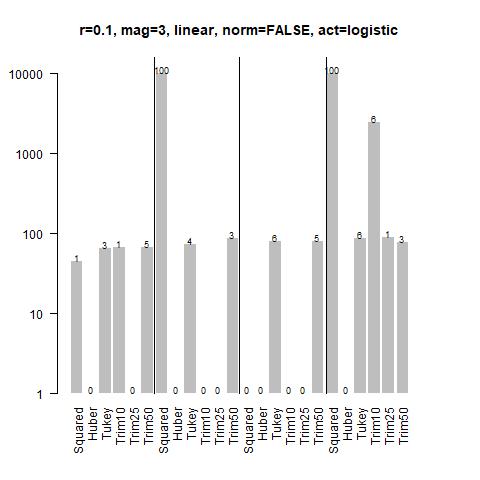} 
\includegraphics[width=6.75cm,height=6.25cm]{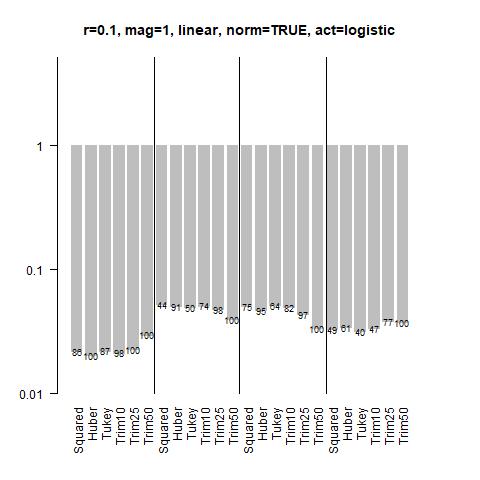}\\
\includegraphics[width=6.75cm,height=6.25cm]{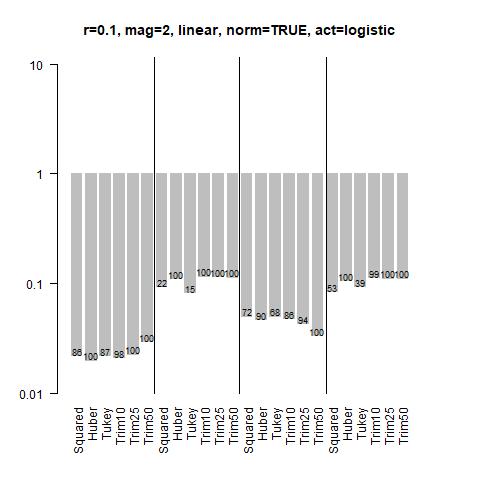} 
\includegraphics[width=6.75cm,height=6.25cm]{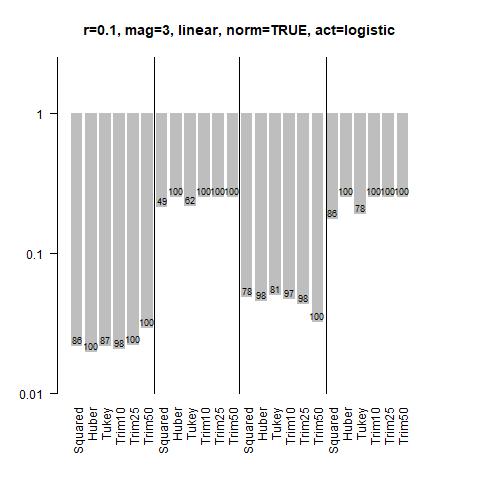} 
\end{center}
\caption{Results for $r=0.1$}
\end{figure}

\begin{figure}[H]
\label{trimnn:n1000p50r25m1linnonlogdeep}
\begin{center}
\includegraphics[width=6.75cm,height=6.25cm]{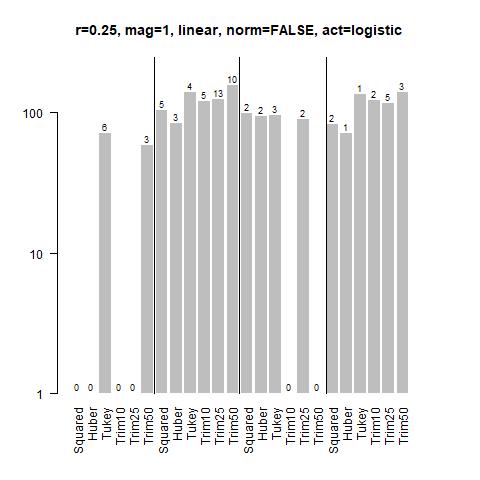}
\includegraphics[width=6.75cm,height=6.25cm]{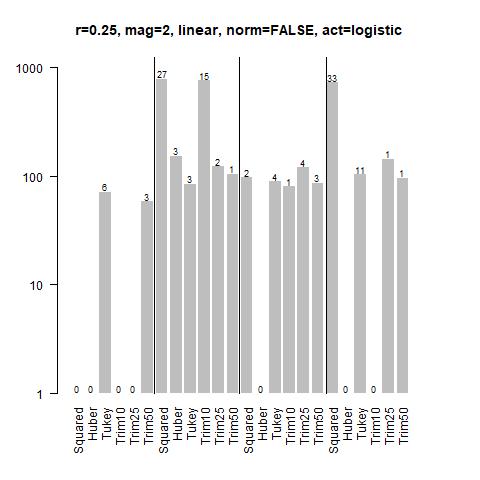} \\
\includegraphics[width=6.75cm,height=6.25cm]{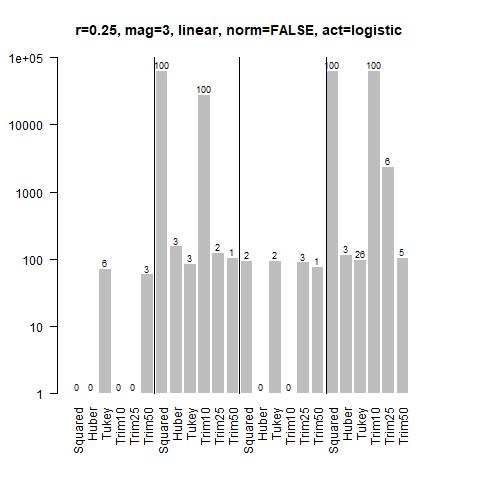} 
\includegraphics[width=6.75cm,height=6.25cm]{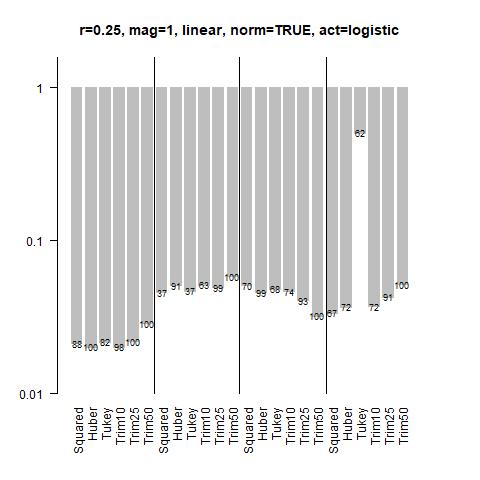}\\
\includegraphics[width=6.75cm,height=6.25cm]{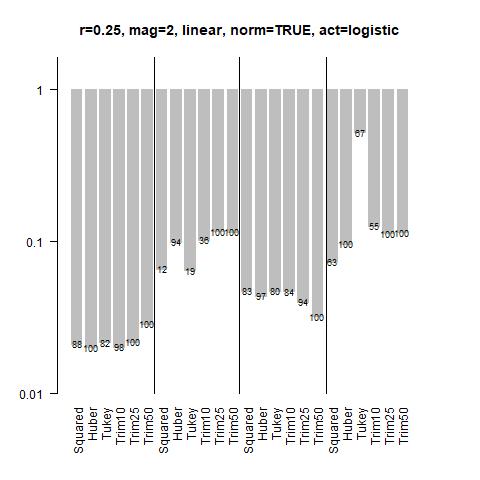} 
\includegraphics[width=6.75cm,height=6.25cm]{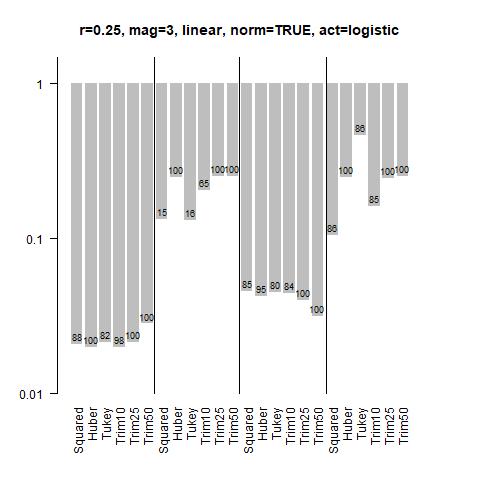} 
\end{center}
\caption{Results for $r=0.25$}
\end{figure}

\begin{figure}[H]
\label{trimnn:n1000p50r40m1linnonlogdeep}
\begin{center}
\includegraphics[width=6.75cm,height=6.25cm]{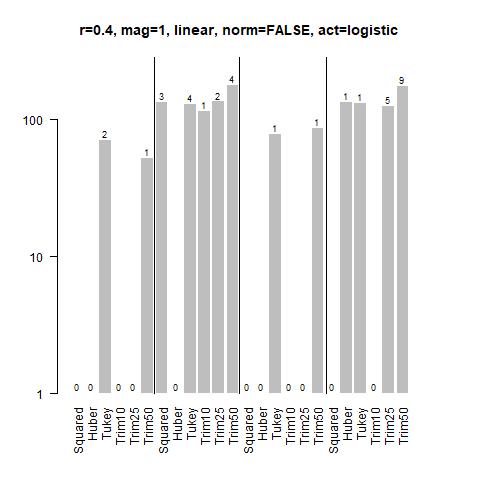}
\includegraphics[width=6.75cm,height=6.25cm]{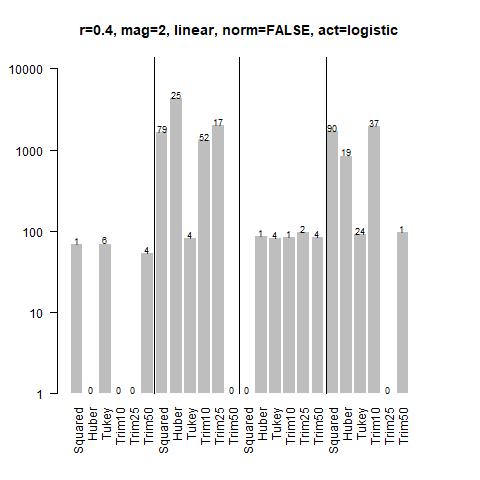} \\
\includegraphics[width=6.75cm,height=6.25cm]{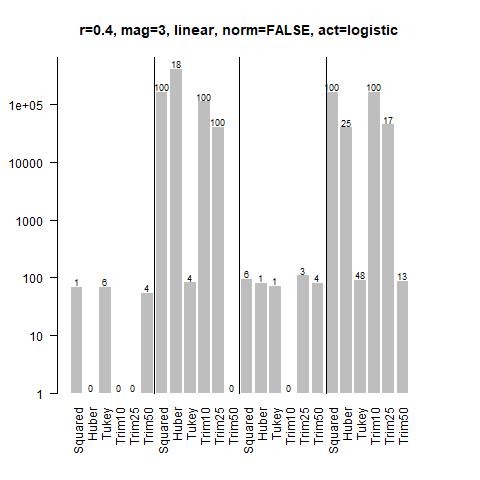} 
\includegraphics[width=6.75cm,height=6.25cm]{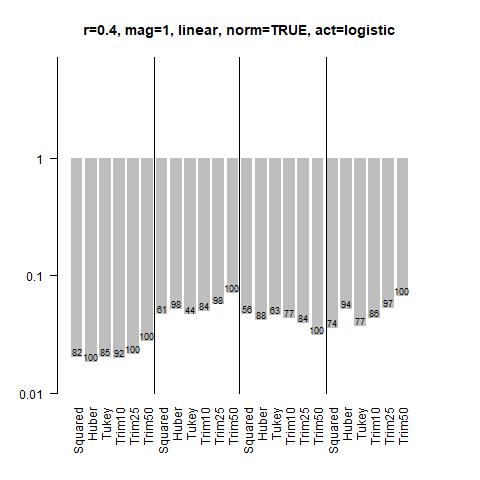}\\
\includegraphics[width=6.75cm,height=6.25cm]{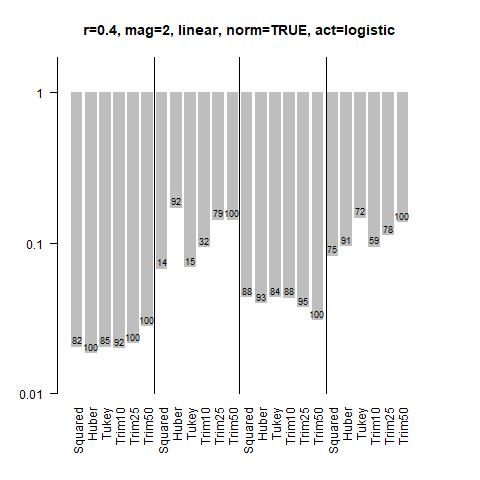} 
\includegraphics[width=6.75cm,height=6.25cm]{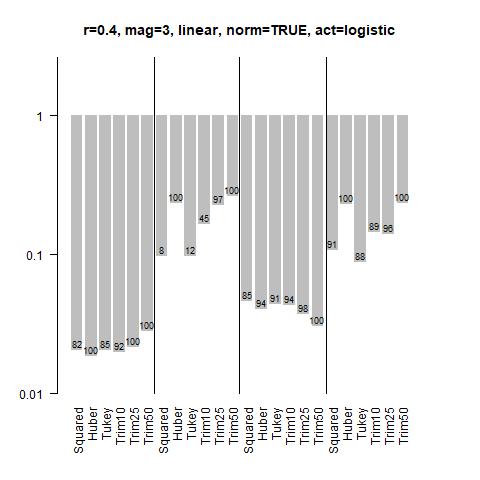} 
\end{center}
\caption{Results for $r=0.4$}
\end{figure}

\subsubsection{Polynomial function}

\begin{figure}[H]
\label{trimnn:n1000p50r10m1polynonlogdeep}
\begin{center}
\includegraphics[width=6.75cm,height=6.25cm]{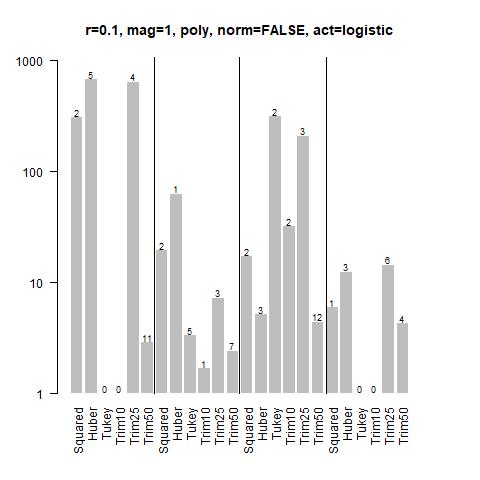}
\includegraphics[width=6.75cm,height=6.25cm]{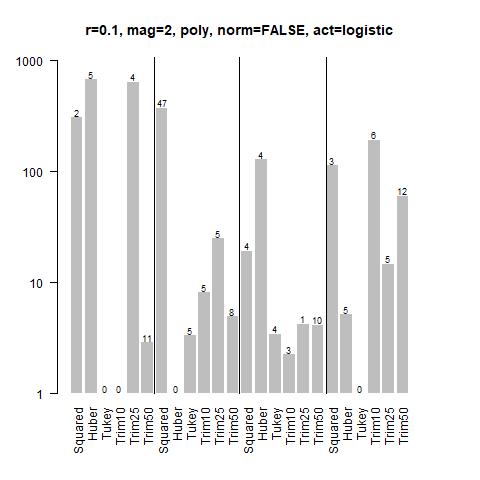} \\
\includegraphics[width=6.75cm,height=6.25cm]{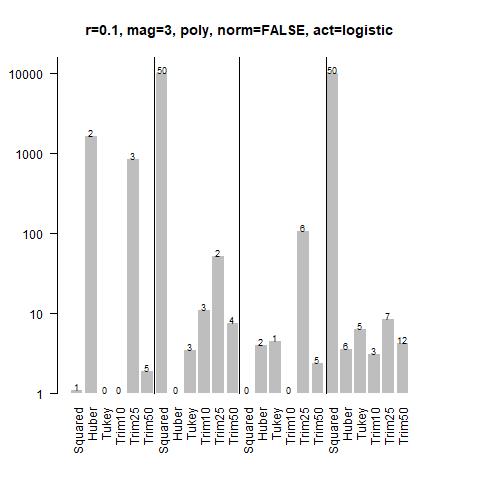} 
\includegraphics[width=6.75cm,height=6.25cm]{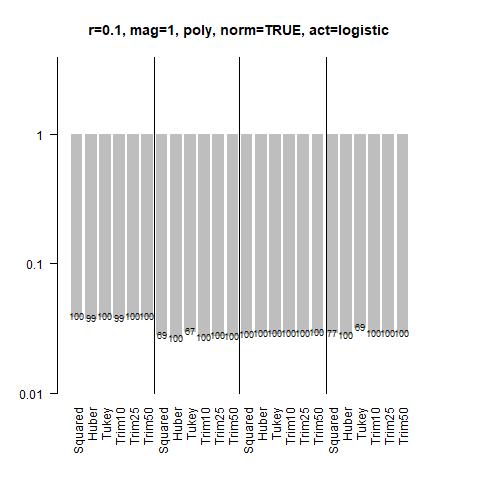}\\
\includegraphics[width=6.75cm,height=6.25cm]{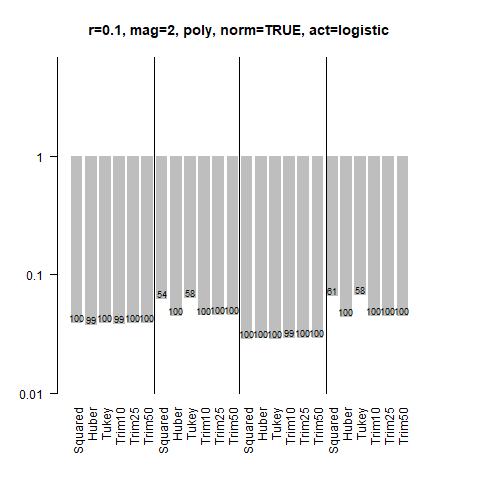} 
\includegraphics[width=6.75cm,height=6.25cm]{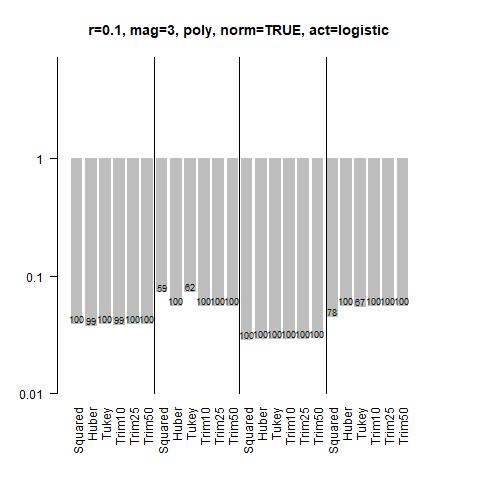} 
\end{center}
\caption{Results for $r=0.1$}
\end{figure}

\begin{figure}[H]
\label{trimnn:n1000p50r25m1polynonlogdeep}
\begin{center}
\includegraphics[width=6.75cm,height=6.25cm]{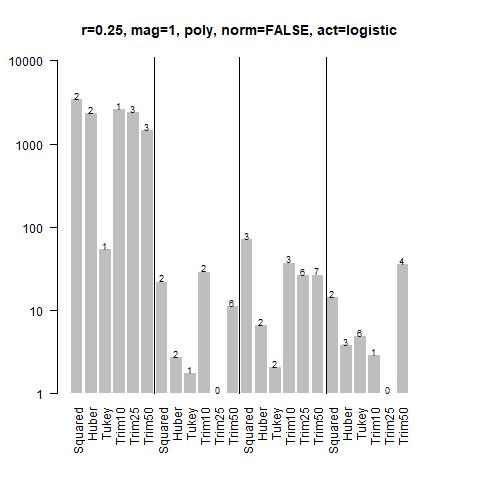}
\includegraphics[width=6.75cm,height=6.25cm]{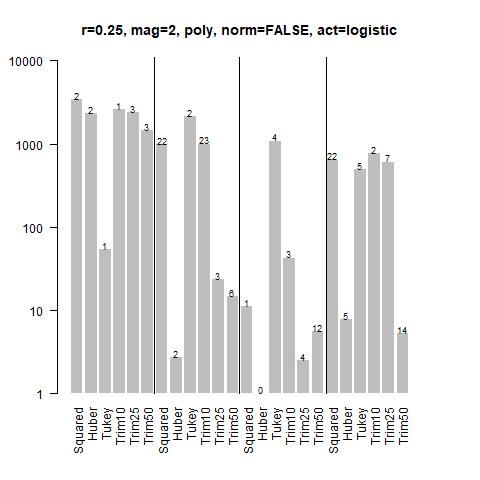} \\
\includegraphics[width=6.75cm,height=6.25cm]{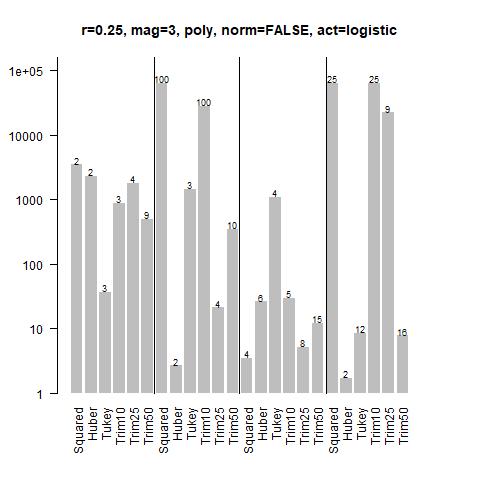} 
\includegraphics[width=6.75cm,height=6.25cm]{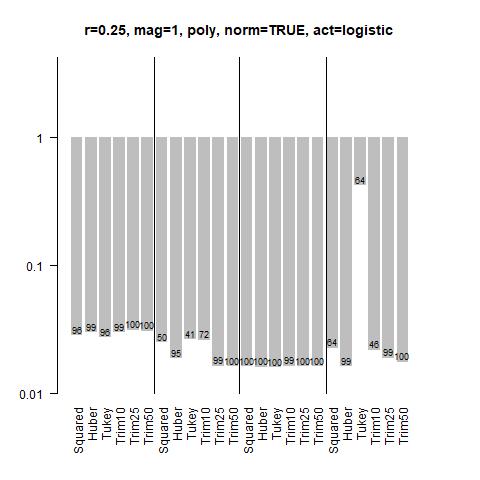}\\
\includegraphics[width=6.75cm,height=6.25cm]{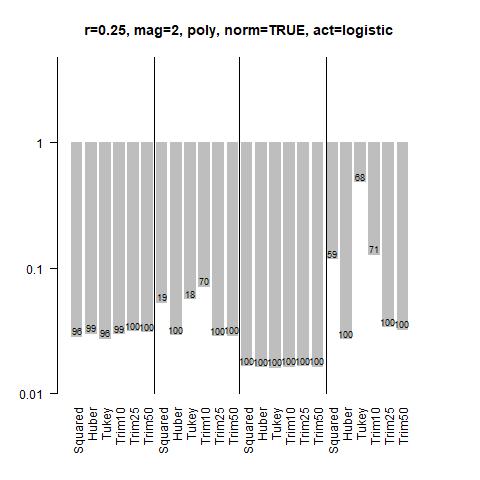} 
\includegraphics[width=6.75cm,height=6.25cm]{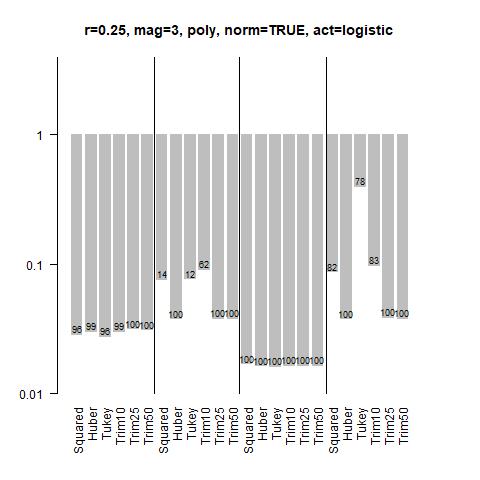} 
\end{center}
\caption{Results for $r=0.25$}
\end{figure}

\begin{figure}[H]
\label{trimnn:n1000p50r40m1polynonlogdeep}
\begin{center}
\includegraphics[width=6.75cm,height=6.25cm]{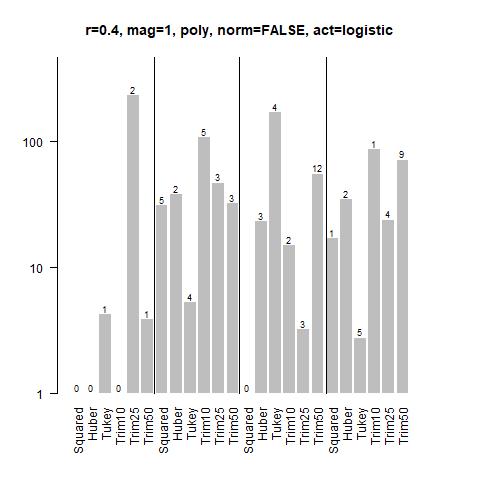}
\includegraphics[width=6.75cm,height=6.25cm]{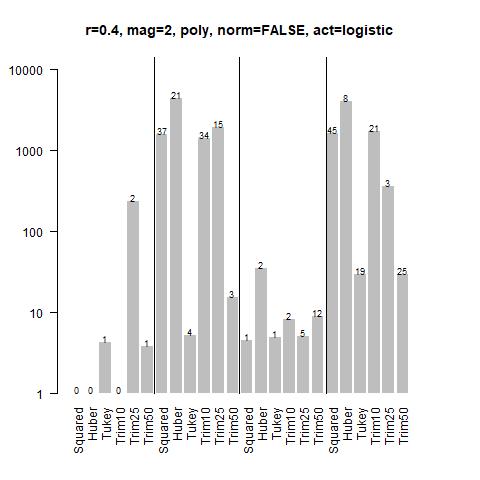} \\
\includegraphics[width=6.75cm,height=6.25cm]{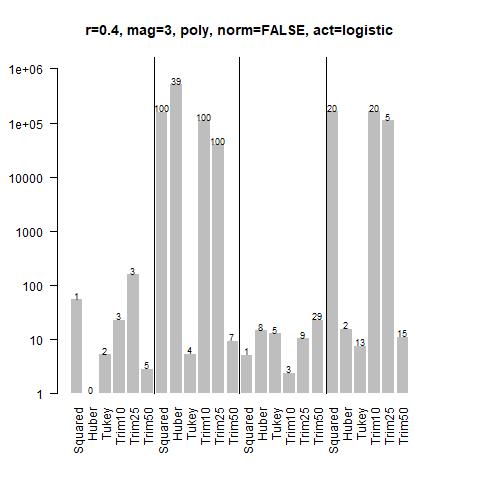} 
\includegraphics[width=6.75cm,height=6.25cm]{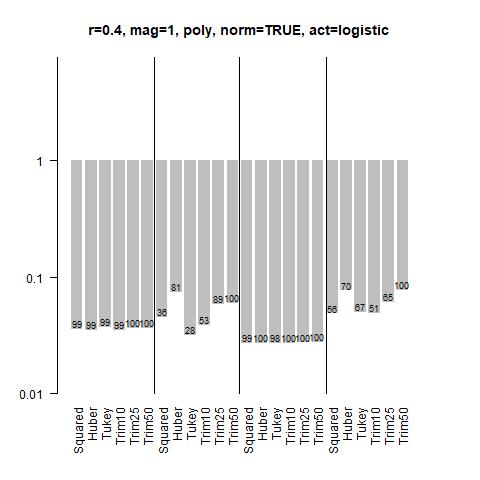}\\
\includegraphics[width=6.75cm,height=6.25cm]{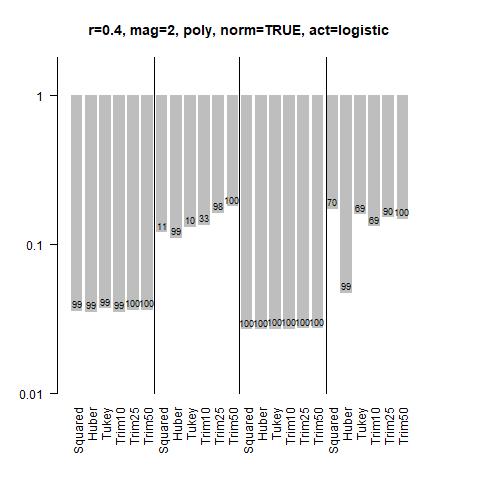} 
\includegraphics[width=6.75cm,height=6.25cm]{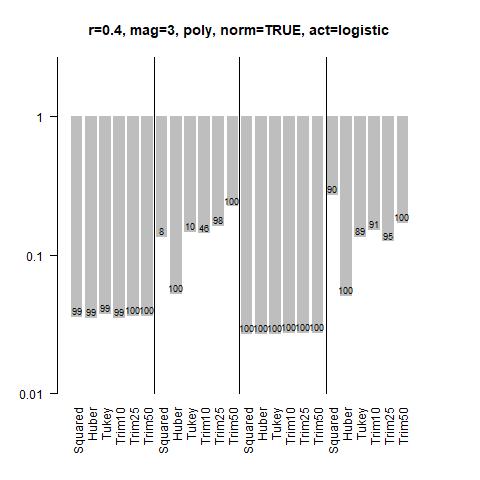} 
\end{center}
\caption{Results for $r=0.4$}
\end{figure}

\subsubsection{Trigonometric function}

\begin{figure}[H]
\label{trimnn:n1000p50r10m1trignonlogdeep}
\begin{center}
\includegraphics[width=6.75cm,height=6.25cm]{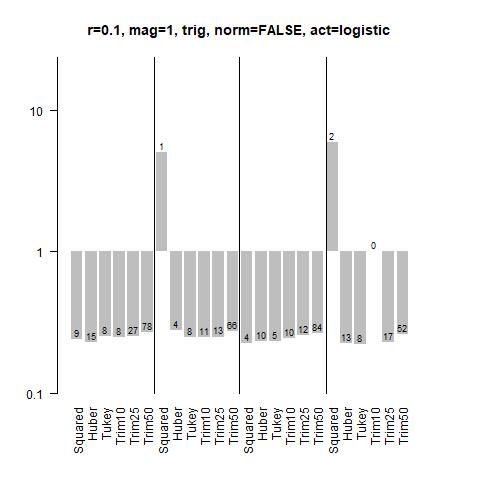}
\includegraphics[width=6.75cm,height=6.25cm]{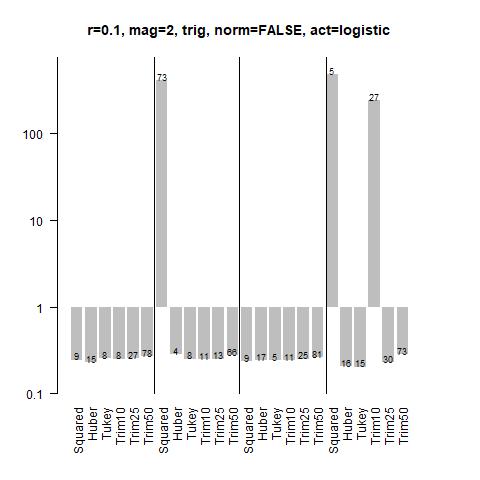} \\
\includegraphics[width=6.75cm,height=6.25cm]{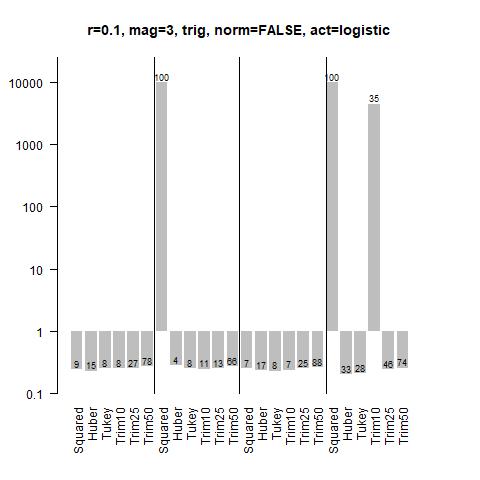} 
\includegraphics[width=6.75cm,height=6.25cm]{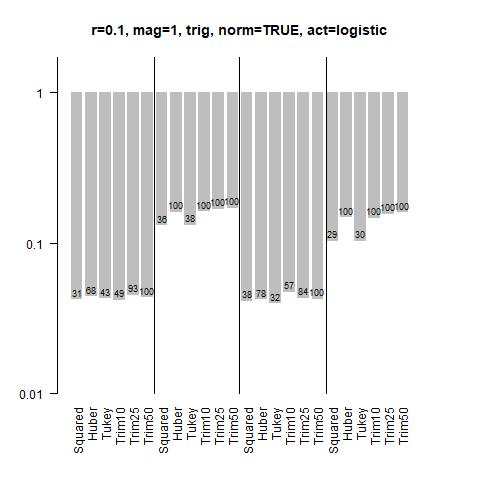}\\
\includegraphics[width=6.75cm,height=6.25cm]{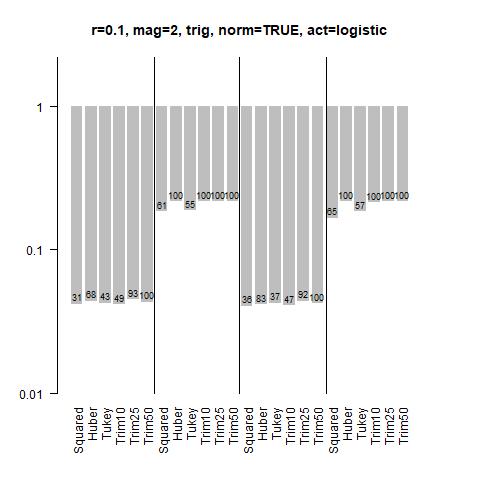} 
\includegraphics[width=6.75cm,height=6.25cm]{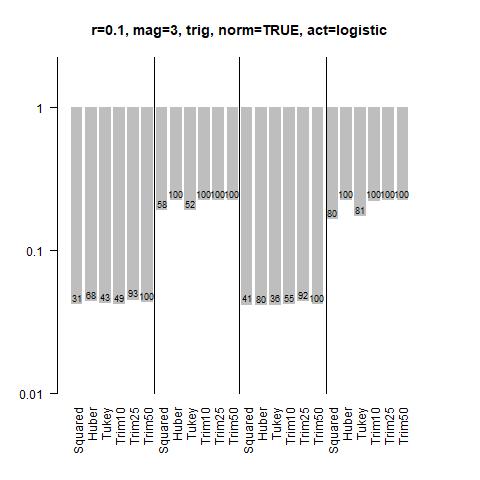} 
\end{center}
\caption{Results for $r=0.1$}
\end{figure}

\begin{figure}[H]
\label{trimnn:n1000p50r25m1trignonlogdeep}
\begin{center}
\includegraphics[width=6.75cm,height=6.25cm]{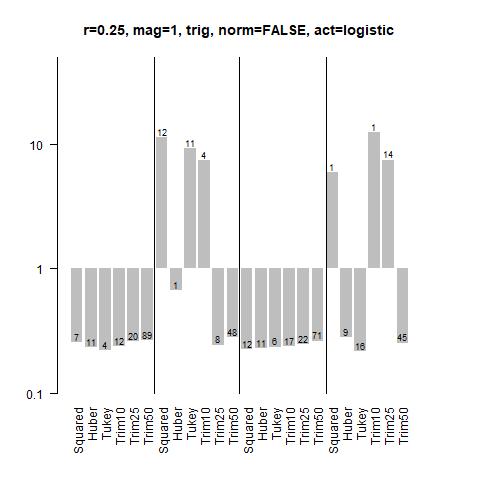}
\includegraphics[width=6.75cm,height=6.25cm]{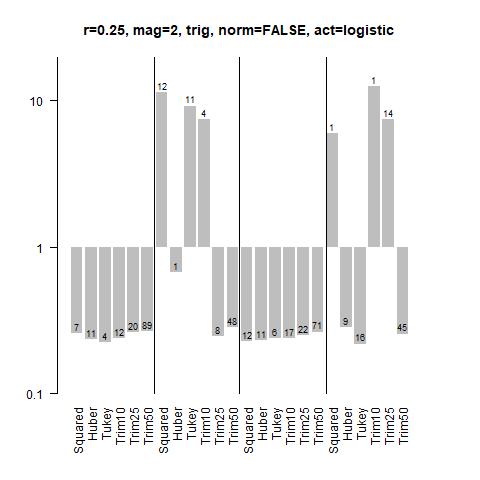} \\
\includegraphics[width=6.75cm,height=6.25cm]{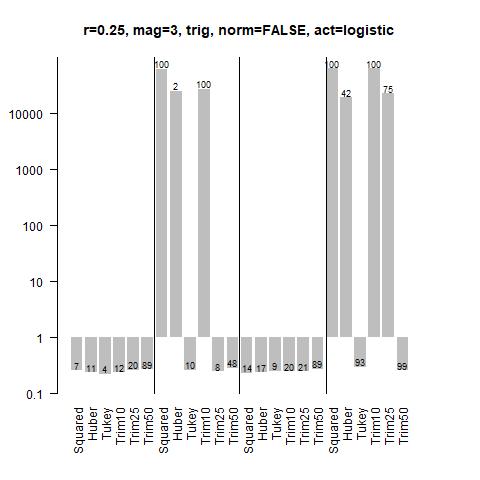} 
\includegraphics[width=6.75cm,height=6.25cm]{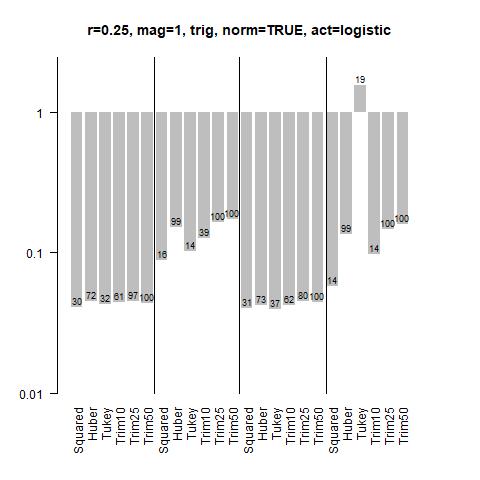}\\
\includegraphics[width=6.75cm,height=6.25cm]{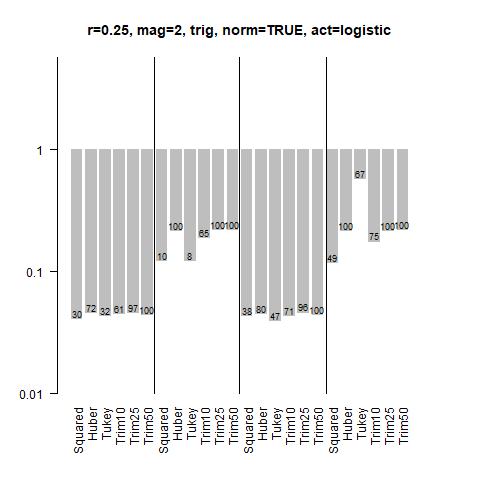} 
\includegraphics[width=6.75cm,height=6.25cm]{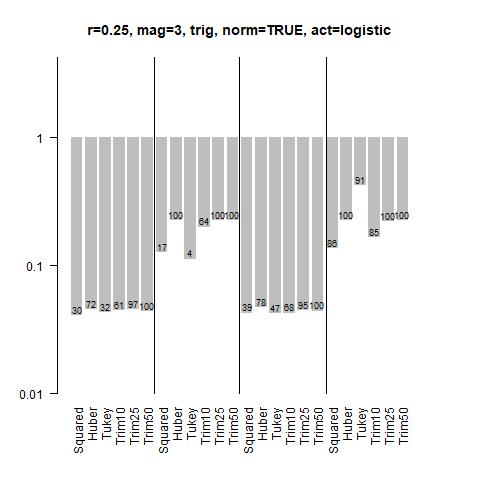} 
\end{center}
\caption{Results for $r=0.25$}
\end{figure}

\begin{figure}[H]
\begin{center}
\includegraphics[width=6.75cm,height=6.25cm]{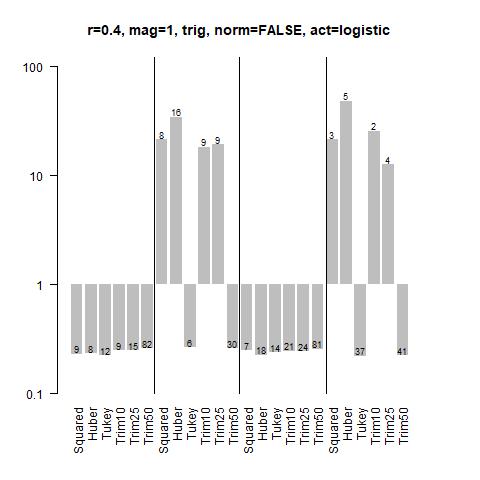}
\includegraphics[width=6.75cm,height=6.25cm]{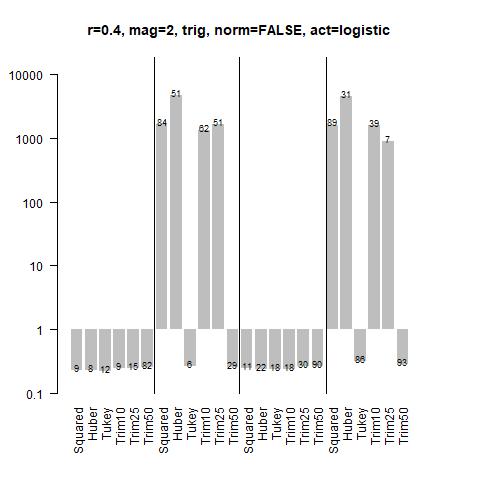} \\
\includegraphics[width=6.75cm,height=6.25cm]{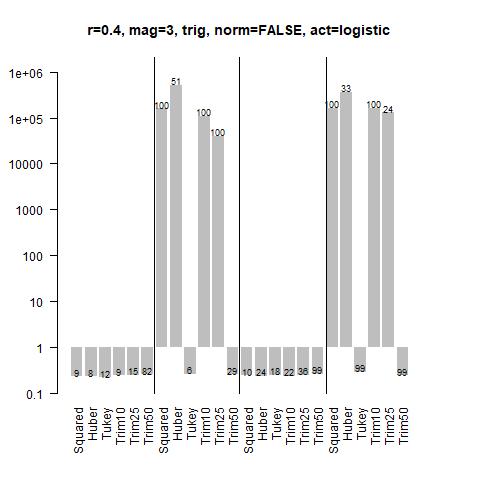} 
\includegraphics[width=6.75cm,height=6.25cm]{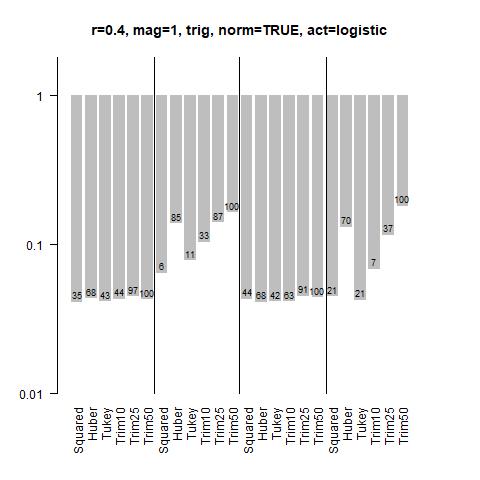}\\
\includegraphics[width=6.75cm,height=6.25cm]{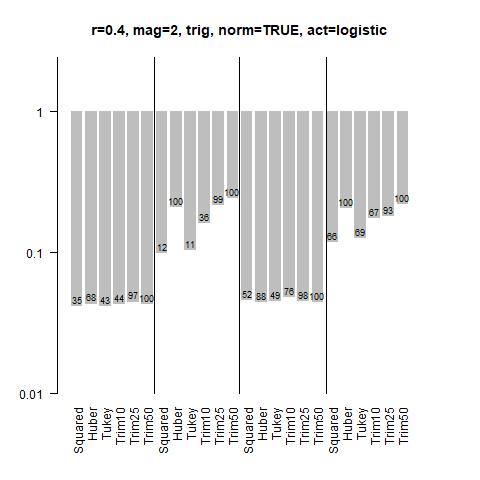} 
\includegraphics[width=6.75cm,height=6.25cm]{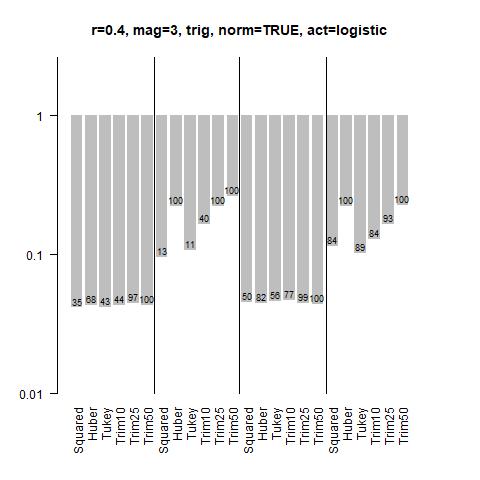} 
\end{center}
\caption{Results for $r=0.4$}\label{trimnn:n1000p50r40m1trignonlogdeep}
\end{figure}

\subsection{Softplus activation function}

\subsubsection{Linear function}

\begin{figure}[H]
\label{trimnn:n1000p50r10m1linnonreludeep}
\begin{center}
\includegraphics[width=6.75cm,height=6.25cm]{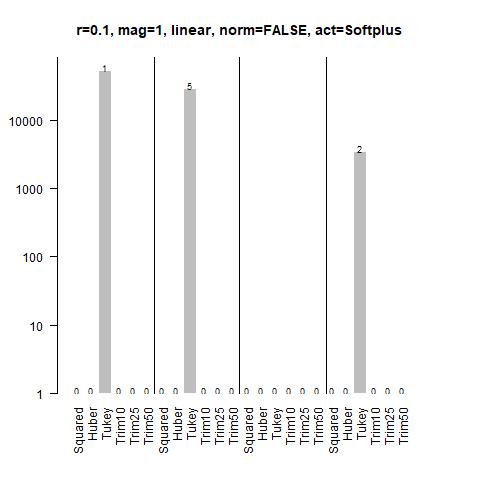}
\includegraphics[width=6.75cm,height=6.25cm]{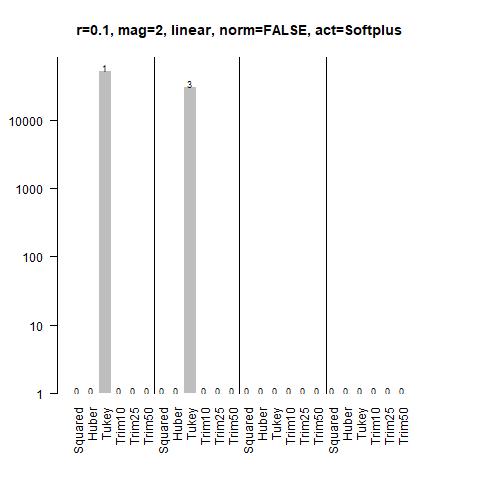} \\
\includegraphics[width=6.75cm,height=6.25cm]{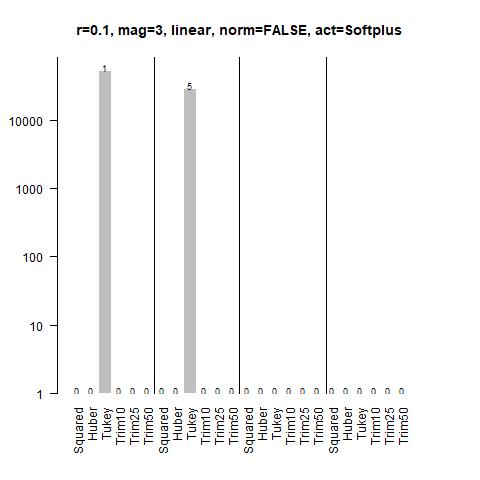} 
\includegraphics[width=6.75cm,height=6.25cm]{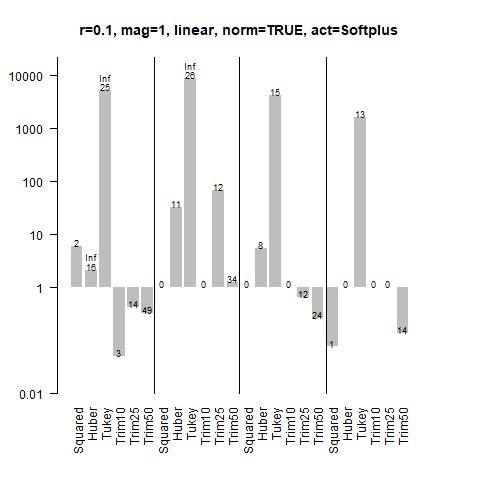}\\
\includegraphics[width=6.75cm,height=6.25cm]{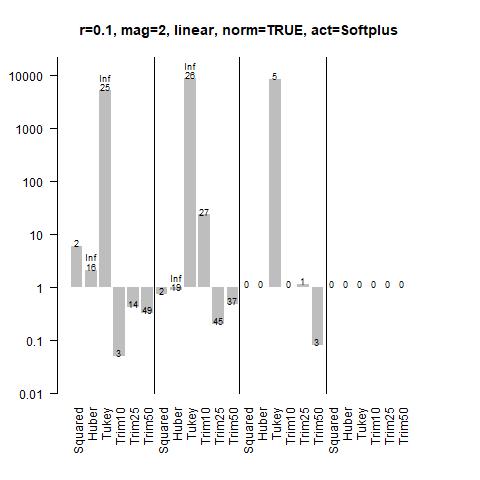} 
\includegraphics[width=6.75cm,height=6.25cm]{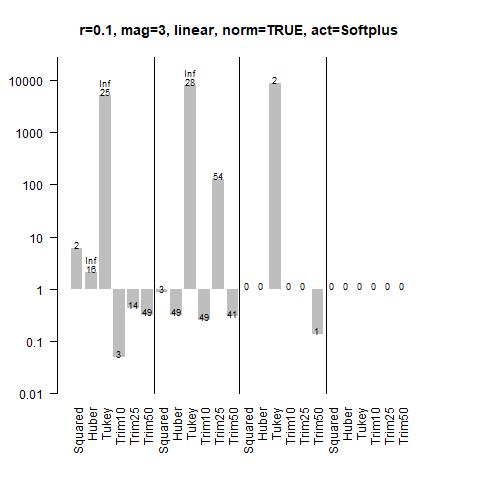} 
\end{center}
\caption{Results for $r=0.1$}
\end{figure}

\begin{figure}[H]
\label{trimnn:n1000p50r25m1linnonreludeep}
\begin{center}
\includegraphics[width=6.75cm,height=6.25cm]{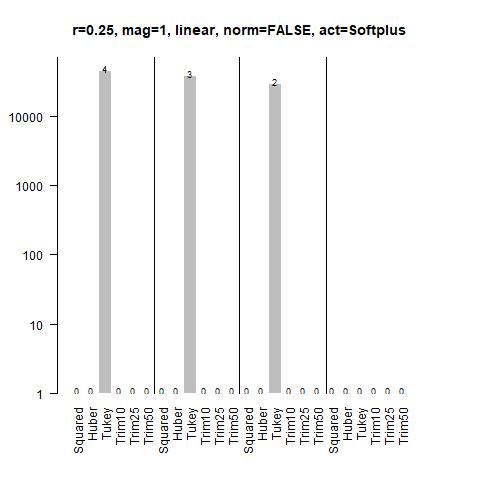}
\includegraphics[width=6.75cm,height=6.25cm]{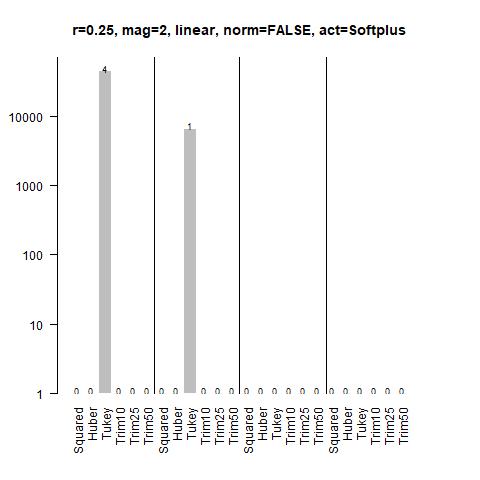} \\
\includegraphics[width=6.75cm,height=6.25cm]{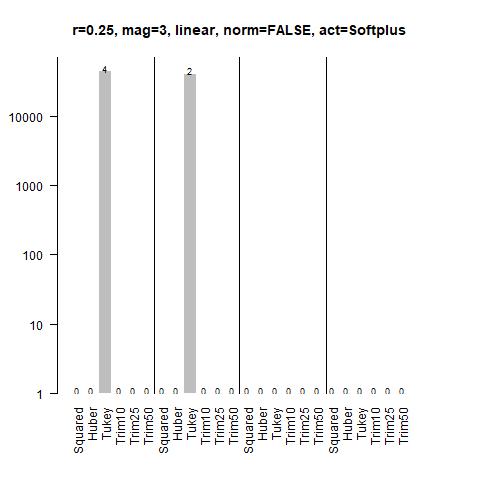} 
\includegraphics[width=6.75cm,height=6.25cm]{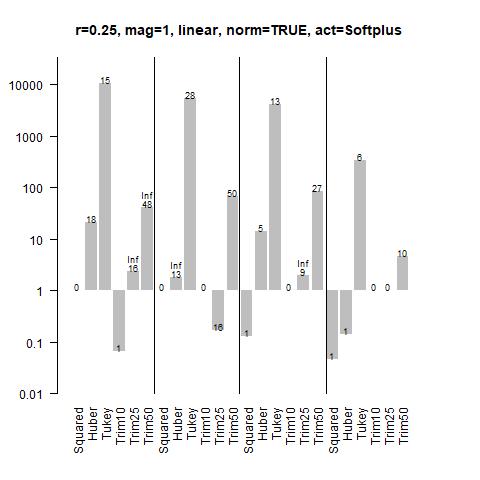}\\
\includegraphics[width=6.75cm,height=6.25cm]{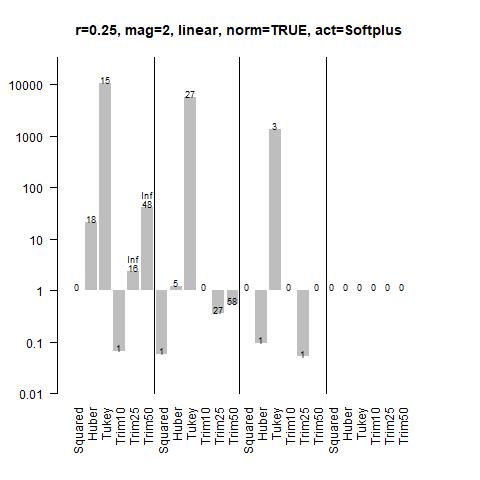} 
\includegraphics[width=6.75cm,height=6.25cm]{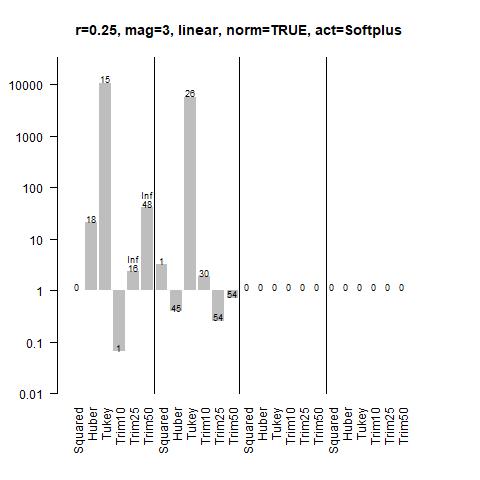} 
\end{center}
\caption{Results for $r=0.25$}
\end{figure}

\begin{figure}[H]
\label{trimnn:n1000p50r40m1linnonreludeep}
\begin{center}
\includegraphics[width=6.75cm,height=6.25cm]{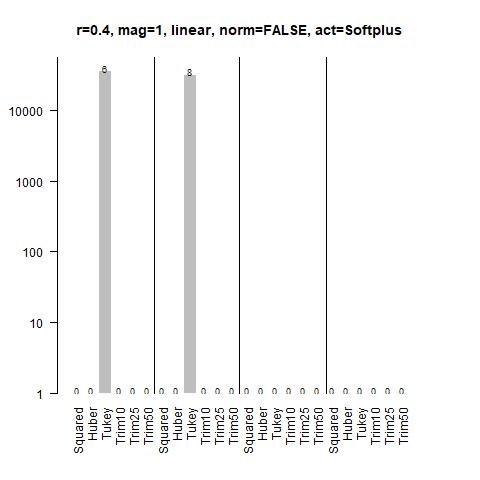}
\includegraphics[width=6.75cm,height=6.25cm]{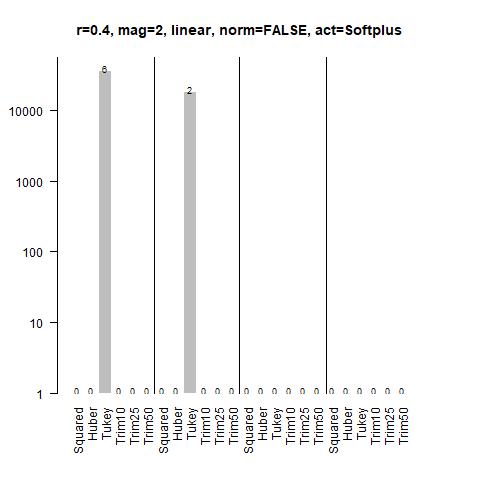} \\
\includegraphics[width=6.75cm,height=6.25cm]{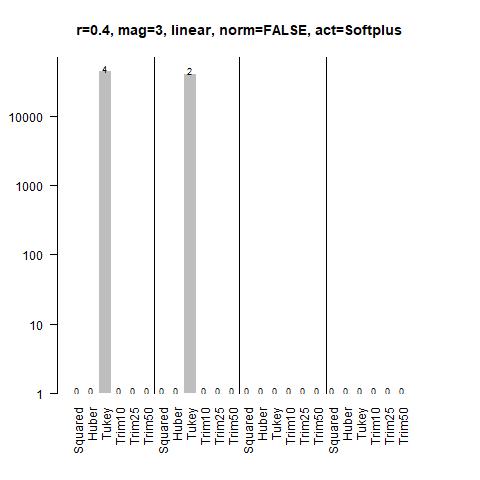} 
\includegraphics[width=6.75cm,height=6.25cm]{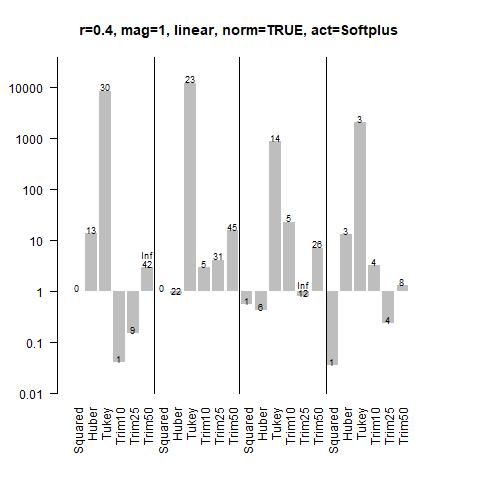}\\
\includegraphics[width=6.75cm,height=6.25cm]{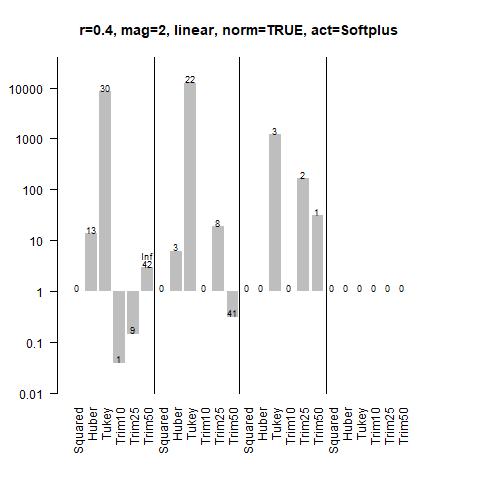} 
\includegraphics[width=6.75cm,height=6.25cm]{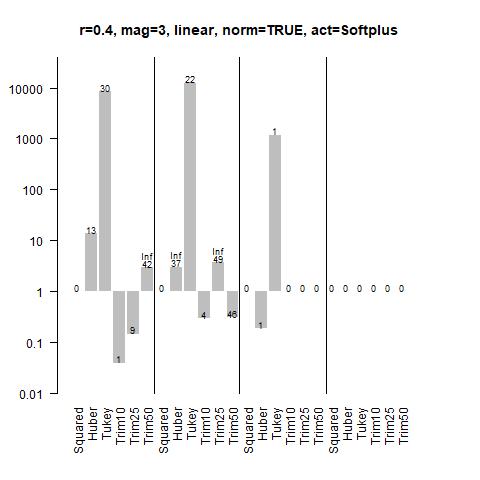} 
\end{center}
\caption{Results for $r=0.4$}
\end{figure}

\subsubsection{Polynomial function}

\begin{figure}[H]
\label{trimnn:n1000p50r10m1polynonreludeep}
\begin{center}
\includegraphics[width=6.75cm,height=6.25cm]{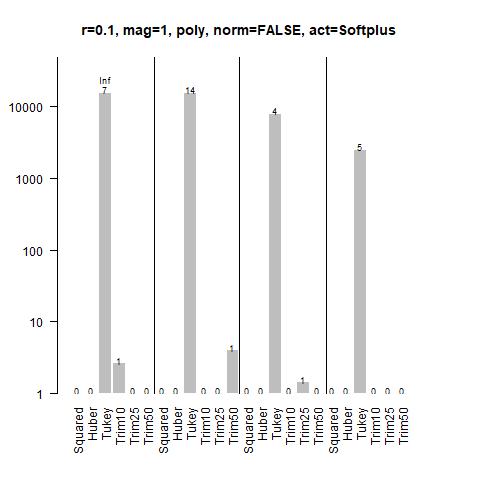}
\includegraphics[width=6.75cm,height=6.25cm]{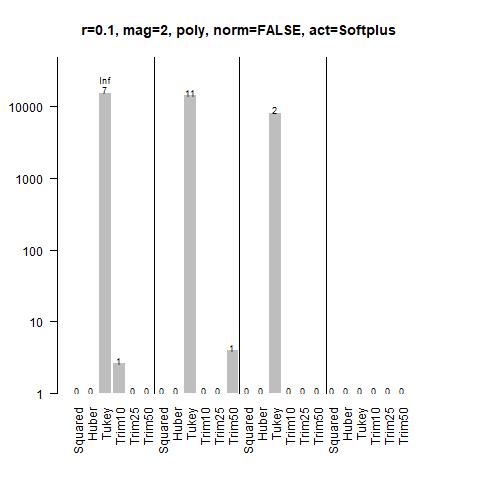} \\
\includegraphics[width=6.75cm,height=6.25cm]{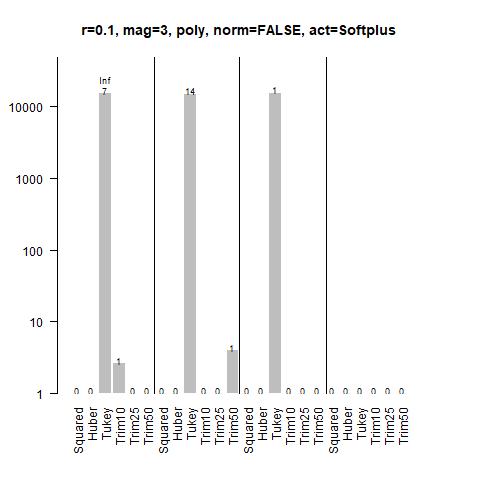} 
\includegraphics[width=6.75cm,height=6.25cm]{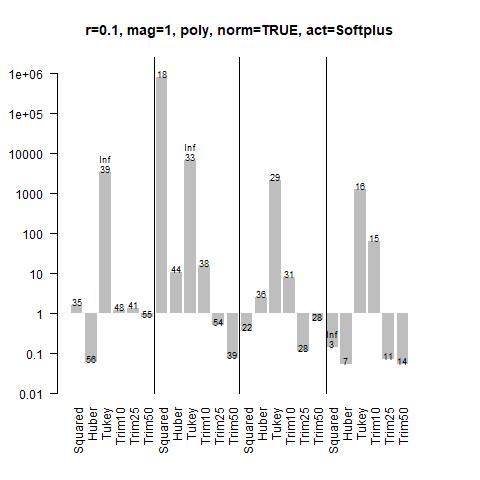}\\
\includegraphics[width=6.75cm,height=6.25cm]{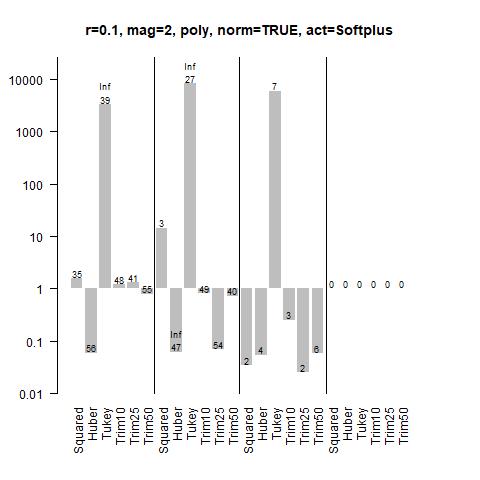} 
\includegraphics[width=6.75cm,height=6.25cm]{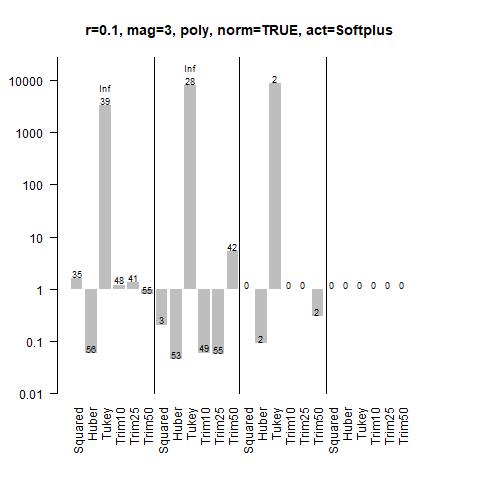} 
\end{center}
\caption{Results for $r=0.1$}
\end{figure}

\begin{figure}[H]
\label{trimnn:n1000p50r25m1polynonreludeep}
\begin{center}
\includegraphics[width=6.75cm,height=6.25cm]{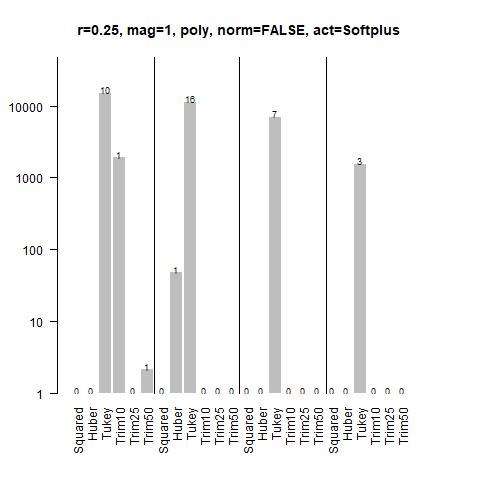}
\includegraphics[width=6.75cm,height=6.25cm]{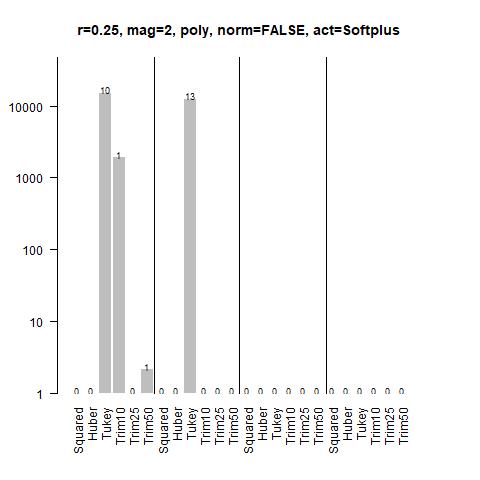} \\
\includegraphics[width=6.75cm,height=6.25cm]{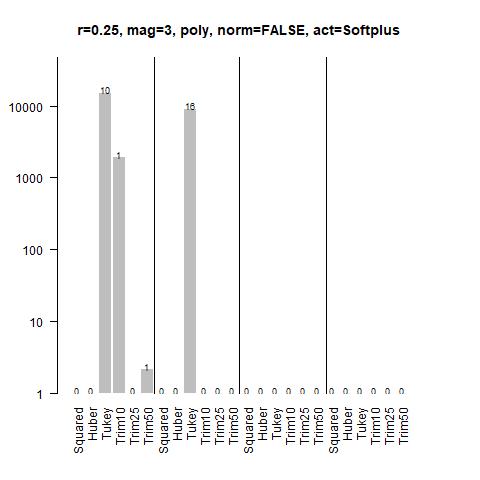} 
\includegraphics[width=6.75cm,height=6.25cm]{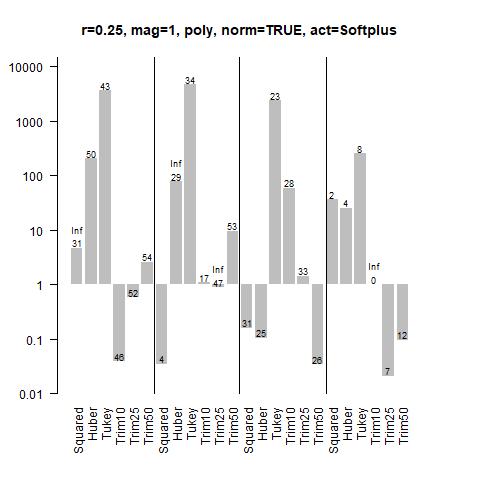}\\
\includegraphics[width=6.75cm,height=6.25cm]{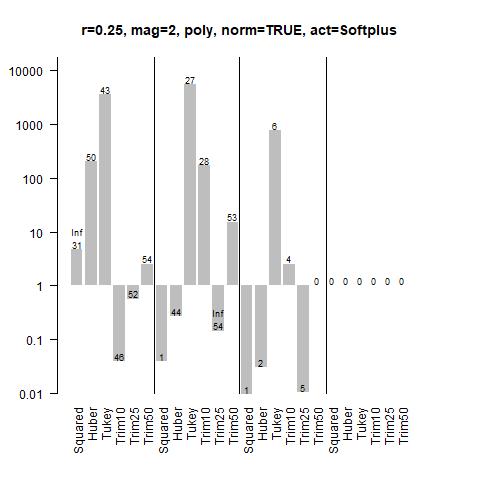} 
\includegraphics[width=6.75cm,height=6.25cm]{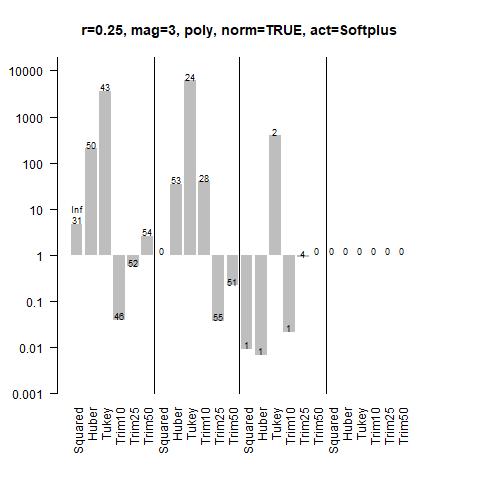} 
\end{center}
\caption{Results for $r=0.25$}
\end{figure}

\begin{figure}[H]
\label{trimnn:n1000p50r40m1polynonreludeep}
\begin{center}
\includegraphics[width=6.75cm,height=6.25cm]{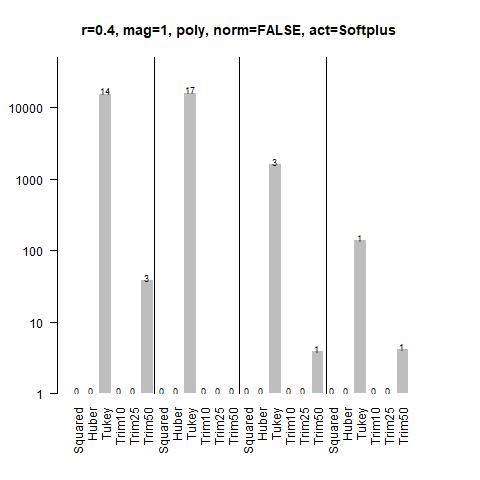}
\includegraphics[width=6.75cm,height=6.25cm]{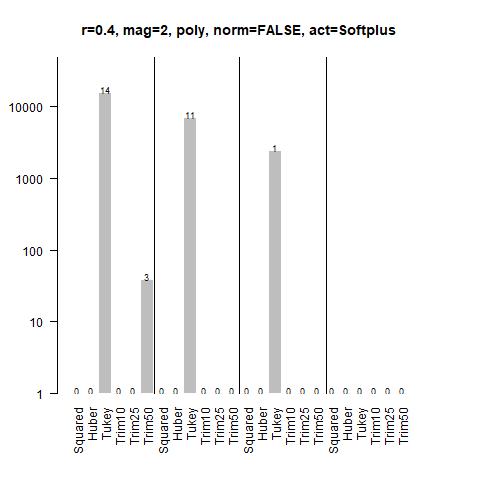} \\
\includegraphics[width=6.75cm,height=6.25cm]{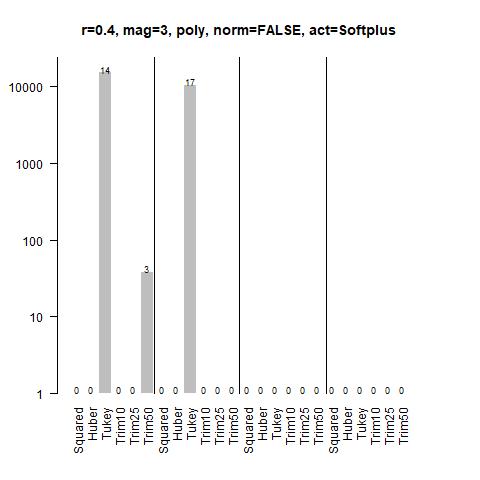} 
\includegraphics[width=6.75cm,height=6.25cm]{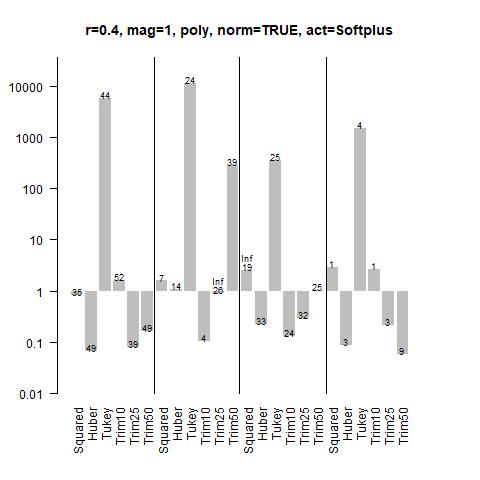}\\
\includegraphics[width=6.75cm,height=6.25cm]{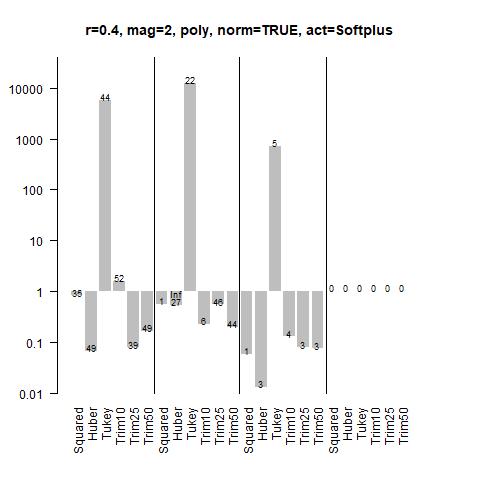} 
\includegraphics[width=6.75cm,height=6.25cm]{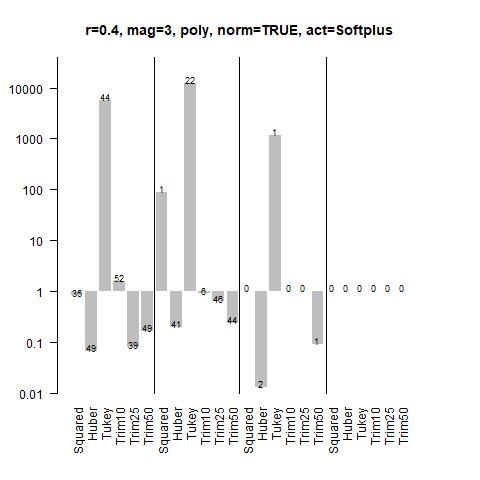} 
\end{center}
\caption{Results for $r=0.4$}
\end{figure}

\subsubsection{Trigonometric function}

\begin{figure}[H]
\label{trimnn:n1000p50r10m1trignonreludeep}
\begin{center}
\includegraphics[width=6.75cm,height=6.25cm]{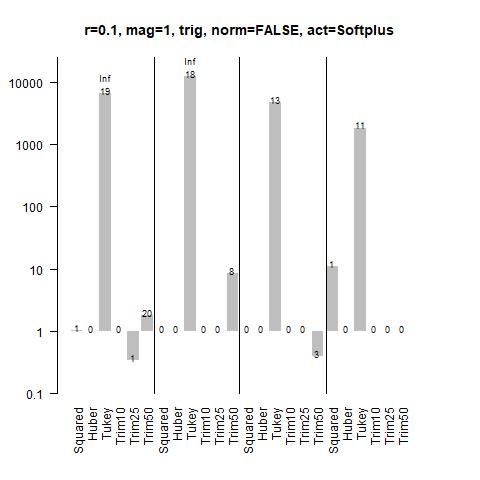}
\includegraphics[width=6.75cm,height=6.25cm]{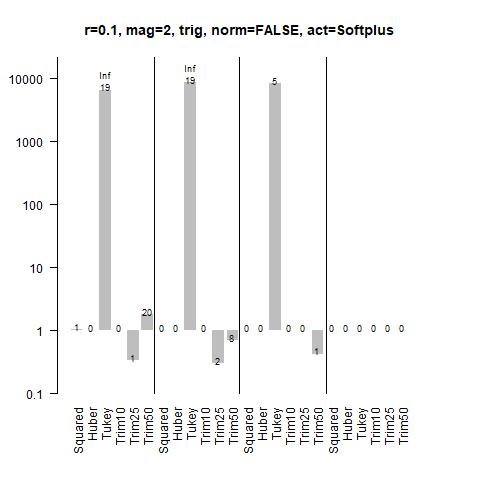} \\
\includegraphics[width=6.75cm,height=6.25cm]{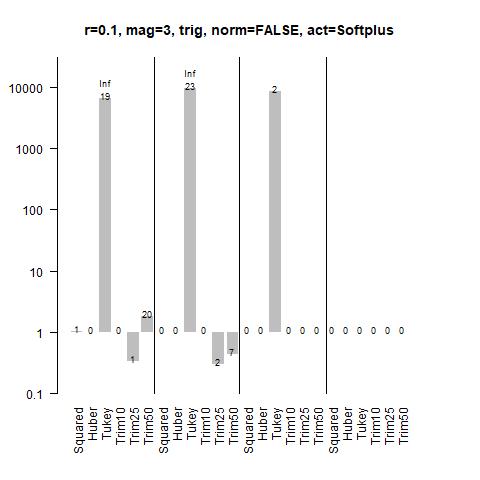} 
\includegraphics[width=6.75cm,height=6.25cm]{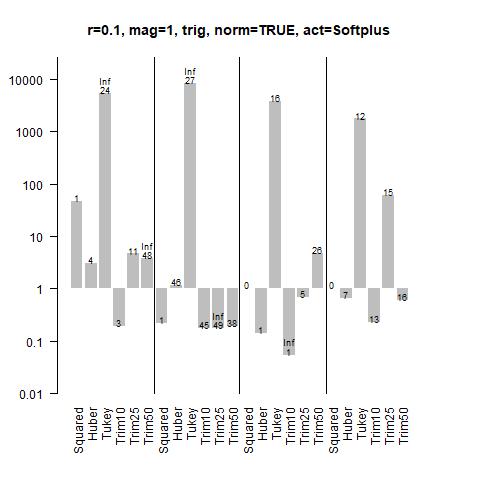}\\
\includegraphics[width=6.75cm,height=6.25cm]{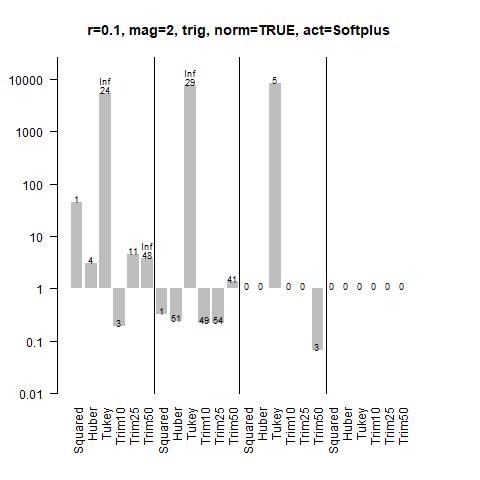} 
\includegraphics[width=6.75cm,height=6.25cm]{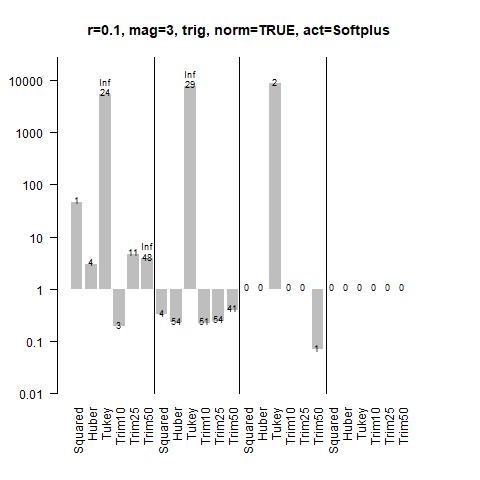} 
\end{center}
\caption{Results for $r=0.1$}
\end{figure}

\begin{figure}[H]
\label{trimnn:n1000p50r25m1trignonreludeep}
\begin{center}
\includegraphics[width=6.75cm,height=6.25cm]{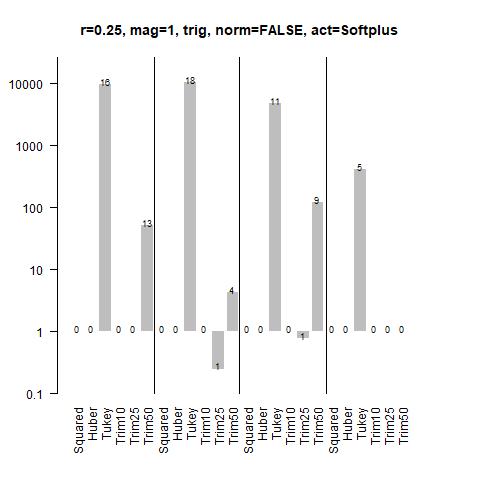}
\includegraphics[width=6.75cm,height=6.25cm]{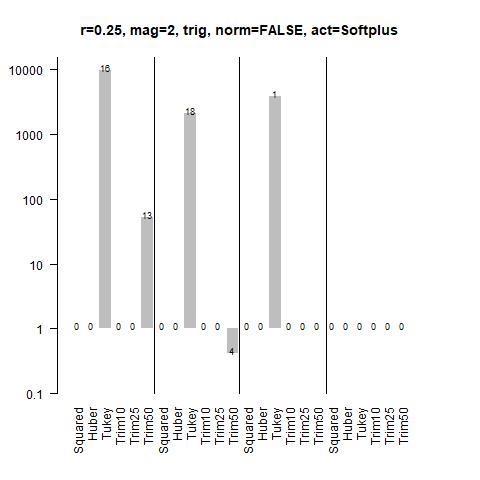} \\
\includegraphics[width=6.75cm,height=6.25cm]{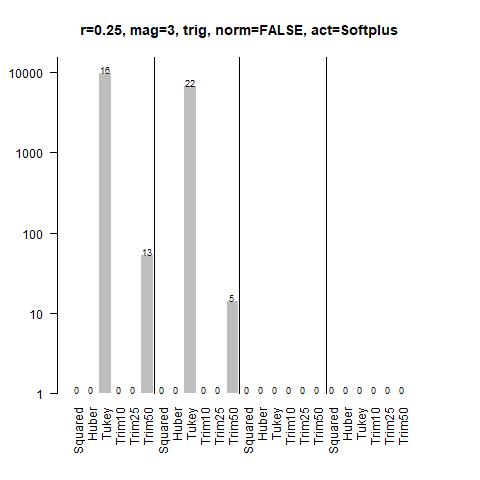} 
\includegraphics[width=6.75cm,height=6.25cm]{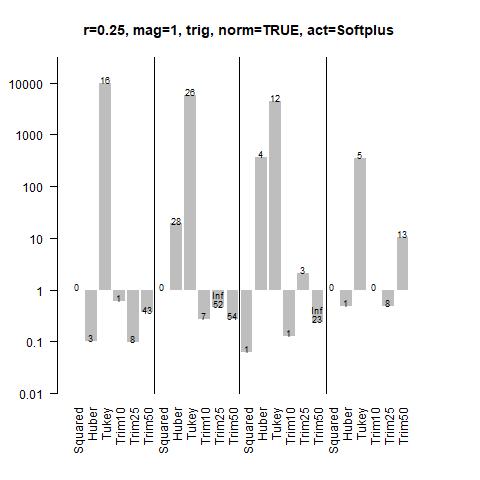}\\
\includegraphics[width=6.75cm,height=6.25cm]{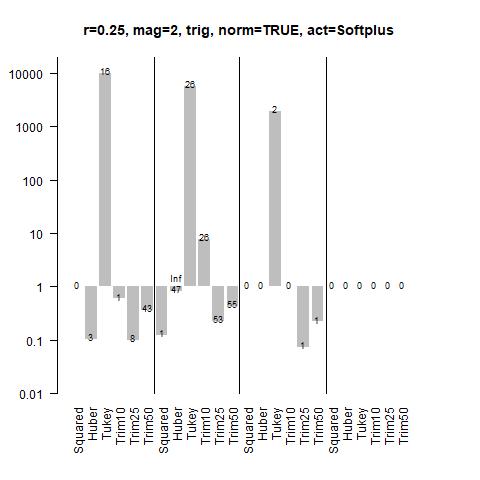} 
\includegraphics[width=6.75cm,height=6.25cm]{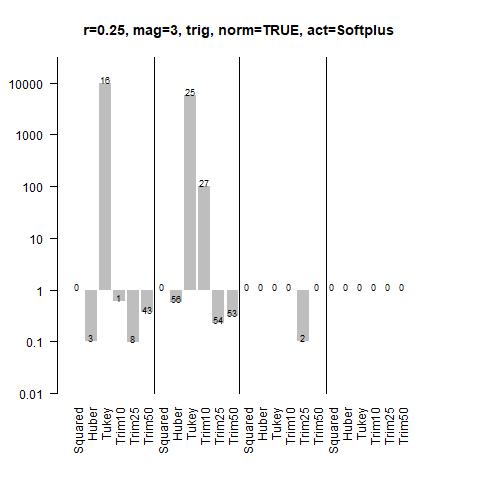} 
\end{center}
\caption{Results for $r=0.25$}
\end{figure}

\begin{figure}[H]
\label{trimnn:n1000p50r40m1trignonreludeep}
\begin{center}
\includegraphics[width=6.75cm,height=6.25cm]{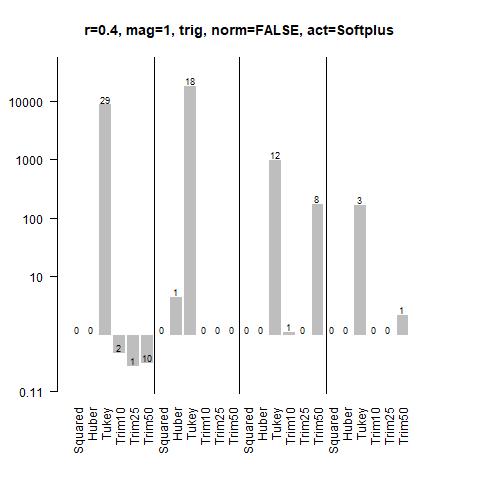}
\includegraphics[width=6.75cm,height=6.25cm]{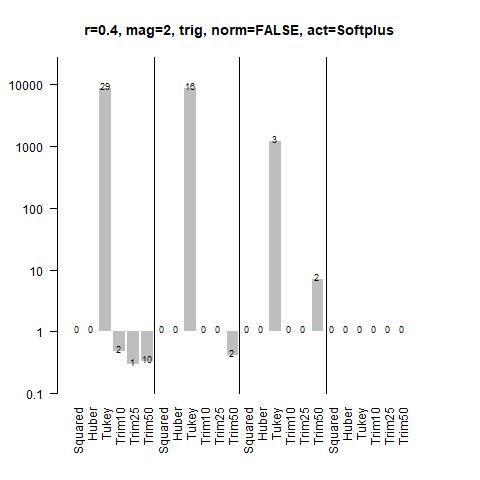} \\
\includegraphics[width=6.75cm,height=6.25cm]{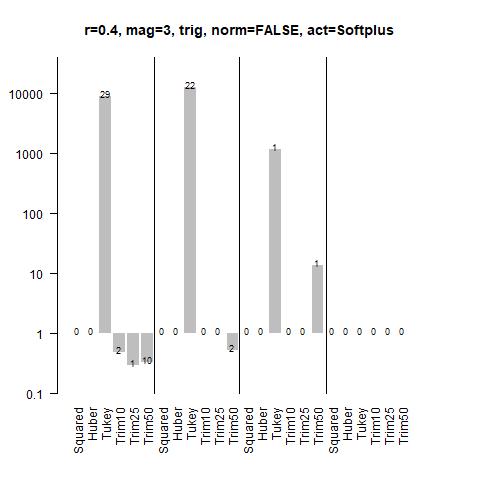} 
\includegraphics[width=6.75cm,height=6.25cm]{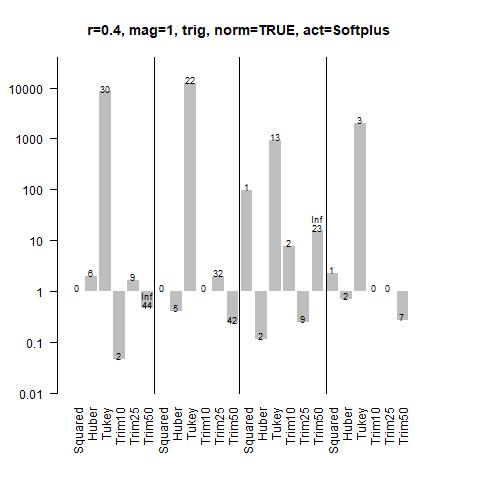}\\
\includegraphics[width=6.75cm,height=6.25cm]{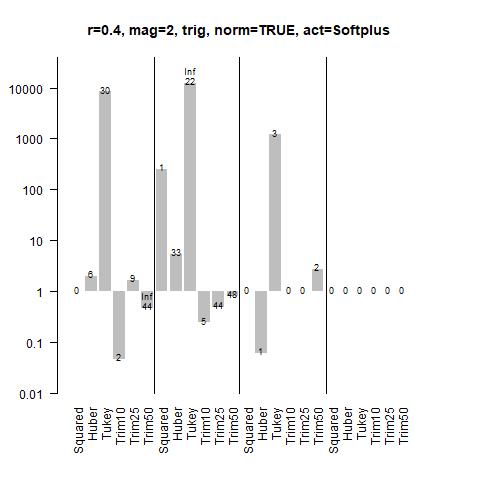} 
\includegraphics[width=6.75cm,height=6.25cm]{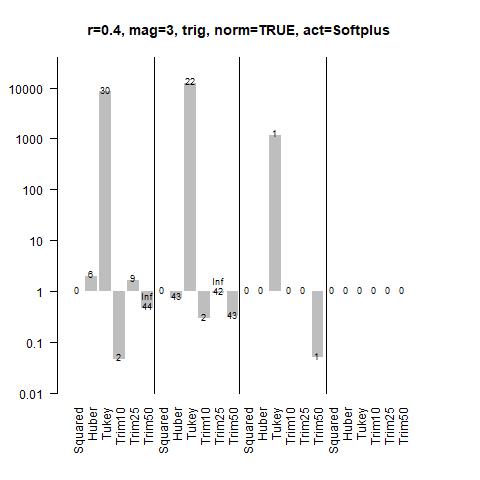} 
\end{center}
\caption{Results for $r=0.4$}
\end{figure}

\section{Simulation results for $n=150$ and $p=5$: Training steps}  \label{trimnn:secstep1505}

\subsection{Logistic activation function}

\subsubsection{Linear function}

\begin{figure}[H]
\label{trimnn:n200p5r10m1linnonlogStep}
\begin{center}
\includegraphics[width=6.75cm,height=6.25cm]{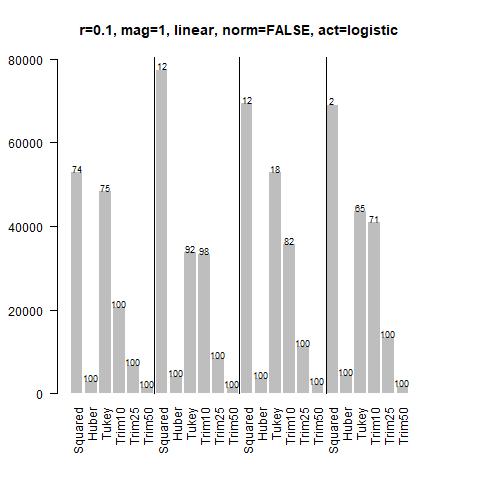}
\includegraphics[width=6.75cm,height=6.25cm]{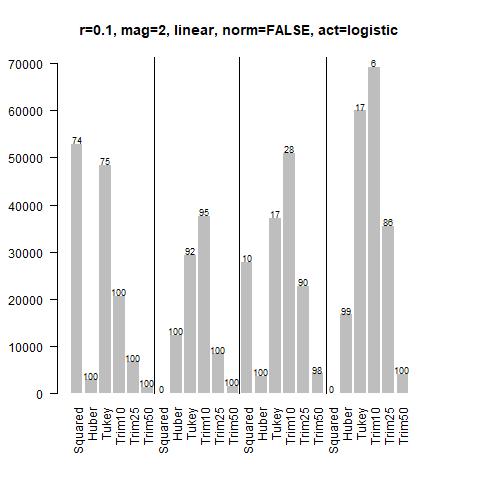} \\
\includegraphics[width=6.75cm,height=6.25cm]{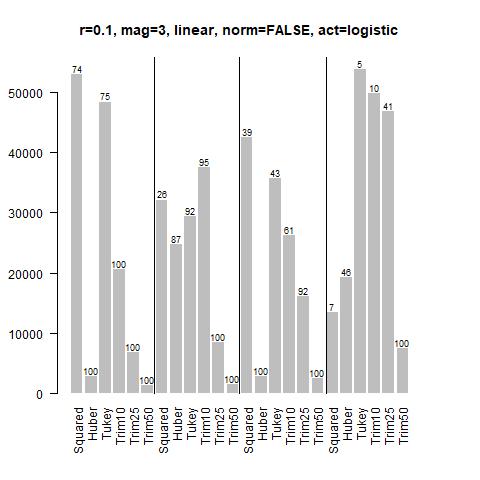} 
\includegraphics[width=6.75cm,height=6.25cm]{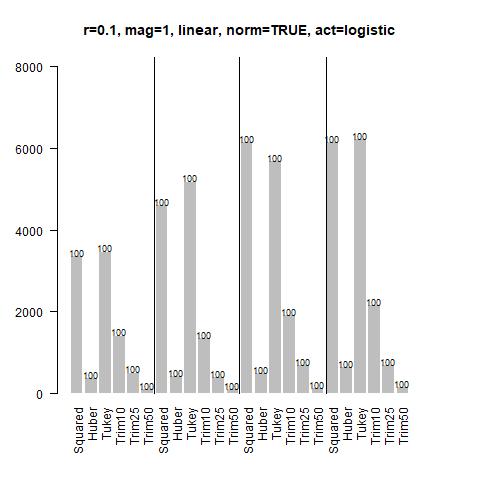}\\
\includegraphics[width=6.75cm,height=6.25cm]{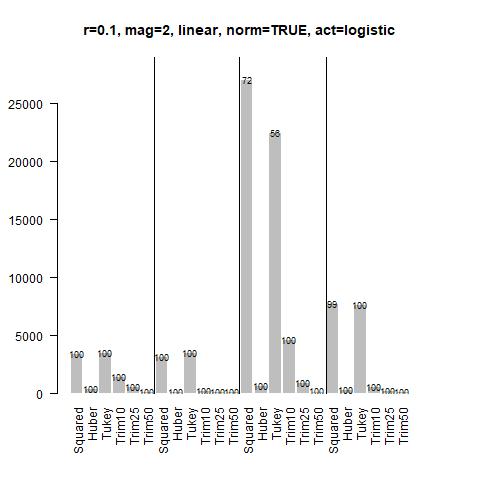} 
\includegraphics[width=6.75cm,height=6.25cm]{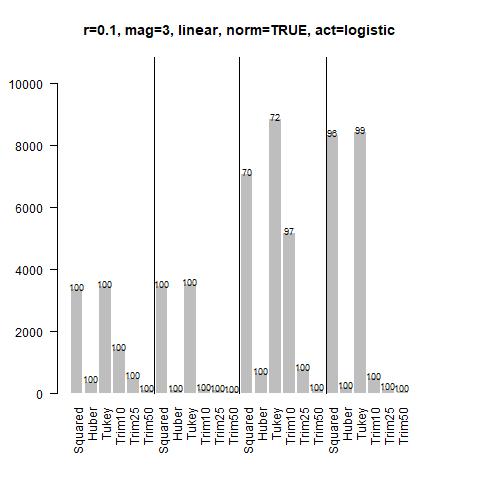} 
\end{center}
\caption{Results for $r=0.1$}
\end{figure}

\begin{figure}[H]
\label{trimnn:n200p5r25m1linnonlogStep}
\begin{center}
\includegraphics[width=6.75cm,height=6.25cm]{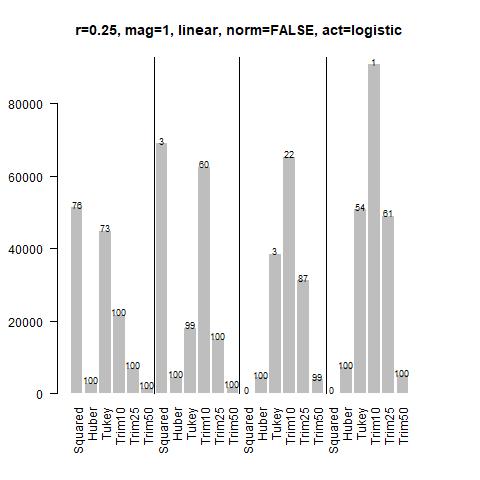}
\includegraphics[width=6.75cm,height=6.25cm]{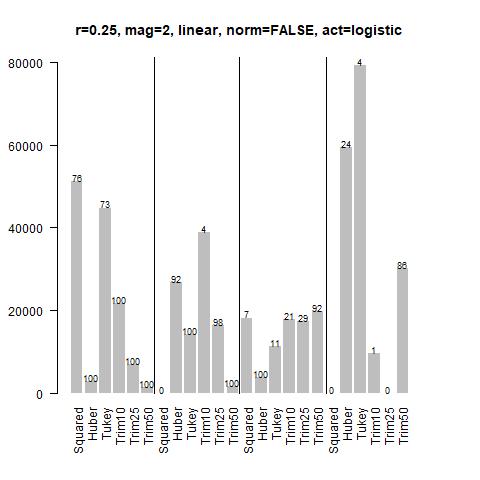} \\
\includegraphics[width=6.75cm,height=6.25cm]{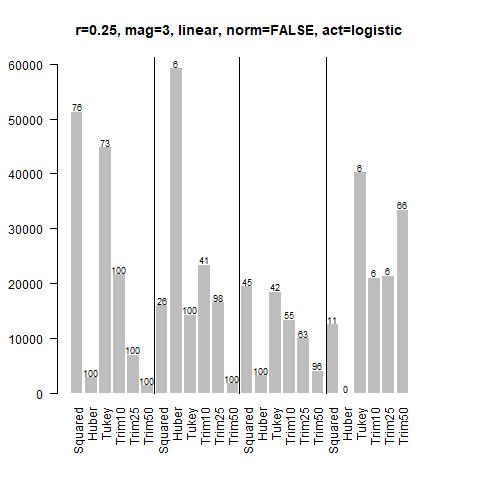} 
\includegraphics[width=6.75cm,height=6.25cm]{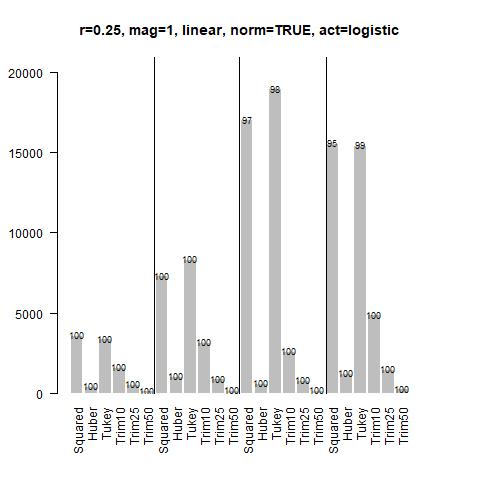}\\
\includegraphics[width=6.75cm,height=6.25cm]{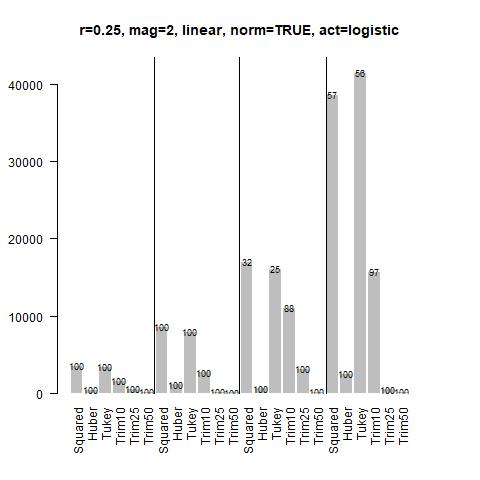} 
\includegraphics[width=6.75cm,height=6.25cm]{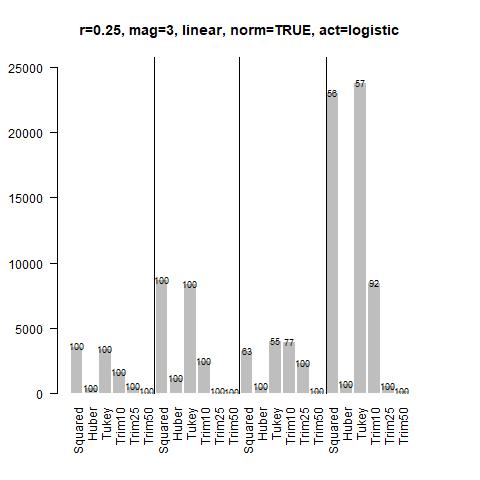} 
\end{center}
\caption{Results for $r=0.25$}
\end{figure}

\begin{figure}[H]
\label{trimnn:n200p5r40m1linnonlogStep}
\begin{center}
\includegraphics[width=6.75cm,height=6.25cm]{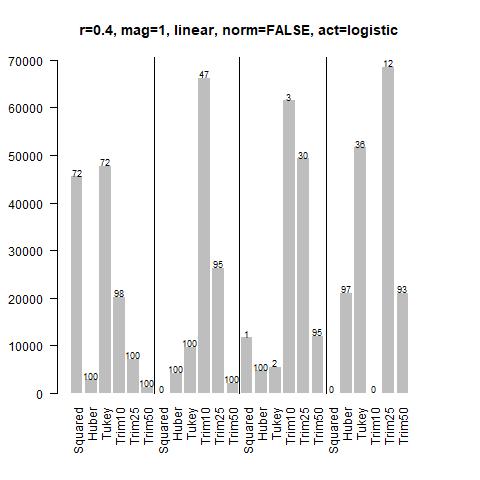}
\includegraphics[width=6.75cm,height=6.25cm]{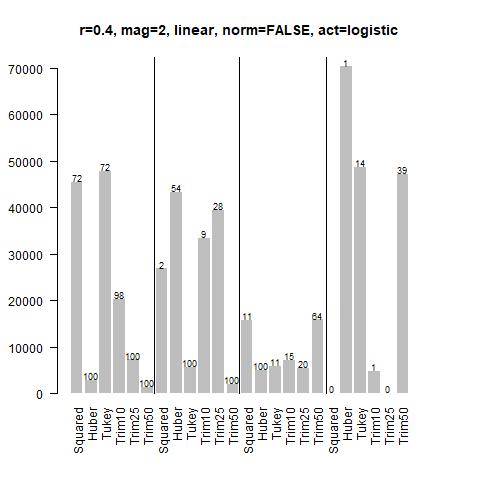} \\
\includegraphics[width=6.75cm,height=6.25cm]{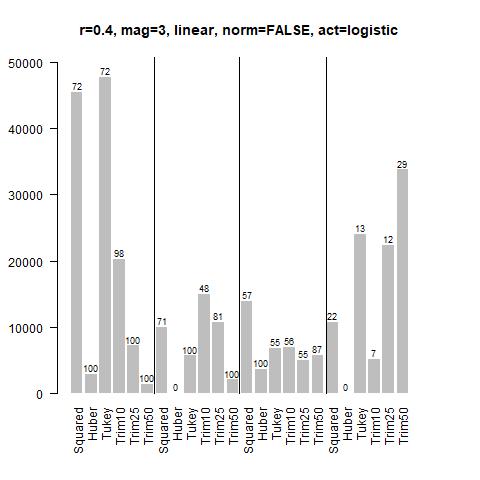} 
\includegraphics[width=6.75cm,height=6.25cm]{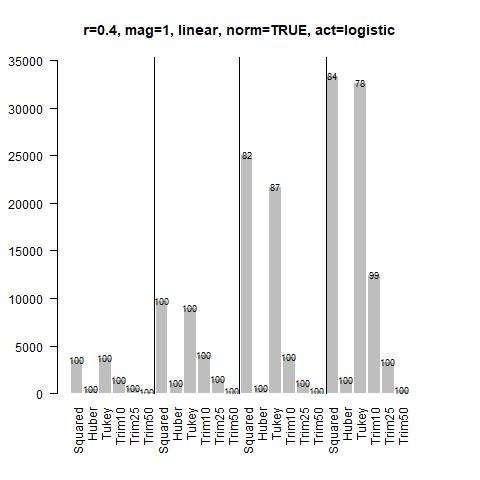}\\
\includegraphics[width=6.75cm,height=6.25cm]{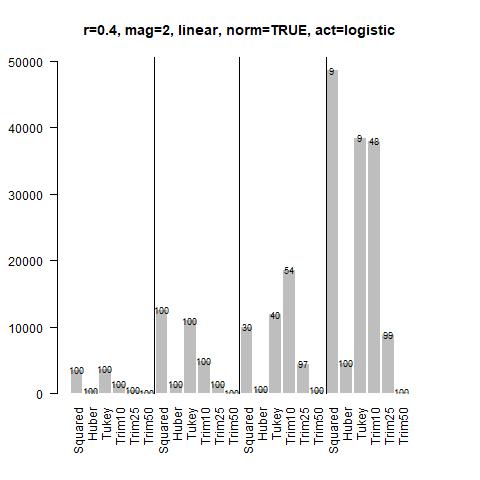} 
\includegraphics[width=6.75cm,height=6.25cm]{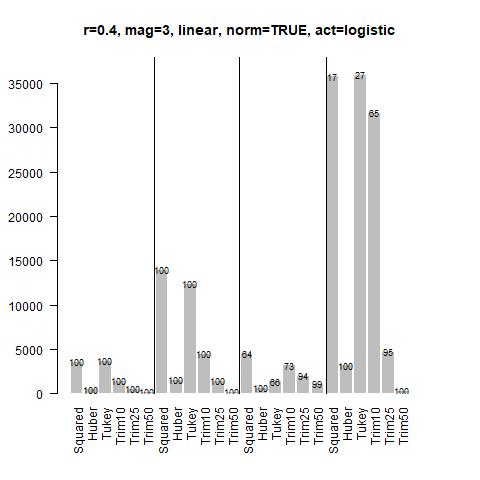} 
\end{center}
\caption{Results for $r=0.4$}
\end{figure}

\subsubsection{Polynomial function}

\begin{figure}[H]
\label{trimnn:n200p5r10m1polynonlogStep}
\begin{center}
\includegraphics[width=6.75cm,height=6.25cm]{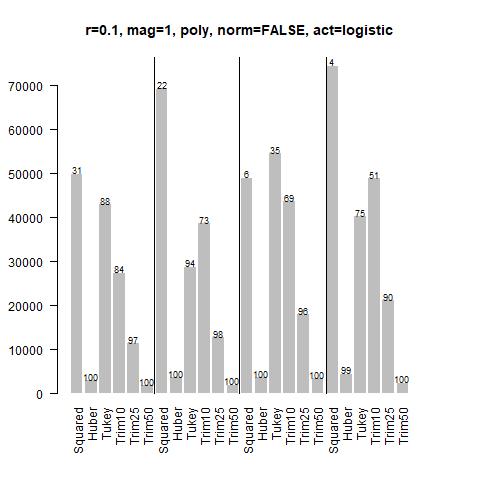}
\includegraphics[width=6.75cm,height=6.25cm]{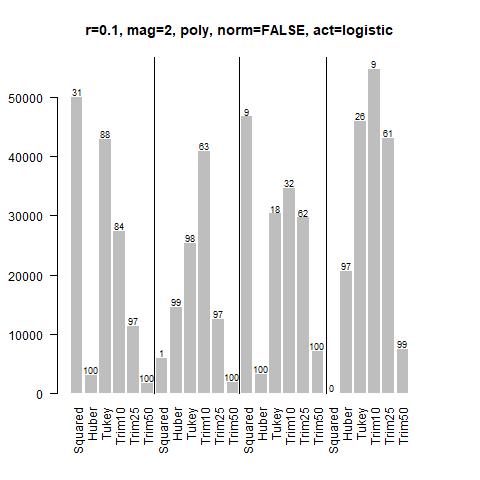} \\
\includegraphics[width=6.75cm,height=6.25cm]{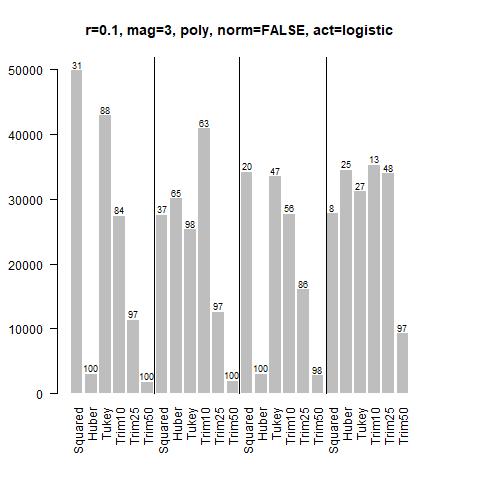} 
\includegraphics[width=6.75cm,height=6.25cm]{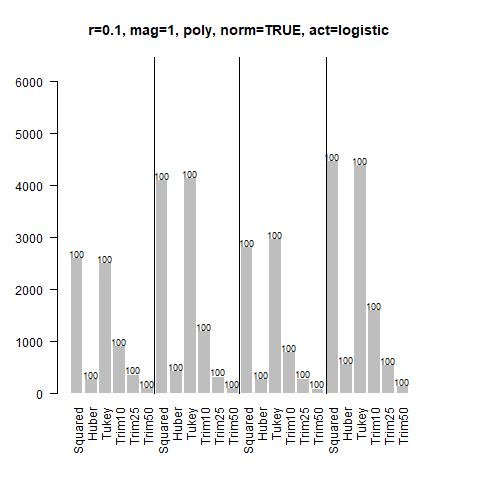}\\
\includegraphics[width=6.75cm,height=6.25cm]{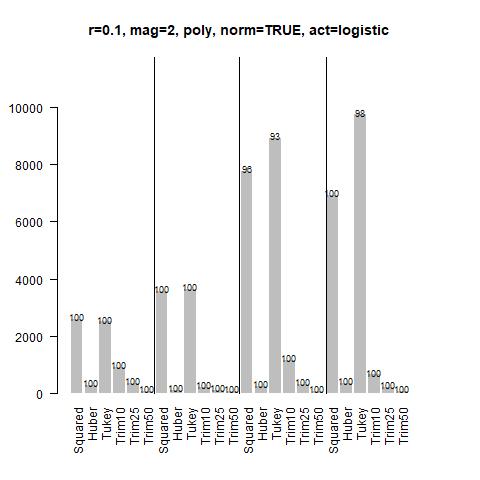} 
\includegraphics[width=6.75cm,height=6.25cm]{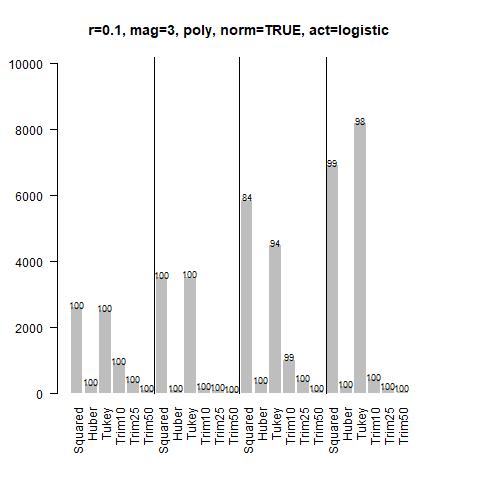} 
\end{center}
\caption{Results for $r=0.1$}
\end{figure}

\begin{figure}[H]
\label{trimnn:n200p5r25m1polynonlogStep}
\begin{center}
\includegraphics[width=6.75cm,height=6.25cm]{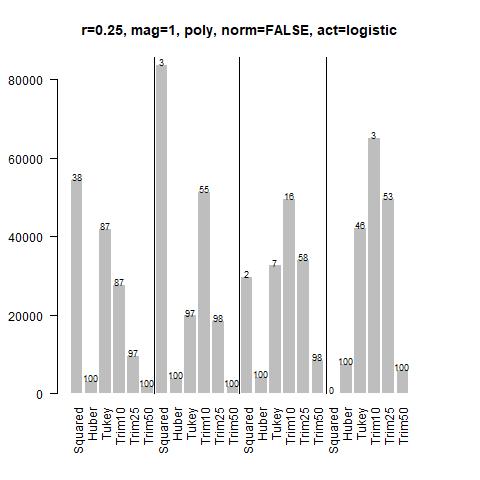}
\includegraphics[width=6.75cm,height=6.25cm]{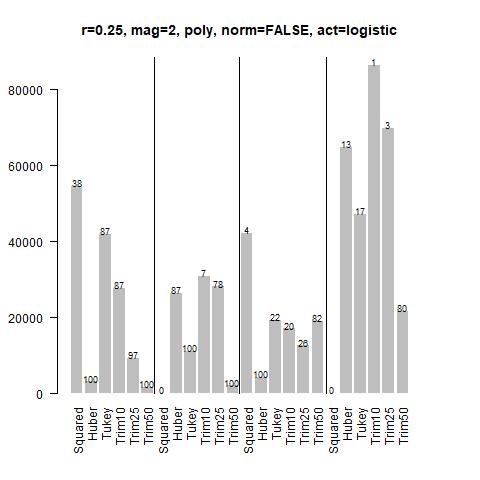} \\
\includegraphics[width=6.75cm,height=6.25cm]{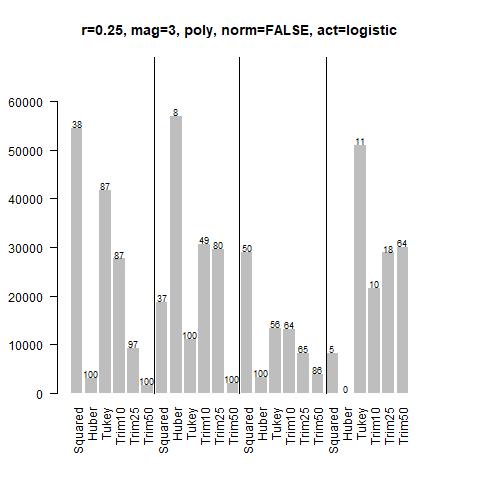} 
\includegraphics[width=6.75cm,height=6.25cm]{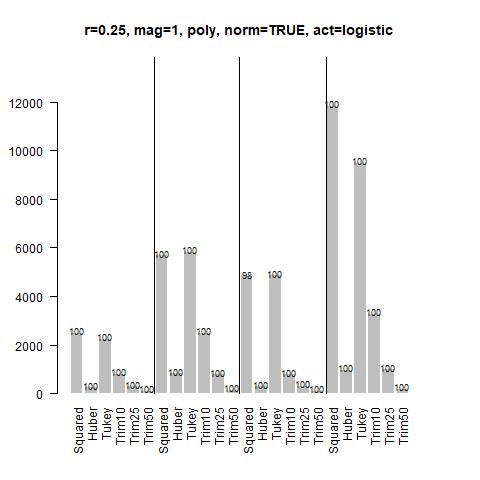}\\
\includegraphics[width=6.75cm,height=6.25cm]{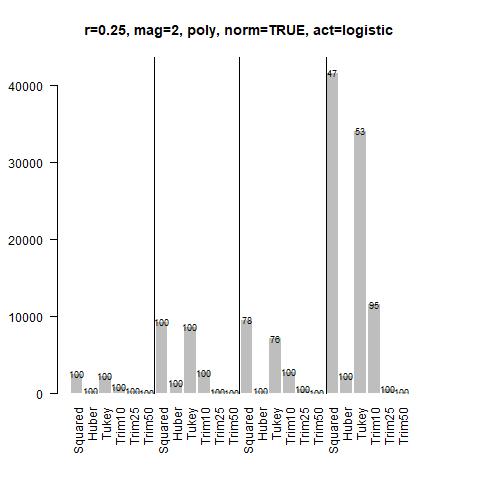} 
\includegraphics[width=6.75cm,height=6.25cm]{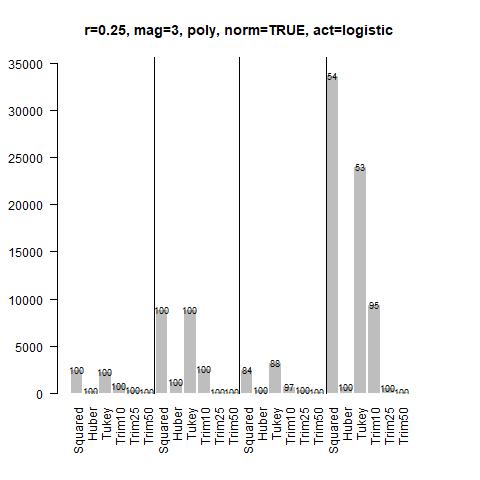} 
\end{center}
\caption{Results for $r=0.25$}
\end{figure}

\begin{figure}[H]
\label{trimnn:n200p5r40m1polynonlogStep}
\begin{center}
\includegraphics[width=6.75cm,height=6.25cm]{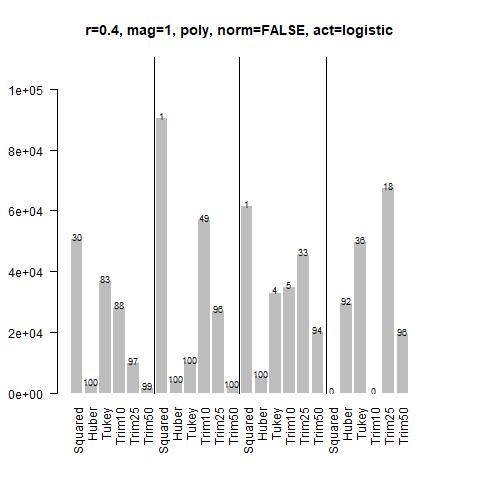}
\includegraphics[width=6.75cm,height=6.25cm]{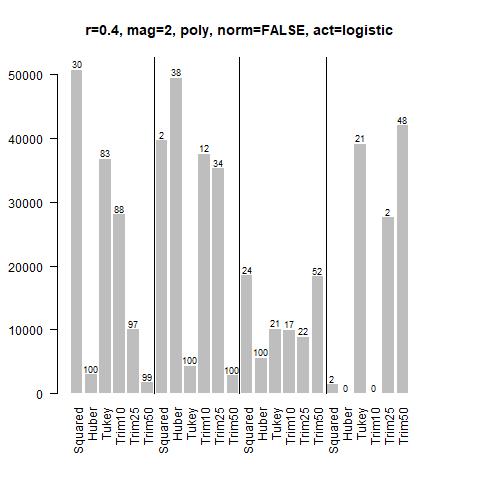} \\
\includegraphics[width=6.75cm,height=6.25cm]{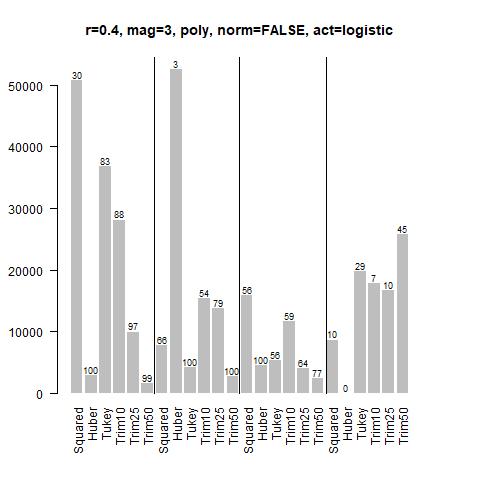} 
\includegraphics[width=6.75cm,height=6.25cm]{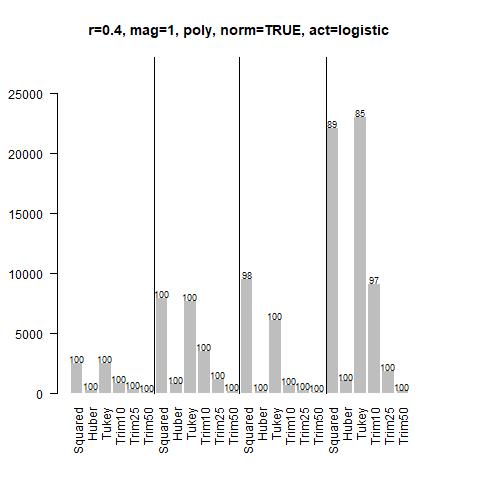}\\
\includegraphics[width=6.75cm,height=6.25cm]{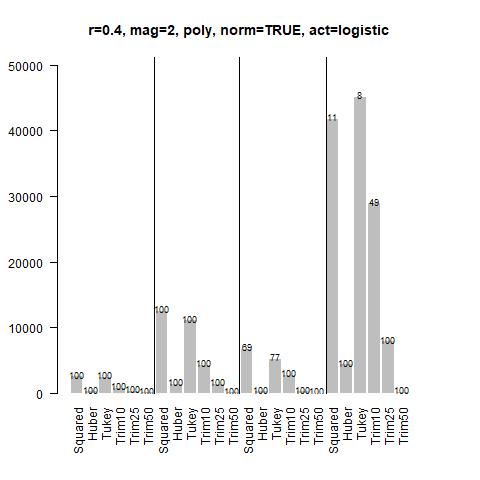} 
\includegraphics[width=6.75cm,height=6.25cm]{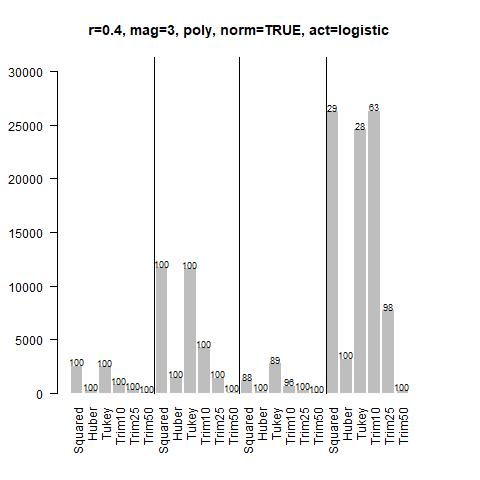} 
\end{center}
\caption{Results for $r=0.4$}
\end{figure}

\subsubsection{Trigonometric function}

\begin{figure}[H]
\label{trimnn:n200p5r10m1trignonlogStep}
\begin{center}
\includegraphics[width=6.75cm,height=6.25cm]{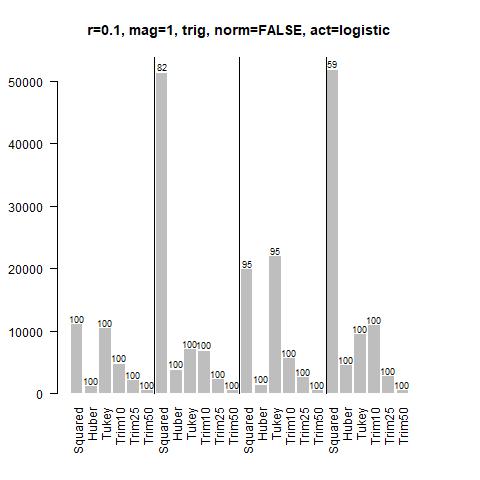}
\includegraphics[width=6.75cm,height=6.25cm]{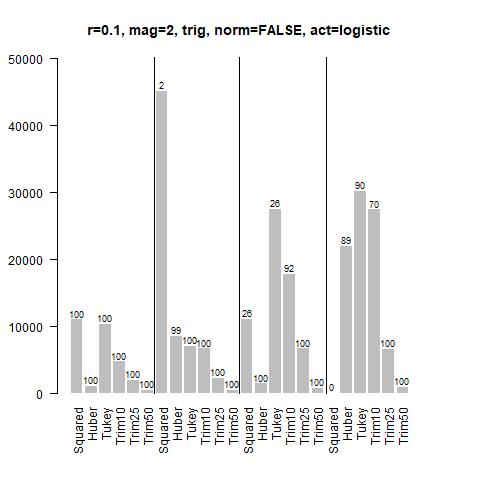} \\
\includegraphics[width=6.75cm,height=6.25cm]{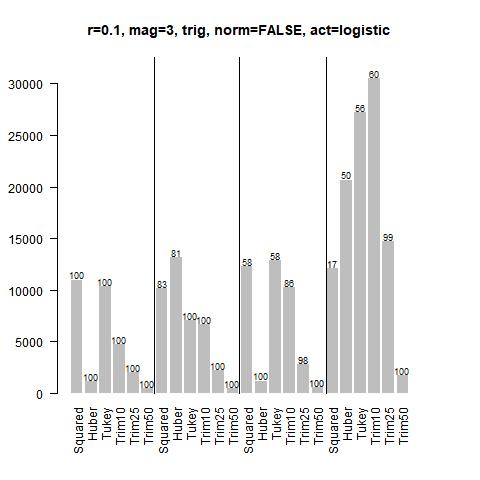} 
\includegraphics[width=6.75cm,height=6.25cm]{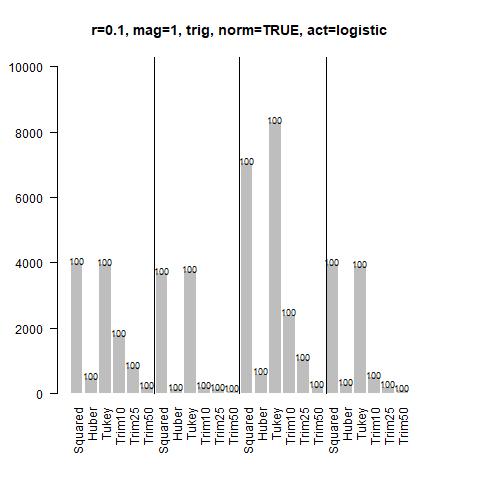}\\
\includegraphics[width=6.75cm,height=6.25cm]{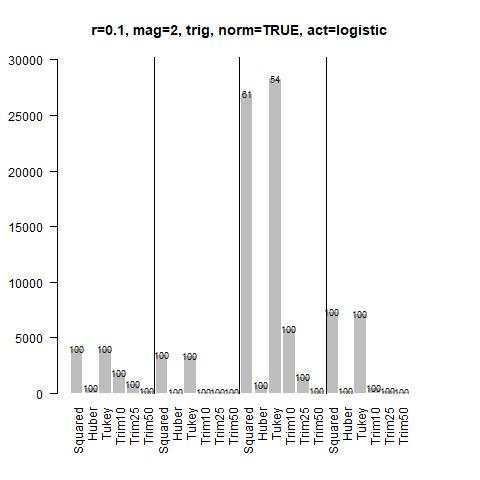} 
\includegraphics[width=6.75cm,height=6.25cm]{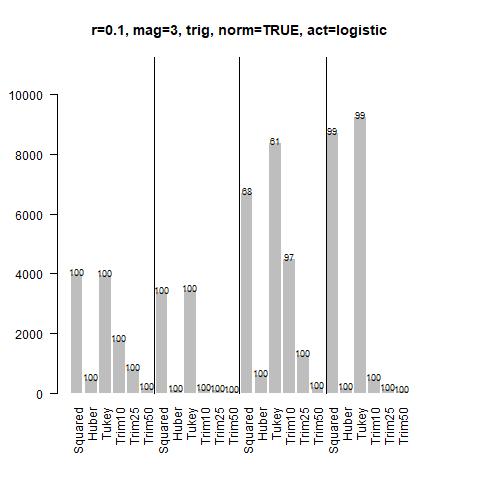} 
\end{center}
\caption{Results for $r=0.1$}
\end{figure}

\begin{figure}[H]
\label{trimnn:n200p5r25m1trignonlogStep}
\begin{center}
\includegraphics[width=6.75cm,height=6.25cm]{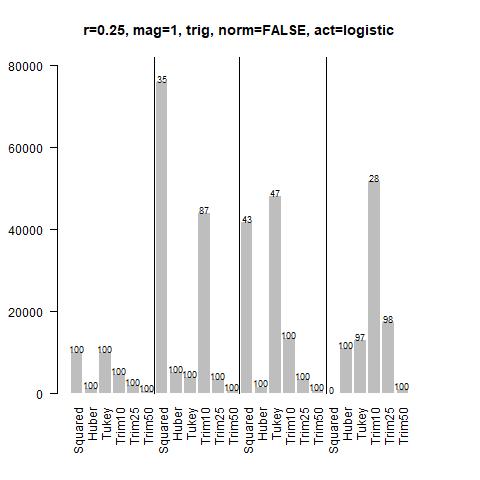}
\includegraphics[width=6.75cm,height=6.25cm]{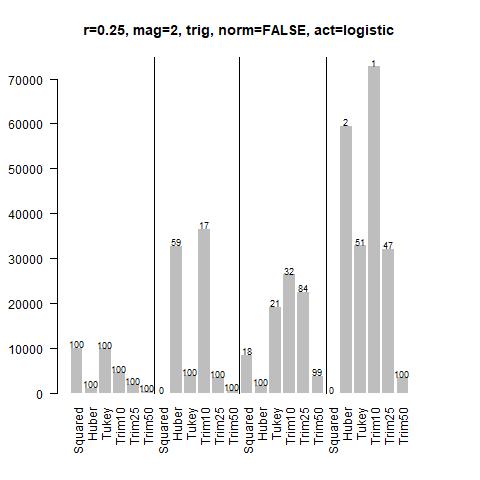} \\
\includegraphics[width=6.75cm,height=6.25cm]{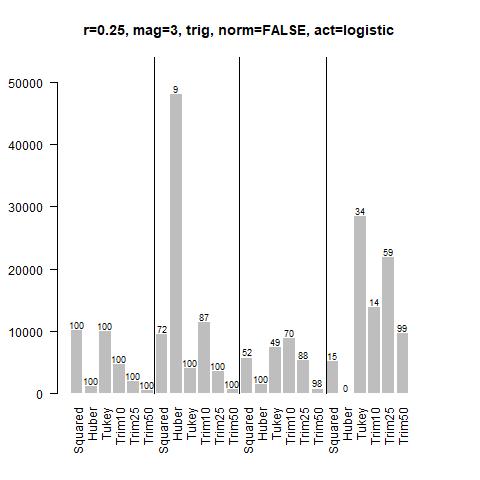} 
\includegraphics[width=6.75cm,height=6.25cm]{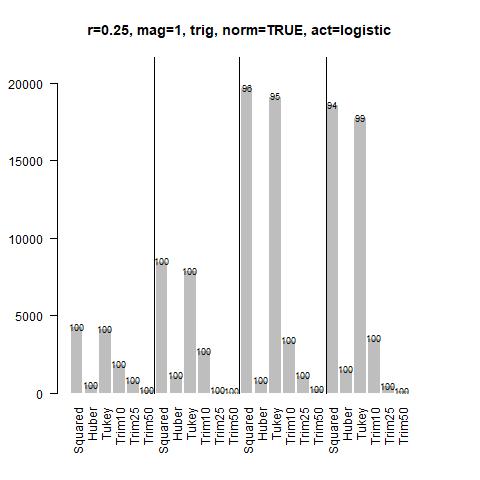}\\
\includegraphics[width=6.75cm,height=6.25cm]{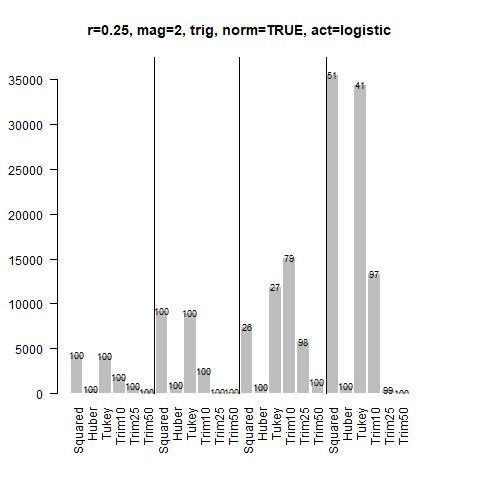} 
\includegraphics[width=6.75cm,height=6.25cm]{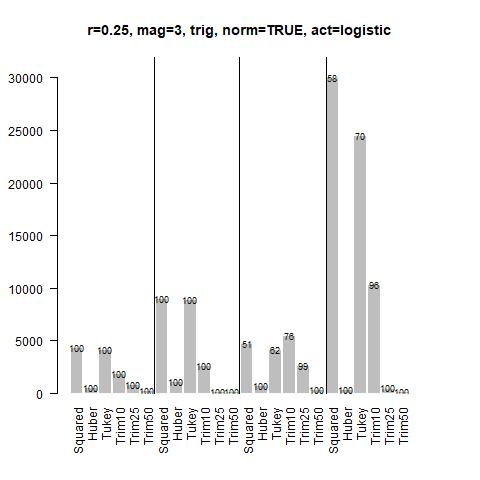} 
\end{center}
\caption{Results for $r=0.25$}
\end{figure}

\begin{figure}[H]
\begin{center}
\includegraphics[width=6.75cm,height=6.25cm]{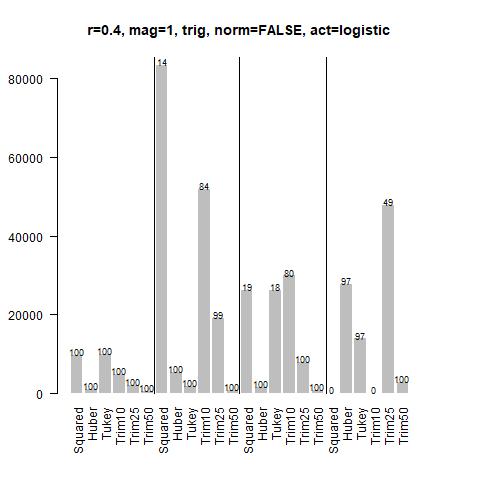}
\includegraphics[width=6.75cm,height=6.25cm]{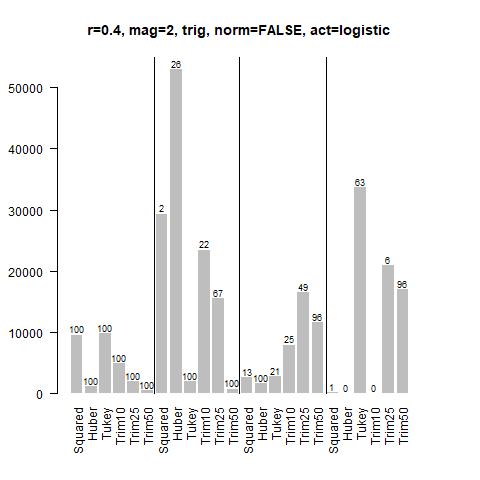} \\
\includegraphics[width=6.75cm,height=6.25cm]{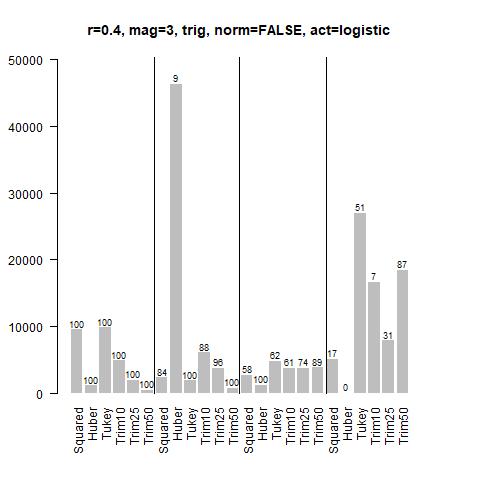} 
\includegraphics[width=6.75cm,height=6.25cm]{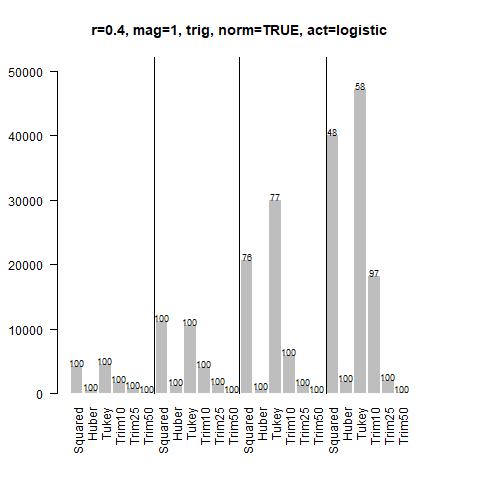}\\
\includegraphics[width=6.75cm,height=6.25cm]{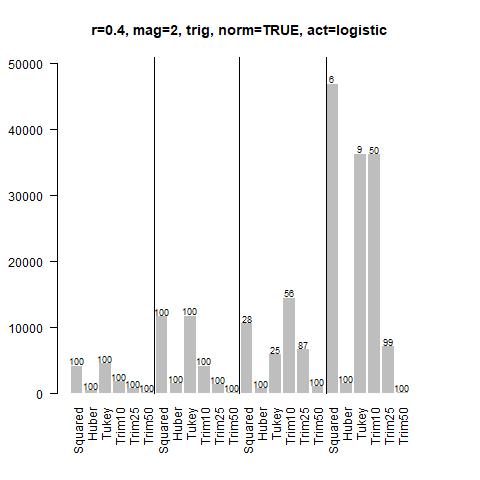} 
\includegraphics[width=6.75cm,height=6.25cm]{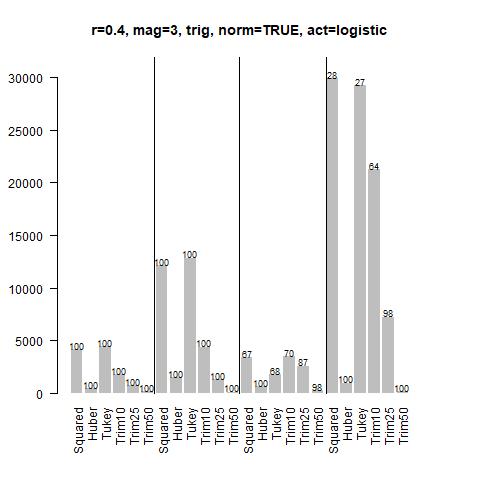} 
\end{center}
\caption{Results for $r=0.4$}\label{trimnn:n200p5r40m1trignonlogStep}
\end{figure}

\subsection{Softplus activation function}

\subsubsection{Linear function}

\begin{figure}[H]
\label{trimnn:n200p5r10m1linnonreluStep}
\begin{center}
\includegraphics[width=6.75cm,height=6.25cm]{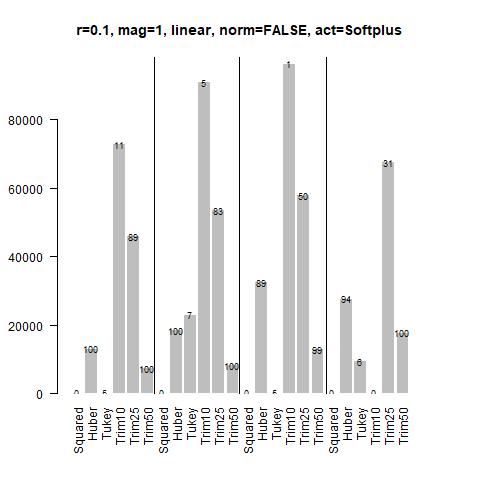}
\includegraphics[width=6.75cm,height=6.25cm]{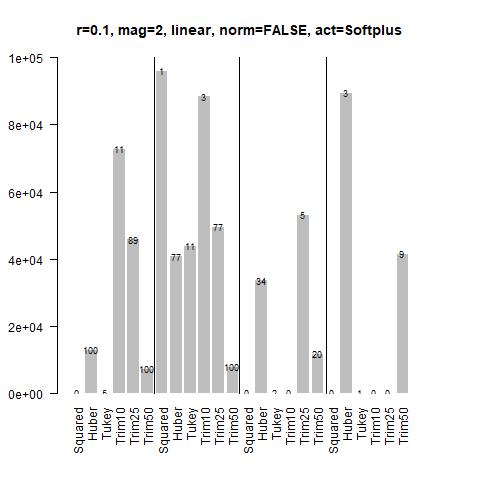} \\
\includegraphics[width=6.75cm,height=6.25cm]{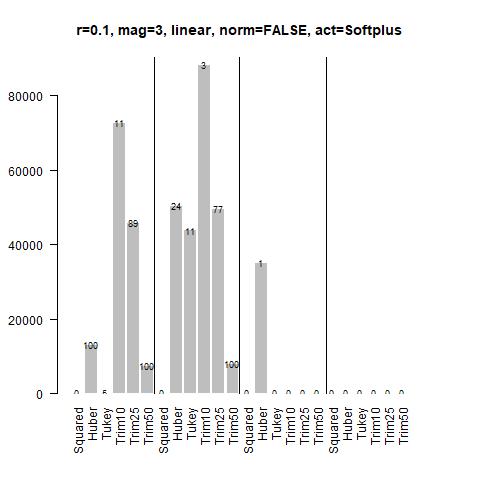} 
\includegraphics[width=6.75cm,height=6.25cm]{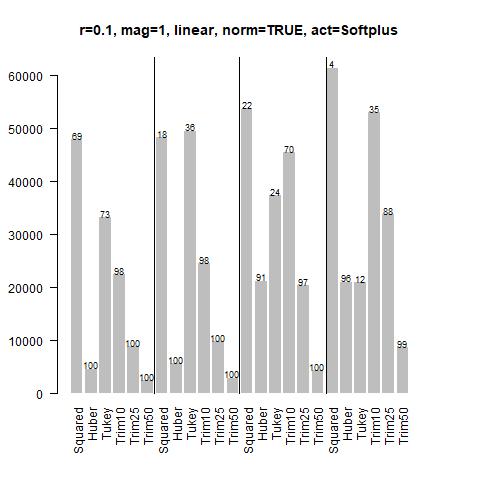}\\
\includegraphics[width=6.75cm,height=6.25cm]{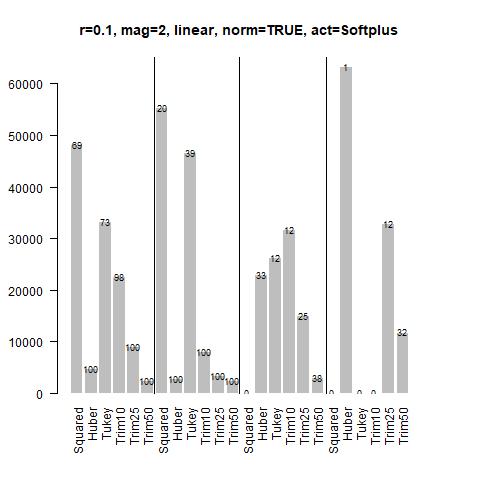} 
\includegraphics[width=6.75cm,height=6.25cm]{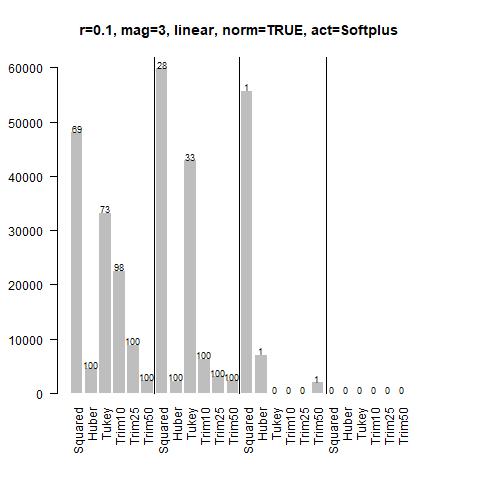} 
\end{center}
\caption{Results for $r=0.1$}
\end{figure}

\begin{figure}[H]
\label{trimnn:n200p5r25m1linnonreluStep}
\begin{center}
\includegraphics[width=6.75cm,height=6.25cm]{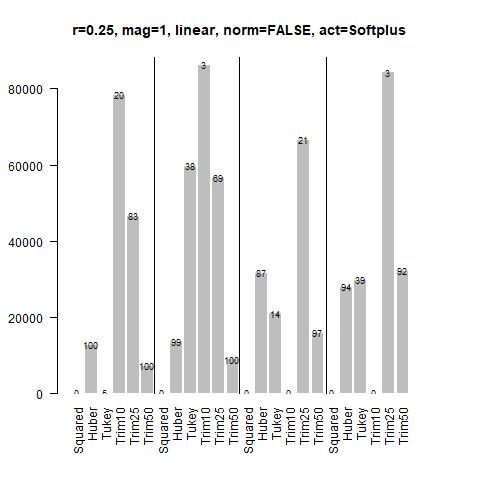}
\includegraphics[width=6.75cm,height=6.25cm]{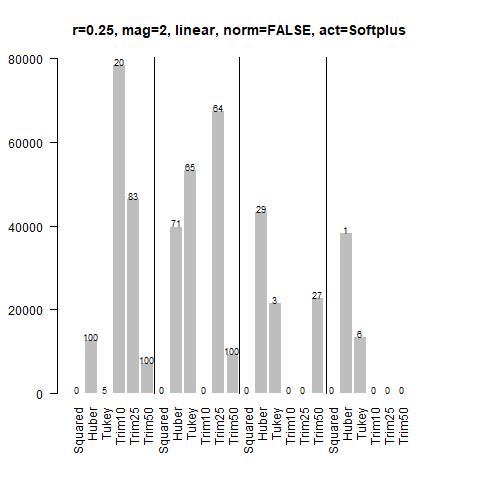} \\
\includegraphics[width=6.75cm,height=6.25cm]{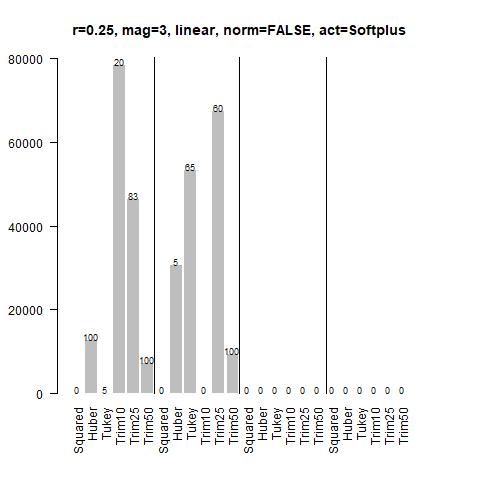} 
\includegraphics[width=6.75cm,height=6.25cm]{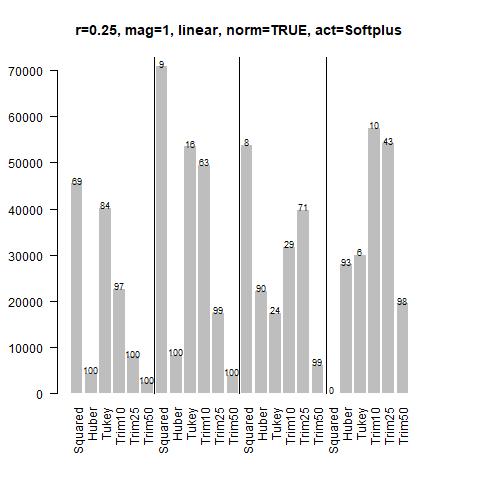}\\
\includegraphics[width=6.75cm,height=6.25cm]{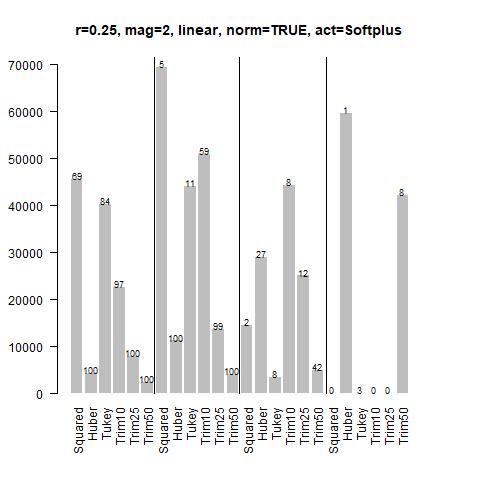} 
\includegraphics[width=6.75cm,height=6.25cm]{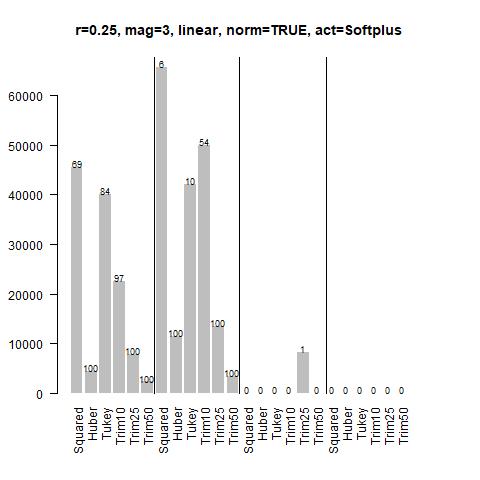} 
\end{center}
\caption{Results for $r=0.25$}
\end{figure}

\begin{figure}[H]
\label{trimnn:n200p5r40m1linnonreluStep}
\begin{center}
\includegraphics[width=6.75cm,height=6.25cm]{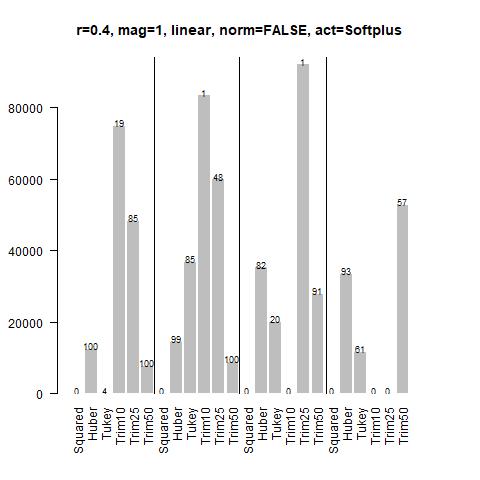}
\includegraphics[width=6.75cm,height=6.25cm]{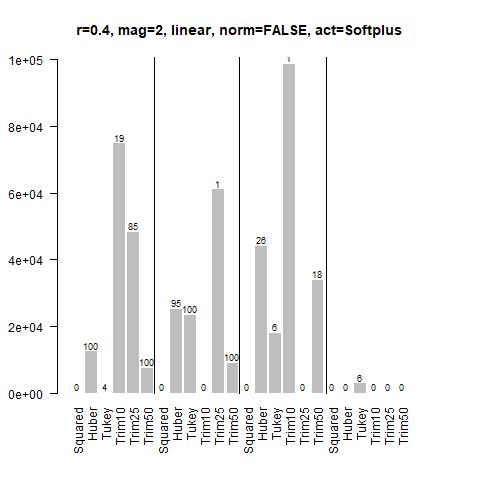} \\
\includegraphics[width=6.75cm,height=6.25cm]{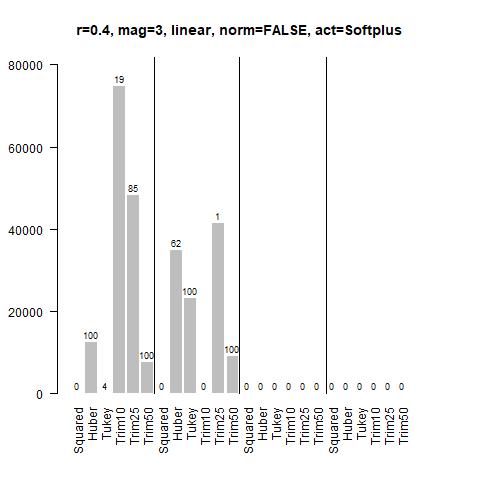} 
\includegraphics[width=6.75cm,height=6.25cm]{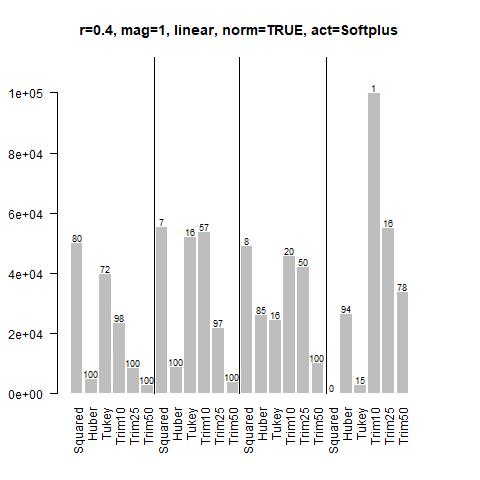}\\
\includegraphics[width=6.75cm,height=6.25cm]{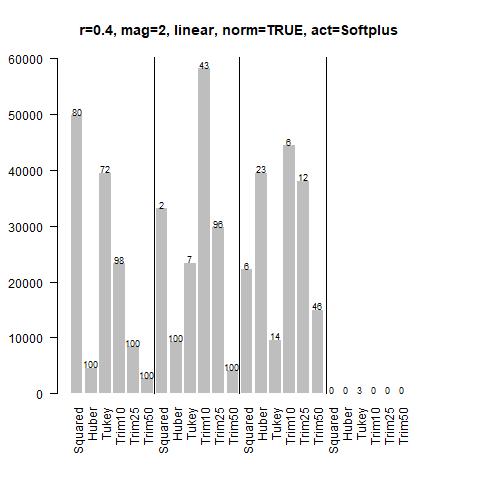} 
\includegraphics[width=6.75cm,height=6.25cm]{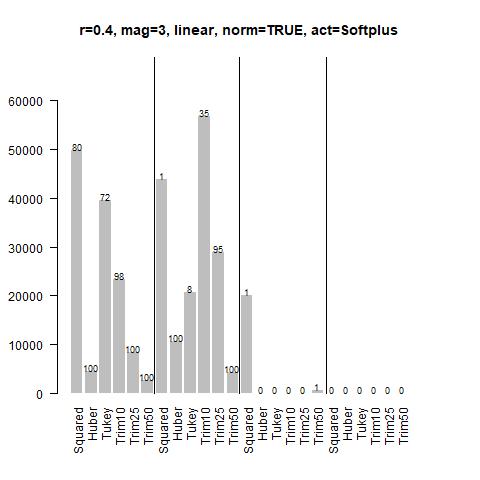} 
\end{center}
\caption{Results for $r=0.4$}
\end{figure}

\subsubsection{Polynomial function}

\begin{figure}[H]
\label{trimnn:n200p5r10m1polynonreluStep}
\begin{center}
\includegraphics[width=6.75cm,height=6.25cm]{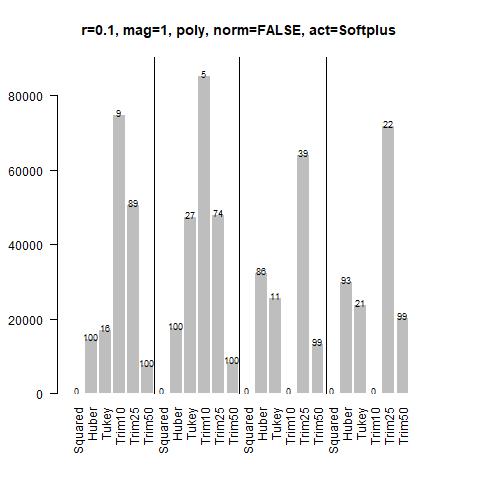}
\includegraphics[width=6.75cm,height=6.25cm]{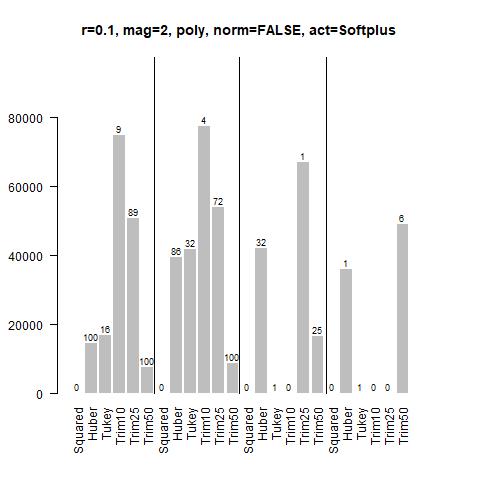} \\
\includegraphics[width=6.75cm,height=6.25cm]{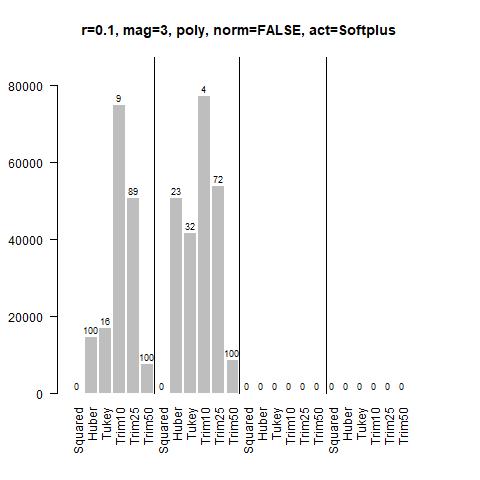} 
\includegraphics[width=6.75cm,height=6.25cm]{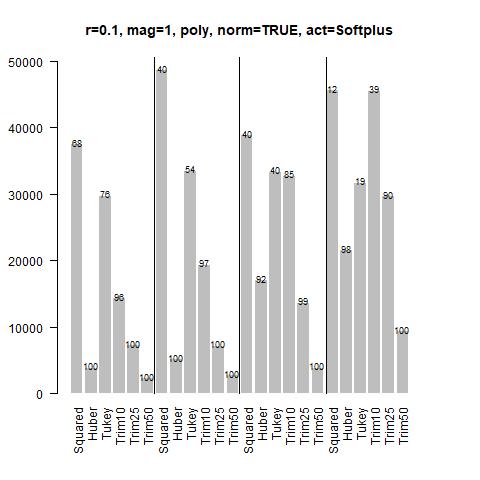}\\
\includegraphics[width=6.75cm,height=6.25cm]{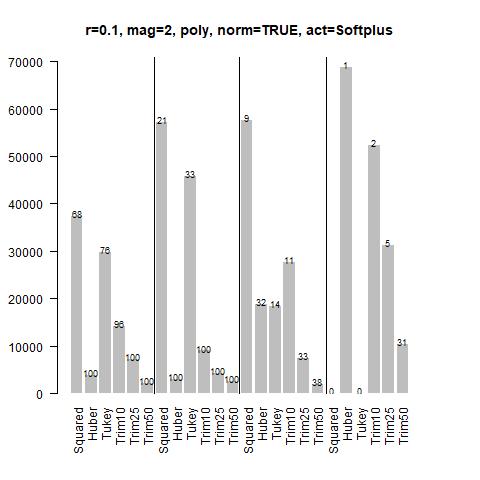} 
\includegraphics[width=6.75cm,height=6.25cm]{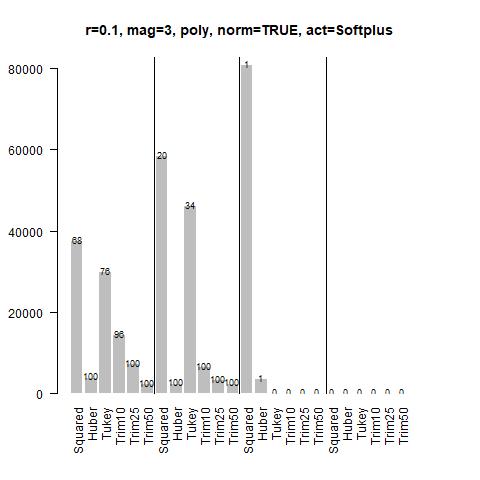} 
\end{center}
\caption{Results for $r=0.1$}
\end{figure}

\begin{figure}[H]
\label{trimnn:n200p5r25m1polynonreluStep}
\begin{center}
\includegraphics[width=6.75cm,height=6.25cm]{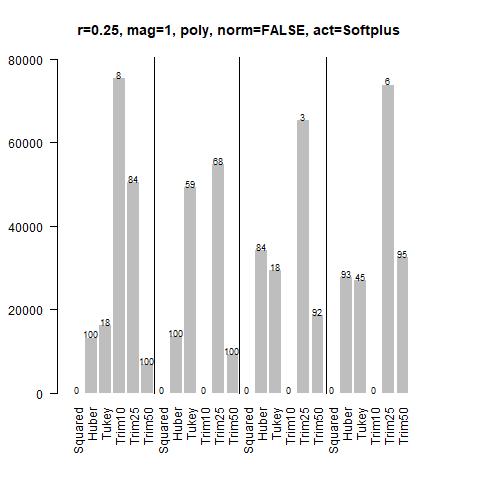}
\includegraphics[width=6.75cm,height=6.25cm]{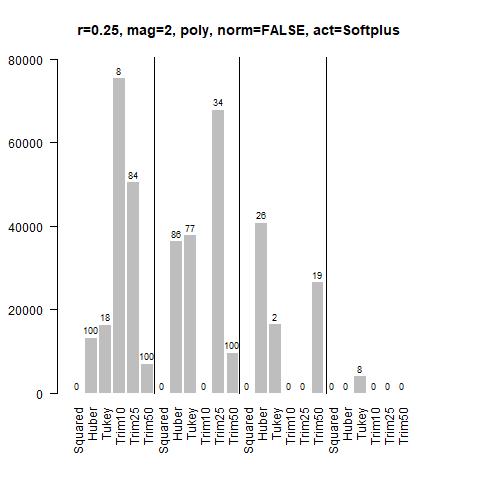} \\
\includegraphics[width=6.75cm,height=6.25cm]{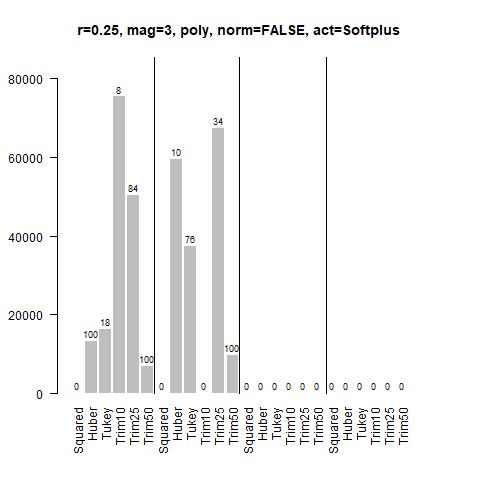} 
\includegraphics[width=6.75cm,height=6.25cm]{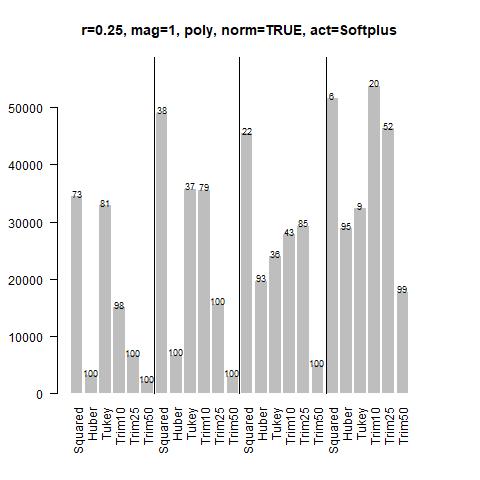}\\
\includegraphics[width=6.75cm,height=6.25cm]{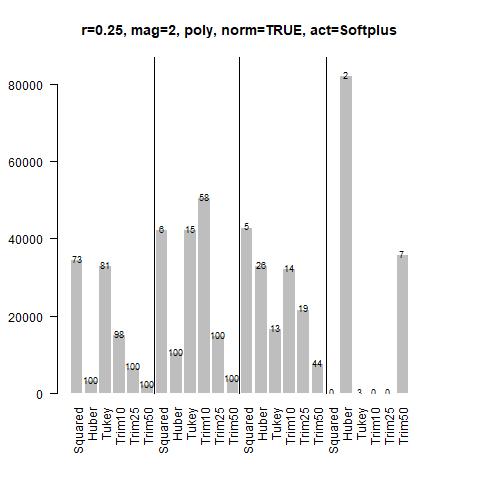} 
\includegraphics[width=6.75cm,height=6.25cm]{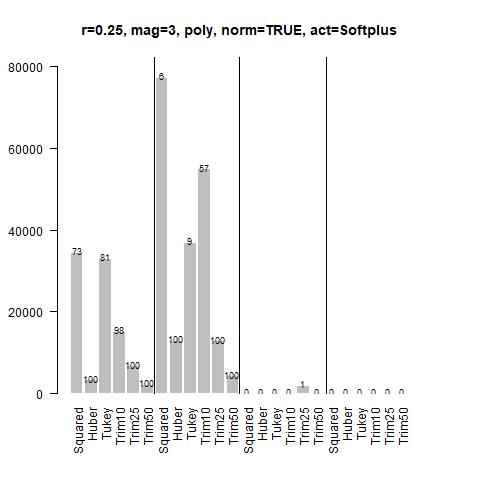} 
\end{center}
\caption{Results for $r=0.25$}
\end{figure}

\begin{figure}[H]
\label{trimnn:n200p5r40m1polynonreluStep}
\begin{center}
\includegraphics[width=6.75cm,height=6.25cm]{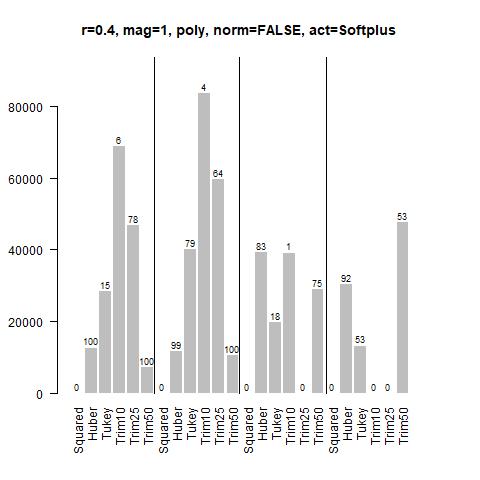}
\includegraphics[width=6.75cm,height=6.25cm]{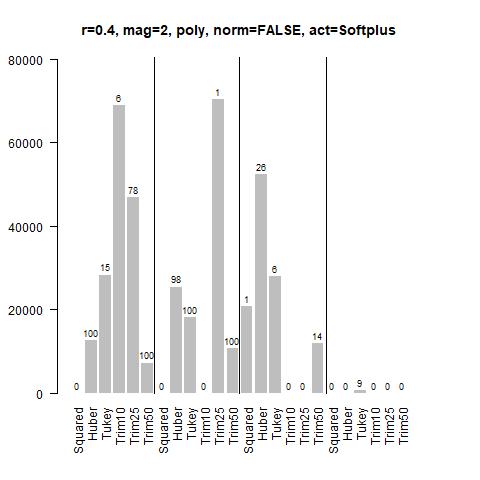} \\
\includegraphics[width=6.75cm,height=6.25cm]{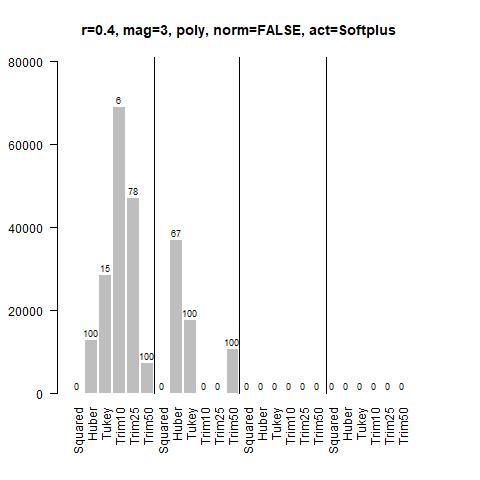} 
\includegraphics[width=6.75cm,height=6.25cm]{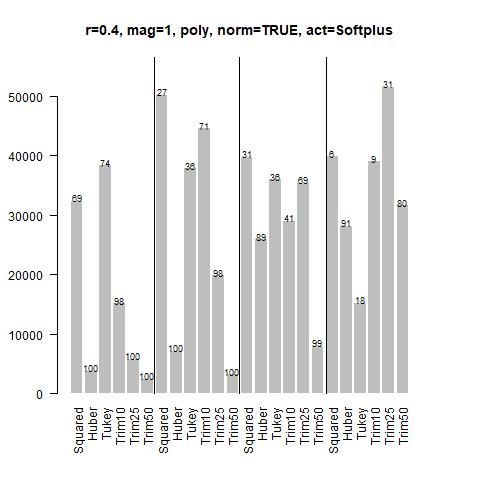}\\
\includegraphics[width=6.75cm,height=6.25cm]{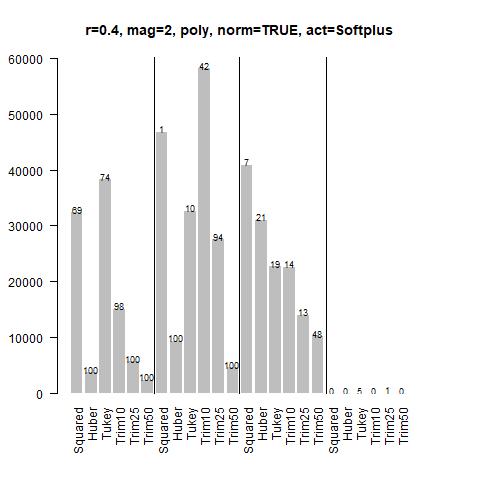} 
\includegraphics[width=6.75cm,height=6.25cm]{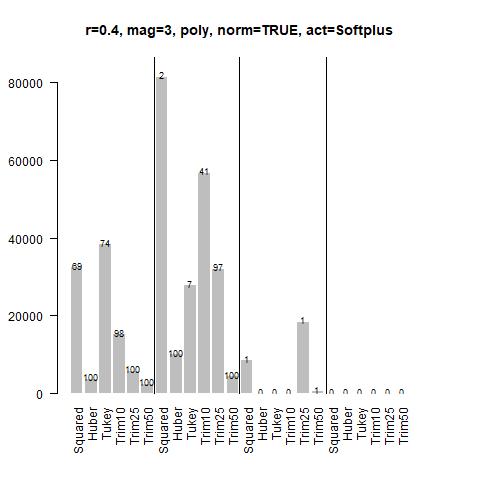} 
\end{center}
\caption{Results for $r=0.4$}
\end{figure}

\subsubsection{Trigonometric function}

\begin{figure}[H]
\begin{center}
\includegraphics[width=6.75cm,height=6.25cm]{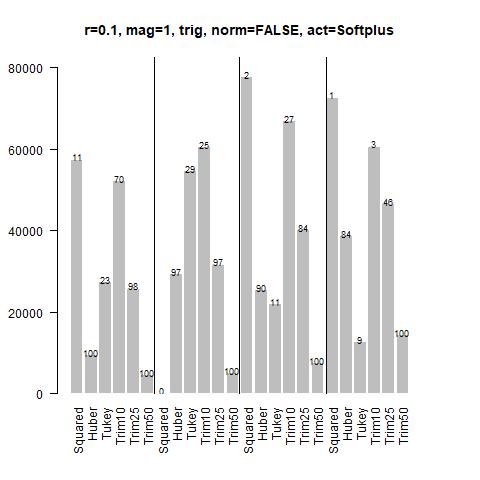}
\includegraphics[width=6.75cm,height=6.25cm]{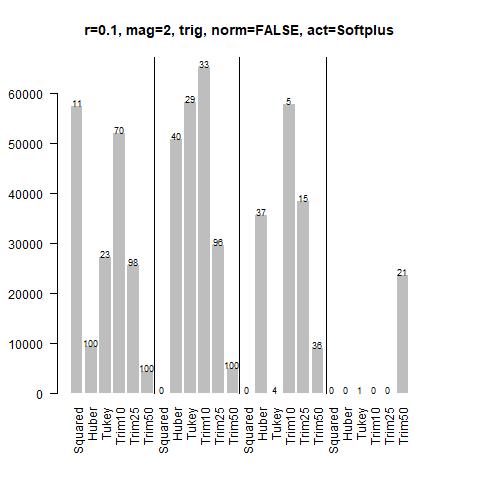} \\
\includegraphics[width=6.75cm,height=6.25cm]{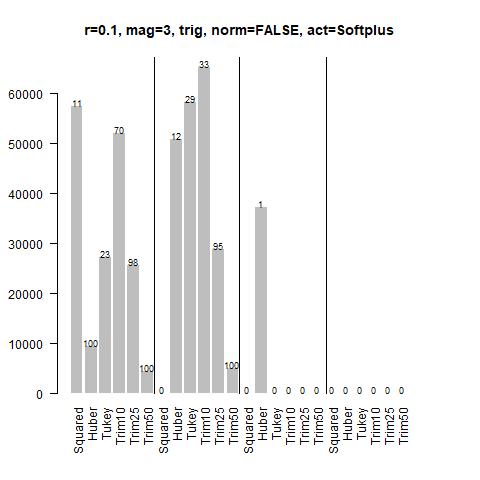} 
\includegraphics[width=6.75cm,height=6.25cm]{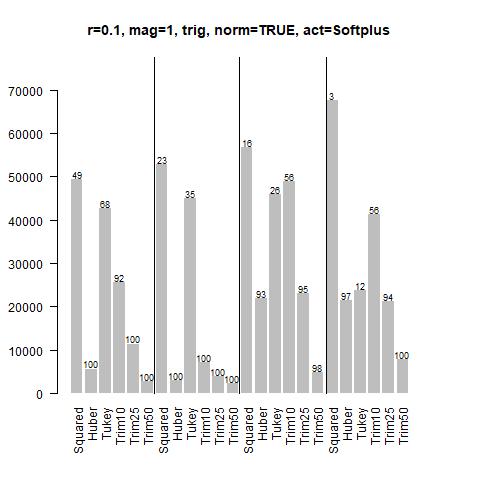}\\
\includegraphics[width=6.75cm,height=6.25cm]{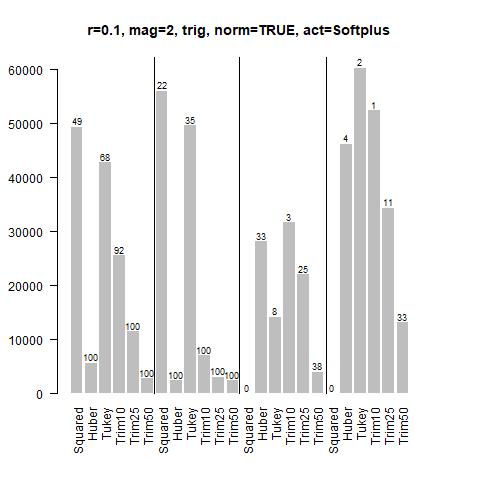} 
\includegraphics[width=6.75cm,height=6.25cm]{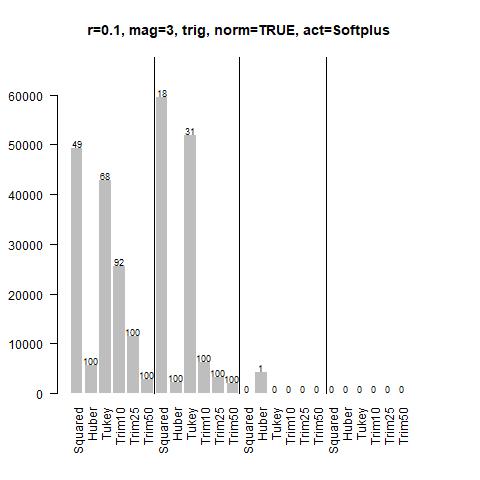} 
\end{center}
\caption{Results for $r=0.1$}\label{trimnn:n200p5r10m1trignonreluStep}
\end{figure}

\begin{figure}[H]
\label{trimnn:n200p5r25m1trignonreluStep}
\begin{center}
\includegraphics[width=6.75cm,height=6.25cm]{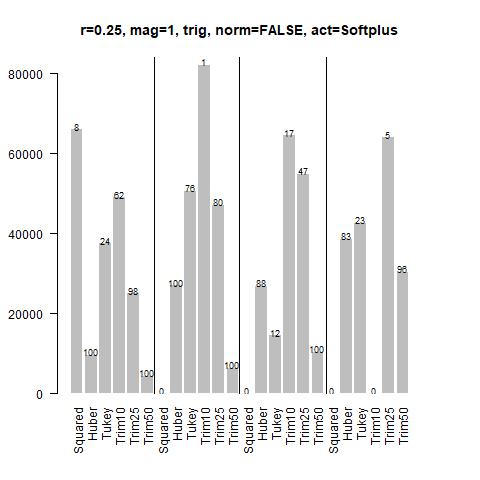}
\includegraphics[width=6.75cm,height=6.25cm]{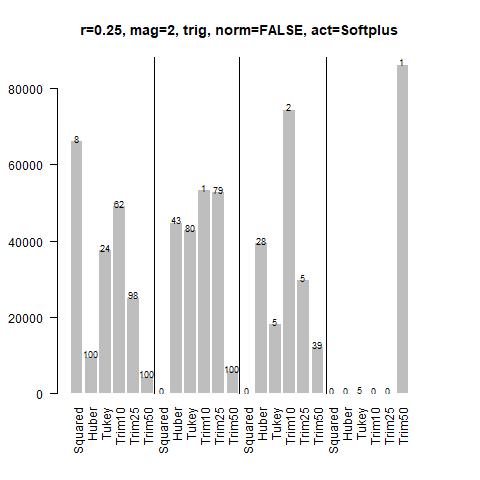} \\
\includegraphics[width=6.75cm,height=6.25cm]{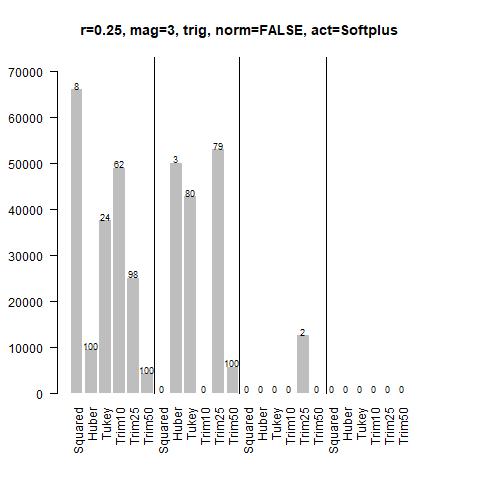} 
\includegraphics[width=6.75cm,height=6.25cm]{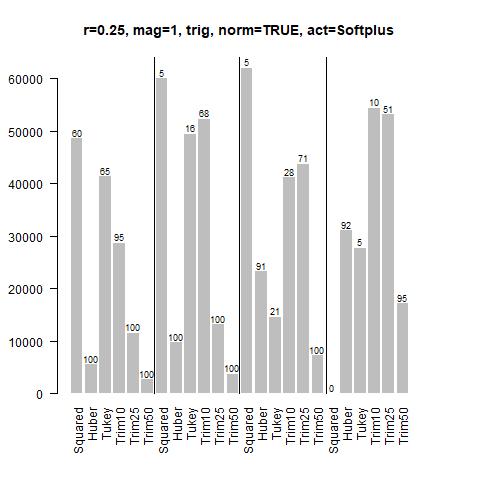}\\
\includegraphics[width=6.75cm,height=6.25cm]{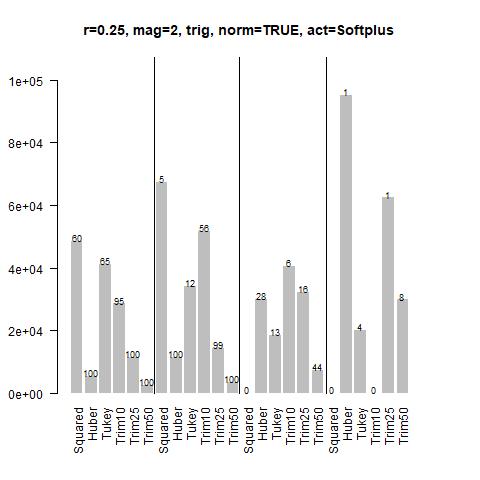} 
\includegraphics[width=6.75cm,height=6.25cm]{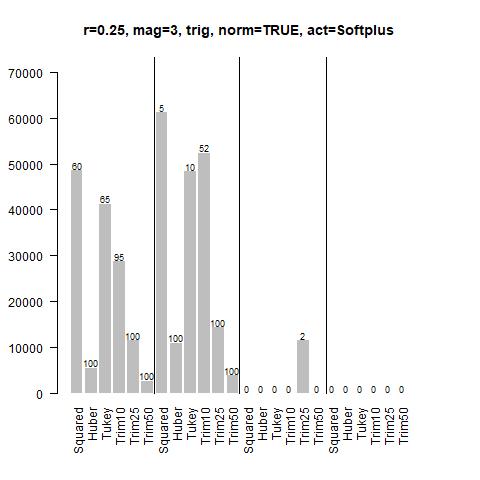} 
\end{center}
\caption{Results for $r=0.25$}
\end{figure}

\begin{figure}[H]
\label{trimnn:n200p5r40m1trignonreluStep}
\begin{center}
\includegraphics[width=6.75cm,height=6.25cm]{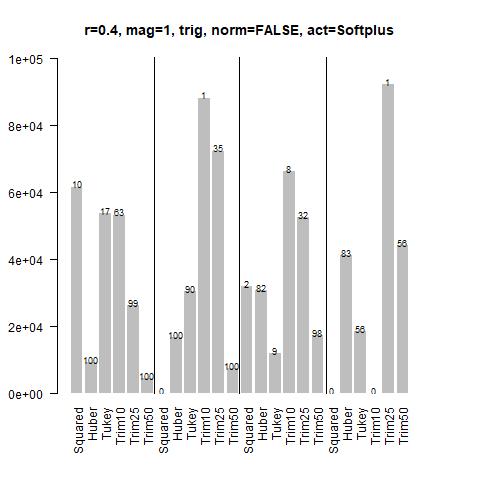}
\includegraphics[width=6.75cm,height=6.25cm]{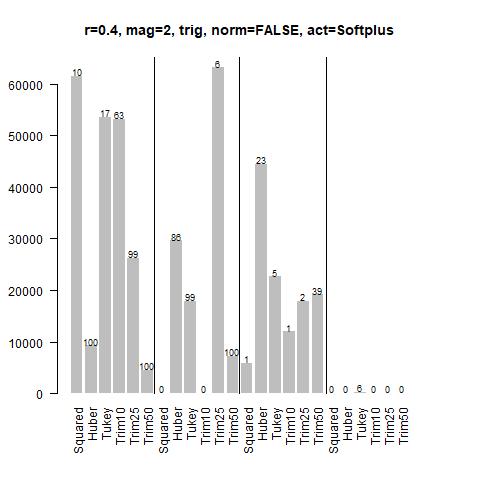} \\
\includegraphics[width=6.75cm,height=6.25cm]{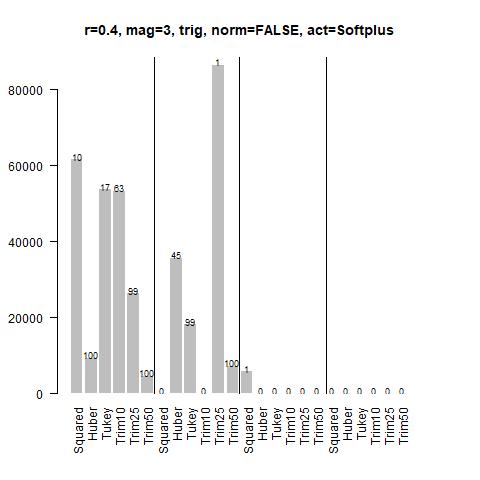} 
\includegraphics[width=6.75cm,height=6.25cm]{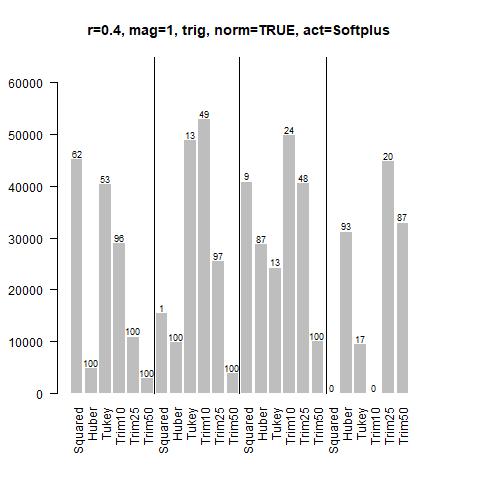}\\
\includegraphics[width=6.75cm,height=6.25cm]{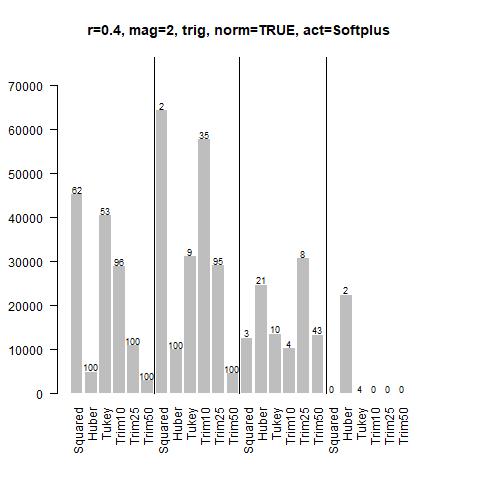} 
\includegraphics[width=6.75cm,height=6.25cm]{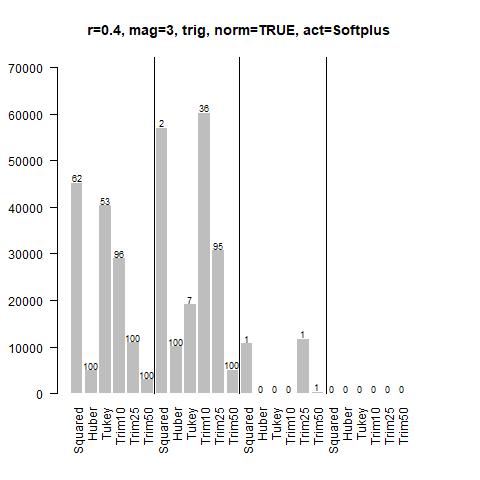} 
\end{center}
\caption{Results for $r=0.4$}
\end{figure}

\section{Simulation results for $n=150$ and $p=5$, deep network: Training steps} \label{trimnn:secstep1505deep}

\subsection{Logistic activation function}

\subsubsection{Linear function}

\begin{figure}[H]
\label{trimnn:n200p5r10m1linnonlogdeepStep}
\begin{center}
\includegraphics[width=6.75cm,height=6.25cm]{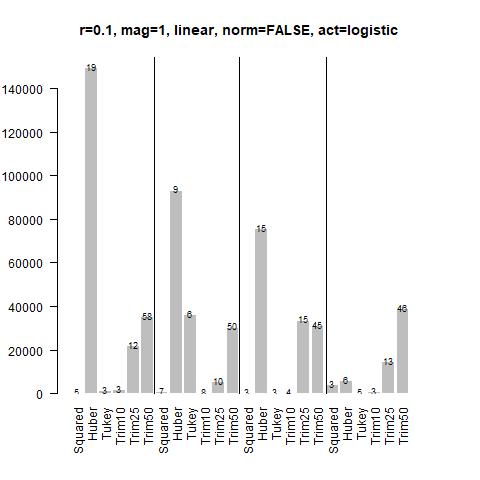}
\includegraphics[width=6.75cm,height=6.25cm]{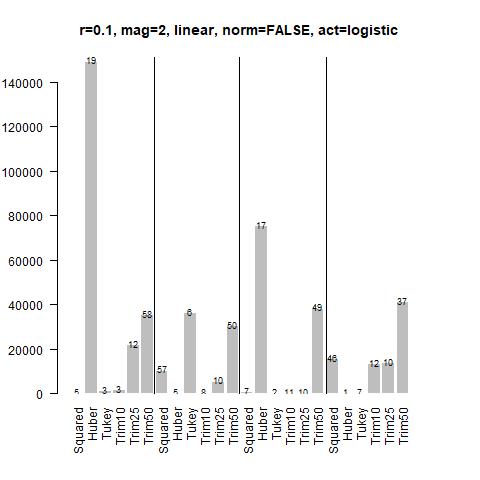} \\
\includegraphics[width=6.75cm,height=6.25cm]{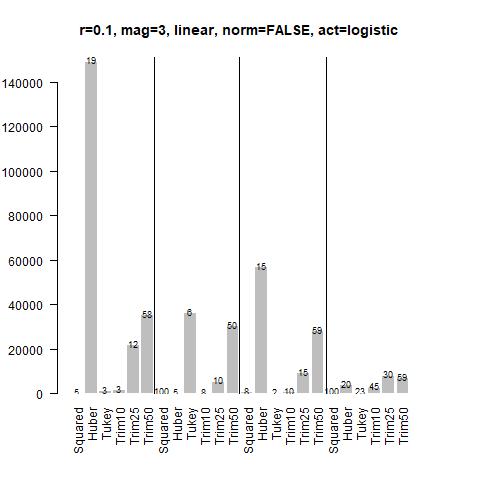} 
\includegraphics[width=6.75cm,height=6.25cm]{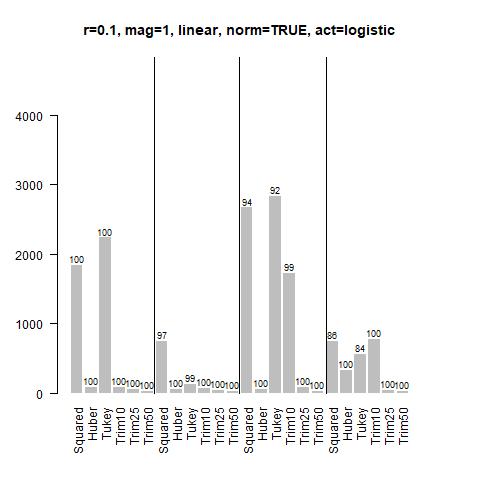}\\
\includegraphics[width=6.75cm,height=6.25cm]{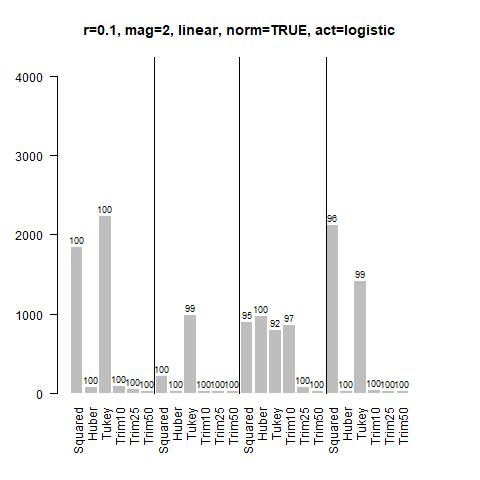} 
\includegraphics[width=6.75cm,height=6.25cm]{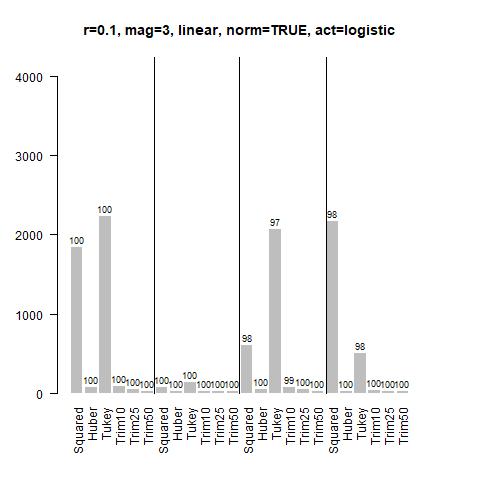} 
\end{center}
\caption{Results for $r=0.1$}
\end{figure}

\begin{figure}[H]
\label{trimnn:n200p5r25m1linnonlogdeepStep}
\begin{center}
\includegraphics[width=6.75cm,height=6.25cm]{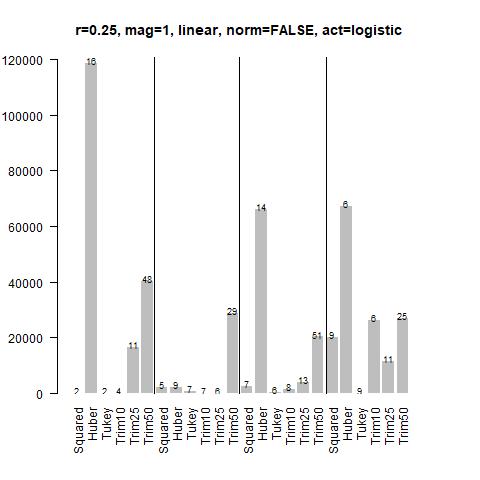}
\includegraphics[width=6.75cm,height=6.25cm]{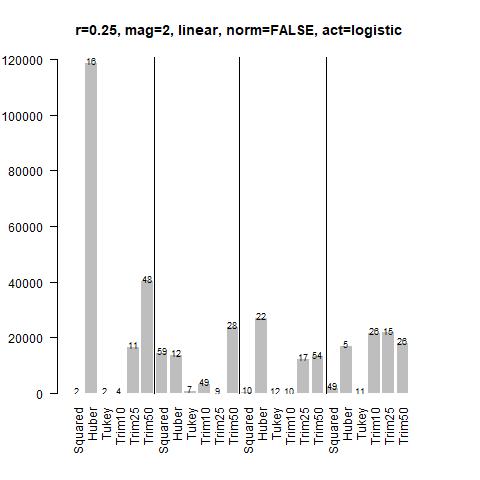} \\
\includegraphics[width=6.75cm,height=6.25cm]{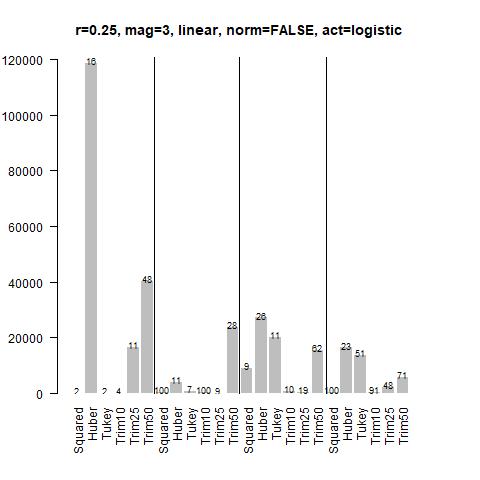} 
\includegraphics[width=6.75cm,height=6.25cm]{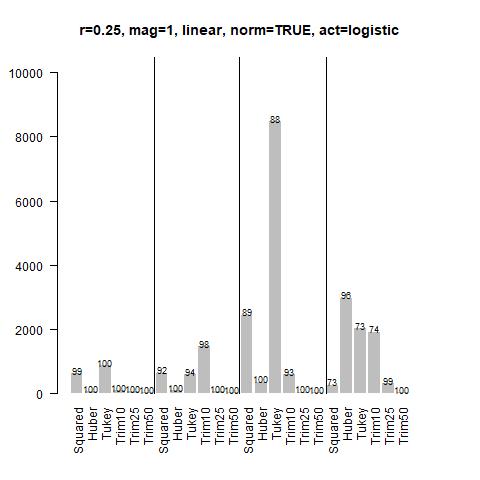}\\
\includegraphics[width=6.75cm,height=6.25cm]{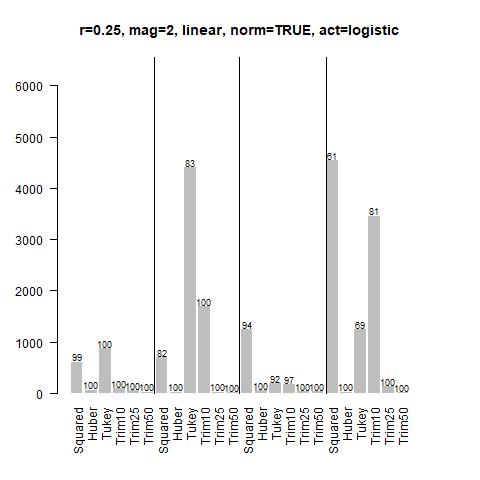} 
\includegraphics[width=6.75cm,height=6.25cm]{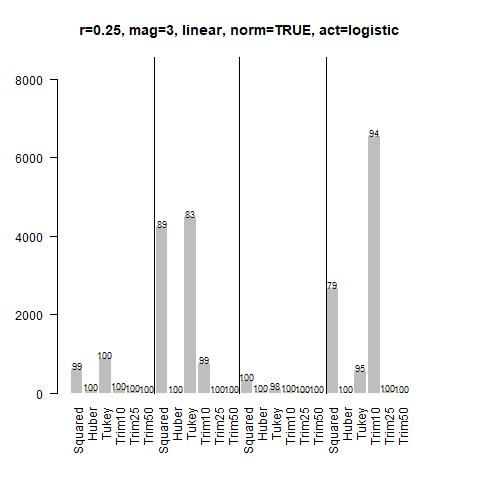} 
\end{center}
\caption{Results for $r=0.25$}
\end{figure}

\begin{figure}[H]
\begin{center}
\includegraphics[width=6.75cm,height=6.25cm]{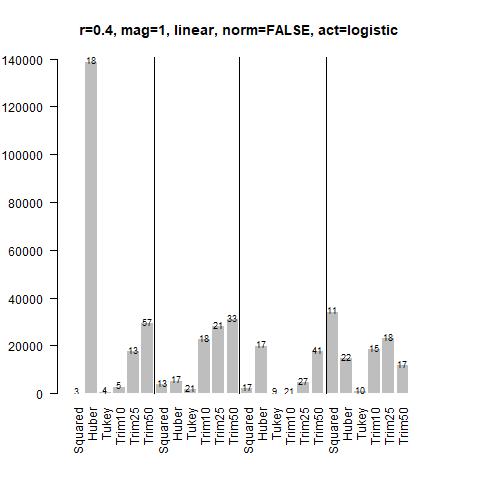}
\includegraphics[width=6.75cm,height=6.25cm]{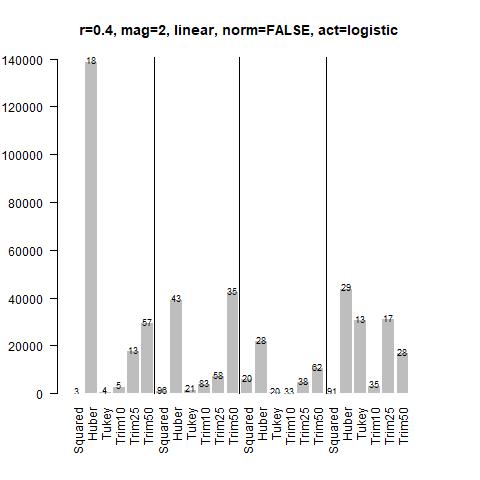} \\
\includegraphics[width=6.75cm,height=6.25cm]{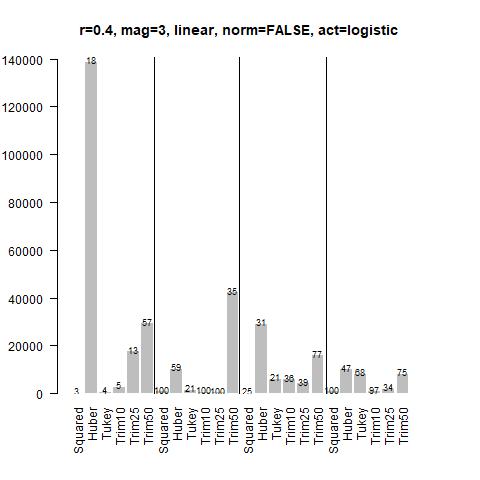} 
\includegraphics[width=6.75cm,height=6.25cm]{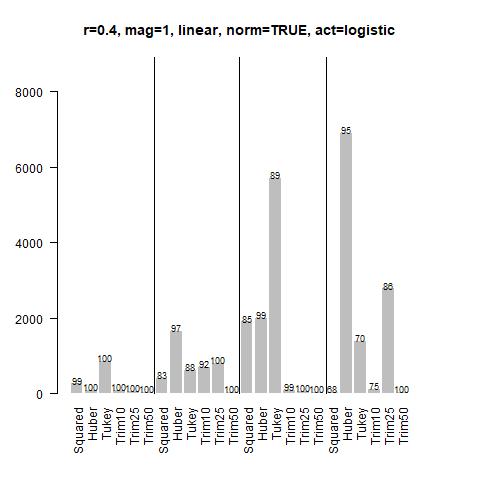}\\
\includegraphics[width=6.75cm,height=6.25cm]{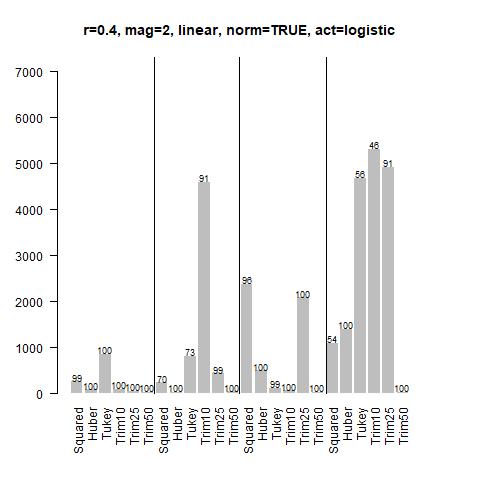} 
\includegraphics[width=6.75cm,height=6.25cm]{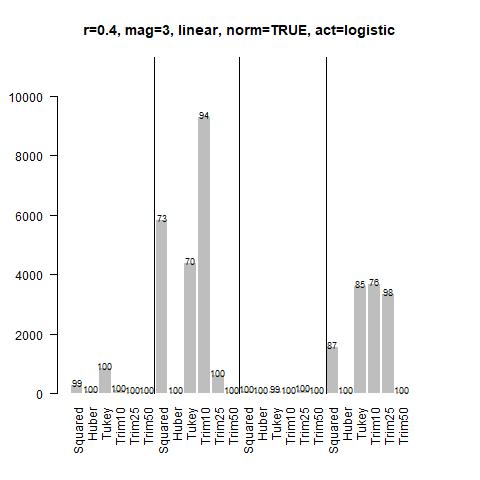} 
\end{center}
\caption{Results for $r=0.4$}\label{trimnn:n200p5r40m1linnonlogdeepStep}
\end{figure}

\subsubsection{Polynomial function}

\begin{figure}[H]
\label{trimnn:n200p5r10m1polynonlogdeepStep}
\begin{center}
\includegraphics[width=6.75cm,height=6.25cm]{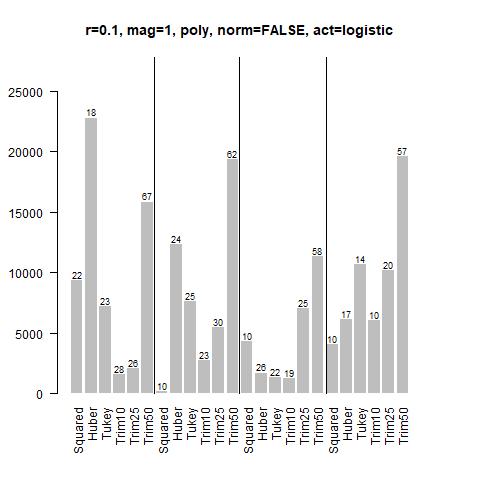}
\includegraphics[width=6.75cm,height=6.25cm]{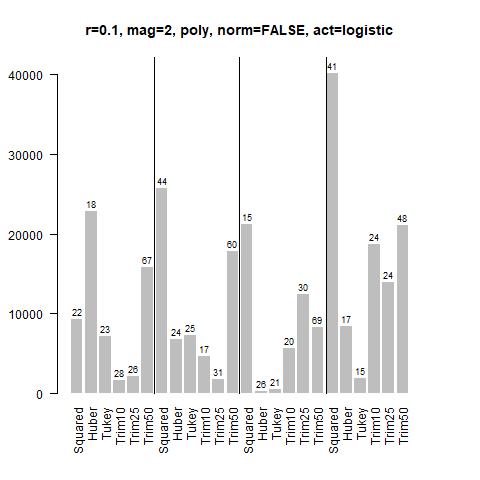} \\
\includegraphics[width=6.75cm,height=6.25cm]{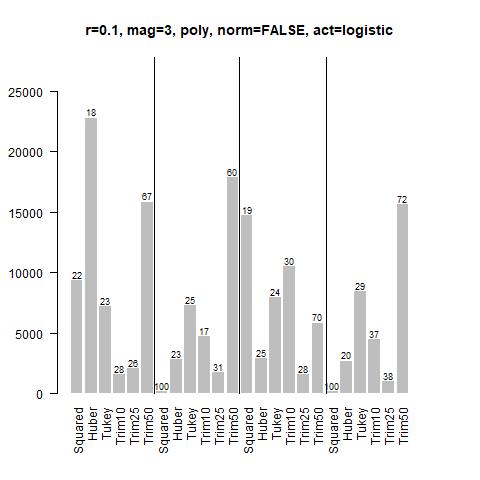} 
\includegraphics[width=6.75cm,height=6.25cm]{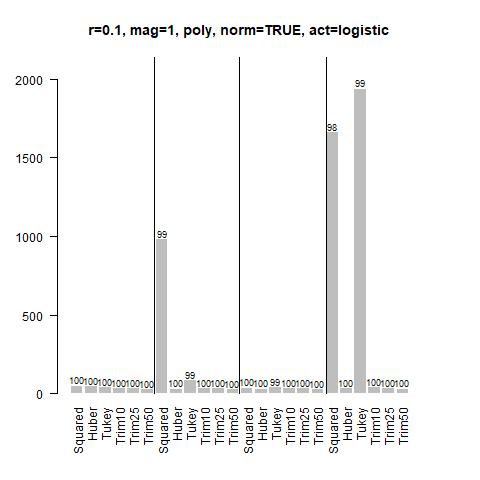}\\
\includegraphics[width=6.75cm,height=6.25cm]{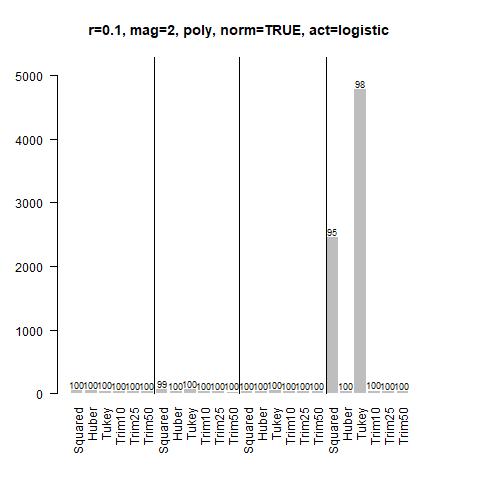} 
\includegraphics[width=6.75cm,height=6.25cm]{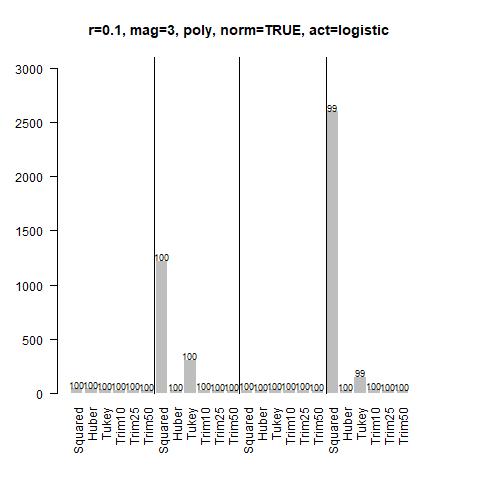} 
\end{center}
\caption{Results for $r=0.1$}
\end{figure}

\begin{figure}[H]
\label{trimnn:n200p5r25m1polynonlogdeepStep}
\begin{center}
\includegraphics[width=6.75cm,height=6.25cm]{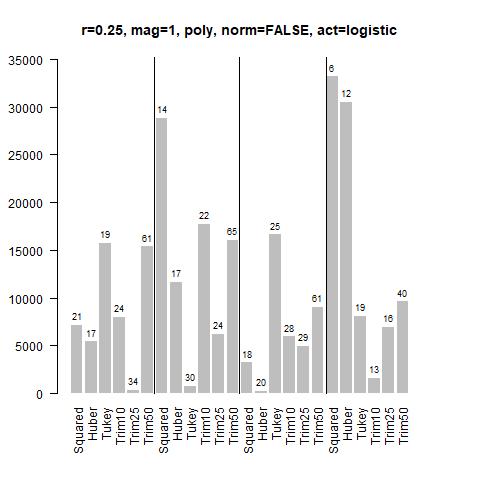}
\includegraphics[width=6.75cm,height=6.25cm]{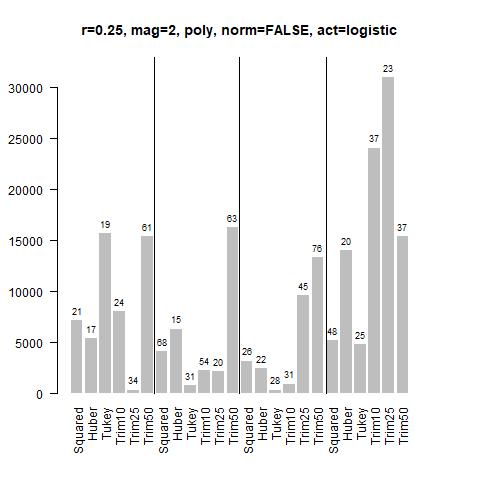} \\
\includegraphics[width=6.75cm,height=6.25cm]{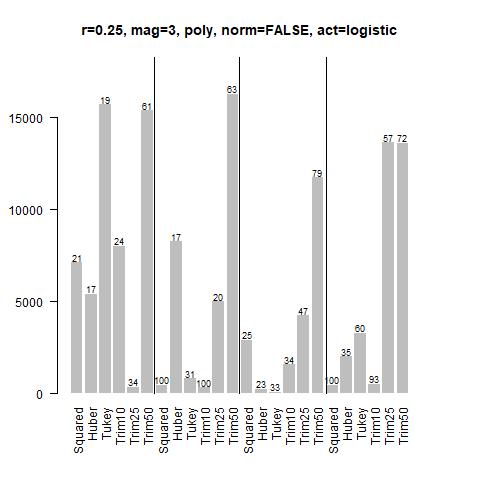} 
\includegraphics[width=6.75cm,height=6.25cm]{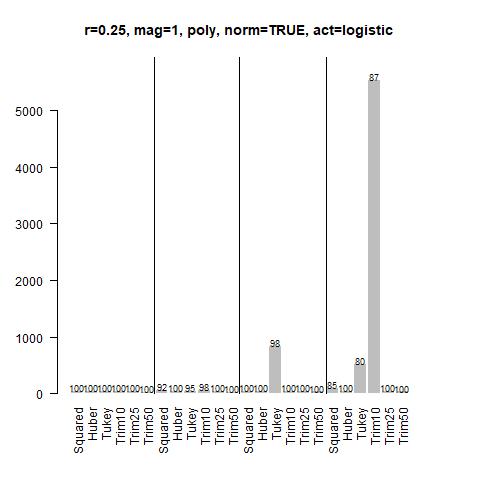}\\
\includegraphics[width=6.75cm,height=6.25cm]{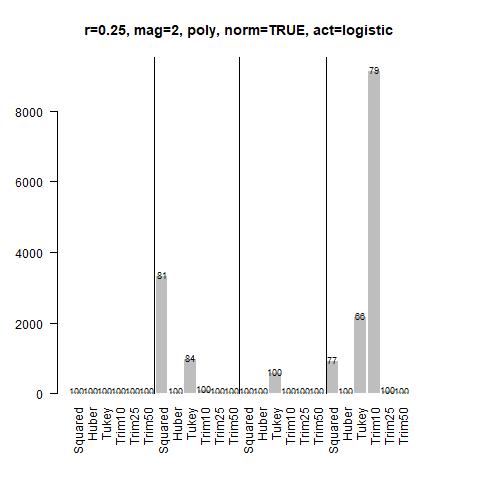} 
\includegraphics[width=6.75cm,height=6.25cm]{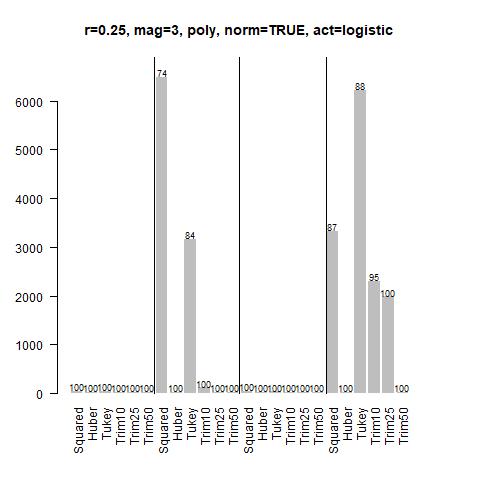} 
\end{center}
\caption{Results for $r=0.25$}
\end{figure}

\begin{figure}[H]
\label{trimnn:n200p5r40m1polynonlogdeepStep}
\begin{center}
\includegraphics[width=6.75cm,height=6.25cm]{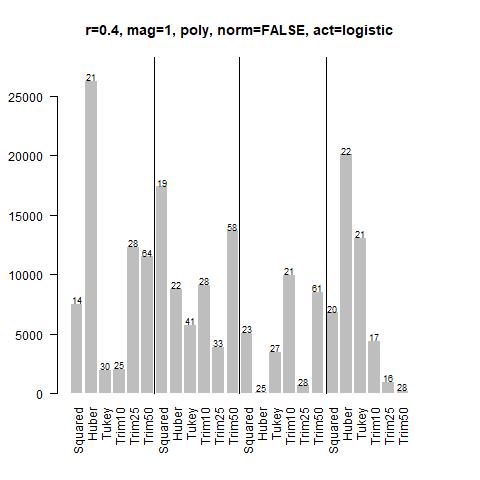}
\includegraphics[width=6.75cm,height=6.25cm]{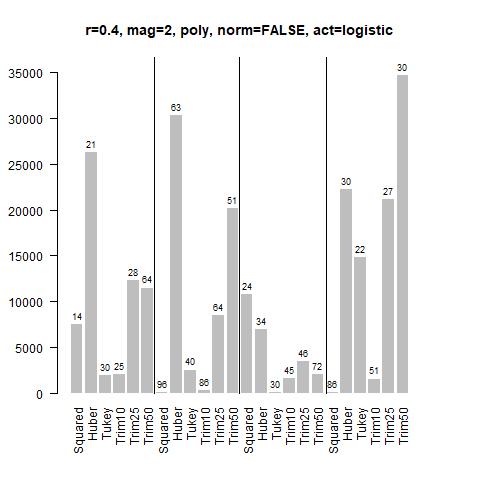} \\
\includegraphics[width=6.75cm,height=6.25cm]{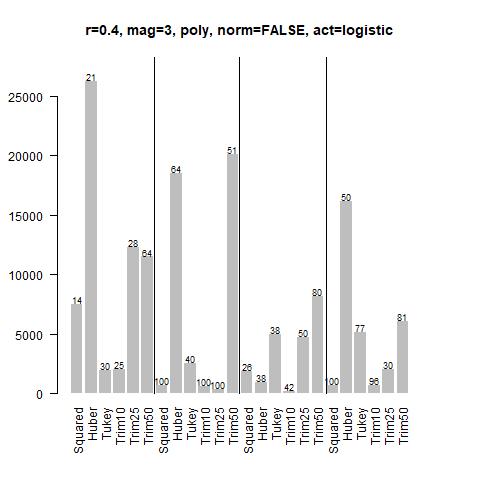} 
\includegraphics[width=6.75cm,height=6.25cm]{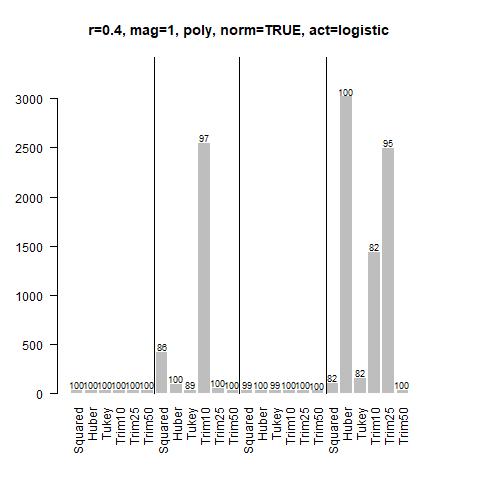}\\
\includegraphics[width=6.75cm,height=6.25cm]{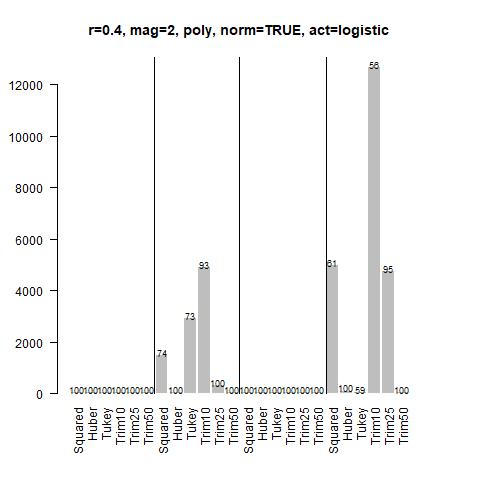} 
\includegraphics[width=6.75cm,height=6.25cm]{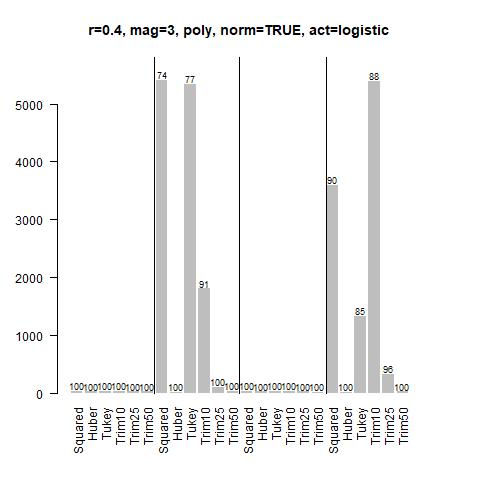} 
\end{center}
\caption{Results for $r=0.4$}
\end{figure}

\subsubsection{Trigonometric function}

\begin{figure}[H]
\label{trimnn:n200p5r10m1trignonlogdeepStep}
\begin{center}
\includegraphics[width=6.75cm,height=6.25cm]{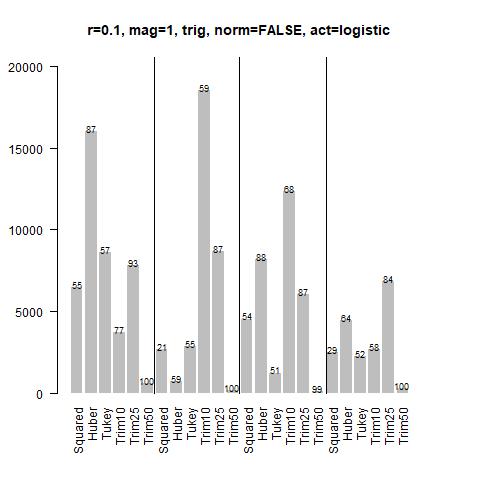}
\includegraphics[width=6.75cm,height=6.25cm]{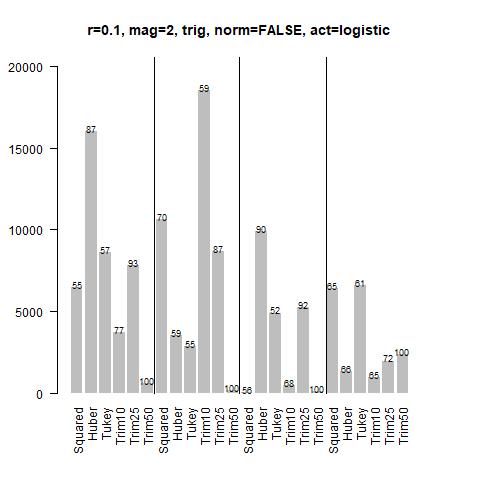} \\
\includegraphics[width=6.75cm,height=6.25cm]{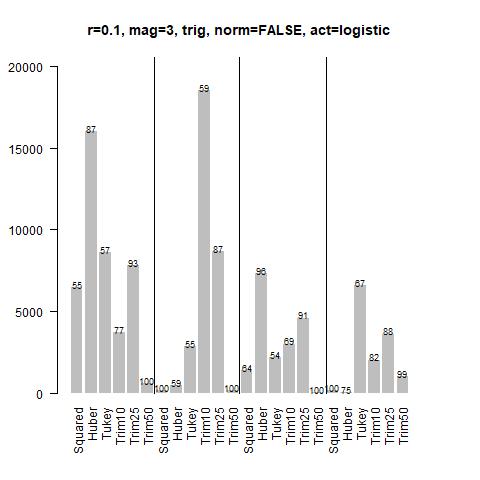} 
\includegraphics[width=6.75cm,height=6.25cm]{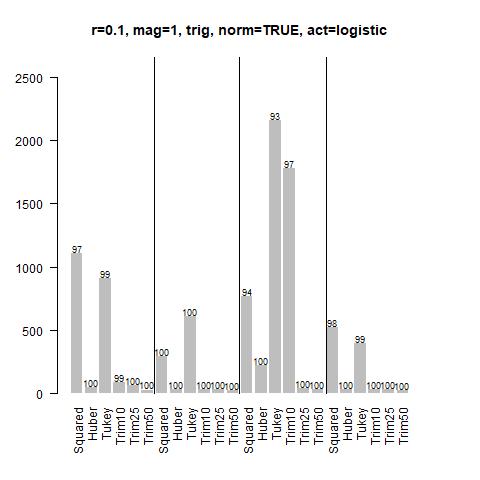}\\
\includegraphics[width=6.75cm,height=6.25cm]{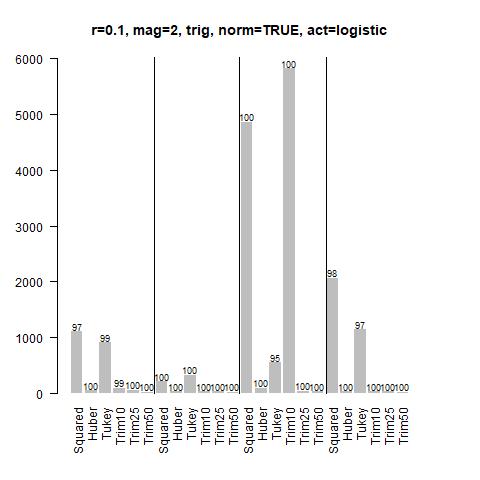} 
\includegraphics[width=6.75cm,height=6.25cm]{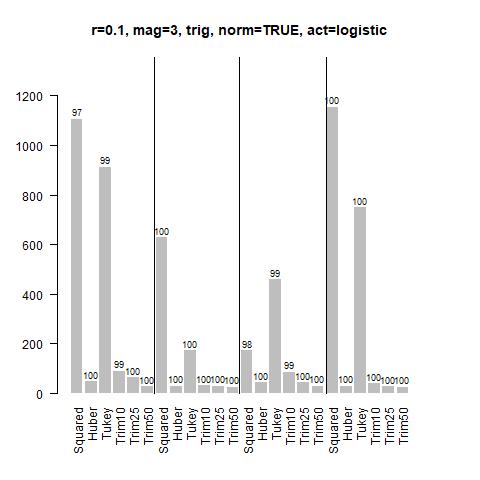} 
\end{center}
\caption{Results for $r=0.1$}
\end{figure}

\begin{figure}[H]
\label{trimnn:n200p5r25m1trignonlogdeepStep}
\begin{center}
\includegraphics[width=6.75cm,height=6.25cm]{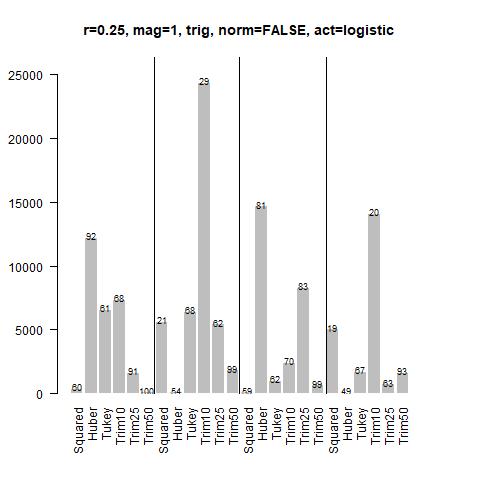}
\includegraphics[width=6.75cm,height=6.25cm]{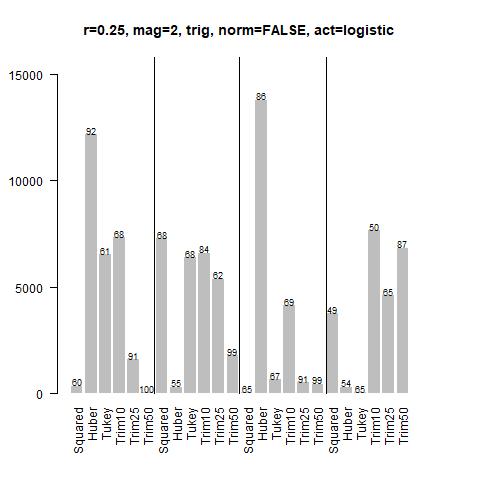} \\
\includegraphics[width=6.75cm,height=6.25cm]{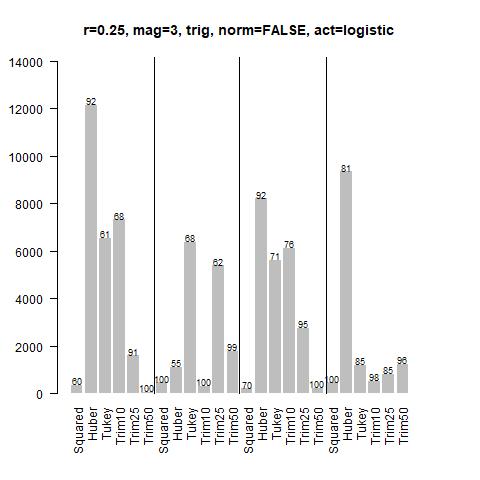} 
\includegraphics[width=6.75cm,height=6.25cm]{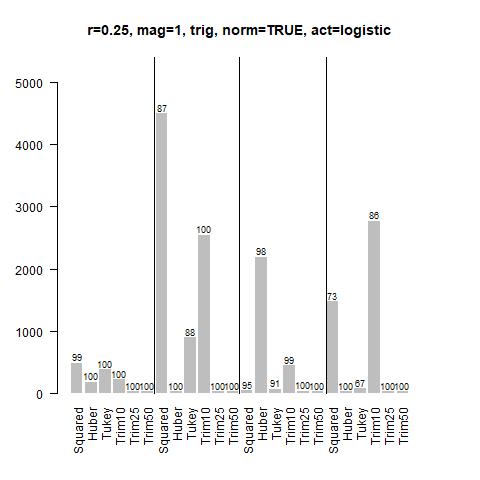}\\
\includegraphics[width=6.75cm,height=6.25cm]{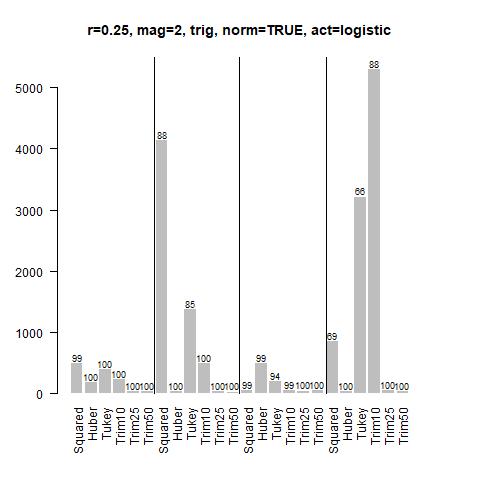} 
\includegraphics[width=6.75cm,height=6.25cm]{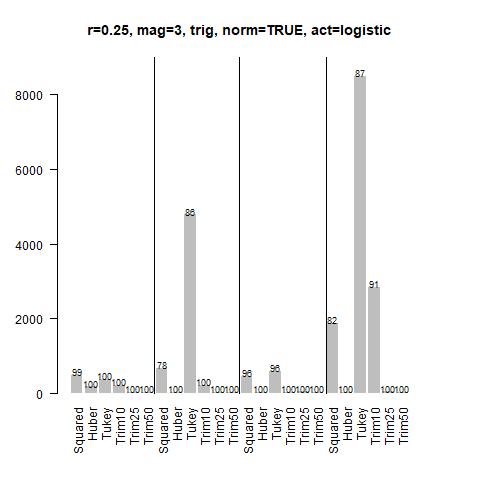} 
\end{center}
\caption{Results for $r=0.25$}
\end{figure}

\begin{figure}[H]
\label{trimnn:n200p5r40m1trignonlogdeepStep}
\begin{center}
\includegraphics[width=6.75cm,height=6.25cm]{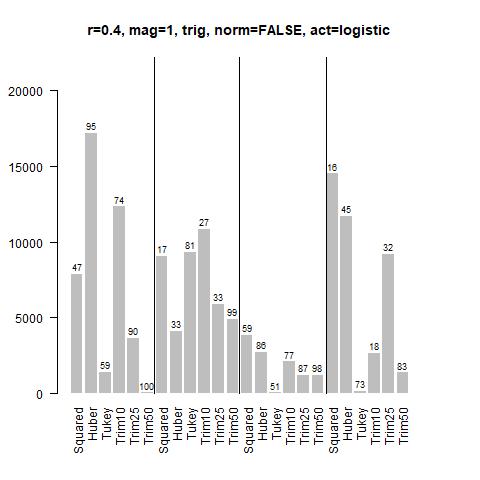}
\includegraphics[width=6.75cm,height=6.25cm]{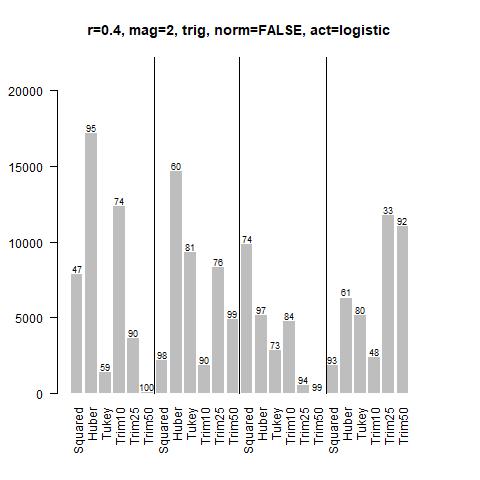} \\
\includegraphics[width=6.75cm,height=6.25cm]{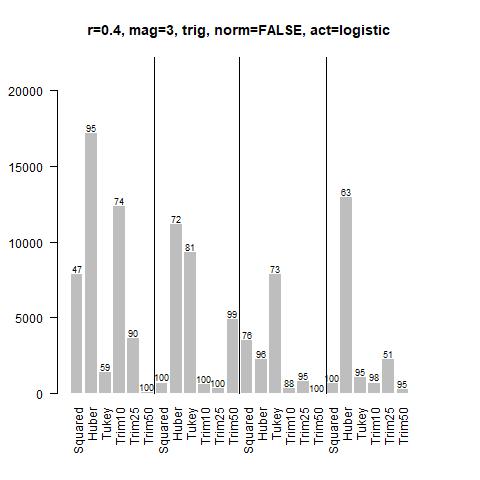} 
\includegraphics[width=6.75cm,height=6.25cm]{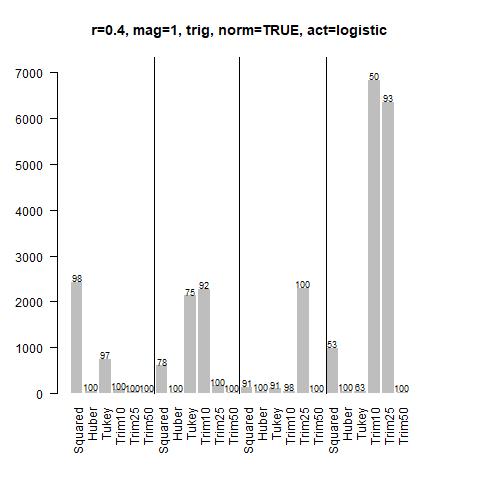}\\
\includegraphics[width=6.75cm,height=6.25cm]{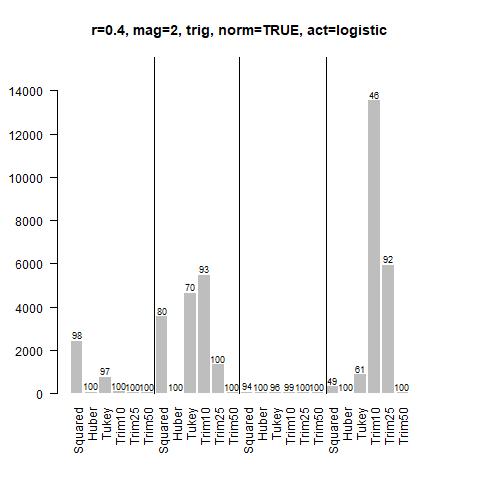} 
\includegraphics[width=6.75cm,height=6.25cm]{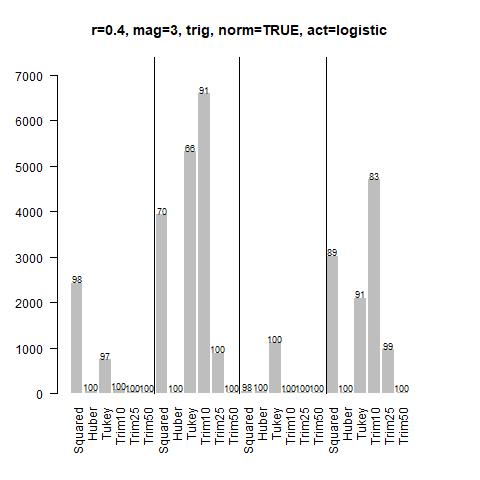} 
\end{center}
\caption{Results for $r=0.4$}
\end{figure}

\subsection{Softplus activation function}

\subsubsection{Linear function}

\begin{figure}[H]
\label{trimnn:n200p5r10m1linnonreludeepStep}
\begin{center}
\includegraphics[width=6.75cm,height=6.25cm]{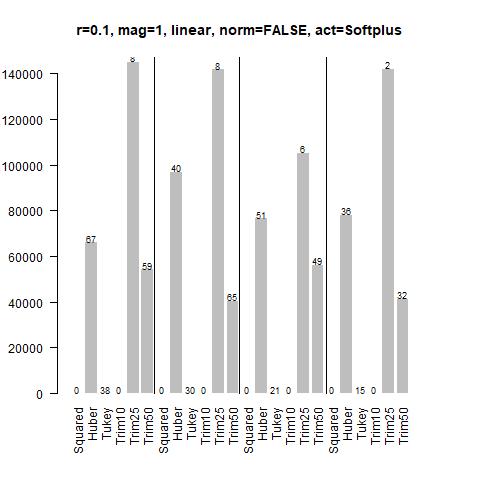}
\includegraphics[width=6.75cm,height=6.25cm]{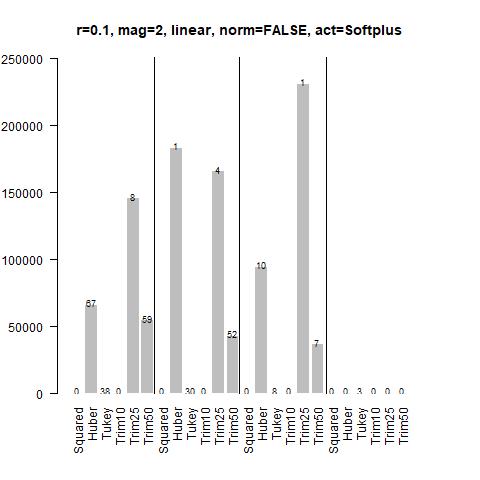} \\
\includegraphics[width=6.75cm,height=6.25cm]{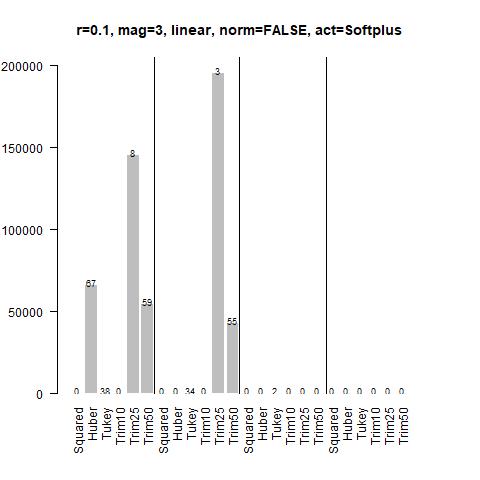} 
\includegraphics[width=6.75cm,height=6.25cm]{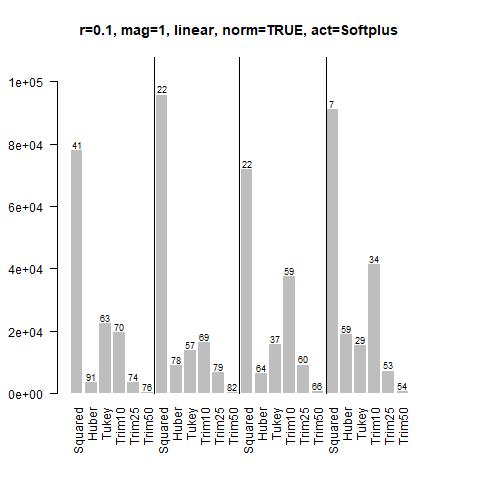}\\
\includegraphics[width=6.75cm,height=6.25cm]{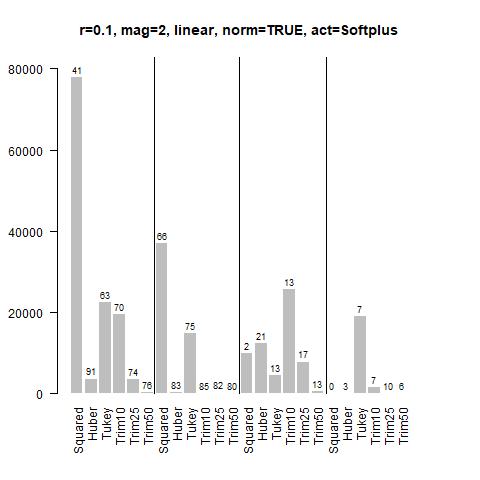} 
\includegraphics[width=6.75cm,height=6.25cm]{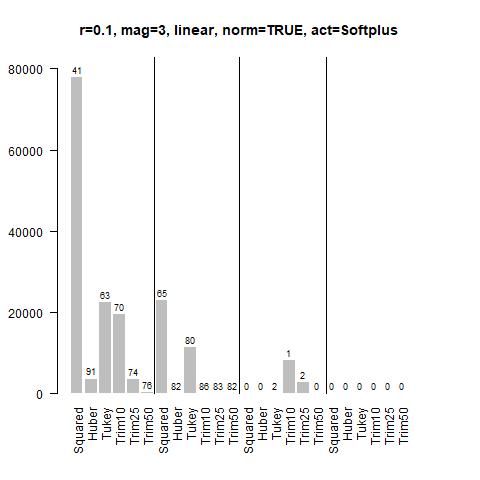} 
\end{center}
\caption{Results for $r=0.1$}
\end{figure}

\begin{figure}[H]
\begin{center}
\includegraphics[width=6.75cm,height=6.25cm]{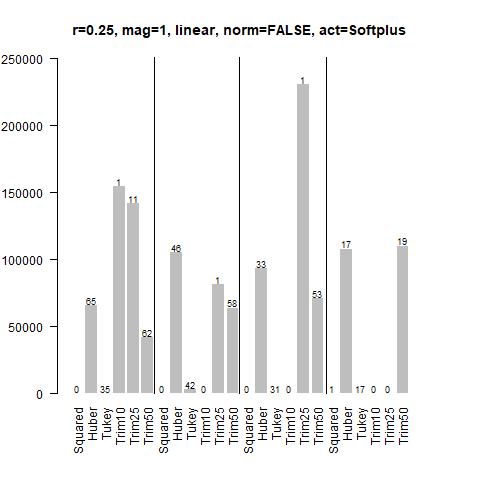}
\includegraphics[width=6.75cm,height=6.25cm]{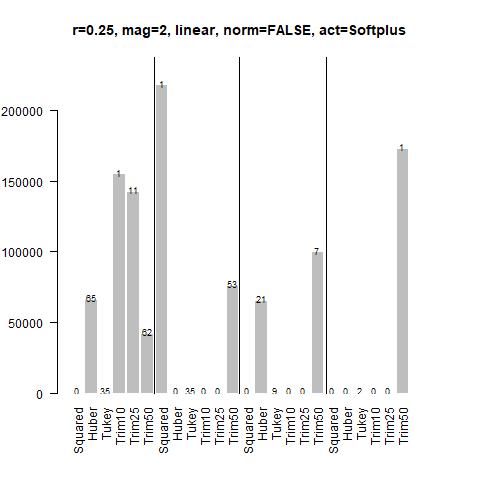} \\
\includegraphics[width=6.75cm,height=6.25cm]{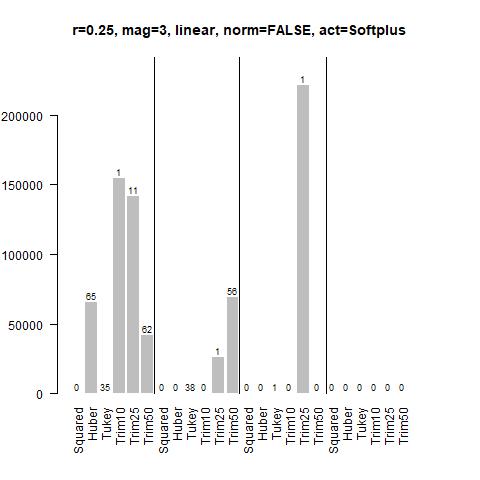} 
\includegraphics[width=6.75cm,height=6.25cm]{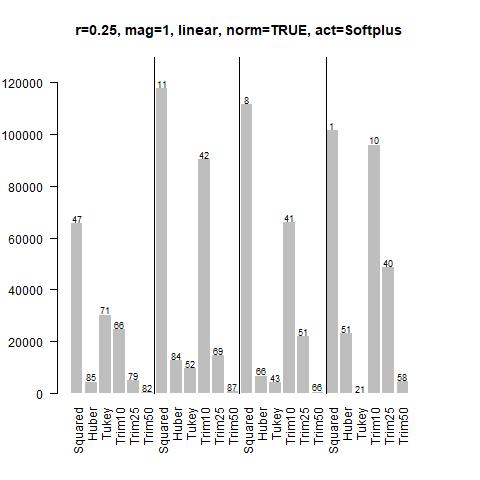}\\
\includegraphics[width=6.75cm,height=6.25cm]{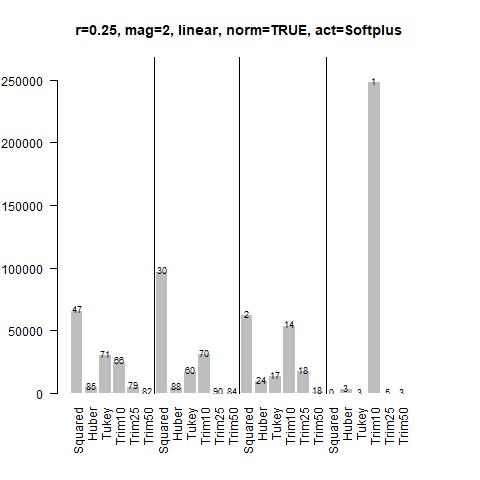} 
\includegraphics[width=6.75cm,height=6.25cm]{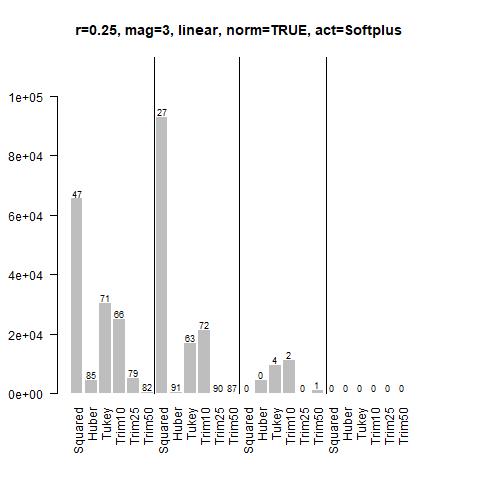} 
\end{center}
\caption{Results for $r=0.25$}\label{trimnn:n200p5r25m1linnonreludeepStep}
\end{figure}

\begin{figure}[H]
\begin{center}
\includegraphics[width=6.75cm,height=6.25cm]{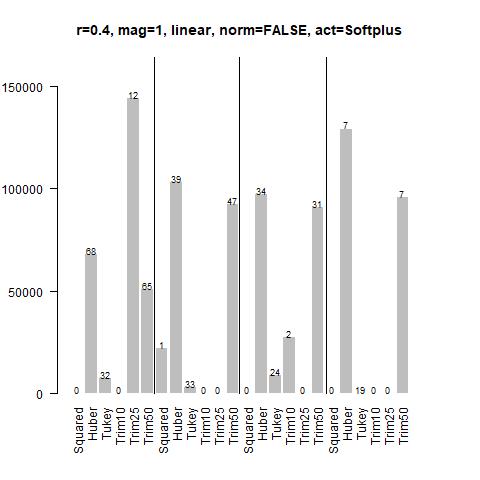}
\includegraphics[width=6.75cm,height=6.25cm]{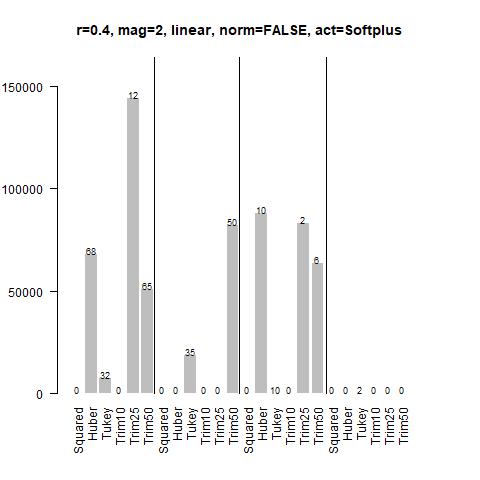} \\
\includegraphics[width=6.75cm,height=6.25cm]{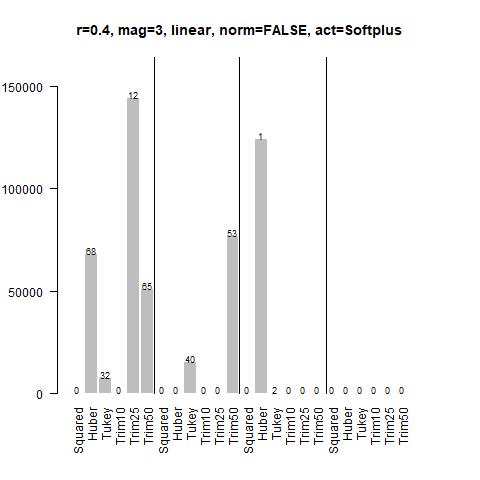} 
\includegraphics[width=6.75cm,height=6.25cm]{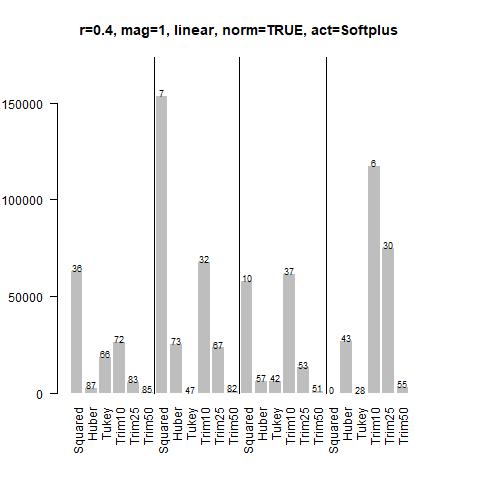}\\
\includegraphics[width=6.75cm,height=6.25cm]{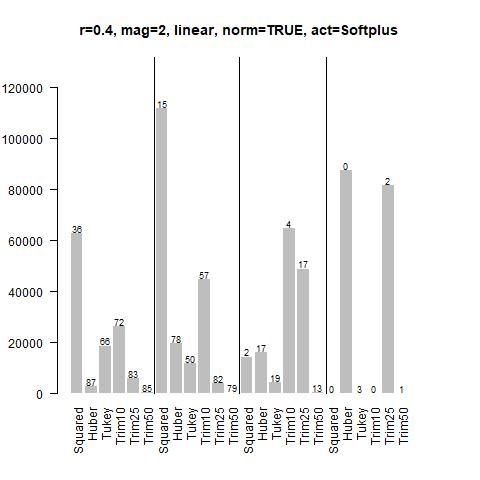} 
\includegraphics[width=6.75cm,height=6.25cm]{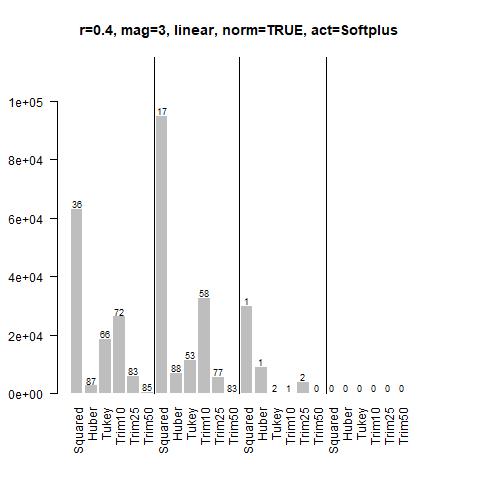} 
\end{center}
\caption{Results for $r=0.4$}\label{trimnn:n200p5r40m1linnonreludeepStep}
\end{figure}

\subsubsection{Polynomial function}

\begin{figure}[H]
\label{trimnn:n200p5r10m1polynonreludeepStep}
\begin{center}
\includegraphics[width=6.75cm,height=6.25cm]{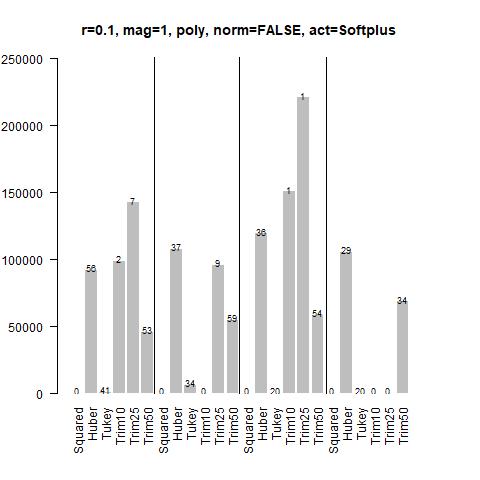}
\includegraphics[width=6.75cm,height=6.25cm]{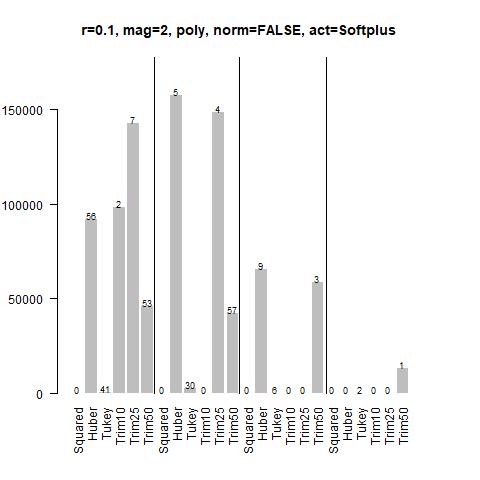} \\
\includegraphics[width=6.75cm,height=6.25cm]{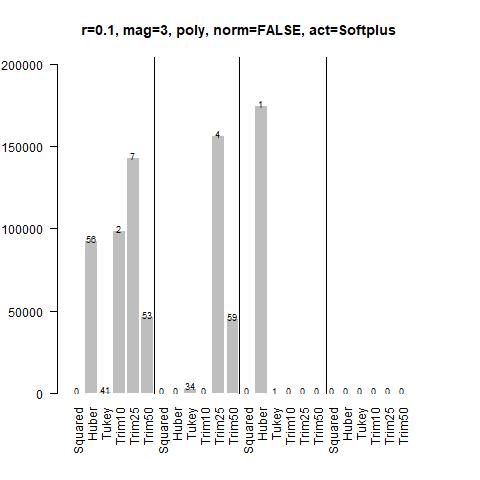} 
\includegraphics[width=6.75cm,height=6.25cm]{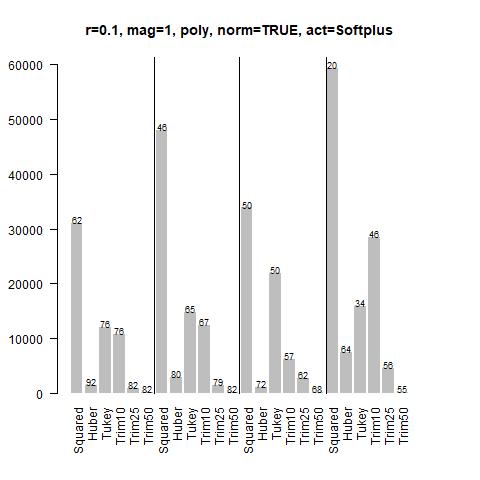}\\
\includegraphics[width=6.75cm,height=6.25cm]{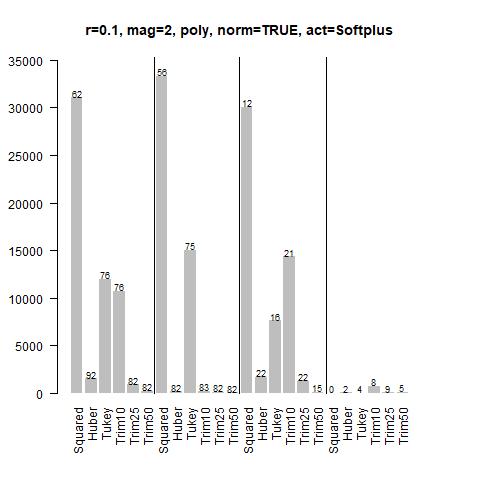} 
\includegraphics[width=6.75cm,height=6.25cm]{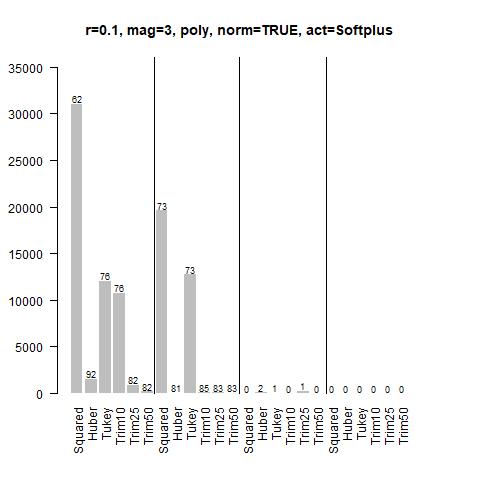} 
\end{center}
\caption{Results for $r=0.1$}
\end{figure}

\begin{figure}[H]
\label{trimnn:n200p5r25m1polynonreludeepStep}
\begin{center}
\includegraphics[width=6.75cm,height=6.25cm]{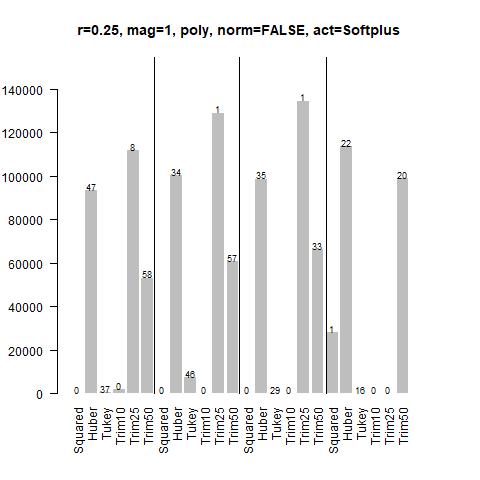}
\includegraphics[width=6.75cm,height=6.25cm]{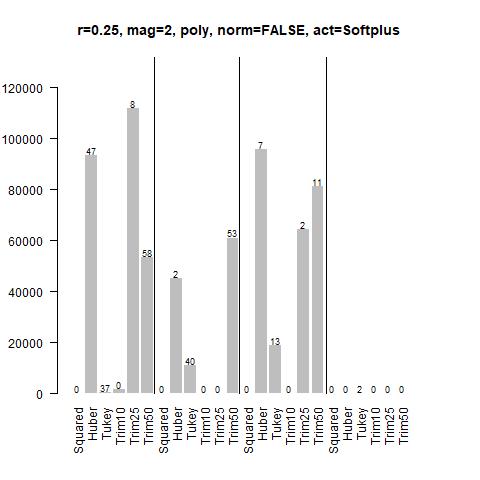} \\
\includegraphics[width=6.75cm,height=6.25cm]{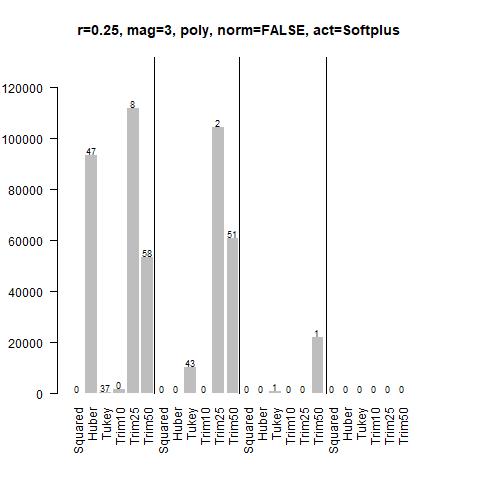} 
\includegraphics[width=6.75cm,height=6.25cm]{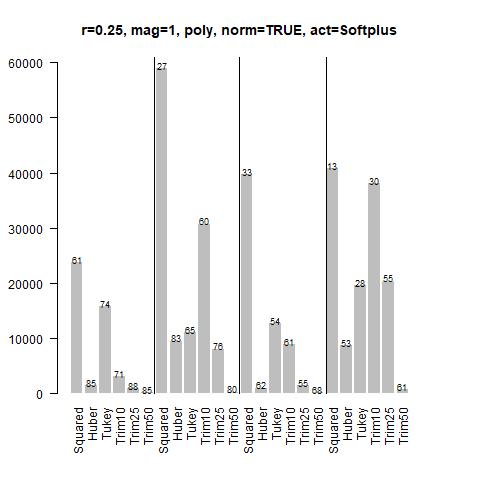}\\
\includegraphics[width=6.75cm,height=6.25cm]{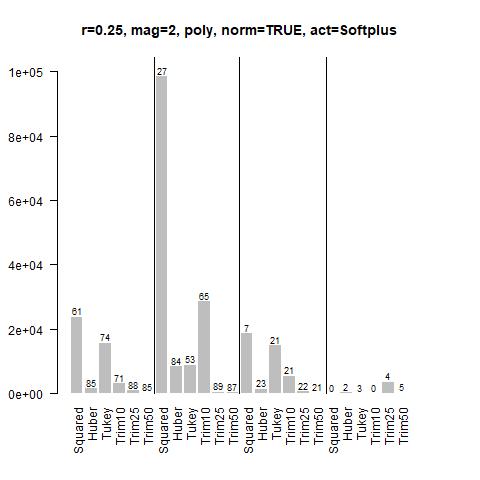} 
\includegraphics[width=6.75cm,height=6.25cm]{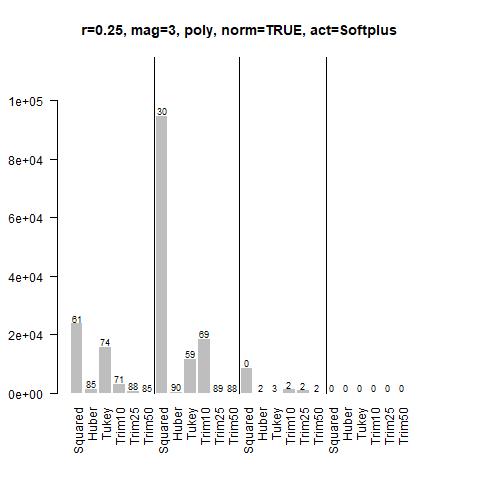} 
\end{center}
\caption{Results for $r=0.25$}
\end{figure}

\begin{figure}[H]
\label{trimnn:n200p5r40m1polynonreludeepStep}
\begin{center}
\includegraphics[width=6.75cm,height=6.25cm]{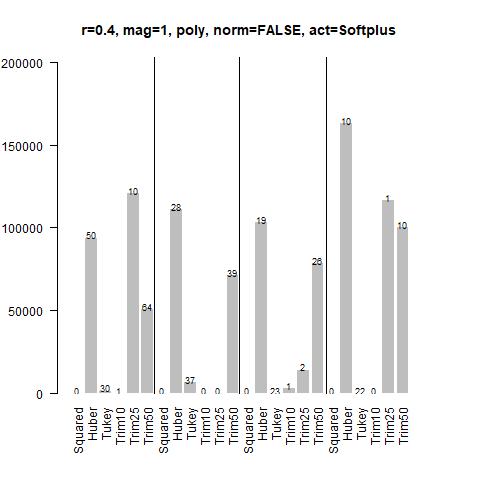}
\includegraphics[width=6.75cm,height=6.25cm]{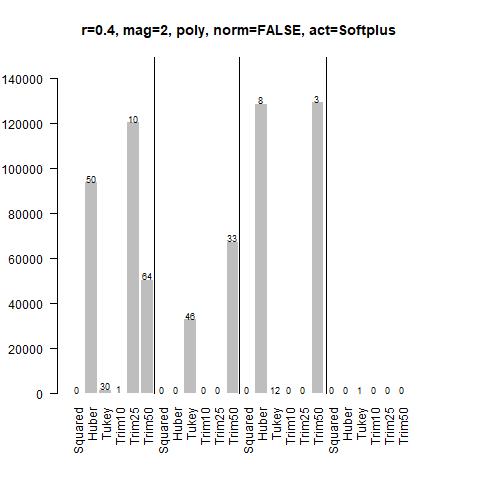} \\
\includegraphics[width=6.75cm,height=6.25cm]{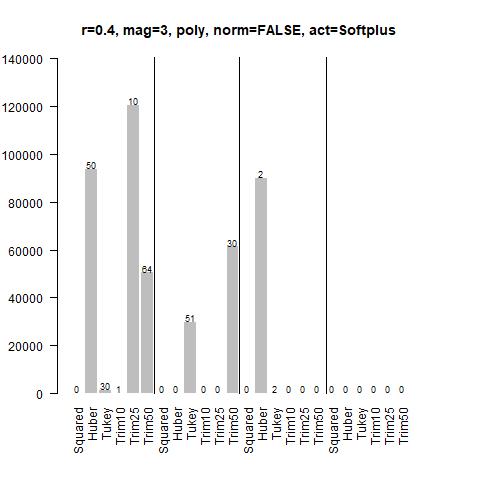} 
\includegraphics[width=6.75cm,height=6.25cm]{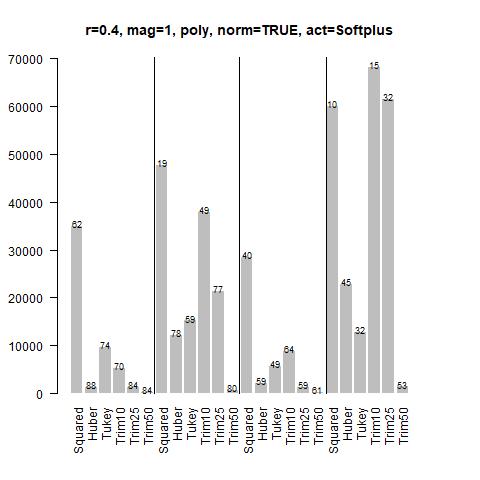}\\
\includegraphics[width=6.75cm,height=6.25cm]{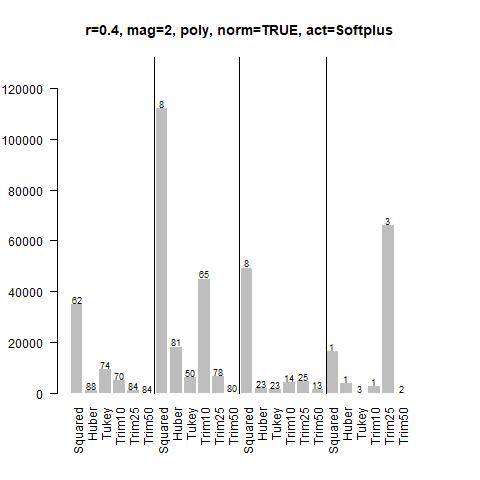} 
\includegraphics[width=6.75cm,height=6.25cm]{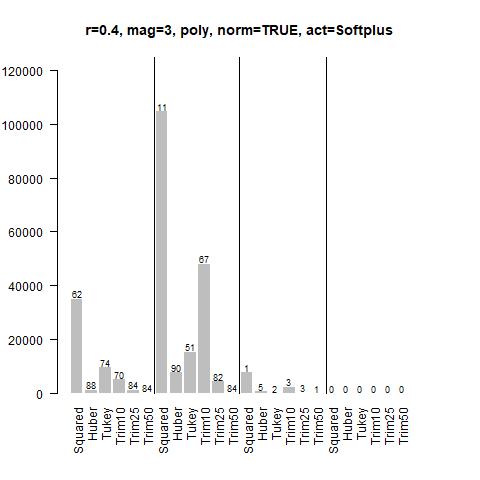} 
\end{center}
\caption{Results for $r=0.4$}
\end{figure}

\subsubsection{Trigonometric function}

\begin{figure}[H]
\label{trimnn:n200p5r10m1trignonreludeepStep}
\begin{center}
\includegraphics[width=6.75cm,height=6.25cm]{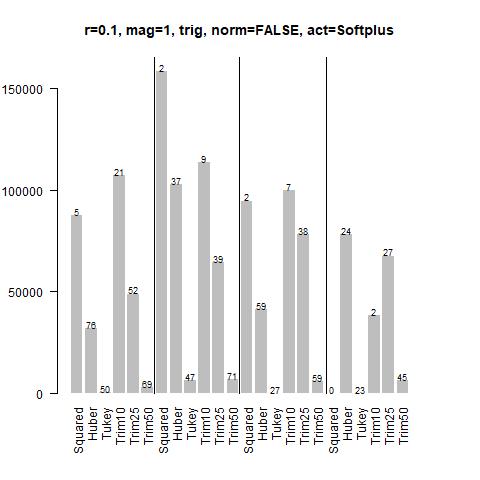}
\includegraphics[width=6.75cm,height=6.25cm]{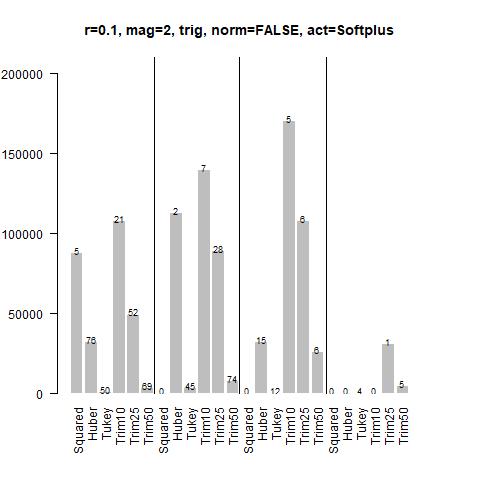} \\
\includegraphics[width=6.75cm,height=6.25cm]{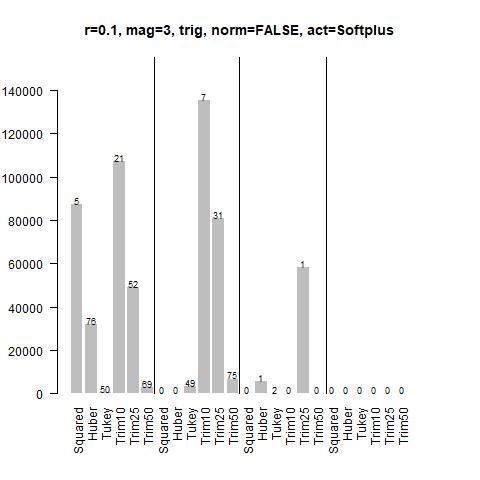} 
\includegraphics[width=6.75cm,height=6.25cm]{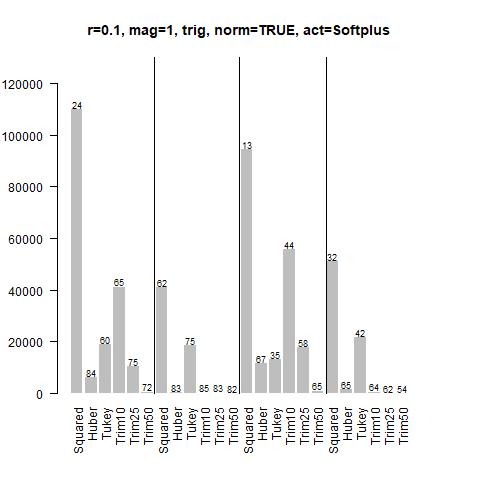}\\
\includegraphics[width=6.75cm,height=6.25cm]{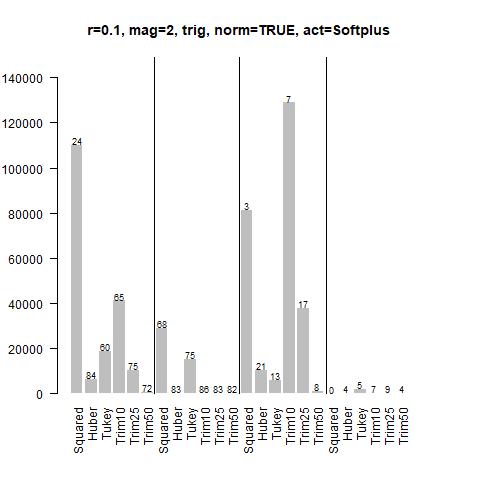} 
\includegraphics[width=6.75cm,height=6.25cm]{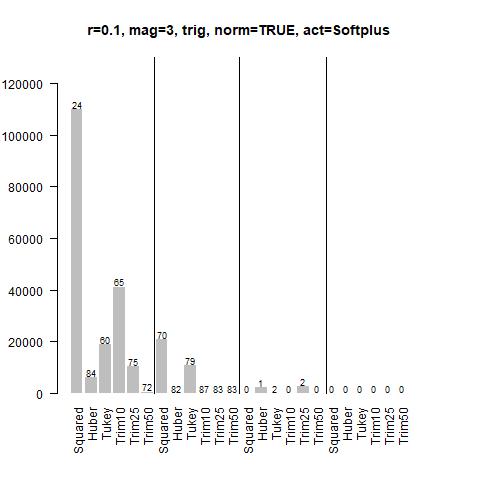} 
\end{center}
\caption{Results for $r=0.1$}
\end{figure}

\begin{figure}[H]
\label{trimnn:n200p5r25m1trignonreludeepStep}
\begin{center}
\includegraphics[width=6.75cm,height=6.25cm]{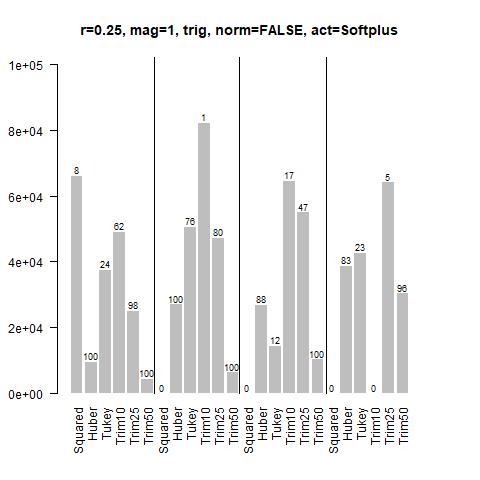}
\includegraphics[width=6.75cm,height=6.25cm]{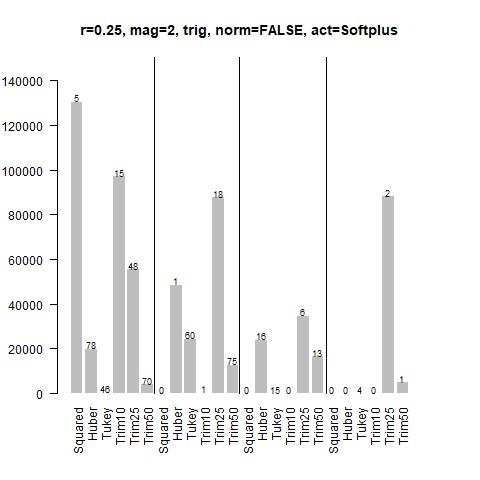} \\
\includegraphics[width=6.75cm,height=6.25cm]{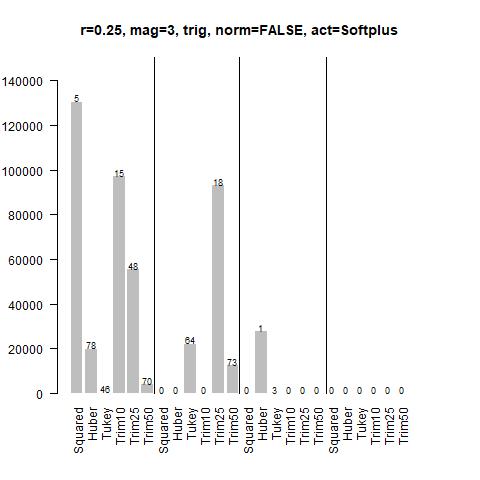} 
\includegraphics[width=6.75cm,height=6.25cm]{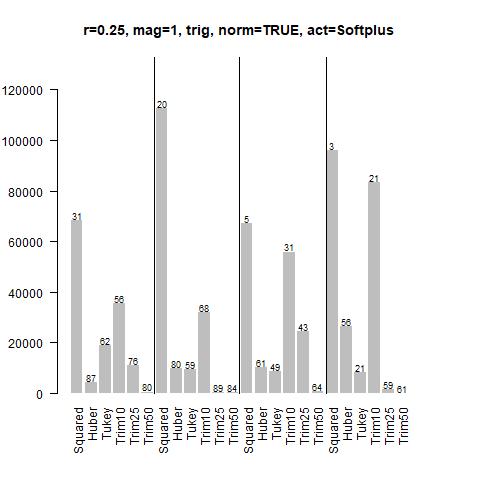}\\
\includegraphics[width=6.75cm,height=6.25cm]{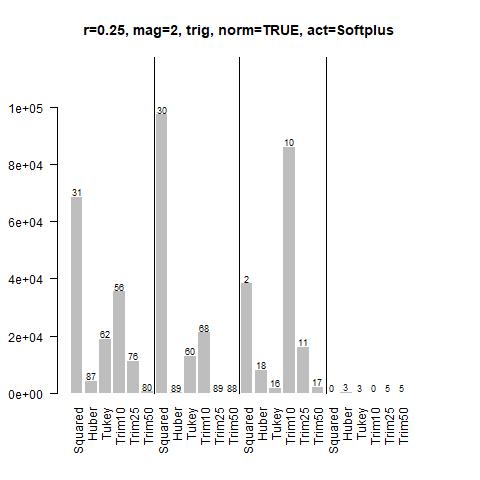} 
\includegraphics[width=6.75cm,height=6.25cm]{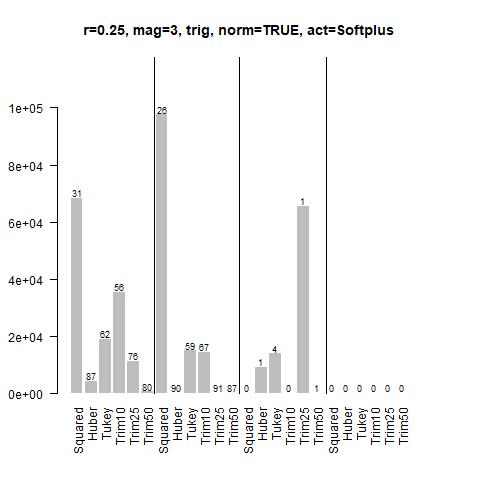} 
\end{center}
\caption{Results for $r=0.25$}
\end{figure}

\begin{figure}[H]
\label{trimnn:n200p5r40m1trignonreludeepStep}
\begin{center}
\includegraphics[width=6.75cm,height=6.25cm]{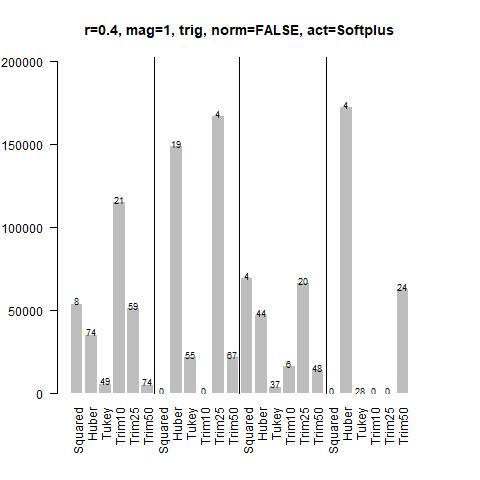}
\includegraphics[width=6.75cm,height=6.25cm]{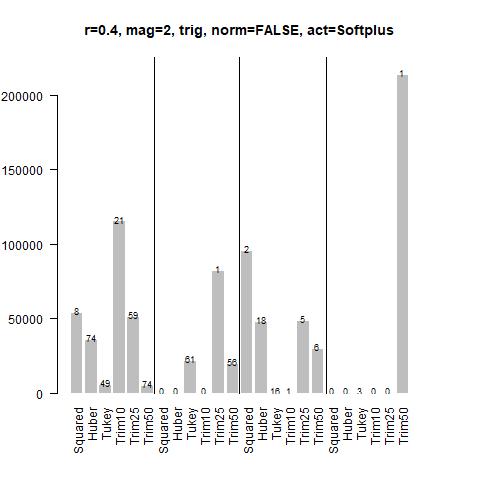} \\
\includegraphics[width=6.75cm,height=6.25cm]{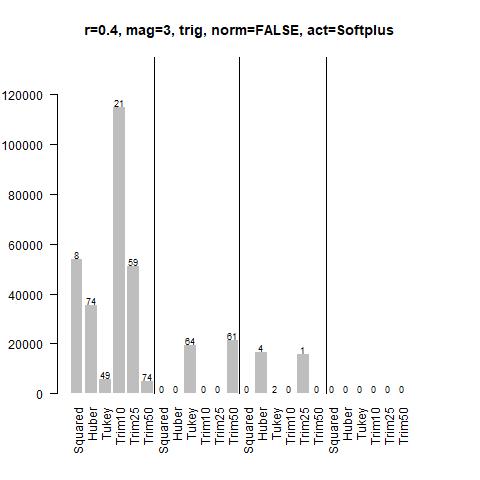} 
\includegraphics[width=6.75cm,height=6.25cm]{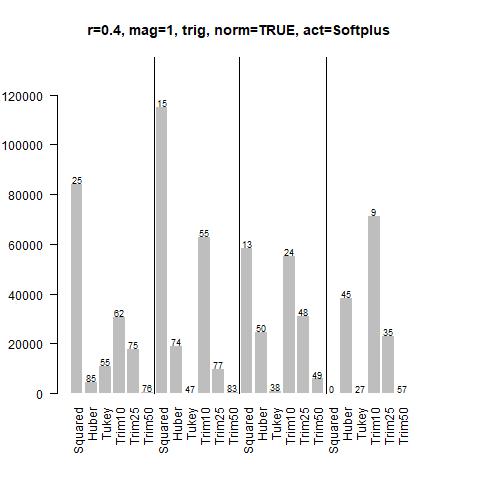}\\
\includegraphics[width=6.75cm,height=6.25cm]{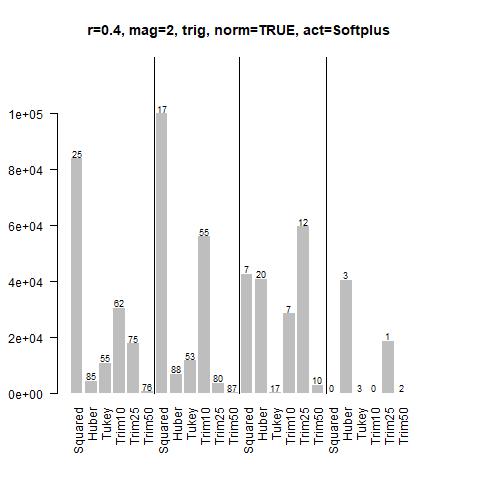} 
\includegraphics[width=6.75cm,height=6.25cm]{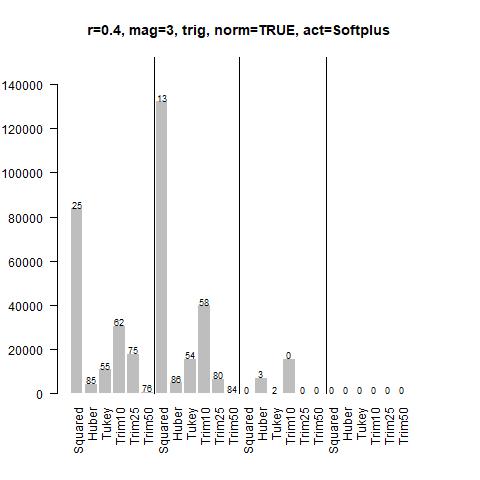} 
\end{center}
\caption{Results for $r=0.4$}
\end{figure}

\section{Simulation results for $n=500$ and $p=20$: Training steps} \label{trimnn:secstep50020}

\subsection{Logistic activation function}

\subsubsection{Linear function}

\begin{figure}[H]
\label{trimnn:n500p20r10m1linnonlogStep}
\begin{center}
\includegraphics[width=6.75cm,height=6.25cm]{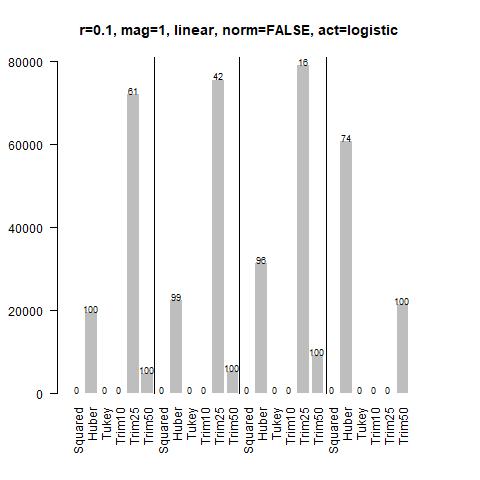}
\includegraphics[width=6.75cm,height=6.25cm]{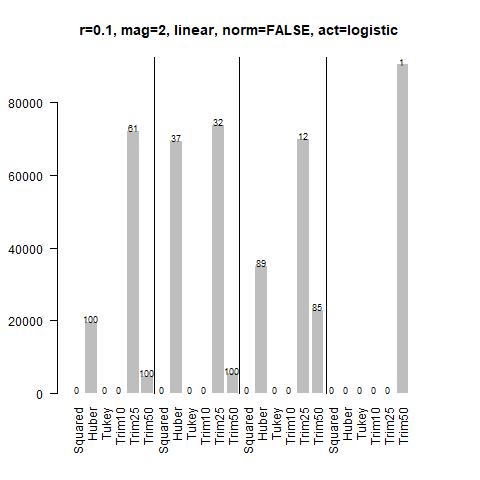} \\
\includegraphics[width=6.75cm,height=6.25cm]{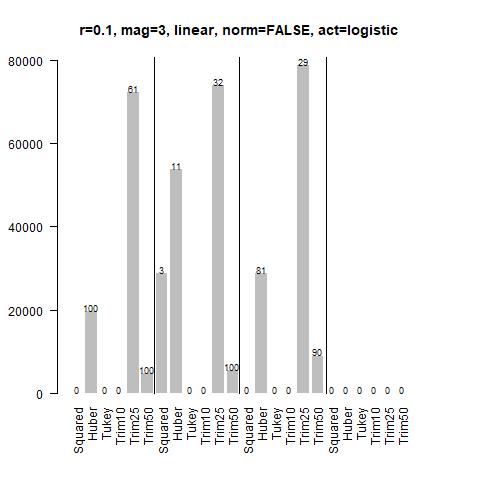} 
\includegraphics[width=6.75cm,height=6.25cm]{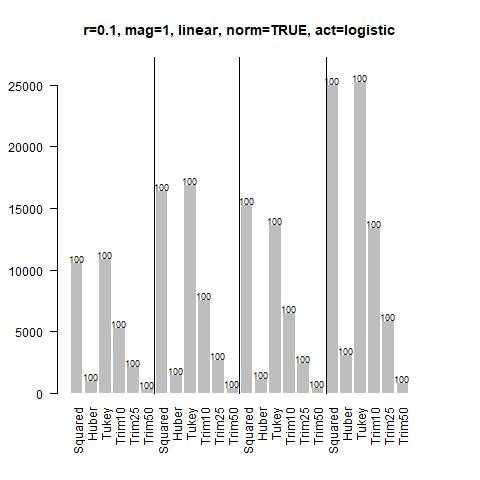}\\
\includegraphics[width=6.75cm,height=6.25cm]{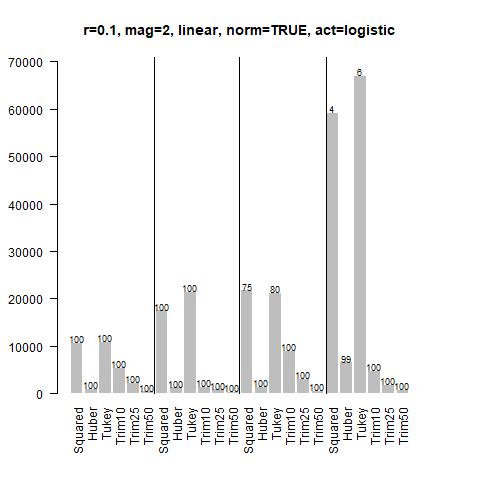} 
\includegraphics[width=6.75cm,height=6.25cm]{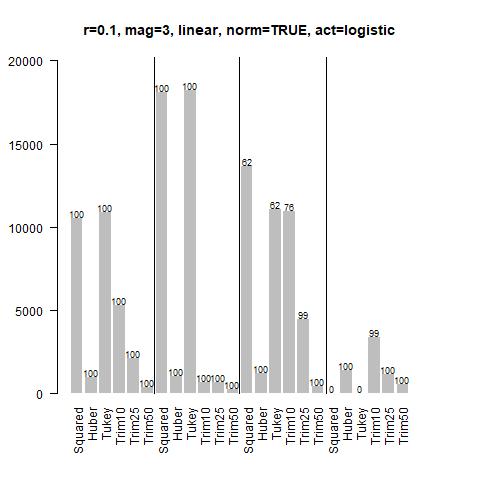} 
\end{center}
\caption{Results for $r=0.1$}
\end{figure}

\begin{figure}[H]
\label{trimnn:n500p20r25m1linnonlogStep}
\begin{center}
\includegraphics[width=6.75cm,height=6.25cm]{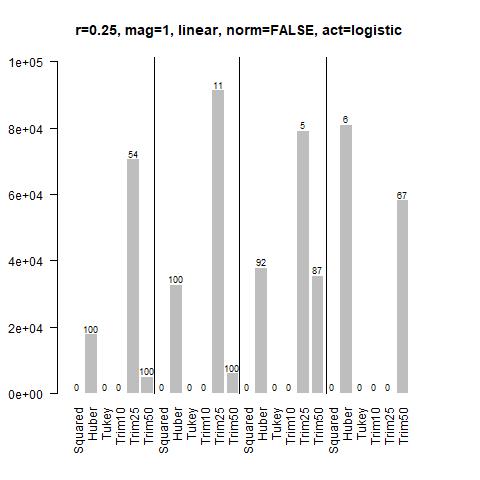}
\includegraphics[width=6.75cm,height=6.25cm]{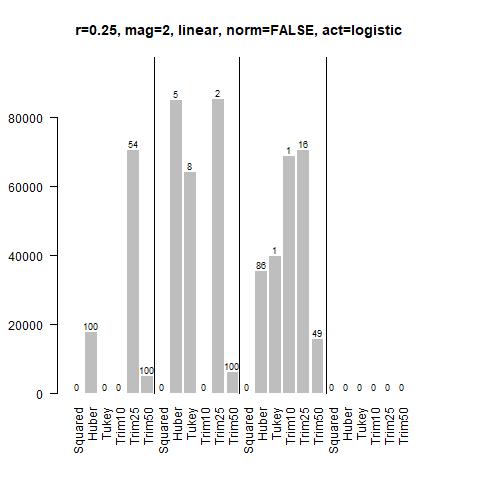} \\
\includegraphics[width=6.75cm,height=6.25cm]{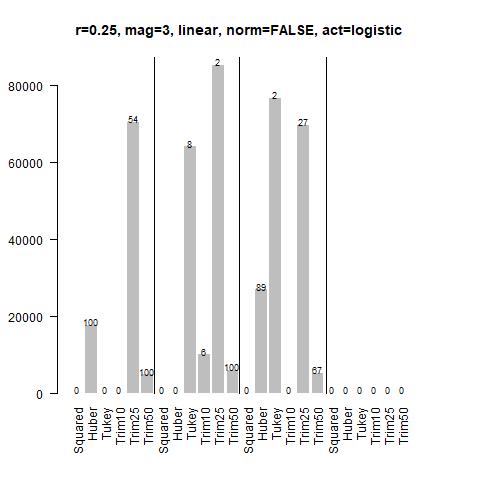} 
\includegraphics[width=6.75cm,height=6.25cm]{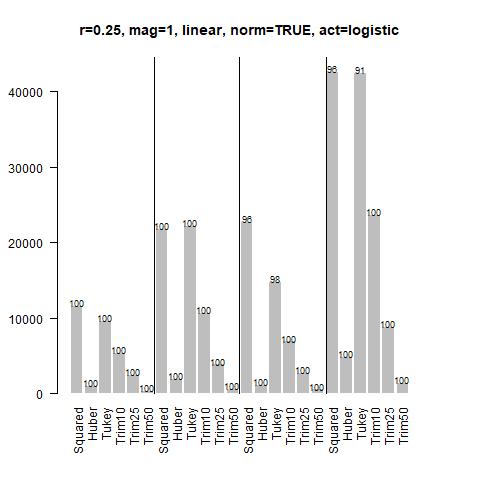}\\
\includegraphics[width=6.75cm,height=6.25cm]{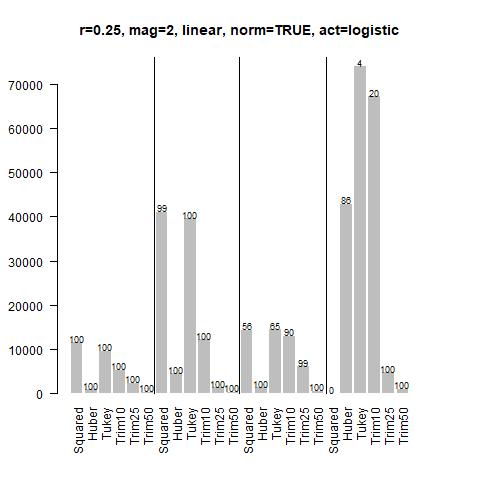} 
\includegraphics[width=6.75cm,height=6.25cm]{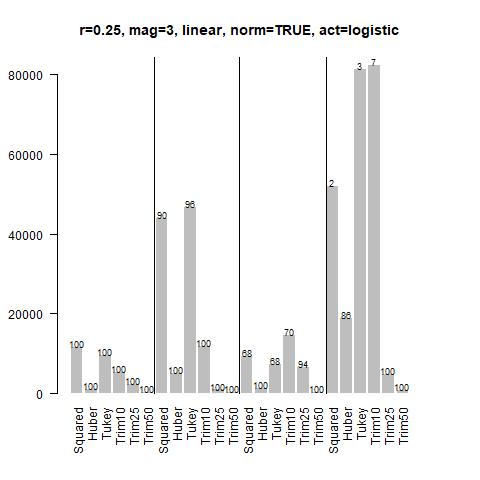} 
\end{center}
\caption{Results for $r=0.25$}
\end{figure}

\begin{figure}[H]
\label{trimnn:n500p20r40m1linnonlogStep}
\begin{center}
\includegraphics[width=6.75cm,height=6.25cm]{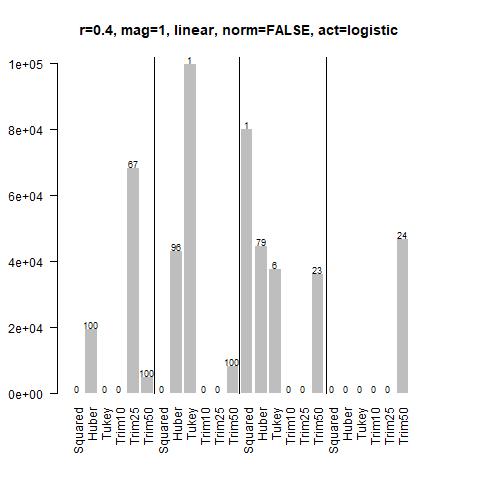}
\includegraphics[width=6.75cm,height=6.25cm]{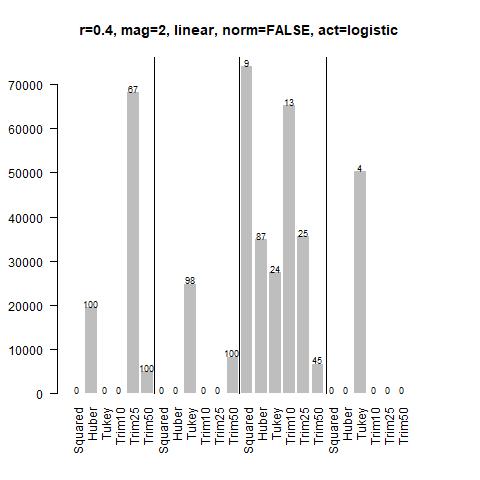} \\
\includegraphics[width=6.75cm,height=6.25cm]{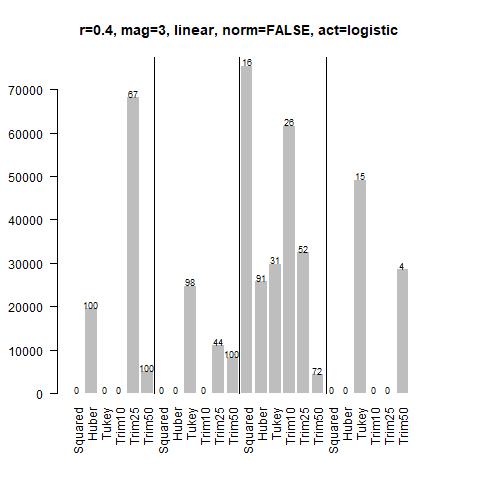} 
\includegraphics[width=6.75cm,height=6.25cm]{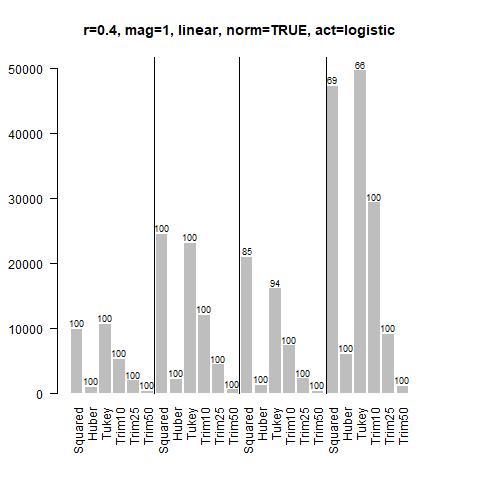}\\
\includegraphics[width=6.75cm,height=6.25cm]{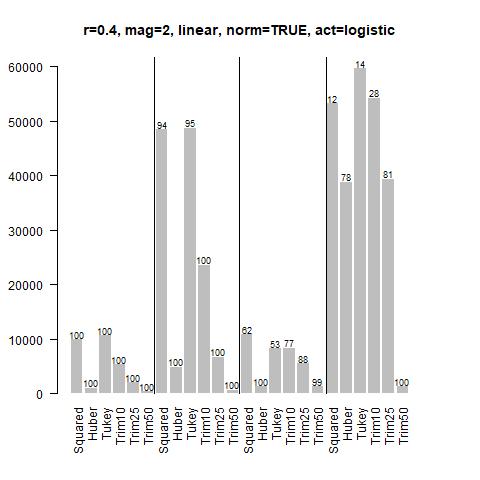} 
\includegraphics[width=6.75cm,height=6.25cm]{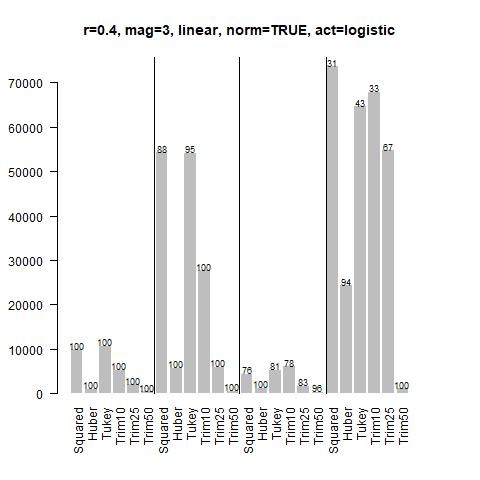} 
\end{center}
\caption{Results for $r=0.4$}
\end{figure}

\subsubsection{Polynomial function}

\begin{figure}[H]
\begin{center}
\includegraphics[width=6.75cm,height=6.25cm]{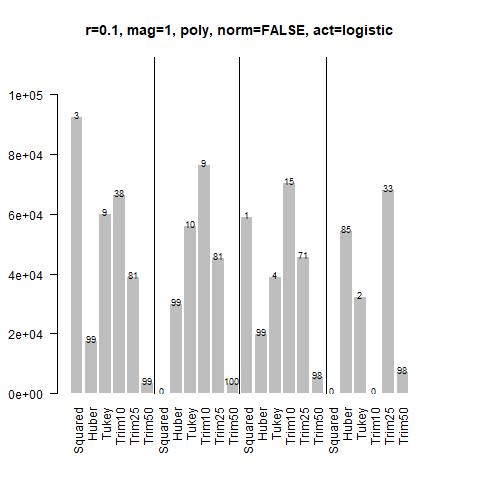}
\includegraphics[width=6.75cm,height=6.25cm]{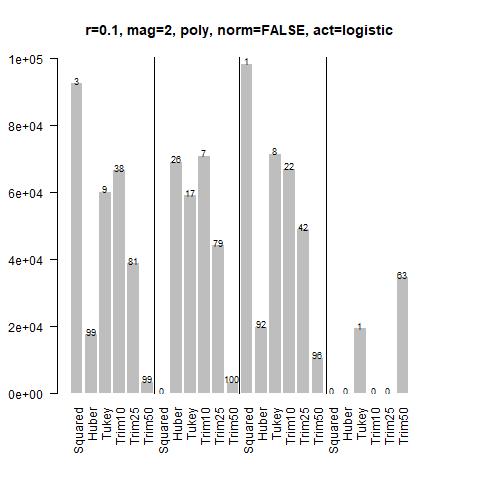} \\
\includegraphics[width=6.75cm,height=6.25cm]{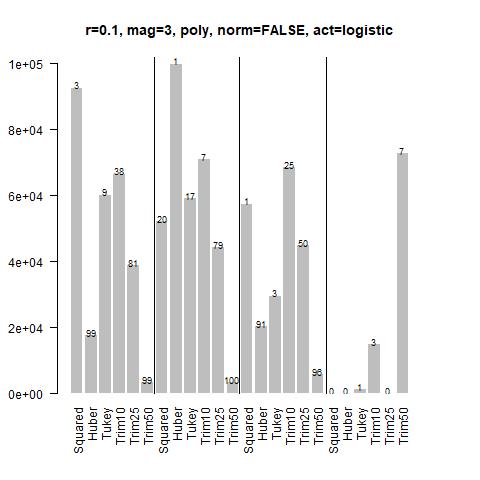} 
\includegraphics[width=6.75cm,height=6.25cm]{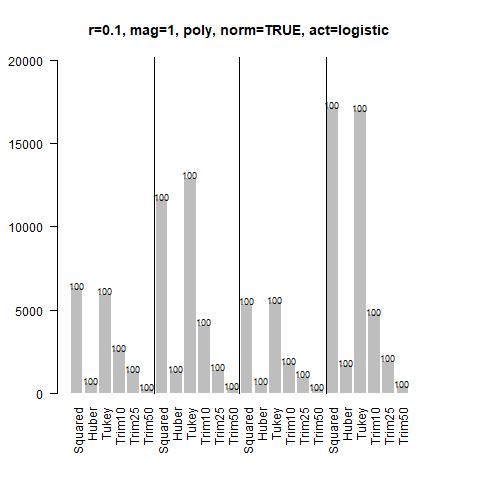}\\
\includegraphics[width=6.75cm,height=6.25cm]{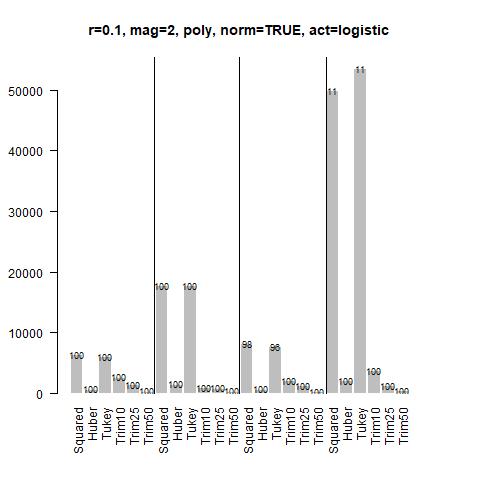} 
\includegraphics[width=6.75cm,height=6.25cm]{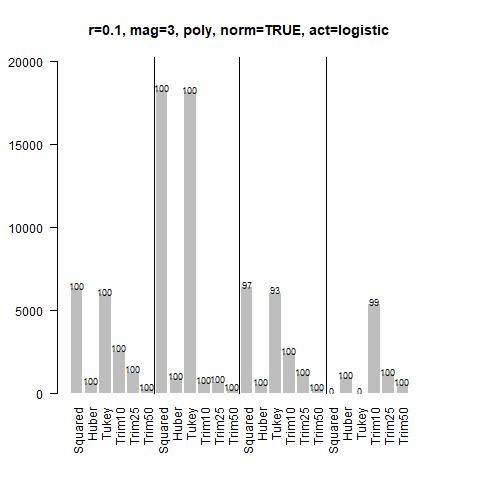} 
\end{center}
\caption{Results for $r=0.1$}\label{trimnn:n500p20r10m1polynonlogStep}
\end{figure}

\begin{figure}[H]
\label{trimnn:n500p20r25m1polynonlogStep}
\begin{center}
\includegraphics[width=6.75cm,height=6.25cm]{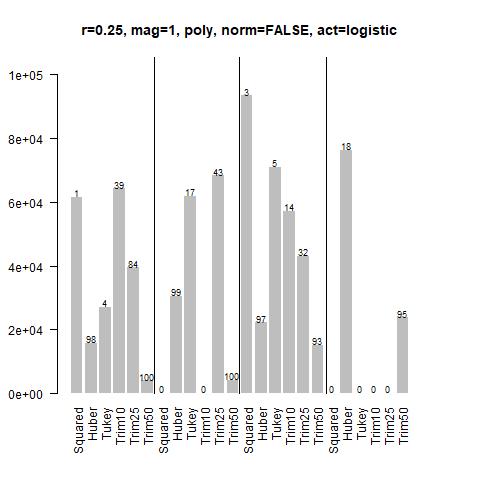}
\includegraphics[width=6.75cm,height=6.25cm]{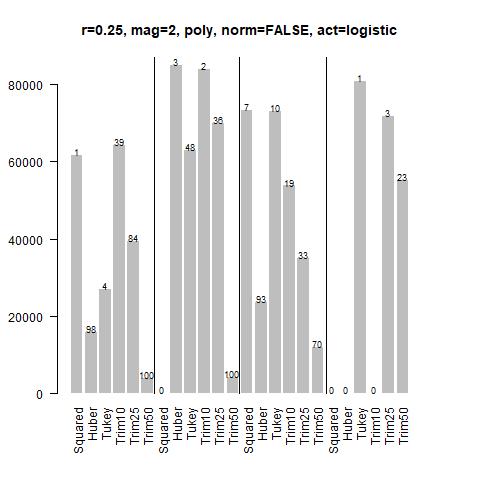} \\
\includegraphics[width=6.75cm,height=6.25cm]{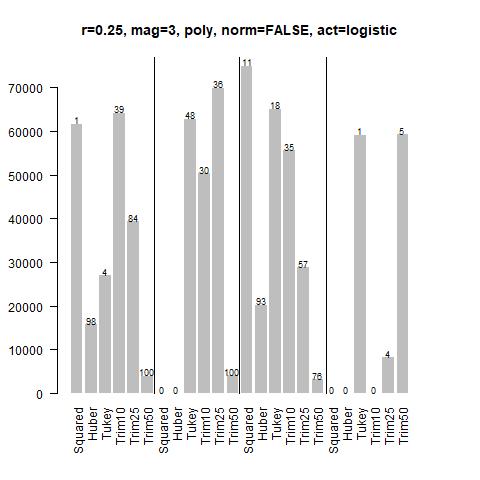} 
\includegraphics[width=6.75cm,height=6.25cm]{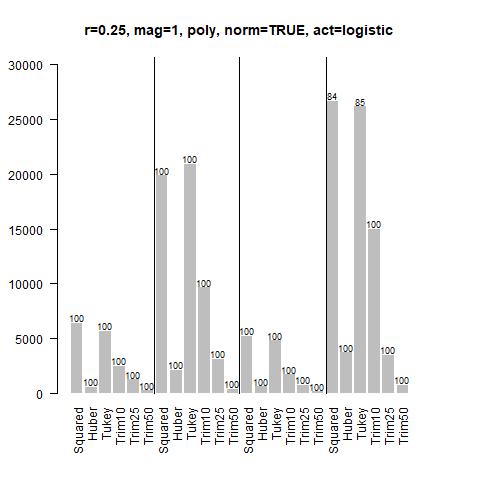}\\
\includegraphics[width=6.75cm,height=6.25cm]{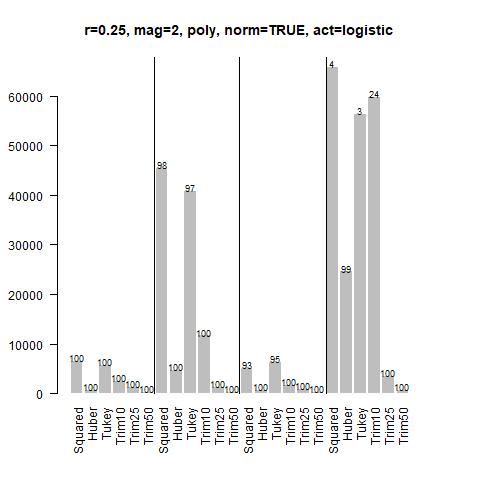} 
\includegraphics[width=6.75cm,height=6.25cm]{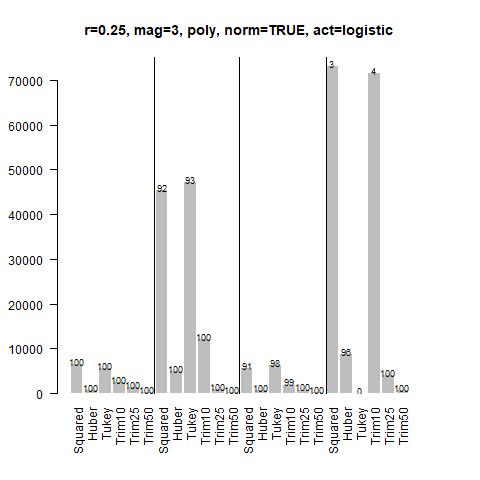} 
\end{center}
\caption{Results for $r=0.25$}
\end{figure}

\begin{figure}[H]
\label{trimnn:n500p20r40m1polynonlogStep}
\begin{center}
\includegraphics[width=6.75cm,height=6.25cm]{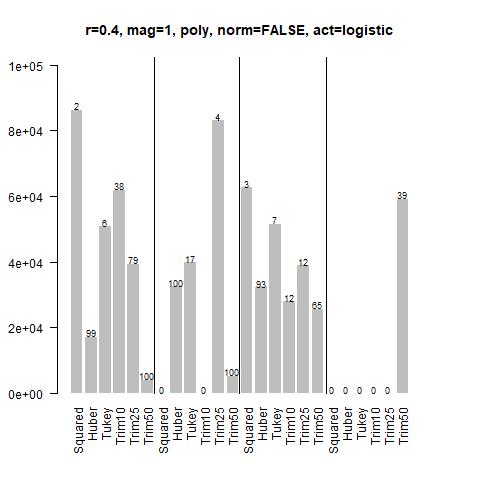}
\includegraphics[width=6.75cm,height=6.25cm]{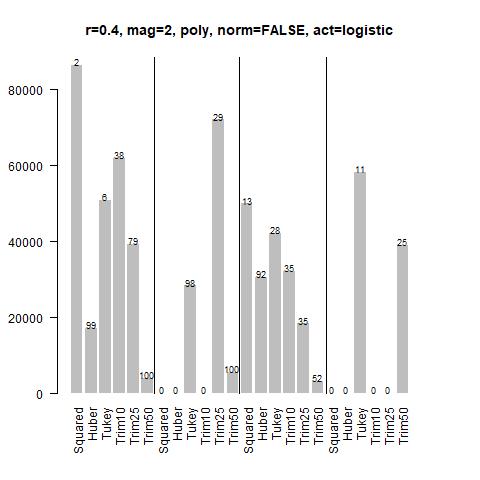} \\
\includegraphics[width=6.75cm,height=6.25cm]{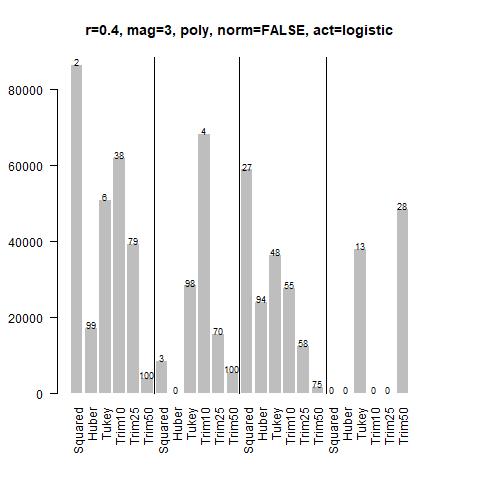} 
\includegraphics[width=6.75cm,height=6.25cm]{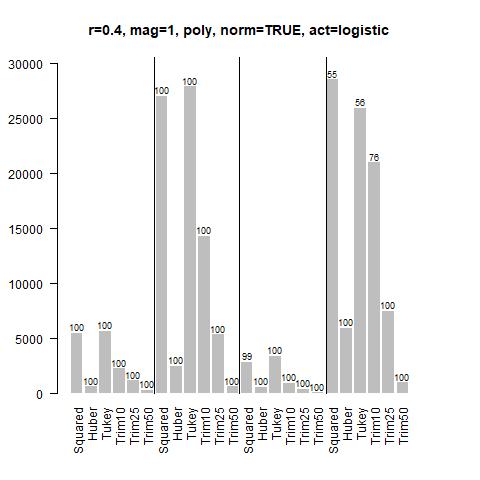}\\
\includegraphics[width=6.75cm,height=6.25cm]{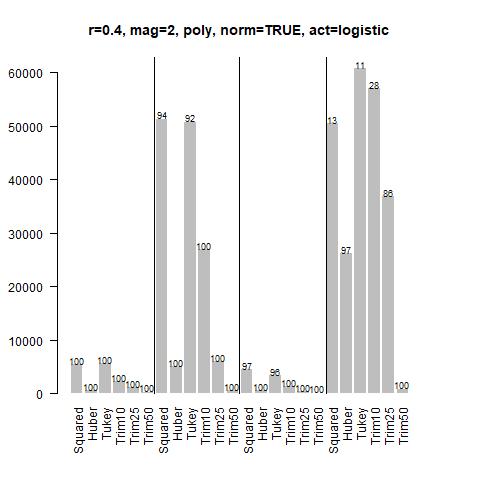} 
\includegraphics[width=6.75cm,height=6.25cm]{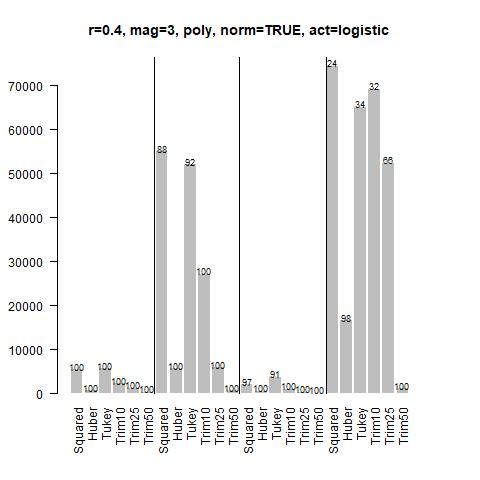} 
\end{center}
\caption{Results for $r=0.4$}
\end{figure}

\subsubsection{Trigonometric function}

\begin{figure}[H]
\begin{center}
\includegraphics[width=6.75cm,height=6.25cm]{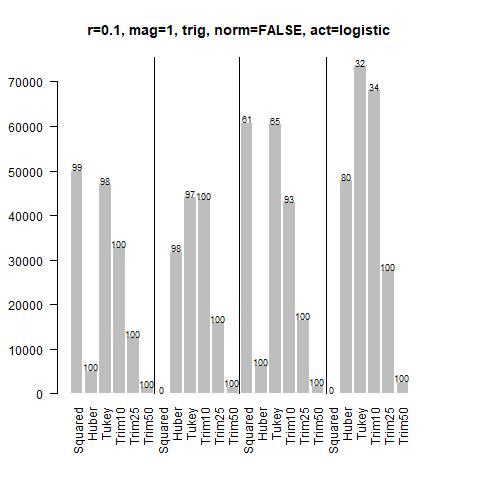}
\includegraphics[width=6.75cm,height=6.25cm]{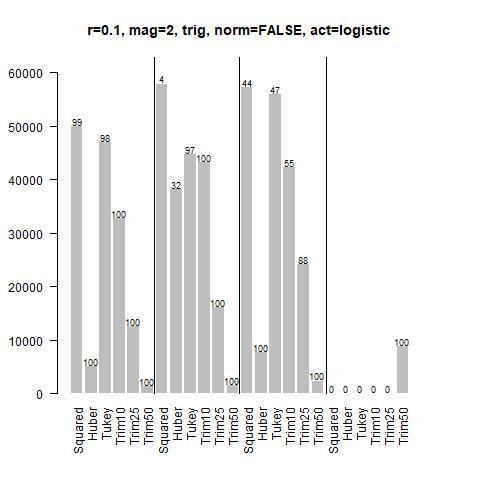} \\
\includegraphics[width=6.75cm,height=6.25cm]{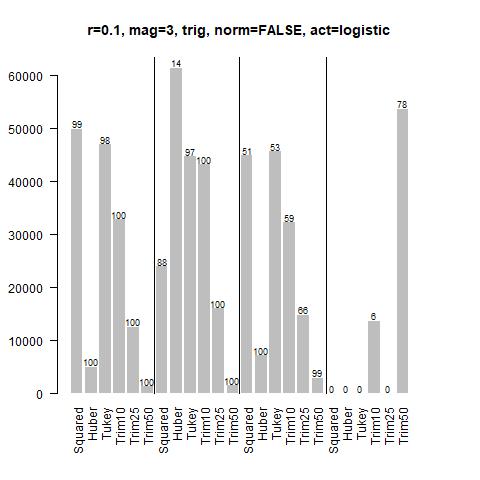} 
\includegraphics[width=6.75cm,height=6.25cm]{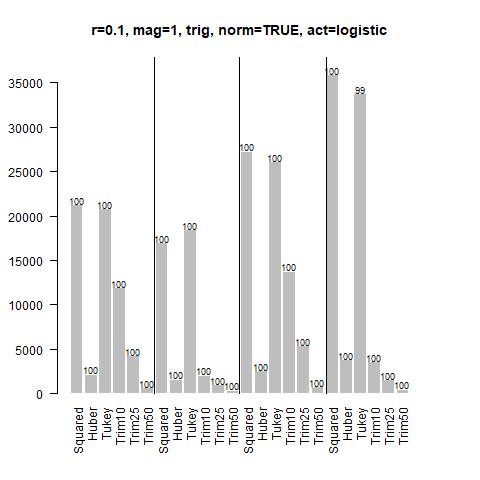}\\
\includegraphics[width=6.75cm,height=6.25cm]{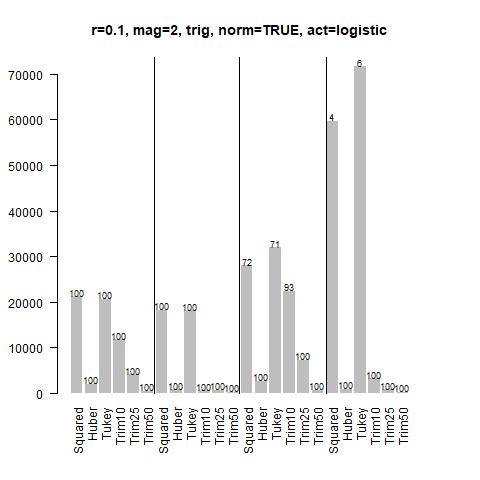} 
\includegraphics[width=6.75cm,height=6.25cm]{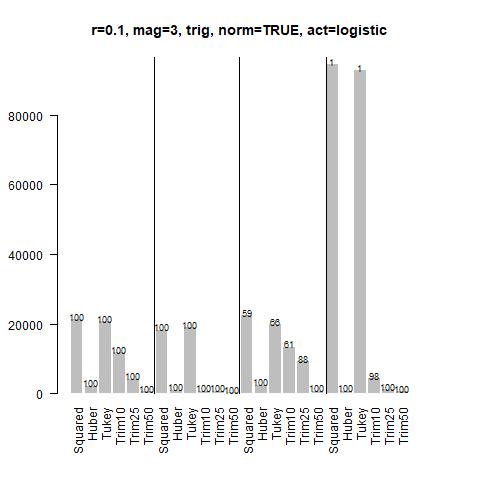} 
\end{center}
\caption{Results for $r=0.1$}\label{trimnn:n500p20r10m1trignonlogStep}
\end{figure}

\begin{figure}[H]
\begin{center}
\includegraphics[width=6.75cm,height=6.25cm]{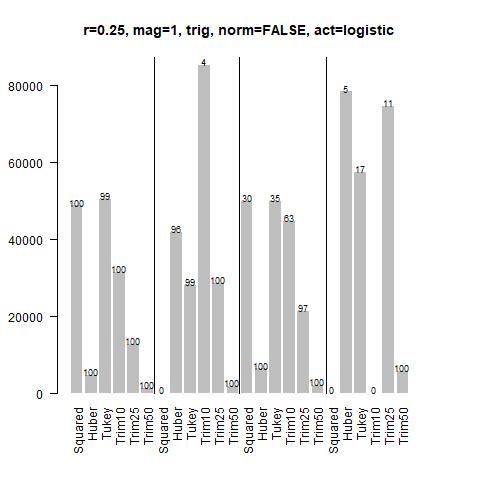}
\includegraphics[width=6.75cm,height=6.25cm]{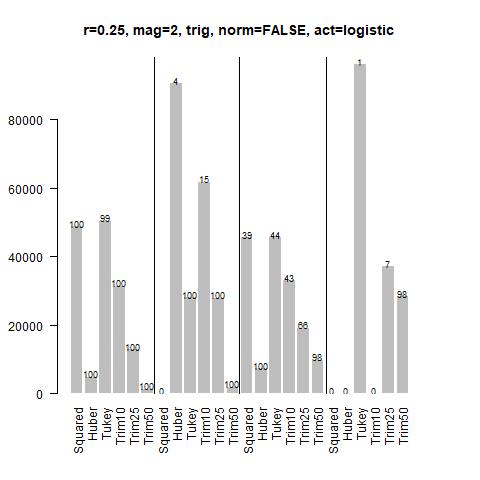} \\
\includegraphics[width=6.75cm,height=6.25cm]{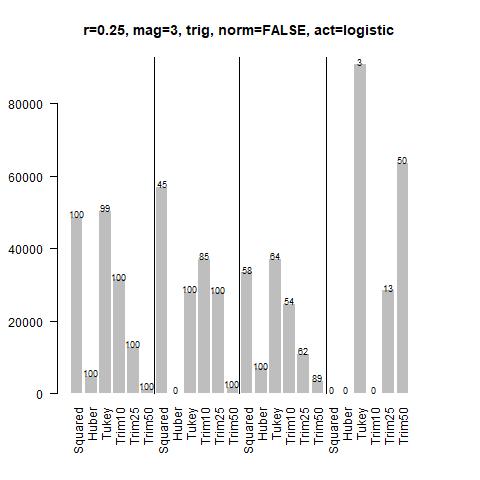} 
\includegraphics[width=6.75cm,height=6.25cm]{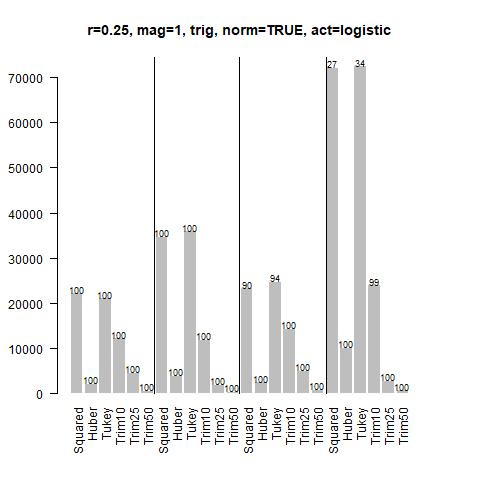}\\
\includegraphics[width=6.75cm,height=6.25cm]{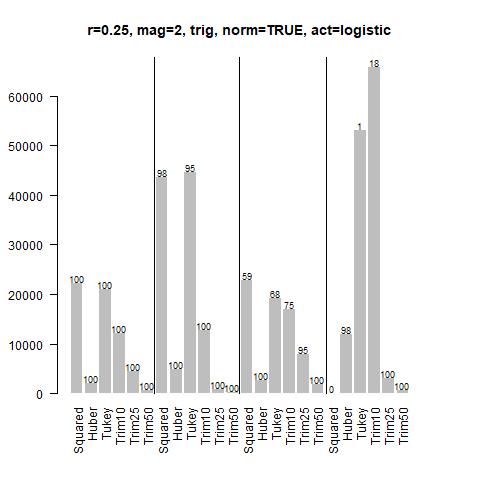} 
\includegraphics[width=6.75cm,height=6.25cm]{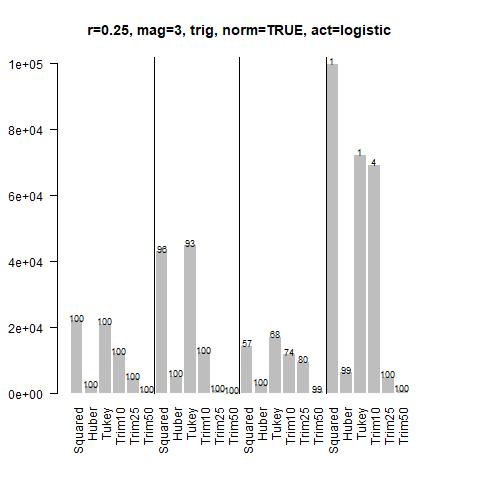} 
\end{center}
\caption{Results for $r=0.25$}\label{trimnn:n500p20r25m1trignonlogStep}
\end{figure}

\begin{figure}[H]
\begin{center}
\includegraphics[width=6.75cm,height=6.25cm]{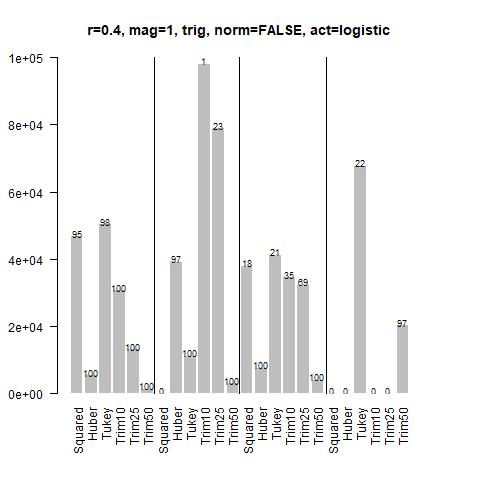}
\includegraphics[width=6.75cm,height=6.25cm]{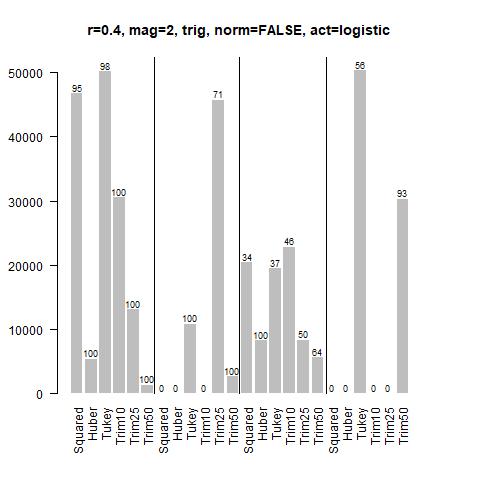} \\
\includegraphics[width=6.75cm,height=6.25cm]{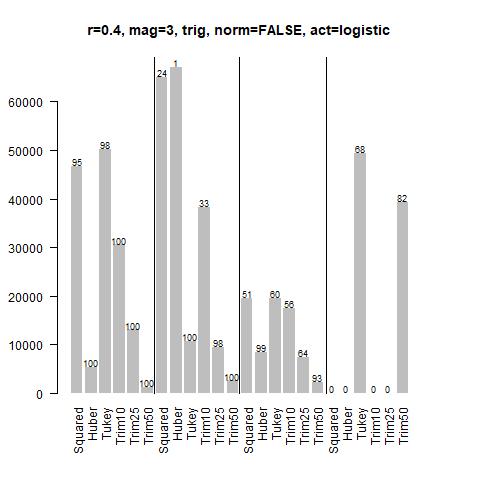} 
\includegraphics[width=6.75cm,height=6.25cm]{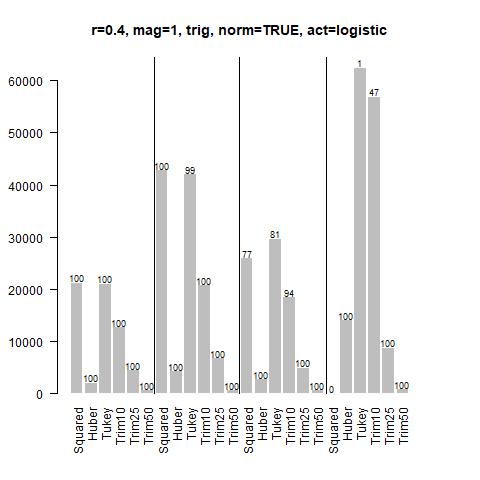}\\
\includegraphics[width=6.75cm,height=6.25cm]{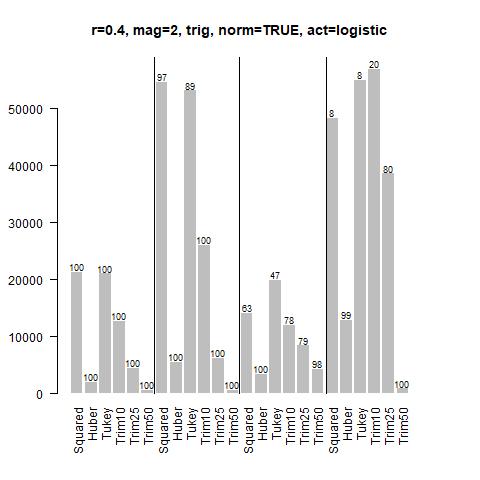} 
\includegraphics[width=6.75cm,height=6.25cm]{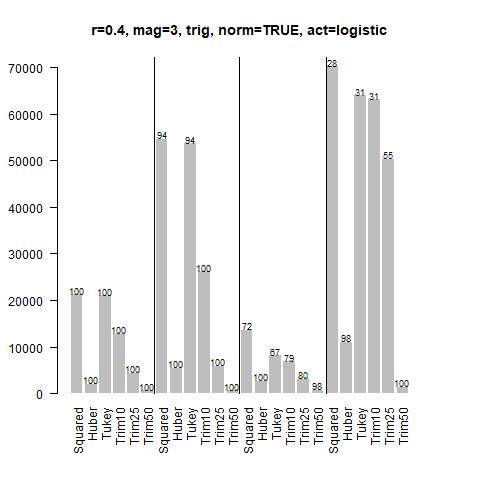} 
\end{center}
\caption{Results for $r=0.4$}\label{trimnn:n500p20r40m1trignonlogStep}
\end{figure}

\subsection{Softplus activation function}

\subsubsection{Linear function}

\begin{figure}[H]
\label{trimnn:n500p20r10m1linnonreluStep}
\begin{center}
\includegraphics[width=6.75cm,height=6.25cm]{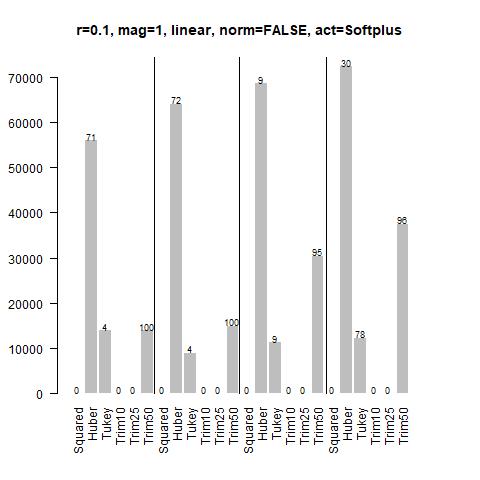}
\includegraphics[width=6.75cm,height=6.25cm]{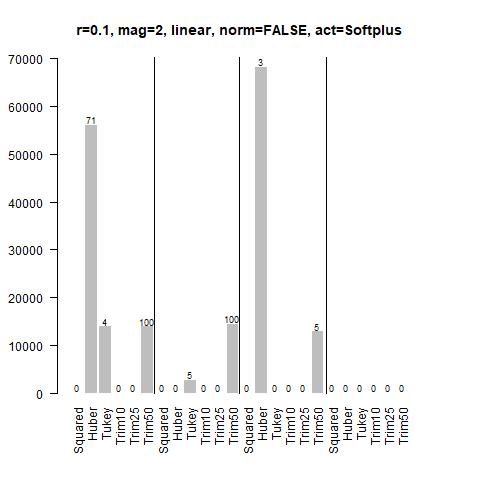} \\
\includegraphics[width=6.75cm,height=6.25cm]{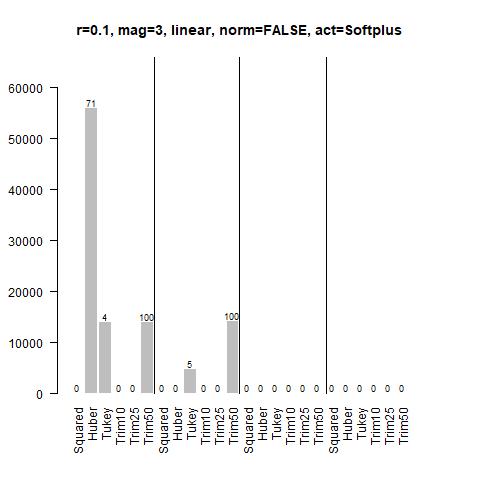} 
\includegraphics[width=6.75cm,height=6.25cm]{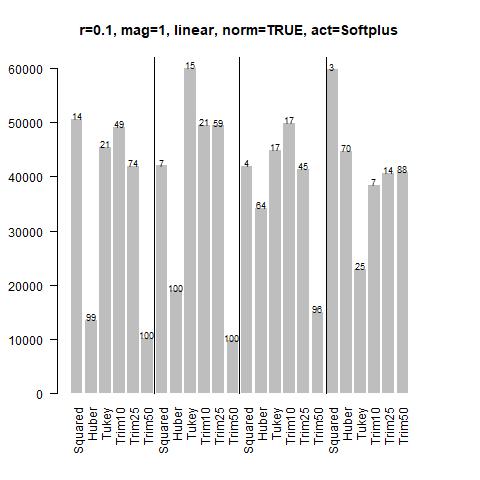}\\
\includegraphics[width=6.75cm,height=6.25cm]{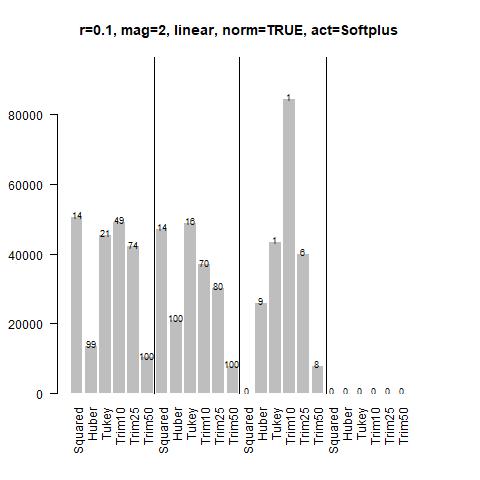} 
\includegraphics[width=6.75cm,height=6.25cm]{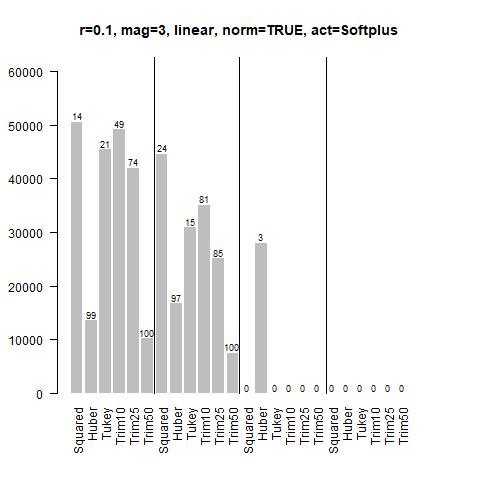} 
\end{center}
\caption{Results for $r=0.1$}
\end{figure}

\begin{figure}[H]
\label{trimnn:n500p20r25m1linnonreluStep}
\begin{center}
\includegraphics[width=6.75cm,height=6.25cm]{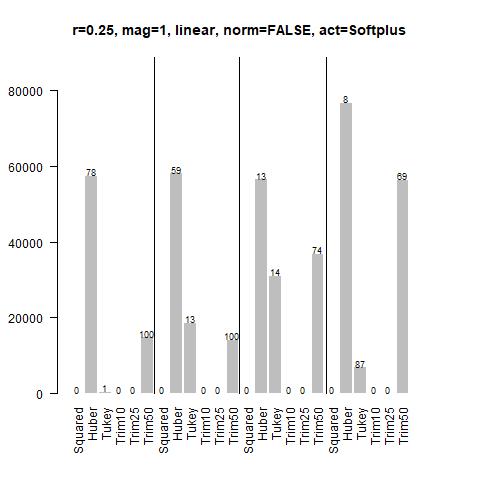}
\includegraphics[width=6.75cm,height=6.25cm]{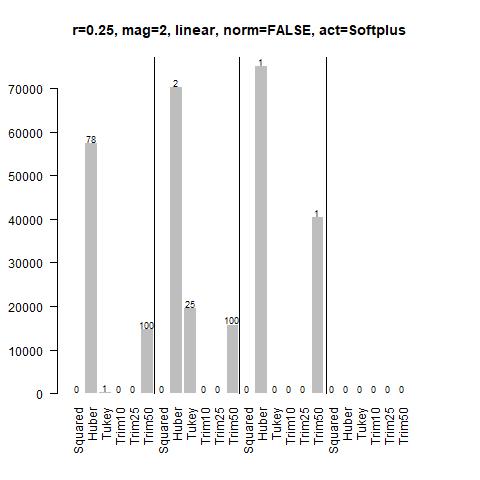} \\
\includegraphics[width=6.75cm,height=6.25cm]{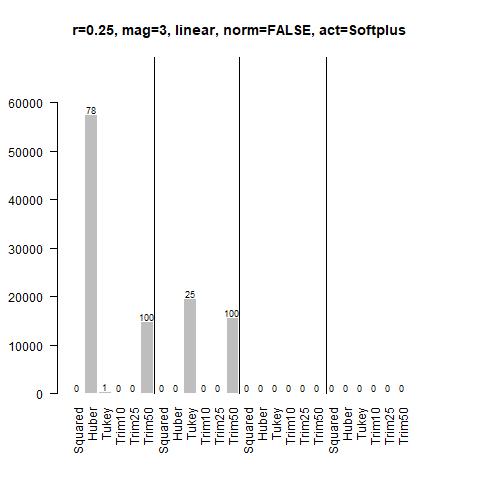} 
\includegraphics[width=6.75cm,height=6.25cm]{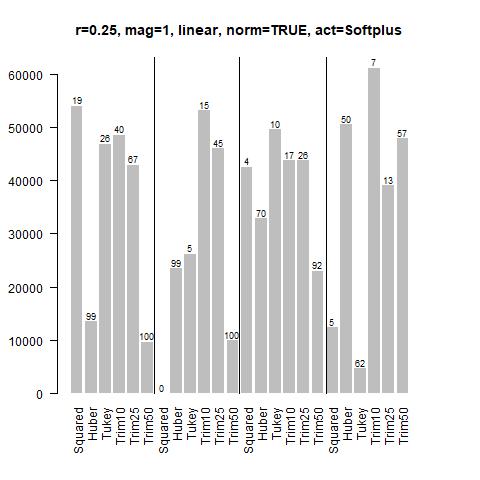}\\
\includegraphics[width=6.75cm,height=6.25cm]{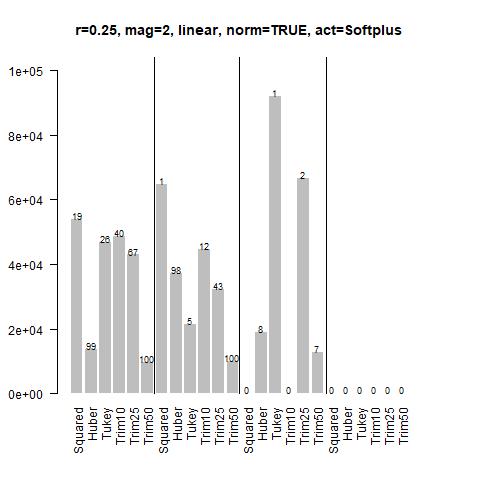} 
\includegraphics[width=6.75cm,height=6.25cm]{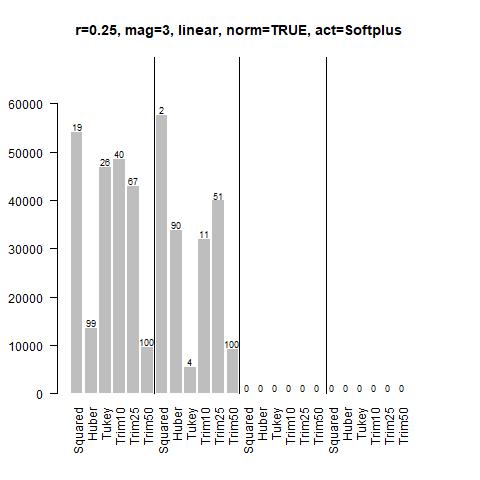} 
\end{center}
\caption{Results for $r=0.25$}
\end{figure}

\begin{figure}[H]
\label{trimnn:n500p20r40m1linnonreluStep}
\begin{center}
\includegraphics[width=6.75cm,height=6.25cm]{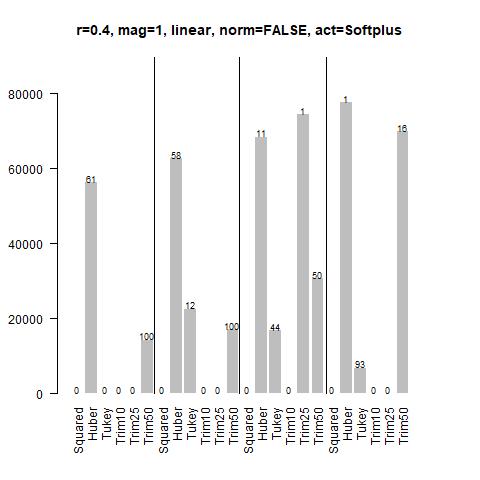}
\includegraphics[width=6.75cm,height=6.25cm]{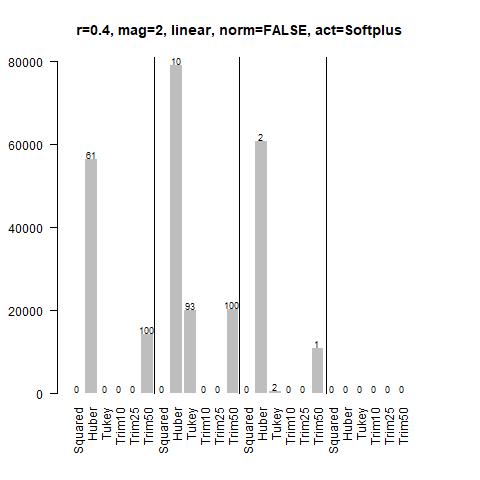} \\
\includegraphics[width=6.75cm,height=6.25cm]{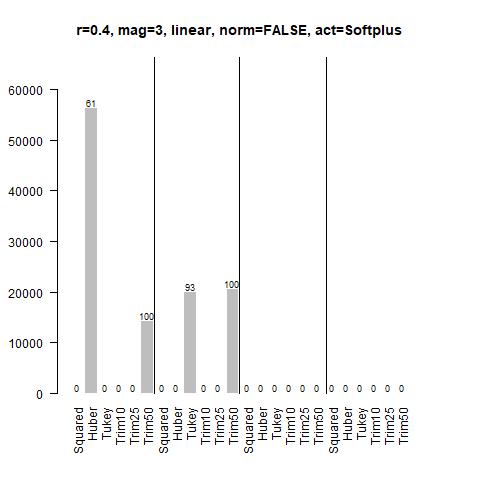} 
\includegraphics[width=6.75cm,height=6.25cm]{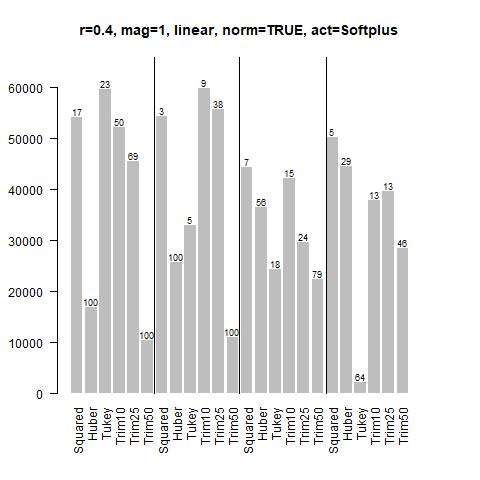}\\
\includegraphics[width=6.75cm,height=6.25cm]{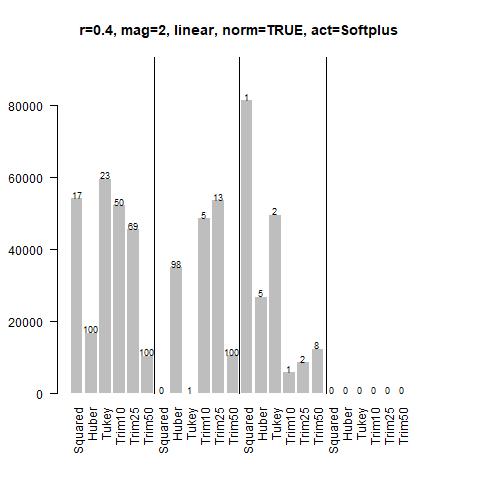} 
\includegraphics[width=6.75cm,height=6.25cm]{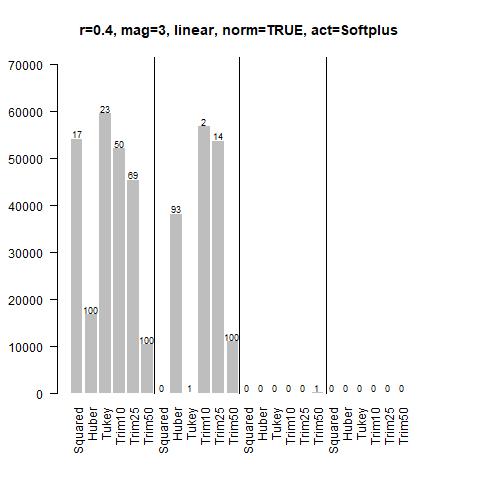} 
\end{center}
\caption{Results for $r=0.4$}
\end{figure}

\subsubsection{Polynomial function}

\begin{figure}[H]
\label{trimnn:n500p20r10m1polynonreluStep}
\begin{center}
\includegraphics[width=6.75cm,height=6.25cm]{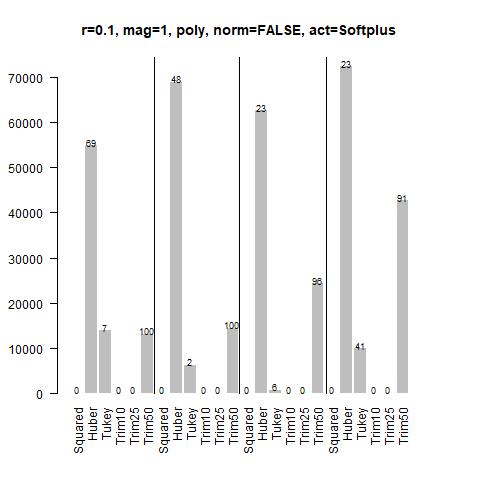}
\includegraphics[width=6.75cm,height=6.25cm]{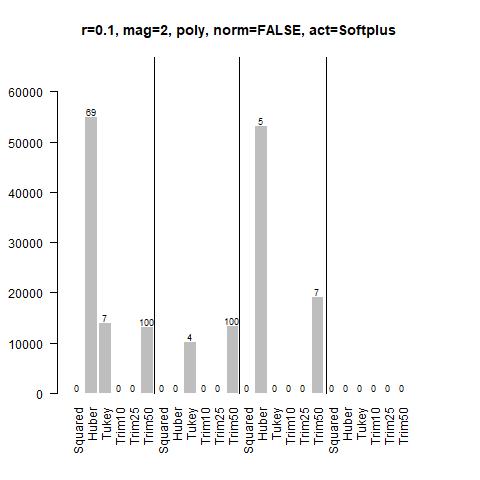} \\
\includegraphics[width=6.75cm,height=6.25cm]{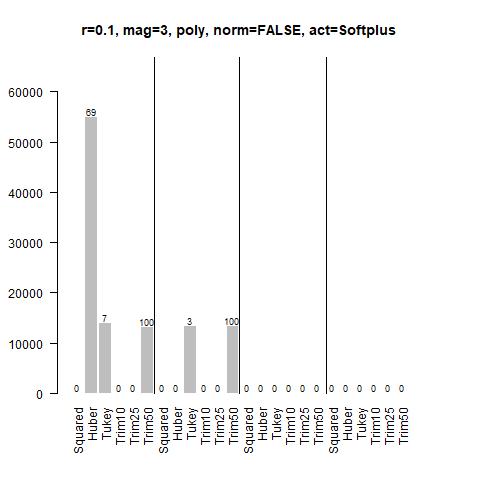} 
\includegraphics[width=6.75cm,height=6.25cm]{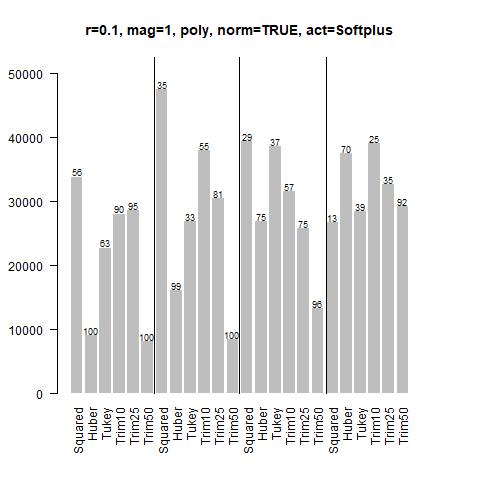}\\
\includegraphics[width=6.75cm,height=6.25cm]{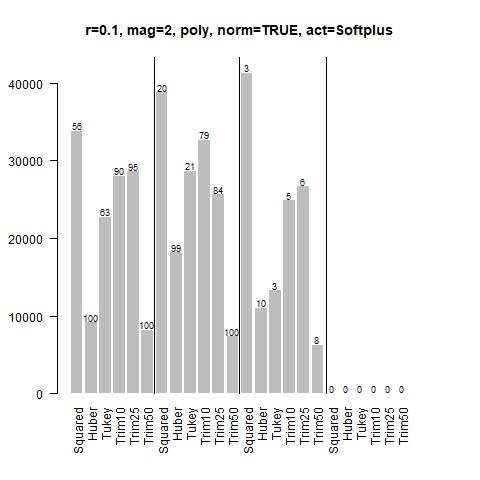} 
\includegraphics[width=6.75cm,height=6.25cm]{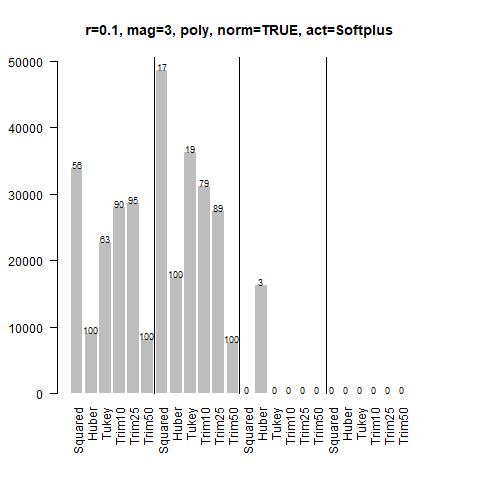} 
\end{center}
\caption{Results for $r=0.1$}
\end{figure}

\begin{figure}[H]
\label{trimnn:n500p20r25m1polynonreluStep}
\begin{center}
\includegraphics[width=6.75cm,height=6.25cm]{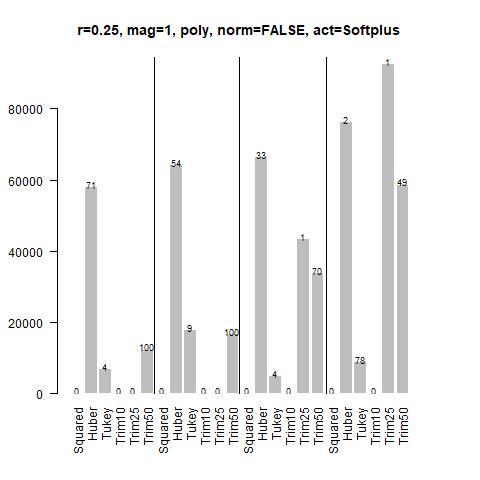}
\includegraphics[width=6.75cm,height=6.25cm]{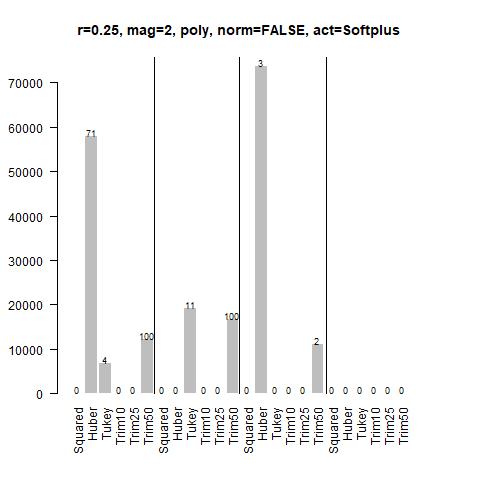} \\
\includegraphics[width=6.75cm,height=6.25cm]{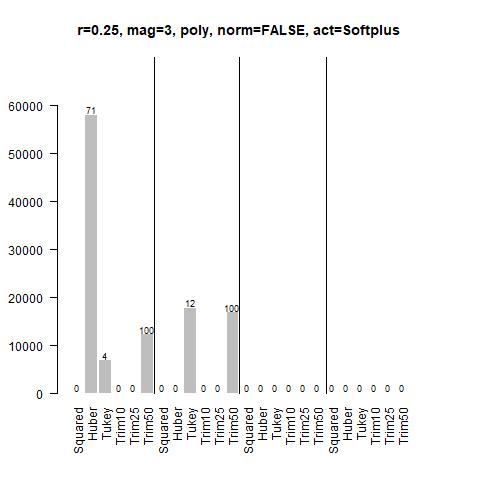} 
\includegraphics[width=6.75cm,height=6.25cm]{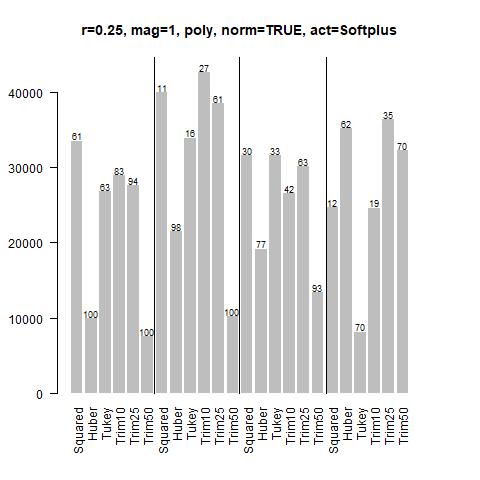}\\
\includegraphics[width=6.75cm,height=6.25cm]{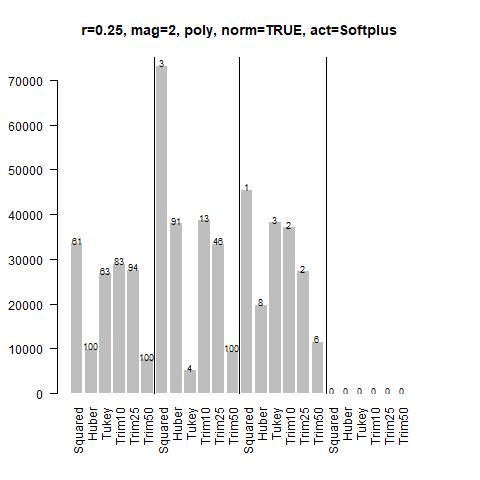} 
\includegraphics[width=6.75cm,height=6.25cm]{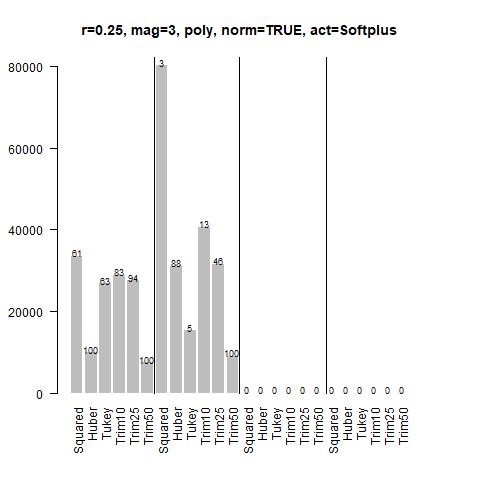} 
\end{center}
\caption{Results for $r=0.25$}
\end{figure}

\begin{figure}[H]
\label{trimnn:n500p20r40m1polynonreluStep}
\begin{center}
\includegraphics[width=6.75cm,height=6.25cm]{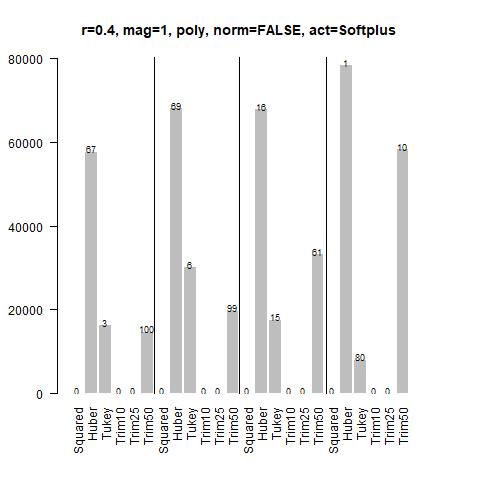}
\includegraphics[width=6.75cm,height=6.25cm]{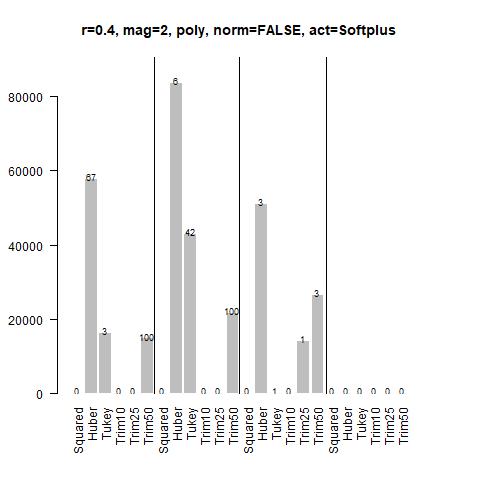} \\
\includegraphics[width=6.75cm,height=6.25cm]{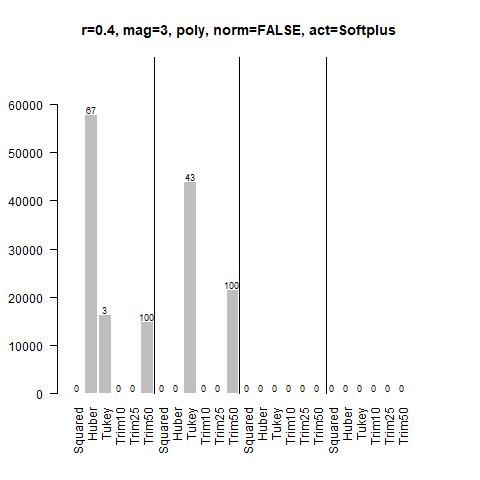} 
\includegraphics[width=6.75cm,height=6.25cm]{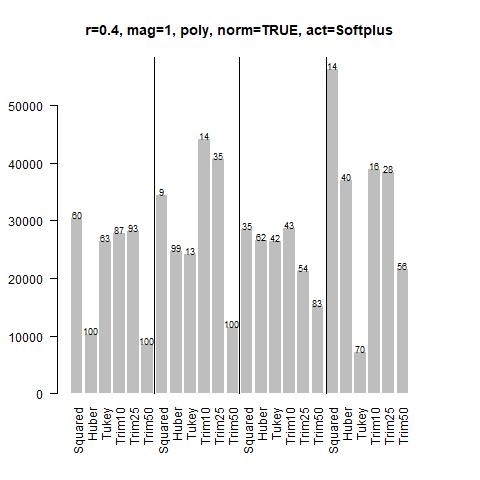}\\
\includegraphics[width=6.75cm,height=6.25cm]{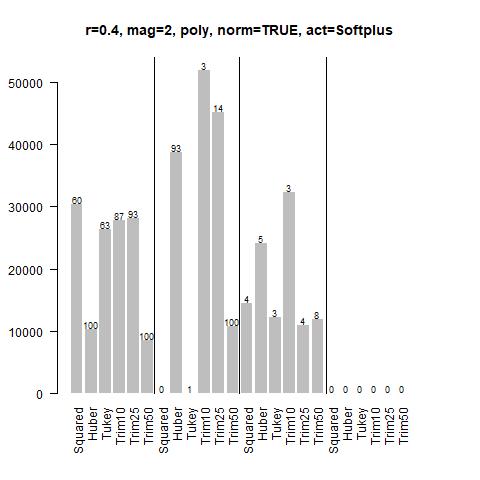} 
\includegraphics[width=6.75cm,height=6.25cm]{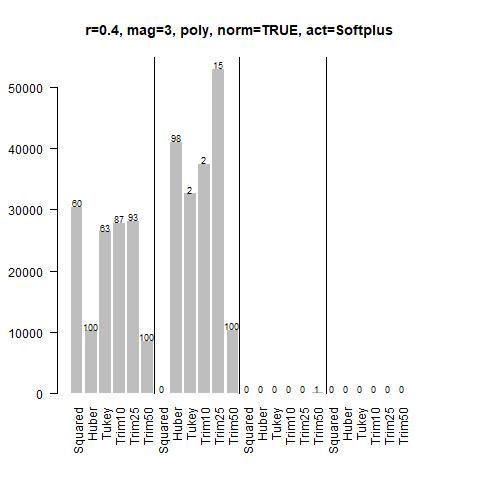} 
\end{center}
\caption{Results for $r=0.4$}
\end{figure}

\subsubsection{Trigonometric function}

\begin{figure}[H]
\label{trimnn:n500p20r10m1trignonreluStep}
\begin{center}
\includegraphics[width=6.75cm,height=6.25cm]{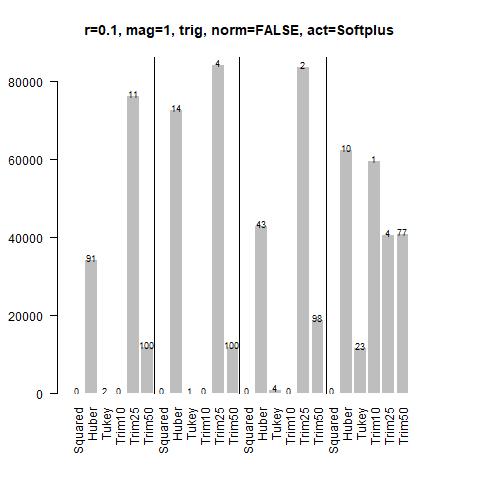}
\includegraphics[width=6.75cm,height=6.25cm]{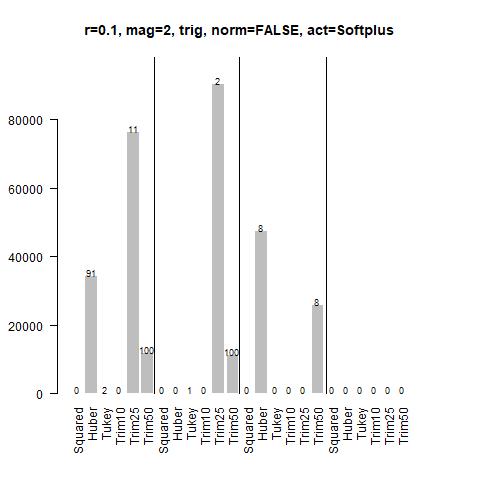} \\
\includegraphics[width=6.75cm,height=6.25cm]{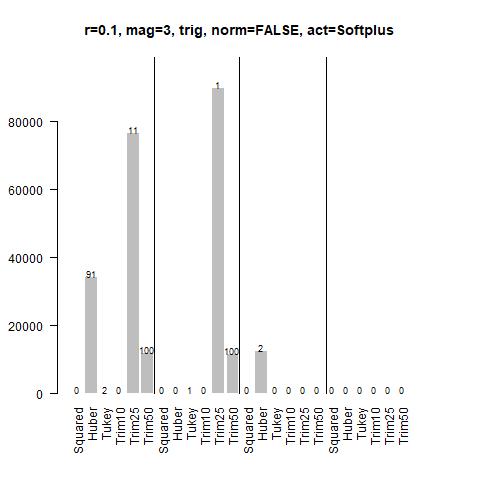} 
\includegraphics[width=6.75cm,height=6.25cm]{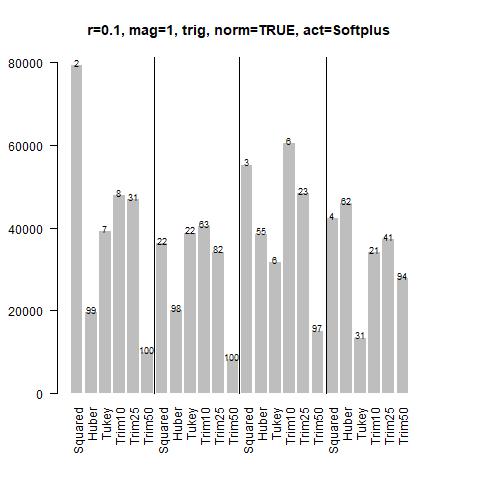}\\
\includegraphics[width=6.75cm,height=6.25cm]{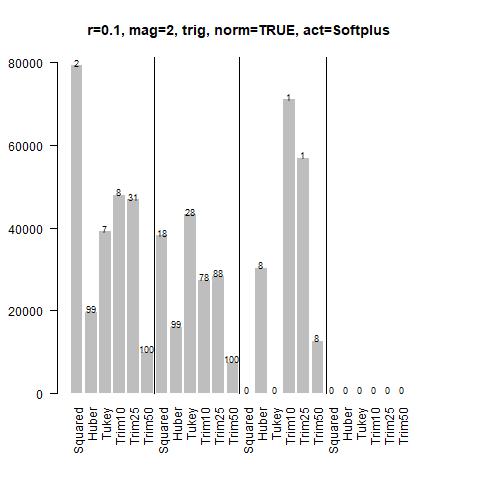} 
\includegraphics[width=6.75cm,height=6.25cm]{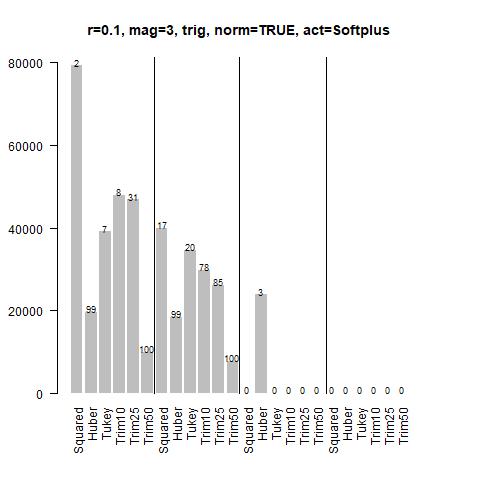} 
\end{center}
\caption{Results for $r=0.1$}
\end{figure}

\begin{figure}[H]
\label{trimnn:n500p20r25m1trignonreluStep}
\begin{center}
\includegraphics[width=6.75cm,height=6.25cm]{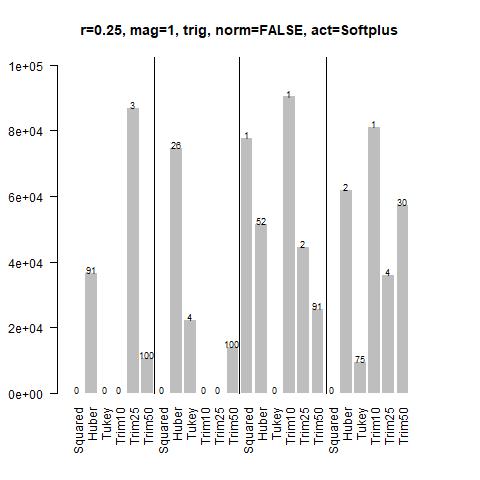}
\includegraphics[width=6.75cm,height=6.25cm]{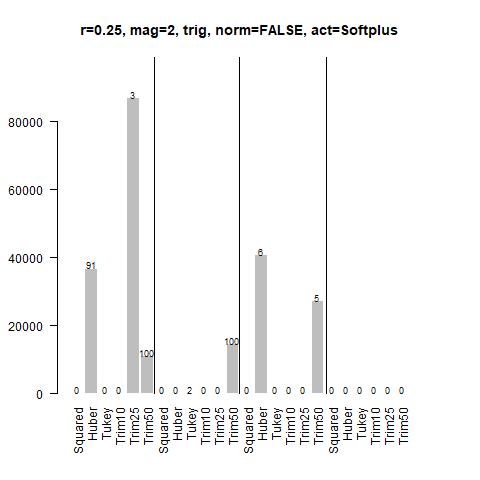} \\
\includegraphics[width=6.75cm,height=6.25cm]{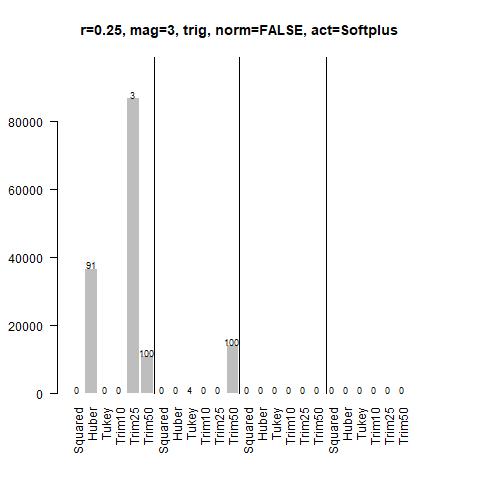} 
\includegraphics[width=6.75cm,height=6.25cm]{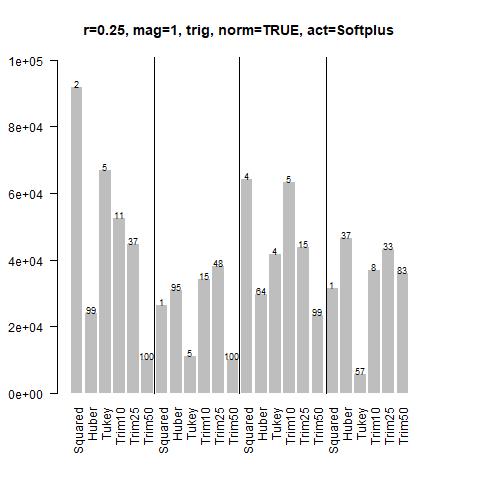}\\
\includegraphics[width=6.75cm,height=6.25cm]{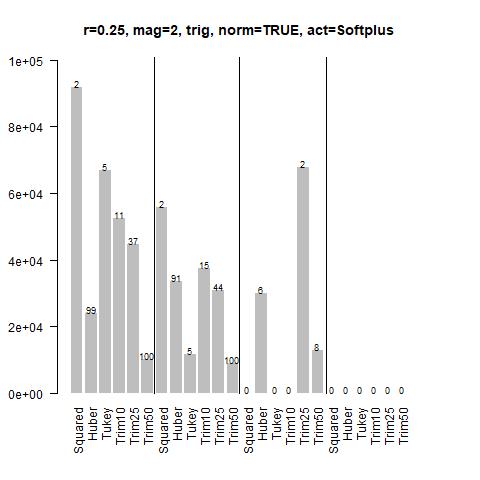} 
\includegraphics[width=6.75cm,height=6.25cm]{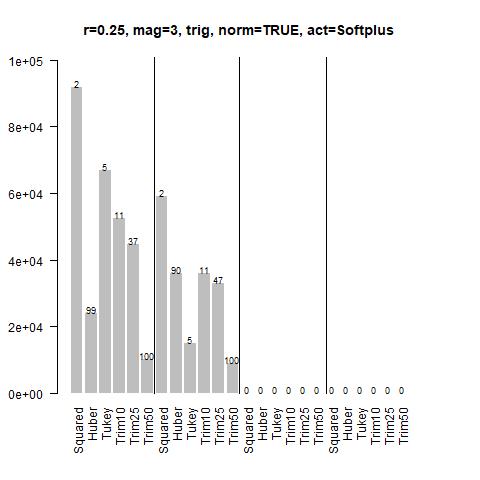} 
\end{center}
\caption{Results for $r=0.25$}
\end{figure}

\begin{figure}[H]
\label{trimnn:n500p20r40m1trignonreluStep}
\begin{center}
\includegraphics[width=6.75cm,height=6.25cm]{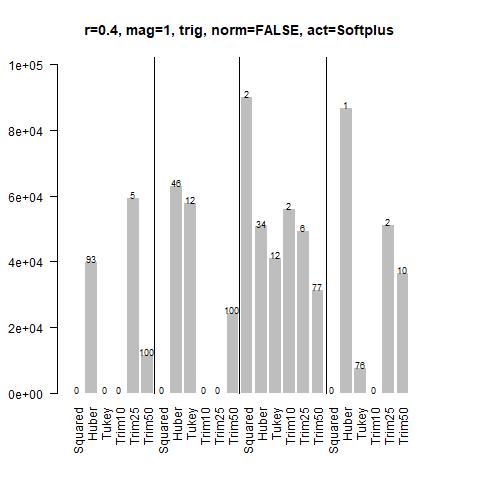}
\includegraphics[width=6.75cm,height=6.25cm]{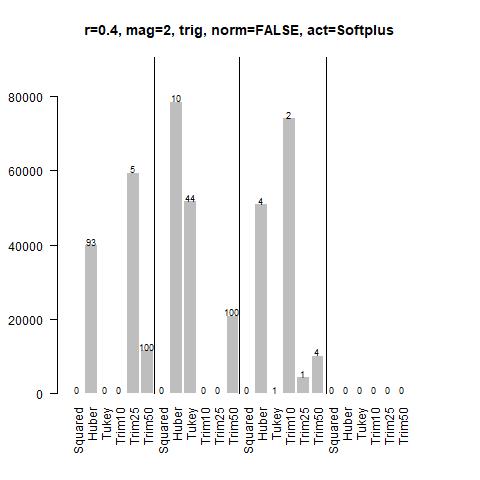} \\
\includegraphics[width=6.75cm,height=6.25cm]{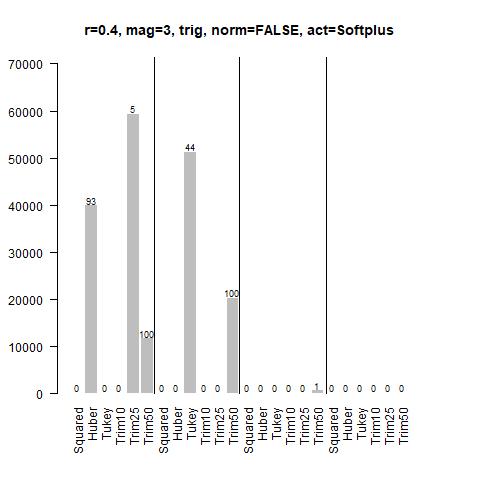} 
\includegraphics[width=6.75cm,height=6.25cm]{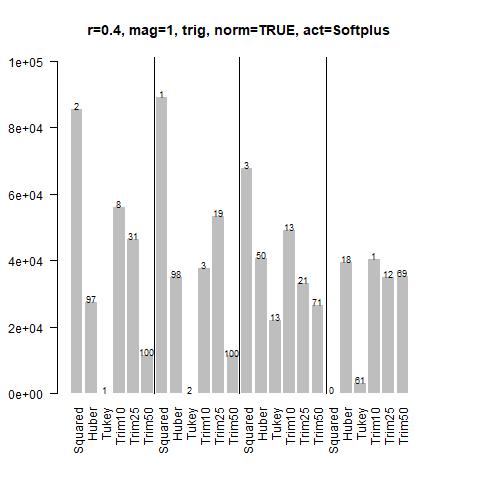}\\
\includegraphics[width=6.75cm,height=6.25cm]{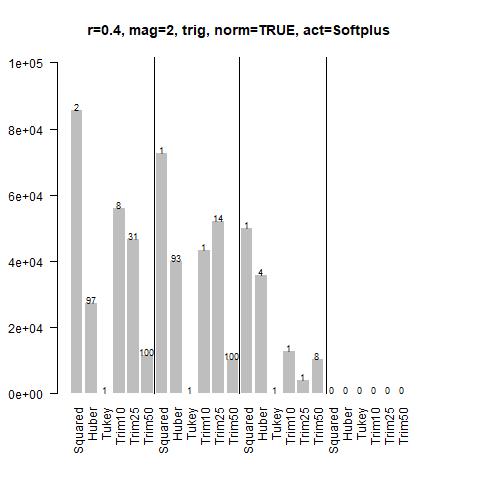} 
\includegraphics[width=6.75cm,height=6.25cm]{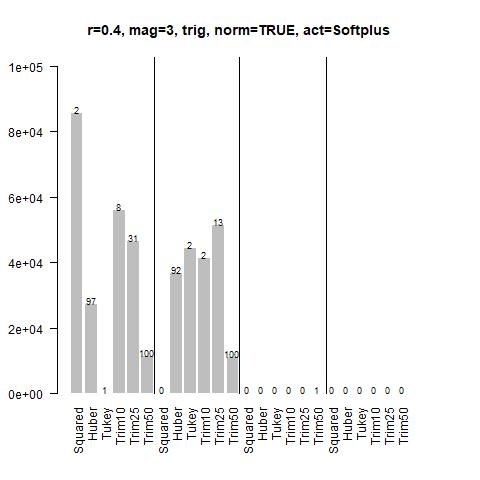} 
\end{center}
\caption{Results for $r=0.4$}
\end{figure}

\section{Simulation results for $n=500$ and $p=20$, deep network: Training steps}  \label{trimnn:secstep50020deep}

\subsection{Logistic activation function}

\subsubsection{Linear function}

\begin{figure}[H]
\label{trimnn:n500p20r10m1linnonlogdeepStep}
\begin{center}
\includegraphics[width=6.75cm,height=6.25cm]{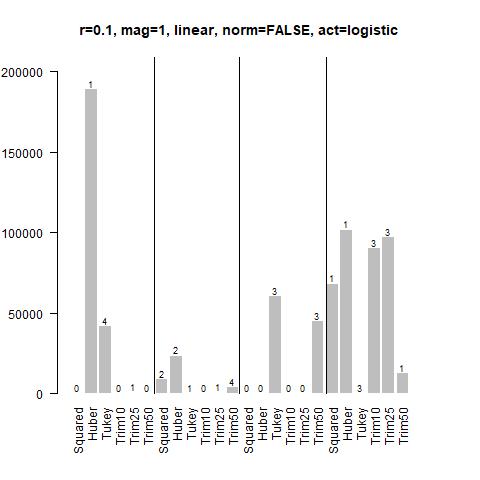}
\includegraphics[width=6.75cm,height=6.25cm]{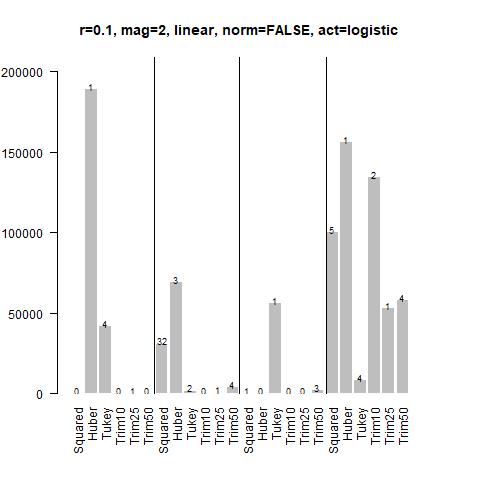} \\
\includegraphics[width=6.75cm,height=6.25cm]{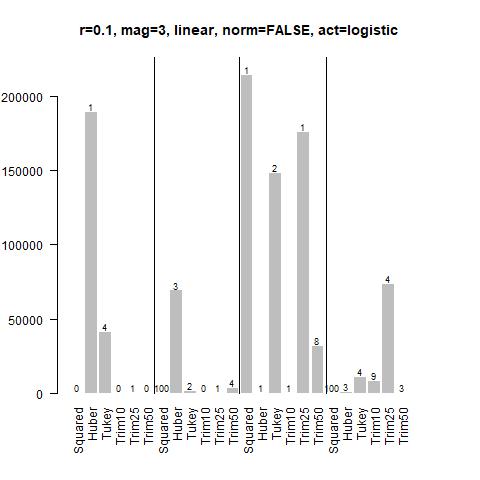} 
\includegraphics[width=6.75cm,height=6.25cm]{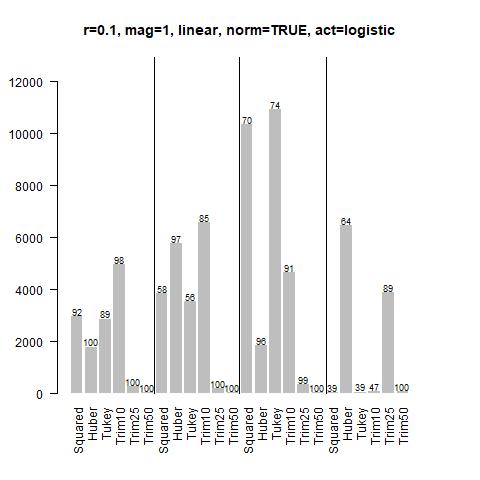}\\
\includegraphics[width=6.75cm,height=6.25cm]{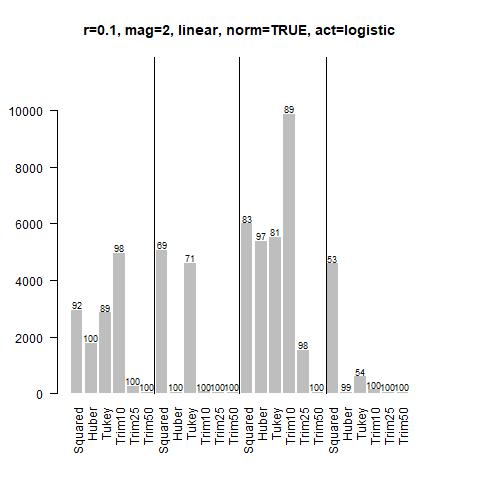} 
\includegraphics[width=6.75cm,height=6.25cm]{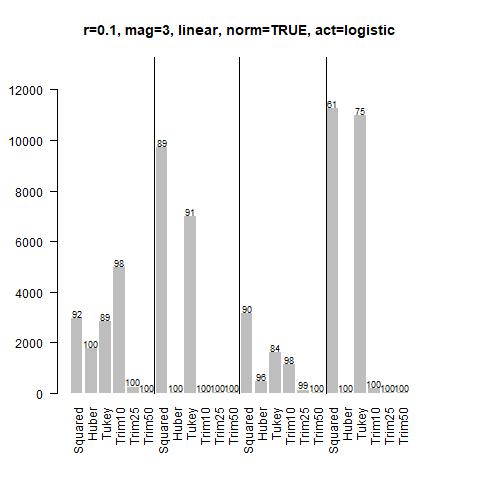} 
\end{center}
\caption{Results for $r=0.1$}
\end{figure}

\begin{figure}[H]
\label{trimnn:n500p20r25m1linnonlogdeepStep}
\begin{center}
\includegraphics[width=6.75cm,height=6.25cm]{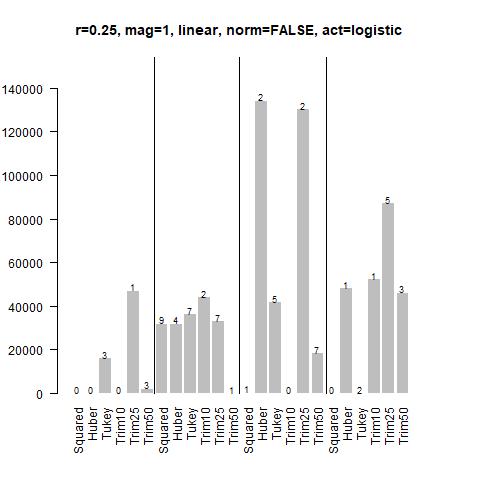}
\includegraphics[width=6.75cm,height=6.25cm]{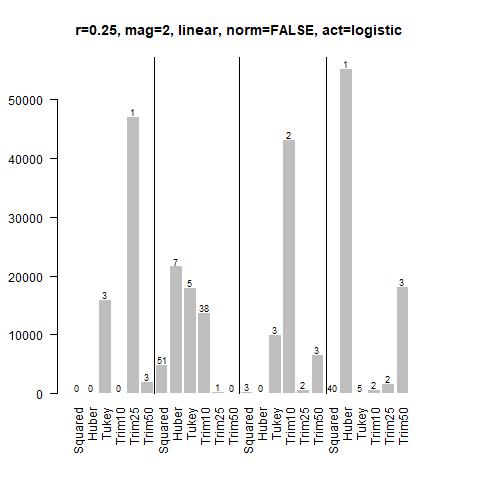} \\
\includegraphics[width=6.75cm,height=6.25cm]{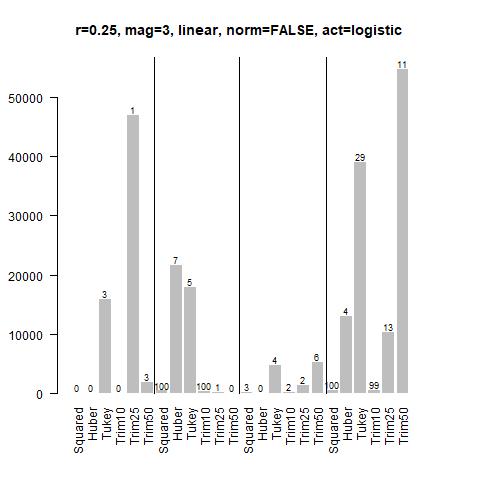} 
\includegraphics[width=6.75cm,height=6.25cm]{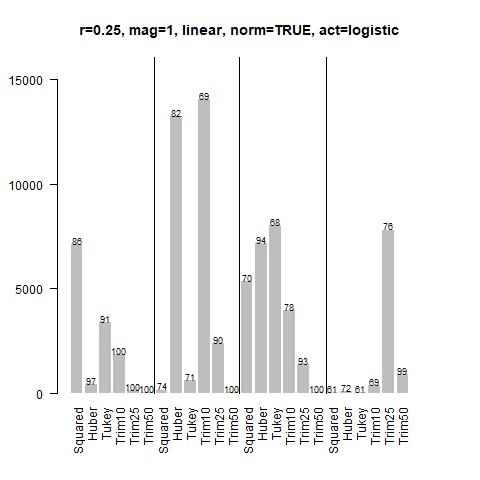}\\
\includegraphics[width=6.75cm,height=6.25cm]{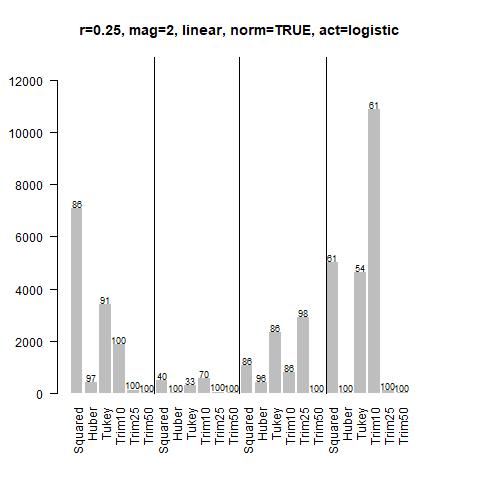} 
\includegraphics[width=6.75cm,height=6.25cm]{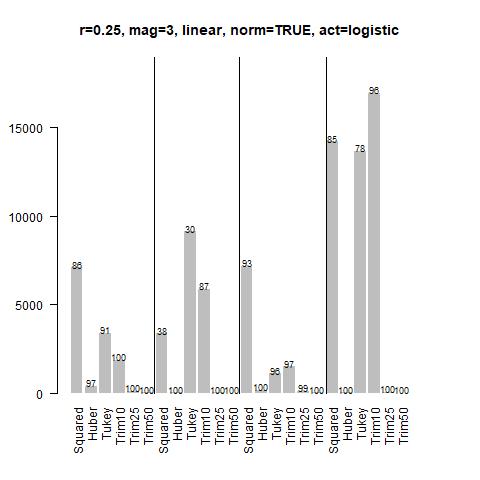} 
\end{center}
\caption{Results for $r=0.25$}
\end{figure}

\begin{figure}[H]
\label{trimnn:n500p20r40m1linnonlogdeepStep}
\begin{center}
\includegraphics[width=6.75cm,height=6.25cm]{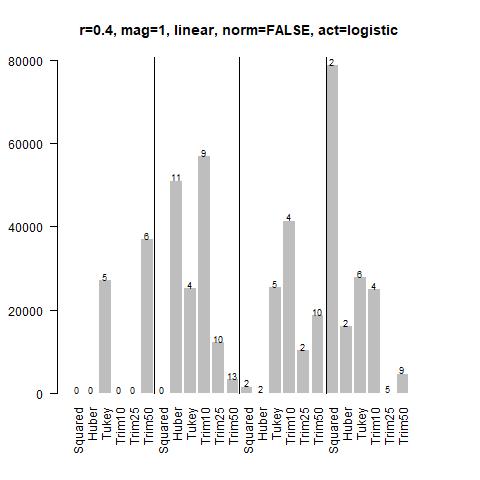}
\includegraphics[width=6.75cm,height=6.25cm]{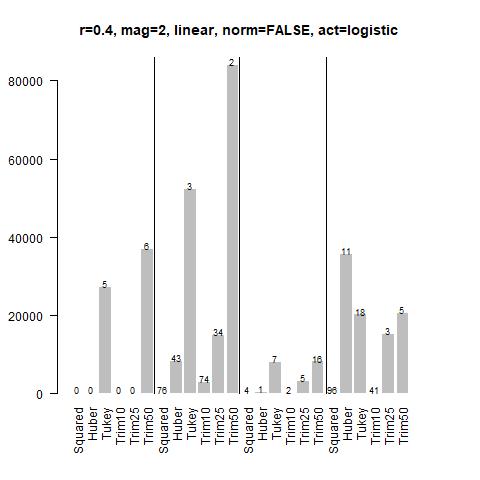} \\
\includegraphics[width=6.75cm,height=6.25cm]{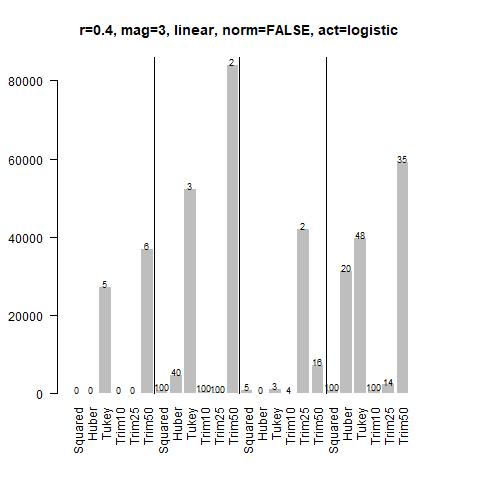} 
\includegraphics[width=6.75cm,height=6.25cm]{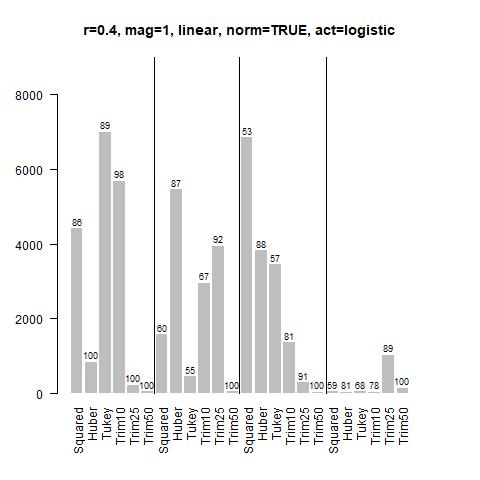}\\
\includegraphics[width=6.75cm,height=6.25cm]{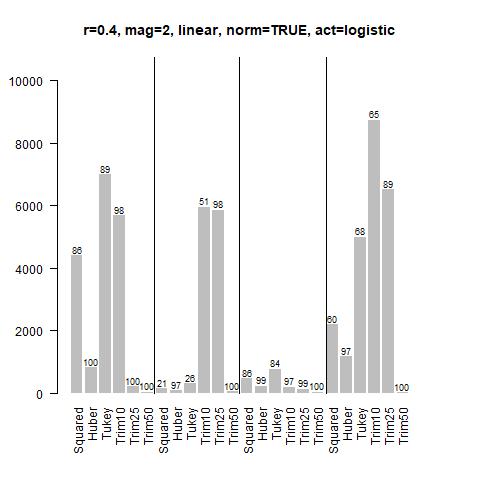} 
\includegraphics[width=6.75cm,height=6.25cm]{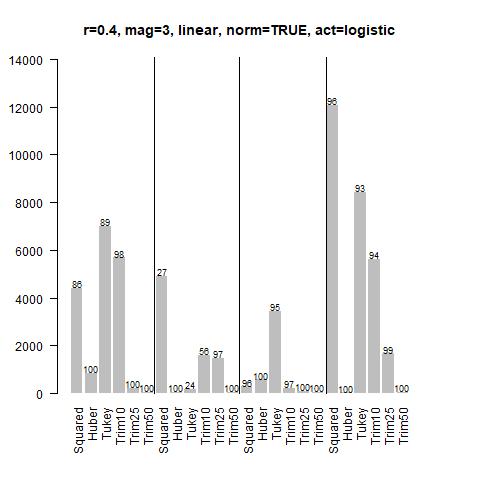} 
\end{center}
\caption{Results for $r=0.4$}
\end{figure}

\subsubsection{Polynomial function}

\begin{figure}[H]
\begin{center}
\includegraphics[width=6.75cm,height=6.25cm]{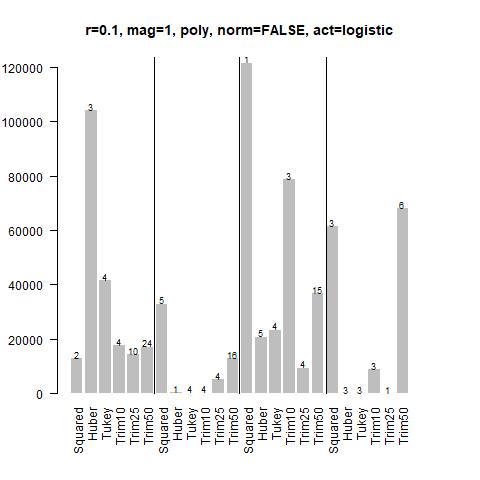}
\includegraphics[width=6.75cm,height=6.25cm]{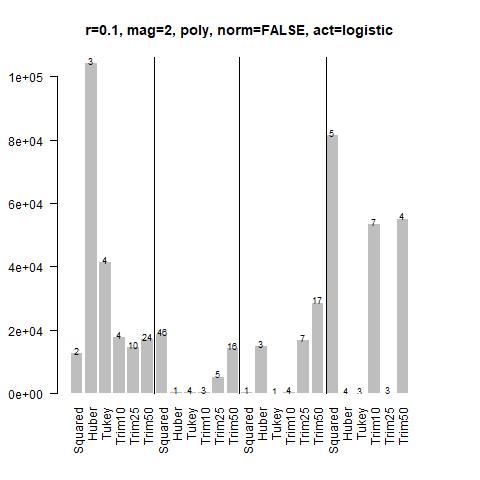} \\
\includegraphics[width=6.75cm,height=6.25cm]{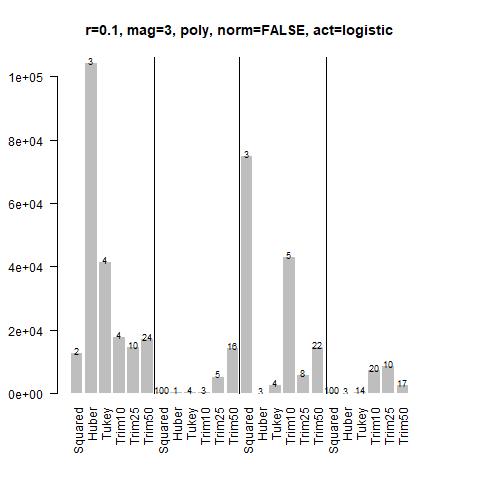} 
\includegraphics[width=6.75cm,height=6.25cm]{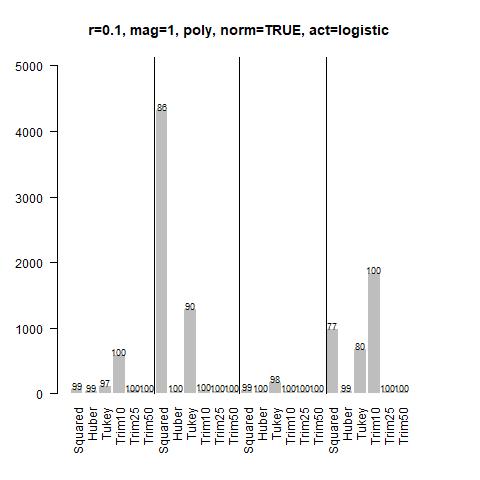}\\
\includegraphics[width=6.75cm,height=6.25cm]{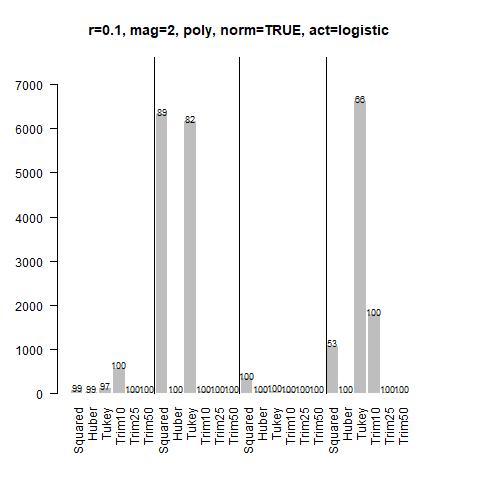} 
\includegraphics[width=6.75cm,height=6.25cm]{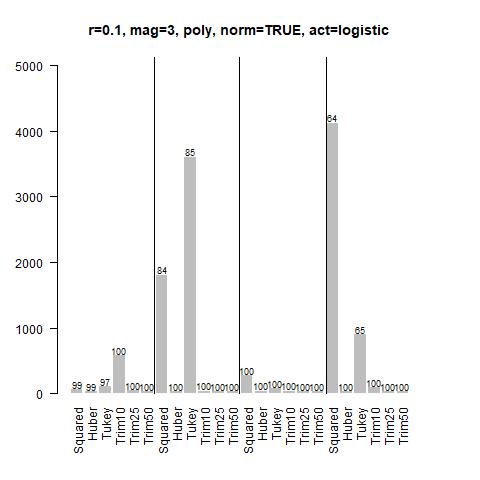} 
\end{center}
\caption{Results for $r=0.1$}\label{trimnn:n500p20r10m1polynonlogdeepStep}
\end{figure}

\begin{figure}[H]
\begin{center}
\includegraphics[width=6.75cm,height=6.25cm]{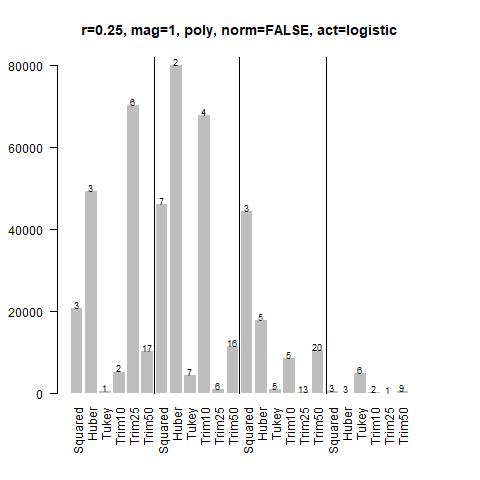}
\includegraphics[width=6.75cm,height=6.25cm]{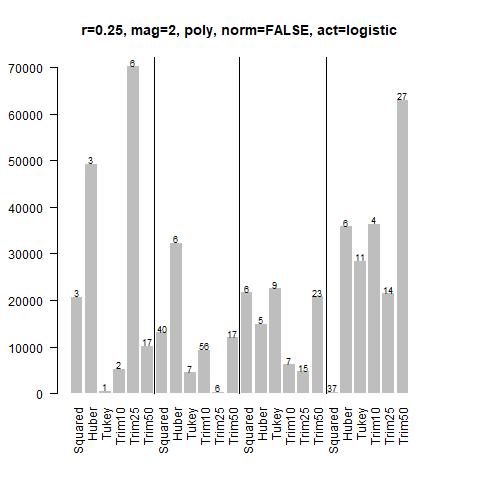} \\
\includegraphics[width=6.75cm,height=6.25cm]{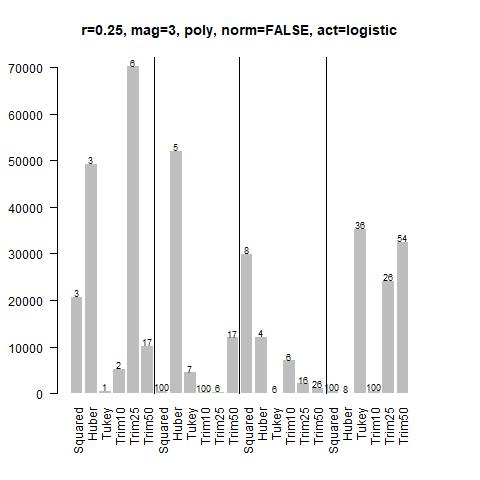} 
\includegraphics[width=6.75cm,height=6.25cm]{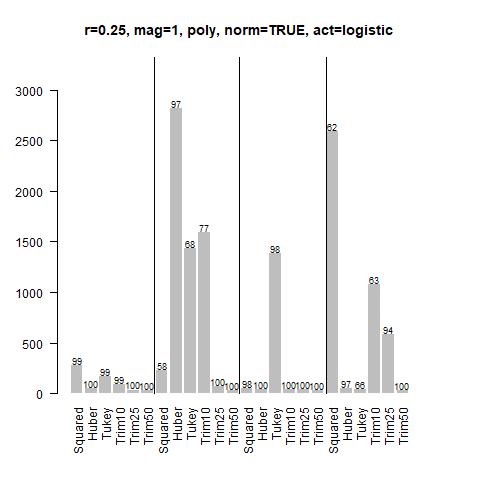}\\
\includegraphics[width=6.75cm,height=6.25cm]{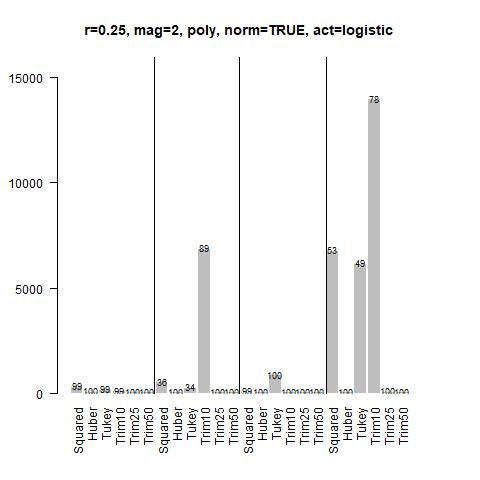} 
\includegraphics[width=6.75cm,height=6.25cm]{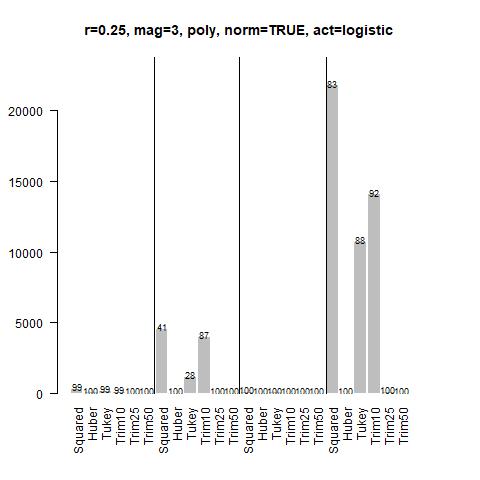} 
\end{center}
\caption{Results for $r=0.25$}\label{trimnn:n500p20r25m1polynonlogdeepStep}
\end{figure}

\begin{figure}[H]
\begin{center}
\includegraphics[width=6.75cm,height=6.25cm]{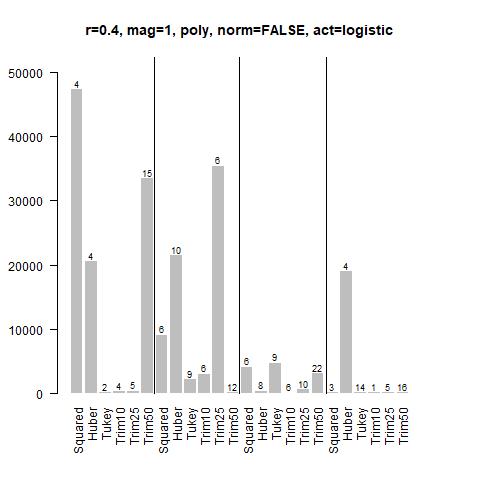}
\includegraphics[width=6.75cm,height=6.25cm]{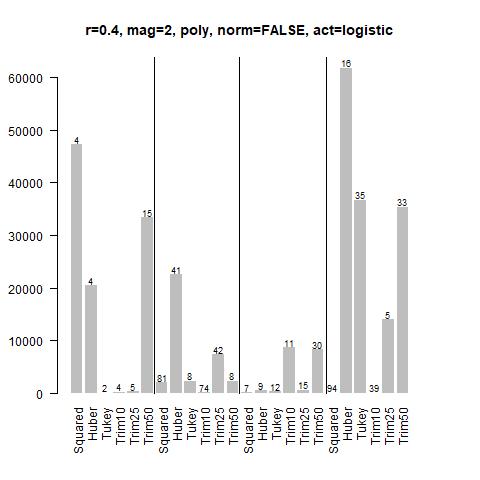} \\
\includegraphics[width=6.75cm,height=6.25cm]{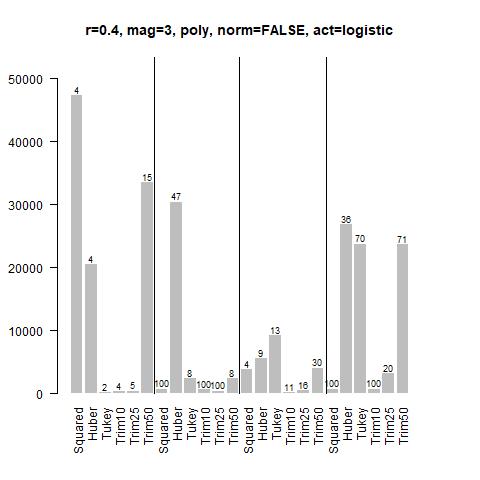} 
\includegraphics[width=6.75cm,height=6.25cm]{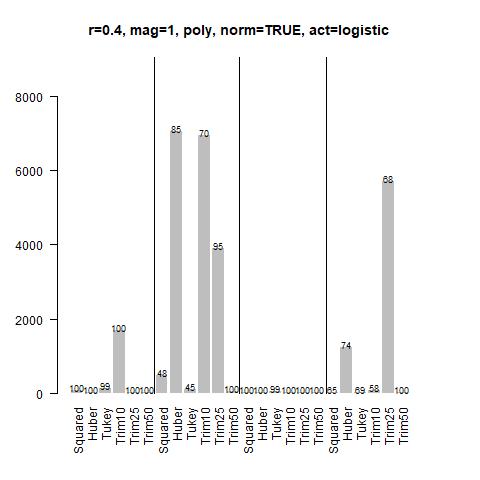}\\
\includegraphics[width=6.75cm,height=6.25cm]{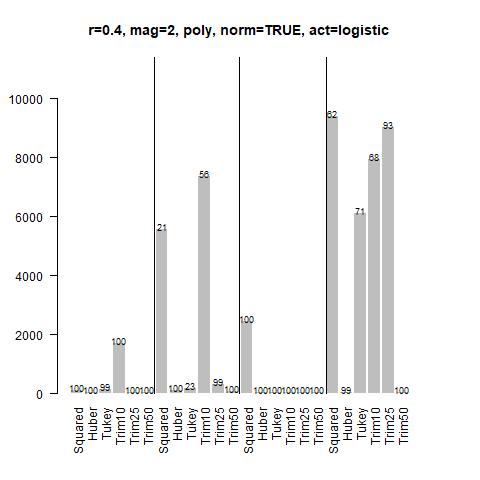} 
\includegraphics[width=6.75cm,height=6.25cm]{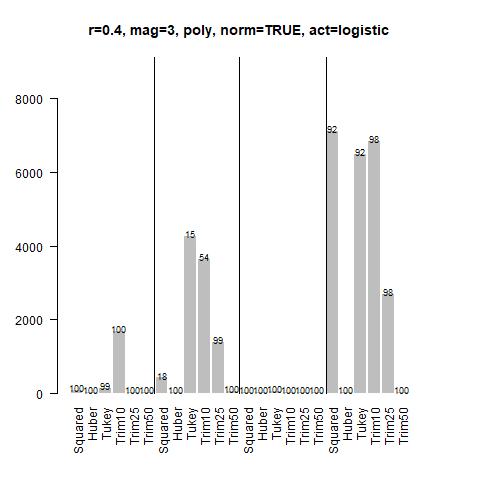} 
\end{center}
\caption{Results for $r=0.4$}\label{trimnn:n500p20r40m1polynonlogdeepStep}
\end{figure}

\subsubsection{Trigonometric function}

\begin{figure}[H]
\label{trimnn:n500p20r10m1trignonlogdeepStep}
\begin{center}
\includegraphics[width=6.75cm,height=6.25cm]{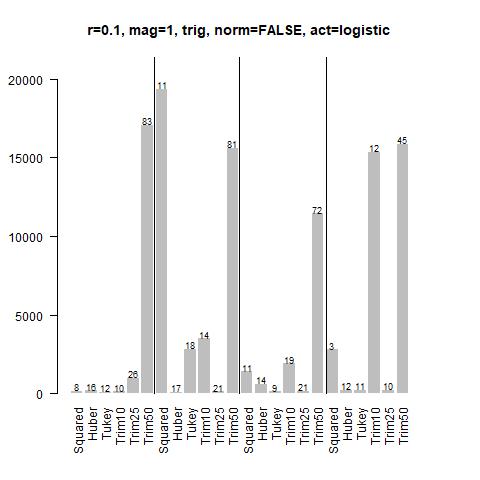}
\includegraphics[width=6.75cm,height=6.25cm]{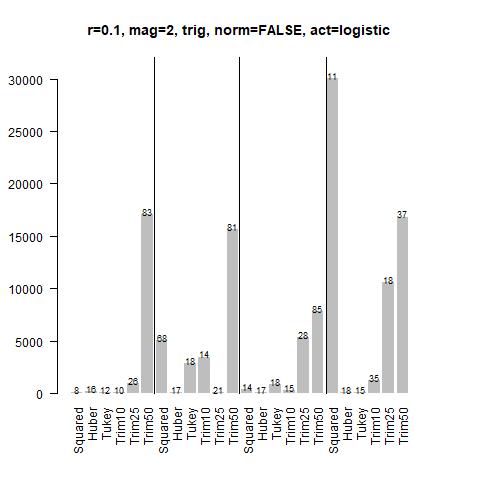} \\
\includegraphics[width=6.75cm,height=6.25cm]{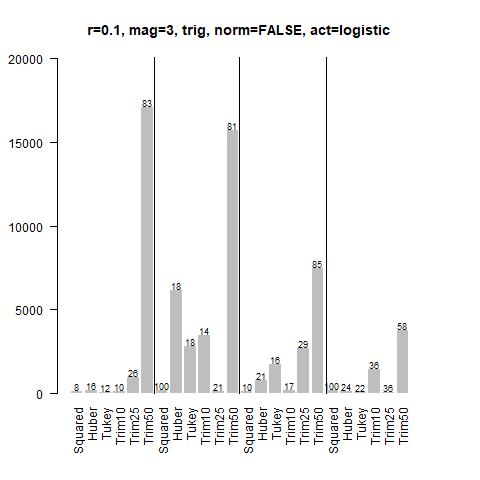} 
\includegraphics[width=6.75cm,height=6.25cm]{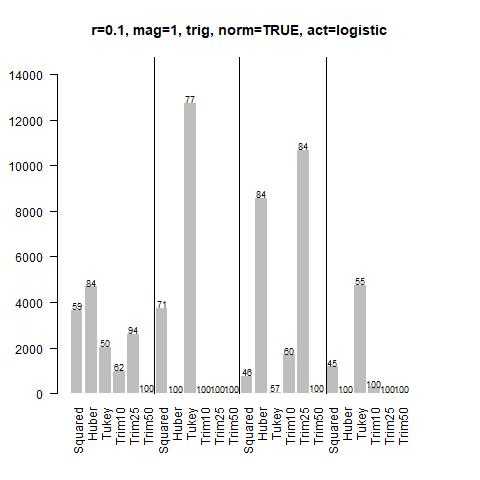}\\
\includegraphics[width=6.75cm,height=6.25cm]{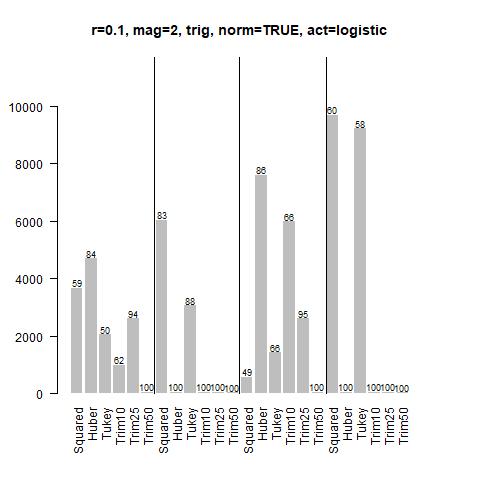} 
\includegraphics[width=6.75cm,height=6.25cm]{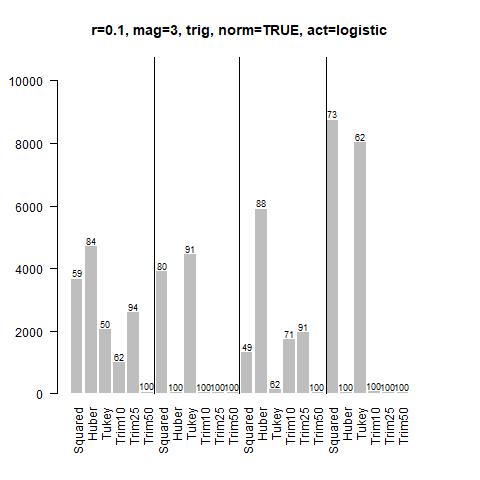} 
\end{center}
\caption{Results for $r=0.1$}
\end{figure}

\begin{figure}[H]
\label{trimnn:n500p20r25m1trignonlogdeepStep}
\begin{center}
\includegraphics[width=6.75cm,height=6.25cm]{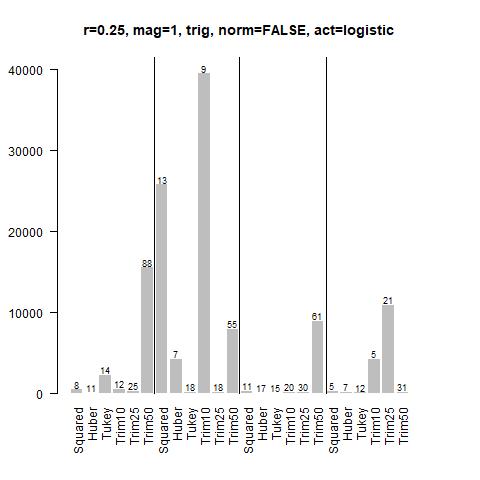}
\includegraphics[width=6.75cm,height=6.25cm]{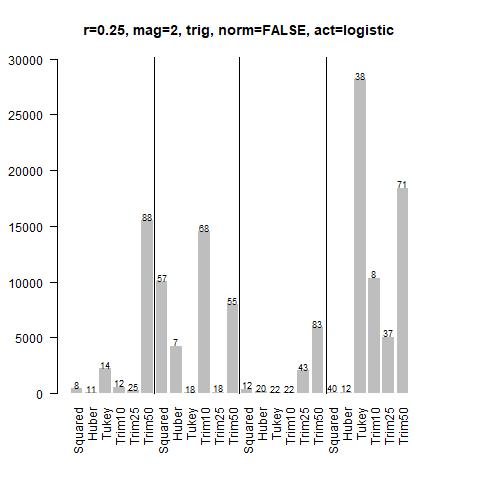} \\
\includegraphics[width=6.75cm,height=6.25cm]{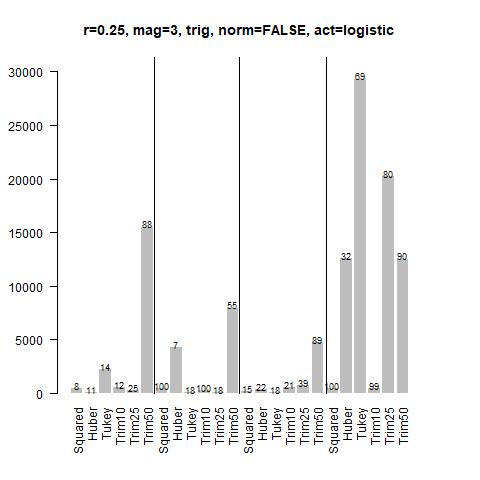} 
\includegraphics[width=6.75cm,height=6.25cm]{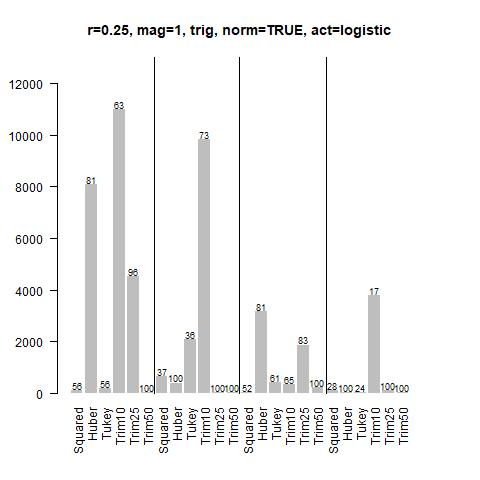}\\
\includegraphics[width=6.75cm,height=6.25cm]{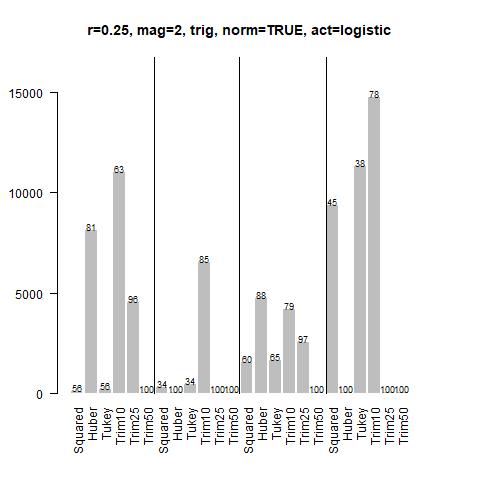} 
\includegraphics[width=6.75cm,height=6.25cm]{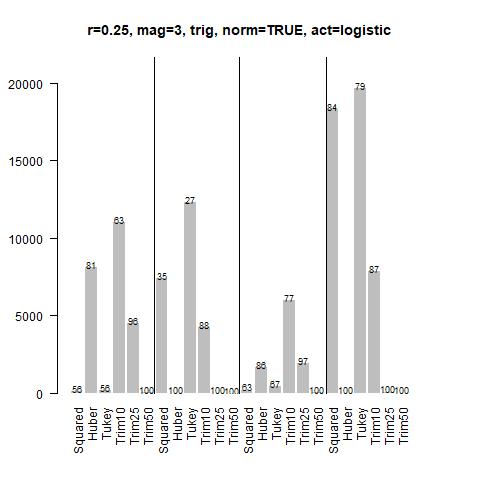} 
\end{center}
\caption{Results for $r=0.25$}
\end{figure}

\begin{figure}[H]
\label{trimnn:n500p20r40m1trignonlogdeepStep}
\begin{center}
\includegraphics[width=6.75cm,height=6.25cm]{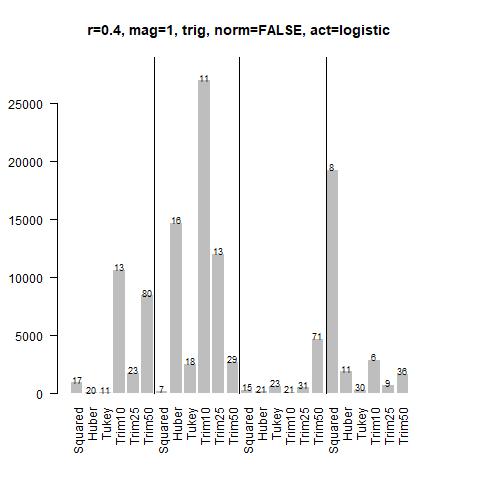}
\includegraphics[width=6.75cm,height=6.25cm]{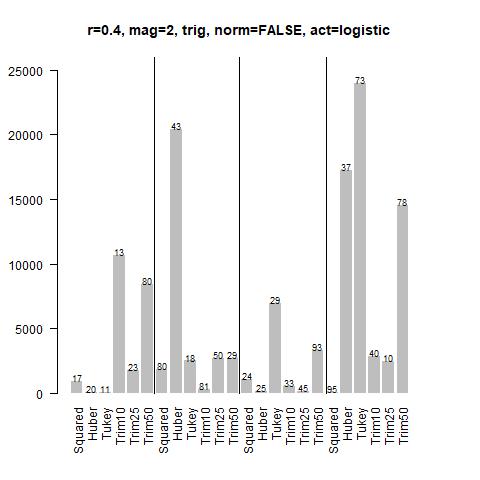} \\
\includegraphics[width=6.75cm,height=6.25cm]{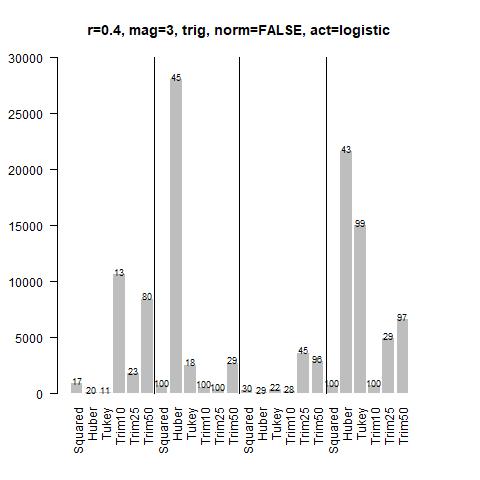} 
\includegraphics[width=6.75cm,height=6.25cm]{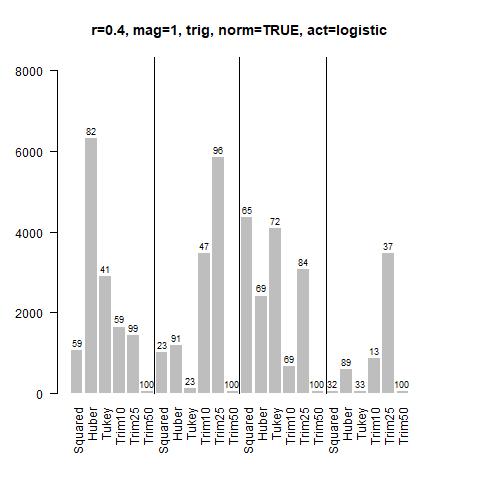}\\
\includegraphics[width=6.75cm,height=6.25cm]{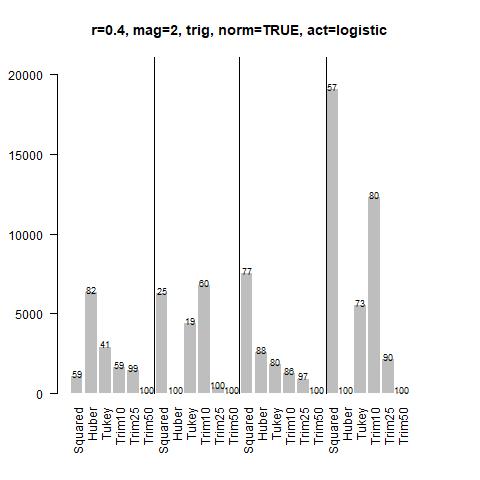} 
\includegraphics[width=6.75cm,height=6.25cm]{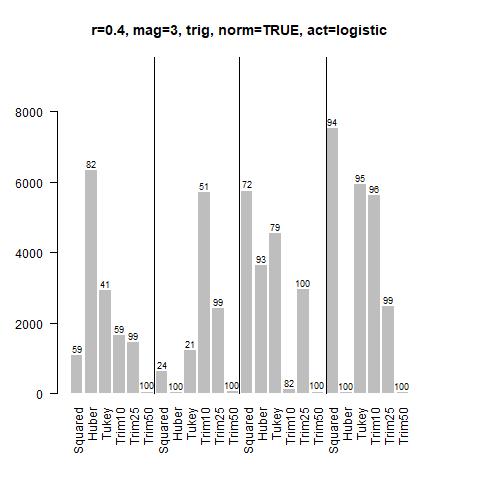} 
\end{center}
\caption{Results for $r=0.4$}
\end{figure}

\subsection{Softplus activation function}

\subsubsection{Linear function}

\begin{figure}[H]
\begin{center}
\includegraphics[width=6.75cm,height=6.25cm]{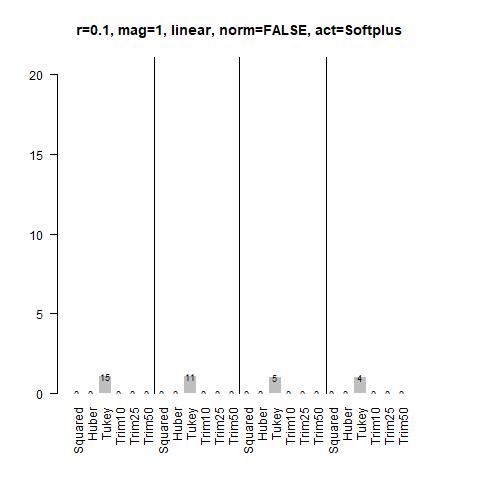}
\includegraphics[width=6.75cm,height=6.25cm]{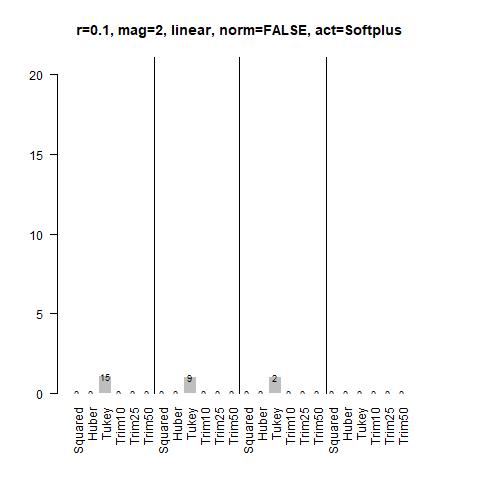} \\
\includegraphics[width=6.75cm,height=6.25cm]{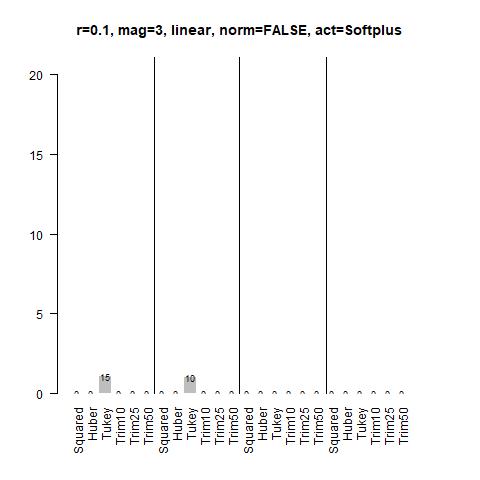} 
\includegraphics[width=6.75cm,height=6.25cm]{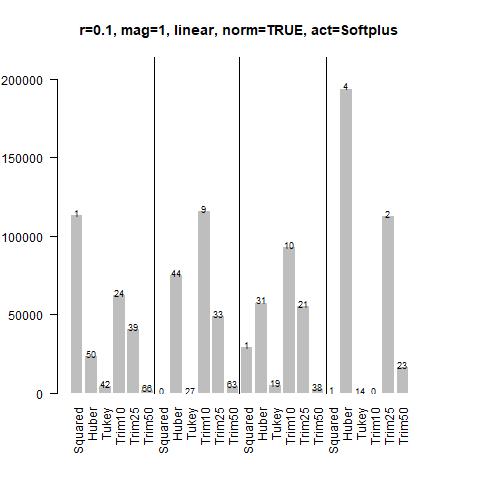}\\
\includegraphics[width=6.75cm,height=6.25cm]{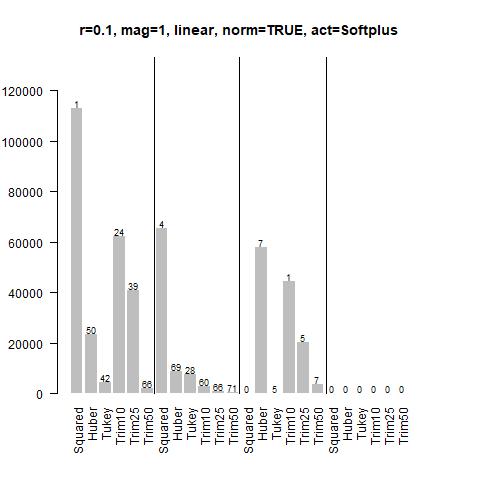} 
\includegraphics[width=6.75cm,height=6.25cm]{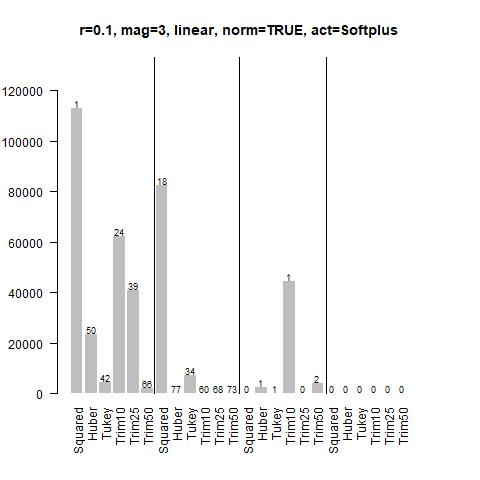} 
\end{center}
\caption{Results for $r=0.1$}\label{trimnn:n500p20r10m1linnonreludeepStep}
\end{figure}

\begin{figure}[H]
\begin{center}
\includegraphics[width=6.75cm,height=6.25cm]{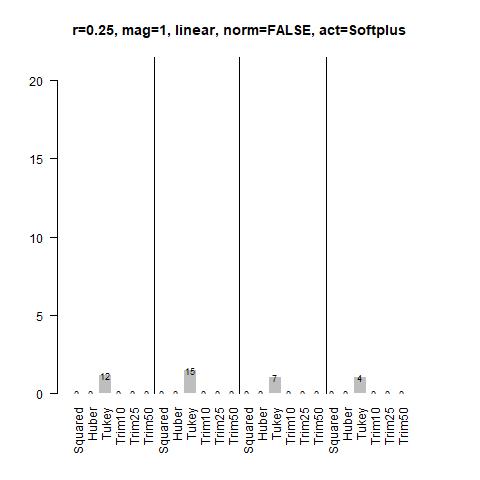}
\includegraphics[width=6.75cm,height=6.25cm]{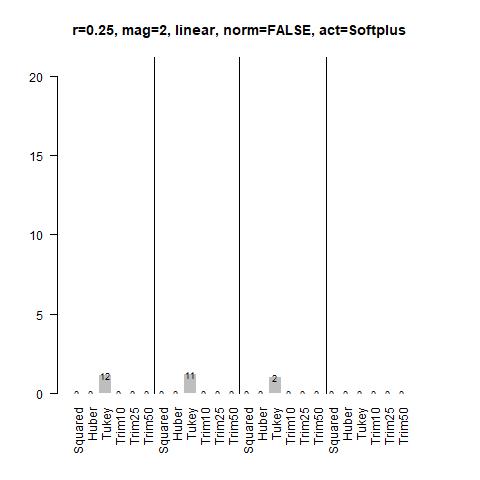} \\
\includegraphics[width=6.75cm,height=6.25cm]{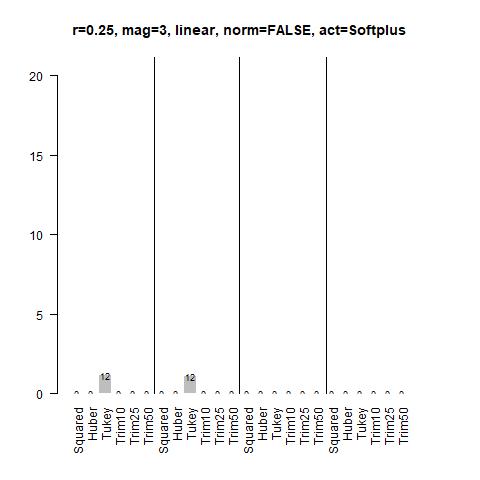} 
\includegraphics[width=6.75cm,height=6.25cm]{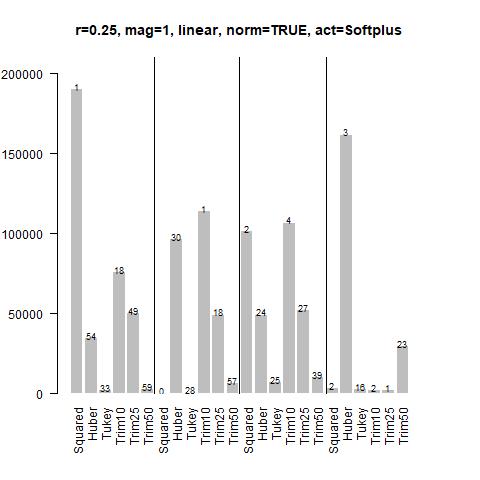}\\
\includegraphics[width=6.75cm,height=6.25cm]{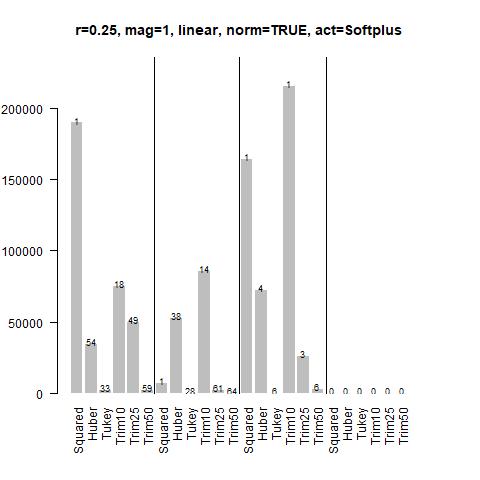} 
\includegraphics[width=6.75cm,height=6.25cm]{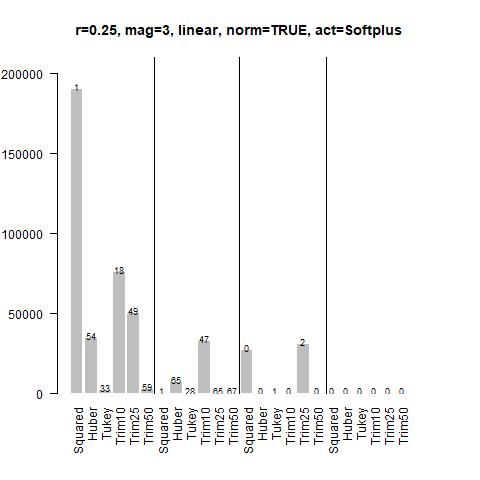} 
\end{center}
\caption{Results for $r=0.25$}\label{trimnn:n500p20r25m1linnonreludeepStep}
\end{figure}

\begin{figure}[H]
\label{trimnn:n500p20r40m1linnonreludeepStep}
\begin{center}
\includegraphics[width=6.75cm,height=6.25cm]{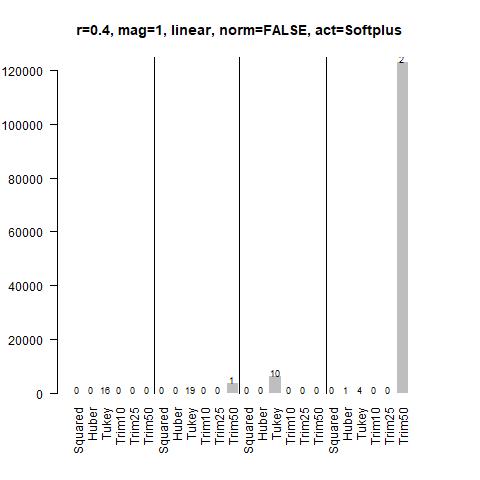}
\includegraphics[width=6.75cm,height=6.25cm]{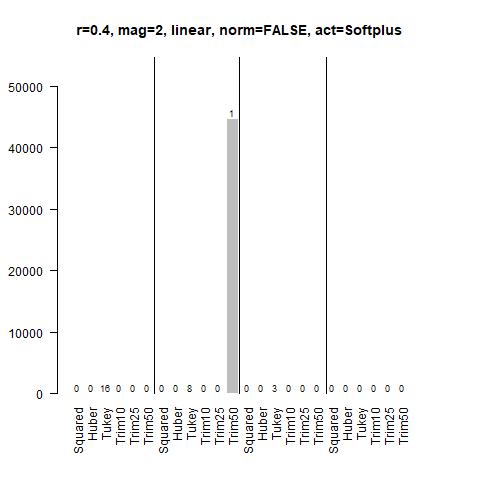} \\
\includegraphics[width=6.75cm,height=6.25cm]{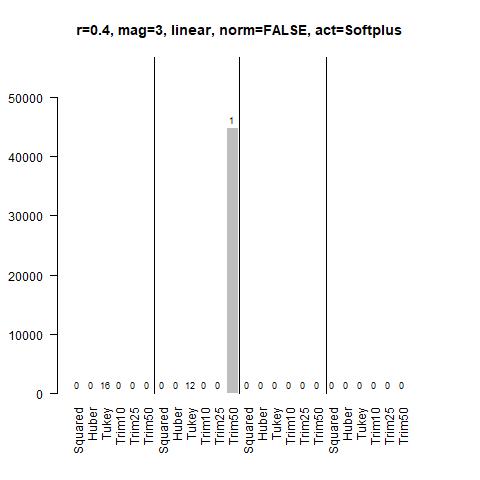} 
\includegraphics[width=6.75cm,height=6.25cm]{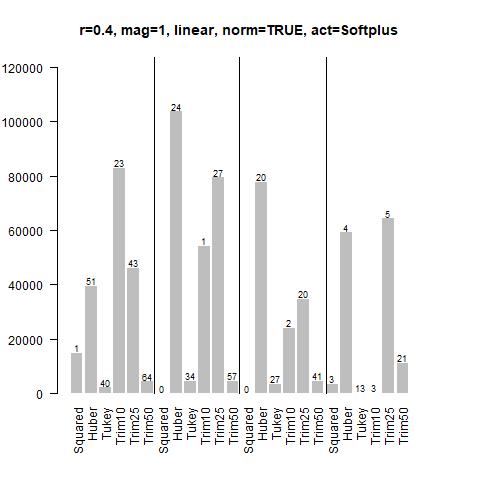}\\
\includegraphics[width=6.75cm,height=6.25cm]{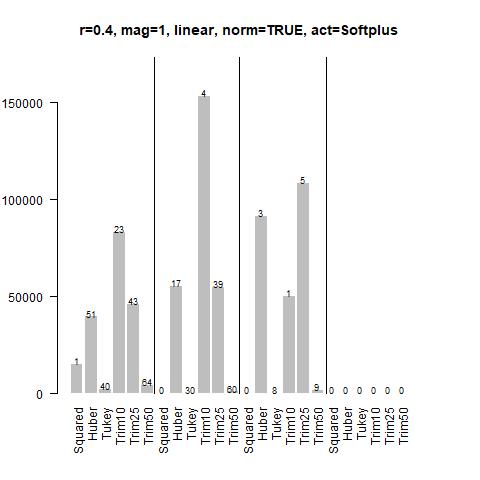} 
\includegraphics[width=6.75cm,height=6.25cm]{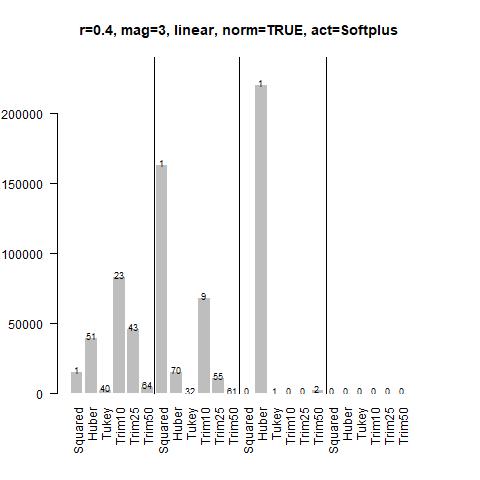} 
\end{center}
\caption{Results for $r=0.4$}
\end{figure}

\subsubsection{Polynomial function}

\begin{figure}[H]
\label{trimnn:n500p20r10m1polynonreludeepStep}
\begin{center}
\includegraphics[width=6.75cm,height=6.25cm]{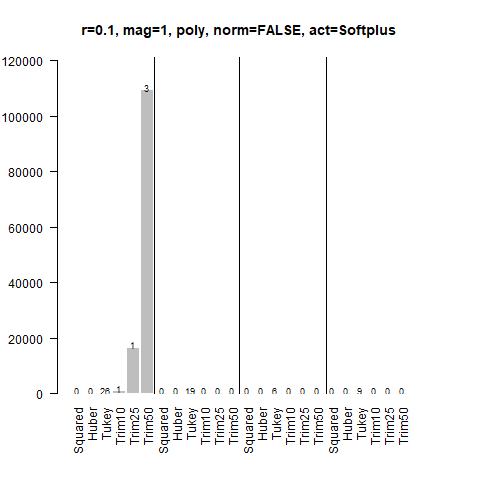}
\includegraphics[width=6.75cm,height=6.25cm]{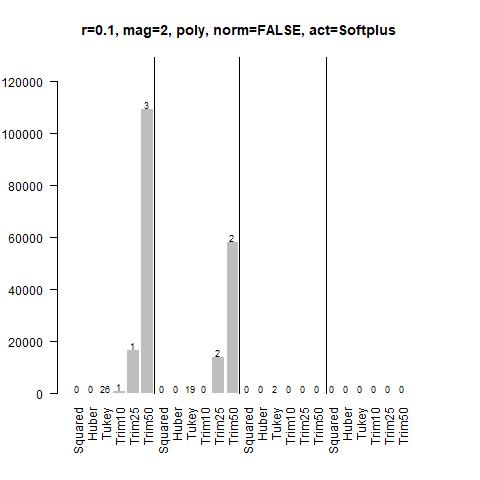} \\
\includegraphics[width=6.75cm,height=6.25cm]{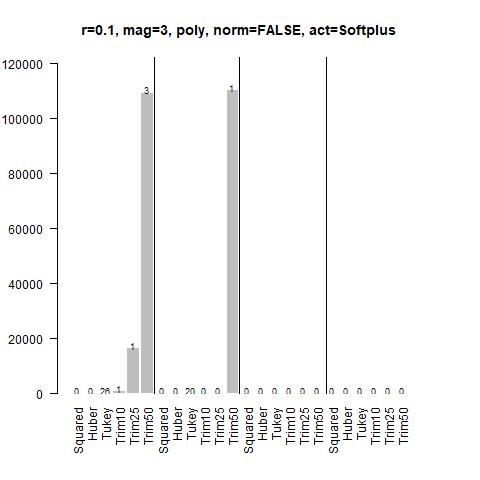} 
\includegraphics[width=6.75cm,height=6.25cm]{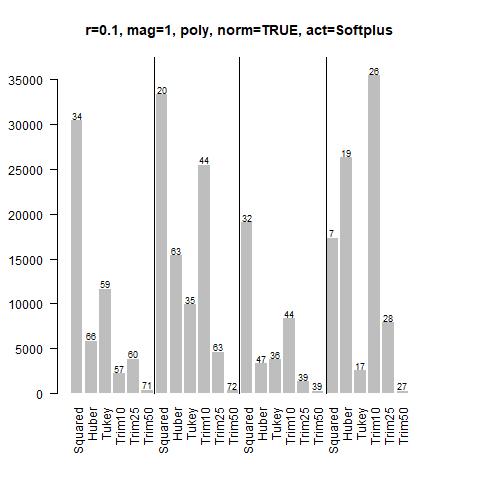}\\
\includegraphics[width=6.75cm,height=6.25cm]{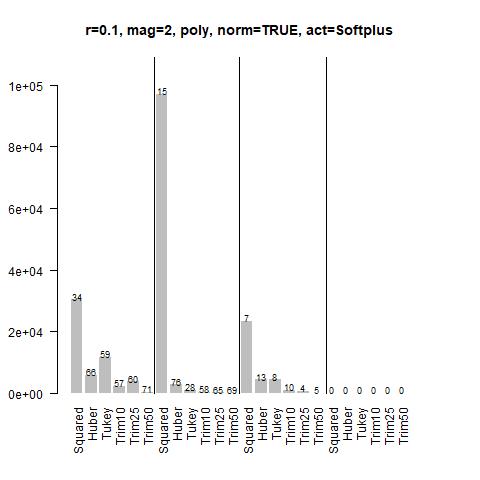} 
\includegraphics[width=6.75cm,height=6.25cm]{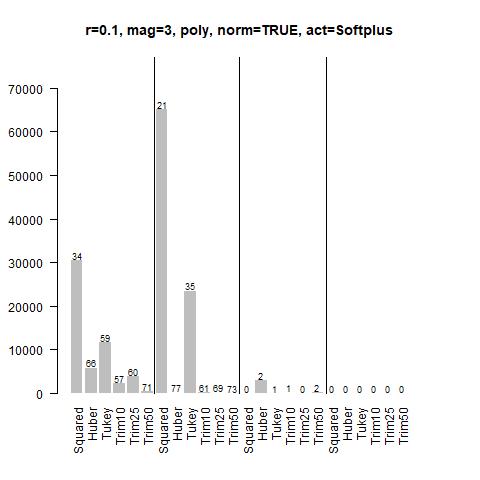} 
\end{center}
\caption{Results for $r=0.1$}
\end{figure}

\begin{figure}[H]
\label{trimnn:n500p20r25m1polynonreludeepStep}
\begin{center}
\includegraphics[width=6.75cm,height=6.25cm]{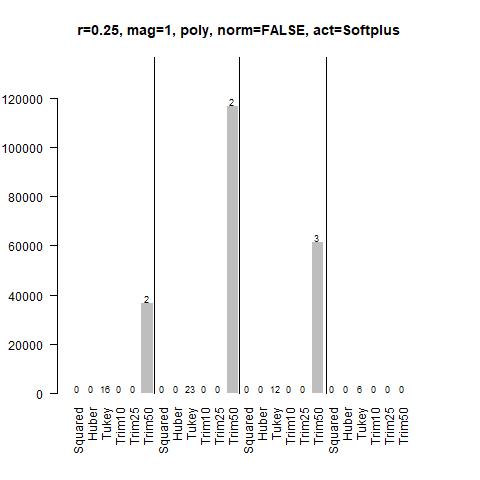}
\includegraphics[width=6.75cm,height=6.25cm]{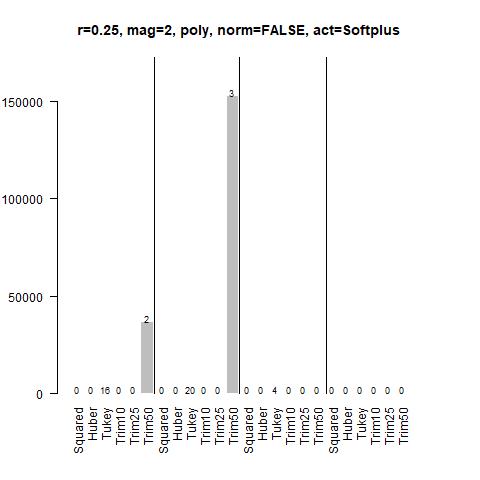} \\
\includegraphics[width=6.75cm,height=6.25cm]{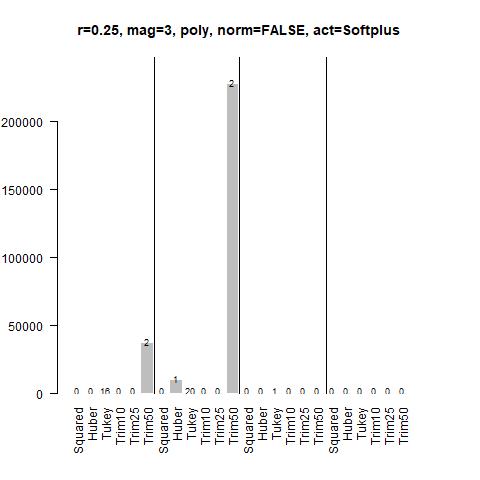} 
\includegraphics[width=6.75cm,height=6.25cm]{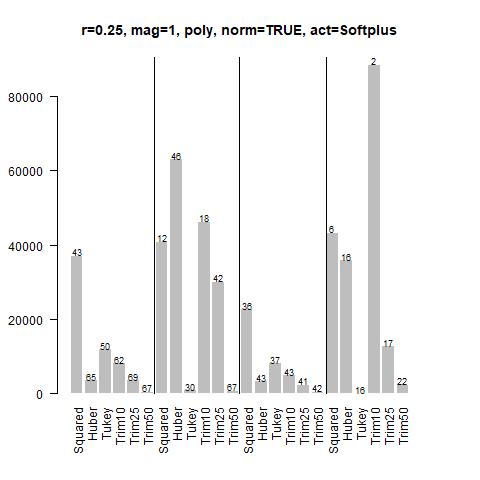}\\
\includegraphics[width=6.75cm,height=6.25cm]{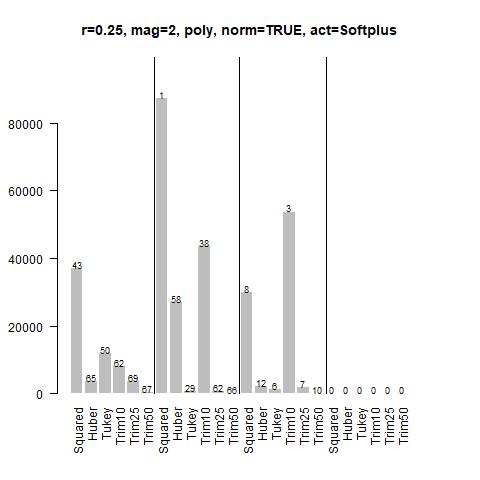} 
\includegraphics[width=6.75cm,height=6.25cm]{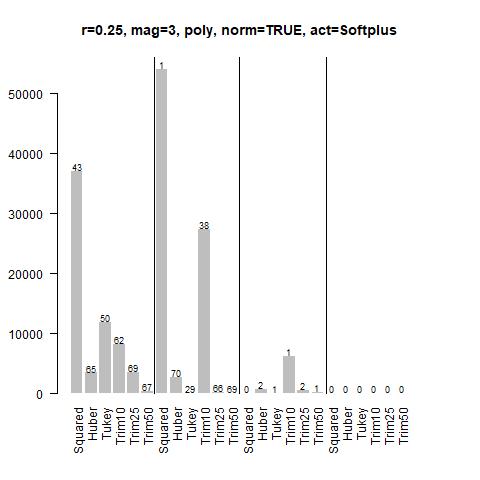} 
\end{center}
\caption{Results for $r=0.25$}
\end{figure}

\begin{figure}[H]
\label{trimnn:n500p20r40m1polynonreludeepStep}
\begin{center}
\includegraphics[width=6.75cm,height=6.25cm]{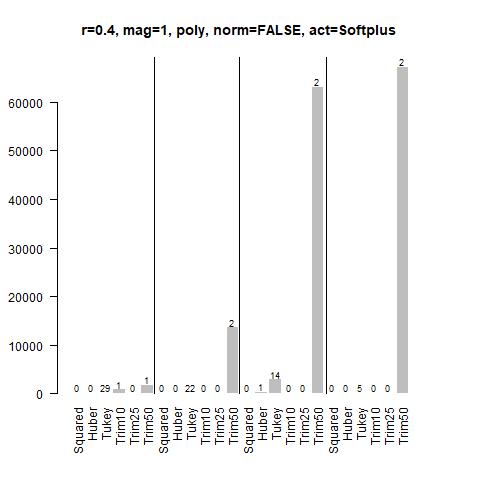}
\includegraphics[width=6.75cm,height=6.25cm]{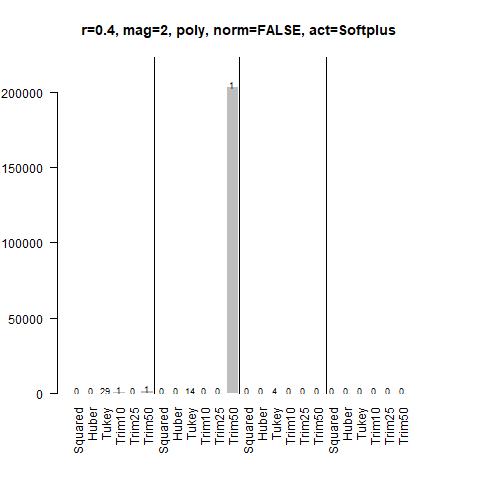} \\
\includegraphics[width=6.75cm,height=6.25cm]{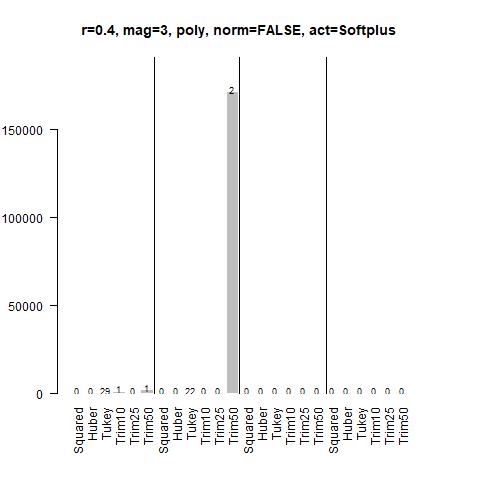} 
\includegraphics[width=6.75cm,height=6.25cm]{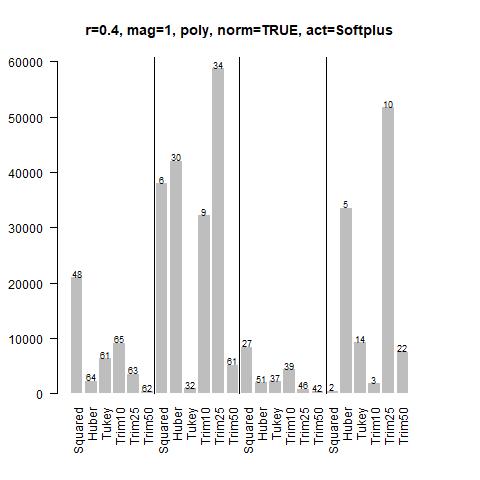}\\
\includegraphics[width=6.75cm,height=6.25cm]{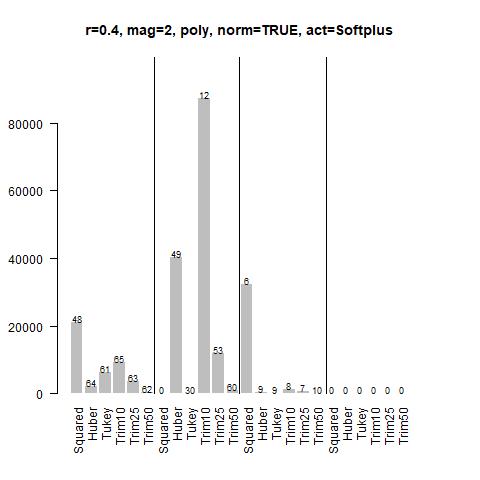} 
\includegraphics[width=6.75cm,height=6.25cm]{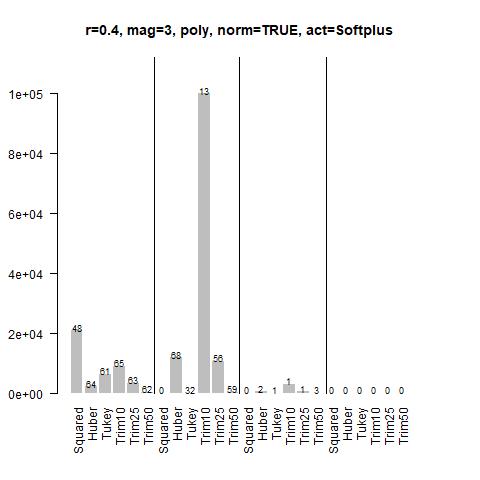} 
\end{center}
\caption{Results for $r=0.4$}
\end{figure}

\subsubsection{Trigonometric function}

\begin{figure}[H]
\label{trimnn:n500p20r10m1trignonreludeepStep}
\begin{center}
\includegraphics[width=6.75cm,height=6.25cm]{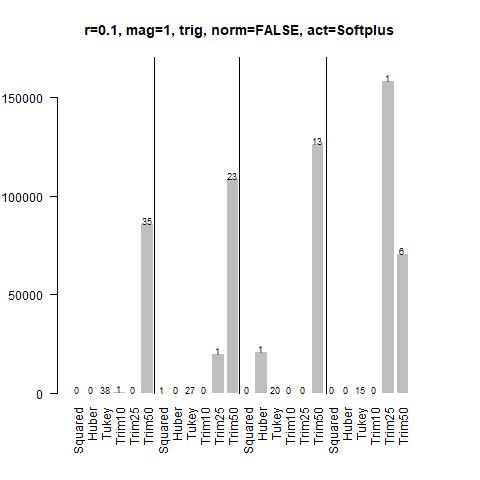}
\includegraphics[width=6.75cm,height=6.25cm]{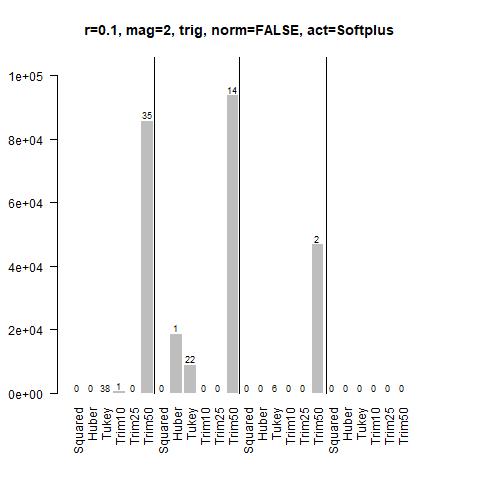} \\
\includegraphics[width=6.75cm,height=6.25cm]{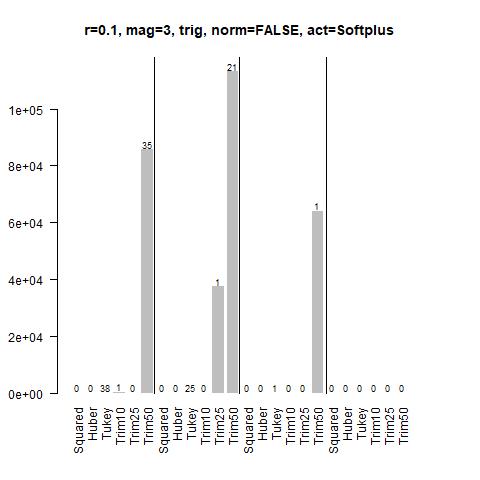} 
\includegraphics[width=6.75cm,height=6.25cm]{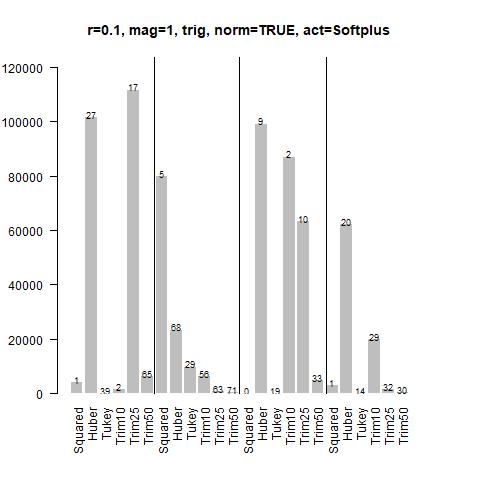}\\
\includegraphics[width=6.75cm,height=6.25cm]{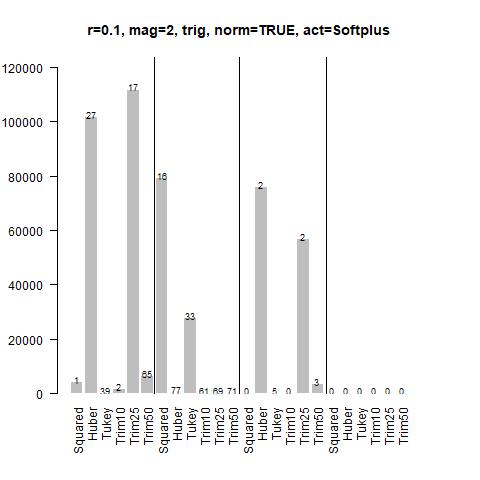} 
\includegraphics[width=6.75cm,height=6.25cm]{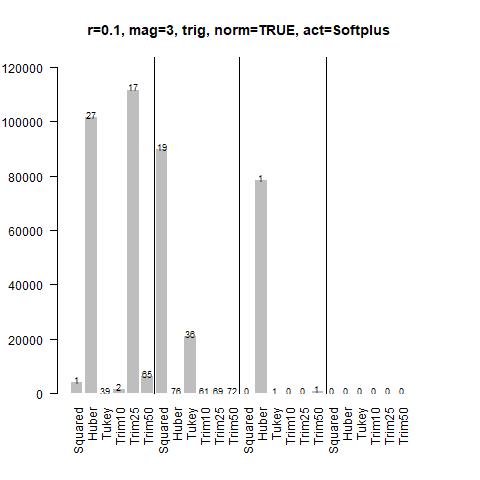} 
\end{center}
\caption{Results for $r=0.1$}
\end{figure}

\begin{figure}[H]
\label{trimnn:n500p20r25m1trignonreludeepStep}
\begin{center}
\includegraphics[width=6.75cm,height=6.25cm]{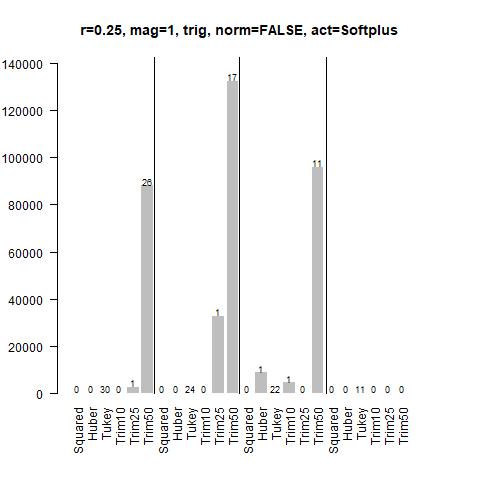}
\includegraphics[width=6.75cm,height=6.25cm]{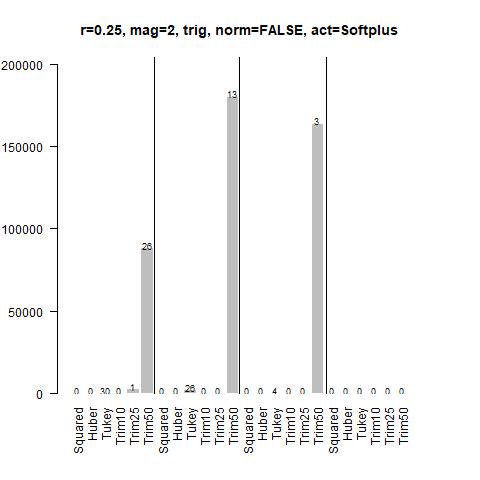} \\
\includegraphics[width=6.75cm,height=6.25cm]{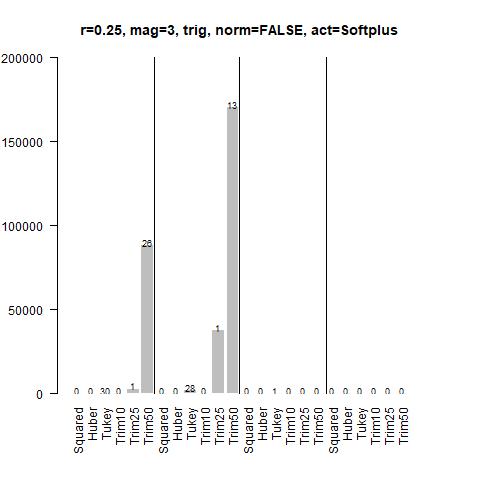} 
\includegraphics[width=6.75cm,height=6.25cm]{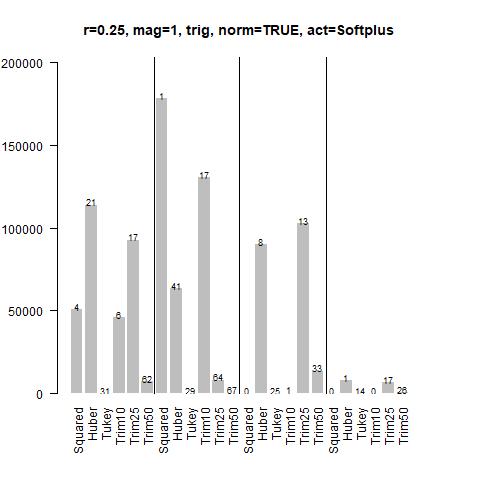}\\
\includegraphics[width=6.75cm,height=6.25cm]{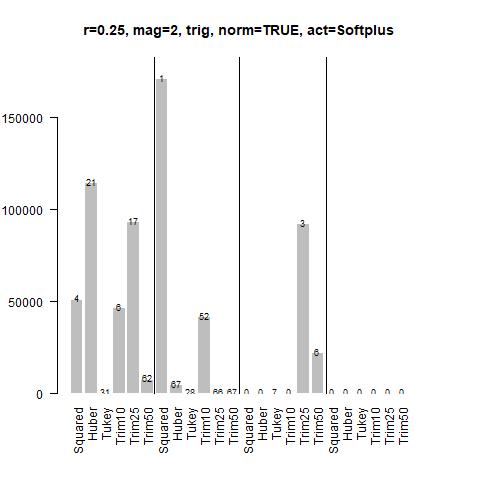} 
\includegraphics[width=6.75cm,height=6.25cm]{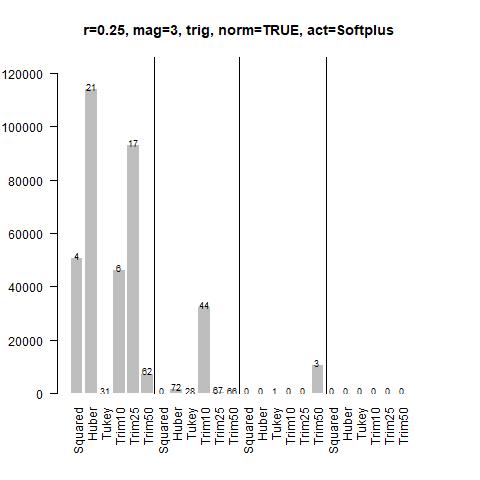} 
\end{center}
\caption{Results for $r=0.25$}
\end{figure}

\begin{figure}[H]
\label{trimnn:n500p20r40m1trignonreludeepStep}
\begin{center}
\includegraphics[width=6.75cm,height=6.25cm]{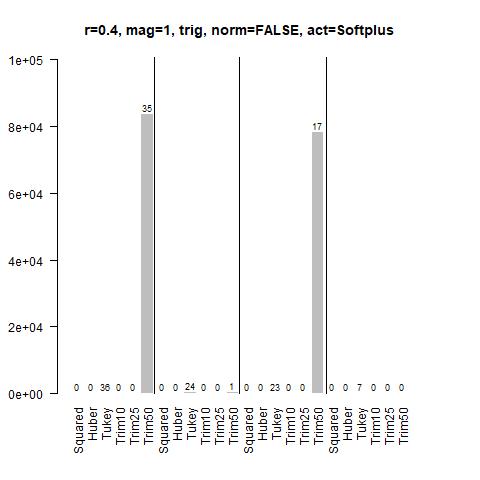}
\includegraphics[width=6.75cm,height=6.25cm]{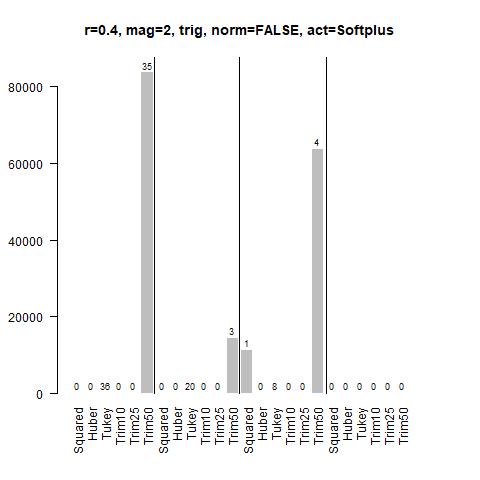} \\
\includegraphics[width=6.75cm,height=6.25cm]{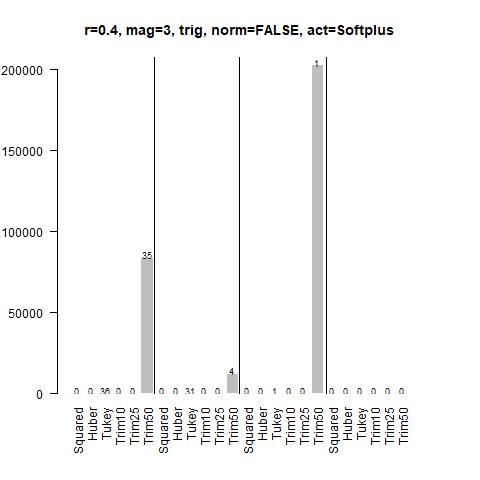} 
\includegraphics[width=6.75cm,height=6.25cm]{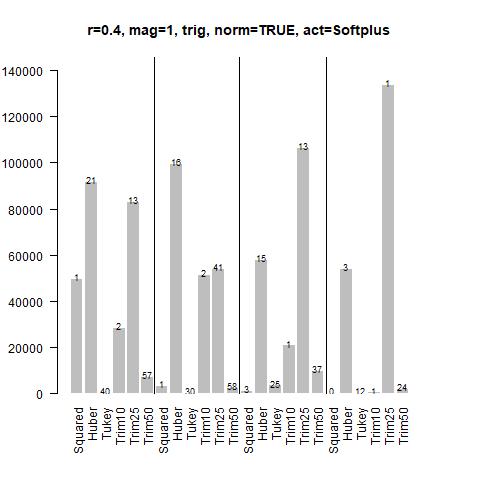}\\
\includegraphics[width=6.75cm,height=6.25cm]{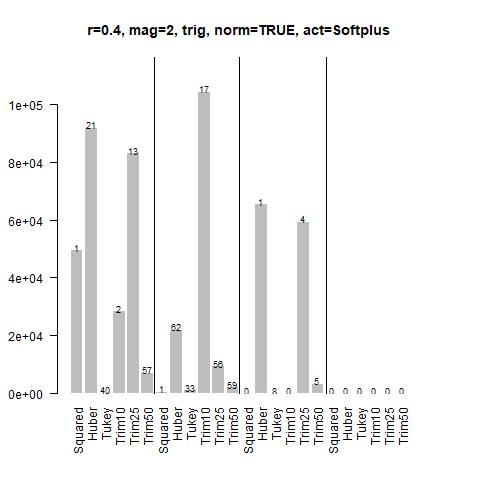} 
\includegraphics[width=6.75cm,height=6.25cm]{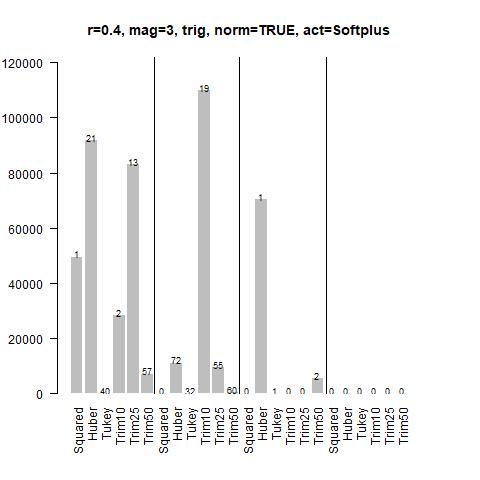} 
\end{center}
\caption{Results for $r=0.4$}
\end{figure}

\section{Simulation results for $n=1000$ and $p=50$: Training steps}  \label{trimnn:secstep100050}

\subsection{Logistic activation function}

\subsubsection{Linear function}

\begin{figure}[H]
\label{trimnn:n1000p50r10m1linnonlogStep}
\begin{center}
\includegraphics[width=6.75cm,height=6.25cm]{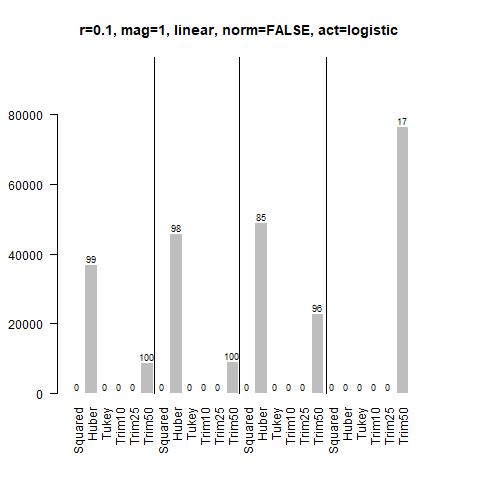}
\includegraphics[width=6.75cm,height=6.25cm]{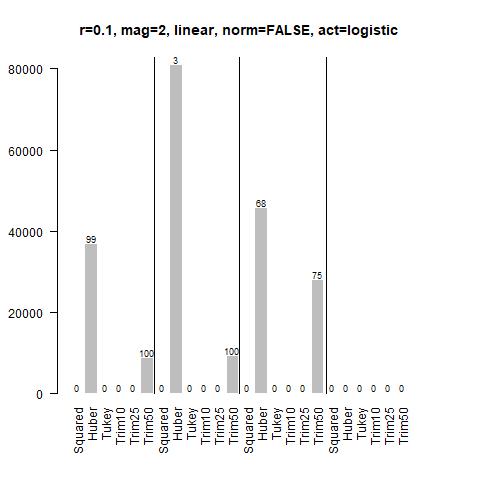} \\
\includegraphics[width=6.75cm,height=6.25cm]{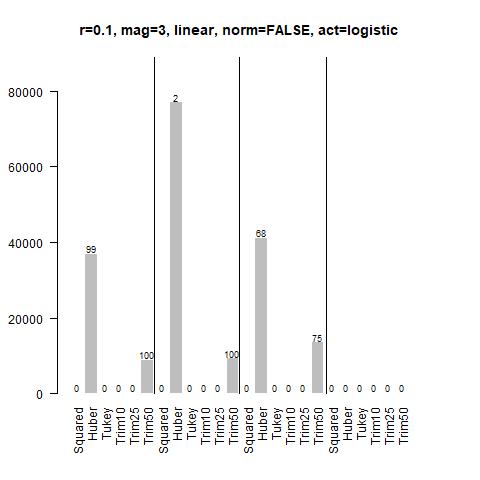} 
\includegraphics[width=6.75cm,height=6.25cm]{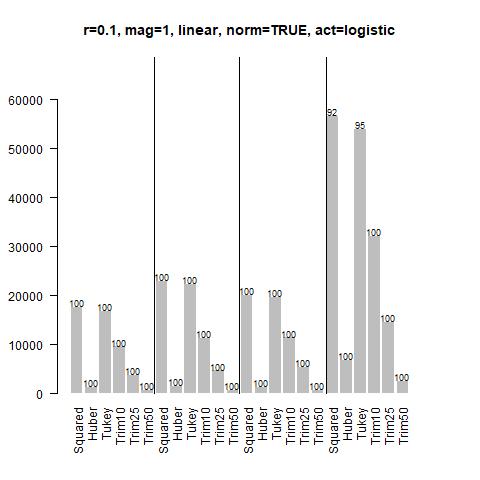}\\
\includegraphics[width=6.75cm,height=6.25cm]{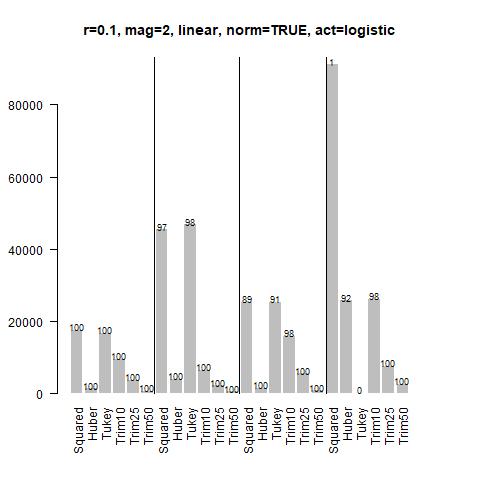} 
\includegraphics[width=6.75cm,height=6.25cm]{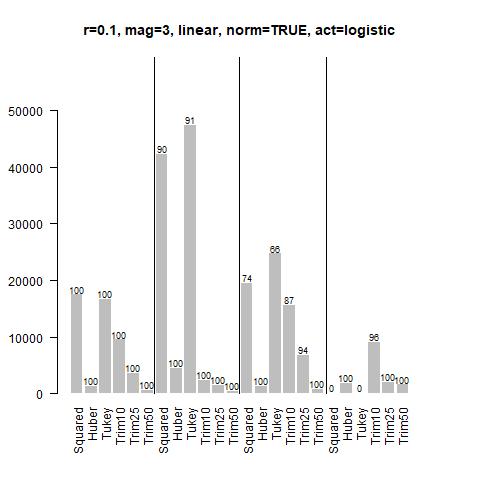} 
\end{center}
\caption{Results for $r=0.1$}
\end{figure}

\begin{figure}[H]
\label{trimnn:n1000p50r25m1linnonlogStep}
\begin{center}
\includegraphics[width=6.75cm,height=6.25cm]{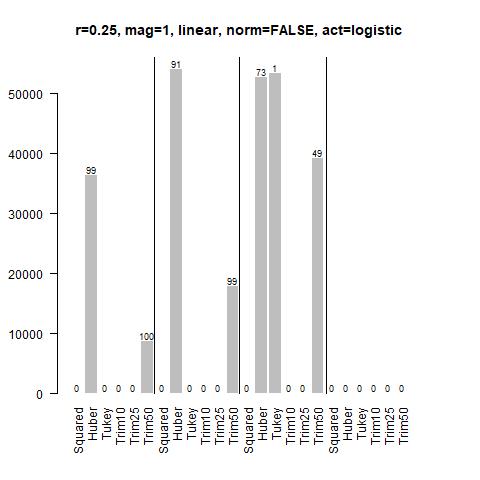}
\includegraphics[width=6.75cm,height=6.25cm]{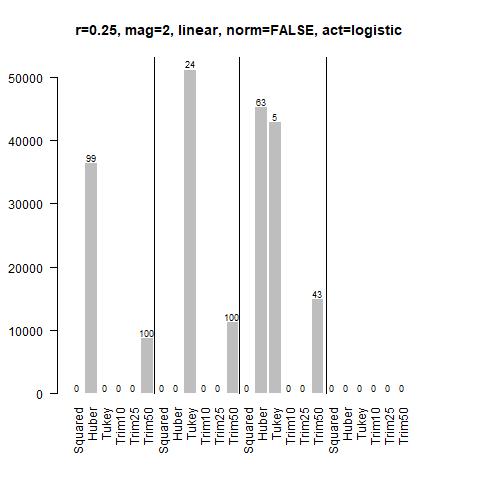} \\
\includegraphics[width=6.75cm,height=6.25cm]{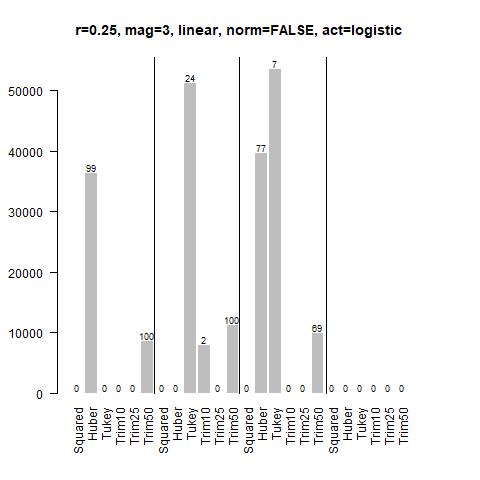} 
\includegraphics[width=6.75cm,height=6.25cm]{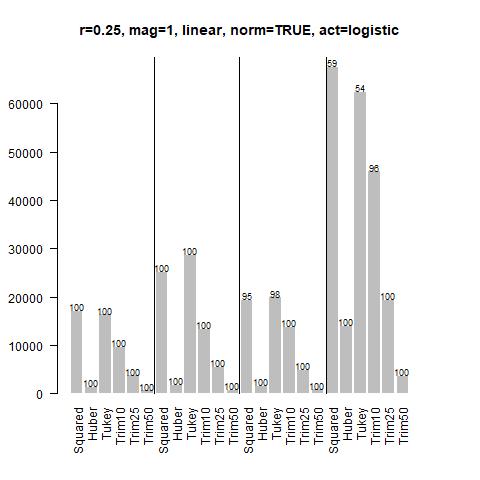}\\
\includegraphics[width=6.75cm,height=6.25cm]{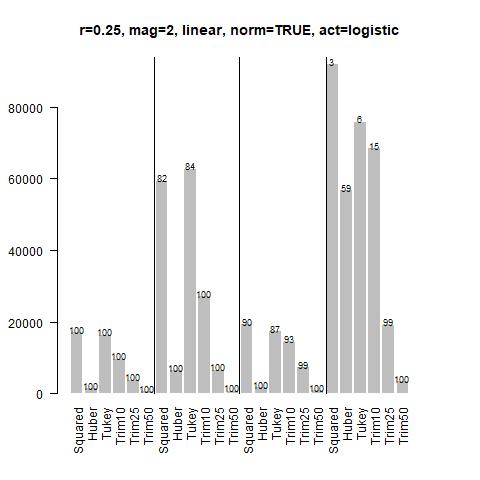} 
\includegraphics[width=6.75cm,height=6.25cm]{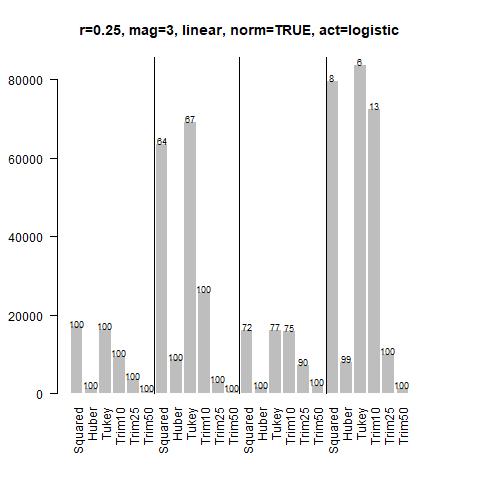} 
\end{center}
\caption{Results for $r=0.25$}
\end{figure}

\begin{figure}[H]
\label{trimnn:n1000p50r40m1linnonlogStep}
\begin{center}
\includegraphics[width=6.75cm,height=6.25cm]{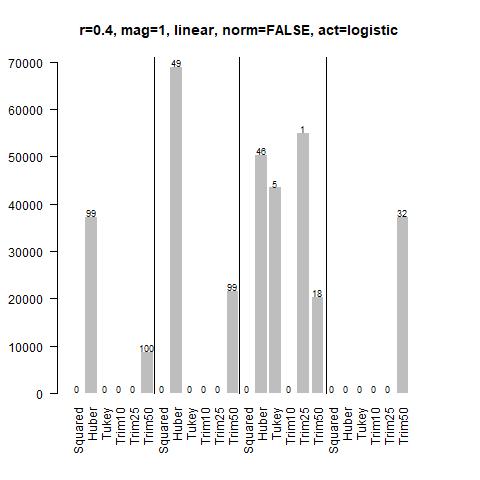}
\includegraphics[width=6.75cm,height=6.25cm]{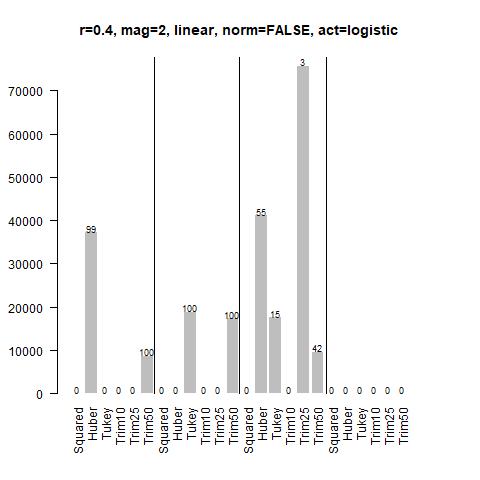} \\
\includegraphics[width=6.75cm,height=6.25cm]{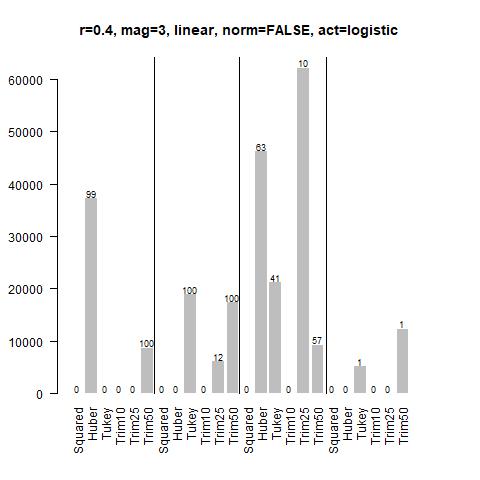} 
\includegraphics[width=6.75cm,height=6.25cm]{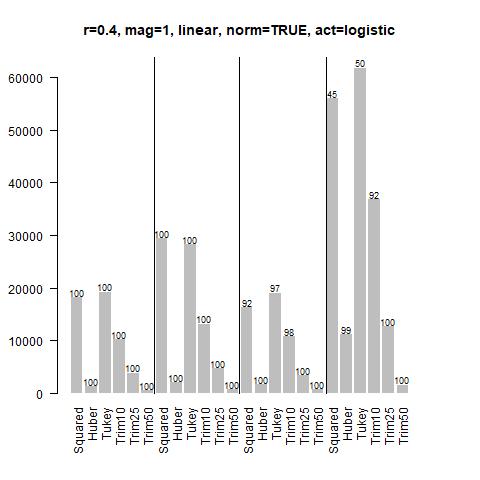}\\
\includegraphics[width=6.75cm,height=6.25cm]{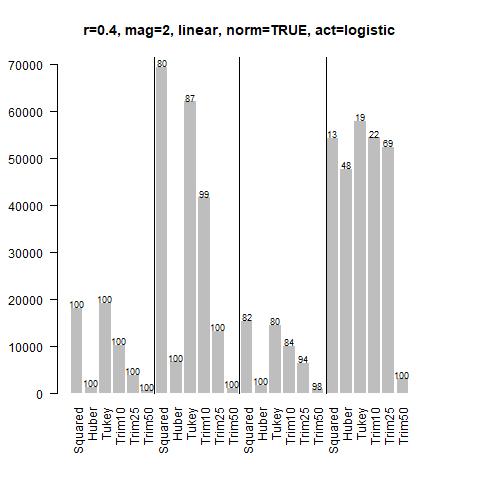} 
\includegraphics[width=6.75cm,height=6.25cm]{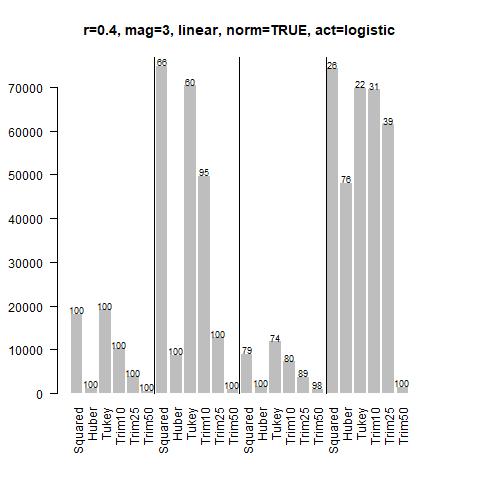} 
\end{center}
\caption{Results for $r=0.4$}
\end{figure}

\subsubsection{Polynomial function}

\begin{figure}[H]
\label{trimnn:n1000p50r10m1polynonlogStep}
\begin{center}
\includegraphics[width=6.75cm,height=6.25cm]{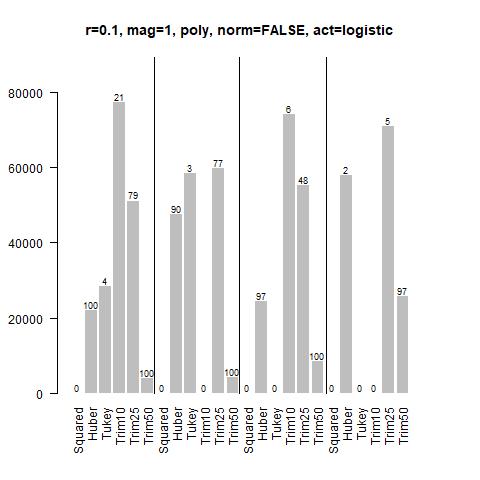}
\includegraphics[width=6.75cm,height=6.25cm]{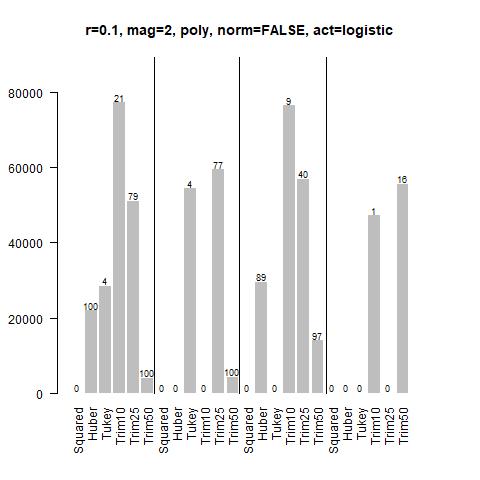} \\
\includegraphics[width=6.75cm,height=6.25cm]{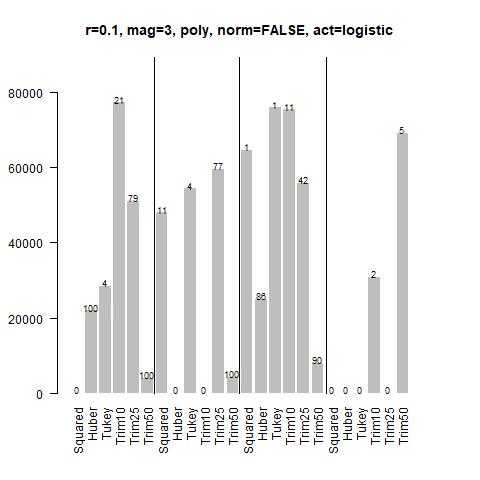} 
\includegraphics[width=6.75cm,height=6.25cm]{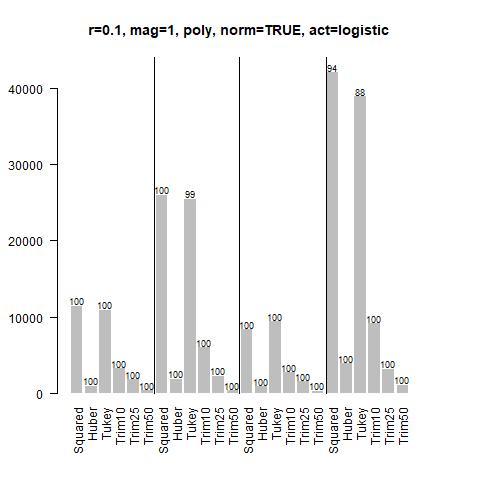}\\
\includegraphics[width=6.75cm,height=6.25cm]{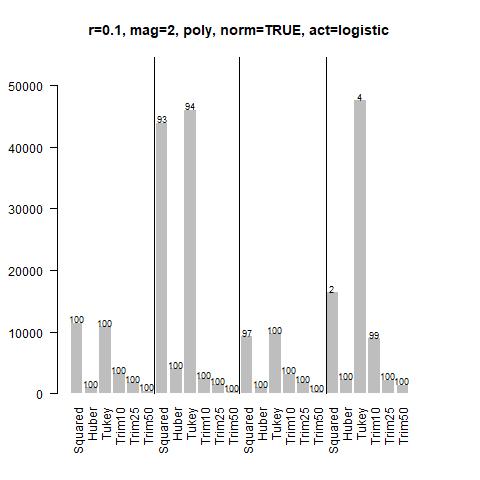} 
\includegraphics[width=6.75cm,height=6.25cm]{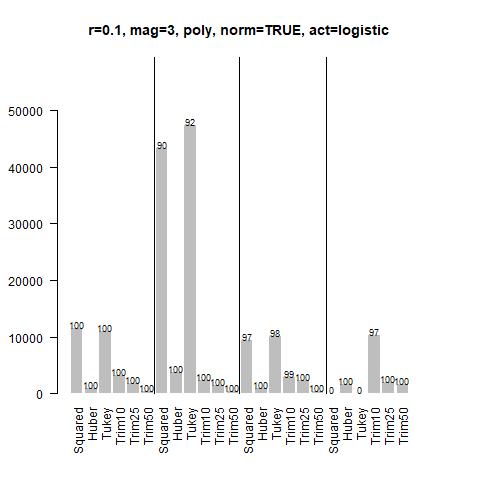} 
\end{center}
\caption{Results for $r=0.1$}
\end{figure}

\begin{figure}[H]
\label{trimnn:n1000p50r25m1polynonlogStep}
\begin{center}
\includegraphics[width=6.75cm,height=6.25cm]{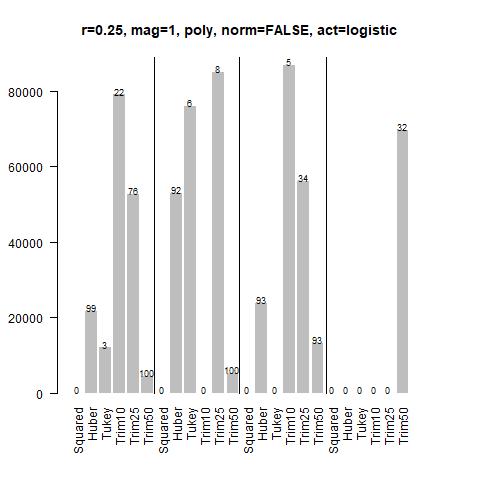}
\includegraphics[width=6.75cm,height=6.25cm]{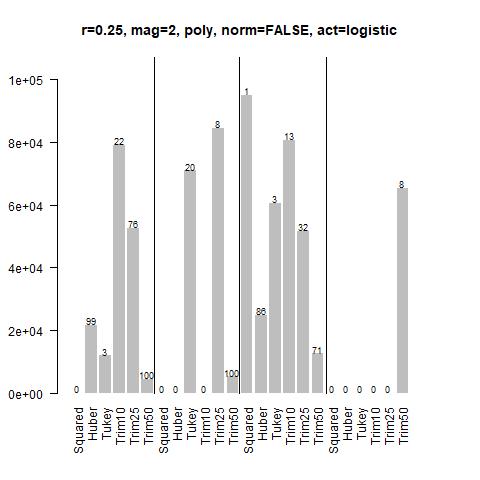} \\
\includegraphics[width=6.75cm,height=6.25cm]{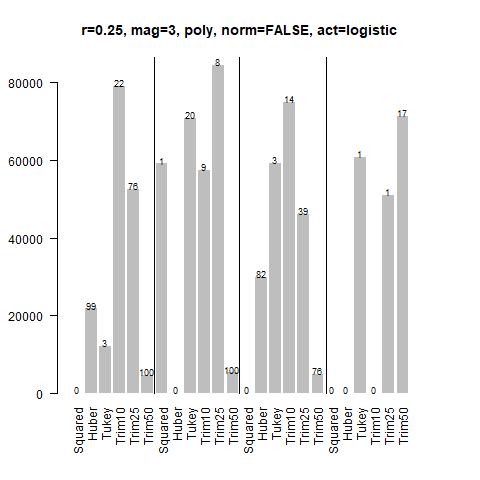} 
\includegraphics[width=6.75cm,height=6.25cm]{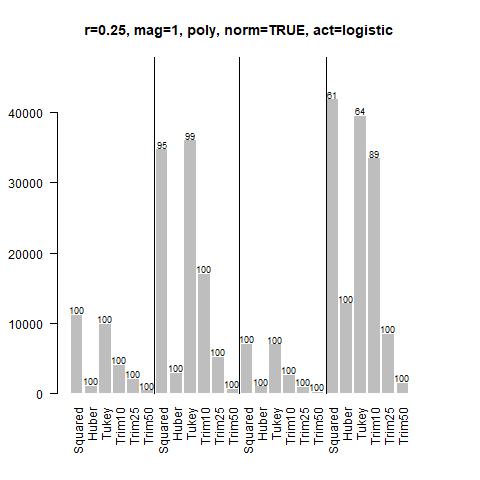}\\
\includegraphics[width=6.75cm,height=6.25cm]{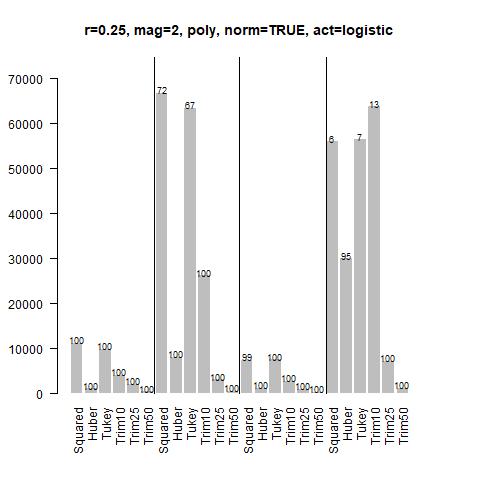} 
\includegraphics[width=6.75cm,height=6.25cm]{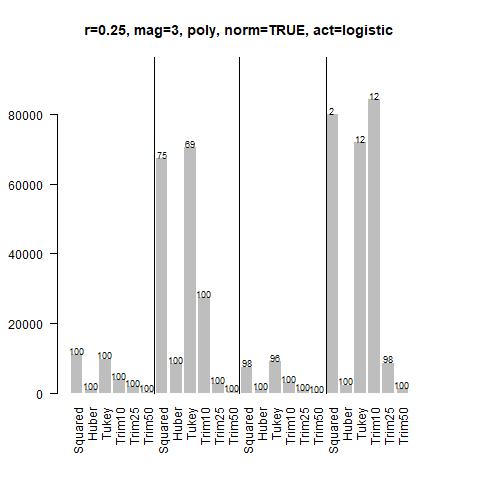} 
\end{center}
\caption{Results for $r=0.25$}
\end{figure}

\begin{figure}[H]
\label{trimnn:n1000p50r40m1polynonlogStep}
\begin{center}
\includegraphics[width=6.75cm,height=6.25cm]{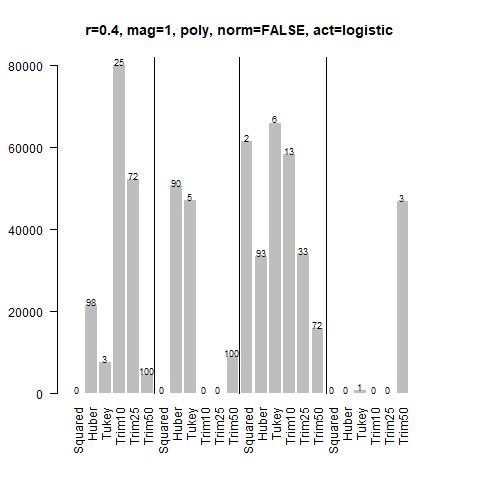}
\includegraphics[width=6.75cm,height=6.25cm]{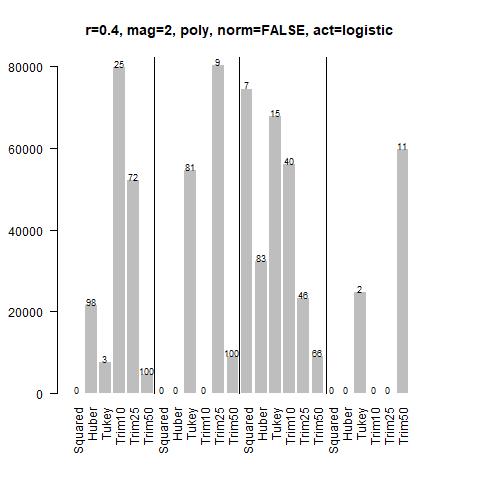} \\
\includegraphics[width=6.75cm,height=6.25cm]{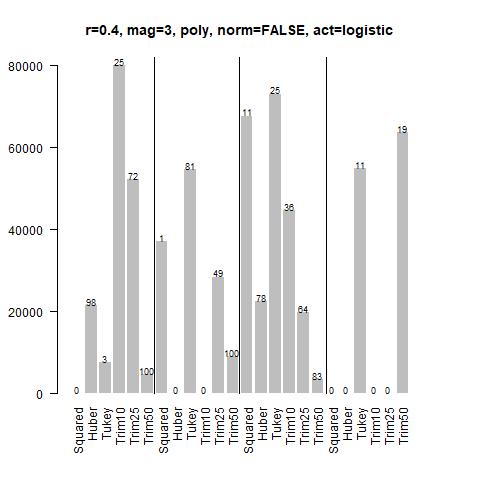} 
\includegraphics[width=6.75cm,height=6.25cm]{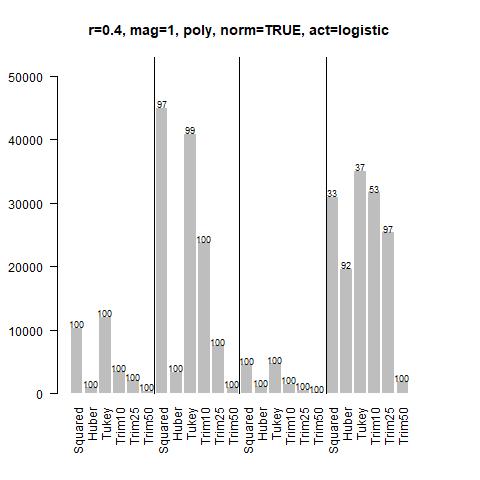}\\
\includegraphics[width=6.75cm,height=6.25cm]{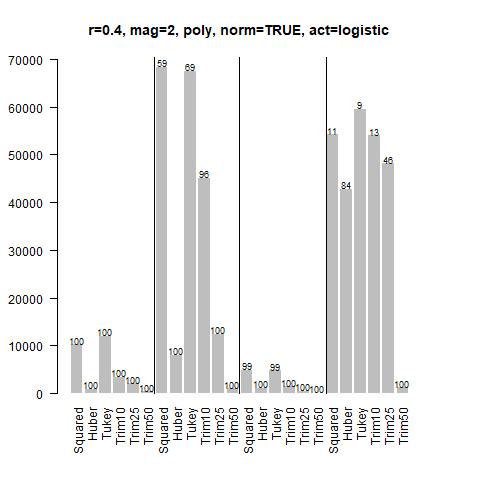} 
\includegraphics[width=6.75cm,height=6.25cm]{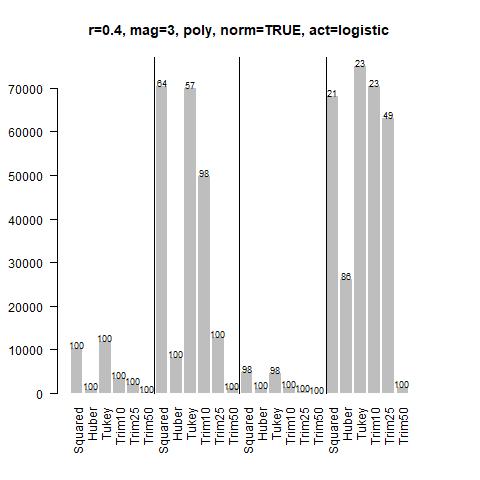} 
\end{center}
\caption{Results for $r=0.4$}
\end{figure}

\subsubsection{Trigonometric function}

\begin{figure}[H]
\label{trimnn:n1000p50r10m1trignonlogStep}
\begin{center}
\includegraphics[width=6.75cm,height=6.25cm]{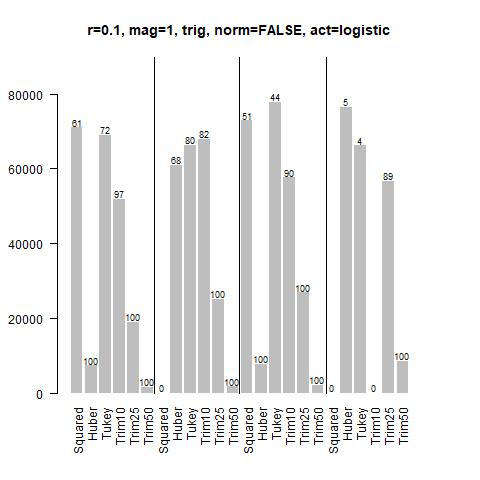}
\includegraphics[width=6.75cm,height=6.25cm]{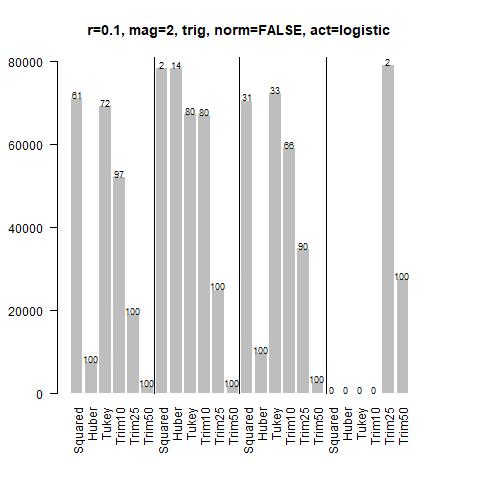} \\
\includegraphics[width=6.75cm,height=6.25cm]{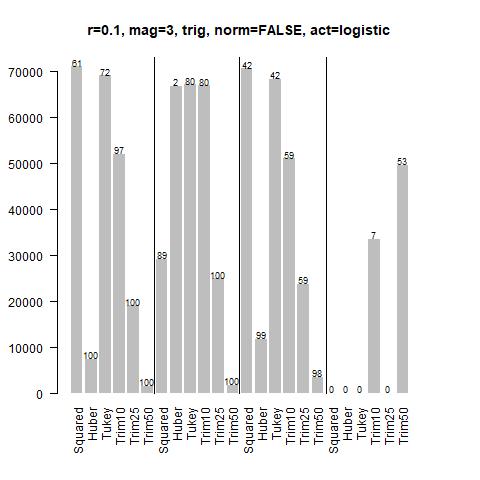} 
\includegraphics[width=6.75cm,height=6.25cm]{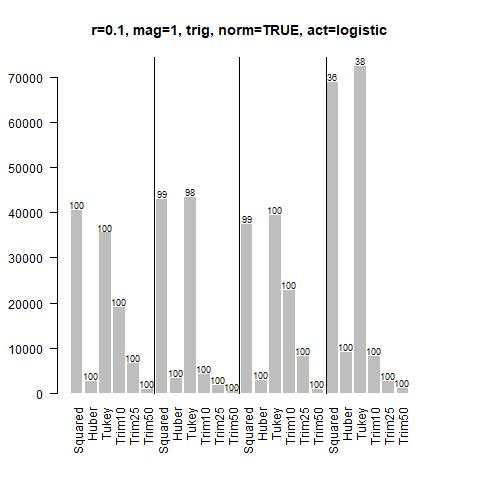}\\
\includegraphics[width=6.75cm,height=6.25cm]{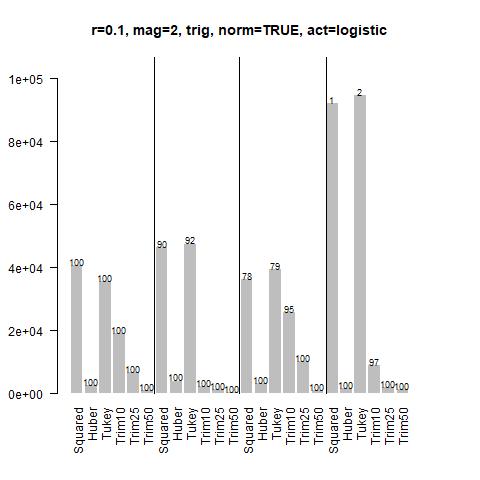} 
\includegraphics[width=6.75cm,height=6.25cm]{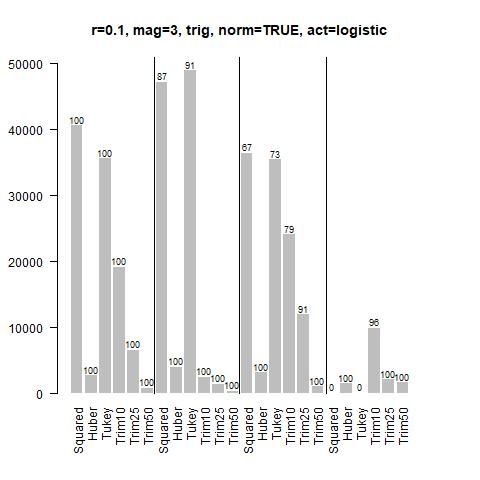} 
\end{center}
\caption{Results for $r=0.1$}
\end{figure}

\begin{figure}[H]
\label{trimnn:n1000p50r25m1trignonlogStep}
\begin{center}
\includegraphics[width=6.75cm,height=6.25cm]{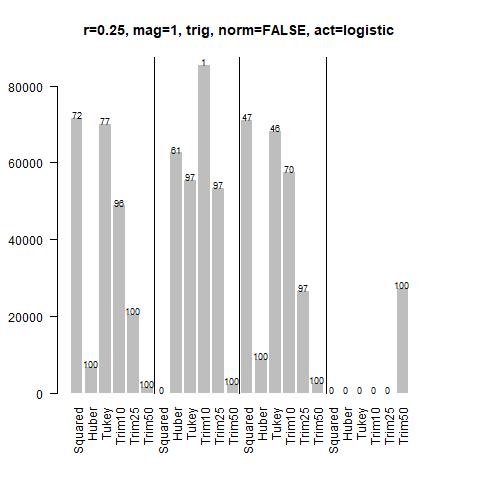}
\includegraphics[width=6.75cm,height=6.25cm]{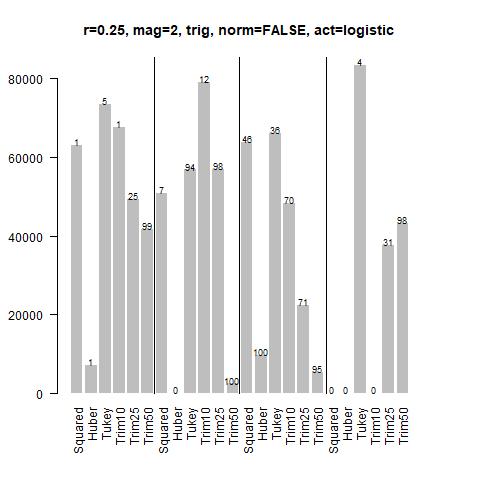} \\
\includegraphics[width=6.75cm,height=6.25cm]{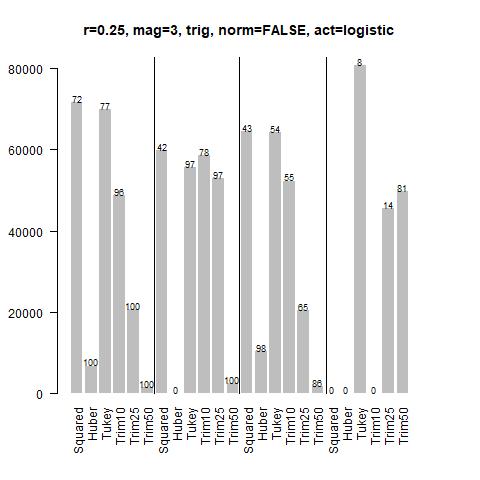} 
\includegraphics[width=6.75cm,height=6.25cm]{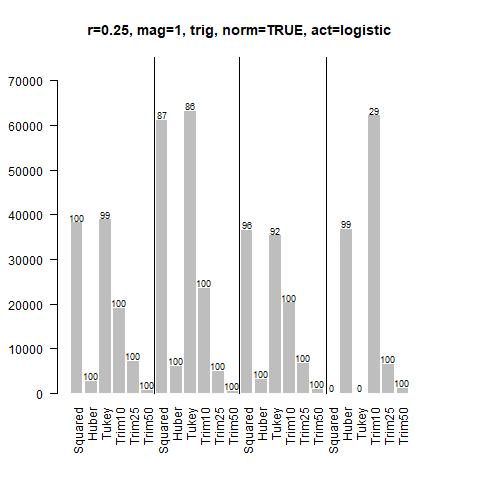}\\
\includegraphics[width=6.75cm,height=6.25cm]{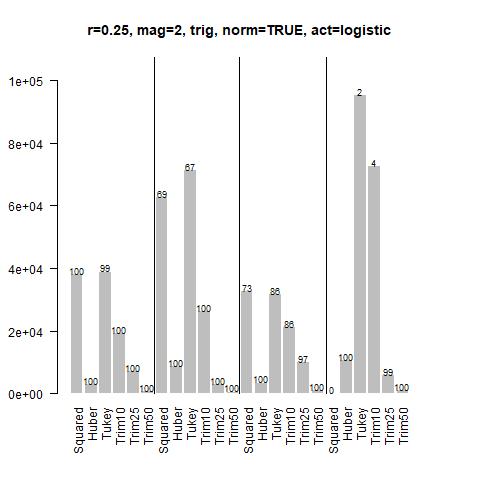} 
\includegraphics[width=6.75cm,height=6.25cm]{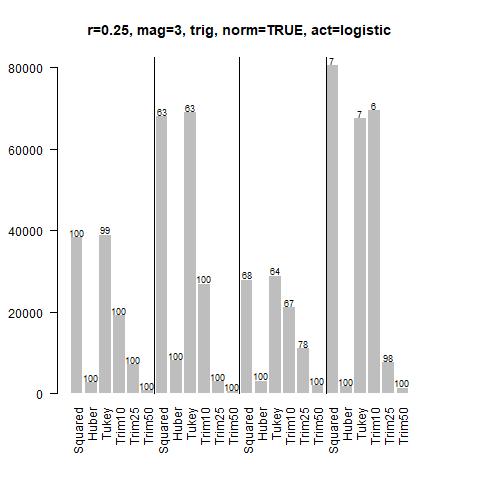} 
\end{center}
\caption{Results for $r=0.25$}
\end{figure}

\begin{figure}[H]
\label{trimnn:n1000p50r40m1trignonlogStep}
\begin{center}
\includegraphics[width=6.75cm,height=6.25cm]{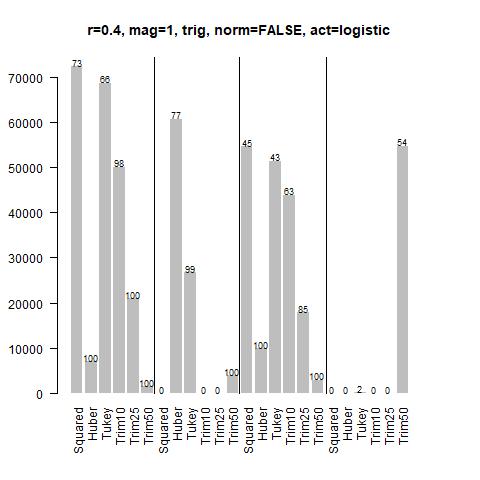}
\includegraphics[width=6.75cm,height=6.25cm]{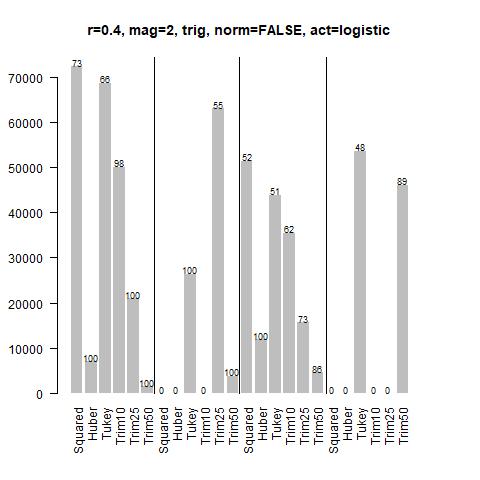} \\
\includegraphics[width=6.75cm,height=6.25cm]{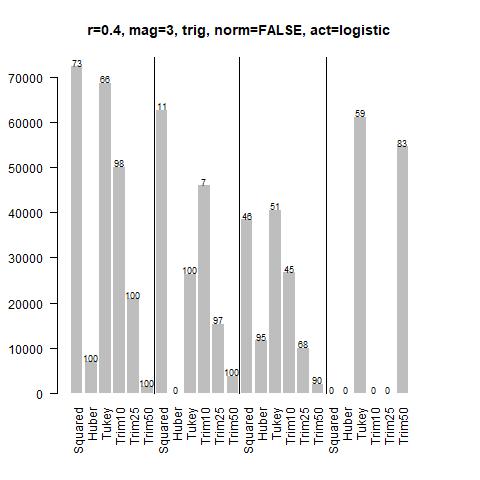} 
\includegraphics[width=6.75cm,height=6.25cm]{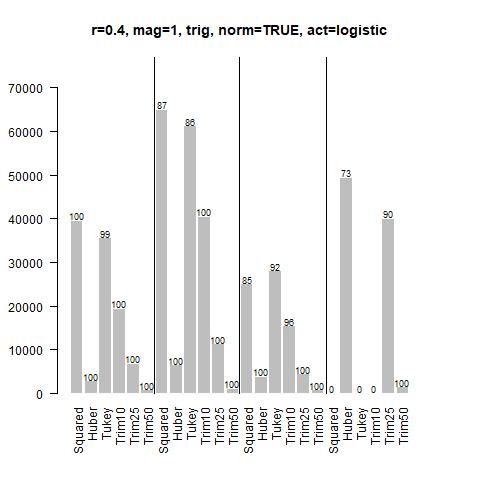}\\
\includegraphics[width=6.75cm,height=6.25cm]{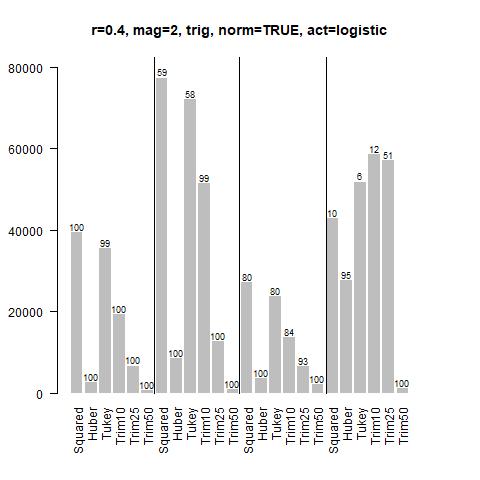} 
\includegraphics[width=6.75cm,height=6.25cm]{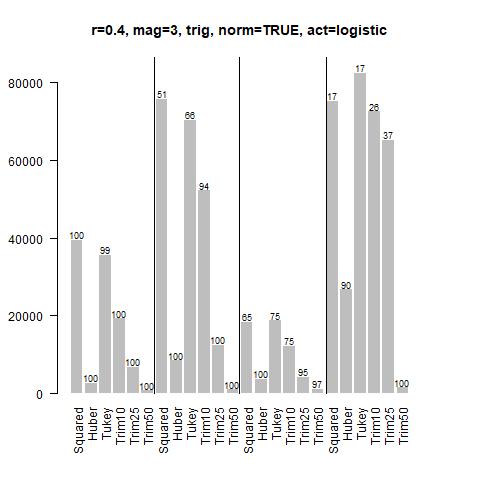} 
\end{center}
\caption{Results for $r=0.4$}
\end{figure}

\subsection{Softplus activation function}

\subsubsection{Linear function}

\begin{figure}[H]
\label{trimnn:n1000p50r10m1linnonreluStep}
\begin{center}
\includegraphics[width=6.75cm,height=6.25cm]{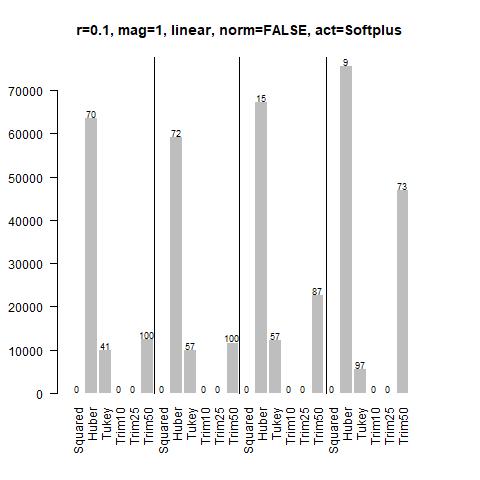}
\includegraphics[width=6.75cm,height=6.25cm]{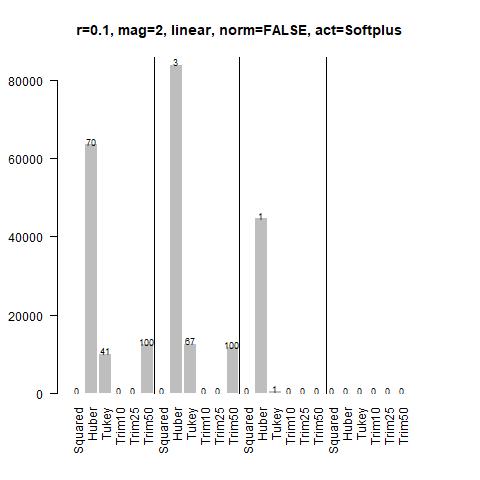} \\
\includegraphics[width=6.75cm,height=6.25cm]{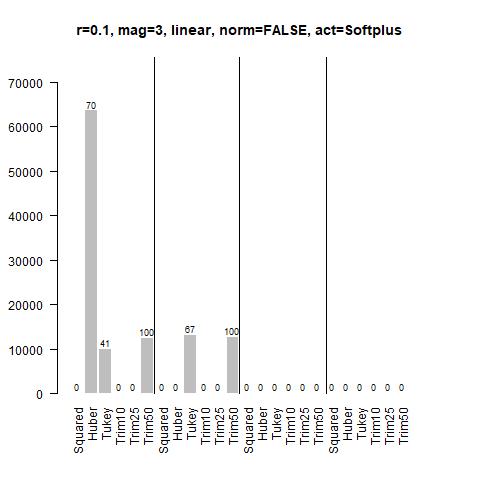} 
\includegraphics[width=6.75cm,height=6.25cm]{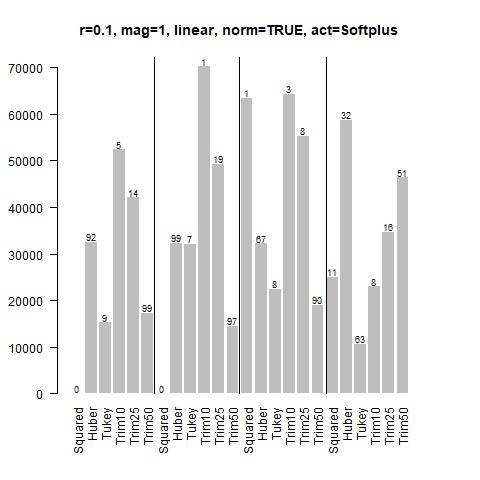}\\
\includegraphics[width=6.75cm,height=6.25cm]{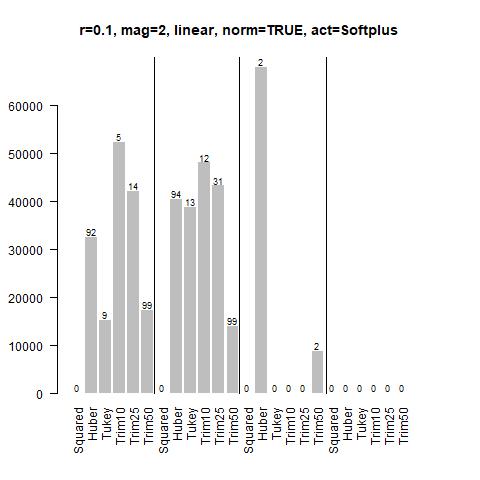} 
\includegraphics[width=6.75cm,height=6.25cm]{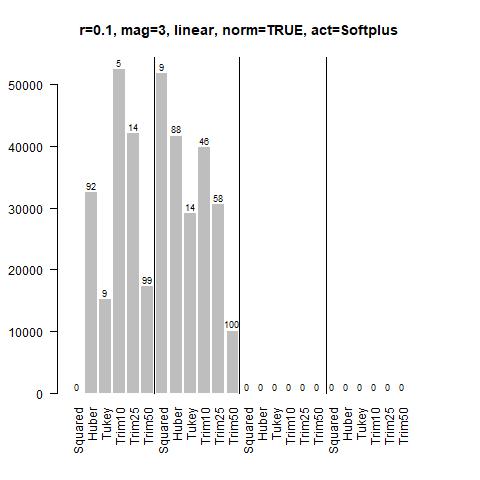} 
\end{center}
\caption{Results for $r=0.1$}
\end{figure}

\begin{figure}[H]
\label{trimnn:n1000p50r25m1linnonreluStep}
\begin{center}
\includegraphics[width=6.75cm,height=6.25cm]{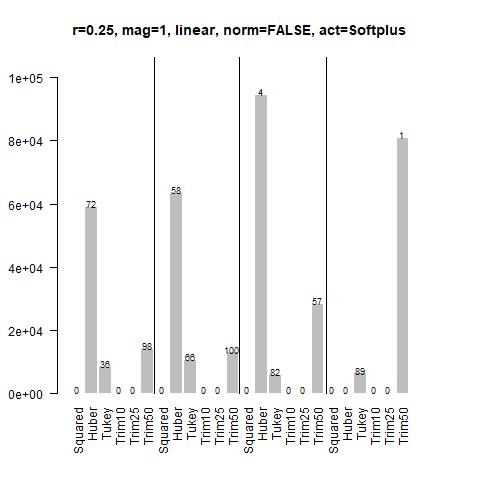}
\includegraphics[width=6.75cm,height=6.25cm]{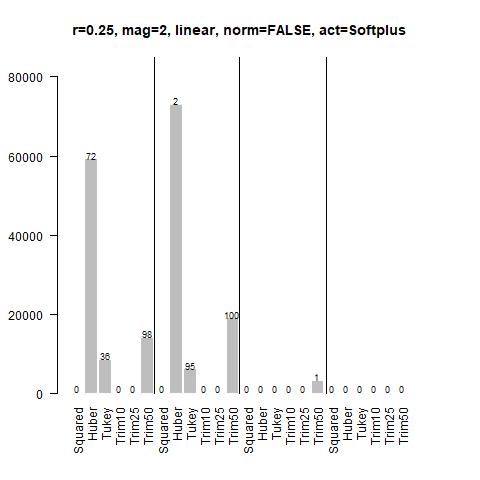} \\
\includegraphics[width=6.75cm,height=6.25cm]{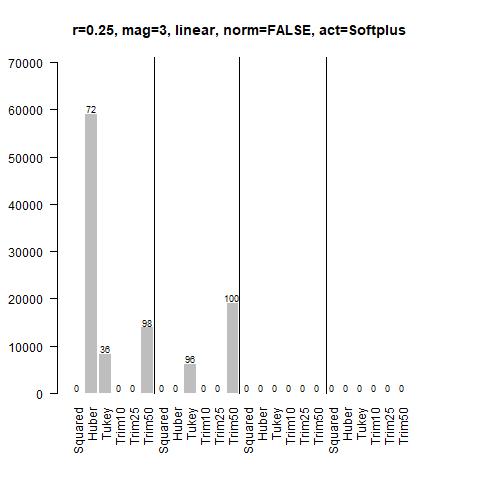} 
\includegraphics[width=6.75cm,height=6.25cm]{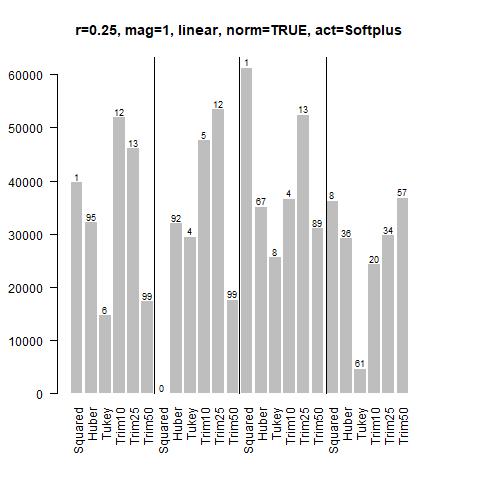}\\
\includegraphics[width=6.75cm,height=6.25cm]{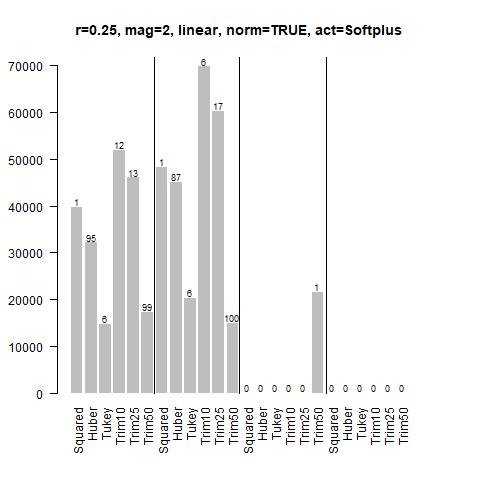} 
\includegraphics[width=6.75cm,height=6.25cm]{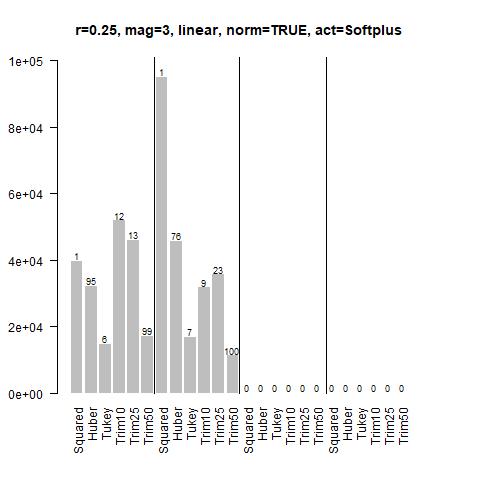} 
\end{center}
\caption{Results for $r=0.25$}
\end{figure}

\begin{figure}[H]
\label{trimnn:n1000p50r40m1linnonreluStep}
\begin{center}
\includegraphics[width=6.75cm,height=6.25cm]{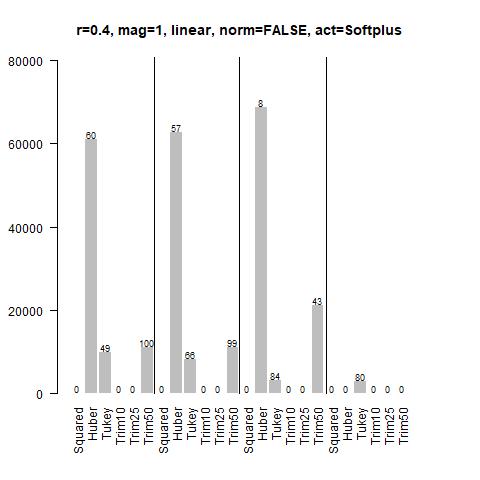}
\includegraphics[width=6.75cm,height=6.25cm]{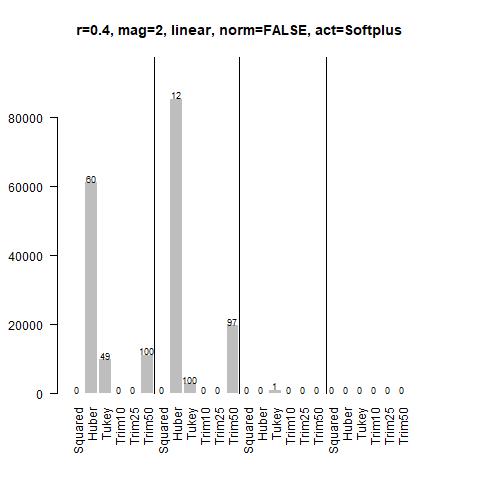} \\
\includegraphics[width=6.75cm,height=6.25cm]{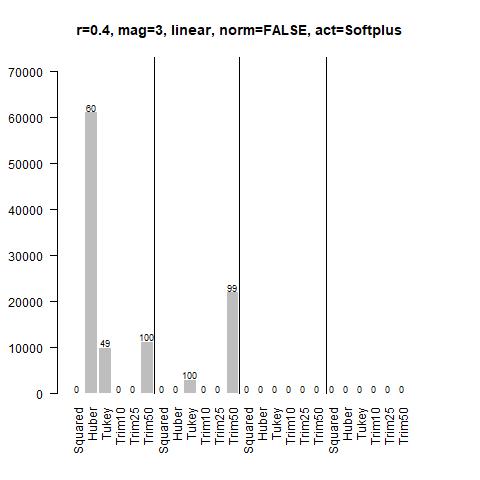} 
\includegraphics[width=6.75cm,height=6.25cm]{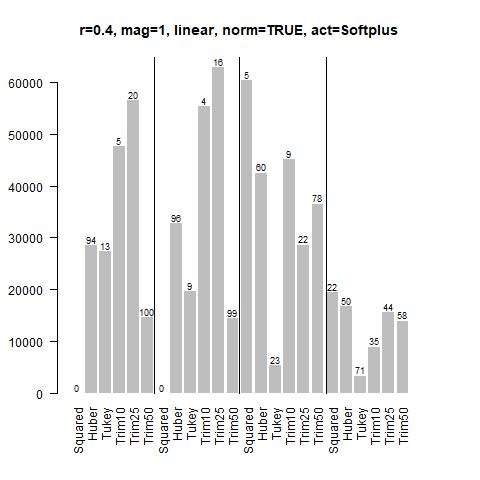}\\
\includegraphics[width=6.75cm,height=6.25cm]{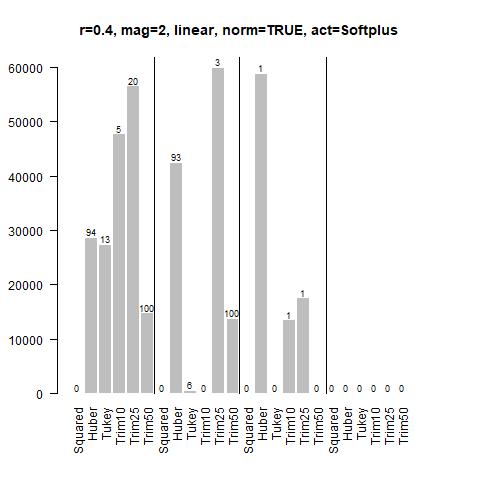} 
\includegraphics[width=6.75cm,height=6.25cm]{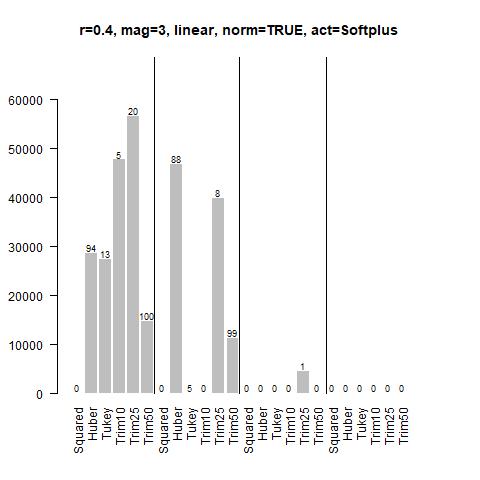} 
\end{center}
\caption{Results for $r=0.4$}
\end{figure}

\subsubsection{Polynomial function}

\begin{figure}[H]
\label{trimnn:n1000p50r10m1polynonreluStep}
\begin{center}
\includegraphics[width=6.75cm,height=6.25cm]{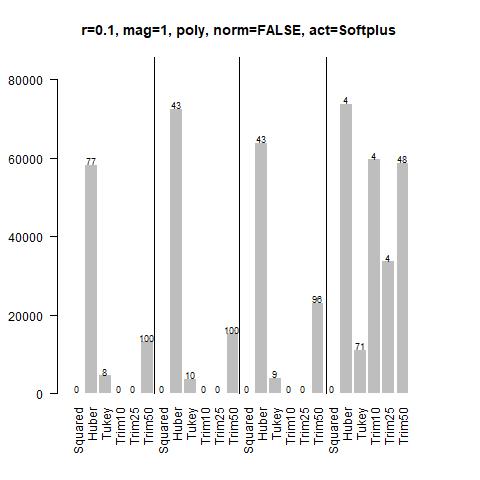}
\includegraphics[width=6.75cm,height=6.25cm]{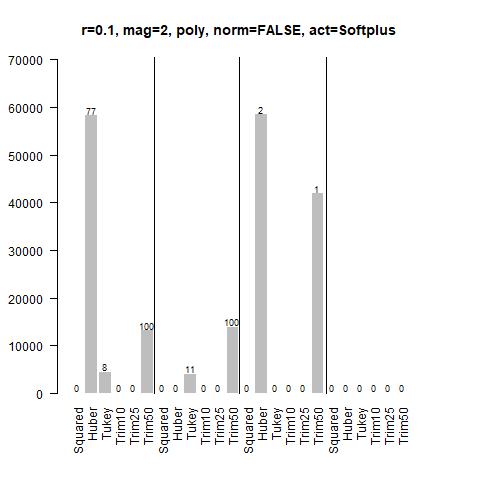} \\
\includegraphics[width=6.75cm,height=6.25cm]{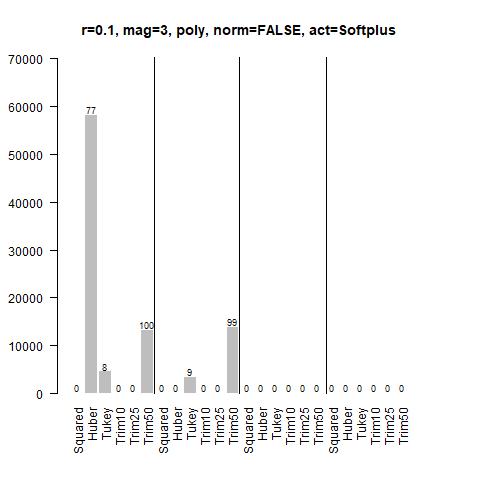} 
\includegraphics[width=6.75cm,height=6.25cm]{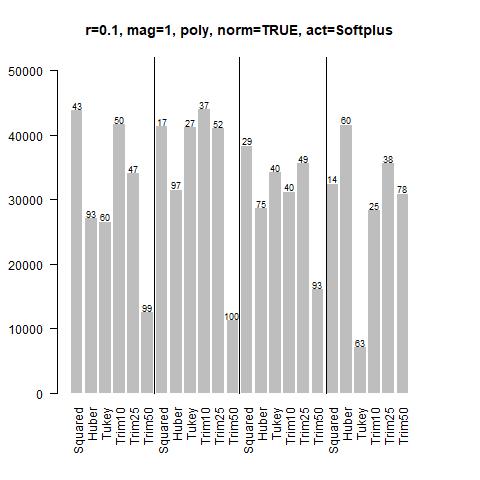}\\
\includegraphics[width=6.75cm,height=6.25cm]{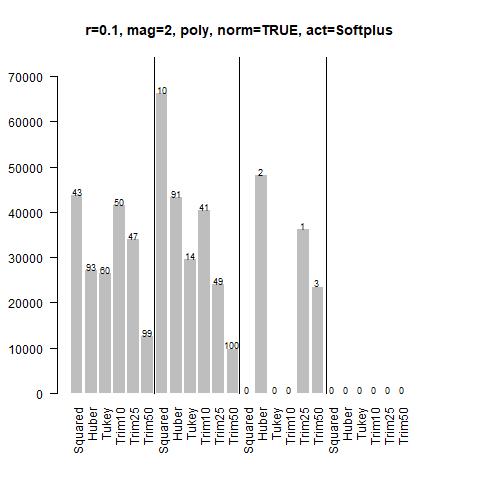} 
\includegraphics[width=6.75cm,height=6.25cm]{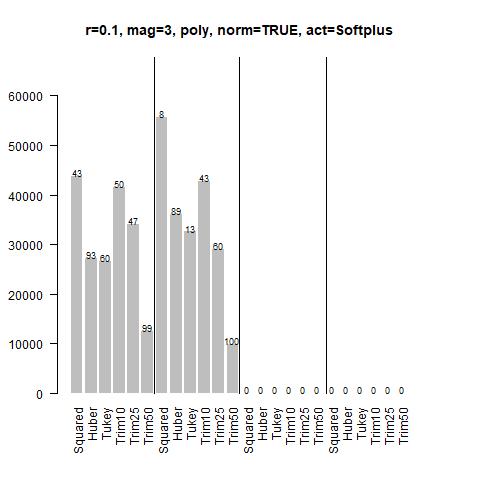} 
\end{center}
\caption{Results for $r=0.1$}
\end{figure}

\begin{figure}[H]
\label{trimnn:n1000p50r25m1polynonreluStep}
\begin{center}
\includegraphics[width=6.75cm,height=6.25cm]{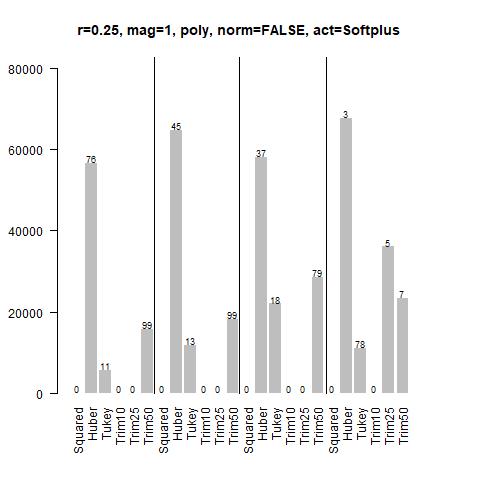}
\includegraphics[width=6.75cm,height=6.25cm]{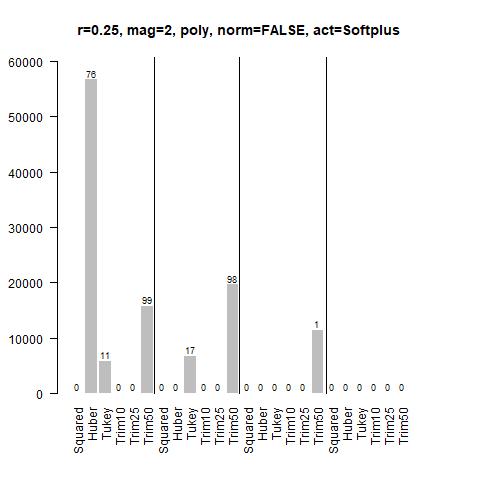} \\
\includegraphics[width=6.75cm,height=6.25cm]{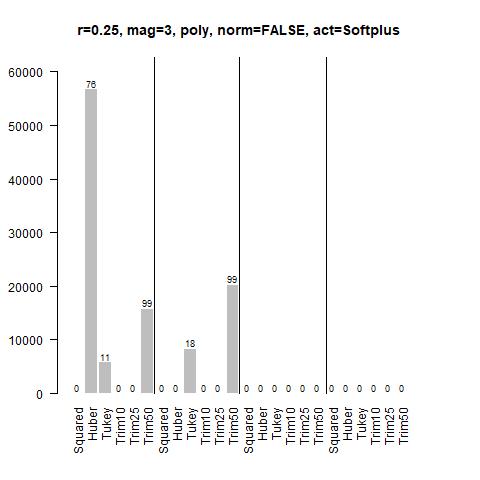} 
\includegraphics[width=6.75cm,height=6.25cm]{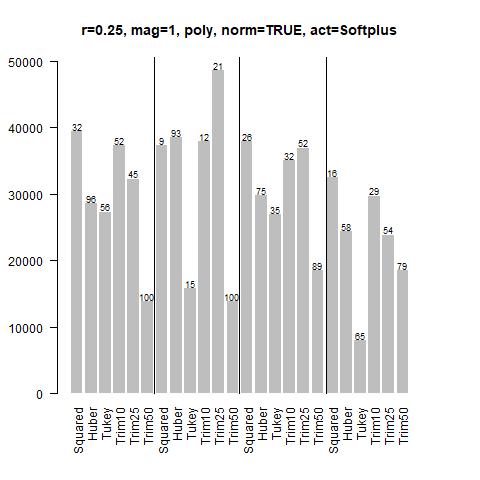}\\
\includegraphics[width=6.75cm,height=6.25cm]{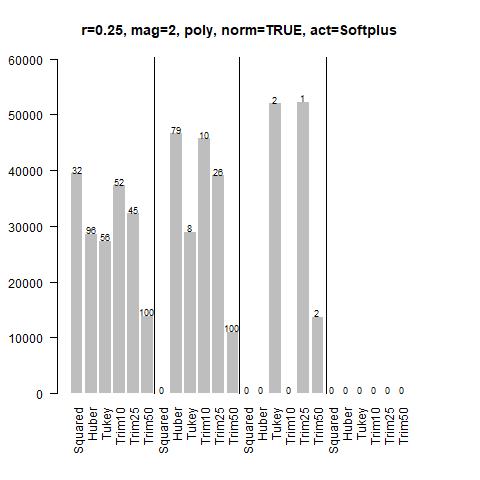} 
\includegraphics[width=6.75cm,height=6.25cm]{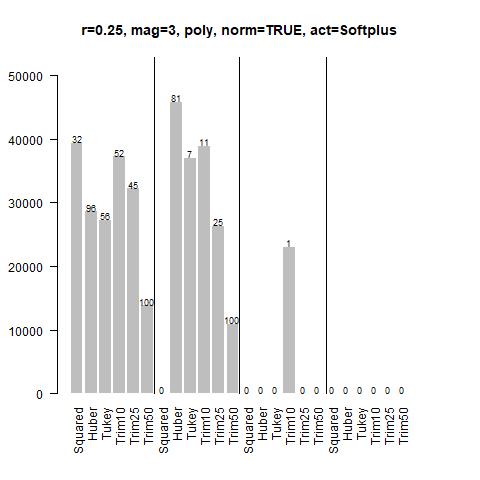} 
\end{center}
\caption{Results for $r=0.25$}
\end{figure}

\begin{figure}[H]
\label{trimnn:n1000p50r40m1polynonreluStep}
\begin{center}
\includegraphics[width=6.75cm,height=6.25cm]{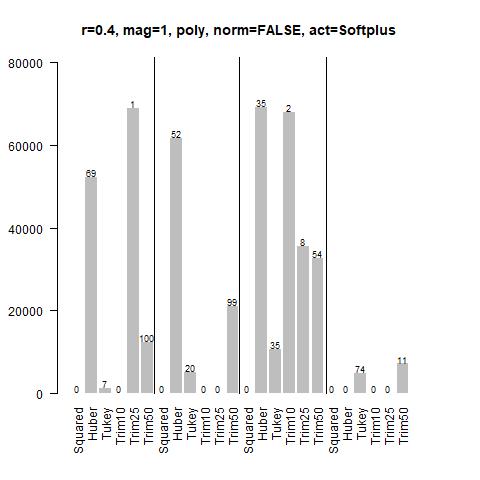}
\includegraphics[width=6.75cm,height=6.25cm]{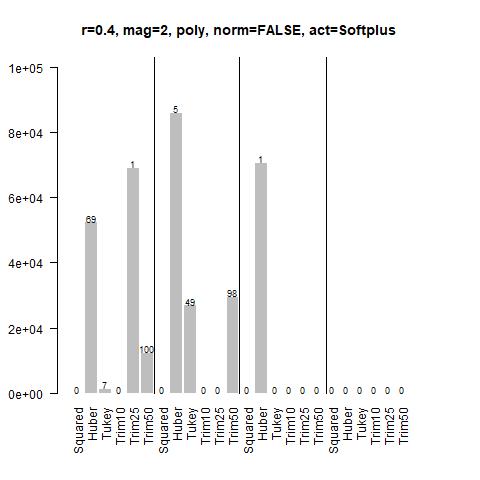} \\
\includegraphics[width=6.75cm,height=6.25cm]{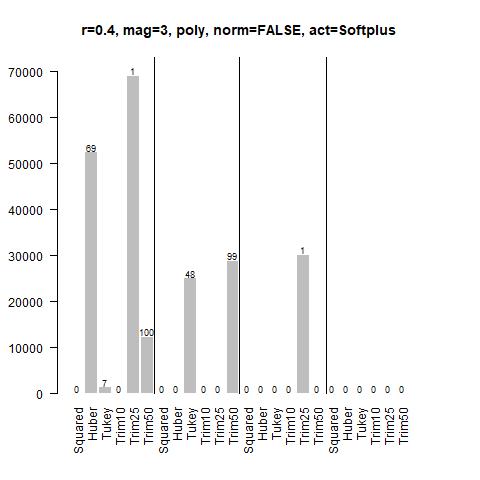} 
\includegraphics[width=6.75cm,height=6.25cm]{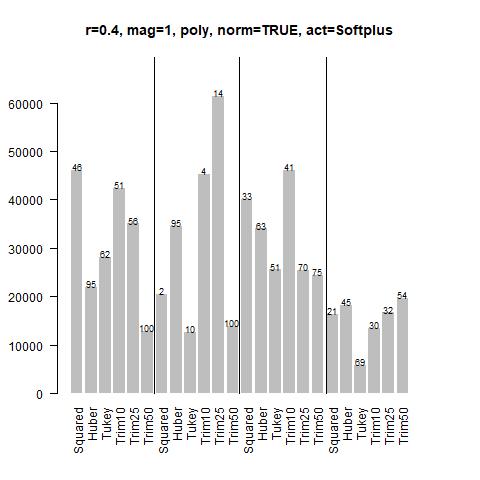}\\
\includegraphics[width=6.75cm,height=6.25cm]{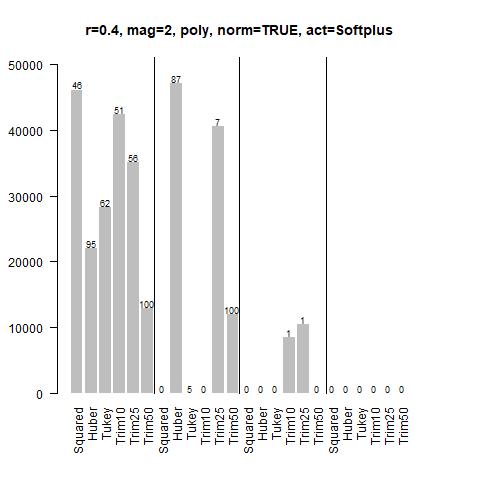} 
\includegraphics[width=6.75cm,height=6.25cm]{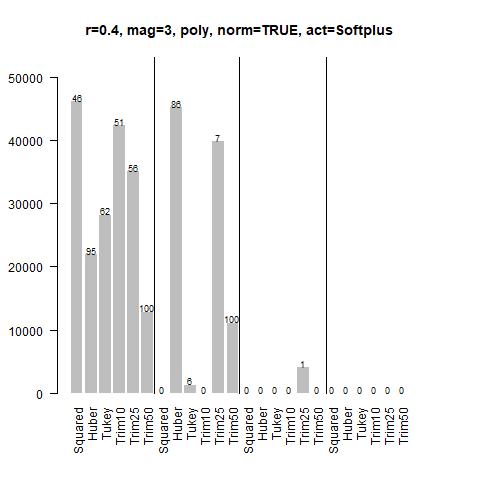} 
\end{center}
\caption{Results for $r=0.4$}
\end{figure}

\subsubsection{Trigonometric function}

\begin{figure}[H]
\label{trimnn:n1000p50r10m1trignonreluStep}
\begin{center}
\includegraphics[width=6.75cm,height=6.25cm]{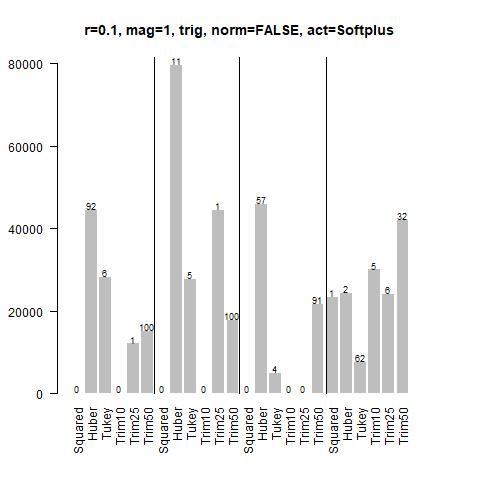}
\includegraphics[width=6.75cm,height=6.25cm]{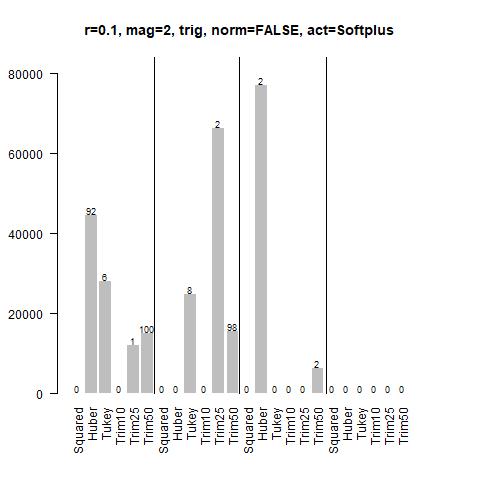} \\
\includegraphics[width=6.75cm,height=6.25cm]{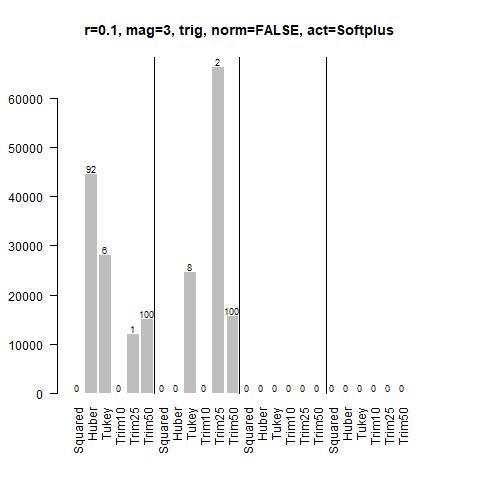} 
\includegraphics[width=6.75cm,height=6.25cm]{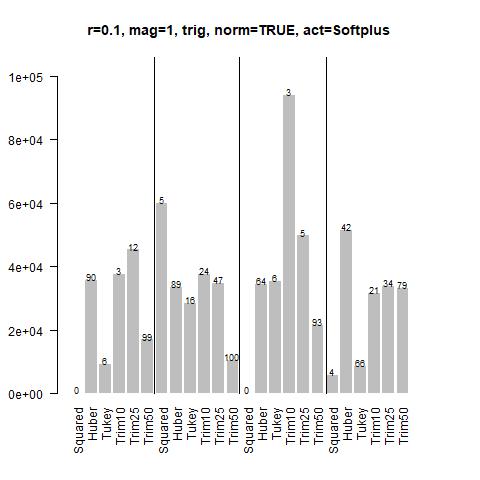}\\
\includegraphics[width=6.75cm,height=6.25cm]{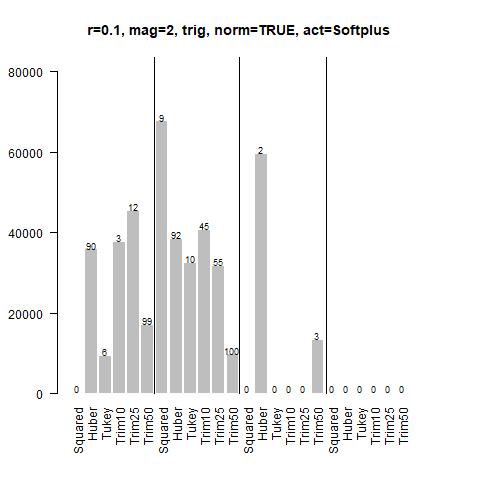} 
\includegraphics[width=6.75cm,height=6.25cm]{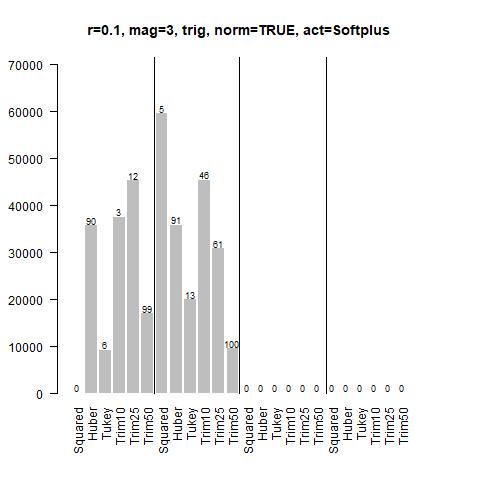} 
\end{center}
\caption{Results for $r=0.1$}
\end{figure}

\begin{figure}[H]
\label{trimnn:n1000p50r25m1trignonreluStep}
\begin{center}
\includegraphics[width=6.75cm,height=6.25cm]{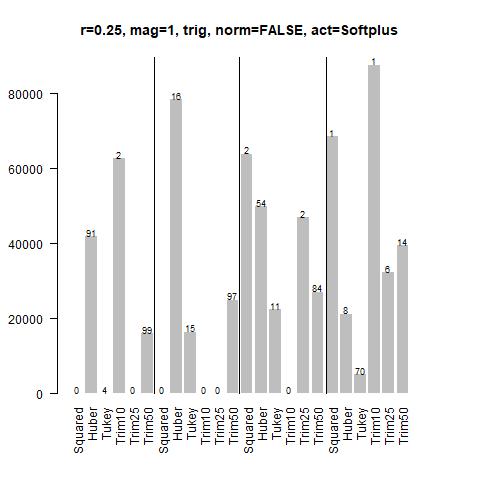}
\includegraphics[width=6.75cm,height=6.25cm]{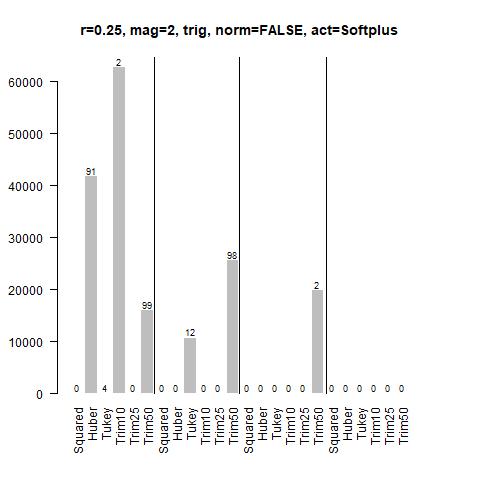} \\
\includegraphics[width=6.75cm,height=6.25cm]{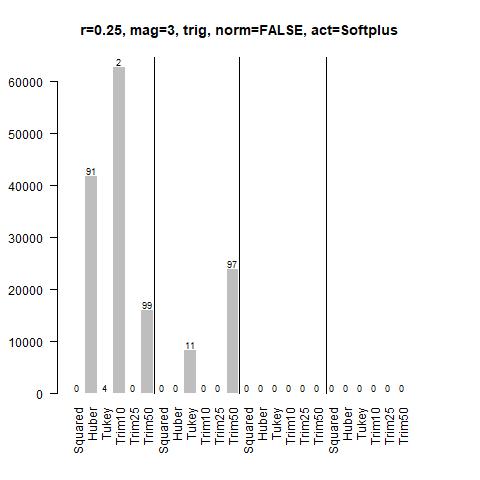} 
\includegraphics[width=6.75cm,height=6.25cm]{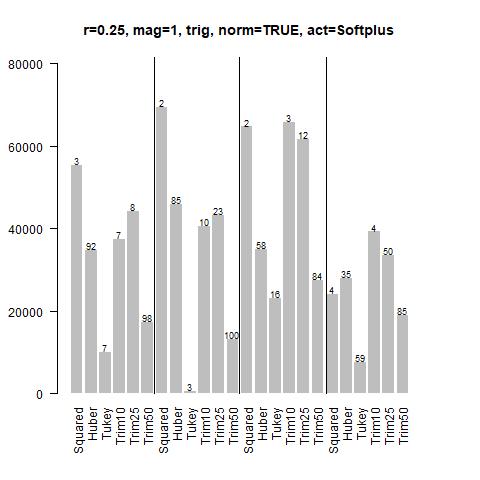}\\
\includegraphics[width=6.75cm,height=6.25cm]{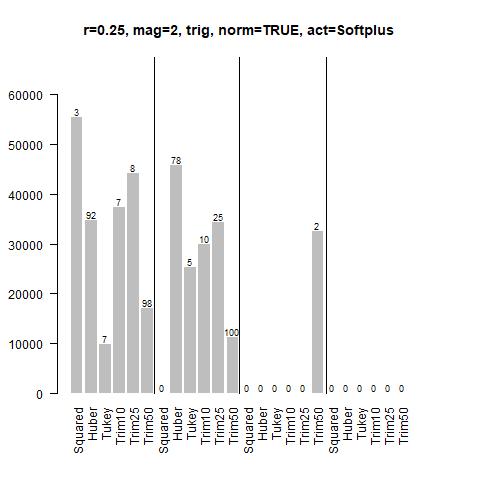} 
\includegraphics[width=6.75cm,height=6.25cm]{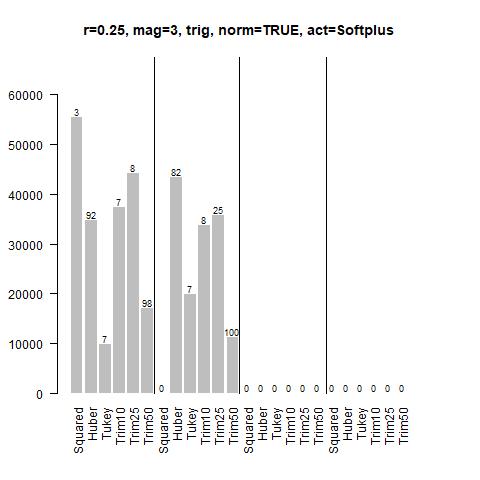} 
\end{center}
\caption{Results for $r=0.25$}
\end{figure}

\begin{figure}[H]
\label{trimnn:n1000p50r40m1trignonreluStep}
\begin{center}
\includegraphics[width=6.75cm,height=6.25cm]{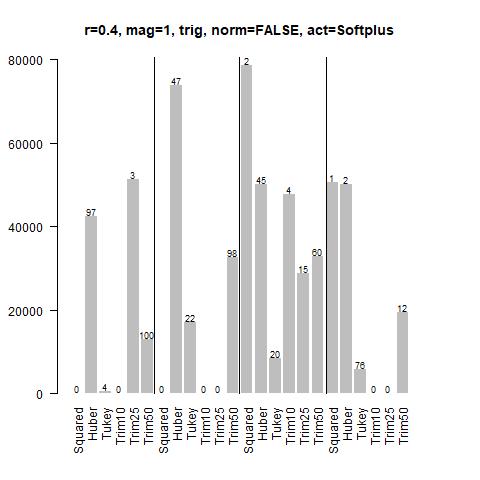}
\includegraphics[width=6.75cm,height=6.25cm]{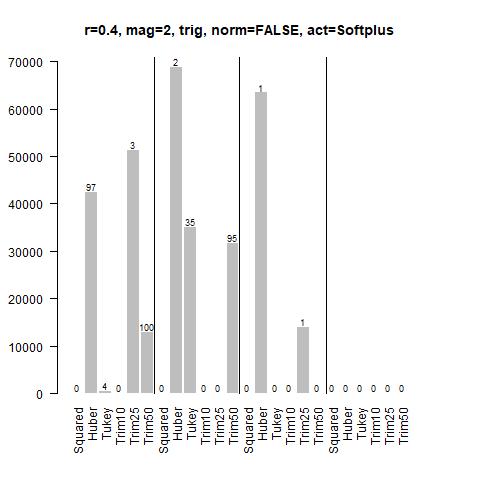} \\
\includegraphics[width=6.75cm,height=6.25cm]{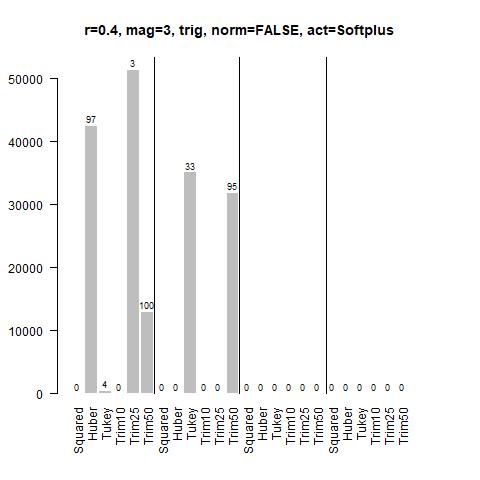} 
\includegraphics[width=6.75cm,height=6.25cm]{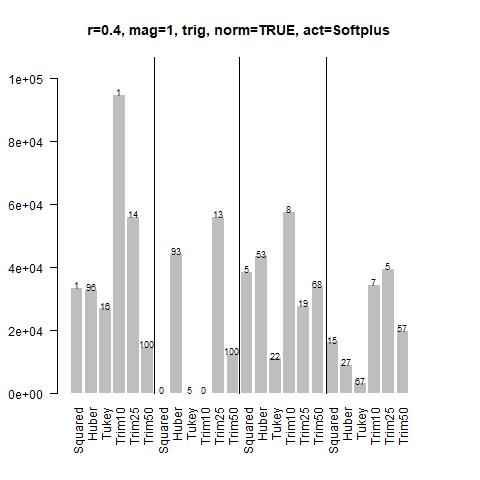}\\
\includegraphics[width=6.75cm,height=6.25cm]{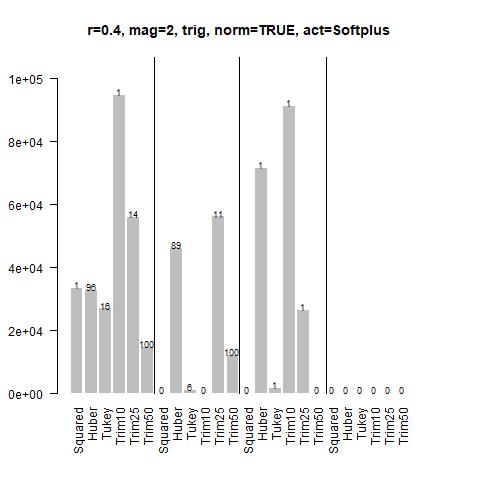} 
\includegraphics[width=6.75cm,height=6.25cm]{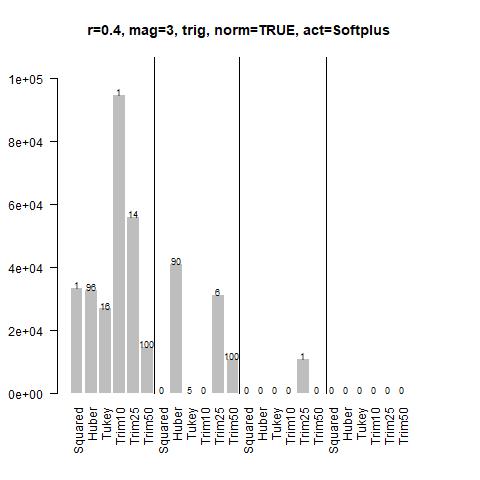} 
\end{center}
\caption{Results for $r=0.4$}
\end{figure}

\section{Simulation results for $n=1000$ and $p=50$, deep network: Training steps} \label{trimnn:secstep100050deep}

\subsection{Logistic activation function}

\subsubsection{Linear function}

\begin{figure}[H]
\label{trimnn:n1000p50r10m1linnonlogdeepStep}
\begin{center}
\includegraphics[width=6.75cm,height=6.25cm]{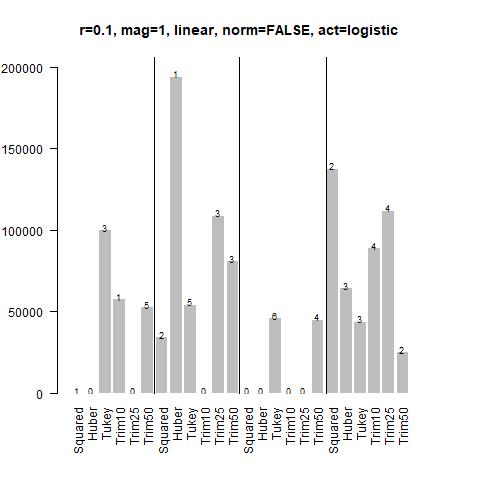}
\includegraphics[width=6.75cm,height=6.25cm]{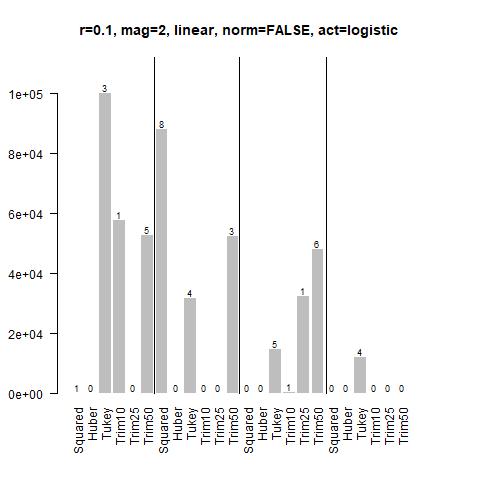} \\
\includegraphics[width=6.75cm,height=6.25cm]{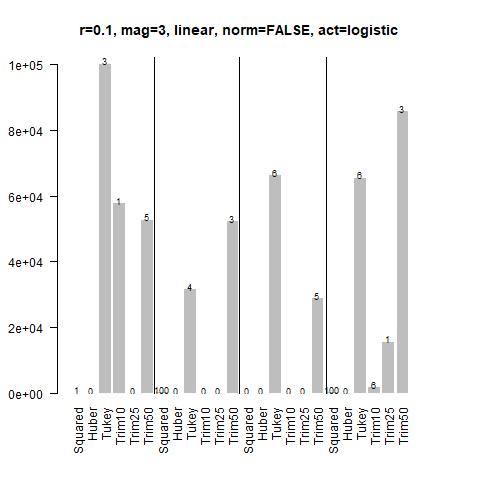} 
\includegraphics[width=6.75cm,height=6.25cm]{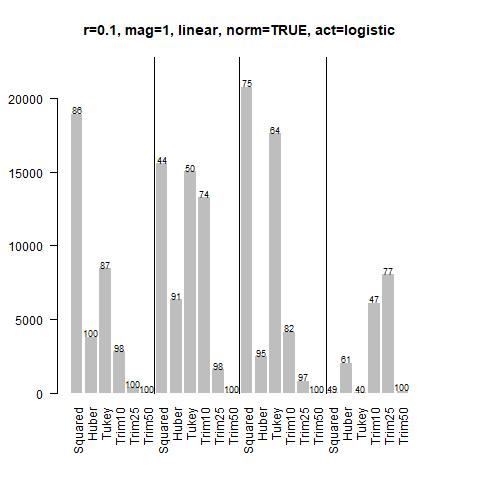}\\
\includegraphics[width=6.75cm,height=6.25cm]{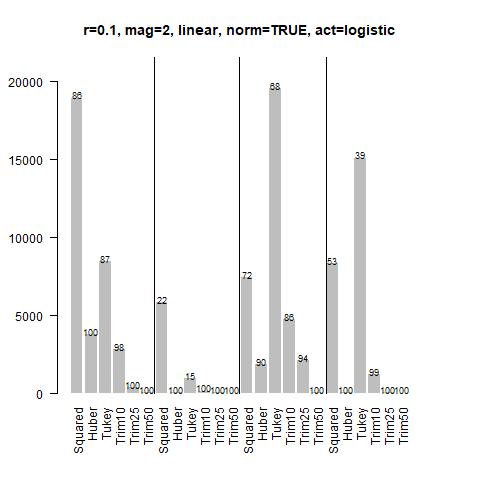} 
\includegraphics[width=6.75cm,height=6.25cm]{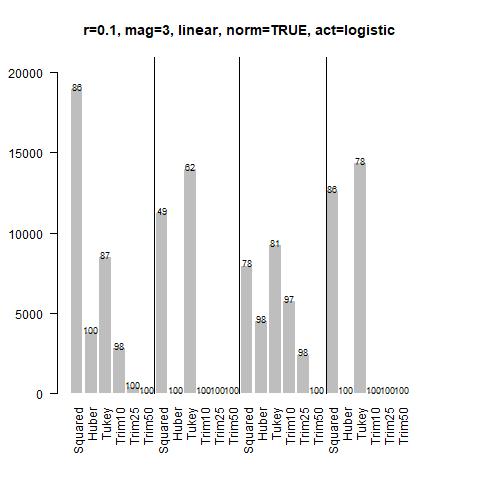} 
\end{center}
\caption{Results for $r=0.1$}
\end{figure}

\begin{figure}[H]
\label{trimnn:n1000p50r25m1linnonlogdeepStep}
\begin{center}
\includegraphics[width=6.75cm,height=6.25cm]{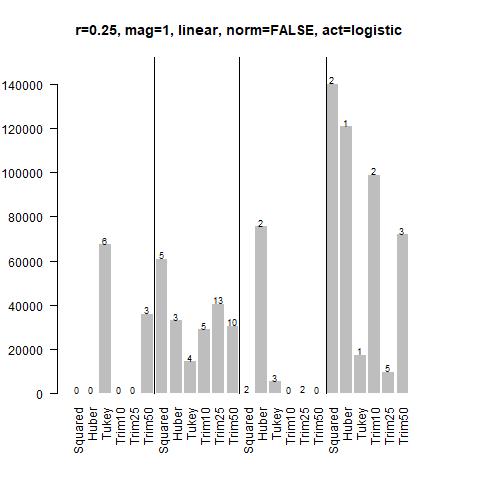}
\includegraphics[width=6.75cm,height=6.25cm]{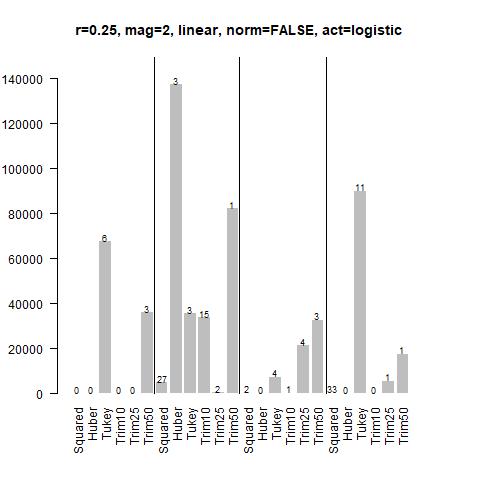} \\
\includegraphics[width=6.75cm,height=6.25cm]{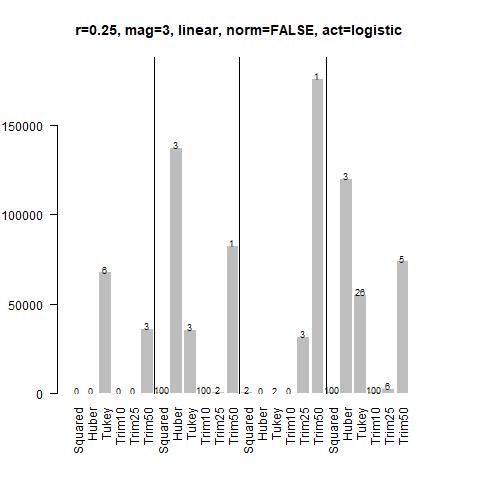} 
\includegraphics[width=6.75cm,height=6.25cm]{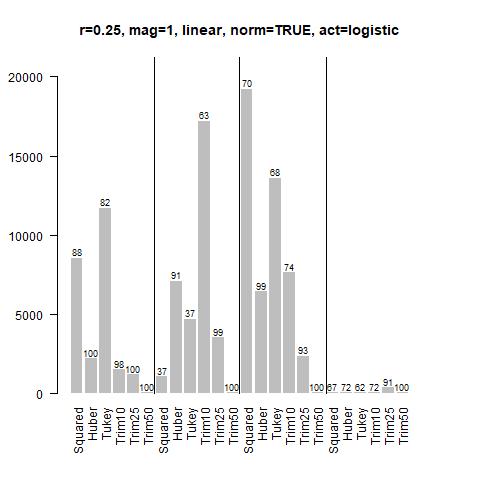}\\
\includegraphics[width=6.75cm,height=6.25cm]{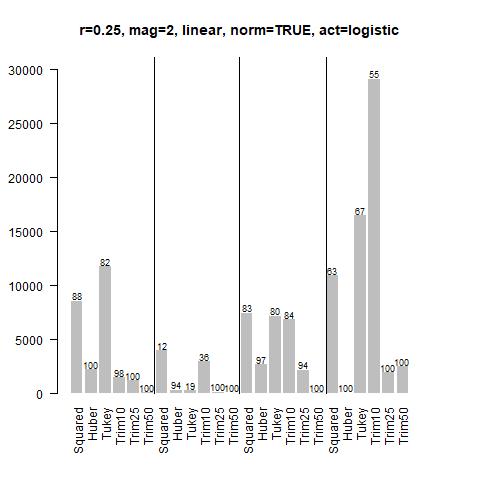} 
\includegraphics[width=6.75cm,height=6.25cm]{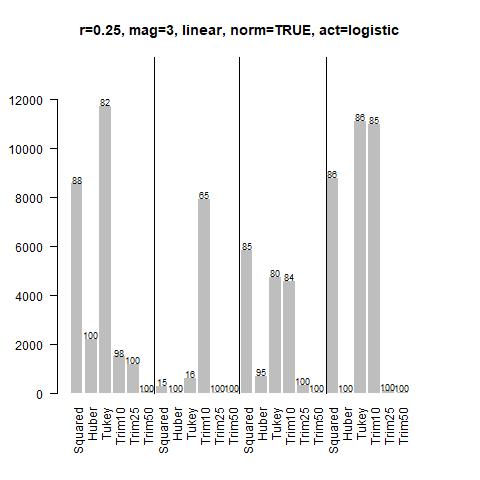} 
\end{center}
\caption{Results for $r=0.25$}
\end{figure}

\begin{figure}[H]
\label{trimnn:n1000p50r40m1linnonlogdeepStep}
\begin{center}
\includegraphics[width=6.75cm,height=6.25cm]{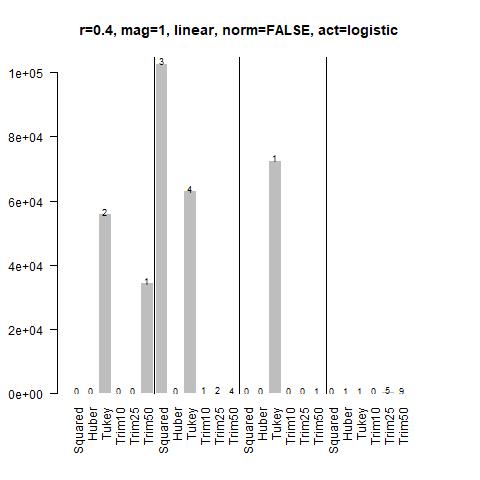}
\includegraphics[width=6.75cm,height=6.25cm]{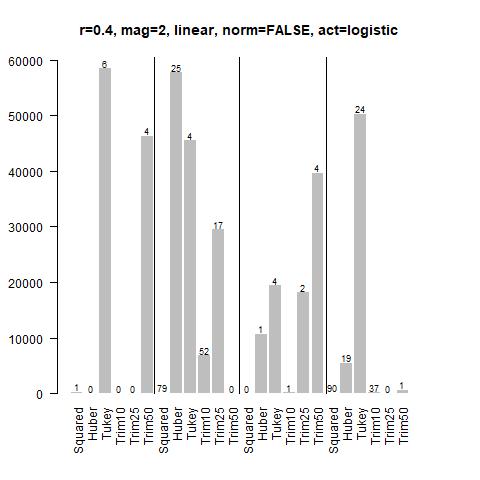} \\
\includegraphics[width=6.75cm,height=6.25cm]{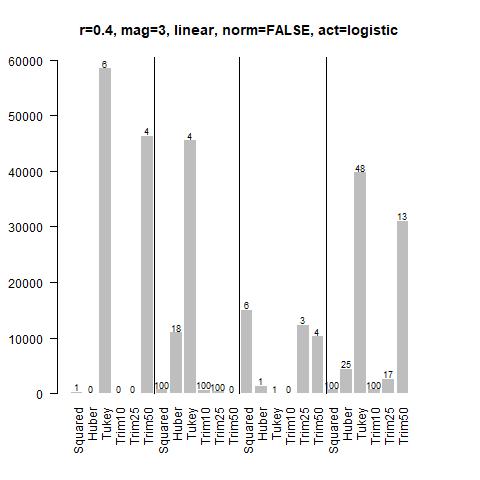} 
\includegraphics[width=6.75cm,height=6.25cm]{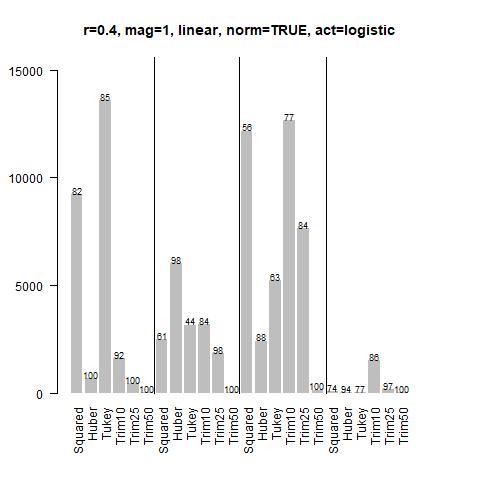}\\
\includegraphics[width=6.75cm,height=6.25cm]{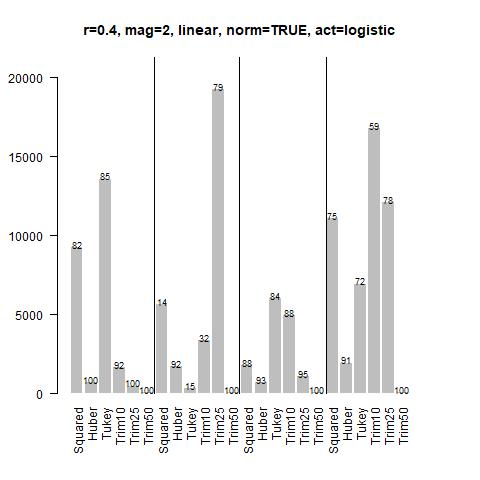} 
\includegraphics[width=6.75cm,height=6.25cm]{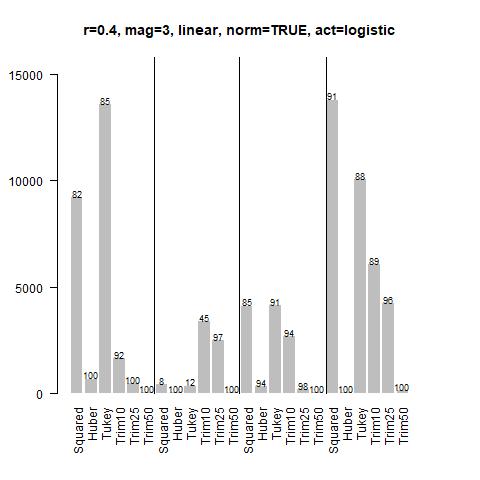} 
\end{center}
\caption{Results for $r=0.4$}
\end{figure}

\subsubsection{Polynomial function}

\begin{figure}[H]
\label{trimnn:n1000p50r10m1polynonlogdeepStep}
\begin{center}
\includegraphics[width=6.75cm,height=6.25cm]{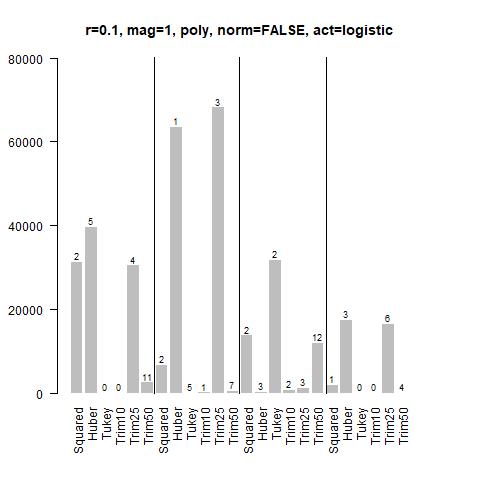}
\includegraphics[width=6.75cm,height=6.25cm]{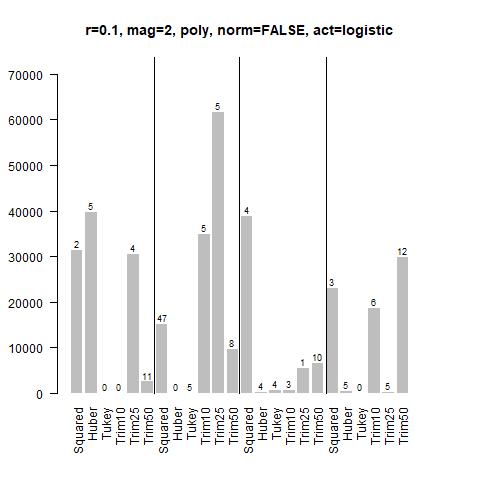} \\
\includegraphics[width=6.75cm,height=6.25cm]{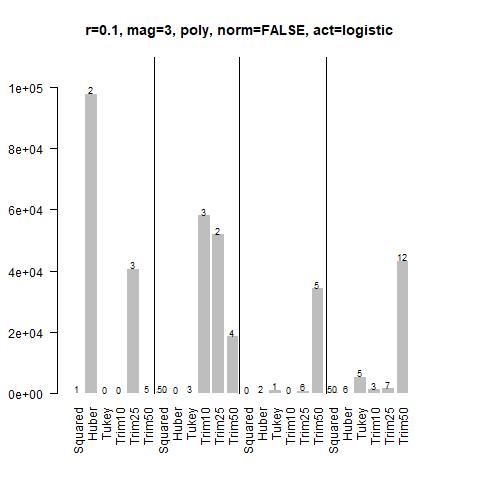} 
\includegraphics[width=6.75cm,height=6.25cm]{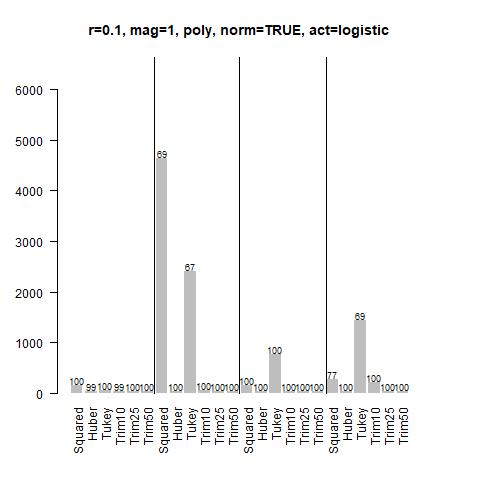}\\
\includegraphics[width=6.75cm,height=6.25cm]{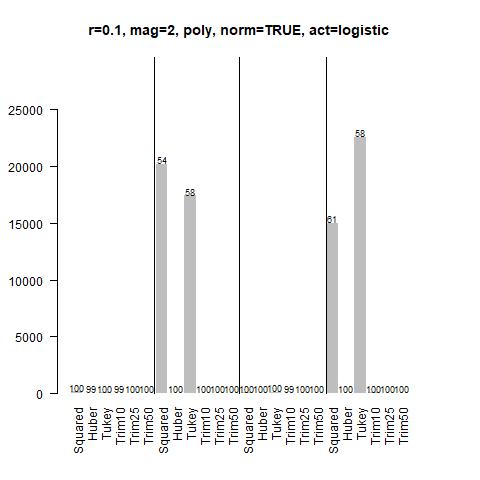} 
\includegraphics[width=6.75cm,height=6.25cm]{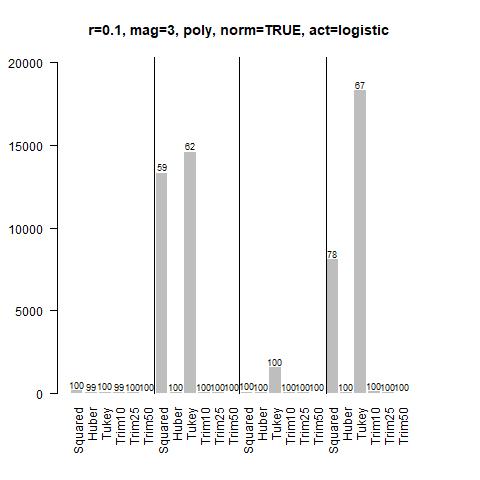} 
\end{center}
\caption{Results for $r=0.1$}
\end{figure}

\begin{figure}[H]
\label{trimnn:n1000p50r25m1polynonlogdeepStep}
\begin{center}
\includegraphics[width=6.75cm,height=6.25cm]{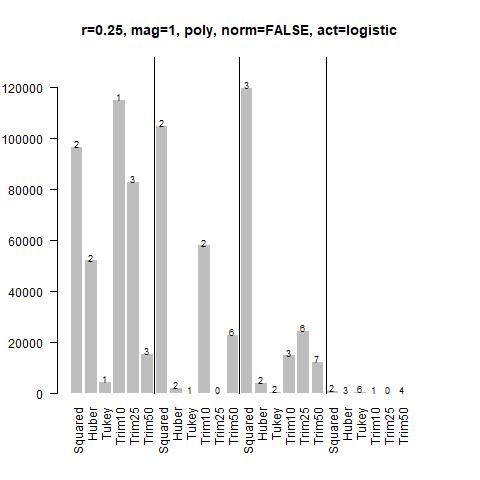}
\includegraphics[width=6.75cm,height=6.25cm]{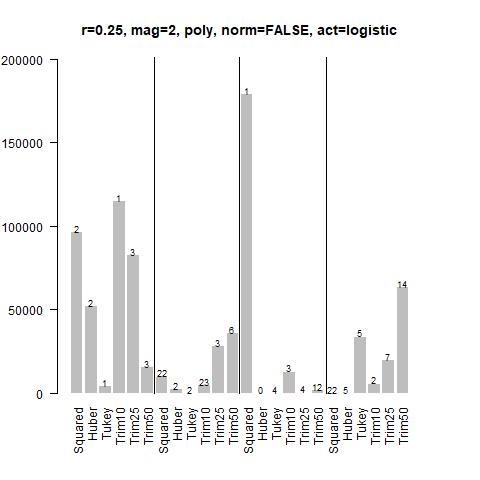} \\
\includegraphics[width=6.75cm,height=6.25cm]{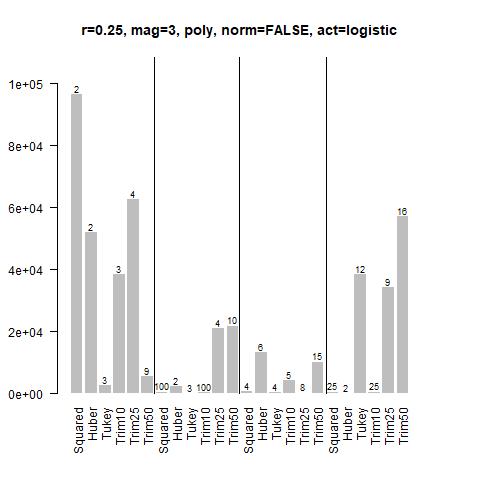} 
\includegraphics[width=6.75cm,height=6.25cm]{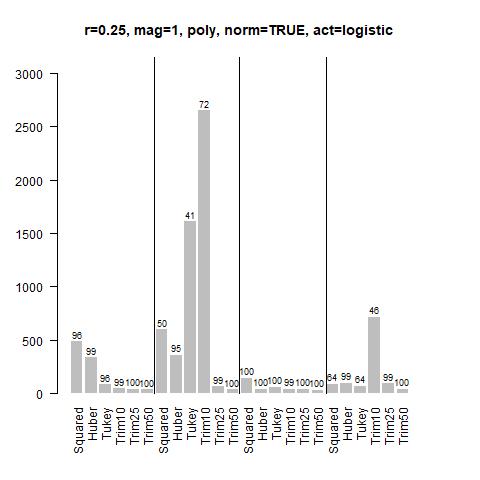}\\
\includegraphics[width=6.75cm,height=6.25cm]{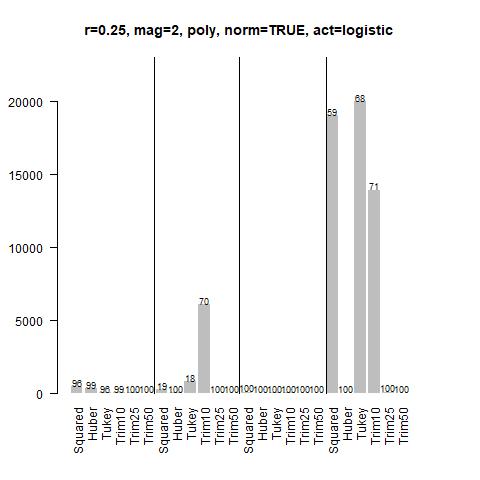} 
\includegraphics[width=6.75cm,height=6.25cm]{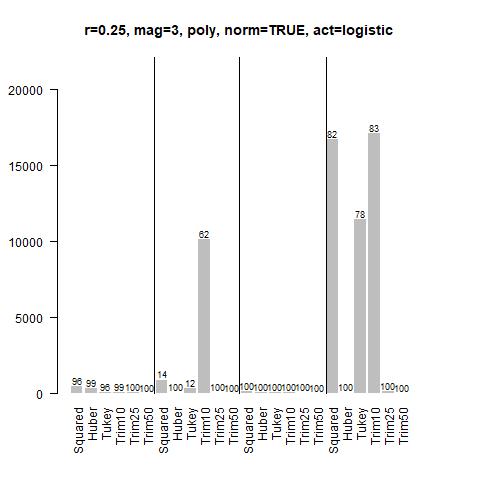} 
\end{center}
\caption{Results for $r=0.25$}
\end{figure}

\begin{figure}[H]
\label{trimnn:n1000p50r40m1polynonlogdeepStep}
\begin{center}
\includegraphics[width=6.75cm,height=6.25cm]{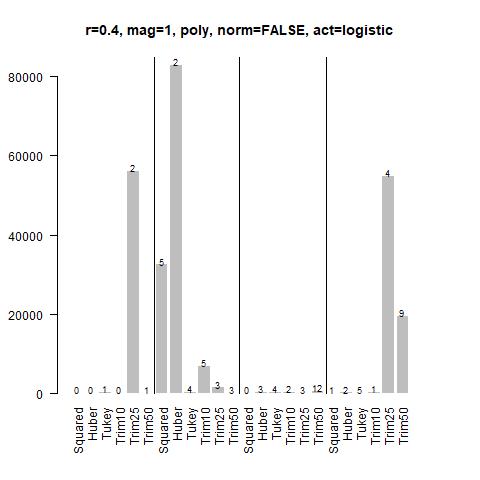}
\includegraphics[width=6.75cm,height=6.25cm]{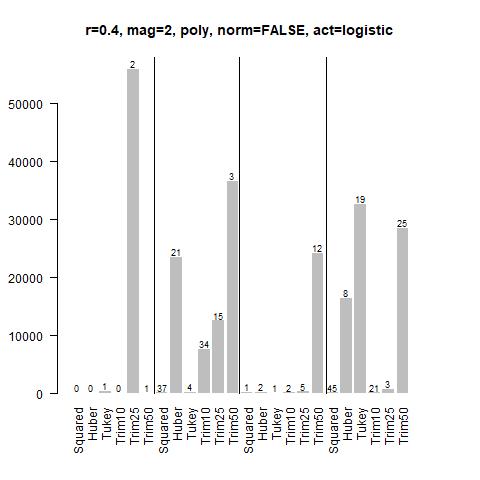} \\
\includegraphics[width=6.75cm,height=6.25cm]{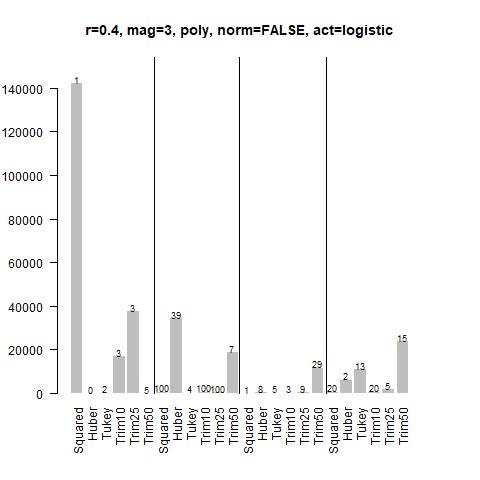} 
\includegraphics[width=6.75cm,height=6.25cm]{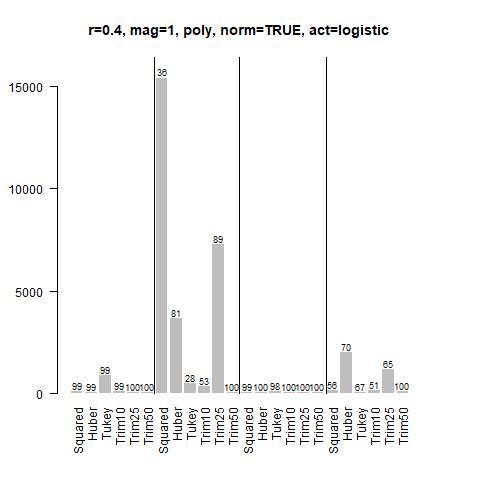}\\
\includegraphics[width=6.75cm,height=6.25cm]{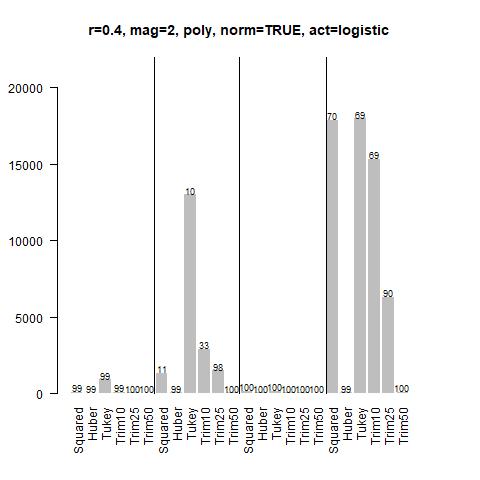} 
\includegraphics[width=6.75cm,height=6.25cm]{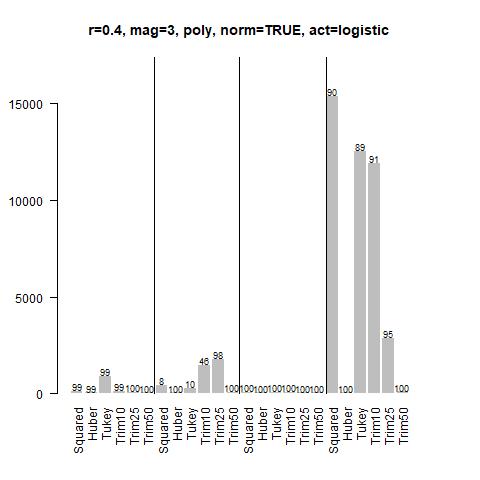} 
\end{center}
\caption{Results for $r=0.4$}
\end{figure}

\subsubsection{Trigonometric function}

\begin{figure}[H]
\label{trimnn:n1000p50r10m1trignonlogdeeStepp}
\begin{center}
\includegraphics[width=6.75cm,height=6.25cm]{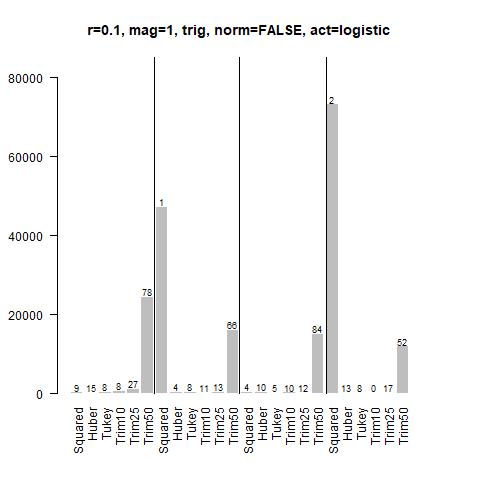}
\includegraphics[width=6.75cm,height=6.25cm]{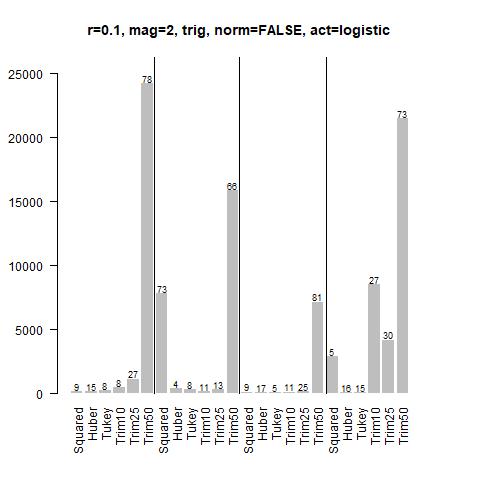} \\
\includegraphics[width=6.75cm,height=6.25cm]{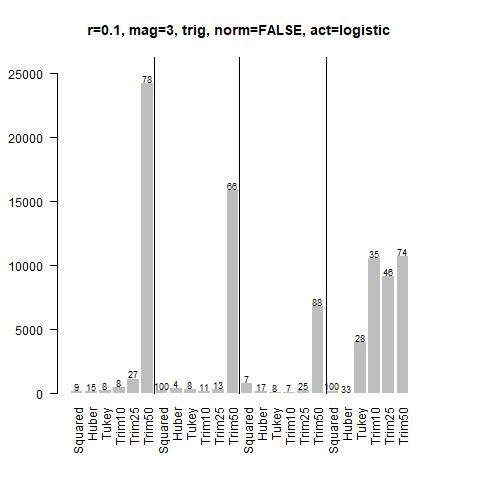} 
\includegraphics[width=6.75cm,height=6.25cm]{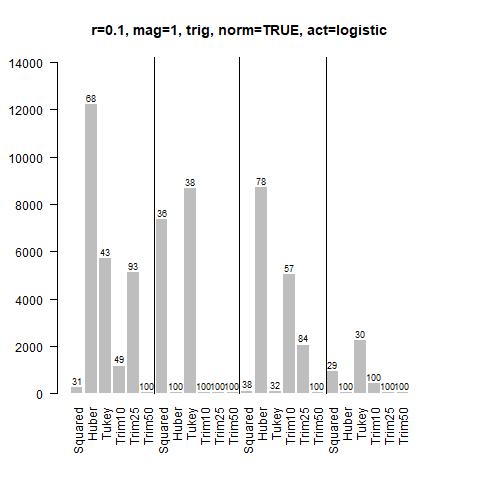}\\
\includegraphics[width=6.75cm,height=6.25cm]{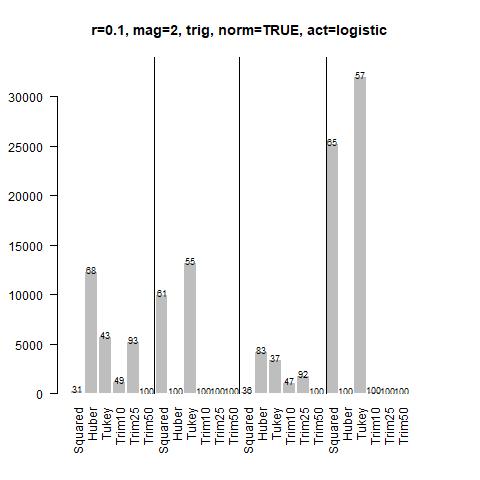} 
\includegraphics[width=6.75cm,height=6.25cm]{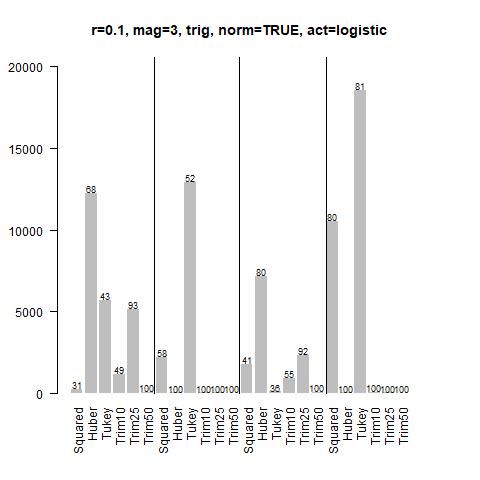} 
\end{center}
\caption{Results for $r=0.1$}
\end{figure}

\begin{figure}[H]
\label{trimnn:n1000p50r25m1trignonlogdeepStep}
\begin{center}
\includegraphics[width=6.75cm,height=6.25cm]{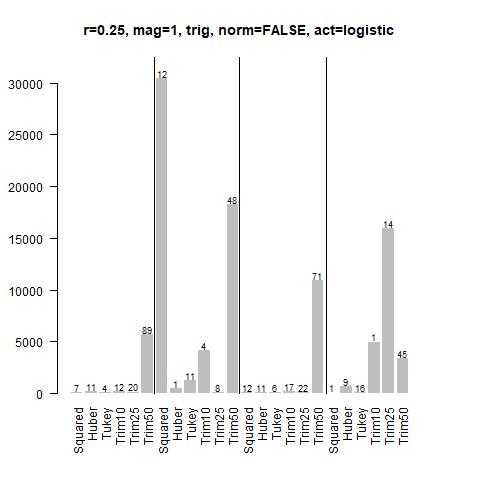}
\includegraphics[width=6.75cm,height=6.25cm]{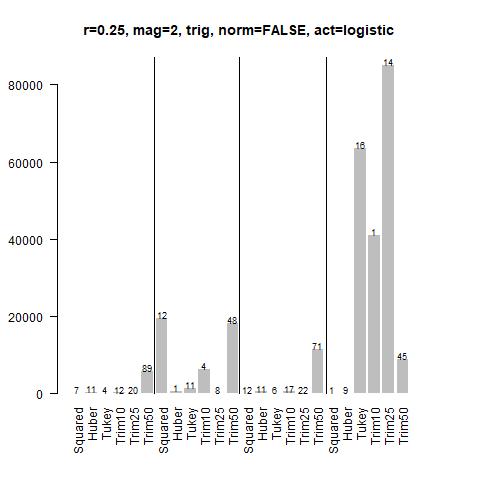} \\
\includegraphics[width=6.75cm,height=6.25cm]{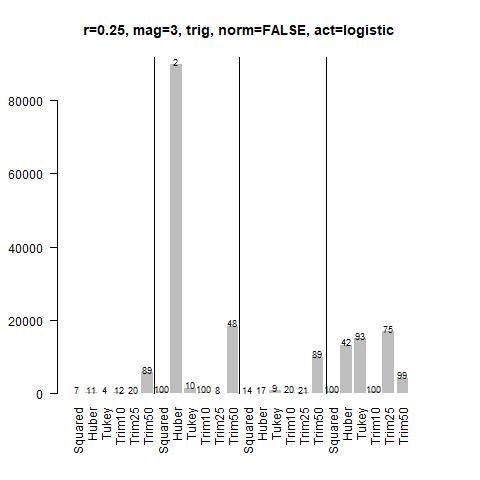} 
\includegraphics[width=6.75cm,height=6.25cm]{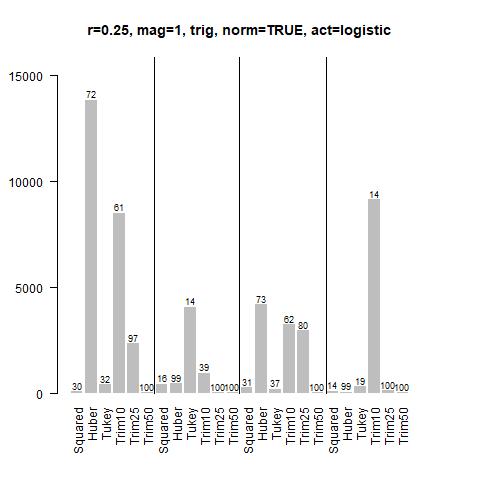}\\
\includegraphics[width=6.75cm,height=6.25cm]{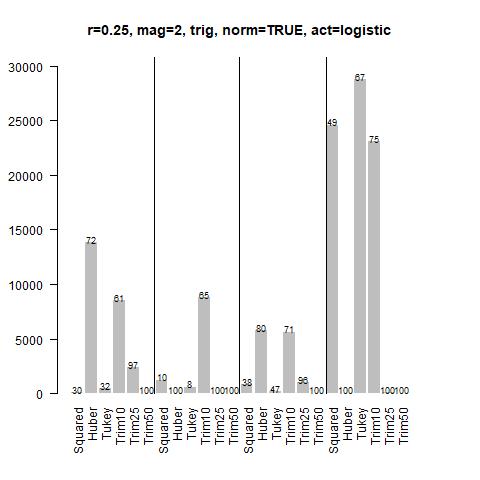} 
\includegraphics[width=6.75cm,height=6.25cm]{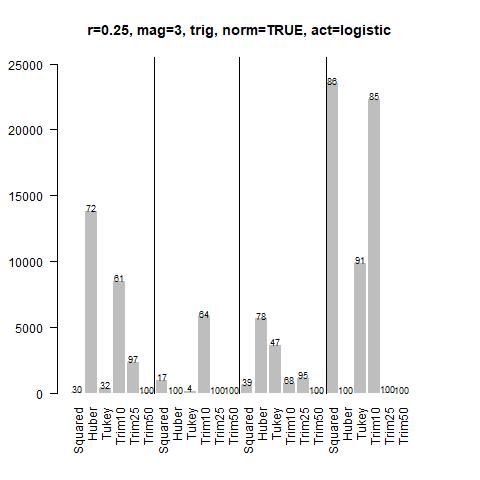} 
\end{center}
\caption{Results for $r=0.25$}
\end{figure}

\begin{figure}[H]
\begin{center}
\includegraphics[width=6.75cm,height=6.25cm]{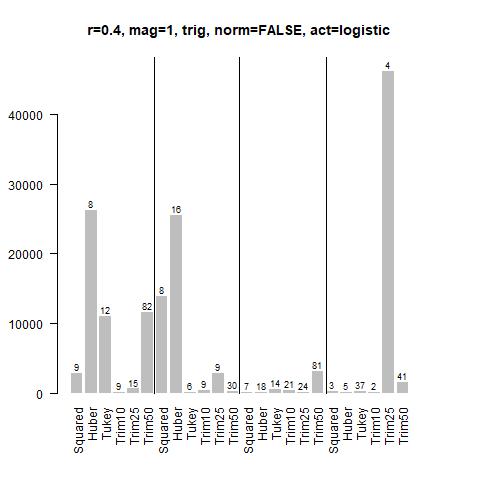}
\includegraphics[width=6.75cm,height=6.25cm]{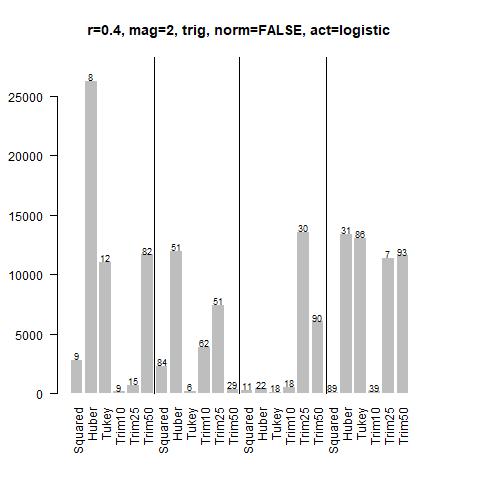} \\
\includegraphics[width=6.75cm,height=6.25cm]{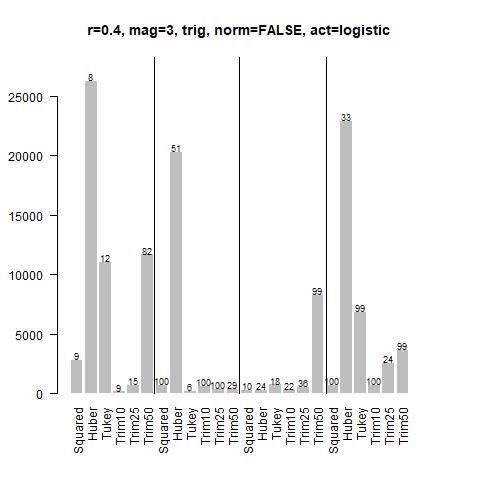} 
\includegraphics[width=6.75cm,height=6.25cm]{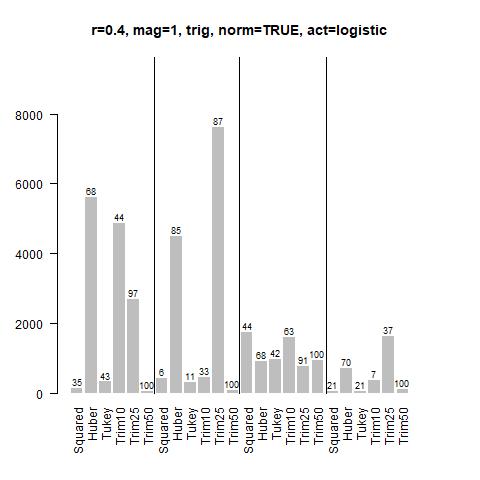}\\
\includegraphics[width=6.75cm,height=6.25cm]{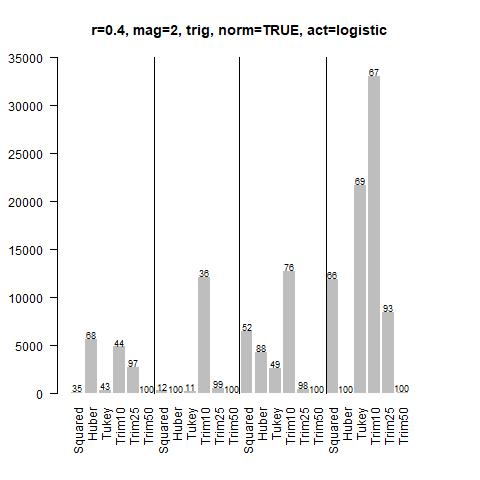} 
\includegraphics[width=6.75cm,height=6.25cm]{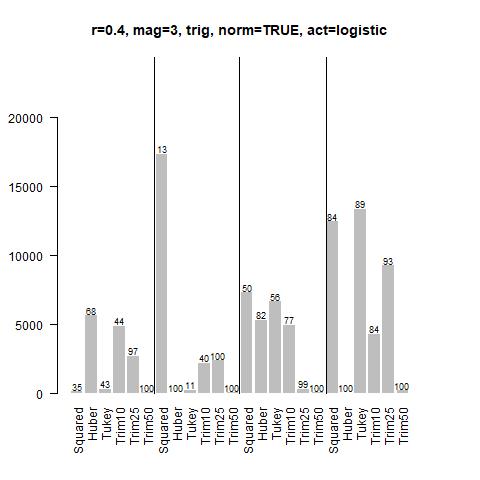} 
\end{center}
\caption{Results for $r=0.4$}\label{trimnn:n1000p50r40m1trignonlogdeepStep}
\end{figure}

\subsection{Softplus activation function}

\subsubsection{Linear function}

\begin{figure}[H]
\begin{center}
\includegraphics[width=6.75cm,height=6.25cm]{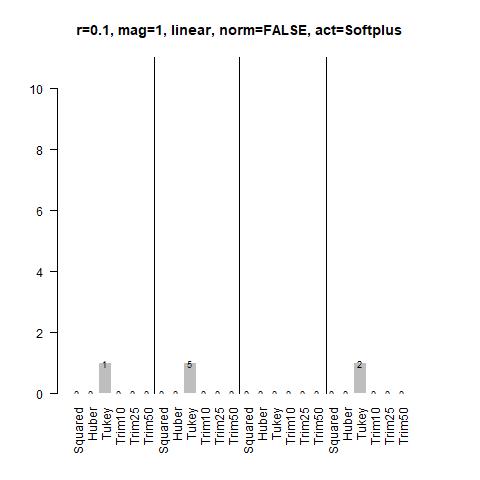}
\includegraphics[width=6.75cm,height=6.25cm]{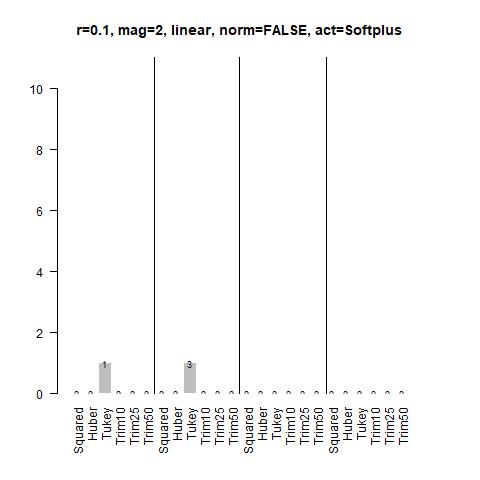} \\
\includegraphics[width=6.75cm,height=6.25cm]{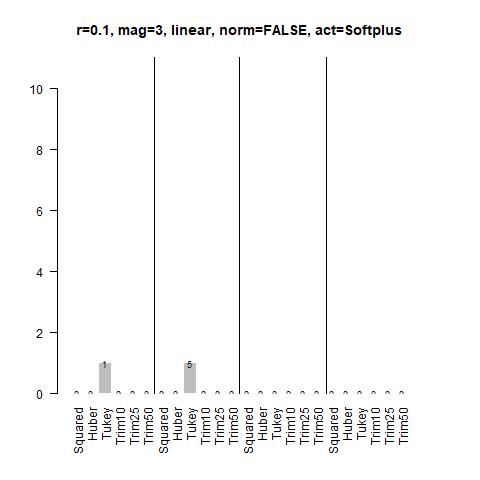} 
\includegraphics[width=6.75cm,height=6.25cm]{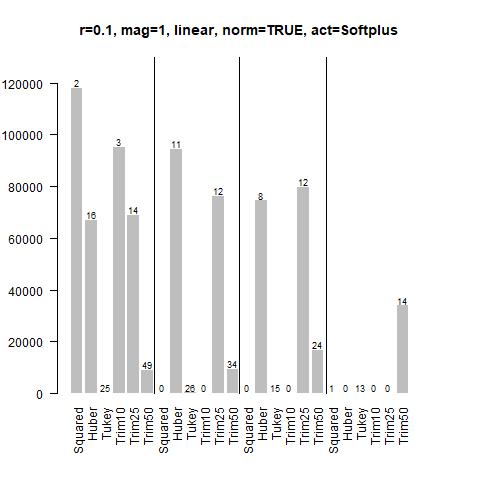}\\
\includegraphics[width=6.75cm,height=6.25cm]{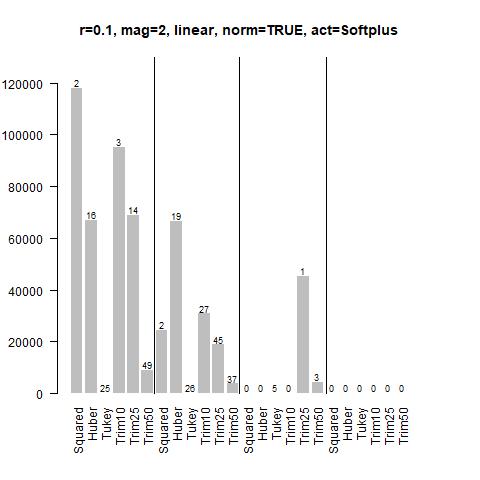} 
\includegraphics[width=6.75cm,height=6.25cm]{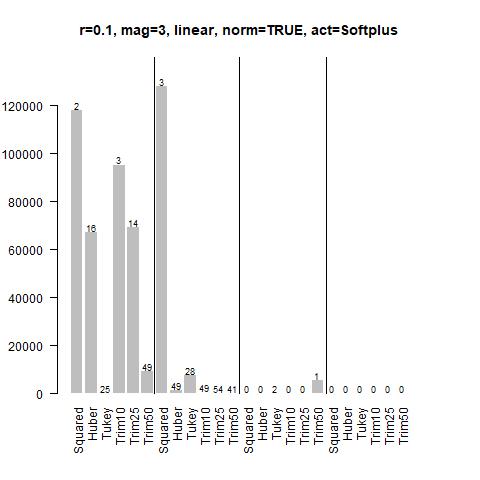} 
\end{center}
\caption{Results for $r=0.1$}\label{trimnn:n1000p50r10m1linnonreludeepStep}
\end{figure}

\begin{figure}[H]
\begin{center}
\includegraphics[width=6.75cm,height=6.25cm]{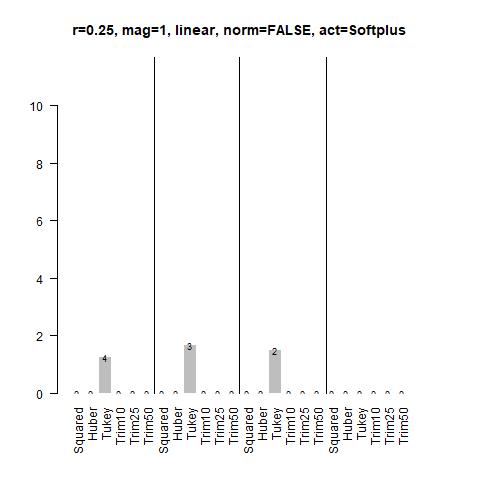}
\includegraphics[width=6.75cm,height=6.25cm]{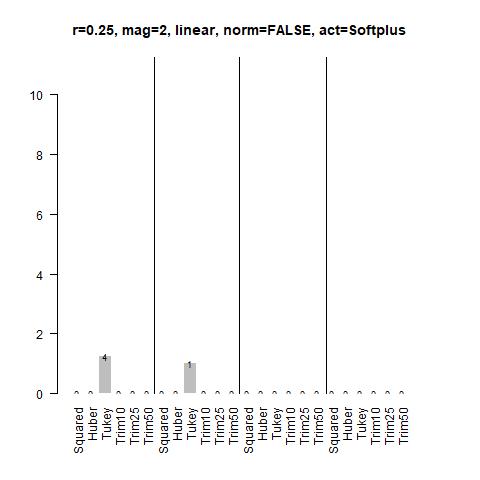} \\
\includegraphics[width=6.75cm,height=6.25cm]{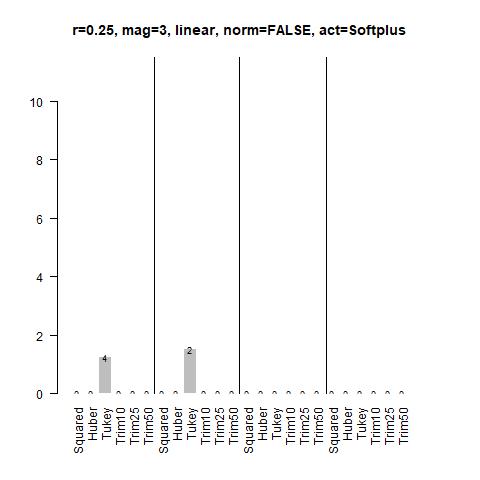} 
\includegraphics[width=6.75cm,height=6.25cm]{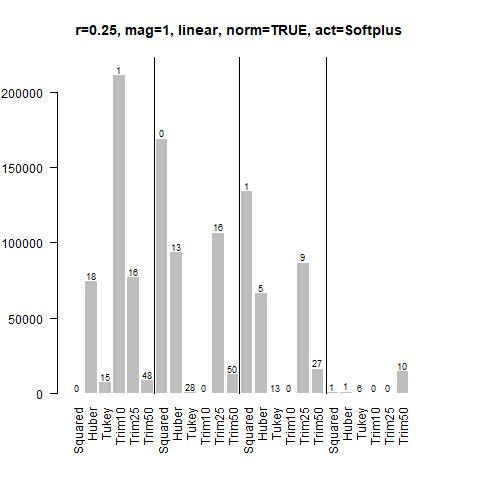}\\
\includegraphics[width=6.75cm,height=6.25cm]{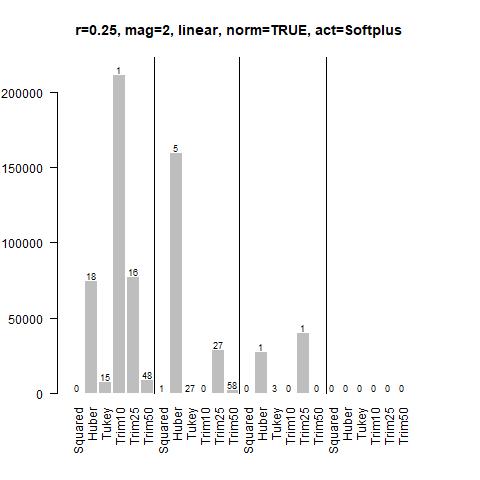} 
\includegraphics[width=6.75cm,height=6.25cm]{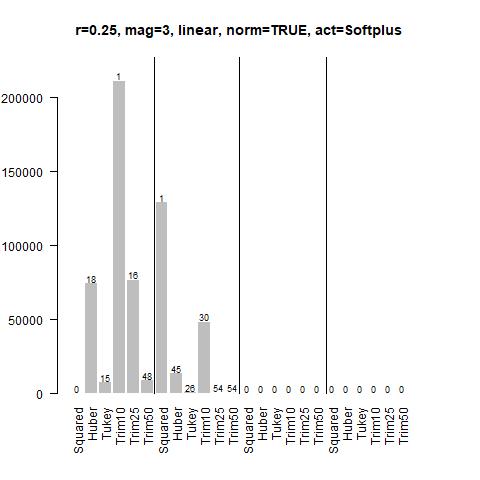} 
\end{center}
\caption{Results for $r=0.25$}\label{trimnn:n1000p50r25m1linnonreludeepStep}
\end{figure}

\begin{figure}[H]
\begin{center}
\includegraphics[width=6.75cm,height=6.25cm]{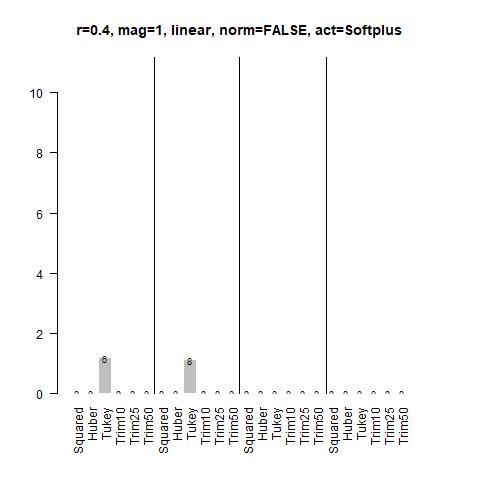}
\includegraphics[width=6.75cm,height=6.25cm]{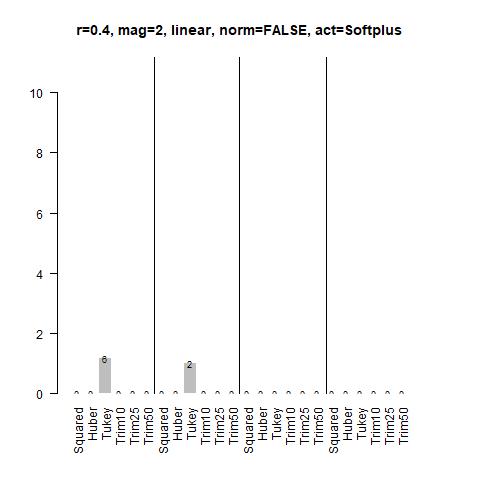} \\
\includegraphics[width=6.75cm,height=6.25cm]{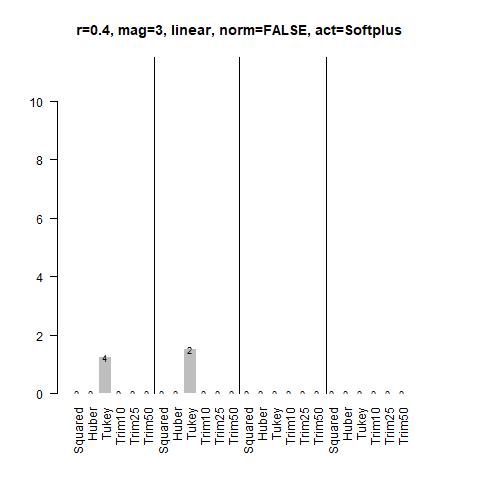} 
\includegraphics[width=6.75cm,height=6.25cm]{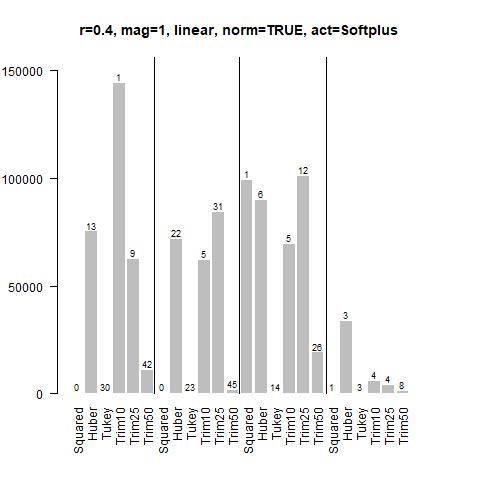}\\
\includegraphics[width=6.75cm,height=6.25cm]{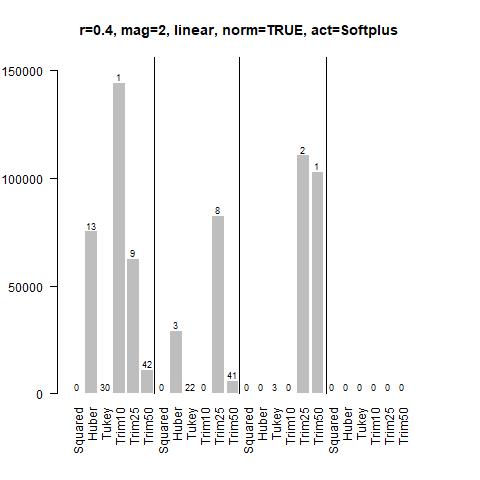} 
\includegraphics[width=6.75cm,height=6.25cm]{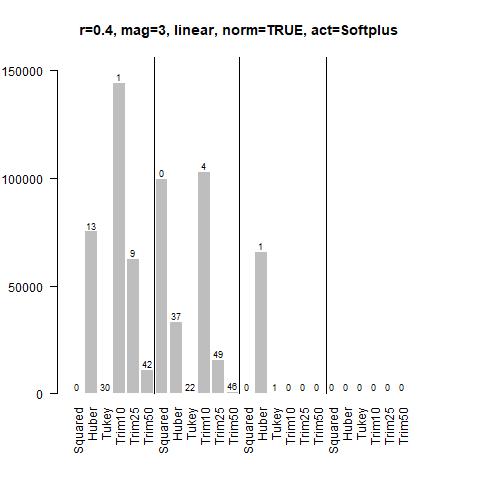} 
\end{center}
\caption{Results for $r=0.4$}\label{trimnn:n1000p50r40m1linnonreludeepStep}
\end{figure}

\subsubsection{Polynomial function}

\begin{figure}[H]
\label{trimnn:n1000p50r10m1polynonreludeepStep}
\begin{center}
\includegraphics[width=6.75cm,height=6.25cm]{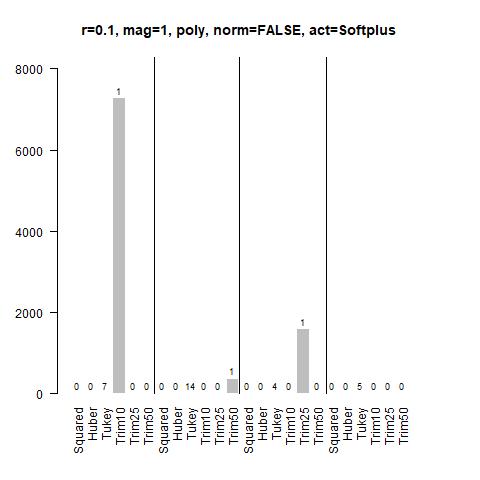}
\includegraphics[width=6.75cm,height=6.25cm]{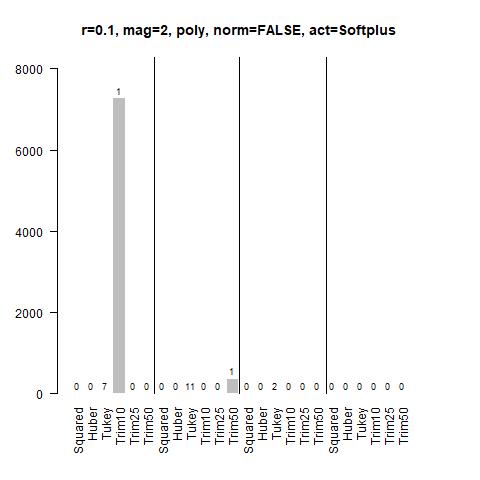} \\
\includegraphics[width=6.75cm,height=6.25cm]{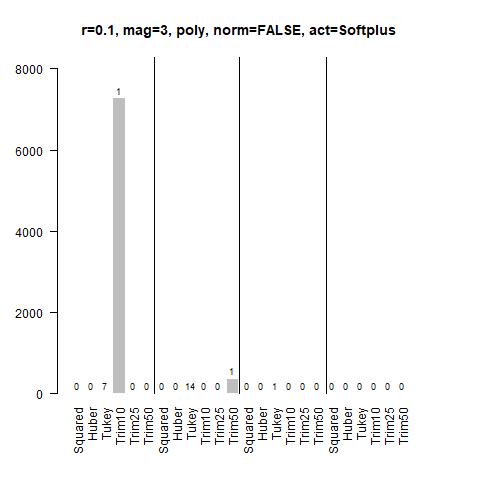} 
\includegraphics[width=6.75cm,height=6.25cm]{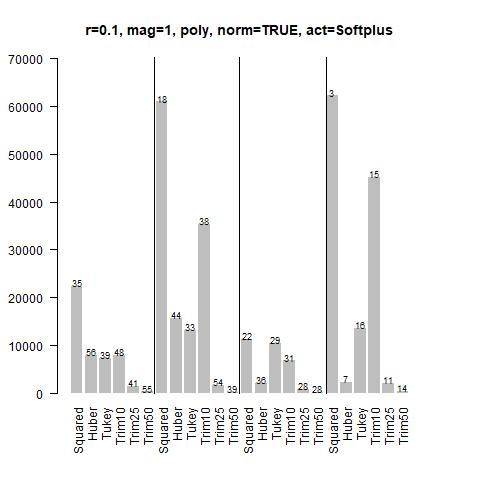}\\
\includegraphics[width=6.75cm,height=6.25cm]{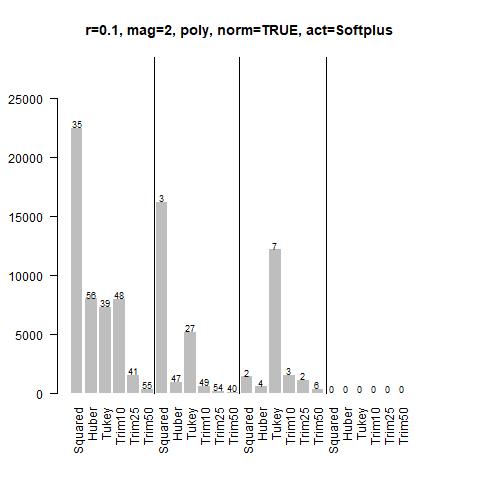} 
\includegraphics[width=6.75cm,height=6.25cm]{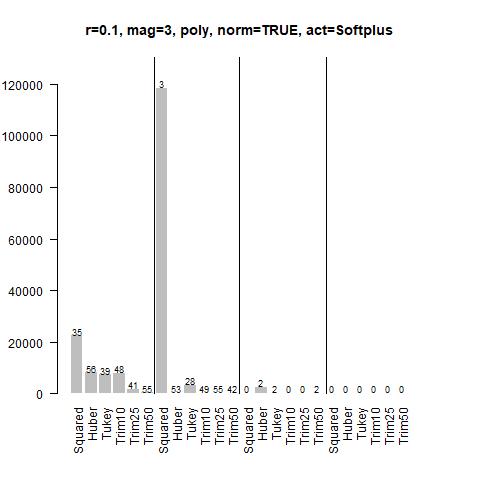} 
\end{center}
\caption{Results for $r=0.1$}
\end{figure}

\begin{figure}[H]
\label{trimnn:n1000p50r25m1polynonreludeepStep}
\begin{center}
\includegraphics[width=6.75cm,height=6.25cm]{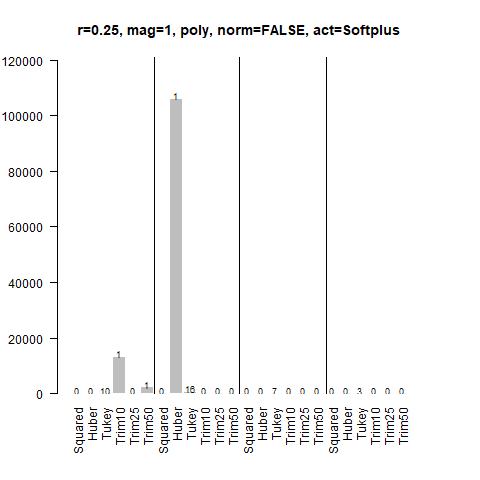}
\includegraphics[width=6.75cm,height=6.25cm]{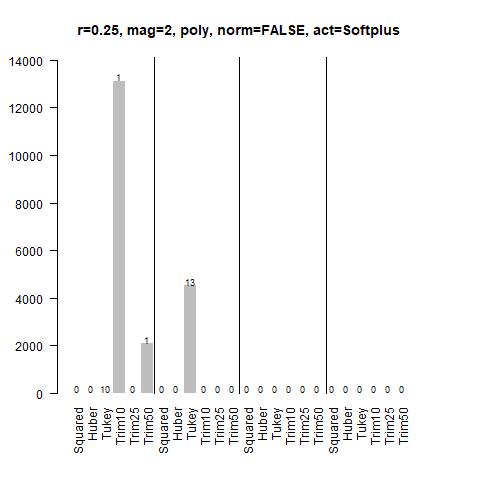} \\
\includegraphics[width=6.75cm,height=6.25cm]{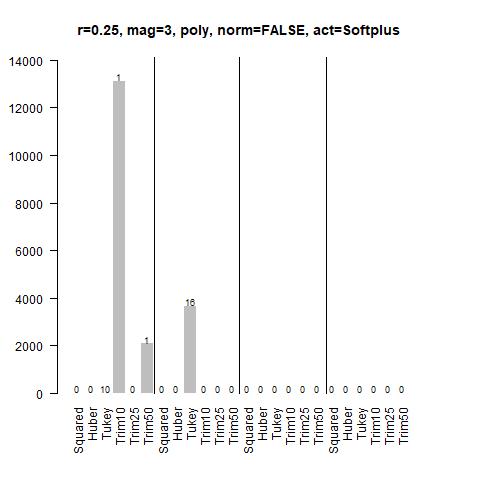} 
\includegraphics[width=6.75cm,height=6.25cm]{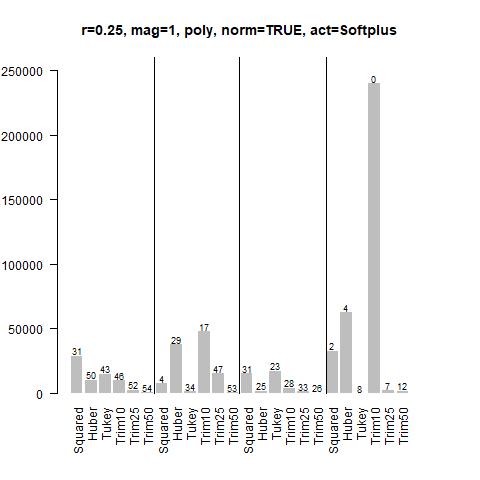}\\
\includegraphics[width=6.75cm,height=6.25cm]{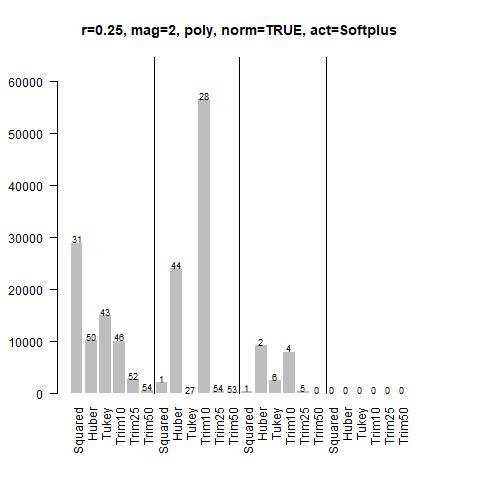} 
\includegraphics[width=6.75cm,height=6.25cm]{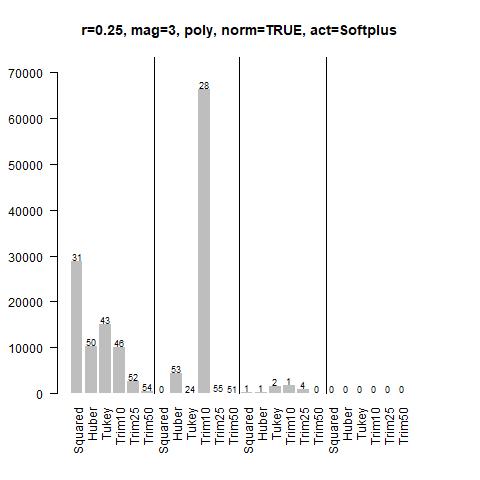} 
\end{center}
\caption{Results for $r=0.25$}
\end{figure}

\begin{figure}[H]
\label{trimnn:n1000p50r40m1polynonreludeepStep}
\begin{center}
\includegraphics[width=6.75cm,height=6.25cm]{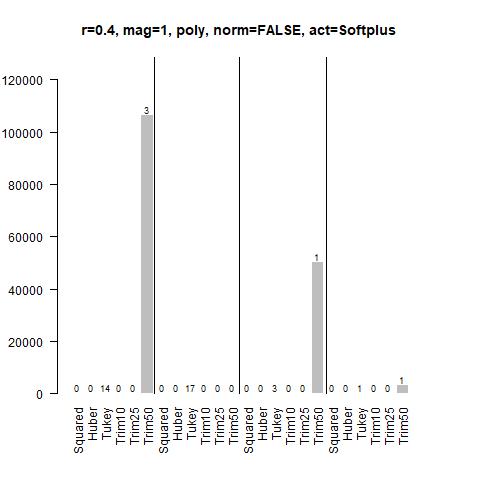}
\includegraphics[width=6.75cm,height=6.25cm]{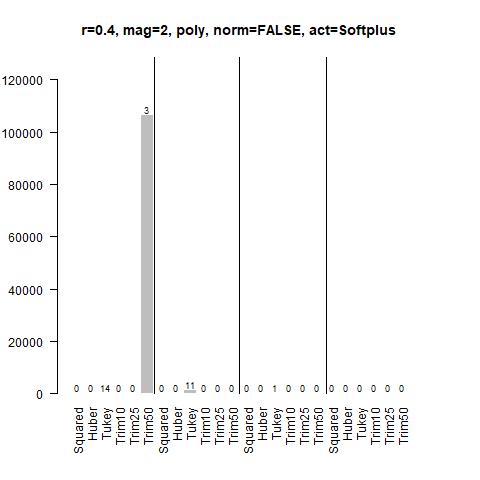} \\
\includegraphics[width=6.75cm,height=6.25cm]{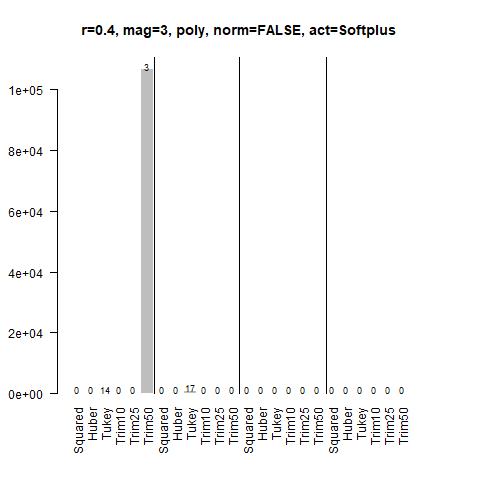} 
\includegraphics[width=6.75cm,height=6.25cm]{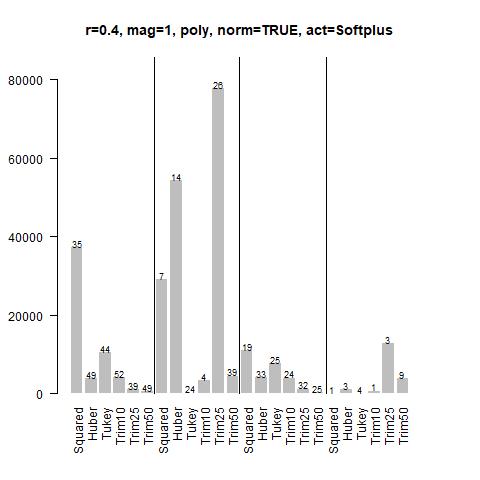}\\
\includegraphics[width=6.75cm,height=6.25cm]{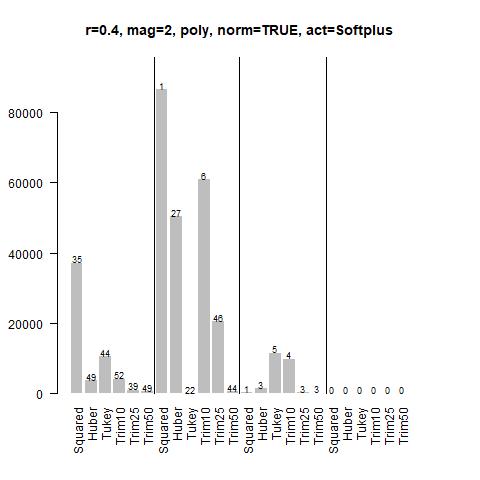} 
\includegraphics[width=6.75cm,height=6.25cm]{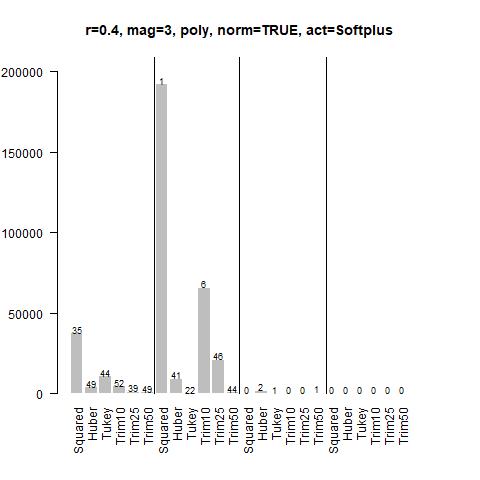} 
\end{center}
\caption{Results for $r=0.4$}
\end{figure}

\subsubsection{Trigonometric function}

\begin{figure}[H]
\label{trimnn:n1000p50r10m1trignonreludeepStep}
\begin{center}
\includegraphics[width=6.75cm,height=6.25cm]{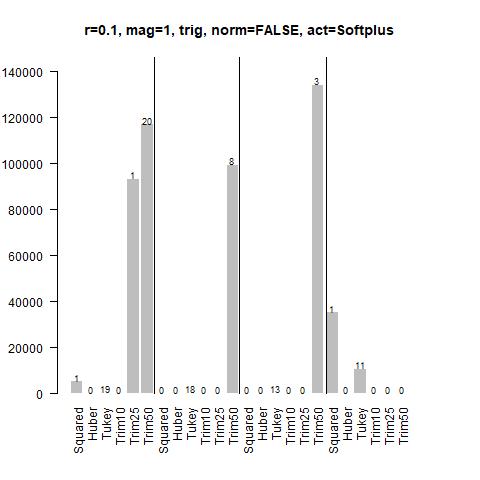}
\includegraphics[width=6.75cm,height=6.25cm]{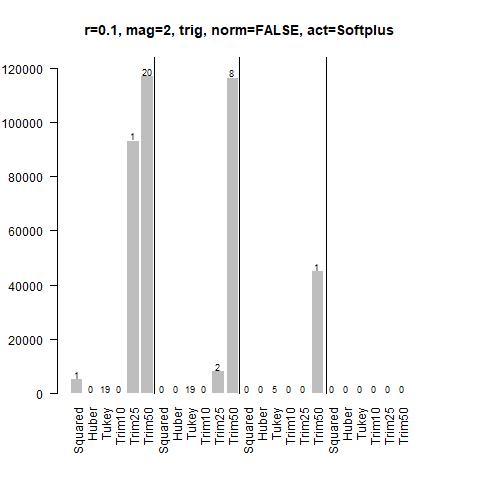} \\
\includegraphics[width=6.75cm,height=6.25cm]{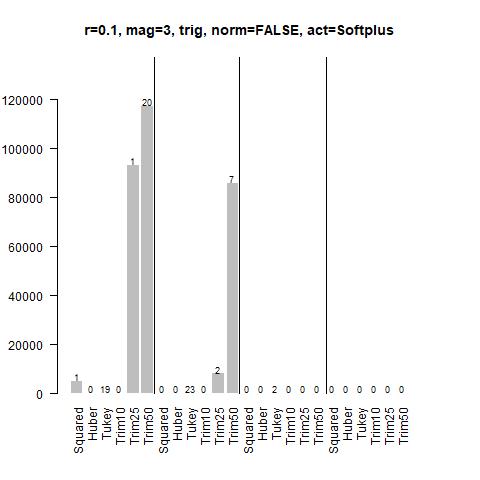} 
\includegraphics[width=6.75cm,height=6.25cm]{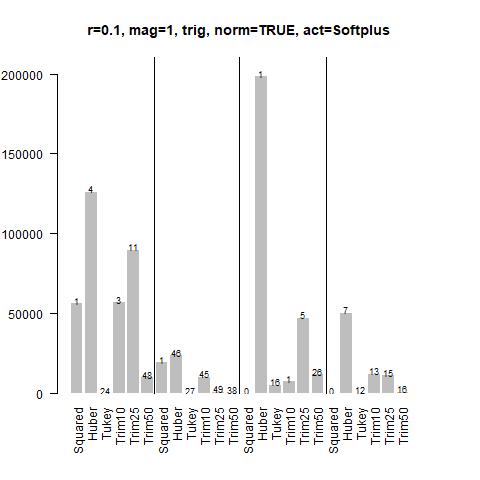}\\
\includegraphics[width=6.75cm,height=6.25cm]{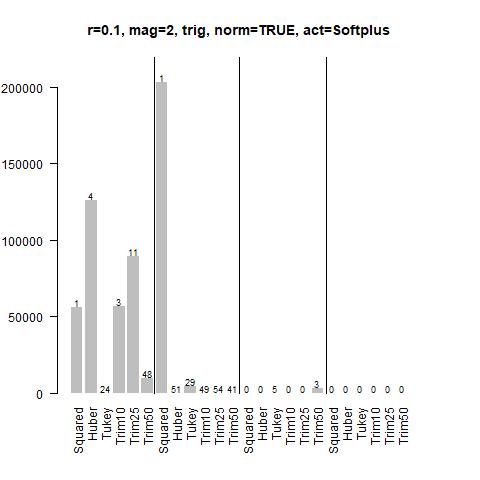} 
\includegraphics[width=6.75cm,height=6.25cm]{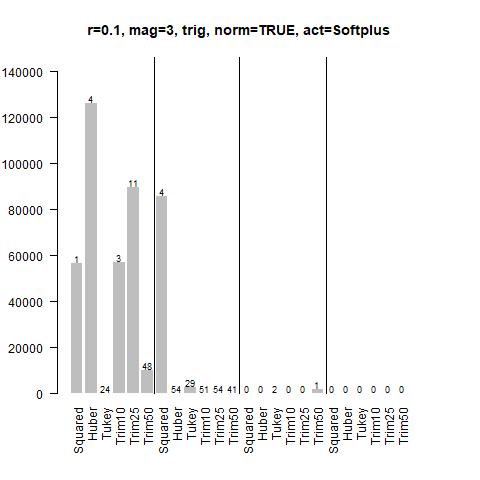} 
\end{center}
\caption{Results for $r=0.1$}
\end{figure}

\begin{figure}[H]
\label{trimnn:n1000p50r25m1trignonreludeepStep}
\begin{center}
\includegraphics[width=6.75cm,height=6.25cm]{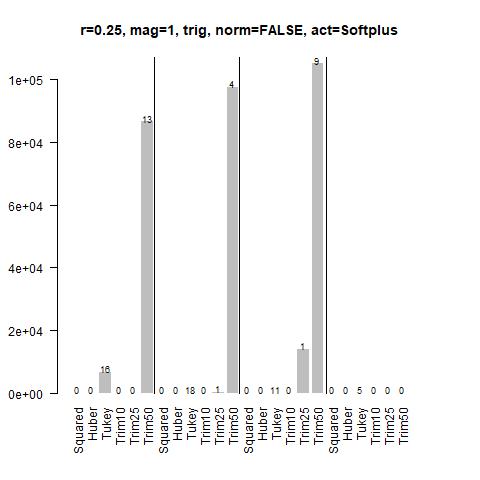}
\includegraphics[width=6.75cm,height=6.25cm]{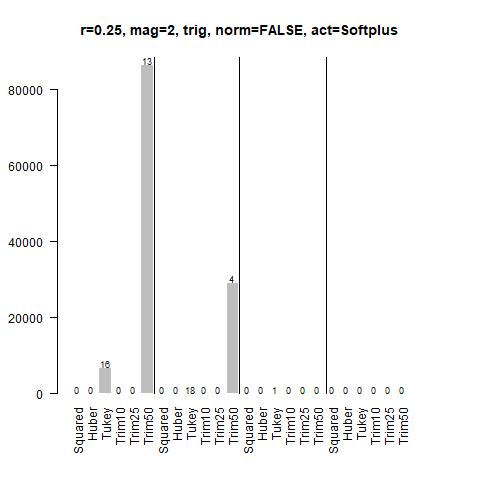} \\
\includegraphics[width=6.75cm,height=6.25cm]{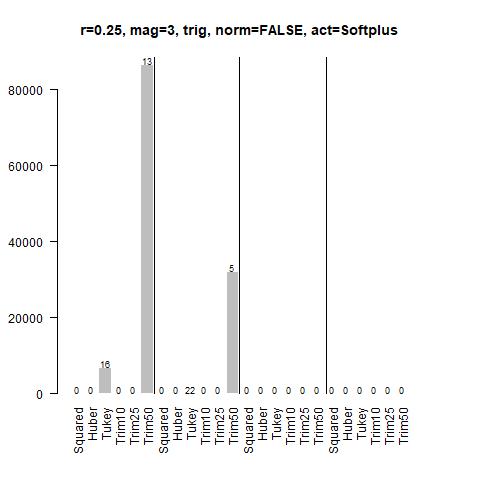} 
\includegraphics[width=6.75cm,height=6.25cm]{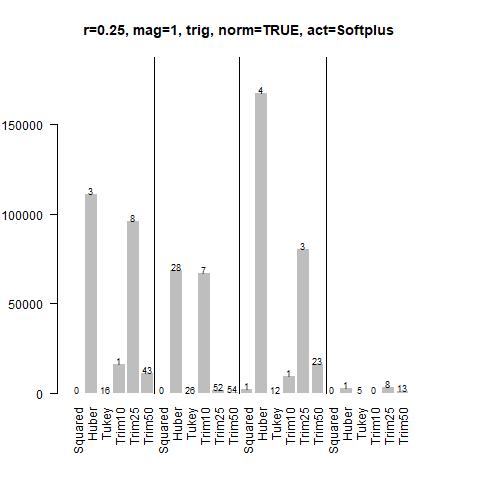}\\
\includegraphics[width=6.75cm,height=6.25cm]{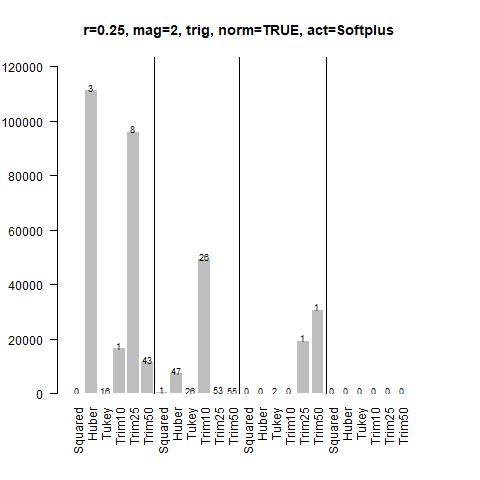} 
\includegraphics[width=6.75cm,height=6.25cm]{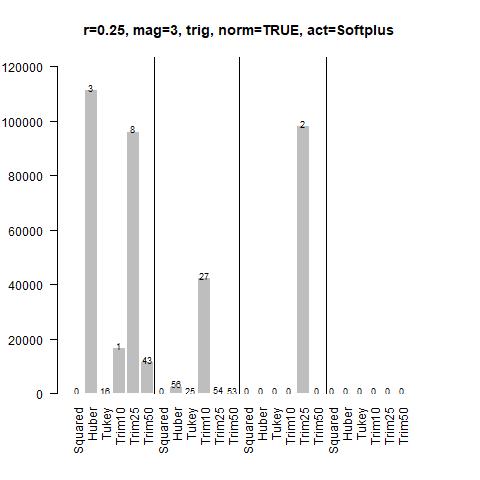} 
\end{center}
\caption{Results for $r=0.25$}
\end{figure}

\begin{figure}[H]
\label{trimnn:n1000p50r40m1trignonreludeepStep}
\begin{center}
\includegraphics[width=6.75cm,height=6.25cm]{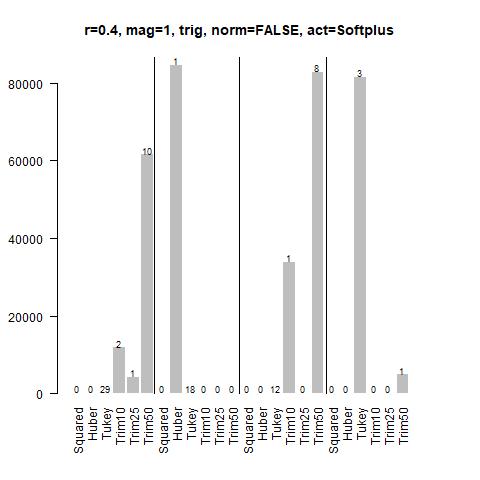}
\includegraphics[width=6.75cm,height=6.25cm]{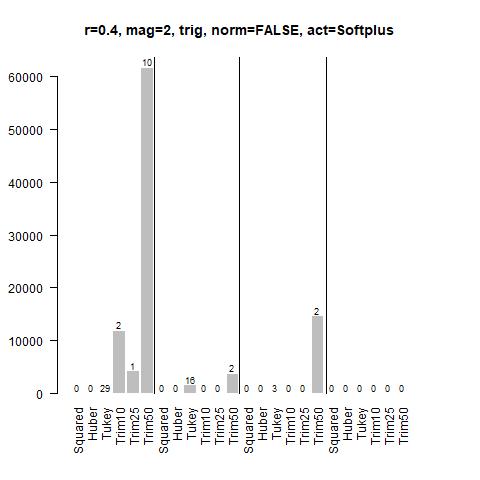} \\
\includegraphics[width=6.75cm,height=6.25cm]{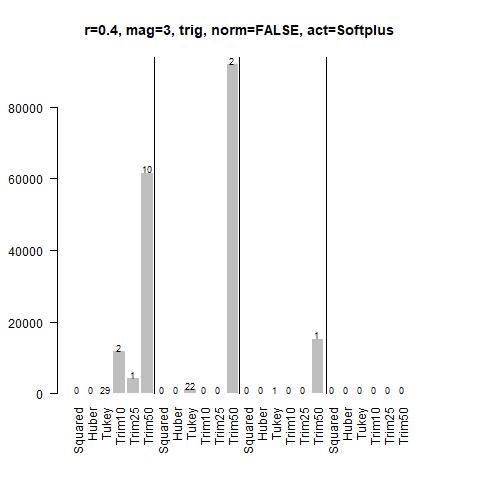} 
\includegraphics[width=6.75cm,height=6.25cm]{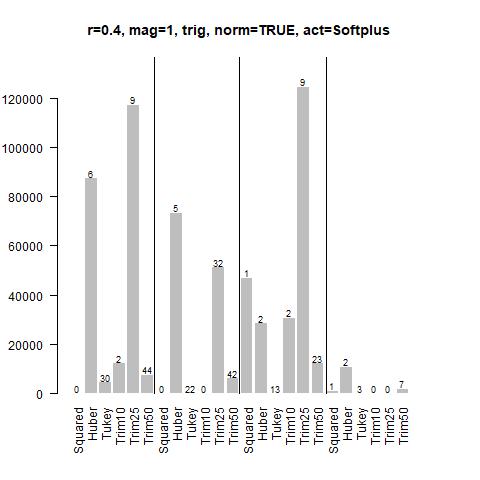}\\
\includegraphics[width=6.75cm,height=6.25cm]{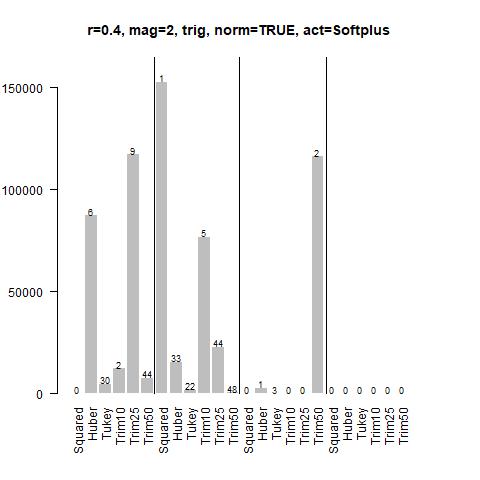} 
\includegraphics[width=6.75cm,height=6.25cm]{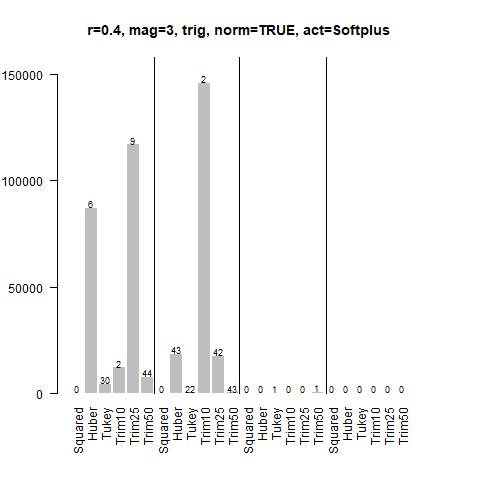} 
\end{center}
\caption{Results for $r=0.4$}
\end{figure}

\end{document}